\documentclass[11pt]{scrreprt}

\usepackage[utf8]{inputenc}
\usepackage[margin=1in]{geometry}
\usepackage[normalem]{ulem}
\usepackage{enumitem}
\usepackage{amsthm,thmtools,mathtools,thm-restate}
\usepackage{amssymb}
\usepackage{amsfonts}
\usepackage{bbm}
\usepackage{lmodern}
\newcommand{\mathsc}[1]{\text{\normalfont\textsc{#1}}}

\usepackage[titletoc]{appendix}

\usepackage{array, booktabs, multirow, tabularx, nicefrac}
\DeclareMathOperator*{\argmax}{arg\,max}
\DeclareMathOperator*{\argmin}{arg\,min}
\DeclareMathOperator{\E}{\mathbb{E}}

\usepackage[dvipsnames]{xcolor}
\definecolor{DarkRed}{rgb}{0.8,0,0.4}
\definecolor{LightRed}{rgb}{1.0,0.6,0.7}
\definecolor{ForestGreen}{rgb}{0.1333,0.5451,0.1333}
\definecolor{aliceblue}{rgb}{0.94, 0.97, 1.0}
\definecolor{blizzardblue}{rgb}{0.67, 0.9, 0.93}
\definecolor{Brown}{rgb}{0.545,0.27,0.07}
\definecolor{darkblue}{rgb}{0,0,0.5}
\definecolor{grey}{rgb}{0.91,0.91,0.91}
\definecolor{darkgreen}{rgb}{0,0.5,0}
\definecolor{paleblue}{rgb}{0.678,0.847,0.902}
\definecolor{llcolor}{RGB}{138,192,165}
\definecolor{hccolor}{RGB}{233,162,128}
\definecolor{dacolor}{RGB}{141,160,203}
\definecolor{skyblue}{RGB}{135, 206, 235}
\definecolor{ForestG}{rgb}{0,0.45,0.4}
\definecolor{DarkR}{rgb}{0.5,0,0.1}

\usepackage[hyphens]{url}
\usepackage[linktocpage=true,
    colorlinks,
    urlcolor=DarkR,
    linkcolor=DarkR,
    citecolor=ForestG,
    bookmarks,
    bookmarksopen,
    bookmarksnumbered]{hyperref}
\usepackage[noabbrev,nameinlink]{cleveref}
\crefname{property}{property}{Property}
\creflabelformat{property}{(#1)#2#3}
\crefname{equation}{eq}{Eq}
\creflabelformat{equation}{(#1)#2#3}
\crefname{lemma}{Lemma}{Lemmas}
\crefname{claim}{Claim}{Claims}

\usepackage[noend]{algpseudocode}
\usepackage[linesnumbered,vlined,ruled,noend,resetcount,algochapter]{algorithm2e}

\SetCommentSty{mycommfont}

\usepackage{pgfplots}
\pgfplotsset{compat=1.10}

\theoremstyle{plain}
\newtheorem{theorem}{Theorem}[chapter]
\newtheorem{definition}{Definition}[chapter]
\newtheorem{lemma}{Lemma}[chapter]
\newtheorem{proposition}{Proposition}[chapter]
\newtheorem{example}{Example}[chapter]
\newtheorem{corollary}{Corollary}[chapter]

\newtheorem{problem}{Problem}[chapter]
\newtheorem{remark}{Remark}[chapter]
\newtheorem{observation}{Observation}[chapter]

\renewcommand{\vec}[1]{\boldsymbol{\mathbf{#1}}}

\newenvironment{numberedtheorem}[1]{%
\renewcommand{\thetheorem}{#1}%
\begin{theorem}}{\end{theorem}\addtocounter{theorem}{-1}}

\newenvironment{numberedlemma}[1]{%
\renewcommand{\thelemma}{#1}%
\begin{lemma}}{\end{lemma}\addtocounter{lemma}{-1}}

\newenvironment{numberedcorollary}[1]{%
\renewcommand{\thecorollary}{#1}%
\begin{corollary}}{\end{corollary}\addtocounter{corollary}{-1}}

\newtheoremstyle{restate}{}{}{\itshape}{}{\bfseries}{~(restated).}{.5em}{\thmnote{#3}}
\theoremstyle{restate}

\usepackage{tcolorbox}
\tcbuselibrary{skins,breakable}
\tcbset{enhanced jigsaw}
\usepackage{mdframed}
\newtheorem{mdresult}{Result}

\usepackage{graphicx}
\usepackage{subcaption}
\let\origsubref\subref
\renewcommand{\subref}[1]{\origsubref{#1}}

\usepackage{wrapfig}
\usepackage{epsfig}

\usepackage{makecell, tabularx, booktabs, multirow}
\usepackage{tikz}
\usetikzlibrary{backgrounds, arrows, shapes, tikzmark, calc, positioning, 
     decorations.markings, pgfplots.groupplots, arrows.meta, fit,
     shadows, trees, mindmap, patterns, decorations.pathmorphing                    
}

\makeatletter
\def\BState{\State\hskip-\ALG@thistlm}
\makeatother

\usepackage{makecell, ragged2e, colortbl}
\newcolumntype{P}[1]{>{\RaggedRight\hspace{0pt}}p{#1}}
\newcolumntype{L}{>{\arraybackslash}m{3cm}}
\newcolumntype{q}{>{\arraybackslash}m{4.5cm}}

\renewcommand{\vec}[1]{\mathbf{#1}}
\newcommand{\points}{\mathcal{P}}
\newcommand{\pfinal}{\mathcal{P}^\text{final}}
\newcommand{\agents}{\mathcal{X}} 
\newcommand{\q}{\mathcal{Q}} 
\newcommand{\graph}{\mathcal{G}}
\newcommand{\f}{f} 
\newcommand{\g}{g} 

\newcommand{\opt}{\textup{\textsc{OPT}}}

\newcommand{\eps}{\varepsilon}

\renewcommand{\qed}{\nobreak \ifvmode \relax \else
    \ifdim\lastskip<1.5em \hskip-\lastskip
    \hskip1.5em plus0em minus0.5em \fi \nobreak
    \vrule height0.75em width0.5em depth0.25em\fi}

\newenvironment{proofsketch}{%
  \proof}{\endproof}

\newcommand{\init}[1]{{\mathbf{#1}}^{\text{init}}}
\newcommand{\true}[1]{{\mathbf{#1}}^{\text{true}}}
\newcommand{\perc}[1]{{\mathbf{#1}}^{\text{perc}}}
\newcommand{\initij}[1]{{\mathbf{#1}}^{\text{init}}_i[j]}

\newcommand{\percij}[1]{{\mathbf{#1}}^{\text{perc}}_i[j]}
\newcommand{\Tau}{\mathcal{T}}

\DeclareMathOperator{\sw}{\text{SW}}

\DeclareMathOperator{\T}{\mathcal{T}}
\DeclareMathOperator{\D}{\Delta}
\DeclareMathOperator{\PDim}{PDim}
\renewcommand{\vec}[1]{\mathbf{#1}}

\newcommand{\vecj}[1]{{\mathbf{#1}}[j]}
\newcommand{\veck}[1]{{\mathbf{#1}}[k]}
\newcommand{\vecone}[1]{{\mathbf{#1}}[1]}
\newcommand{\vectwo}[1]{{\mathbf{#1}}[2]}
\newcommand{\vecstarj}[1]{{\mathbf{#1}}^{\star}[j]}
\newcommand{\vecstark}[1]{{\mathbf{#1}}^{\star}[k]}

\DeclareMathOperator{\Xs}{\mathcal{X}}
\DeclareMathOperator{\Ts}{\mathcal{T}}

\DeclareMathOperator{\mneg}{\mathit{m}^{-}}
\DeclareMathOperator{\mpos}{\mathit{m}^{+}}
\DeclareMathOperator{\qneg}{\mathit{q}^{-}}
\DeclareMathOperator{\qpos}{\mathit{q}^{+}}
\DeclareMathOperator{\splan}{\text{social planner}}

\allowdisplaybreaks

\usepackage{soul}

\SetKwInput{KwInput}{Input}
\SetKwInput{KwOutput}{Output}

\newcounter{para}

\newcommand{\R}{\ensuremath{\mathbb{R}}}

\newcommand{\floor}[1]{{\left\lfloor{#1}\right\rfloor}}

\DeclareMathOperator*{\Prob}{\ensuremath{\mathbb{P}}}
\renewcommand{\Pr}{\Prob}

\newenvironment{tbox}{\begin{tcolorbox}[
		enlarge top by=5pt,
		enlarge bottom by=5pt,
		 breakable,
		 boxsep=0pt,
                  left=4pt,
                  right=4pt,
                  top=10pt,
                  arc=0pt,
                  boxrule=1pt,toprule=1pt,
                  colback=white
                  ]
	}
{\end{tcolorbox}}

\newcommand{\II}{\ensuremath{\mathbb{I}}}

\newcommand{\mireal}[1][]{
  \ifx\relax#1\relax%
    \II(\mione \,; \mitwo)%
  \else%
    \II(\mione \,; \mitwo\mid #1)%
  \fi
}

\newcommand{\norm}[1]{\ensuremath{\left\lVert #1 \right\rVert}}

\newcommand{\lrset}[1]{\mathopen{}\left\{#1\right\}}

\newcommand{\clos}{\mathrm{CLOS}}

\newcommand{\ir}{\mathrm{IR}}
\newcommand{\bs}{\mathrm{BS}}

\usepackage[
backref=true,
natbib=true,
style=apa,
maxalphanames=8,
sorting=nyt
]{biblatex}
\begin{document}

\pagenumbering{gobble}

\title{Shaping Human-AI Interactions to Provide Improvement Pathways and Balance Competing Objectives}

\date{\today}
\author{KEZIAH NAGGITA}

\begin{titlepage}
   \begin{center}
       \vspace*{2cm}
        {\LARGE \textbf{Shaping Human-AI Interactions to Provide 
        \\ \vspace{0.1cm}
       Improvement Pathways and Balance 
        \\ \vspace{0.3cm}
        Competing Objectives}}

        \vspace{2cm}
        By\\
        Keziah Naggita
        
        \vspace{2cm}   
        A thesis submitted
        \\
        in partial fulfillment of the requirements for the degree of
        
        \vspace{0.11cm}
        DOCTOR OF PHILOSOPHY IN COMPUTER SCIENCE 
        
        \vspace{0.11cm}
        at the

        \vspace{0.11cm}
        TOYOTA TECHNOLOGICAL INSTITUTE AT CHICAGO
        \\
        Chicago, Illinois

        \vspace{0.11cm}
        June, 2026
        
        \vspace{1cm}

        \begin{figure}[ht!]
        \centering
            \includegraphics[totalheight=2.5cm]{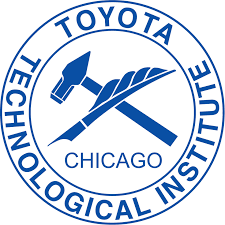}
        \end{figure}

        \vspace{1cm}
            
        {\large\textbf{Thesis Committee:}}\\
        Avrim Blum (Thesis Advisor)\\
        Matthew Walter (Thesis Advisor)\\
        Hedyeh Beyhaghi\\
        Sarah Sebo  

   \end{center}

    \vspace*{\fill}
    \begin{center}
    © 2026 Keziah Naggita. All rights reserved.
    \end{center}

    \clearpage

    \begin{center}
        {\Large \textbf{Shaping Human-AI Interactions to Provide Improvement \\ \vspace{0.3cm} Pathways and Balance Competing Objectives}}      
        \vspace{1cm}
    \end{center}

    \subsubsection*{\makebox[\textwidth]{Abstract}}
    {
    \small
    \setlength{\parskip}{0pt}
        When an AI system is deployed, the individuals who use and or are evaluated by it form beliefs about how the system operates and use those beliefs to strategically present their preferences, behaviors, or attributes. The system then responds with feedback or a decision outcome, thereby creating a human-AI interaction loop. This thesis studies how to design and shape such interactions to achieve three goals: (1) help individuals develop accurate beliefs about the AI systems so they can improve and or secure favorable outcomes at minimal cost, (2) encourage improvement and or discourage gaming behaviors, and (3) ensure that the AI system continues to achieve its intended objectives, such as maximizing accuracy.
        
        To address these goals, the thesis is organized into three complementary parts that examine and study human-AI interactions from the perspectives of both evaluated individuals and AI systems.  
        Through an online human-subject experiment, Part~I investigates how parents perceive and respond to children’s aggressive behavior towards embodied and disembodied AI-driven home devices, with the goal of informing the design of AI-mediated developmental pathways for children.
        
        In Part~II, we study how to design and shape human-human and human-AI interactions to support individual improvement. Focusing on people’s interactions with AI-driven decision systems, we develop methods for setting reachable targets that maximize improvement, generating actionable guidance to help individuals reverse unfavorable decision outcomes, and revealing information that maximizes the expected number of individuals who emulate positive role models.
        
        Part~III studies human-AI interactions from the perspective of the AI-driven decision system, aiming to balance competing objectives such as enabling individuals to achieve favorable outcomes at minimal cost while maximizing the system’s predictive accuracy. We analyze how individuals' capacity for improvement affects learnability and the design of algorithms for achieving accurate classification. Lastly, to maximize true positives while minimizing false positives, we develop theoretical and empirical foundations for accurately classifying all individuals, penalizing gaming, and incentivizing genuine improvement.
        
        Together, the work presented in this thesis advances human-centered machine learning by providing principles and methods for designing AI systems that align with human needs, values, and capabilities. Methodologically, this thesis integrates theoretical analysis, data-driven modeling, human-subject experimentation, and empirical evaluations on real-world and semi-synthetic datasets.
    }

    \vspace*{\fill}
    
    {\footnotesize
    \noindent\textbf{Keywords:} Agents Improvement, Algorithmic Fairness, Child-Robot Abuse, Child-Robot Interaction, Child-Technology Interaction, Counterfactual Explanations, Counterfactual Interventions, Data-Driven Algorithm Design, Incentivizing Improvement, Intervention, Learning for Strategic Behavior, Learning from Graphs, Linear Programming, PAC Learning, Parenting, Recourse, Sample Complexity, Social welfare, Strategic Classification, Submodular optimization
    }

\end{titlepage}

\pagenumbering{roman}

\newpage 
\topskip0pt
\vspace*{\fill}
\begin{center}
    \emph{For mom, Bwenene \(<\!3\)}\\
    \emph{N'okutuusa kati, Mukama akyatubedde}
\end{center}
\vspace*{\fill}

\newpage
\section*{\Huge{Acknowledgments}}
I am deeply grateful to everyone who has made this journey and every season of it meaningful.

First and foremost, I would like to express my deepest gratitude to my PhD advisors, Avrim Blum and Matthew Walter (Matt), for taking a chance on me and for their support, patience, and guidance throughout my PhD. 
Avrim, thank you so much for always being able to distill my long and often unfocused thoughts into intuitive and compelling ideas that became the foundation of the papers comprising this thesis.  
I am deeply grateful for the many ways you have advocated for me throughout my studies and for introducing me to various academic communities that have profoundly enriched my professional growth. 
Matt, thank you so much for continuing to serve as my advisor, even after my dissertation research shifted from robotics to responsible and trustworthy AI.  I am deeply grateful for the many technical and nontechnical conversations we have shared over the years, as well as for the thoughtful and detailed feedback you have provided on my writing, which has strengthened my work in countless ways.

I would like to thank Hedyeh Beyhaghi and Sarah Sebo for serving on my thesis committee, and for the opportunity to collaborate with them both on projects that have profoundly enriched and shaped this thesis. 
Hedyeh, thank you so much for introducing me to theoretical research, for sharing your expertise, and for the chance to learn from your insight and perspective. 
Sarah, thank you so much for giving me the opportunity to explore HRI research in your lab, and for empowering me to take risks, learn from costly experimentation mistakes, and grow as a researcher both independently and as part of the team.

Beyond my thesis committee, I am deeply grateful to the many researchers I have had the privilege of working with. In alphabetical order, thank you so much to S. Ahmadi, J. C. Aguma, E. Athiley, I. Attias, B. Desta, J. Finocchiaro, M. Juarez, J. LaChance, F. Monachou, A. Ritchie, D. Saless, J. Schoeffer, D. Sharma, J. Wang, and A. Xiang. Working with you all profoundly shaped my PhD experience and strongly influenced my research and career interests.

My internship at SONY AI was both timely and deeply enriching. I am incredibly grateful to my internship mentors, Julienne LaChance (Julie) and Alice Xiang, as well as the entire SONY AI Ethics team, for a fun, collaborative, and inspiring summer. Julie, thank you so much for continuing to be my collaborator and for championing me long after the internship ended.

I have been incredibly fortunate to be part of several vibrant academic communities, including TTIC, UChicago, MD4SG, TOC4Fairness, and CRA-WP, among others, where I had the opportunity to engage with students, postdocs, and faculty across all levels. I am deeply grateful for the many thought-provoking conversations, whether in classes, seminars, spontaneous hallway chats, or moments of (in)formal guidance, that brightened my days and shaped my research and career aspirations. Thank you to the following students (at the time):  Aida R., Akilesh T., Alexandra D., Ana S., Ankita P., Ceasar A., Chara P., Claire Z., David Y., Elaine K., Falcon D., Freda S., Harvineet S., Jessica F., Jonas N., Jerry M., Jerry T., Justyna K., Kaoru S., Kavya R., Kevin I., Kevin S., Kshitij P., Melissa D., Naren M., Omar M., Pushkar S., Rachel H., Shengjie L., Sudarshan B., Takuma Y., Yanan L., Zoe L., and so many others. 

The administrative staff at TTIC has been exceptional and very kind, consistently going above and beyond to make the PhD experience both smooth and positively memorable. In alphabetical order, thank you so much to the following administrative staff I had the pleasure of interacting with: Adam B., Alicia M., Amy M., Brandie J., Chrissy C., Deree K., Erica C., Liz C., Mary M., Matthew T., Randall L., and Rose B. I am also deeply grateful to the facilities staff, Aisha C., and Jerry R., for the warm and cheerful banter that always brightened my days.

My PhD journey would have been very different without Charles Earl, whom I was lucky to be matched with as a mentor through the Black in AI graduate mentorship program. Thank you so much, Charles, for going far beyond your role as a graduate applications mentor. You generously critiqued my application essays, encouraged me to apply to TTIC, and introduced me to Janos Simon, together with whom you offered invaluable support and guidance throughout my application process and transition to TTIC and Chicago.  

Above all, to my family, boyfriend, and friends: your unwavering love, support, and prayers sustain me, without which I wouldn’t be who I am today. 
I am forever grateful for the many ways you enrich and brighten my life. In alphabetical order, thank you so much to Ceasar K., Christopher N., Edwin M., Evelyn N., Herbert A., Irene N., Kendra T., Monica N., Nicholas M., Nnaji I., Priscilla N., Samallie N., Samuel K., Shafik K., Tezra N., and Timothy N.  
{\footnotesize XOXO}

I have spent a great deal of time reflecting on the journey that brought me to this moment and on the many people who made it easier to navigate, enriched it with countless positive memories, and ultimately made the various seasons brighter. For example, I think of my teachers, Madam Nulu, Mr. Semakula, and Mr. Kilabiriza, who made it possible for me to attend Kyambogo tuition-free, and of the various Uber/Lyft drivers who spoke life into me in my first year in Chicago. I could go on and on because the greatest fortune of my life has been the people who have seen me. However, since this space is far too small to properly name and honor all the incredible people who have shaped my journey, I would like to end this acknowledgment section by expressing my deepest gratitude to the various influential people in my life by group.
To my family, both given and chosen, who make every day worth living; to the staff, teachers, lecturers, and professors who took me under their wing;  to the mentors, both formal and informal, who offered me guidance and generously shared their wisdom;  to the generous letter writers and recommenders who opened doors for me;  to genuine friendships forged along the way; and  to the people, sometimes strangers, who spoke life into me when I needed it the most,  each of you has played an invaluable part in my journey and I will forever be grateful to you. While these words are less specific and far too few to capture the depth of my gratitude, know I will always carry your impact with me and will do my best to pay it forward. Thank you so much!

\begin{center}
    \emph{Mukama kyaterekera omunaku tekivunda}\\
    ... \emph{Always sonder\footnote{\url{https://www.youtube.com/watch?v=AkoML0_FiV4}} + ubuntu\footnote{\url{https://en.wikipedia.org/wiki/Ubuntu_philosophy}}} ...
\end{center}

\newpage
\tableofcontents
\newpage
\listoffigures
\newpage
\newpage
\listofalgorithms
\newpage

\setkomafont{chapter}{\huge\bfseries}
\RedeclareSectionCommand[
  beforeskip=5pt,  
  afterskip=20pt    
]{chapter}
\renewcommand*{\chapterlinesformat}[3]{%
 \Huge\bfseries Chapter \thechapter\par\nobreak
  \vskip 5pt          
 \LARGE\bfseries #3\par\nobreak
}

\pagenumbering{arabic}

\chapter{Introduction}
\label{chap:overview}
The proliferation and integration of artificial intelligence (AI) systems in both home settings (e.g., smart speakers and robots) and high-stakes decision-making contexts (e.g., hiring and admissions) is transforming how individuals interact with one another, perform daily tasks, and engage with these systems. Because these systems can materially affect people's livelihoods, society rightfully demands transparency and the right to explanations, as reflected in Articles 13-15 of the \citet{EuropeanParliament2016a} General Data Protection Regulation and Article 13 of the \citet{EuropeanParliament2025} AI Act.

However, providing detailed explanations of how the AI systems operate might encourage strategic behavior from those being evaluated. Individuals could use that knowledge to influence the system's feedback or decisions at minimal cost, for instance, by selectively modifying observable features, withholding information, or opting out of the interaction altogether. 
For example, in content moderation, creators who learn how classifiers identify prohibited content may adopt forms of ``algospeak'' or coded expressions \citep{lorenz2022algospeak} to evade demonetization. Similarly, in hiring contexts, applicants may increase chances of a favorable decision outcome either through genuine \textit{improvements}, such as additional training, or through \textit{gaming} actions, such as tailoring resumes to keyword-based AI applicant screening systems without corresponding gains in underlying qualifications.

The classification of such strategic individuals (agents)\footnote{In this thesis, individuals evaluated by AI systems are sometimes referred to as \textit{agents}. This terminology reflects the interactive nature of human-AI decision settings, in which individuals may act as strategic agents who truthfully or fraudulently alter their observable features or behavior so as to obtain more favorable outcomes from AI-driven decision systems.} was formalized in the literature as \textit{strategic classification} \citep{hardt2016strategic,Kleinberg2018HowDC,revealed_preferences}.
Beyond agent-side responses, the AI-driven decision system (decision-maker) may also act strategically by anticipating how agents will optimally respond and incorporating these expectations into the design of the AI-driven decision system. Relatedly, to help agents form more accurate beliefs about AI models and respond optimally, research on \textit{algorithmic recourse} and \textit{counterfactual explanations} (CFEs) \citep{Karimi22,Verma2020CounterfactualEF} aims to provide individuals with actionable guidance on how to modify their features  to reverse  unfavorable decision outcomes.

In three parts, this dissertation examines and shapes human-AI interactions from both the agents' and AI systems' perspectives. 
In particular, 
Part~I focuses on understanding parent-child and child-AI interactions to help inform design decisions that can encourage fewer instances of aggressive behavior from children and help parents mitigate any forms of child aggression towards AI-driven home devices.
Part~II focuses on shaping human-human and human-AI interactions to support agents' improvement, for example, by helping individuals develop accurate beliefs about AI-driven decision systems so they can more effectively improve their features or choices and achieve better outcomes.
Then, from the perspective of the AI-driven decision system, Part~III studies how to shape agent-AI interactions to jointly optimize the objectives of the individuals being evaluated and those of the AI-driven decision systems, such as securing a favorable classification at minimal cost while maximizing the system accuracy. Below is an overview of the three parts.

Part~\ref{sec:intro_part1} provides an in-depth discussion of the first component of this thesis. It examines parental responses to aggressive child behavior towards robots, smart speakers, and tablets. Through an online between-subjects experimental study, we investigate how parents perceive and respond to children's aggressive behavior towards robots in the home compared with other technological devices (smart speakers and tablets). We also examine whether parents' perceptions of the device's \emph{anthropomorphic} and \textit{animacy} characteristics affect how they perceive and respond to their child's aggressive behavior.

The second part of this thesis (Part~\ref{sec:intro_part11}) studies settings in which agents respond to AI-driven decision systems strategically but truthfully, taking actions that improve their underlying qualifications rather than merely adjusting observable features to influence how they are perceived by the decision-maker.  To support such individual improvement, we first develop algorithms for computing (near-)optimal target placements that incentivize and maximize agents' improvement, with and without fairness considerations. Second, we introduce linear programming and data-driven methods for searching the real-world-like action spaces to identify the least-cost set of actions an individual can take to reverse an unfavorable decision outcome. Finally, we provide algorithmic and hardness results for maximizing the expected number of agents who emulate positive role models in a social network, in a setting where agents take actions by following adjacent role models but cannot distinguish positive from negative ones.

In Part~\ref{sec:intro_part111}, we study settings in which agents can both game and improve; that is, they may modify their observable features either truthfully, deceptively, or through a combination of the two in order to obtain favorable outcomes from AI-driven decision systems. The dual capacity for gaming and genuine improvement presents both challenges and opportunities. Gaming behavior can increase the rate of false positives, whereas genuine improvement can increase the number of true positives. Accordingly, this part of the thesis investigates how agents' capacity for improvement affects learnability, sample complexity, and algorithm design for accurate classification. Lastly, to maximize true positives while minimizing false positives, we develop theoretical and empirical foundations for correctly classifying agents, discouraging gaming, and incentivizing agents with the capacity to improve to become qualified.

Put together, the three parts advance human-centric machine learning by deepening our understanding of human-AI interactions and informing the design of AI systems that better account for human needs, values, and capabilities. Across this thesis, we aim to link rigorous analysis with actionable insights that support trustworthy and responsible human-AI interactions. 
Methodologically, the chapters spans human-subject experimentation (Chapter~\ref{chap:parental_concern}), theoretical analysis (Chapters~\ref{chap:setfair}, \ref{chap:revealrm}, \ref{chap:paclearn}, and \ref{chap:ocagi_theory}), and empirical computations on both benchmark real-world datasets and task-specific (semi-)synthetic datasets (Chapters~\ref{chap:cfes} and \ref{chap:ocagi_exp1}).

\section{Part I: Investigating Human-AI Interactions Through Human-Subject Experiments} 
\label{sec:intro_part1}

In Part~I, we examine how parents respond to children's aggressive interactions with embodied and disembodied AI-driven home devices, and how parents' perceptions of the devices' anthropomorphism and animacy shape these responses. Our goal is to inform the design of effective AI-mediated interventions that promote positive behavioral development in children.

\subsection{Parental Responses to Aggressive Child Behavior Towards Robots, Smart speakers and Tablets}
\label{sec:intro_parental}

AI-driven home devices such as smart speakers and robots are increasingly reshaping how children learn, play, and interact with family members~\citep{Michaeliseaat5999, KONIJN2020103970, Scassellatieaat7544, Othman17}. Alongside these benefits, prior work shows children may behave aggressively toward such devices. For example, \citet{Brscic15} documented instances of children abusing robots in public spaces by obstructing their movement, verbally insulting them, and physically attacking them. Relatedly, anecdotal concerns have been raised about children's verbal aggression toward smart speakers~\citep{alexa_making_us, alexa_making_kids_rude}, which has led companies to introduce politeness-oriented features such as Amazon's Magic Word~\citep{magic_word} and Google's Pretty Please~\citep{pretty_please}.

Although prior work shows that children may behave aggressively toward various AI-driven home devices, it remains unclear whether parents respond differently to such behavior depending on how anthropomorphic and animate they perceive the device to be. To address this gap, we conducted an online between-subjects experiment with a \(2\) (aggression: aggressive vs. \ neutral) \(\times 2\) (device type: robot vs. \ smart speaker/tablet) \(\times 3\) (interaction modality: audio, physical, audio + physical) design to answer the following question:

\begin{center}
\emph{How do parents perceive and respond to children's aggression toward robots compared with other AI-driven home devices, and how do perceptions of a device's anthropomorphism and animacy shape those responses?}
\end{center}

The findings, detailed in Chapter~\ref{chap:parental_concern}, demonstrate that parents' responses and interventions vary according to their perceptions of a device's anthropomorphism and animacy, as well as the form and context of children's aggressive behavior toward AI-driven home devices. These results underscore the importance of attending not only to children's aggressive behavior but also to how it manifests across device types.

Insights from Chapter~\ref{chap:parental_concern} can inform design strategies that reduce child aggression towards AI-driven home devices and support parents to be better equipped to recognize and address device-directed aggression in developmentally appropriate ways.

\section{Part II: Shaping Human-Human and Human-AI Interactions to Provide Improvement Pathways} 
\label{sec:intro_part11}

In Part~II, we study how to shape human-human and human-AI interactions to foster individual improvement. In particular, we develop methods for setting reachable targets that incentivize and maximize agent improvement (Section~\ref{sec:intro_reachable}), generating actionable insights in large state spaces to help agents reverse unfavorable decision outcomes (Section~\ref{sec_intro_cfes}), and selectively revealing information about role models to maximize the expected number of agents who emulate adjacent positive role models (Section~\ref{sec_intro_steer}).

\subsection{Setting Reachable Targets to Maximize People's Improvement}
\label{sec:intro_reachable}

Consider a policymaking scenario where setting appropriate targets for beneficiaries is crucial. In contexts such as education or career development, targets that are too low (easily attainable) may prevent agents from realizing their full potential, while targets that are too high (out of reach) can be discouraging, potentially inhibiting any attempt at improvement.

Assuming that agents will attempt to improve from their initial skill level to the nearest reachable target, or do nothing if no target is attainable, this part of the thesis, detailed in Chapter~\ref{chap:setfair}, investigates how short-term targets can be set to incentivize and maximize agent improvement. 
This problem formulation presents two key challenges. Target placement is non-monotonic, so adding a target can reduce total improvement because agents redirect effort from a distant goal to a closer one, and with multiple agent groups, simultaneous target placement can further induce interference that lowers improvement for some or all groups.
Given these challenges, the central research question is:

\begin{center}
\emph{What is the optimal placement of short-term goals (i.e., targets) that maximizes agents' total improvement (i.e., social welfare), both with and without fairness considerations?}
\end{center}

This line of inquiry is most closely related to \citet{Diana-etal}, who study the assignment of agents to portfolios with risk below their tolerance, minimizing the gap between each agent's tolerance and the risk they are assigned. Unlike our model, theirs is a minimization problem, and adding any new target can only help with the objective function. Our work is also related to \citet{steering-user-beha}, who analyze optimal badge placement to influence agent behavior under a single agent type. Unlike our setting, their model allows multi-action effort without conflicts across agents, and additional badges weakly increase the desired behavior.

Suppose the policymaker has full access to agents' initial skill levels and improvement capacities, which may be common (the same for all agents) or individualized (unique to each agent). In this setting, we develop a polynomial-time algorithm to optimally place at most $k$ targets that maximize total improvement for $n$ agents and a pseudo-polynomial algorithm to construct the Pareto frontier of group social welfare for $n$ agents across $g$ groups. Chapter~\ref{chap:setfair} also provides a fully polynomial-time approximation scheme for the max-min objective when groups have distinct improvement capacities, and show that max-min solutions do not guarantee a constant approximation simultaneously for all groups; in fact, no solution can achieve an approximation factor better than $1/g$ across all groups when agents have a common improvement capacity. For agents with common improvement capacity, we design an algorithm that achieves a simultaneously $\Omega(1/g^3)$-approximately optimal total improvement per group, while, for individualized capacities, no approximation factor can be defined solely based on the number of groups. Lastly, when the policymaker only has sampled access to agents, we provide generalization guarantees for both total improvement maximization and fairness objectives.

The key takeaway from Chapter~\ref{chap:setfair} is that although setting achievable targets is essential for effective policymaking, doing so is inherently difficult because target placement is non-monotonic, and adding new targets can paradoxically reduce overall agent improvement. This challenge is amplified in multi-group settings where targets that are optimal when placed independently for each group can, when combined, lead to arbitrarily poor improvement outcomes for both.
Several open questions remain, e.g., the design of algorithms for nonhomogeneous improvement models and closing the gap between the $\Omega(1/g^3)$ and $1/g$ approximation guarantees.

\subsection{Generating Actionable Insights in Large State Spaces to Help People Reverse Unfavorable Decision Outcomes} \label{sec_intro_cfes}

As mentioned in the introduction of Chapter~\ref{chap:overview}, merely informing agents of the system's decision outcome is insufficient. Decision-makers are also required to provide unsuccessful applicants with actionable guidance that enables them to make necessary changes to overturn an unfavorable decision outcome. Unlike Section~\ref{sec:intro_reachable} that focused on developing algorithms for incentivizing agents to optimally improve, here, the assumption is that agents already have an incentive to improve but don't have knowledge of how to best respond.

In particular, we assume that the decision-maker knows where agents are in the large state space and also what personalized actions agents could take to make feature-level changes that can reverse an unfavorable decision outcome. In contrast, agents are assumed to lack knowledge of the best action to take and therefore need the personalized recommendations to direct them toward states associated with favorable outcomes. This problem setting falls within the broader area of algorithmic recourse or (counterfactual explanation) CFE generation.  Although substantial progress has been made in this area~\citep{Wachter2017, Ustun19, Shalmali19, Dandl2020, Mothilal20, Karimi21, Karimi22},  existing approaches largely underutilize similarities among agent states during CFE generation, and producing CFEs that are easy to execute remains a challenge. The difficulty of the latter challenge arises in part from the absence of clear real-world action analogs and well-defined notions of recourse cost. Motivated by these limitations, Chapter~\ref{chap:cfes} addresses the following question:

\begin{center}
\emph{ How can decision-makers efficiently generate actionable insights that resemble real-world actions to help agents truthfully modify their input features (i.e., state) and reverse unfavorable decision outcomes?}
\end{center}

To address this question, Chapter~\ref{chap:cfes} proposes both single-agent CFE formulations and data-driven approaches for generating actionable and real-world-like CFEs. The single-agent CFE generation methods bridge the gap between feature-based and action-based action spaces, enabling the production of more real-world-like actionable insights. We also redefine the price of recourse to better reflect real-world costs and experimentally demonstrate that our proposed forms of recourse have fewer easy-to-interpret actions, result in higher levels of agent improvements, and modify more features than traditional feature-level recourse. In addition, we propose data-driven approaches that learn from agent-CFE training data, allowing the efficient generation of optimal CFEs for new agents while leveraging similarities between agent states. We further analyze how restricting agents to specific sets of actions or varying feature satisfiability under threshold classifiers influences both the quality of recourse and the performance of learned data-driven CFE generators across different groups.

The key takeaway from Chapter~\ref{chap:cfes} is that moving beyond a narrow focus on classifier-specific details (e.g., system parameters and training data) and instead leveraging similarities among agents' initial states can substantially improve the scalability and efficiency of automated CFE generation. A promising direction for future research is to examine the robustness of data-driven CFE generators under model drift, and to assess whether CFE generators based on real-world actions exhibit greater resilience to system changes than feature-level approaches.

\subsection{Selectively Revealing Positive and Negative Role Models to Help People Make Good Decisions}
\label{sec_intro_steer}

Although Sections~\ref{sec_intro_cfes} and~\ref{sec_intro_steer} both develop mechanisms for helping agents achieve favorable outcomes, they operate under different modeling assumptions and information structures. Section~\ref{sec_intro_cfes} assumes that the social planner (or decision-maker) has full access to agents' initial feature states and action spaces, and can compute a minimal-cost set of actions that lead to a desirable decision outcome. In contrast, Section~\ref{sec_intro_steer} considers a setting with limited observability, where the social planner lacks access to agents' initial feature states and actions. Instead, the planner provides limited signals indicating whether nearby targets, such as role models or exemplars, are positive or negative, thereby steering agents toward improved outcomes through emulating adjacent positive targets.

Consider a tax filing scenario where a taxpayer (i.e., agent) facing a challenging situation has various strategies (i.e., targets) to choose from, but doesn't know which are compliant or non-compliant. A tax-compliance body that wants to increase the expected number of individuals who choose compliant strategies, subject to a limited outreach budget, distributes materials such as short booklets that highlight a subset of strategies, illustrating how to handle common tax situations that most taxpayers will likely encounter. These examples may include both positive or compliant strategies (e.g., disclosing reportable transfers) and non-compliant or negative ones (e.g., misclassifying gifts to avoid taxes). When filing, the taxpayer consults the booklet and decides on which strategy to follow from their collection. They avoid strategies presented negatively, follow those presented positively if any are available, and otherwise select randomly from the remaining options. We therefore address the following research question in Chapter~\ref{chap:revealrm}:

\begin{center}
\textit{Which targets should the social planner highlight (i.e., reveal their labels) to maximize social welfare, defined as the expected number of agents who choose to emulate positive targets?}
\end{center}

While addressing this question with a greedy approach, we found that the disclosure policy affected the submodularity of the social welfare function and, consequently, the approximation guarantees of the approach. That is, when the social planner is restricted to revealing positive targets, then submodularity is preserved, and the greedy algorithm achieves the classic  $(1-1/e)$-approximation guarantee. Once negative targets are allowed, submodularity may fail, and the algorithm can perform arbitrarily poorly. To restore the guarantee, we introduce a proxy welfare function that remains submodular even when negative targets can be revealed. Additionally, when all agents have at most $c$ negative target neighbors, we show that this proxy achieves a constant-factor approximation to the true optimal welfare gain. Unlike the approximation guarantees, which are affected by the disclosure policy, we found that the problem of maximizing social welfare remains NP-Hard regardless of the restriction on the nature of targets that the social planner can reveal. Lastly, we extend the standard model to settings in which agents may belong to multiple groups, have negative local neighborhoods, or be unaware of positive targets in their neighborhood. Here, we formalize fairness guarantees and introduce algorithms for two complementary interventions: (1) directly linking at-risk agents most likely to emulate negative targets with positive targets, and (2) expanding the reach of positive targets so that more nearby agents can discover and emulate them.

The key takeaways from Chapter~\ref{chap:revealrm} are that performance guarantees depend critically on the chosen disclosure policy, and that greedy target-selection strategies are not always sufficient, as they can leave some agents underserved while disproportionately benefiting others. Despite these limitations, greedy approaches often perform well in practice when label disclosure is unconstrained, possibly because real-world social networks tend to be more balanced.

\section{Part~III: Shaping Human-AI  Interactions to Jointly Optimize Competing Objectives}
\label{sec:intro_part111}

In Part~III, we study how to shape human-AI interactions to jointly optimize the objectives of both the agents being evaluated and of the AI-driven decision systems. First, we analyze how the agents' capacity for improvement affects learnability, sample complexity, and algorithm design for achieving accurate classification (Section~\ref{sec:intro_pac}). 
Second, we develop a theoretical framework for the binary classification of improving and gaming agents, introducing efficient learning-theoretic formulations and algorithms for general discrete and linear models (Section~\ref{sec:intro_ocagi_theory}). 
Lastly, we empirically evaluate the effectiveness of algorithmic strategies that account for both improvement and gaming behavior, using a (weighted) utility function defined as the (weighted) number of true positives minus false positives (Section~\ref{sec:intro_ocagi_exp}).

\subsection{PAC Learning with Improvements}
\label{sec:intro_pac}

Thus far, we have examined how agents can be incentivized to improve and supported in achieving favorable decision outcomes. In this section (further developed in Chapter~\ref{chap:paclearn}), we shift focus to how agent-AI interactions can be shaped from the perspective of the decision maker. 
In particular, we study how agents' capacity for improvement impacts learnability, sample complexity, and algorithm design for accurate classification.

Although each agent's response set is constrained by its local neighborhood in Sections~\ref{sec_intro_steer} and~\ref{sec:intro_pac}, Section~\ref{sec:intro_pac} considers a stronger informational setting in which the learner directly controls the labeling function and can make it fully transparent to agents, rather than indirectly revealing information about it through guidance on who to emulate. Since in this section we assume that agents can observe the AI-driven decision model and have the incentive and ability to improve their features and alter their true labels, this naturally raises the following question:

\begin{center}
\emph{How does agents' capacity for improvement impact learnability, sample complexity, and algorithm design for accurate classification?}
\end{center}

To address this research question, Chapter~\ref{chap:paclearn} begins by comparing PAC learning with improvements to standard and strategic PAC learning \citep{strategicPAC}, and then investigates the conditions under which agents' capacity for improvement can either reduce the sample complexity of learning or, conversely, make learning more challenging. Specifically, we analyze several illustrative cases, including one-sided thresholds under uniform and arbitrary distributions, intersection-closed hypothesis classes, and homogeneous halfspaces under the uniform distribution on the unit ball. In these settings, PAC learning with improvement can achieve substantial error reductions (even zero error), whereas standard PAC models remain constrained by a constant error bound when learning the same target. Finally, we empirically study improvement-aware algorithms that explicitly account for agents' limited ability to improve strategically when doing so serves their interests.

The key takeaways from Chapter~\ref{chap:paclearn} are that agents' capacity for improvement tends to favor algorithms that employ more risk-averse decision-making strategies. This is because while there is less concern about false negatives since agents initially classified as negative may later improve and be correctly classified as positive, there is greater concern about false positives, which could inadvertently encourage agents to ``improve''  incorrectly.
Importantly, agents' capacity for improvement allows for the possibility of achieving a zero classification error, often unattainable under standard models.

\subsection{Theory of Classification of Improving-and-Gaming Agents} 
\label{sec:intro_ocagi_theory} 

This section develops a theoretical framework for the binary classification of strategic agents who can both game and improve. This setting differs from prior work on the classification of strategic and improvable agents~\citep{harris2021stateful, Haghtalab2020MaximizingWW, Kleinberg2018HowDC} in two important ways. First, rather than optimizing for total improvement, our objective is to maximize the number of true positives while minimizing false positives. Second, in the linear setting, we adopt a different model for agents' movement and costs. Within this framework, we address the following question:

\begin{center}
\emph{When agents can both game and improve, how can decision-makers design classifiers that accurately identify which of an agent's revealed features should contribute to a favorable classification, thereby maximizing true positives (e.g., approve good loans and incentivize improvable agents to improve) while minimizing false positives (e.g., reject bad loans)?}
\end{center}

To answer this question, Chapter~\ref{chap:ocagi_theory} studies general discrete and linear models of strategic classification. In the general discrete model, the decision problem is represented by a weighted, colored bipartite graph, with agents on the left-hand side and criteria on the right. Each edge corresponds to an action taken by an agent to satisfy a criterion, and edge weights capture the associated costs. Blue edges represent genuine improvements, red edges represent gaming behavior. The graph includes only actions whose costs are below the agent's value for receiving a positive classification. The objective is to select a subset of criteria such that, when each agent chooses a minimum-cost action, the resulting classification maximizes true positives while minimizing false positives, for example, approving many good loans and few bad ones. To achieve this goal, we develop algorithms both with and without limiting the number of selected criteria. We show that identifying a set of criteria that maximizes true positives while achieving zero false positives is NP-hard. In both the full- and partial-information learning settings, we derive bounds on the number of agents needed to ensure that, with high probability, the probability mass of true positives is close to the maximum achievable under the constraint of zero false positives, and the probability mass of false positives is small.

The linear model represents each agent by a feature vector and uses a linear threshold function with non-negative weights to separate truly qualified agents from unqualified ones.  Agents may be initially qualified, unqualified but unimprovable, or unqualified and improvable. Agents may increase feature values at a cost, while decreases are free, and receive a value of one if classified as positive. Some features correspond to genuine improvements, while others enable gaming. When agents modify their features, true qualification depends on the combination of genuine improvements and gaming, whereas the classifier observes only reported feature values. The objective remains to maximize true positives while minimizing false positives. We show that a linear classifier exists that correctly classifies all agents while also qualifying all improvable agents. In a 2D feature space, we provide algorithms that compute the optimal linear classifier when both features are improvable or gameable, and to maximize true positives while ensuring minimal or zero false positives when one feature is gameable and the other is improvable.

The key takeaway from Chapter~\ref{chap:ocagi_theory} is that maximizing the number of true positives while maintaining a nonzero bound on false positives is NP-hard, a complexity that persists even in a finite point version of the linear model.

\subsection{Empirical Study on the Classification of Improving-and-Gaming Agents} 
\label{sec:intro_ocagi_exp} 

The work presented in this section, and examined in greater detail in Chapter~\ref{chap:ocagi_exp1}, complements the theoretical analysis of Section~\ref{sec:intro_ocagi_theory} with empirical evidence. We empirically study the classification of agents who can truthfully (improve) and fraudulently (game) move within an $\ell_{\infty}$ or $\ell_{2}$ ball of radius $r$, with $r$ representing the movement budget. 
We evaluate algorithmic designs that account for improvement-and-gaming behavior, focusing on two practical risk-averse approaches: loss-based and threshold-based strategies. These practical strategic classification models align with the theoretical linear modeling framework introduced in Section~\ref{sec:intro_ocagi_theory}, which seeks to design classifiers that accurately classify all individuals, penalize gaming, and incentivize all improvable agents to become qualified. Consistent with these objectives, we investigate the following research question:

\begin{center}
\emph{What kind of practical improvement-gaming-aware algorithmic designs effectively maximize the decision-maker's weighted utility, defined as the weighted sum of true positives minus the weighted sum of false positives, in contexts where agents can both game and improve?}
\end{center}

Chapter~\ref{chap:ocagi_exp1} addresses this question through extensive empirical evaluation on three real-world tabular benchmarks (Adult, OULAD, and Law School) and a synthetic eight-dimensional binary classification dataset.
The empirical findings reveal an important interplay between classification accuracy, weighted utility, the degree of risk aversion, and the geometric constraints governing agent movement. Specifically, increasing risk aversion significantly reduces classification error, but excessive risk aversion can lower overall utility scores. Second, achieving zero classification error does not guarantee a perfect utility score. Third, 
(weighted)utility scores generally improve as agents utilize their available improvement budgets, with risk-averse models consistently outperforming non-strategic models. Fourth, unlike unweighted utility, the weighted utility score is highly sensitive to the ratio of true positives to false positives after agents move. Finally, the geometric constraints induced by the $\ell_{\infty}$ and $\ell_{2}$ norms play a significant role in shaping the direction and extent of feasible agent movements.

The main takeaway of Chapter~\ref{chap:ocagi_exp1} is that improvement-gaming-aware algorithmic designs, particularly loss-based risk-averse strategies, can effectively maximize (weighted)utility. However, stronger forms of risk aversion and geometric constraints on agent movement may reduce (weighted)utility scores. These findings highlight the importance of carefully balancing robustness to strategic behavior with overall utility, emphasizing the role of cost-benefit analysis in the design of improvement-gaming-aware algorithms.

\section{Bibliographical Remarks}
This thesis presents collaborative work with several researchers I had the privilege of working with, most of whom I met through my thesis committee. Unless marked with an asterisk (*), co-authors are listed alphabetically. 
Chapter~\ref{chap:parental_concern} is based on *Naggita, Athiley, Desta, and Sebo (\citeyear{parentalresponse}).
Chapter~\ref{chap:setfair} is based on Ahmadi, Beyhaghi, Blum, and Naggita (\citeyear{ahmadi2022settingfairincentivesmaximize}). 
Chapter~\ref{chap:cfes} is based on *Naggita, Walter, and Blum (\citeyear{learningactionablecounterfactualexplanations}).
Chapter~\ref{chap:revealrm} is based on Blum, Naggita, Walter, and Wang (\citeyear{Blum2026RevealingPA}).
Chapter~\ref{chap:paclearn} is based on Attias, Blum, Naggita, Saless, Sharma, and Walter (\citeyear{Attias2025PACLW}).
Chapter~\ref{chap:ocagi_theory} is based on Ahmadi, Beyhaghi, Blum, and Naggita (\citeyear{ahmadi2022classificationstrategicagentsgame}), and 
Chapter~\ref{chap:ocagi_exp1} is based on thesis complementary work by *Naggita, Walter, and Blum.

\clearpage
\thispagestyle{empty} 
\begin{center}
    {\Huge\bfseries
    Part~I \\[1.5em]
    Investigating Human-AI Interactions \\[0.3em]
    Through Human-Subject \\[0.6em]
    Experiments
    }
\end{center}
\chapter{Parental Responses to Aggressive Child Behavior Towards Robots, Smart speakers and Tablets}
\label{chap:parental_concern}
\section{Introduction} 
\label{sec:parental_intro}

The advancement and abundance of technological devices (e.g., smartphones, tablets, smart speakers, robots) has dramatically changed how children learn, play, and engage with family members. The use of these devices in the home can greatly benefit children by, for example, providing them with access to tutoring and enhanced learning \citep{Michaeliseaat5999,KONIJN2020103970,Ramachandran16} and tools that can enhance social skill use in children with Autism \citep{Scassellatieaat7544,Othman17}. The benefits of technological devices in the home should be considered, however, in light of their potential to change a family's social dynamics \citep{parenting_alexa,forlizzi2006service, forlizzi2007robotic}. 
The adoption of an Amazon Alexa smart speaker may, for example, have negative impacts on family dynamics by replacing some important parent-child interactions (e.g., singing lullabies, telling stories) with device-child interactions \citep{parenting_alexa}. However, it is also possible for a device to have a positive influence on family interactions, for instance, families that started using a Roomba vacuum robot incorporated more family members into the cleaning process than were involved before, especially men and children \citep{forlizzi2006service, forlizzi2007robotic}. 

When considering the incorporation of robots into homes, it is also important to examine the possibility of children behaving aggressively towards robots. The work of \citet{Brscic15} underscores the real possibility of children abusing a robot, who observed children in a shopping mall abusing a patrolling robot by obstructing its movements, calling it names (e.g., ``\textit{you idiot}''), and exhibiting physical violence (e.g., hitting the robot). Further highlighting the possibility of children expressing aggressive behavior towards devices in the home, there is growing anecdotal evidence of children expressing verbal aggression towards smart speakers \citep{alexa_making_us,alexa_making_kids_rude}. In response, some companies have developed politeness features for voice assistants' interactions with children (e.g., Amazon's Magic Word \citep{magic_word}, Google's Pretty Please \citep{pretty_please}). Although robots have similarities with more well-studied home devices (e.g., smart speakers), it is possible that a robot's distinct attributes (e.g., a human-like physical embodiment) may make child aggression towards a robot and parents' concern unique from other types of devices.

In this chapter, we seek to understand how parents view child aggressive behavior towards robots in the home compared with other technological devices. With the current limitations of the COVID-$19$ pandemic, we asked parents, recruited on a crowdsourcing platform, how they would respond to their child expressing aggressive behavior towards different devices. This study had a $2$ (presence of aggression: aggressive, neutral) $\times$ $3$ (device type: robot, smart speaker, tablet) $\times$ $3$ (interaction modality: audio, physical, audio+physical) between-subjects design. 
We asked participants to watch a video clip of an adult actress interacting with a technological device, according to the experimental condition, and imagine that their child exhibited the same behavior. 
We assess the influence of the presence of aggression, device type, and interaction modality on parents' questionnaire responses that reveal their concern, responses to their child's behavior, and perceptions of the device. 

\section{Background and Related Work}
\label{sec:parental_related}

Currently, researchers determine the level of closeness of technological devices to humans on the dimensions: anthropomorphism \citep{antrop_human,anthrop_social_robos}, human-like dialogue \citep{alexa_humanlike,bad_pa}, reciprocity/responsiveness \citep{phantom_friend}, among others.  \citet{antrop_human} describes anthropomorphism as the ``tendency for people to attribute human characteristics to non-lifelike artifacts.''  Researchers usually based the anthropomorphism of robots on their embodiment \citep{humanoid_robot,gen_emotions} and conversational agents on their ability to hold a human-like conversation \citep{phantom_friend}. Humanness varies across devices, and humanoid robots are usually the highest. \citet{Carlson2019PerceivedMA} showed that participants believed Nao had more capabilities for emotions than a laptop, and after assessing forum posts on devices on dimensions: life-likeness, emotional states, gender/personality, name, socially integrated, and metaphorical-ways, \citet{anthropolanguages} found that people anthropomorphize robotic pet (AIBO) significantly more than a functional robot (Roomba) or a tablet computer (iPad). In addition, \citet{owner_perception}'s analysis on the dimensions of anthropomorphism, animacy, likability, and perceived intelligence revealed lower ratings for the dog-like smart speaker than humanoid robots Nao and Pepper.  
How anthropomorphic a device is viewed also influences how people interact with it. As device anthropomorphism increases people treat them with higher agency and display higher levels of sympathy \citep{Carlson2019PerceivedMA} and frustration \citep{fyou_chatbot}.
However, to the best of our knowledge, no work has investigated how the human-like nature of devices affects parents' level of concern and likelihood to intervene with a repairing action when their child acts aggressively towards the robots, smart speakers, and tablets.

When considering interactions between children and technological devices in the home with varying levels of anthropomorphism (tablets, smart speakers, and robots), it is essential to address the possibility of the children being aggressive toward such devices. In this chapter, we define aggression as ``intentional harm to others''  \citep{Stangor2014PrinciplesOS}, and robot abuse as the ``persistent offensive action, verbal, nonverbal or physical violence that violates the role of the robot or its human-like (or animal-like) nature'' \citep{Brscic15}. Even though people of all ages harass and abuse technological devices, for the case of robots, several HRI studies have observed that unsupervised children seem to be the most inclined to display aggression towards robots, blocking, pushing and kicking them, and speaking rudely to them \citep{Brscic15, SalviniSafe}. 
Anecdotally, several parents have reported their children being rude, verbally abusive, and demanding towards smart speakers: Siri, Alexa, and Google assistants in a home \citep{alexa_making_kids_rude,alexa_making_us}. 
Additionally, \citet{behavior_time} noted a link between increased screen time and children's aggressive behavior. 
Despite increasing evidence that children do display aggressive behavior towards a wide array of technological devices, no work to our knowledge has systematically investigated differences in child aggressive behavior and parental reactions towards that behavior between several different device types.

\section{Methodology} 
\label{sec:parental_methodology}

To examine how parents perceive aggressive interactions of their children with technological devices in the home, we designed an online study in which parents watched an actress interacting with these devices, imagining it was their child. We chose an online format for this study due to the limitations of the COVID-19 pandemic. The study has a $2$ (presence of aggression: aggressive or neutral) $\times$ $3$ (interaction modality: physical, audio, or audio+physical) $\times$ $3$ (device type: robot, smart speaker, or tablet) between-subjects design.
This study was approved by the University of Chicago Social \& Behavioral Sciences Institutional Review Board (IRB20-1856).

\begin{table}[b!]
    \begin{center}
        \begin{tabular}{lccc}
        & \textbf{Robot} & \textbf{Smart Speaker} & \textbf{Tablet}\\
        \small{\textbf{a)}} & \includegraphics[width=0.29\textwidth]{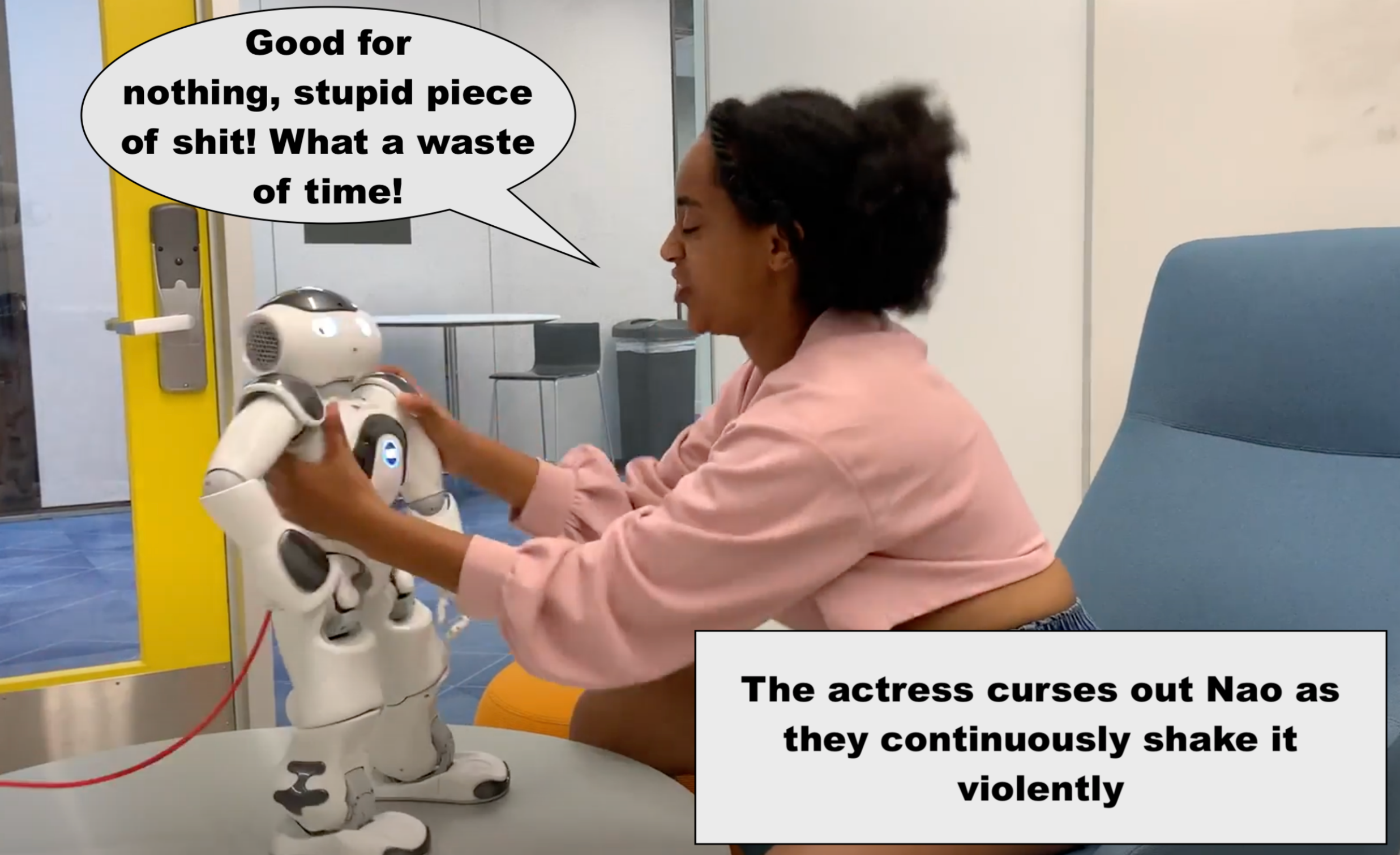}  & \includegraphics[width=0.29\textwidth]{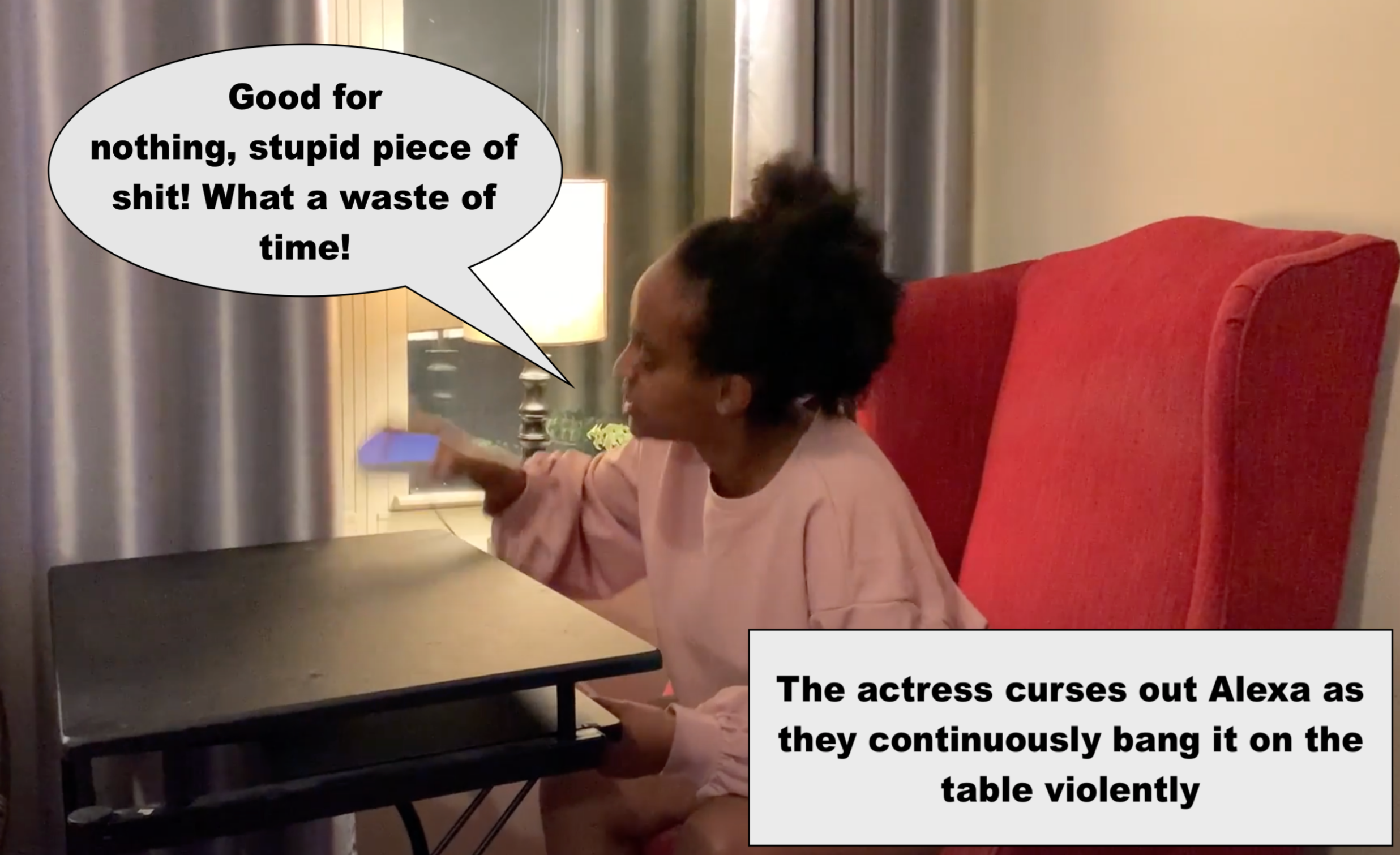} & \includegraphics[width=0.29\textwidth]{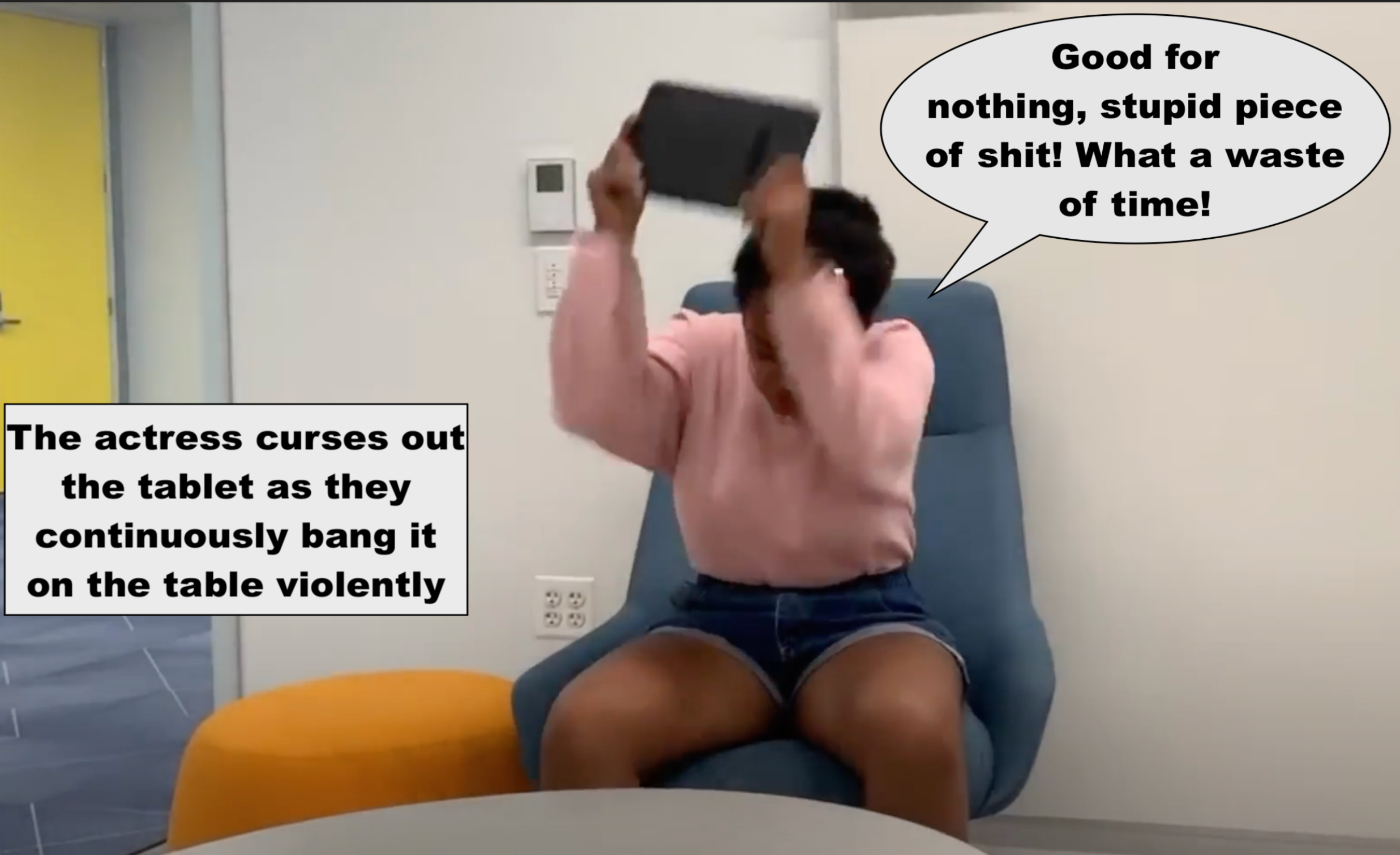}\\
        \small{\textbf{b)}} & \includegraphics[width=0.29\textwidth]{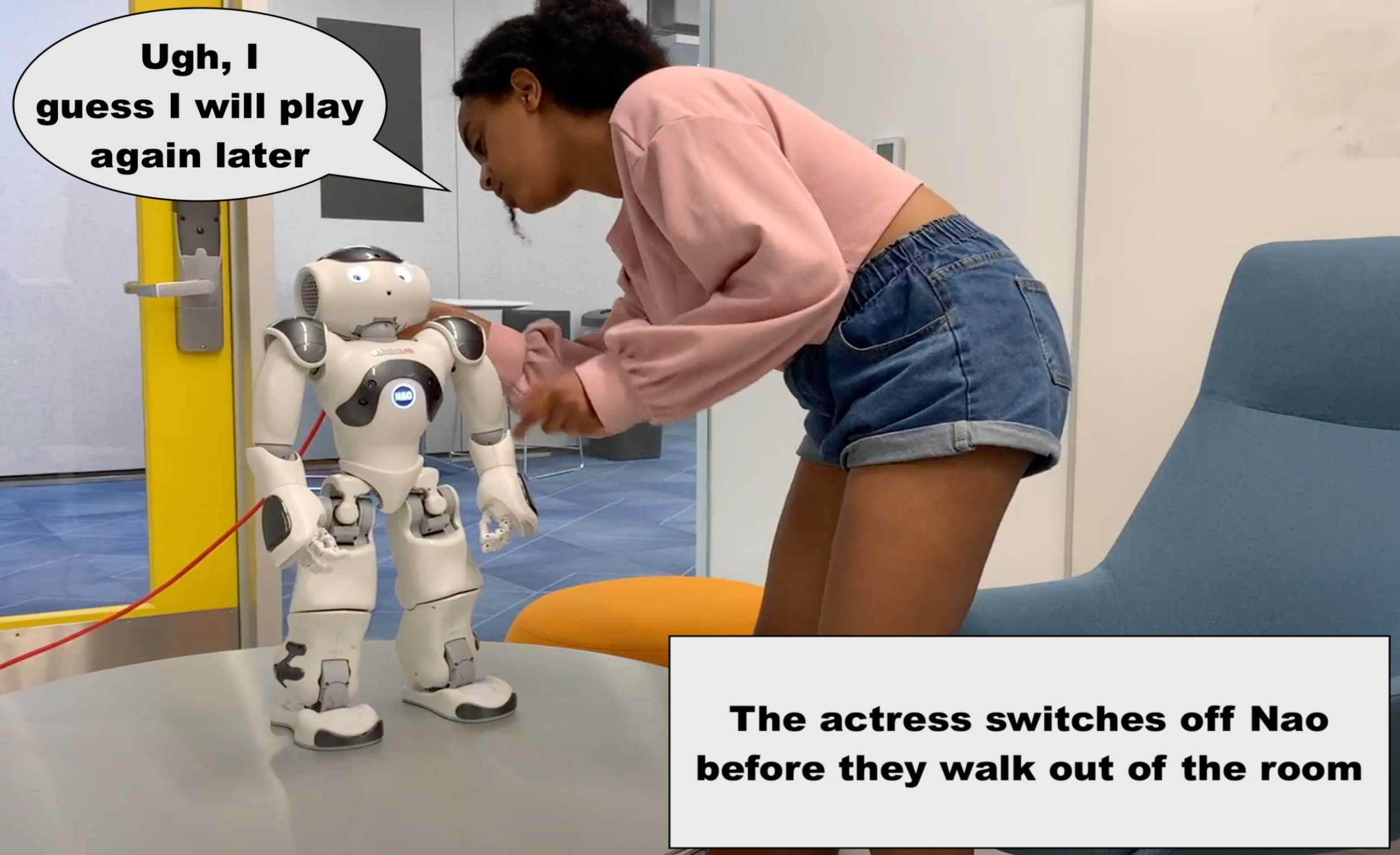}  & \includegraphics[width=0.29\textwidth]{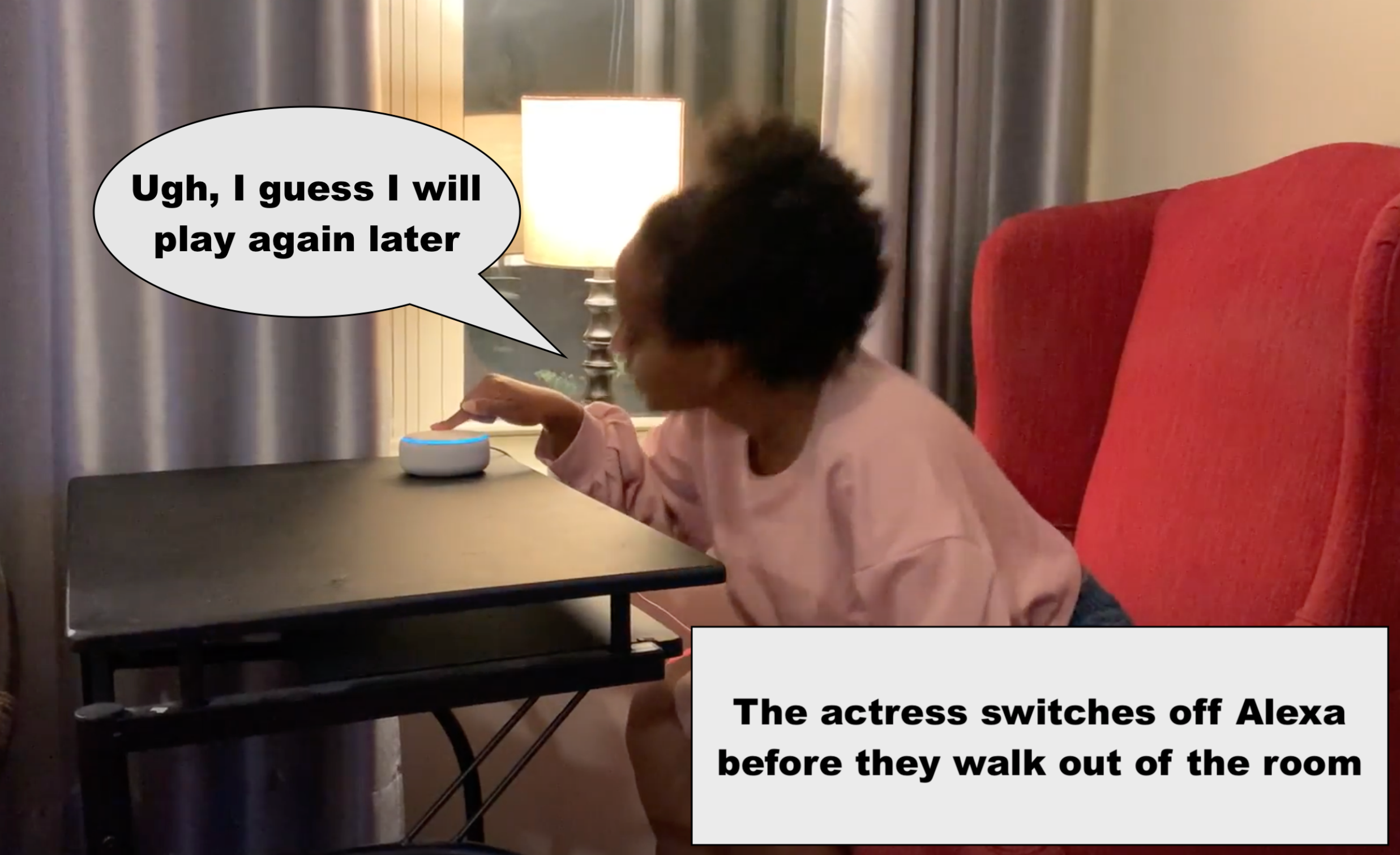} & \includegraphics[width=0.29\textwidth]{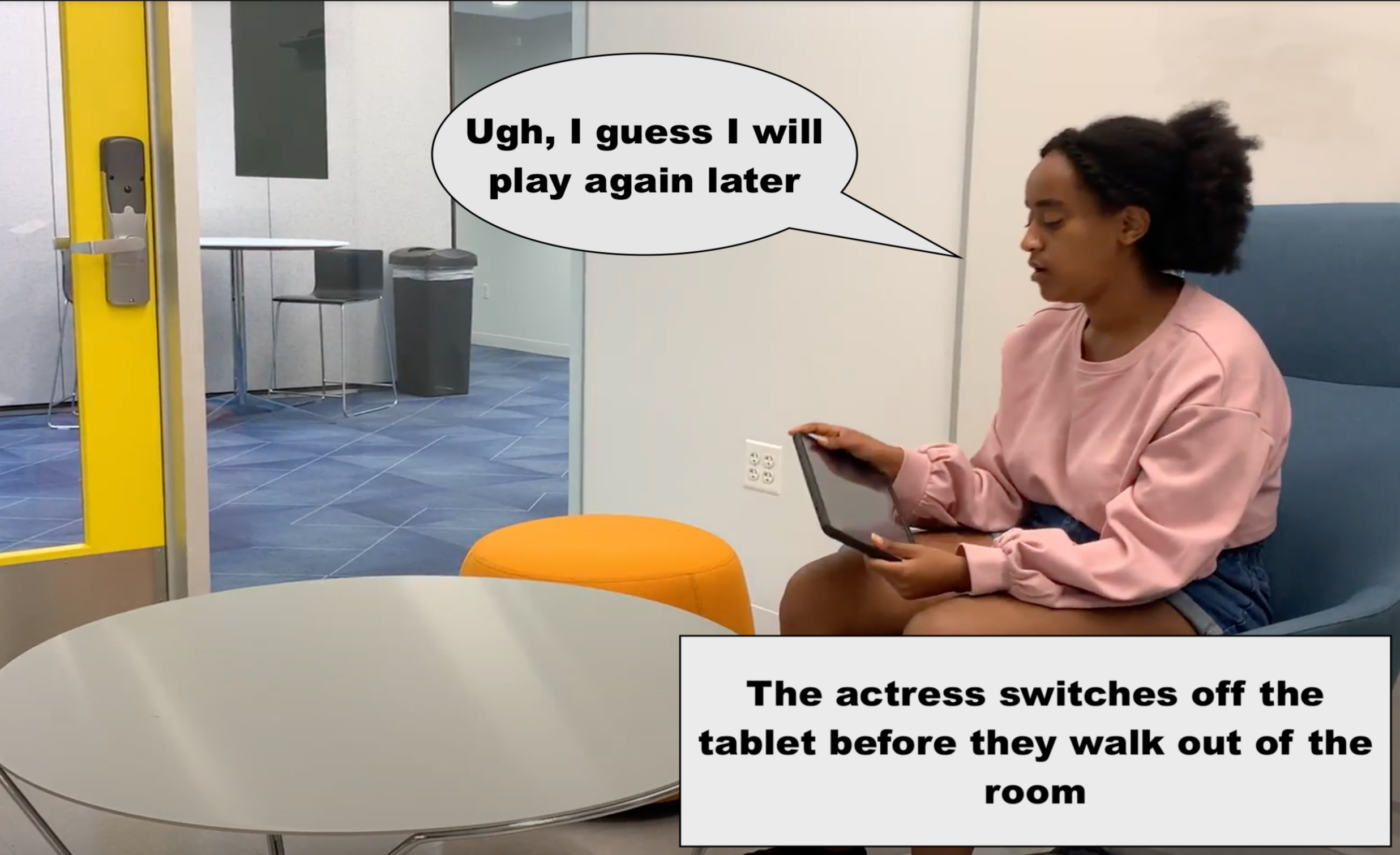}\\
        \end{tabular}
    \end{center}
    \caption[Differences in parental reactions between aggressive and neutral child behavior toward 3 device types]{This work examines differences in parental reactions between aggressive (a) and neutral (b) child behavior toward three device types (robot, smart speaker, tablet). These examples showcase the audio+physical interaction modality.}
    \label{table:inanimateobjs}
\end{table}

\subsection{Video Scenario} 
\label{subsec:parental_videos}

We used Amazon's pre-existing family-friendly ``Name That Animal'' game as the context for studying parent reactions and perceptions of child-robot behavior. We chose this game because it could be played on all three device types (robot, smart speaker, and tablet) and provided opportunities for aggressive behavior displays after a series of incorrect guesses. We filmed video clips of a $ 20$-year-old female, who we introduced as Rachael, playing the animal guessing game for each experimental condition (see Table \ref{table:inanimateobjs}). Each video began with Rachael walking into the room and initiating the animal guessing game by switching on the device and asking the device to play the ``Name That Animal'' game. The device then begins giving Rachael clues, that each contains more information to aid her in guessing the correct animal. For all of the experimental conditions, Rachael receives the same four clues from the devices and answers incorrectly after each clue. 
After the devices start to give her the fifth clue, Rachael's behavior then changes depending on the experimental condition.

\subsection{Experimental Conditions} 
\label{subsec:parental_exp_cond}

The study examines three factors that we hypothesize will shape parent reactions to child-device interactions: the presence of aggressive behavior (aggressive or neutral), the technological device type (robot, smart speaker, or tablet), and the interaction modality (audio, physical, or audio+physical). A human actress called Rachael plays an animal guessing game with a technological device. After four incorrect animal guesses, Rachel responds with aggressive or neutral behavior in one of three interaction modalities: audio, physical, or audio and physical.  
In the aggressive physical modality, Rachael angrily says ``\textit{ughh!}'' as she violently shakes (robot) or slams the device (smart speaker or tablet) three times before storming out of the room. In the aggressive audio modality, Rachael loudly curses at the device ``\textit{good for nothing, stupid piece of shit, what a waste of time!}'' before storming out of the room. The aggressive audio+physical modality combines the aggressive physical and audio modalities. 
In the neutral condition, Rachael loses interest in playing. She defeatedly says ``\textit{ughh!}'' as she switches off the device in the physical modality and in the audio modality, defeatedly says ``\textit{ugh, I guess I will again later}'' as she walks out of the room. The neutral audio+physical modality combines the neutral physical and audio modalities. 

We used the Softbank Robotics Nao robot for this study, programming it using the NAOqi Python API. We programmed the robot to make gestures while speaking and used its built-in text-to-speech. We used the Amazon Echo Dot (third generation) as the smart speaker for this study and its preexisting ``Name That Animal'' game. For the tablet condition, we used a Samsung Galaxy Tab A tablet and developed visual tablet screen displays for the game that included game icons, text instructions and clues, and speech button. The tablet condition utilized an online text-to-speech generator \citep{tablet_voice} with a female voice.

\subsection{Hypotheses} 
\label{subsec:parental_hypotheses}

We predict that parents would express concern (\textbf{H1a}) and be more likely to intervene (\textbf{H2a}) when imagining their child exhibiting aggressive behavior compared with neutral behavior. Supporting this prediction, anecdotal evidence demonstrates parental concern about children acting rudely towards smart speakers (e.g., Google, Alexa, Siri) \citep{alexa_making_us,alexa_making_kids_rude} and prior work has shown that families employ a variety of speech and language modifications in their attempts to repair communication breakdowns with devices \citep{parenting_alexa,parents_learn}.
Additionally, since prior work has demonstrated that people perceive a higher level of mistreatment for a robot than a computer \citep{Carlson2019PerceivedMA}, we anticipate that parental concern (\textbf{H1b}) and intervention (\textbf{H2b}) will be highest for a robot, then a smart speaker, and lowest for a tablet.

Since \citet{Carlson2019PerceivedMA} showed that people perceive higher levels of mistreatment of and had greater sympathy for a robot compared with a computer, we predict that participants will be more likely to perceive their child's behavior as mistreatment (\textbf{H3a}) and exhibit more sympathy (\textbf{H4a}) when the child exhibits aggressive behavior, versus neutral behavior, towards the devices. Additionally, we hypothesize that participants will perceive mistreatment (\textbf{H3b}) and exhibit more sympathy (\textbf{H4b}) for the devices in the following order: robot (highest) $>$ smart speaker (middle) $>$ tablet (least).

Lastly, \citet{Carlson2019PerceivedMA} has demonstrated that people view robots with more sympathy and as more emotionally capable than a computer and \citet{bad_pa} has shown higher ratings of anthropomorphism and animacy for a Nao robot compared with a Google Home. Therefore, we predict (\textbf{H5}) that participants will rate the devices on the dimensions of anthropomorphism, animacy, warmth, competence, and discomfort in the following order: robot (highest) $>$ smart speaker (middle) $>$ tablet (lowest).

\subsection{Protocol} 
\label{subsec:parental_protocol}
Participants were recruited on Prolific and directed to take a Qualtrics survey where they provided consent, viewed the video, and completed questionnaire items with interspersed attention checks. Participants first provided consent and their demographic information. Then they watched the video clip on their experimental condition, completed an attention check and answered questions specific to the video.
Each participant received $\$1.85$ for completion of the survey, which took approximately $17$ minutes.

\subsection{Measures}
\label{subsec:parental_measure}
In the questionnaire completed by participants, we collected participant demographics, participant reactions to and perceptions of the video they watched - imagining their child exhibited the same behavior, and how their children interact with technological devices in the home.

\begin{enumerate}[label=\arabic*)]
    \item \textbf{Demographics.}
    We gathered participants' age, gender, ethnicity, marital status, education, how many children they take care of, the ages and genders of their children, and whether or not they live with parents or older adults. 

    \item \textbf{Parental Concern.}
    We assessed parental concern by asking participants to indicate their agreement with ``I would be concerned if my child acted the way Rachael did'' on a Likert scale from 1 (\textit{Strongly Disagree}) to $5$ (\textit{Strongly Agree}). If they indicated agreement ($4$ - \textit{Somewhat Agree} or $5$ - \textit{Strongly Disagree}) we then asked the open-ended question, ``What specifically concerned you and why?'' Otherwise, we asked the open-ended question ``Why?''
    
    \item \textbf{Parental Responses and Interventions.}
    We measured parental responses by asking participants the open-ended question ``If your child were to act in that same way, how would you react? Why?'' We also asked them to indicate their agreement a Likert scale from $1$ (\textit{Strongly Disagree}) to $5$ (\textit{Strongly Agree}) for the following questions: ``If my child acted in that same way, I would take away [the device]'' and ``If my child acted in that same way, I would reprimand them.''  

    \item \textbf{Perceptions of Mistreatment and Sympathy.}
    To evaluate parent's perception device mistreatment and sympathy toward the device, we used the same measures as \citep{Carlson2019PerceivedMA}.  Parents answered yes/no to ``Do you feel [the device] was mistreated?'' Then on a Likert scale from $1$ (\textit{Strongly Disagree}) to $7$ (\textit{Strongly Agree}), parents rated ``If mistreatment is defined as verbal or physical behavior that meant to damage, insult, or belittle another, do you feel [the device] was mistreated?'' Finally, parents rated ``\textit{How sympathetic did you feel towards [the device]?}'' on a $7$-point Likert scale: $1$ (\textit{No Sympathy}), $3$ (\textit{Some Sympathy}), $7$ (\textit{Very Sympathetic}). 
    
    \item \textbf{Perceptions of the Devices.}
    To capture participant impressions of the device they witnessed in the video, we administered the RoSAS questionnaire \citep{Carpinella2017TheRS} to assess the devices' perceived warmth, competence, and discomfort using a $7$-point Likert scale. We also assessed participant's views of the device using the anthropomorphism and animacy components of the Godspeed questionnaire \citep{Godspeed} with a $5$-point Likert scale.

\end{enumerate}

\subsection{Participants} 
\label{subsec:parental_participants}

To determine our sample size, we conducted an a-priori power analysis using G*Power and the reported effect size of $\eta_p^2 = 0.06$ (computed from \citet{Carlson2019PerceivedMA}'s results of the interaction between agent and presence of aggressive behavior on the perception of mistreatment using the operational definition) and a power of $0.95$, resulting a target sample size of $296$ participants ($16.4$ participants / condition). Anticipating lower data quality using a crowdsourcing platform, we decided to recruit approximately $20$ participants for each of our $18$ conditions, totaling $360$ participants. 

We recruited a total of $370$ participants on Prolific who 
satisfied the following criteria: live in the U.S., speak fluent English, have a stable internet connection to watch the video, children, and a minimum approval rating of $95\%$ on the platform. We discarded the responses of $38$ participants who did not have children between ages $3$ and $12$ years. Among the $332$ remaining participants, there were between $17$ and $22$ participants in each condition, with an average of $18.44$ participants per condition ($SD = 1.54$). Participants ranged in age from $20$ to $68$ years ($M  = 34.34, SD  = 6.80$). $159$ were female, $170$ male, and $3$ were non-binary/other genders. $215$ participants indicated ownership of the device they viewed.

\section{Results} 
\label{sec:parental_res}

We examined participant impressions of the actor's interaction with a technological device as influenced by the presence of aggressive behavior by the actress (aggressive or neutral), the type of technological device (robot, smart speaker, or tablet), and the interaction modality (audio, physical, or audio and physical). We analyzed the data using an analysis of variance (ANOVA) examining each of independent variables of interest (presence of aggressive behavior, interaction modality, and type of technological device), their $2$-way interactions, and a set of covariates (gender, ethnicity, and whether or not the participant owns the device type they witnessed in the video) as fixed effects. Pairwise comparisons were evaluated using Tukey's honest significant difference tests and the effect size of each ANOVA is reported as partial eta squared ($\eta_p^2$).

\begin{figure}[b!]
\centering
  \subfloat{\includegraphics[width=5.2cm]{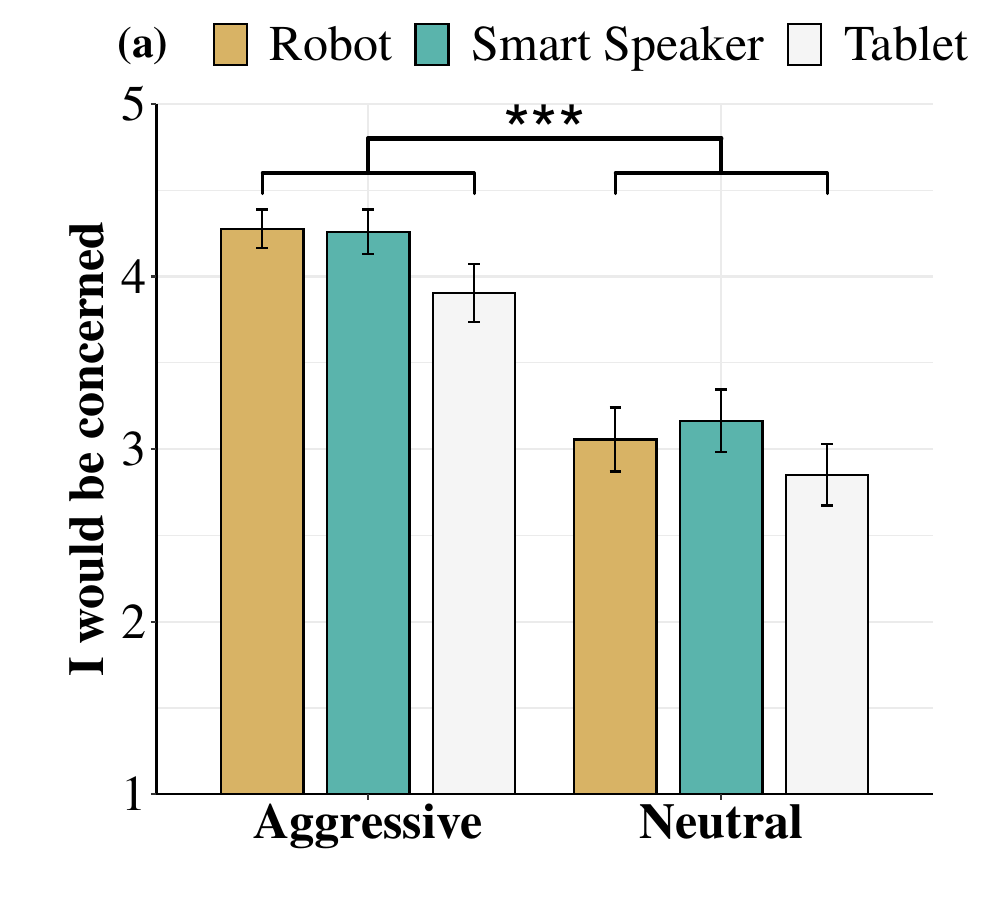}\label{fig:concern}}
  \subfloat{\includegraphics[width=5.2cm]{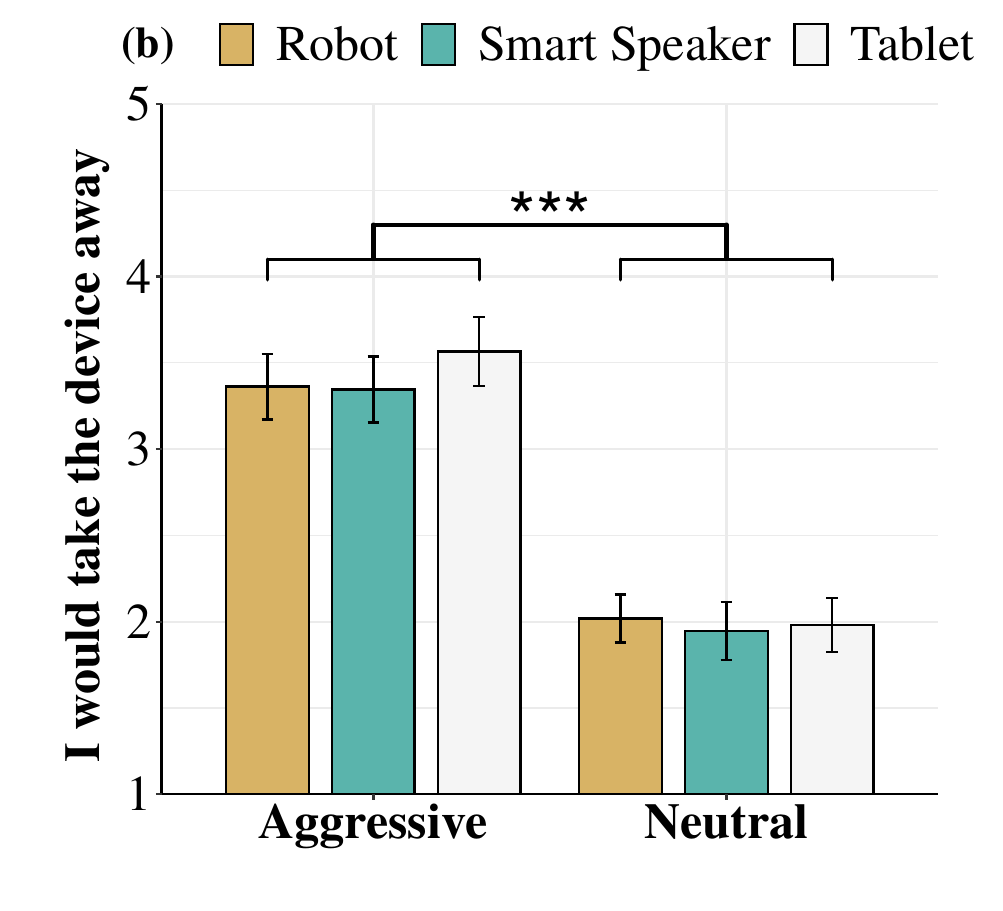}\label{fig:takeaway}}
  \subfloat{\includegraphics[width=5.2cm]{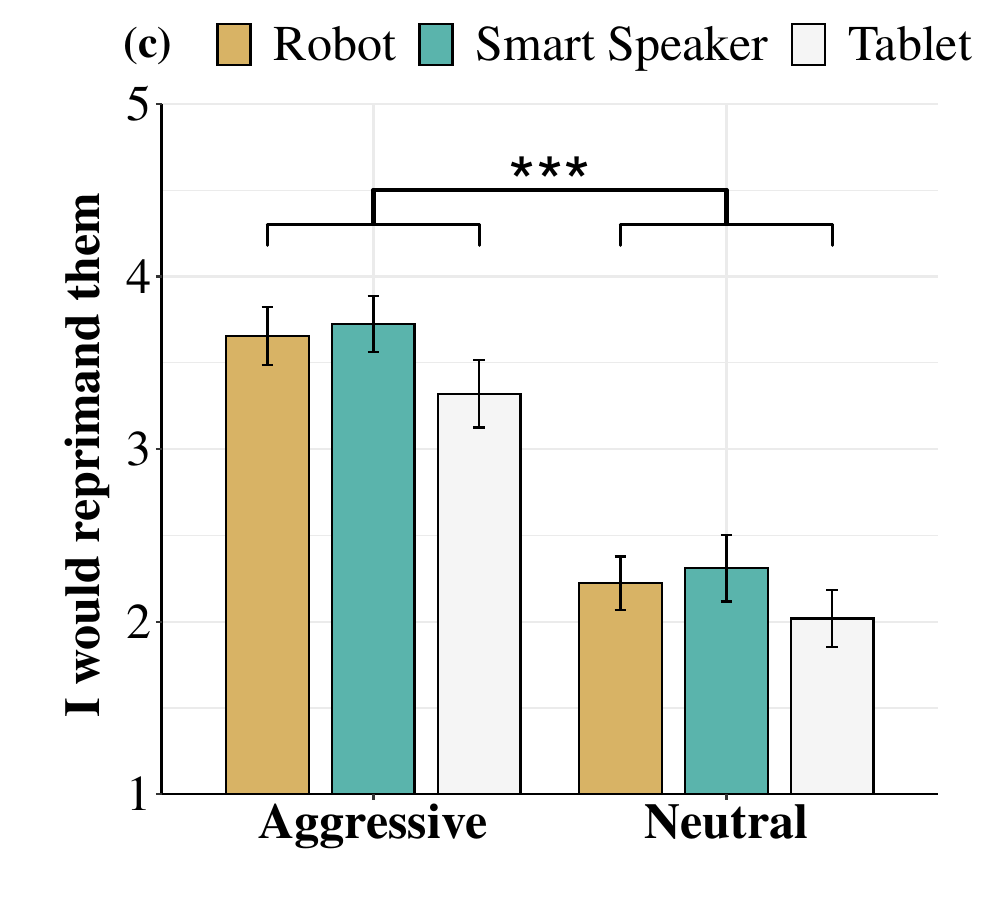} \label{fig:reprimand}}
\caption[Participants displayed higher levels of concern and likelihood of taking the device]{Imagining that their child expressed the same behavior as what they saw in the video, participants displayed (a) higher levels of concern, (b) a greater likelihood of taking the device away, and (c) a greater likelihood of reprimanding their child when aggressive behavior was expressed, compared with neutral behavior (*$p < 0.05$, **$p <0.01$ , ***$p < 0.001$).}
\label{fig:concern_takeaway_reprimand}
\end{figure}
\raggedbottom

\subsection{Parental Concern}
\label{subsec:parental_concern}

Parents ratings of concern about their child's behavior, imagining their child to have acted in the same way as the actress in the video, revealed a significant main effect for the presence of aggressive behavior ($F = 74.84, \eta_p^2 = 0.20, p < 0.001$), see Figure~\ref{fig:concern_takeaway_reprimand}(a). Participants that observed aggressive behavior expressed a significantly greater level of concern ($M = 4.15, SD = 1.04$) than those who observed neutral behavior ($M = 3.02, SD = 1.34$), \textbf{supporting H1a}. 

Additionally, we discovered a significant main effect for the interaction modality on parental concern ($F = 3.28, \eta_p^2 = 0.02, p = 0.039$) and a marginally significant main effect for the type of technological device on parental concern ($F = 2.90, \eta_p^2 = 0.01, p = 0.057$). However, pairwise comparisons did not reveal statistically significant differences between either interaction modalities or device types. 
For example, although participants indicated higher concern in interactions with both the robot ($M = 3.69, SD = 1.28$) and the smart speaker ($M = 3.73, SD = 1.29$) than compared with the tablet ($M = 3.37, SD = 1.37$), pairwise comparisons did not reveal significant differences between the three device types. Therefore, \textbf{H1b was not supported}.

For the $214$ participants who indicated concern ($4$ - \textit{somewhat agree} or $5$ - \textit{strongly agree} to ``I would be concerned if my child acted the way the person in the video clip did''), we examined their responses to the open-ended question, ``what specifically concerned you and why?'' We coded participant responses into $7$ categories of concern, verified by an independent coder on an overlap set of 20/214 responses with an average inter-rater reliability Cohen's kappa value of 0.78 across all categories. We found the following types of concern to be the most commonly mentioned by parents: concern about the lack of patience and perseverance ($29.4\%$), concern about the lack of emotional control ($24.3\%$), concern about the display of aggressive or rude behavior ($23.8\%$), and concern about the use of inappropriate language ($13.6\%$).

\subsection{Parental Reactions and Interventions} 
\label{subsec:parental_intervene}

Analysis of participants' indication to intervene by taking away the device reveal a significant main effect for the presence of aggressive behavior ($F = 102.40, \eta_p^2 = 0.23, p < 0.001$), see Figure~\ref{fig:concern_takeaway_reprimand}(b). Participants indicated a higher likelihood to take away the device in response to aggressive behavior ($M = 3.42, SD = 1.45$) than neutral behavior ($M = 1.98, SD = 1.14$), \textbf{supporting H2a}. Additionally, we found a significant interaction between the device type and interaction modality on participants' likelihood to intervene ($F = 2.56, \eta_p^2 = 0.03, p = 0.039$), however, pairwise comparisons did not reveal any statistically significant differences between specific combinations of device types and interaction modalities. There was also no significant main effect for device type, indicating no difference in the likelihood to take away the robot ($M=2.71, SD=1.42$), smart speaker ($M=2.66, SD=1.52$), and tablet ($M=2.77, SD=1.53$), demonstrating \textbf{a lack of support for H2b}.
\begin{figure}[b!]
\centering
\includegraphics[width=14cm]{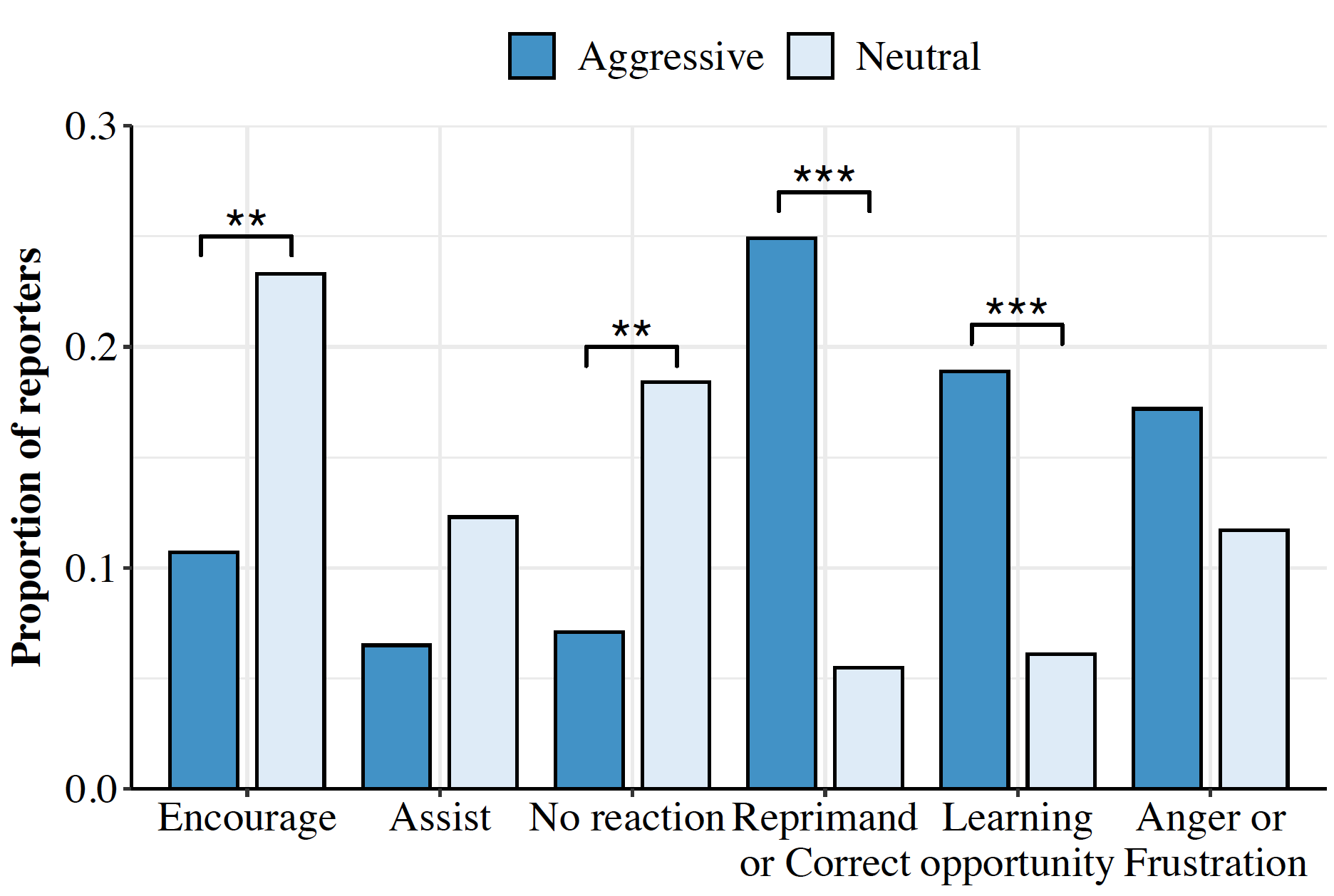}
\caption[Participants indicated significant differences in their responses]{Participants indicated significant differences in their responses to their child exhibiting aggressive versus neutral behavior towards the devices (**$p <0.01$ , ***$p < 0.001$).}
\label{fig:reactions}
\end{figure}

Similarly, intervening by reprimanding their child indicated a significant main effect for the presence of aggressive behavior ($F = 97.99, \eta_p^2 = 0.22, p < 0.001$), see Figure~\ref{fig:concern_takeaway_reprimand}(c). Parents indicated a higher likelihood to reprimand their child if they were aggressive ($M = 3.57, SD = 1.32$) than if they reacted neutrally ($M = 2.18, SD = 1.26$), further \textbf{supporting H2a}. 
We also found a marginally significant main effect for the type of device on participants' likelihood to reprimand their child ($F = 2.60, \eta_p^2 = 0.01, p = 0.075$), however, pairwise comparisons did not reveal any significant differences between the robot ($M = 2.96, SD = 1.41$), smart speaker ($M = 3.04, SD = 1.50$), and tablet ($M = 2.66, SD = 1.47$), showing further \textbf{lack of support for H2b}. 

To further understand how participants would respond if their child acted in similar ways to the actress in the video, we examined responses to the question ``if your child were to act in that same way, how would you react? why?'' We coded participant responses into the following categories: provide encouragement, provide assistance with the game, use the situation as a learning opportunity, express anger and frustration at child's reaction, reprimand or correct the child, and no reaction. These codings were verified by an independent coder on 21 of the 332 responses with an average inter-rater reliability Cohen's kappa value of 0.96 across all categories. Analysis of this data revealed the presence of aggressive behavior to be the largest determinant of the reactions participants listed (Figure~\ref{fig:reactions}). Participants who watched a video where the person displayed \textit{neutral} behavior were significantly or marginally significantly more likely to say they would respond by providing encouragement ($F = 10.56, \eta_p^2 = 0.02, p = 0.001, agg = 10.7\%, neu = 23.3\%$), providing assistance with the game ($F = 3.46, \eta_p^2 = 0.01, p = 0.064, agg = 6.5\%, neu = 12.3\%$), and providing no reaction ($F = 10.14, \eta_p^2 = 0.03, p = 0.002, agg = 7.1\%, neu = 18.4$). 

In contrast, participants who watched a video where the person displayed \textit{aggressive} behavior were significantly more likely to say that they would respond by using the situation as a learning opportunity ($F = 12.55, \eta_p^2 = 0.04, p < 0.001, agg = 18.9\%, neu = 6.1\%$) and reprimanding or correcting the child  ($F = 24.87, \eta_p^2 = 0.08, p < 0.001, agg = 24.9\%, neu = 5.5\%$). And while more participants in the aggressive condition than the neutral condition reported that they would respond by expressing anger and frustration at the child's reaction ($F = 1.94, \eta_p^2 = 0.01, p = 0.165, M_{agg} = 0.17, SD_{agg} = 0.38, M_{neu} = 0.12, SD_{neu} = 0.32$), the difference was not statistically significant. These responses highlight the different methods parents use to help their children in frustrating situations, and how these methods differ based on the level of aggression in the child's behavior.

\subsection{Perceptions of Mistreatment and Feelings of Sympathy} 
\label{subsec:parental_mistreat}

\begin{figure}[t!]
\centering
\subfloat{\includegraphics[width=7cm]{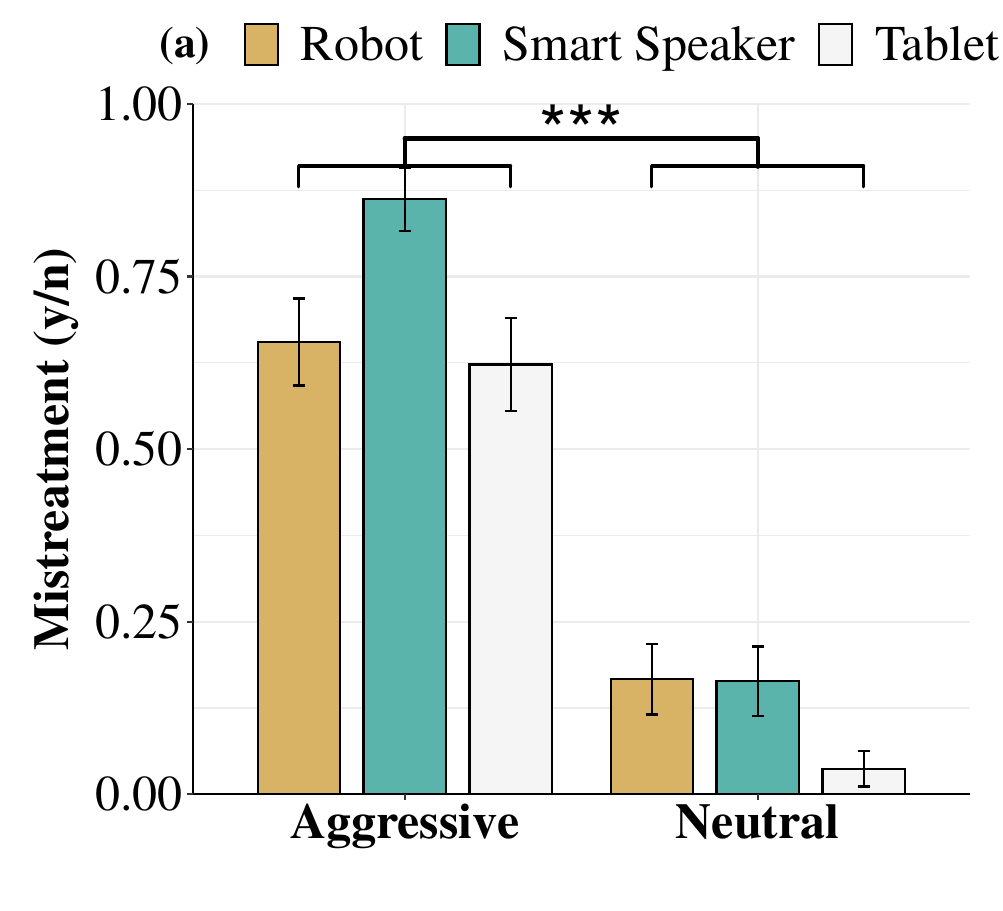}\label{fig:mistreated}}
~~
\subfloat{\includegraphics[width=7cm]{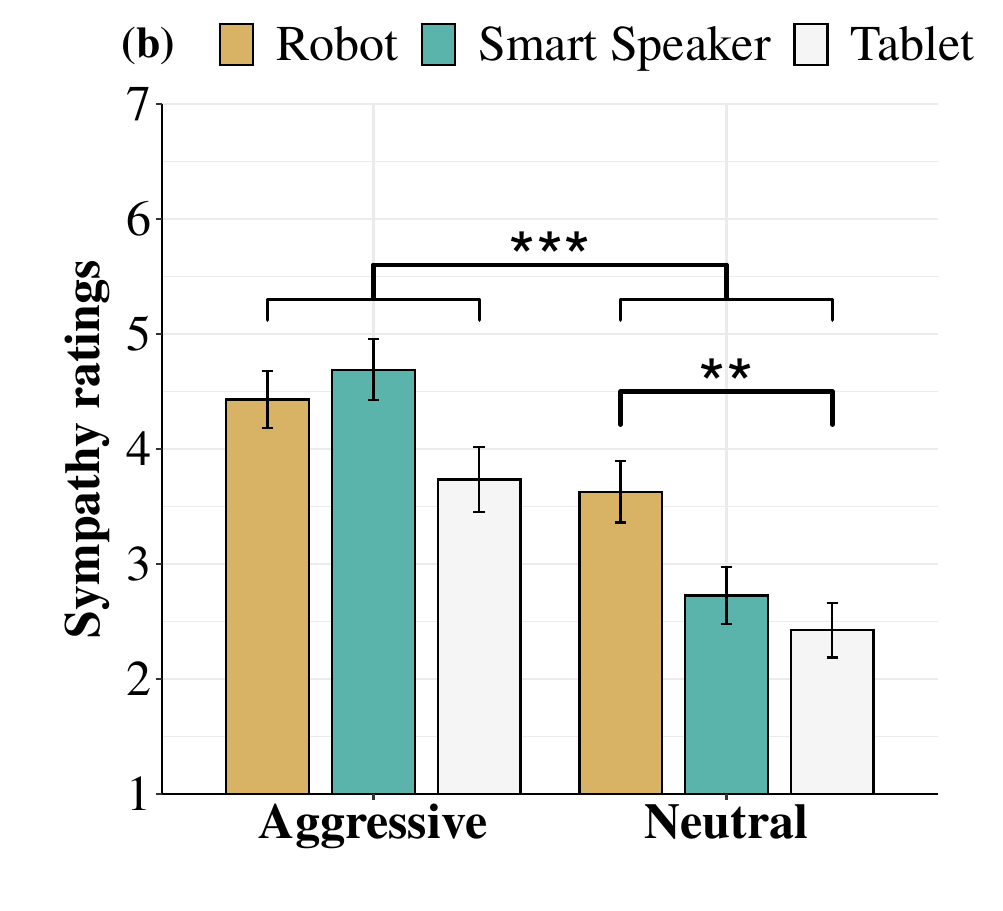}\label{fig:sympathy}}
\caption[Parents (a) perceived greater device mistreatment and (b) felt more sympathetic]{Parents (a) perceived greater device mistreatment and (b) felt more sympathetic towards the devices when aggressive behavior was exhibited (**$p <0.01$, ***$p < 0.001$).}
\label{fig:mistreat_sympathy}
\end{figure}

Participants' perceptions of device mistreatment indicated a significant main effect for the presence of aggressive behavior ($F = 198.19, \eta_p^2 = 0.38, p < 0.001$), see Figure~\ref{fig:mistreat_sympathy}(a), where participants had a higher perception of mistreatment in the aggressive condition ($M = 0.72, SD = 0.45$) than in neutral condition ($M = 0.12, SD = 0.33$), \textbf{supporting H3a}. We also found a significant main effect for device type on perceptions of mistreatment ($F = 7.16, \eta_p^2 = 0.04, p < 0.001$), where participants viewed the smart speaker as the most mistreated ($M = 0.52, SD = 0.50$), then the robot ($M = 0.42, SD = 0.50$), and lastly the tablet ($M = 0.33, SD = 0.47$). Pairwise comparisons revealed only a significant difference between the smart speaker and tablet ($p<0.001$), thus, \textbf{H3b was only partially supported}. Additionally, there was a significant interaction between device type and interaction modality on perceptions of mistreatment ($F=2.51, \eta_p^2 = 0.03, p = 0.042$), where pairwise comparisons revealed that the tablet-audio condition ($M = 0.15, SD = 0.36$) had significantly lower perceptions of mistreatment than both the robot-audio+physical ($M = 0.44, SD = 0.50, p = 0.010$) and the smart speaker-audio+physical conditions ($M = 0.61, SD = 0.50, p < 0.001$). Finally, there was also a significant interaction between the presence of aggression and interaction modality ($F=3.77, \eta_p^2 = 0.02, p = 0.024$), where within the aggressive/neutral conditions, the only significant pairwise comparison was between the aggressive-audio ($M = 0.61, SD = 0.49$) and aggressive-audio+physical conditions ($M = 0.85, SD = 0.36, p = 0.008)$.

When mistreatment was defined as damage, insult, or belittling, participants' ratings for mistreatment of the device indicated a significant main effect for the presence of aggressive behavior ($F = 251.22, \eta_p^2 = 0.43, p < 0.001$), where participants had higher ratings of mistreatment in response to aggressive behavior ($M = 5.43, SD = 1.66$) than neutral behavior ($M = 2.56, SD = 1.67$), \textbf{supporting H3a}. We also found a significant main effect for device type on ratings of mistreatment ($F = 3.92, \eta_p^2 = 0.02, p = 0.021$), where participants viewed the smart speaker as the most mistreated ($M = 4.28, SD = 2.30$), then the robot ($M = 4.109, SD = 2.17$), and lastly the tablet ($M = 3.67, SD = 2.10$). Pairwise comparisons revealed only a significant difference between the smart speaker and tablet, showing \textbf{partial support for H3b}. Finally, we found a significant main effect for the interaction modality on perceptions of mistreatment ($F = 3.37, \eta_p^2 = 0.02, p = 0.036$), where pairwise comparisons revealed that precipitants viewed the device as more mistreated in the audio+physical modality ($M = 4.32, SD = 2.27$) than the audio modality ($M = 3.78, SD = 2.18, p = 0.035$). No pairwise comparisons with the physical modality ($M = 3.96, SD = 2.14$) were statistically significant.

Responses for feeling sympathetic towards the devices revealed significant main effect for the presence of aggressive behavior ($F = 47.57, \eta_p^2 = 0.13, p < 0.001$), see Figure~\ref{fig:mistreat_sympathy}(b), where participants felt more sympathetic when aggressive behavior was displayed ($M = 4.30, SD = 2.01$) compared with neutral behavior ($M = 2.93, SD = 1.92$), \textbf{supporting H4a}. We also found a significant main effect for the device type on device sympathy ($F = 8.28, \eta_p^2 = 0.04, p < 0.001$), where pairwise comparisons revealed that participants had significantly less sympathy for the tablet ($M = 3.07, SD = 2.01$) than both the robot ($M = 4.04, SD = 1.95, p < 0.001$) and smart speaker ($M = 3.73, SD = 2.17, p = 0.019$). There was no significant difference between sympathy towards the robot and smart speaker, thus \textbf{H4b was only partially supported}. We also observed a interaction effect between the presence of aggression and device type on device sympathy ($F = 3.13, \eta_p^2 = 0.02, p = 0.045$), where pairwise comparisons within the aggressive and neutral behavior types reveal a significant difference between the robot-neutral condition ($M = 3.63, SD = 1.96$) and the tablet-neutral condition ($M = 2.43, SD = 1.74, p = 0.004$), where all other pairwise comparisons were not statistically significant.

\begin{figure}[t!]
\centering
\subfloat{\includegraphics[width=7cm]{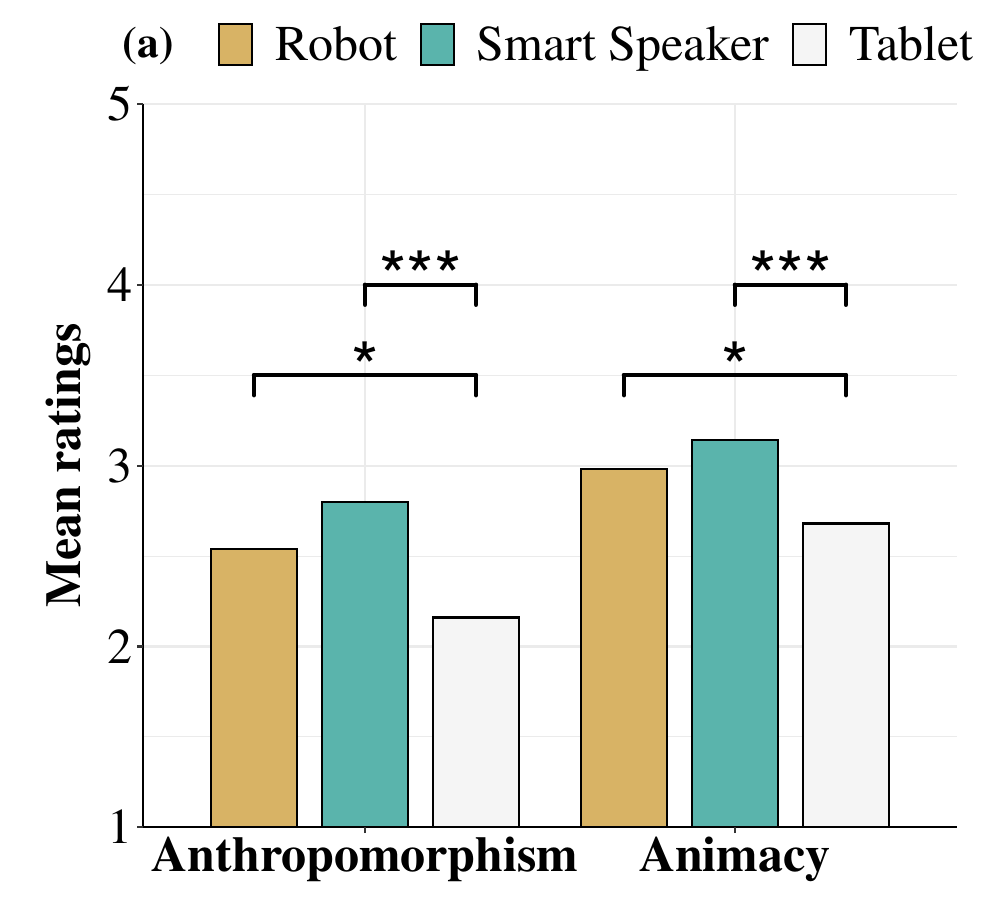}\label{fig:anthropo_anim}}
~~
\subfloat{\includegraphics[width=7cm]{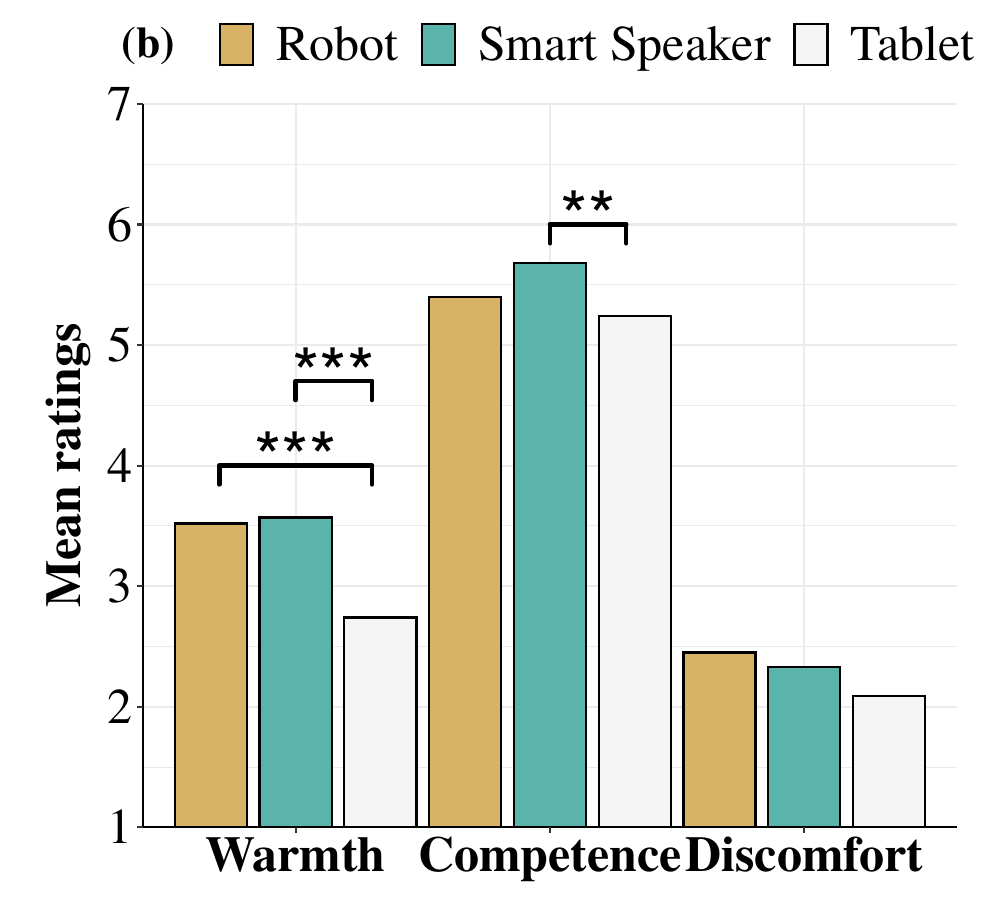}\label{fig:rosas_scale}}
\caption[Participants rated the three device types on (a) anthropomorphism and animacy]{Participants rated the three device types on (a) anthropomorphism and animacy, and (b) warmth, competence, and discomfort (*$p < .05$, **$p < .01$, ***$p < .001$).}
\label{fig:attributes}
\end{figure}

\subsection{Perceptions of the Devices}
We analyzed participant ratings of the devices' attributes of anthropomorphism, animacy, warmth, competence, and discomfort (Figure~\ref{fig:attributes}).
Participants' ratings of \textit{anthropomorphism} revealed significant main effect for the type of device ($F = 11.45, \eta_p^2 = 0.08, p < 0.001$), where they rated the tablet ($M = 2.16, SD =1.00$) as significantly less anthropomorphic than both the robot ($M = 2.54, SD = 1.17, p = 0.013$) and smart speaker ($M = 2.80, SD = 1.16, p < 0.001$), with no significant difference between the robot and smart speaker. 
Participants' ratings of \textit{animacy} also exhibited significant main effect for the type of device ($F = 7.23, \eta_p^2 = 0.05, p < 0.001$), where they rated the tablet ($M = 2.68, SD =1.01$) as significantly less animate than both the robot ($M = 2.98, SD = 0.98, p = 0.045$) and smart speaker ($M = 3.14, SD = 1.04, p < 0.001$), with no significant difference between the robot and smart speaker.

\newpage
The \textit{Warmth} ratings showed significant main effect for the type of device ($F = 11.93, \eta_p^2 = 0.08, p < 0.001$). Participants also rated the tablet ($M = 2.74, SD =1.63$) as significantly less warm than the robot ($M = 3.52, SD = 1.49, p < 0.001$) and the smart speaker ($M = 3.57, SD = 1.56, p < 0.001$), with no significant difference between the robot and smart speaker. 
\textit{Competence} ratings indicated statistical significance main effect for the type of device ($F = 4.63, \eta_p^2 = 0.03, p = 0.010$), where participants rated the smart speaker ($M = 5.68, SD =0.97$) and the robot ($M = 5.40, SD = 0.97$) as more competent than the tablet ($M = 5.24, SD =1.31$), however, only the pairwise comparison between the tablet and the smart speaker was significant ($ p<0.01$). Additionally, we found a significant interaction between the presence of aggression and device type on competence ratings ($F = 4.53, \eta_p^2 = 0.03, p = 0.012$), where pairwise comparisons revealed that when aggressive behavior is displayed, participants see the smart speaker as significantly more competent ($M = 5.93, SD = 0.70$) than the tablet ($M = 5.08, SD = 1.41, p < 0.001$). All other pairwise comparisons were not statistically significant.
Lastly, parents ratings of \textit{discomfort} did not indicate any significant main effects or interaction effects. 

Taking all of the device perception ratings together, we had hypothesized that the robot would be highest in anthropomorphism, animacy, warmth, and competence and lowest in discomfort. Instead, we observed similar ratings between the smart speaker and robot, where both were viewed as more anthropomorphic, animate, and warm than the tablet. Thus, we \textbf{do not have support for H5}.

\section{Discussion and Conclusion} 
\label{sec:parental_disc}

In this chapter, we examined the influence of the presence of aggression (aggressive or neutral), the type of device (robot, smart speaker, or tablet), and the interaction modality (audio, physical, or audio and physical) on reported parent responses if their child displayed that behavior and parents' perceptions of the technological device. Confirming our predictions, our results indicate that parents have greater concern, are more likely to intervene, and perceive the device as mistreated and attribute more sympathy to it after imagining that their child exhibited aggressive behavior.

We also observed differences in how parents reported they would respond if their child exhibited the same behavior they saw in the video. Parents who watched a video where the actress displayed neutral behavior were more likely to respond by encouraging their child to keep trying and by providing assistance in the game. On the other hand, parents who watched a video where the actress displayed aggressive behavior were more likely to respond by treating the situation as a learning opportunity and reprimanding or correcting their child.  
These results demonstrate that parents intervene and assist their children differently depending on the severity of the aggressive behavior displayed by their child.

Although our results showed strong support for parents' concern and likelihood to intervene when aggressive behavior is displayed, we did not find results supporting a higher level of parental concern and intervention when the device was a robot compared with the smart speaker and the tablet. It is possible that the lack of difference in parental concern and intervention between the robot and the other two device types could be driven by the lack of perceived differences between some of the device types. 
While participants viewed the robot and smart speaker as being more anthropomorphic, animate, and warm when compared with the tablet, we did not observe significant differences between the robot and smart speaker on these dimensions.
We were surprised by this lack of distinction between the robot and smart speaker, because we had hypothesized that the increased human-like physical appearance of the robot would result in higher ratings of sympathy, anthropomorphism, animacy, and warmth; and therefore, also result in a higher amount of parental concern and likelihood to intervene when aggressive behavior is displayed toward the device. Although we did control for whether or not the participant owned the device they viewed in the video in our statistical analysis, it is possible that the greater use and exposure to smart speaker devices in the United States, compared with robots like Nao, could have led to similar ratings between robots and smart speakers. 
It is also possible that if the robot had more human-like movement, anthropomorphism and animacy ratings might have been higher for the robot than the smart speaker.

Additionally, it is important to recognize that there could be differences between A) parents watching their children acting aggressively toward a robot in-person and B) watching a video of an actress acting aggressively toward a robot and imagining their child exhibited the same behavior, as we implemented in this work.  Prior work has demonstrated significant differences in how people respond to a physical robot embodiment compared with a telepresent or virtual robot embodiment \citep{bainbridge2011benefits}.
While we were not able to explore in-person parental reactions to aggressive child behavior towards robots due to the COVID-$19$ pandemic, future work is needed to ascertain whether parental reactions are similar, as our results suggest, between child aggressive actions towards different devices in the home.

The increased incorporation of tablets, smart speakers, and robots into the home are changing family interactions and leading to instances of aggressive child behavior towards these devices. This work underscores the importance of critically examining the causes and potential types of aggression children might have towards different types of devices to inform design decisions that can encourage fewer instances of aggressive behavior from children and help parents mitigate any instances of child aggression towards home devices.

\clearpage
\thispagestyle{empty} 
\begin{center}
    {\Huge\bfseries
    Part~II \\[1.5em]
    Shaping Human-Human and \\[0.3em]
    Human-AI  Interactions to \\[0.6em]
    Provide Improvement Pathways
    }
\end{center}
\chapter{Setting Reachable Targets to Maximize People's Improvement}
\label{chap:setfair}
\section{Introduction}
One of the principal practices in policy making is setting reasonable expectations for the groups or individuals involved in the policy. Whether it is in the context of public policy, education, or career development, a too low expectation (i.e., an easy-to-achieve goal) may be a cause for not improving at one's capacity, while a too high expectation (i.e., an out-of-reach goal) may be discouraging to even make an attempt. To accommodate different levels of participants' abilities, the policy-maker may consider different levels of expectations and design goals at various levels so that all (or most of) the participants have a goal within their reach but not too easily achieved. At the other side from the policy-maker are the participants who may view their goal as a mere requirement for accessing other benefits, e.g., increasing their chances of promotion or gaining freedom to pursue other opportunities. In this case, the individuals may choose an easy-to-achieve goal rather than aim for maximal improvement. Examples of such expectations include reading goals for youth, language proficiency goals for applicants, outreach activities for employers, etc.

In this chapter, we consider the problem of helping agents improve by setting reachable targets that maximize improvement. Given a set of target ``skill levels'', we assume each agent will try to improve from their initial skill level to the closest target level within reach (or do nothing if no target level is within reach). The designer's goal is to maximize the total improvement both with and without fairness considerations.

Mathematically, we formulate this problem as follows. There are $n$ agents belonging to $g$ distinct groups. Agent $i$ has an initial skill level, $p_i \in \mathbb{Z}_{\geq 0}$, and can increase their skill by at most $\Delta_i$ which is called its ``improvement capacity''. Given a set of target levels $\T \subset \mathbb{Z}_{\geq 0}$, agent $i$ improves to the closest target $\tau \in \T$ such that $\tau > p_i$ and $\tau \leq p_i+\Delta_i$ if such target exists; otherwise it stays at $p_i$\footnote{We assume the policy-maker can disallow agent $i$ from choosing a target $\tau \leq p_i$.
For example, in designing reading goals for grade-school students, the fifth graders are not allowed to choose materials from the second-grade level, although the reverse is allowed. Setting the base for each agent at their true skill level is an abstraction of our mathematical model.}.

\begin{figure}[ht!]
    \centering
    \begin{subfigure}[b]{0.5\textwidth}
        \centering
        \includegraphics[height=1.6cm]{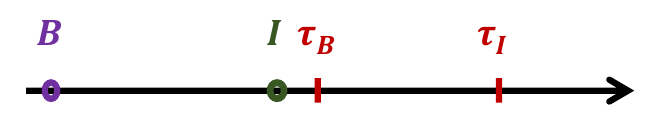}
        \caption{Conflict in optimizing improvement.}
        \label{fig:interfere}
    \end{subfigure}
    \hfill
    \begin{subfigure}[b]{0.45\textwidth}
        \centering
        \includegraphics[height=1.6cm]{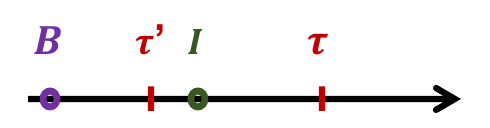}
        \caption{Non-monotonicity in set of target levels.}
        \label{fig:nonmonotone_sub}
    \end{subfigure}
    \caption{Challenges in designing optimal target levels.}
    \label{fig:conflict_and_nonmonotone}
\end{figure}

This problem formulation gives rise to multiple challenges. First, optimizing improvement for a set of agents may conflict with another set. Consider a beginner-level agent (skill level $B$) and an intermediate-level (skill level $I$). Agent $I$ finds any level up to $\tau_I$ within reach. Therefore, we need to design a project at level $\tau_I$ for this agent to improve maximally. On the other hand, $B$ has the capacity to improve until $\tau_B$, where $I < \tau_B < \tau_I$ --- See Figure~\ref{fig:interfere}. 
Now, consider both target levels $\tau_B$ and $\tau_I$. Since agent $I$ now has a closer target of $\tau_B$, this agent no longer achieves its maximum improvement, {and only reaches skill level $\tau_B$.} 
Secondly, there is non-monotonicity in the placement of target levels, i.e., adding a new target to the current placement may decrease the total amount of improvement. Consider a beginner-level ($B$) and an intermediate-level ($I$) agent and a target, $\tau$, achievable by both agents --- See Figure~\ref{fig:nonmonotone_sub}. 
Designing a new project at level $\tau'$ between $B$ and $\tau$ decreases the total amount of improvement since one agent (if $B <\tau'\leq I$) or both agents (if $I<\tau'<\tau$) switch from improving to $\tau$ to improving to $\tau'$, which requires less improvement.

\paragraph{Main Results} 
In this chapter, we consider algorithmic, fairness, and learning-theoretic formulations, where a set of optimal target levels must be found in the presence of effort-bounded agents. {We use \textit{social welfare} as the notion of efficiency and define it as the total amount of improvement. Also, we define \textit{social welfare for a given group} as the amount of improvement that group achieves.} We consider two models: (1) the common improvement capacity model, where agents have the same limit $\Delta$ on how much they can improve, and (2) the individualized improvement capacity model, where agents have individualized limits $\Delta_i$. 

The main results of the chapter are:
\begin{enumerate}
    \item An efficient algorithm for placement of target levels to maximize social welfare. (Section~\ref{sec:setfair_max_total_improvement})
    \item An efficient algorithm for outputting the Pareto-optimal outcome for the social welfare of multiple groups. In particular, this can output the max-min fair solution that maximizes the minimum total improvement across groups. (Section~\ref{sec:setfair_max_min}) 
    \item A structural result on Pareto-optimal solutions: there exists a placement of target levels that simultaneously is approximately optimal for each group. {More explicitly,  when there are a constant number of groups, the total improvement for each group is a constant-factor approximation of the maximum  improvement that we could provide that group if it were the only group under consideration. \textbf{This is our main contribution.}} (Section~\ref{sec:setfair_approx_optimality})     
    \item An efficient learning algorithm for near-optimal placement of target levels. (Section~\ref{sec:setfair_gen_guarantees})
\end{enumerate}

The algorithmic results work for both the common and individualized improvement capacity models. However, the structural result only holds in the common improvement capacity model, and we illustrate examples where achieving any nontrivial fraction of optimal for all groups is not possible in the individualized capacity model.

\paragraph{Related Work.}
Our work broadly falls under two general research areas: social welfare maximization in mechanism design and algorithmic fairness. Specifically, the closest topics to this chapter are designing portfolios for consumers to minimize loss of returns \citep{Diana-etal}, designing badges to steer users' behavior \citep{steering-user-beha}, and the literature on strategic classification.

Closest to our work is \citet{Diana-etal} who consider a model where each agent has a risk tolerance, observed as a real number, and must be assigned to a portfolio with risk lower than what they can tolerate. The goal of the mechanism designer is to design a small number of portfolios that minimizes the sum of the differences between the agent's risk tolerance and the risk of the portfolio they take; in other words, it minimizes the loss of returns. Since this is a minimization problem where each agent selects the closest target (portfolio) below their risk tolerance, adding any new target can only help with the objective function. Therefore, unlike our model, there is no conflict between targets, and the objective function is monotone in the set of targets.

Designing targets to incentivize agents to take specific actions is also a common feature of online communities and social media sites. In these platforms, there is a mechanism for rewarding user achievements based on a system of \emph{badges} (similar to targets in our model) \citep{steering-user-beha, gamification, babaioff2012bitcoin, burke2009feed, burke2011plugged}. Among such papers, the closest to ours is \citet{steering-user-beha} who investigate how to optimally place badges in order to induce particular user behaviors, among other things. They consider a dynamic setting with a single user type interested in a particular distribution of actions and a mechanism designer whose objective is to set badges to motivate a different distribution of actions. Compared to our work, their model is more general in the sense that users can spend effort on different actions (improve in multiple dimensions), but also more specific, in the sense that there is only one user type; therefore, unlike our model there is no conflict between different users and adding more badges for the desired action always helps with steering the users in that direction (it is a monotone setting).

Another line of work that is relevant is strategic classification. In most cases, agents are fraudulently strategic, that is to say, game the decision-making model to get desired outcomes (see \citep{hardt2016strategic,revealed_preferences,Hu:2019:,Milli2018TheSC,strategicperceptron,adversarial_games_pred,Frankel2019ImprovingIF,braverman_et_al} among others). In other cases, in addition to actions only involving gaming the system, agents can also perform actions that truthfully change themselves to become truly qualified (see \citep{Kleinberg2018HowDC, harris2021stateful,Alon2020MultiagentEM,xiao2020optimal,Miller2019StrategicCI,Haghtalab2020MaximizingWW, Bechavod2020CausalFD,Shavit2020LearningFS} among others). In this chapter, we assume agents only truthfully change themselves and, therefore, focus on incentivizing agents to improve as much as they can.

\paragraph{Organization of this Chapter.}
{Section~\ref{sec:setfair_model} formally introduces the general model settings and definitions used in the chapter, and Section~\ref{sec:setfair_max_total_improvement} provides an efficient algorithm for the problem of maximizing total improvement.
In Section~\ref{sec:setfair_max_min}, we provide algorithms that output Pareto optimal solutions for groups' social welfare, including a solution that maximizes the minimum improvement per group.
In Section~\ref{sec:setfair_approx_optimality}, we provide an algorithm that finds the best simultaneously approximately optimal improvement per group and show it provides a constant approximation when the number of groups is constant.
In Section~\ref{sec:setfair_gen_guarantees}, we provide efficient learning algorithms which generalize the previous results to a setting where there is only sample access to agents, and Section~\ref{sec:setfair_extensions} provides further extensions to our main problems.
All missing proofs are deferred to the appendix.}

\section{Model and Preliminaries}
\label{sec:setfair_model}

There are $n$ agents $1, \ldots, n$. Agent $i$ is associated with two quantifiers: initial skill level, $p_i$, and \textit{improvement capacity}, $\Delta_i$, which determines the maximum amount agent $i$ can improve its skill. For the majority of the chapter, we assume $p_i$ and $\Delta_i$ belong to $\mathbb{Z}_{\geq 0}$; however, some of our results hold more generally for real numbers\footnote{All our examples that do not use integer numbers can be converted to integer numbers by scaling.}.

We consider two different models. The \textit{common} and the \textit{individualized} improvement capacity models. In the first model, all agents have the same improvement capacity, i.e., $\Delta_i$ are equal across agents; we substitute $\Delta_i$ with $\Delta$ in this case. The second model is a generalization where $\Delta_i$ may have different values. We use \[\Delta_{max} = \max\{\Delta_1,\cdots,\Delta_n\}.\]

Our solution is a finite set of target levels $\T \subset \mathbb{Z}_{\geq 0}$. We assume we are given a maximum number of allowed target levels $k$ (if $k=n$, this is equivalent to allowing an unbounded number of target levels).

\paragraph{Agents Behavior.} Given target levels $\T \subset \mathbb{Z}_{\geq 0}$, agent $i$ aims for the closest target above its initial skill if it can reach to that target given its improvement capacity. More formally, agent $i$ aims for $\min \{\tau \in \T : p_i<\tau\leq p_i+\Delta_i\}$ if such $\tau$ exists and improves from $p_i$ to $\tau$. If no such target exists, agent $i$ does not improve and its final skill level remains the same as the initial skill level $p_i$. 

We use \textit{social welfare} ($\sw$) as our notion of efficiency and define it as the total amount of improvement of agents.

\paragraph{Groups and Fairness Notion.} Each agent belongs to one of $g$ distinct groups {$G_1,\cdots, G_g$}. Given any set of target levels, the social welfare of group $\ell$, $\sw_\ell$, is defined as the total amount of improvement for agents in that group\footnote{Although the results are presented for the \emph{total} improvement objective, they also hold for the \emph{average} improvement objective.}.
We are interested in Pareto-optimal solutions for groups' social welfare. A solution $\T$ is Pareto-optimal (is on the Pareto frontier) if there does not exist $\T'$ in which all groups gain at least as much social welfare, and one group gains strictly higher. In particular, the Pareto frontier includes the max-min solution that maximizes the minimum social welfare across groups.  In this chapter, we focus on two natural fairness notions: one is the max-min solution described above, and the other is the notion of simultaneous approximate optimality given below.

\begin{definition}[Simultaneous $\alpha$-approximate optimality] 
A solution with at most $k$ targets is simultaneously approximately optimal for each group with approximation factor $0 \leq \alpha \leq 1$ if, for each group $\ell$, the social welfare of group $\ell$ is at least an $\alpha$ fraction of the maximum social welfare achievable for group $\ell$ using at most $k$ targets.
\end{definition}

\subsection{Basic Properties of Optimal Target Sets}
\label{sec:setfair_basics}

This section provides a simple structural result on optimal set of target levels. The following observation determines the potential positions of the targets in an optimal solution.

\begin{observation} \label{obs:potential_targets}
Without loss of optimality, the targets in an optimal solution are either at positions $p_i + \Delta_i$ or $p_i$ for some $i \in \{1, 2, \ldots, n\}$. Consider a solution where target $\tau$ does not satisfy this condition. By shifting $\tau$ to the right as long as it does not cross $p_i + \Delta_i$ or $p_i$ for any $i$, the total amount of improvement weakly increases: This transformation does not change the sets of agents that reach each target, and only increases the improvement of agents aiming for $\tau$.
\end{observation}

Observation~\ref{obs:potential_targets} motivates the following definition. 

\begin{definition}[$\T_p$]
The set of potential optimal target levels, $\T_p$, is defined as \\ $\bigcup_{i=1}^n\{p_i, p_i+\Delta_i\}$.
\end{definition}

\section{Maximizing Total Improvement }
\label{sec:setfair_max_total_improvement}

In this section, we provide an efficient dynamic programming algorithm for finding a set of $k$ target levels that maximizes total improvement 
for a collection of $n$ agents. Algorithm~\ref{alg:recurrence-dp-one-group} 
{provides the details} of the dynamic programming algorithm. We bound its time-complexity in Theorem~\ref{thm:total-improvement}.

In Algorithm~\ref{alg:recurrence-dp-one-group}, the recursion function $T(\tau,\kappa)$ finds the best set of at most $\kappa$ target levels {for} agents on or to the right of $\tau$. 
Recall that any target $\tau$ only affects the agents on its left, and agent $i$ such that $p_i < \tau$ never selects $\tau' > \tau$ in presence of $\tau$. Utilizing these properties, the 
main idea for the recursive step (item $3$ in Algorithm~\ref{alg:recurrence-dp-one-group}) is to first consider the potential leftmost targets $\tau' > \tau$ and use the smaller subproblem of finding the optimal targets for agents on or to the right of $\tau'$ with one less available target level; i.e., $T(\tau', \kappa-1)$.  
To optimize over the potential leftmost target levels, $\tau'$, we first evaluate the performance of each potential target by improvement of agents who reach it; i.e., $i$ such that $\tau \leq p_i < \tau'$ and $\tau'-p_i\leq \Delta_i$, where agent $i$ improves by $\tau'-p_i$. Next, we add the performance of each potential leftmost target to the optimal improvement of the remaining subproblem and pick the leftmost target that maximizes this summation.

\paragraph{}
\makebox[\textwidth][c]{%
\begin{minipage}{1.06\textwidth} 
\setlength{\algoheightrule}{0pt}
\setlength{\algotitleheightrule}{0pt}
\begin{algorithm}[H]
  \caption[Find a set of \(k\) target levels to maximize total improvement for \(n\) agents]
  {Run dynamic program based on function $T$, defined below, that takes $\cup_i \{p_i\}$ and $k$ as input and outputs $T(\tau_{\min},k)$, as the optimal improvement, and $S(\tau_{\min},k)$, as the optimal set of targets; where $\tau_{\min}= \min\{\tau \in \T_p\}$ and $\tau_{\max}= \max\{\tau \in \T_p\}$. $T(\tau,\kappa)$ captures the maximum improvement possible for agents on or to the right of $\tau\in \T_p$ when at most $\kappa$ target levels can be selected. Function $T$ is defined as follows.}
  \SetAlgoLined
    \label{alg:recurrence-dp-one-group}

    \paragraph{}
    \begin{enumerate}[leftmargin=1em, rightmargin=2cm]
    \justifying
    \item[1)]  For any $\tau\in \T_p$, 
    $ \; T(\tau,0)=0$.
    
    \item[2)]  For any $1\leq \kappa\leq k$, $ \; T(\tau_{\max},\kappa)=0$.
    
    \item[3)]  For any $\tau\in \T_p, \tau< \tau_{\max}$ and $1\leq \kappa\leq k$:
    
    \[T(\tau,\kappa) = \max_{\tau'\in \Tau_p \ \text{s.t} \ \tau' > \tau}\Bigg(T(\tau',\kappa-1) \ + \sum_{\tau\leq p_i <\tau'\text{ s.t. } \tau'-p_i \leq\Delta_i}(\tau'-p_i)\Bigg)\]
    \end{enumerate}
    
    $S(\tau,\kappa)$ keeps track of the optimal set of targets corresponding to $T(\tau,\kappa)$.
\end{algorithm}
\setlength{\algoheightrule}{1pt} 
\setlength{\algotitleheightrule}{1pt} 
\end{minipage}
}
\paragraph{}

The following theorem proves the correctness of the dynamic programming algorithm and bounds its time-complexity.
\begin{theorem}
Algorithm~\ref{alg:recurrence-dp-one-group} finds a set of targets that achieves the optimal social welfare (maximum total improvement)
that is feasible using at most $k$ targets given $n$ agents. The algorithm runs in $\mathcal{O}(n^3)$.
\label{thm:total-improvement}
\end{theorem}
\begin{proof}
{See Appendix~\ref{app:setfair_missing_max_tot}.}
\end{proof}

\section{Pareto Optimality and Maximizing Minimum Improvement}
\label{sec:setfair_max_min}

In this section, we provide a dynamic programming algorithm that constructs the Pareto frontier for groups' social welfare. By iterating through all Pareto-optimal solutions, we can find the solution that maximizes minimum improvement across all groups in pseudo-polynomial time. Next, we provide a Fully Polynomial Time Approximation Scheme (FPTAS) for this objective. 

In Algorithm~\ref{alg:recurrence-exact-fairness-objective}, we provide a dynamic program that constructs the Pareto frontier for groups' social welfare. 
In contrast to Algorithm~\ref{alg:recurrence-dp-one-group} where the algorithm only needs to store an optimal solution for each subproblem, here for each subproblem the algorithm stores a set containing \emph{all} $g$-tuples of groups' improvements $(I_1, I_2,\cdots, I_g)$ that are simultaneously achievable for groups $\{G_1,\cdots, G_g\}$. Similar to the recurrence in Algorithm~\ref{alg:recurrence-dp-one-group}, we consider the potential left-most targets $\tau'$ and subproblems for agents on or to the right of $\tau'$ with one less available target level; i.e., $T(\tau', \kappa-1)$. Particularly, in item $3$ of Algorithm~\ref{alg:recurrence-exact-fairness-objective}, we consider all combinations of potential left-most targets $\tau'$ and their corresponding subproblems. To evaluate the performance, for any potential leftmost target $\tau' \in \T_p$ and $\tau' > \tau$, we compute the improvement of all agents 
reaching to $\tau'$ from each group separately, i.e., $i \in G_\ell$ such that $\tau \leq p_i < \tau'$ and $\tau'-p_i\leq \Delta_i$, and measure their improvement to reach $\tau'$, i.e., $\tau' - p_i$. Then, we add this tuple to any tuples $(I_{\ell})_{\ell=1}^{g}\in T(\tau',\kappa-1)$, and store all the dominating resulted tuples (the Pareto frontier) in $T(\tau,\kappa)$. 

\paragraph{}
\makebox[\textwidth][c]{%
\begin{minipage}{1.06\textwidth} 
\setlength{\algoheightrule}{0pt}
\setlength{\algotitleheightrule}{0pt}
\begin{algorithm}[H]
  \caption[Construct the Pareto frontier for groups' social welfare]
  {Run dynamic program based on function $T$, defined below, that takes $\forall \ell \; \cup_{i \in G_\ell} \{p_i\}$ and $k$ as input and outputs $T(\tau_{\min},k)$, as the Pareto-frontier improvement tuples, and $S(\tau_{\min},k)$, as the Pareto-frontier sets of targets; where $\tau_{\min}= \min\{\tau \in \T_p\}$ and $\tau_{\max}= \max\{\tau \in \T_p\}$. $T(\tau,\kappa)$ constructs the Pareto frontier for groups' social welfare for agents on or to the right of $\tau\in \T_p$ when at most $\kappa$ target levels can be selected. Function $T$ is defined as follows.}
  \SetAlgoLined
    \label{alg:recurrence-exact-fairness-objective}

    \paragraph{}
    \begin{enumerate}[leftmargin=1em, rightmargin=2cm]
    \justifying
    \item[1)]  For any $\tau\in \T_p$, $ \; T(\tau,0)=\vec{0}_g$.
    
    \item[2)]  For any $1\leq \kappa\leq k$, $ \; T(\tau_{\max},\kappa)=\vec{0}_g$.
    
    \item[3)]  For any $\tau\in \T_p, \tau< \tau_{\max}$ and $1\leq \kappa\leq k$:
    {\footnotesize
    \[T(\tau,\kappa) = \Bigg\{\Bigg(I_{\ell}+\Big(\sum_{\substack{\tau\leq p_i <\tau'\\\text{ s.t. } \\ \tau'-p_i \leq\Delta_i}} \!\!\!\!\! \mathbbm{1}\Big\{i\in G_{\ell}\Big\}(\tau'-p_i)\Big)\Bigg)_{\ell=1}^{g}  \Bigg| \ (I_{\ell})_{\ell=1}^{g}\in T(\tau',\kappa-1), \tau'\in \mathcal{T}_p, \tau'> \tau \Bigg\}\]
    }

    \noindent\item[]   $S(\tau,\kappa)$ stores the sets of targets corresponding to the improvement tuples in $T(\tau,\kappa)$. After the above computations, the algorithm removes all the dominated solutions. 
    \end{enumerate}
\end{algorithm}
\setlength{\algoheightrule}{1pt} 
\setlength{\algotitleheightrule}{1pt} 
\end{minipage}
}
\paragraph{}

When all $p_i, \Delta_i$ values are integral, the running time of~Algorithm~\ref{alg:recurrence-exact-fairness-objective} gets bounded as follows.

\begin{theorem}
\label{prop:running-time-fairness-exact}
Algorithm~\ref{alg:recurrence-exact-fairness-objective} constructs the Pareto frontier for groups' social welfare using at most $k$ targets given $n$ agents in $g$ groups, and has a running time of $\mathcal{O}(n^{g+2}kg\Delta_{\max}^g)$, where $\Delta_{\max}$ is the maximum improvement capacity.
\end{theorem}
\begin{proof}
See Appendix~\ref{app:setfair_missing_max_min}.
\end{proof}

\begin{corollary}
\label{cor:exact-fairness}
There is an efficient algorithm that finds a set of at most $k$ targets that maximizes minimum improvement across all groups, i.e., maximizing $\min_{1 \leq \ell \leq g} \sw_\ell$.
\end{corollary}
\begin{proof}
See Appendix~\ref{app:setfair_missing_max_min}.
\end{proof}

\paragraph{A Fully Polynomial Time Approximation Scheme for the Max-Min Objective.}

The algorithm mentioned in Colloray~\ref{cor:exact-fairness} is pseudo-polytime since its time-complexity depends on the numeric value of $\Delta_{\max}$. We present a Fully Polynomial Time Approximation Scheme (FPTAS) to maximize the minimum improvement across all groups for the setting where each group $G_{\ell}$ has its own improvement capacity $\Delta_{\ell}$. The algorithm finds a set of at most $k$ targets that approximates the max-min objective within a factor of $1-\eps$ for any arbitrary value of $\eps>0$. Here, we relax the assumption that $p_i, \Delta_i$ values need to be integral, and suppose all $p_i, \Delta_i$ values are real numbers.
Similar to the dynamic program based on Algorithm~\ref{alg:recurrence-exact-fairness-objective}, for each subproblem, a set containing all $g$-tuples of improvements $(I_1, I_2,\cdots, I_g)$ that are simultaneously achievable for all groups is stored. However, computing all such tuples takes exponential time since $\sum_{i=1}^k \binom{2n}{i}$ possible cases of targets' placements need to be considered. Therefore, we discretize the set of all possible improvements for this problem by rounding all the improvement tuples, and develop an FPTAS algorithm. The recurrence for the dynamic program is given in Appendix~\ref{sec:setfair_appendix-FPTAS}. The algorithm runs efficiently when the number of groups is a constant. We defer the technical details to Appendix~\ref{sec:setfair_appendix-FPTAS}.

\section{Simultaneous Approximate Optimality}
\label{sec:setfair_approx_optimality}

In this section, we establish a structural result about the Pareto optimal solutions, and show there exists a simultaneously approximately optimal solution on the Pareto frontier, where the approximation factor depends on the number of groups. More specifically, given $g$ groups, and limit $k \geq g$ on the number of target levels, we provide Algorithm~\ref{alg:approx} whose improvement per group is simultaneously an $\Omega(1/g^3)$ approximation of the optimal $k$-target solution for each group; implying a constant approximation when the number of groups is constant. This result is of significance because natural outcomes such as the max-min fair solution and the union of group-optimal targets may lead to arbitrarily poor performance in terms of simultaneous approximate optimality --- See Examples~\ref{ex:max_min_suboptimal} and \ref{ex:interference}. This result only holds for the common improvement capacity model, and in Example~\ref{ex:approx_two_groups}, we show such a solution does not exist for the individualized improvement capacity model.

\begin{theorem} \label{thm:approx}
Algorithm~\ref{alg:approx}, given limit $k \geq g$ on the number of target levels, outputs a solution that is simultaneously $\Omega(1/g^3)$-approximately optimal for each group. More specifically, it provides a solution such that for all $1\leq \ell \leq g$, $\sw_\ell \geq 1/(16g^3) \opt_{\ell}^{k}$, where $\opt_{\ell}^{k}$ is the optimal social welfare of group $\ell$ using at most $k$ target levels.
\end{theorem} 

\begin{corollary} \label{cor:approx}
There is an efficient algorithm to find a simultaneously $\alpha^{\star}$-approximately optimal solution for each group, where $\alpha^{\star}$, defined as the best approximation factor possible, is $\Omega(1/g^3)$. 
\end{corollary}

We are not aware if $\Omega(1/g^3)$ is the best possible ratio, however, the following example shows there are no simultaneously approximately optimal solutions with approximation factor $>1/g$. 

\begin{example}\label{ex:lowerbound}
Let $\Delta = 1$. Suppose group $\ell \in \{1, 2, \ldots, g\}$ has a single agent at position $(\ell-1)/g$; i.e., the agents are at $0, 1/g, \ldots, (g-1)/g$. For each group, the optimal total improvement is $1$ in isolation (independent of the limit on the number of targets). However, using any number of targets in total there are no solutions with $> 1/g$ improvement for all groups. \end{example}

The following example shows that the max-min fair solution does not satisfy a simultaneous constant approximation per group even when there are only two groups.

\begin{example}\label{ex:max_min_suboptimal}
Let $\Delta = 1$. Group $A$ has $n$ agents; one agent at each position $1, 2, \ldots, n$. Group $B$ has $n$ agents in $k$ bundles of size $n/k$. The bundles of agents are at positions $n+1-k^2/n, \ldots, n+k-k^2/n$. The unique max-min solution has targets at $n-k+1, n-k+2, \ldots, n+1$, and leads to $k$ total improvement for each group which is $k/n$ of the optimal total improvement for group $B$.
\end{example}

The following example shows solving the optimization problem separately per group and outputting the union of the targets can lead to arbitrarily low group improvement compared to the optimum.

\begin{example}\label{ex:interference}
Suppose there are two groups $A$ and $B$ and no limit on the number of targets. Group $A$ has $n$ agents at positions $1, 3, 5, \ldots, 2n-1$.  Group $B$ has $n$ agents at positions $2-\eps, 4-\eps, \ldots, 2n-\eps$. {First, consider the common capacity model, where $\Delta = 1$.} In this case, the optimal solution for group $A$ in isolation consists of targets at positions $\{2, 4, \ldots, 2n\}$ and the optimal solution for group $B$ is isolation is $\{3-\eps, 5-\eps, \ldots, 2n+1-\eps\}$. Now, consider a solution that is the union of the targets in the two separate solution. Since each agent in group $B$ is in $\eps$ proximity of a target from group $A$, the total improvement in group $B$ is $n \eps$. Therefore, the total improvement in group $B$ can be arbitrarily close to $0$. Next, consider the individualized capacity model, where agents in group $A$ have $\Delta_A = 1$, and agents in group $B$ have $\Delta_B = 1+2\eps$. The optimal set of targets in isolation for group $A$ is $\{2, 4, \ldots, 2n\}$, and for group $B$ is $\{3+\eps, 5+\eps, \ldots, 2n+1+\eps\}$. The union of these solutions result in $1+(n-1)\eps$ for group $A$, and $n\eps$ for group $B$ which are arbitrarily low compared to the optimum, which is simultaneously $\geq n(1-\epsilon)$ for group $A$ and $\geq n$ for group $B$.
\end{example}

The following example shows that if agents can improve by different amounts (the individualized improvement capacity model), then no approximation factor only as a function of $g$ of optimal improvement per group is  possible. 

\begin{example}\label{ex:approx_two_groups}
Suppose groups $A$ and $B$ each have a single agent at position $0$. The agent in group $A$ has improvement capacity $\Delta_A = \eps < 1$ and the agent in group $B$ has improvement capacity $\Delta_B = 1$. The optimal total improvement in isolation for group $A$ is $\epsilon$, and for group $B$ is $1$. However, when considering both groups, no placement of targets with positive improvement for group $A$ leads to $> \eps$ improvement for group $B$.
\end{example}

First, we describe a high-level overview of Algorithm~\ref{alg:approx}. The algorithm proceeds in the following four main steps. 

\begin{enumerate}[label=\arabic*)]
    \item \textbf{Optimal targets in isolation.} Run Algorithm~\ref{alg:recurrence-dp-one-group} separately for each group to find an optimal allocation of at most $\lceil k/g \rceil$ targets\footnote{Although the total number of targets used in this step can be more than $k$, after the algorithm ends at most $k$ targets are being used in total.}. Let $\T_\ell$ be the output for group $\ell$. 
    \item \textbf{Distant targets in isolation.} Delete $3/4$ fraction of each set of target levels, $\T_\ell$, such that (1) the distance between every two consecutive targets in each set is at least $2\Delta$ and (2) the new $\T_\ell$  (after deletion) guarantees an $\Omega(1)$ approximation of the previous step when the targets for each group are considered in isolation. {Section~\ref{sec:setfair_step2} below shows this is possible.}
    \item \textbf{Locally optimized distant targets in isolation.} For each $\ell$ and $\tau \in \T_\ell$, consider the agents in group $\ell$ that afford to reach $\tau$ (agents in $G_{\ell}$ $\cap[\tau-\Delta, \tau)$). Optimize $\tau$ to maximize the total improvement for this set of agents.
    \item \textbf{Resolve interference of targets.} Consider sets of interfering targets. Relocate these targets locally to guarantee $\Omega(1/g^2)$ approximation per group compared to the previous step where each group was considered in isolation. {Section~\ref{sec:setfair_step4} below shows this is possible.}
\end{enumerate}

\begin{algorithm}[!ht]
\caption{Simultaneous approximate optimality per group.}
\label{alg:approx}
    \SetNoFillComment
    \SetAlgoLined
    \DontPrintSemicolon
    \For{ $\ell= 1 \text{ to } g$}{
        \tcc{Step 1}
        Let $\T_\ell: \tau_1 <\tau_2< \ldots$  be the output of Algorithm~\ref{alg:recurrence-dp-one-group} for agents in $G_{\ell}$ and limit $\lceil k/g \rceil$ on the number of targets.\;

        \tcc{Step 2}
        Partition $\T_\ell$ to $4$ parts $P_1, P_2, P_3, P_4$, where $P_i:= \tau_i, \tau_{4+i}, \tau_{8+i}, \ldots$.\; 
        Update $\T_\ell$ by keeping the part with the highest improvement and deleting the rest.\;
        
        \tcc{Step 3}
        Delete agents in $G_\ell$ that do not improve given $\T_\ell$.\;
        For all $\tau \in \T_\ell$, replace $\tau$ with the output of Algorithm~\ref{alg:recurrence-dp-one-group} for agents in $[\tau-\Delta, \tau) \cap G_{\ell}$ and limit $1$ on the number of targets.\; 
    }
    \tcc{Step 4}
    $\T: \tau_1 < \tau_2 < \ldots = \cup_\ell \T_\ell$\\
    $S, \T^{\star} = \emptyset$\\
    \For{$\tau_j \in \T$}{
        $s_j = \tau_j - \Delta$\\
        $S = S \cup \{s_j\}$\\
    }
    {Partition $S : s_1 < s_2 < \ldots$ into the least number of parts of consecutive points: $S_1, S_2, \ldots$, such that in each part, $S_i$, each two consecutive points are at distance less than $\Delta/g$.}\;
    \For{all $S_i : s_u < s_{u+1} < \ldots < s_v$}{
        $\tau^{\star}_i = \min\{\tau_u, s_{v+1}\}$.\\
        $\T^{\star} = \T^{\star} \cup \tau^{\star}_i$.\\
    }
    \Return $\T^{\star}$\\
\end{algorithm}

Now, we describe and analyze these steps in more detail. 

\subsection{Step \texorpdfstring{$1$}{1}: Optimal Targets in Isolation}

At the end of step $1$, $\T_\ell$ is the optimal set of targets for $G_\ell$ in isolation. The following observation shows that without loss of optimality, we may assume the distance between every other target level is at least $\Delta$\footnote{Example~\ref{ex:consecutive_less_than_delta}, however, shows the distance between two \emph{consecutive} targets may be arbitrarily smaller than $\Delta$.}.

\begin{observation}\label{obs:delta_apart}
Consider a set of target levels $\T: \tau_1 < \tau_2 < \ldots$.  Suppose $\tau_{j+2} < \tau_j + \Delta$. By removing $\tau_{j+1}$, any agent with $\tau_j \leq p_i < \tau_{j+1}$ improves strictly more, and other agents improve the same amount. This weakly increases social welfare. 
\end{observation}

\subsection{Step \texorpdfstring{$2$}{2}: Distant Targets in Isolation}
\label{sec:setfair_step2}

Step $2$ of the algorithm runs the following procedure for $\T_\ell$.

\begin{definition}[Distant targets procedure] 
\label{def:distant_procedure}
Consider solution $\T: \tau_1< \tau_2< \ldots$, where for all $j$,  $\tau_{j+2}-\tau_j \geq \Delta$ as input to the following procedure.
\begin{itemize}
    \item Partition $\T$ into $4$ parts, $P_1, P_2, P_4, P_4$, where $P_i=: \tau_i, \tau_{4+i}, \tau_{8+i}, \ldots$. Consider the part $P_i$ that introduces the highest improvement. Update $\T$ to $P_i$ (and delete the rest).
\end{itemize}
\end{definition}

The following lemma shows that at the end of this step, target levels in $\T_\ell$ are $2\Delta$ apart, this step provides a $4$-approximation compared to the previous step, and the number of targets designated to each group is at most $\lfloor k/g \rfloor$.

\begin{lemma}\label{lm:make_distant}
Consider solution $\T: \tau_1< \tau_2< \ldots$ with total improvement $I$ such that for all $j$,  $\tau_{j+2}-\tau_j \geq \Delta$. Consider the procedure in Definition~\ref{def:distant_procedure}. This procedure results in a solution $\T': \tau'_1< \tau'_2< \ldots$ where $\forall j \; \tau'_{j+1}-\tau'_j \geq 2\Delta$, has total improvement at least $ I/4$, and $|\T'| \leq \lceil |\T|/4 \rceil$. Particularly, for $|\T|\leq \lceil k/g \rceil$ where $k \geq g$, the number of final targets, $|\T'|$, is at most $\lfloor k/g \rfloor$.
\end{lemma}
\begin{proof} See Appendix~\ref{app:setfair_approximation}.
\end{proof}

\subsection{Step \texorpdfstring{$3$}{3}: Locally Optimized Distant Targets in Isolation}
At the end of step $2$, every two targets in $\T_\ell$, the set of targets for group $\ell$, are at distance at least $2\Delta$. Consider only the targets and agents in group $\ell$. For each $\tau \in \T_\ell$, agents in $[\tau-\Delta, \tau)$ improve to $\tau$ and the remaining agents do not improve. 
To continue with the algorithm, we first delete the agents that do not improve. Then, we optimize $\T_\ell$ for the set of agents that do improve. This modification is necessary for the next step. To do the optimization, we use Algorithm~\ref{alg:recurrence-dp-one-group} for agents in $[\tau-\Delta, \tau)$ for any $\tau \in \T_\ell$ and limit $1$ on the number of targets, and replace $\tau$ with the output of the algorithm.

\begin{lemma}\label{lm:step_three}
At the end of step $3$ in Algorithm~\ref{alg:approx}, (i) the distance between every two targets in $\T_\ell$ is at least $\Delta$; (ii) each target $\tau \in \T_\ell$ is optimal, i.e., maximizes total improvement for agents in $G_\ell \cap [\tau-\Delta, \tau)$; and (iii) the total amount of improvement of $G_\ell$ using solution $\T_\ell$ does not decrease compared to the previous step.
\end{lemma}
\begin{proof} See Appendix~\ref{app:setfair_approximation}.
\end{proof}

Now, we extract properties about optimal solutions. Since at the end of step $3$, $\T_\ell$ is optimal for $G_\ell$ we take advantage of these properties in the remaining steps of the algorithm. 

The following lemma shows that if $\tau$ is optimal for agents in $[\tau-\Delta, \tau)$, a considerable fraction of these agents reside in the left-most part of the interval.

\begin{lemma}\label{lm:p_x}
Consider optimal target $\tau$ for the set of agents $A$ in $[\tau-\Delta, \tau)$ in absence of other targets. For each $0\leq x \leq 1$, at least $x$ fraction of $A$ belong to $[\tau-\Delta,\tau-\Delta + x\Delta)$. In particular, at least $1/(2g)$ fraction of the agents are in $[\tau-\Delta,\tau-(2g-1)/(2g) \Delta)$. 
\end{lemma}
\begin{proof}
Let $p_x$ be the fraction of agents in $A$ in $[\tau-\Delta, \tau-\Delta+x\Delta)$. Each of these agents is improving by at least $(1-x) \Delta$. Therefore, the contribution of these agents to total improvement of $A$ is at least $p_x|A| (1-x)\Delta$. Since $\tau$ is the optimal target, it introduces at least as much improvement as any other target, and in particular a target at $\tau' = \tau+x$. Consider the total improvement introduced by $\tau'$ compared to $\tau$ (in absence of target $\tau$). The contribution of the agents in $[\tau-\Delta, \tau-\Delta+x\Delta)$ to total improvement reduces to $0$, but the contribution of the agents in $[\tau-\Delta+x\Delta, \tau)$ increases by $(1-p_x)|A|x\Delta$. Since $\tau$ is the optimal target, the loss of substituting it with $\tau'$ is at least as much as the gain. Therefore, $p_x (1-x)\Delta \geq (1-p_x)x\Delta$; which implies $p_x \geq x$. 
\end{proof}

The following lemma shows that if $\tau$ is optimal for agents in $[\tau-\Delta, \tau)$, substituting $\tau$ with another target in this interval, far enough from the left endpoint, $\tau-\Delta$, guarantees a considerable fraction of the optimal improvement.

\begin{lemma}\label{lm:shift}
Consider optimal target $\tau$ for agents $A$ in $[\tau-\Delta, \tau)$ in absence of other targets. By relocating $\tau$ to any point in $[\tau-\Delta+x\Delta, \tau]$, for $0 \leq x \leq 1$, the total improvement of $A$ is at least $x^2/4$ of the optimum. In particular, by relocating $\tau$ to any point in $[\tau-\Delta+\Delta/g, \tau]$, the total improvement is at least $1/(4g^2)$ of the optimum.
\end{lemma}
\begin{proof}
Similar to the previous lemma, let $p_{x/2}$ be the fraction of agents in $[\tau-\Delta,\tau-\Delta+(x/2)\Delta)$. After the relocation, each such agent improves by at least $(x/2)\Delta$; therefore, the contribution of these agents to total improvement is at least $p_{x/2}|A|(x/2)\Delta$. The optimal total improvement is bounded by $|A|\Delta$. Therefore, using $p_{x/2} \geq x/2$, by Lemma~\ref{lm:p_x}, the total improvement after relocation is at least $x^2/4$ of the optimum.
\end{proof}

\subsection{Step \texorpdfstring{$4$}{4}: Resolve Interference of Targets}\label{sec:setfair_step4}

In this step, we consider the solutions for all groups together and resolve the interference of targets designed for different groups. As illustrated in Example~\ref{ex:interference}, this interference can lead to arbitrarily low social welfare. To resolve this issue, we take advantage of sparsity of the targets designed for the same group (step $2$) and optimality of $\T_\ell$ for $G_\ell$ (step $3$). 

The main purpose of this step is to recover an approximation guarantee of the total improvement of \emph{each target in isolation} at the end of step $3$ by removing the interference among the targets. Particularly, for each target $\tau \in \T_\ell$ in isolation, we consider agents in $G_\ell$ reaching to that, i.e., agents in interval $[\tau-\Delta, \tau)$. By Lemma~\ref{lm:p_x}, a considerable fraction of these agents are on the left-most side of the interval. And as shown in Lemma~\ref{lm:shift}, as long as there exists a target far enough from the left endpoint we are in good shape. More precisely, if for all $\tau$ at the beginning of this step, there is a target in the final solution in $[\tau-\Delta+\Delta/g, \tau]$ (property $1$), and no targets in $(\tau-\Delta, \tau-\Delta+\Delta/g)$ (property $2$), a $1/(4g^2)$ fraction is achievable. The set of targets at the end of step $3$ may fail to satisfy these properties, because there may be targets $\tau' < \tau$ such that $\tau'$ is not far enough from the left endpoint of the interval corresponding to $\tau$; i.e., for $s=\tau-\Delta, \; s < \tau' < s + \Delta/g$. 

To resolve the interference among the targets, in step $4$, we work as follows. First, we consider the left endpoints of improvement intervals corresponding to the targets; i.e., $\forall \tau_j$, at the end of step $3$, consider $s_j=\tau_j-\Delta$. Then, we partition these left endpoints into maximal parts $S_1, S_2, \ldots$, such that in each part, the distance between every two consecutive points is small, particularly, less than $\Delta/g$. Using the sparsity of the targets (step $2$) the number of points in each part is bounded. Finally, we design a new target $\tau^{\star}_i$ (defined formally below) corresponding to part $S_i$, such that $\tau^{\star}_i$ is to the left of any $S_j$ with $j > i$, and at distance between $\Delta/g$ and $\Delta$ to the right of the points in $S_i$ (satisfying properties $1$ and $2$). Using optimality of $\T_\ell$ for $G_\ell$ (step $3$) this results in the desired approximation factor.

More formally, this step proceeds as follows.

\begin{enumerate}[label=\arabic*)]
    \item Let $\T: \tau_1 < \tau_2 < \ldots$ be the union of the set of targets found at the end of step $3$.
    \item Construct $S: s_1 < s_2 < \ldots$ from $\T$, such that $\forall \tau_j \in \T$, include $s_j = \tau_j - \Delta$ in $S$.
    \item Partition $S$ into the least number of parts of consecutive points: $S_1, S_2, \ldots$, such that in each part $S_i : s_u < s_{u+1} < \ldots < s_v$, each two consecutive points are at distance less than $\Delta/g$; i.e., $\forall s_r,s_{r+1} \in S_i, s_{r+1}-s_r < \Delta/g $. By construction of the first three steps (and as shown in the proof of Lemma~\ref{lm:step_four}), the number of points in each part is at most $g$.
    \item For each $S_i : s_u < s_{u+1} < \ldots < s_v$, consider new target $\tau^{\star}_i = \min\{\tau_u, s_{v+1}\}$. 
    \item Output the set of new targets.
\end{enumerate}

\begin{lemma}\label{lm:step_four}
Consider $\T$ as the union of all solutions at the end of step $3$. For all $\tau \in \T$, consider the interval $[\tau-\Delta, \tau)$ which consists of agents that improve to target $\tau$ if it were the only target available. At the end of step $4$, (i) there will be a target in $[\tau-\Delta+\Delta/g, \tau]$, and (ii) there will be no targets in $(\tau-\Delta, \tau-\Delta+\Delta/g)$.
\end{lemma}

\begin{proof} 
    See Appendix~\ref{app:setfair_approximation}.
\end{proof}

\subsection{Putting Everything Together}

\begin{theorem} \label{thm:modified}
Algorithm~\ref{alg:approx}, given $k \geq g$, provides a solution with at most $k$ number of targets, such that for all $1\leq \ell \leq g$, $\sw_\ell \geq 1/(16g^2) \opt_{\ell}^{\lceil k/g \rceil}$, where $\opt_{\ell}^{k}$ is the optimal social welfare of group $\ell$ using at most $k$ target levels.
\end{theorem}

\begin{proof}
By Observation~\ref{obs:delta_apart} and Lemma~\ref{lm:make_distant}, when the targets designed for each group are considered separately and in isolation, at the end of step $2$, there are at most $\lfloor k/g \rfloor$ targets designed for group $\ell$ and the total improvement in this group is $1/4$-approximation of $\opt_{\ell}^{\lceil k/g \rceil}$. By Lemma~\ref{lm:step_three}, Lemma~\ref{lm:shift}, and Lemma~\ref{lm:step_four}, we lose another $4g^2$ factor compared to step $2$. In total, Algorithm~\ref{alg:approx} results in $\sw_\ell \geq 1/(16g^2) \opt_{\ell}^{\lceil k/g \rceil}$, for all groups $1\leq \ell \leq g$. Also, when $k \geq g$, the total number of targets is at most $g \lfloor k/g \rfloor \leq k$.
\end{proof}

\begin{proof}[Proof of Theorem~\ref{thm:approx}]
Given Theorem~\ref{thm:modified}, it suffices to argue $\opt_{\ell}^{\lceil k/g \rceil} \geq  \opt_{\ell}^{k}/g$; i.e., when the number of targets increases by a factor, here $g$, the optimal total improvement increases by at most that factor. This statement is straightforward using subadditivity of total improvement as a function of the set of targets. Specifically, consider the optimal $k$-target solution and an arbitrary partition with $g$ parts of size $\lceil k/g \rceil$ or $\lfloor k/g \rfloor$; by subadditivity, one of the parts provides at least $1/g$ of the total improvement. 
\end{proof}

\begin{proof}[Proof of Corollary~\ref{cor:approx}] Algorithm~\ref{alg:recurrence-exact-fairness-objective} in Section~\ref{sec:setfair_max_min} outputs the Pareto frontier for groups' social welfare. By definition, the solution provided in Algorithm~\ref{alg:approx} is dominated by a solution on the Pareto frontier. By computing the factor of simultaneous approximate optimality of each solution on the Pareto frontier, we find the solution that achieves the best simultaneous approximation factor $\alpha^3$, and by Theorem~\ref{thm:approx}, this solution is simultaneously $\Omega(1/g^3)$-approximately optimal.
\end{proof}

\begin{remark}[a weaker benchmark and a tighter gap]
In contrast with Theorem~\ref{thm:approx} that measures the performance of Algorithm~\ref{alg:approx} with respect to the optimal $k$-target solution for each group (the notion of simultaneous approximate optimality), Theorem~\ref{thm:modified} measures the performance with respect to the optimal $\lceil k/g \rceil$-target solution for each group. Since the lower bound provided in Example~\ref{ex:lowerbound} shows achieving better than $1/g$ of either of these benchmarks is not possible, there is only a factor $g$ gap in the performance of the algorithm and the lower bound with respect to the optimal $\lceil k/g \rceil$-target solution. 
\end{remark}

\section{Generalization Guarantees}
\label{sec:setfair_gen_guarantees}

In this section, we generalize our results to a setting where we only have sample access to agents and provide sample complexity results. Section~\ref{sec:setfair_gen_single} provides a guarantee for the maximization objective in absence of fairness, and Section~\ref{sec:setfair_gen_multiple} provides a guarantee for the fairness objectives.

\subsection{Generalization Guarantees For the Maximization Objective}
\label{sec:setfair_gen_single}
Suppose there is a distribution $\mathcal{D}$ over agents' positions. Our goal is to find a set of $k$ targets $\T$ that maximizes expected improvement of an agent when we only have access to $n$ agents sampled from $\mathcal{D}$. For any distribution $\mathcal{D}$ over agents' positions, we define $I_{\mathcal{D}}(\mathcal{T}) = \E_{p\sim \mathcal{D}}[I_p(\mathcal{T})]$, where $I_p(\mathcal{T})$ captures the improvement of agent $p$ given the targets in $\mathcal{T}$. 
In Theorem~\ref{thm:generalization-absent-fairness}, we provide a generalization guarantee that shows if we sample a set $S$ of size $n\geq \eps^{-2}\big(\Delta_{\max}^2(k\ln(k)+\ln(1/\delta))\big)$ drawn \emph{i.i.d} from $\mathcal{D}$, then with probability at least $1-\delta$, for all sets $\T$ of $k$ targets, we can bound the difference between average performance over $S$ and actual expected performance, such that $\big|I_S(\mathcal{T})-I_{\mathcal{D}}(\mathcal{T})\big|\leq \mathcal{O}(\eps)$. Formally, we show the following theorem holds:
\begin{theorem}
(Generalization of the maximization objective) Let $\mathcal{D}$ be a distribution over agents' positions. For any $\eps>0$, $\delta>0$, and number of targets $k$, if $S=\{p_i\}_{i=1}^n$ is drawn \emph{i.i.d}.\ from $\mathcal{D}$ where 
$n\geq \eps^{-2}\Delta_{\max}^2\big(k\ln(k)+\ln(1/\delta)\big)$,
then with probability at least $1-\delta$, for all sets $\T$ of $k$ targets, $\big|I_S(\mathcal{T})-I_{\mathcal{D}}(\mathcal{T})\big|\leq \mathcal{O}(\eps)$.
\label{thm:generalization-absent-fairness}
\end{theorem}
In particular, the solution $\T^{\star}$ that maximizes improvement on $\mathcal{S}$, also maximizes improvement on $\mathcal{D}$ within an additive factor of $\mathcal{O}(\eps)$.

In order to prove  Theorem~\ref{thm:generalization-absent-fairness}, we use two main ideas. First, using a framework developed by \citet{Balcan-STOC21}, we bound the \emph{pseudo-dimension} complexity of our improvement function. Then, using classic results from learning theory~\citep{pollard1984convergence}, we show how to
translate \emph{pseudo-dimension} bounds into generalization guarantees. The framework proposed by \citet{Balcan-STOC21} depends on the relationship between primal and dual functions. When the dual function is piece-wise constant, piece-wise linear or generally piece-wise structured, they show a general theorem that bounds the \emph{pseudo-dimension} of the primal function.
Formally \emph{pseudo-dimension} is defined as following:

\begin{definition}[Pollard's Pseudo-Dimension] 
A class $\mathcal{F}$ of real-valued functions $P$-shatters a set of points $\mathcal{X} = \{x_1, x_2,\cdots, x_n\}$ if there exists a set of thresholds $\gamma_1, \gamma_2,\cdots, \gamma_n$ such that for every subset $T\subseteq \mathcal{X}$, there exists a function $f_T\in \mathcal{F}$ such that $f_T(x_i)\geq \gamma_i$ if and only if $x_i\in T$. In other words, all $2^n$ possible above/below patterns are achievable for targets $\gamma_1,\cdots, \gamma_n$. The pseudo-dimension of $\mathcal{F}$, denoted by $\PDim(\mathcal{F})$, is the size of the largest set of points that it $P$-shatters.
\end{definition}

\citet{Balcan-STOC21} show when the dual function is piece-wise structured, the \emph{pseudo-dimension} of the primal function gets bounded as following:
\begin{theorem}
(Bounding Pseudo-Dimension~\citep{Balcan-STOC21})
Let $\mathcal{U}=\{u_{\vec{\rho}}\mid \vec{\rho}\in \mathcal{P}\subseteq \mathbb{R}^d\}$ be a class of utility functions defined over a $d$-dimensional parameter space. Suppose the dual class $\mathcal{U}^{\star}$ is $(\mathcal{F},\mathcal{G},m)$-piecewise decomposable, where the boundary functions $\mathcal{G}=\{f_{\vec{a}, \theta}: \mathcal{U}\rightarrow \{0,1\}\mid \vec{a}\in \mathbb{R}^d, \theta\in \mathbb{R}\}$ are halfspace indicator functions $g_{\vec{a},\theta}:u_{\rho}\rightarrow \mathbb{I}_{\vec{a}\cdot\vec{\rho}\leq \theta}$ and the piece functions $\mathcal{F}=\{f_{\vec{a},\theta}:\mathcal{U}\rightarrow \mathbb{R}\mid \vec{a}\in \mathbb{R}^d, \theta\in \mathbb{R}\}$ are linear functions $f_{\vec{a},\theta}:u_{\rho}\rightarrow \vec{a}\cdot \vec{\rho}+\theta$, and $m$ shows the number of boundary functions. Then, $\PDim(\mathcal{U}) = \mathcal{O}(d\ln(dm))$.
\label{thm-Balcan-Pdim}
\end{theorem}

We use Theorem~\ref{thm-Balcan-Pdim} to bound the \emph{pseudo-dimension} of the improvement function.

\begin{lemma}
\label{lem:bounding-pdim}
Let $\mathcal{U}=\{u_{\mathcal{T}}:p\rightarrow u_{\mathcal{T}}(p) \mid \mathcal{T}\in \mathbb{R}^k, p\in \mathbb{R}\}$ be a set of functions, where each function defined by a set of $k$ targets, takes as input a point $p\in \mathbb{R}$ that captures an agent's position, and outputs a number showing the improvement that the agent can make. Then, $\PDim(\mathcal{U})=\mathcal{O}(k\ln(k))$.
\end{lemma}

\begin{proof}
We use Theorem~\ref{thm-Balcan-Pdim} to bound $\PDim(\mathcal{U})$. First, we define the dual class of $\mathcal{U}$ denoted by $\mathcal{U}^{\star}$. The function class $\mathcal{U}^{\star}=\{u^{\star}_p:\mathcal{T}\rightarrow u_{p}(\mathcal{T}) \mid \mathcal{T}\in \mathbb{R}^k, p\in \mathbb{R}\}$ is a set of 
functions, where each function defined by an agent $p$, takes as input a set $\mathcal{T}\in \mathbb{R}^k$ of $k$ targets\footnote{If the input consists of $k'$ targets where $k'<k$, it resembles the case where $k$ targets are used and $k-k'$ of them are ineffective, i.e., are put at position $\tau_{\min}$.}, and outputs the improvement that $p$ can make given $\mathcal{T}$. 
Geometrically, in the dual space, there are $k$ dimensions $\tau_1,\cdots,\tau_k$, and each dimension is corresponding to one target. In order to use Theorem~\ref{thm-Balcan-Pdim}, we show that $\mathcal{U}^{\star} = (\mathcal{F},\mathcal{G},k)$ is piecewise-structured. The boundary functions in $\mathcal{G}$ are defined as follows. 
If agent $p$ improves to a target $\tau_i$, then $0< \tau_i - p \leq \Delta$, where $\Delta$ is the improvement capacity of $p$. Additionally, between all the targets within a distance of at most $\Delta$, $p$ improves to the closest one. For each pair of integers $(i,j)$, where $1\leq i,j\leq k$, we add the hyperplane $\tau_i -\tau_j =0$ to $\mathcal{G}$. Above this hyperplane is the region where $\tau_i >\tau_j$, implying that $\tau_i$ comes after $\tau_j$. Below the hyperplane is the region where the ordering is reversed. In addition, for each target $\tau_i$, we add the boundary functions $\tau_i = p$ and $\tau_i = p+\Delta$ to $\mathcal{G}$. In the region between $\tau_i = p$ and $\tau_i = p+\Delta$, $\tau_i$ is effective and the agent can improve to it.
Now, the dual space is partitioned into a set of regions. In each region, either there exists a unique closest effective target $(\tau_r)$, or all the targets are ineffective. In the former case, the improvement that the agent makes is a linear function of its distance from the closest effective target $(f = \tau_r-p)$. In the later case, the agent makes no improvement ($f = 0$). Therefore, the piece functions in $\mathcal{F}$ are either constant or linear. Now, since the total number of boundary functions is $m=\mathcal{O}(k^2)$ and the space is $k$-dimensional, using Theorem~\ref{thm-Balcan-Pdim},  $\PDim(\mathcal{U})$ is $\mathcal{O}(k\ln(k^3)) = \mathcal{O}(k\ln(k))$.
\end{proof}

Now, we are ready to prove Theorem~\ref{thm:generalization-absent-fairness}.

\begin{proof}[Proof of Theorem~\ref{thm:generalization-absent-fairness}]
Classic results from learning theory~\citep{pollard1984convergence} 
show the following generalization guarantees: Suppose $[0,H]$ is the range of functions in hypothesis class $\mathcal{H}$. For any $\delta\in(0,1)$, and any distribution $\mathcal{D}$ over $\mathcal{X}$, with probability $1-\delta$ over the draw of $\mathcal{S}\sim \mathcal{D}^n$, for all functions $h\in \mathcal{H}$, the difference between the average value of $h$ over $\mathcal{S}$ and its expected value gets bounded as follows:
\[\Big|\frac{1}{n}\sum_{x\in \mathcal{S}}h(x)-\E_{y\sim \mathcal{D}}[h(y)]\Big| = \mathcal{O}\Big(H\sqrt{\frac{1}{n}\Big(\PDim(\mathcal{H})+\ln(\frac{1}{\delta})\Big)}\Big)\]

In the case of maximizing improvement, $H=\Delta_{max}$ and $\PDim(\mathcal{H}) = \mathcal{O}(k\ln(k))$. By setting $n\geq \eps^{-2}\Delta_{max}^2\big(k\ln(k)+\ln(1/\delta)\big)$, with probability at least $1-\delta$, the difference between the average performance over $\mathcal{S}$ and the expected performance on $\mathcal{D}$ gets upper-bounded by $\mathcal{O}(\eps)$. 
\end{proof}

\subsection{Generalization Guarantees For Fairness Objectives}\label{sec:setfair_gen_multiple}
Suppose there is a distribution $\mathcal{D}_{\ell}$ of agents' positions for each group $\ell$. Let $\mathcal{D}=\sum_{\ell=1}^g \alpha_{\ell}\mathcal{D}_{\ell}$ be a weighted mixture of distributions $\mathcal{D}_1,\cdots,\mathcal{D}_g$. Let $\alpha_{\min} = \min_{1\leq \ell \leq g}\alpha_{\ell}$. Suppose we have sampling access to $\mathcal{D}$ and cannot directly sample from $\mathcal{D}_1,\cdots,\mathcal{D}_g$. Our goal is to derive generalization guarantees for different objective functions across multiple groups 
when we only have access to a set $S$ of $n$ agents sampled from distribution $\mathcal{D}$.
Let $I_{G_{\ell}}(\mathcal{T})$ denote the average improvement of agents in group $G_{\ell}\subseteq S$ given a set $\mathcal{T}$ of $k$ targets. Let $I_{\mathcal{D}_{\ell}}(\mathcal{T}) = \E_{p\sim \mathcal{D}_{\ell}}[I_p(\mathcal{T})]$, where $I_p(\mathcal{T})$ captures the improvement of agent $p$ given $\mathcal{T}$.
In Theorem~\ref{thm:generalization-fairness}, we show if we sample a set $S$ of $\mathcal{O}\Big(\alpha_{\min}^{-1}\Big(\eps^{-2}\Delta_{\max}^2\big(k\ln(k)+\ln(g/\delta)\big)+\ln(g/\delta)\Big)\Big)$ examples drawn \emph{i.i.d.}\ from $\mathcal{D}$, then for all sets $\mathcal{T}$ of $k$ targets and for all groups $\ell$, $\big|I_{G_{\ell}}(\mathcal{T})-I_{\mathcal{D}_{\ell}}(\mathcal{T})\big|\leq \mathcal{O}(\eps)$.

\begin{theorem}
(Generalization across multiple groups) Let $\mathcal{D}$ be a distribution over agents' positions. For any $\eps>0$, $\delta>0$, and number of targets $k$, if $S=\{p_i\}_{i=1}^n$ consisting of $g$ groups $\{G_{\ell}\}_{\ell=1}^g$ is drawn i.i.d.\ from $\mathcal{D}$,
where $n\geq (2/\alpha_{\min})\big(\eps^{-2}\Delta_{\max}^2(k\ln(k)+\ln(2g/\delta))+4\ln(2g/\delta)\big)$,
then with probability at least $1-\delta$, for all sets $\mathcal T$ of $k$ targets, for all groups $\ell$, $\big|I_{G_{\ell}}(\mathcal{T})-I_{\mathcal{D}_{\ell}}(\mathcal{T})\big|\leq \mathcal{O}(\eps)$.
\label{thm:generalization-fairness}
\end{theorem}

\begin{proof}
Let $S$ be partitioned into $g$ groups where each group $G_{\ell}$ has size $n_{\ell}$. 
First, for each group $\ell$, let $A_{\ell}$ denote the event that $n_{\ell}\geq (n\alpha_{\ell})/2$. Using Chernoff-Hoeffding bounds we have $\Pr[n_{\ell}<(n\alpha_{\ell})/2]\leq e^{(-n\alpha_{\ell})/8}\leq \delta/(2g)$. The last inequality holds since $n\geq 8\ln(2g/\delta)/\alpha_{\ell}$. Next, for each group $\ell$, let $B_{\ell}$ denote the event that  $\big|I_{G_{\ell}}(\mathcal{T})-I_{\mathcal{D}_{\ell}}(\mathcal{T})\big|\leq \mathcal{O}(\eps)$, then: 
\begin{align}
\Pr[B_{\ell}]\geq \Pr[B_{\ell}\cap A_{\ell}] = \Pr[B_{\ell}\mid A_{\ell}]\cdot \Pr[A_{\ell}]\geq (1-\delta/(2g))(1-\delta/(2g))\geq (1-\delta/g)
\label{bound-prob-B-ell}
\end{align}
In the above statement, inequality $\Pr[B_{\ell}\mid A_{\ell}]\geq (1-\delta/(2g))$ holds since given $A_{\ell}$ happens, then $n_{\ell}\geq \eps^{-2}\Delta_{max}^2(k\ln(k)+\ln(2g/\delta))$, and by Theorem~\ref{thm:generalization-absent-fairness}, event $B_{\ell}$ happens with probability at least $1-\delta/(2g)$. Now, by Equation~\ref{bound-prob-B-ell}, $\Pr[B_{\ell}]\geq 1-\delta/g$. By applying a union bound, event $B_{\ell}$ happens with probability at least $1-\delta$ for any group $\ell$. 
\end{proof}

In particular, solution $\T^{\star}$ satisfying one of the fairness notions considered in this chapter, e.g., simultaneous approximate optimality or maximizing minimum improvement across groups, on input $S$, achieves a performance guarantee within an additive factor of $\mathcal{O}(\eps)$ on inputs drawn from $\mathcal{D}$.

\section{Extensions and Open Problems}
\label{sec:setfair_extensions}

This section provides two extensions to our objective function: 1) maximizing social welfare subject to a lower bound on the number of improving agents, and 2) optimizing the number of target levels. The section concludes with our main open problem of optimizing the factor of simultaneous approximate optimality and tightening the gap between the upper and lower bounds.

\subsection{Extension 1: Lower Bound on the Number of Agents that Improve}
Consider Algorithm~\ref{alg:recurrence-dp-one-group} whose goal is to find a set of at most $k$ target levels that maximizes the total improvement for a collection of $n$ agents. It is possible that the solution of this algorithm focuses on a small fraction of the agents and does not help many agents to improve. In Algorithm~\ref{alg:recurrence-dp-extension}, we show how to modify Algorithm~\ref{alg:recurrence-dp-one-group} to ensure at least $n_{\ell b}$ agents improve. 
The main idea for the recursive step (item $4$ in Algorithm~\ref{alg:recurrence-dp-extension}) is to first consider the potential leftmost targets $\tau' > \tau$, let $x$ denote the number of agents that are within reach to $\tau'$,  and use the smaller subproblem of finding the optimal targets for agents on or to the right of $\tau'$ with one less available target level and an updated lower bound of $\eta-x$, i.e., $S(\tau', \kappa-1, \eta-x)$. We add the performance of each potential leftmost target to the optimal improvement of the remaining subproblem and pick the leftmost target that maximizes this summation.

\paragraph{}
\makebox[\textwidth][c]{%
\begin{minipage}{1.06\textwidth} 
\setlength{\algoheightrule}{0pt}
\setlength{\algotitleheightrule}{0pt}
\begin{algorithm}[H]
  \caption[Find at most $k$ target levels that maximizes the total improvement for $n$ agents]
  {Run dynamic program based on function $S$, defined below, that takes $\cup_i \{p_i\}$ and $k$ as input and outputs $S(\tau_{\min},k,n_{\ell b})$, as the optimal improvement, and $S'(\tau_{\min},k,n_{\ell b})$, as the optimal set of targets; where $\tau_{\min}= \min\{\tau \in \T_p\}$ and $\tau_{\max}= \max\{\tau \in \T_p\}$. $S(\tau,\kappa,\eta)$ captures the maximum improvement possible for agents on or to the right of $\tau\in \T_p$ when $\kappa$ target levels can be selected and at least $\eta$ agents need to improve. If $S(\tau_{\min},k,n_{\ell b})=-\infty$ then incentivizing at least $n_{lb}$ agents to improve is impossible. Function $S$ is defined as follows.}
  \SetAlgoLined
    \label{alg:recurrence-dp-extension}
    \paragraph{}

    \begin{enumerate}[leftmargin=1em, rightmargin=2cm]
    \justifying
    \item[1)] For any $\tau\in \T_p, \eta\geq 1$, we have $S(\tau,0,\eta)=-\infty$.
    \item[2)] For any $1\leq \kappa\leq k, \eta\geq 1$, $S(\tau_{max},\kappa,\eta)=-\infty$, where $\tau_{max}=\max\{\tau\in \T_p\}$. This holds since no agents can improve to $\tau_{max}$, however at least $\eta$ agents to the right of $\tau_{max}$ need to improve which is a contradiction.
    \item[3)] For any $\tau\in \T_p, 0\leq \kappa\leq k, \eta\leq 0$, $S(\tau,\kappa,\eta)=T(\tau,\kappa)$ where function $T$ is defined in Algorithm~\ref{alg:recurrence-dp-one-group}.
    \item[4)] For any $\tau\in \T_p, \tau< \tau_{max}$, $1\leq \kappa\leq k$, and $1\leq \eta\leq n$:
    {\footnotesize
        \begin{gather*} 
        S(\tau,\kappa,\eta) = 
        \max_{\substack{\tau'\in \Tau_p \\ \text{s.t} \\ \tau' > \tau}}
        \Bigg(S\Big(\tau',\kappa-1,\eta-\mathbbm{1}\big[i\mid \tau\leq p_i <\tau'  \ \text{s.t} \  \tau'-p_i \leq\Delta_i\big]\Big) 
        + \sum_{\substack{\tau\leq p_i <\tau' \\  \text{s.t} \\  \tau'-p_i \leq\Delta_i}}\!\!\!(\tau'-p_i)\Bigg)
        \end{gather*} 
    }
    \end{enumerate}

    $S'(\tau,\kappa,\eta)$ keeps track of the optimal set of targets corresponding to $S(\tau,\kappa,\eta)$.
\end{algorithm}
\setlength{\algoheightrule}{1pt} 
\setlength{\algotitleheightrule}{1pt} 
\end{minipage}
}

\subsection{Extension 2: Optimizing the Number of Target Levels}
The nonmonotonicity property may make adding a new target level to the current placement reduce the maximum improvement (see~Figure~\ref{fig:nonmonotone_sub}), or wasteful if we place the new target level somewhere no agent can reach or on top of an existing target. Therefore, when considering $k = 1, 2, \ldots, n$, it is possible that the maximum total improvement is achieved at $k < n$. Using the dynamic program based on Algorithm~\ref{alg:recurrence-dp-one-group} we can find the minimum value of $k$ that satisfies this property and minimizes the number of targets subject to achieving maximum total improvement. Furthermore, by finding the total amount of improvement for different values of $k$, the principal can decide how many targets are sufficient to achieve a desirable total improvement (bi-criteria objective).

\subsection{Open Problem: Tightening the Approximation Gap}
Algorithm~\ref{alg:approx}, as stated in Theorem~\ref{thm:approx}, provides an $\Omega(1/g^3)$-approximation simultaneous guarantee compared to the optimal solution for each group using at most $k$ targets; and as stated in Theorem~\ref{thm:modified}, provides an $\Omega(1/g^2)$-approximation simultaneous guarantee compared to the optimal solution for each group using at most $\lceil k/g \rceil$ targets. Example~\ref{ex:lowerbound}, on the other hand, shows an instance where no solutions with $> 1/g$ simultaneous approximation for the groups is possible for either of the benchmarks. Therefore, there is a gap of $\mathcal{O}(g^2)$ for the first, and a gap of $\mathcal{O}(g)$ for the second benchmark. Finding the optimal order of approximation guarantees for these benchmarks and tight lower bounds are the main problems left open by our work.

\chapter{Generating Actionable Insights in Large State Spaces to Help People Reverse Unfavorable Decision Outcomes}
\label{chap:cfes}
\section{Introduction}
\label{sec:cfe_intro}

Machine learning models are increasingly being used to guide high-stakes decision-making processes. Given the potential impact on individuals' livelihoods, society demands transparency and the right to an explanation, as outlined in Articles 13–15 of the \citet{EuropeanParliament2016a} General Data Protection Regulation and Article 13 of the \citet{EuropeanParliament2025} AI Act.
A critical aspect of this transparency is understanding how individuals (agents) can modify their input features to achieve a desired outcome, such as a positive label in a binary classification setting. 
Recourse or counterfactual explanation (CFE) generators that provide actionable insights offer one such solution~\citep{Wachter2017, Ustun19, Shalmali19, Dandl2020, Mothilal20, Karimi21, Karimi22}\footnote{Following prior work ~\citep{pawelczyk2022exploring, Rasouli24, Jiang2024, Verma2020CounterfactualEF}, we use the terms recourse and CFE interchangeably. For a detailed discussion and a nuanced differentiation of these terms, we refer the reader to Section 2 of \citet{Karimi22} and Section 3.3 of \citet{Verma2020CounterfactualEF}.}.

A popular form of recourse generation, actionable recourse~\citep{Ustun19}, generates feature-based CFEs, specifying precise adjustments to features (state) to ensure that the new features collectively result in a positive classification.
For comparison with our work, we refer to these as \emph{low-level} CFEs.  While helpful, low-level CFEs are overly specific and might be challenging to translate into real-world-like actions. To address this limitation, we introduce three novel forms of recourse that align with real-world actions: high-level continuous (\textit{hl-continuous}), high-level discrete (\textit{hl-discrete}), and high-level ID (\textit{hl-id}) CFEs.
\begin{figure}[t!]
\centering
    \begin{tikzpicture}[
        node distance=1cm and 2cm, 
        every node/.style={font=\scriptsize},  
        box/.style={draw, text width=4cm, align=left, rounded corners, fill=gray!15, line width=0.8mm, draw=gray!30}, 
        thick,
    ]

    \definecolor{llcolor}{RGB}{138,192,165}
    \definecolor{hccolor}{RGB}{233,162,128}
    \definecolor{dacolor}{RGB}{141,160,203}
    
    \node[box, fill=gray!15, text width=4.5cm] (agent1) at (0, 7.5) {
        Calcium (mg): \(\mathbf{309}\)\\
        Carbohydrate (gm): \(\mathbf{109.45}\)\\
        Copper (mg): \(\mathbf{0.425}\)\\
        Dietary fiber (gm): \(\mathbf{4.1}\)\\
        Iron (mg): \(\mathbf{4.08}\)\\
        Magnesium (mg): \(\mathbf{96}\)\\
        \parbox{\linewidth}{\centering \(\vdots\)}
    };
    \node[box, fill=gray!15, text width=3.66cm] (agent2) at (5.2, 7.2) {
        PhysActivity: \(\mathbf{1}\)\\
        Fruits: \(\mathbf{0}\)\\
        Veggies: \(\mathbf{0}\)\\
        AnyHealthcare: \(\mathbf{1}\)\\
        LowBP: \(\mathbf{0}\)\\
        NoSmoke: \(\mathbf{1}\)\\
        LowChol: \(\mathbf{0}\)\\
        HealthBMI: \(\mathbf{0}\)\\
        \parbox{\linewidth}{\centering \(\vdots\)}
    };
    \node[box, fill=gray!15, text width=4.77cm] (agent3) at (10.5, 5) {
        Education: BSc in computer science\\
        Coding experience: \(\mathbf{4}\) years\\
        Leadership: \textbf{none}\\ 
        Published research: \textbf{none}
    };

    \node[box, fill=llcolor!80, text width=4.5cm] (low1) at (0, 3.5) {
        Calcium (mg): \(\mathbf{309 \to 113}\) \\
        Carbohydrate (gm): \(\mathbf{109.45 \to 43.376000000000005}\) \\
        Copper (mg): \(\mathbf{0.425 \to 0.21295000000000001}\) \\
        Dietary fiber (gm): \(\mathbf{4.1 \to 50.113950000000024}\) \\
        Iron (mg): \(\mathbf{4.08 \to 42.729460000000002}\) \\
        Magnesium (mg): \(\mathbf{96 \to 57}\) \\
        \parbox{\linewidth}{\centering \(\vdots\)}
    };
    \node[box, fill=llcolor!80, text width=3.77cm] (low2) at (5.2, 3.3) {
        Fruits: \(\mathbf{0 \to 1}\)\\
        Veggies: \(\mathbf{0 \to 1}\)\\
        LowBP: \(\mathbf{0 \to 1}\)\\
        LowChol: \(\mathbf{0 \to 1}\)\\
        HealthBMI: \(\mathbf{0 \to 1}\)\\
        \parbox{\linewidth}{\centering \(\vdots\)}
    };  
    \node[box, fill=llcolor!80, text width=4.77cm] (low3) at (10.5, 2.5) {
        Coding experience: \(\mathbf{4 \to 5+}\) years\\
        Leadership: \(\mathbf{none \to 1+}\) years \\ 
        Published research: \(\mathbf{none \to 1+}\)
    };

    \node[box, fill=hccolor!80, text width=4.5cm] (high1) at (0, 0) {
        Take ``\textit{leavening agents:  cream of tartar}''  (\(ca_1\)) \\ 
        Take ``\textit{fish, tuna, light, canned in water, drained solids}''  (\(ca_2\))
    };
    \node[box, fill=dacolor!80, text width=3.77cm] (high2) at (5.2, 0.05) {
        Pickup the ``\textit{lisinopril and atorvastatin prescription from the pharmacy}'' (\(da_1\))\\ 
        Adopt ``\textit{a keto diet}'' (\(da_2\))
    };
    \node[box, fill=blue!5, text width=4.77cm] (high3) at (10.5, -0.3) {
         Complete the Google AI residency program
    };

    \node[above=0.05cm of agent1, black] {\textbf{The agent profile (app1)}};
    \node[above=0.05cm of agent2, black] {\textbf{The agent profile (app2)}};
    \node[above=0.05cm of agent3, black] {\textbf{The agent profile (app3)}};
    
    \node[below=0.05cm of low1, black] {\textbf{The low-level CFE}};
    \node[below=0.05cm of low2, black] {\textbf{The low-level CFE}};
    \node[below=0.05cm of low3, black] {\textbf{The low-level CFE}};
    
    \node[below=0.05cm of high1, black] {\textbf{The hl-continuous CFE}};
    \node[below=0.05cm of high2, black] {\textbf{The hl-discrete CFE}};
    \node[below=0.05cm of high3, black] {\textbf{The hl-id CFE}};

    \draw [red, dashed, thick, -stealth] (agent1.south) -- (low1.north);
    \draw [red, dashed, thick, -stealth] (agent1.west) -- ++(-0.33cm, 0) |- (high1.west);
    
    \draw [red, dashed, thick, -stealth] (agent2.south) -- (low2.north);
    \draw [red, dashed, thick, -stealth] (agent2.west) -- ++(-0.33cm, 0) |- (high2.west);
    
    \draw [red, dashed, thick, -stealth] (agent3.south) -- (low3.north);
    \draw [red, dashed, thick, -stealth] (agent3.west) -- ++(-0.33cm, 0) |- (high3.west);
    
    \end{tikzpicture}
    \caption[A comparative analysis of low-level CFEs with three high-level types]{A comparative analysis of low-level CFEs with three high-level types: hl-continuous, hl-discrete, and hl-id, offering actionable insights to help negatively classified agents (with initial profiles/states in grey) achieve positive outcomes from binary classifiers:  dietary changes to achieve a healthy waist-to-hip ratio (app1), guide an agent (app2) to meet wellness check criteria, and help an agent (app3) qualify for an AI junior research engineer role. In all cases, the low-level CFE precisely specifies which features to modify and by how much. The hl-continuous CFE adjusts multiple features numerically, e.g., corresponding to features in the profile (app3), \(ca_1 = [8.0, 61.5, 0.195, 0.2, 3.72, 2.0, \cdots], ca_2 = [17.0, 0.0, 0.05, 0.0, 1.63, 23.0, \cdots]\)  but the agent does not need to understand the exact changes to follow the CFE. The hl-discrete CFE ensures features meet specific eligibility thresholds without the agent needing to know the exact adjustments to take the CFE, e.g., \(da_1 = [0, 0, 0, 0, 1, 0, 1, 0, \cdots],  da_2 = [0, 1, 1, 0, 0, 0, 0, 1, \cdots]\). The hl-id CFE provides a single overarching high-level action that favorably modifies all the features it should.
    This figure illustrates how different types of CFEs might impact the agent's ability to interpret and act on given recourse.}
    \label{fig:main_figure1}
    \vspace{-0.2cm}
\end{figure}
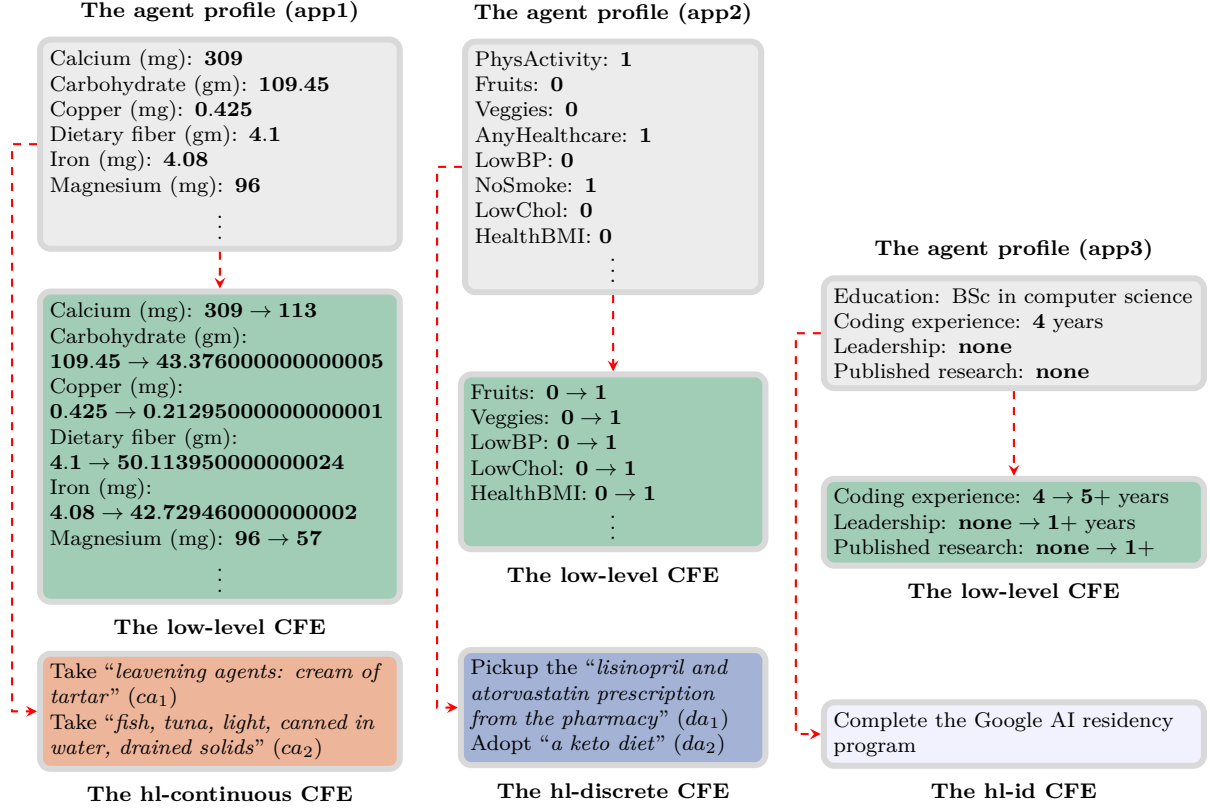

Figure~\ref{fig:main_figure1} presents an illustrative example of hypothetical recourse across three binary classification tasks, highlighting the distinctions between low- and high-level CFEs. Specifically, the hl-continuous CFEs involve general and predefined real-world-like actions, simultaneously modifying multiple features through numerical adjustments. Similarly, hl-discrete CFEs also stem from real-world actions and might modify several features simultaneously. However, unlike hl-continuous CFEs, which adjust feature values, hl-discrete CFEs modify feature eligibility.
The hl-discrete CFEs use binary vector actions to ensure all features meet a predefined threshold, reducing decisions to unit simple yes/no questions. It is particularly efficient in scenarios where feature satisfiability depends on a set threshold, such as level-one decision-making in wellness checks (see Figure~\ref{fig:main_figure1} 
and Appendix~\ref{sec:cfe_app_hld-hlc-diff}).
Lastly, the hl-id CFE encapsulates either hl-continuous or hl-discrete CFEs into a single overarching action, encompassing all necessary changes without detailing specific feature modifications. This level of abstraction relies on domain knowledge and often conveys significant implicit information.

\paragraph{Our Contributions.}
First, we introduce three forms of recourse: hl-continuous, hl-discrete, and hl-id CFEs, that bridge the gap between feature-based and real-world actions (Sections~\ref{sec:cfe_intro} and \ref{sec:cfe_background}). 
Second, we propose single-agent CFE generation methods that leverage predefined, real-world-like actions to generate optimal CFEs. 
Specifically, we formulate single-agent hl-discrete CFE generation as a weighted set cover problem and single-agent hl-continuous CFE generation as an integer linear programming (ILP) problem (Section~\ref{sec:cfe_single_agent_gen}).

Third, we propose data-driven approaches that, given instances of agents and their optimal CFEs (agent–CFE dataset), learn a CFE generator that will quickly provide optimal hl-continuous, hl-discrete, and hl-id CFEs for new agents. Unlike the expensive single-agent CFE generation approach, the data-driven methods are more computationally friendly. Additionally, these methods are especially favorable when historical instances of agents and their optimal CFE data are available or aggregatable, recourse generators can operate independently of decision-makers, query access to the classifier is restricted or unavailable, and the actions along with their costs and explicit effects on features are unknown (Section~\ref{sec:cfe_data_driven_gen}). 
To the best of our knowledge, these alternative forms of recourse and the data-driven approach for generating CFEs are novel contributions.

Finally, we conduct extensive experiments on \(30\) agent–CFE datasets derived from real-world healthcare datasets: BRFSS, Foods, and NHANES. We chose these datasets due to the availability of a wealth of publicly accessible data, which allows us to effectively demonstrate the benefits of incorporating real-world-like actions in CFE generation and sufficiently explore the impact of a larger action space (or action grid according to \citet{Ustun19}), enabled by the high number of actionable features with broad value ranges. Alongside these real-world datasets, we also include \(34\) fully-synthetic agent–CFE datasets for comparison. 
We extensively compare the low-level CFEs with the hl-continuous and hl-discrete CFEs, and provide an in-depth analysis of the performance of the data-driven hl-continuous, hl-discrete, and hl-id CFE generators under various settings (Sections~\ref{sec:cfe_exp_setup} and \ref{sec:cfe_results}). Our code can be accessed here: \href{https://anonymous.4open.science/r/DataDrivenCFEGenerators-27D9/README.md}{link}.

\section{Background}
\label{sec:cfe_background}

We consider a binary classification setting, where an agent in the state \(\mathbf{x} \in \mathcal{X}\) receives either a positive (desirable) or negative (undesirable) outcome under a model \(f(\mathbf{x})\). The state space \(\mathcal{X}\) consists of all valid agent states, each represented by a feature vector capturing attributes, such as age and calcium(mg). 
The model operates over this space, and the CFE generator searches within it to identify alternative states with favorable model outcomes. Although we focus on binary classification, our data-driven CFE framework generalizes to other settings. Given an undesirable outcome, the CFE generator suggests actionable changes to move the agent at state \(\mathbf{x}\) to a new state \(\mathbf{x}' \in \mathcal{X}\) where the model prediction is desirable. Actionable recourse~\citep{Ustun19} provides low-level CFEs that specify feature-level changes needed to reach such a state.

\paragraph{The low-level CFE generator.} \citet{Ustun19} proposed an ILP-based low-level CFE generator (Equation~\ref{eq:act_recourse}) that generates a low-level CFE to help an agent change an undesirable model outcome to a desirable one. 
\begin{equation}
    \begin{aligned}
    \min \quad &\text{cost}(\mathbf{a}; \mathbf{x}) \\
    \text{s.t.} \quad &f(\mathbf{x} + \mathbf{a}) =  \hat{y}^{\star}\\
    &\mathbf{a} \in A(\mathbf{x}),
    \end{aligned}
    \label{eq:act_recourse}
\end{equation}
where \(\hat{y}^{\star}\) is the desired model outcome, \(A(\mathbf{x})\) denotes the set of feasible actions given the input \(\mathbf{x}\), and the function \(\text{cost}(\cdot ; \mathbf{x}) : A(\mathbf{x}) \to \mathbb{R}_+\) encodes the preferences between these actions. When Equation~\ref{eq:act_recourse} is feasible, the optimal actions that modify the features (i.e., \(\mathbf{x} + \mathbf{a}\)) and lead to a desirable model outcome are recommended to the agent (cf. Figure~\ref{fig:main_figure1}).
We refer the reader to \citet{Ustun19} for a more detailed description and to Appendix~\ref{subsec:cfe_app_lowlevel_cfe} for dataset-specific experimental setup and supplementary examples of low-level CFEs.

\paragraph{Shortcomings of Low-level CFEs and their Generators.} 
We note two limitations of low-level CFEs in comparison to our proposed forms of recourse (hl-continuous, hl-discrete, and hl-id CFE), and two, as we compare the low-level CFE generator to the proposed data-driven CFE generation approach \citep{hiddenSolon,Karimi22,Verma2020CounterfactualEF}. 

First, low-level CFEs are feature-based and highly specific (e.g., in Figure~\ref{fig:main_figure1}, app1, calcium (mg): \(309 \to 113\)), which may overwhelm agents and introduce additional costs to translate the CFE into implementable steps. 
In contrast, our proposed CFEs are better aligned with real-world scenarios, offering \textit{what you see is what you get} actionable insights (see Figure~\ref{fig:main_figure1}). Furthermore, with low-level CFEs, details about the actions the agent implements (the number of them needed, which features they would simultaneously modify, and costs to incur) are often unknown beforehand, potentially leading to a misleading price of recourse and related metrics such as sparsity (few modified features) and proximity (closeness of final state to initial state) \citep{hiddenSolon}.

Second, although a CFE (e.g., complete the Google AI residency program in Figure~\ref{fig:main_figure1}) could have been optimal for several agents with different but close profiles (e.g., one has coding experience: \(4\) and another \(3\)), the low-level CFE being too specific, would give the agents different CFEs (coding experience: \(4 \to 5+\) years and coding experience: \(3 \to 5+\) years). In contrast, our proposed recourse generation approaches are both agent-specific (tailored to an agent's initial state) and generalizable (providing similar recommendations to agents with comparable profiles).

Lastly, while most low-level CFE generators operate on a single-agent basis \citep{Karimi22,Verma2020CounterfactualEF}, recent work by \citet{Pedapati2020,Rawal2020,pmlr-v151-kanamori22a,Ley2023} and \citet{Carrizosa24} propose global low-level CFE generators capable of producing CFEs for multiple agents. However, these methods generate feature-based CFEs and require, at a minimum, query access to the classifier. In contrast, we propose data-driven CFE generation approaches that, given historical mappings between agents and their optimal CFEs, such as healthcare intervention records or high school counselors’ past successful college recommendations, can learn to generate CFEs with real-world-like actions for multiple new agents without requiring re-optimization. Additionally, the proposed data-driven approaches work well in settings where access to critical information, such as sufficient classification training data, classifier, or a comprehensive list of actions and their costs, is restricted or inaccessible.

\section{The Proposed Single-Agent CFE Generators}
\label{sec:cfe_single_agent_gen}

This section outlines the single-agent CFE generators for the proposed hl-continuous and hl-discrete CFEs. Each generator relies on predefined, real-world-like actions to solve optimization problems for CFE generation: the weighted set cover problem for hl-discrete CFEs and ILP for hl-continuous CFEs.

\subsection{The Single-Agent hl-continuous CFE Generation}
\label{subsec:cfe_hlc_sa}

Below, we formally define hl-continuous actions and the single-agent hl-continuous CFE generation process for outcomes predicted by linear classification models.

\begin{definition}[The hl-continuous action]  
An hl-continuous action is a signed (\(\pm\)) and predefined real-world action whose cost and varied effects on an agent's input features are predefined and known. For example, in Figure~\ref{fig:main_figure1}, the hl-continuous action: \(ca_1\): take ``\textit{leavening agents: cream of tartar}'' modifies \(6+\) features by a known amount and the agent incurs a cost (e.g., estimated average price in USD) that is known apriori.
\end{definition}

\paragraph{The singe-agent hl-continuous CFE generator.} This generator produces an hl-continuous CFE by solving an integer linear program (ILP). Given the profile of a negatively classified agent \(\mathbf{x}\) and a set of hl-continuous actions with known costs (defined above), the objective is to identify the lowest-cost subset of  hl-continuous actions that, when taken, modify the agent's features to achieve a positive classification. The ILP is of the form:
\begin{equation}
    \begin{aligned}
    \text{minimize} \quad &\sum_{j \in J} \text{cost}_{j}a_{j} \\
    \text{s.t.} \quad &\mathbf{c}^{T} \sum_{j \in J} a_j \cdot (2\epsilon_j - 1) \cdot \mathbf{v}_j \geq - (\mathbf{c}^{T}\mathbf{x} + b) + \delta\\
    \quad &\epsilon_j \in \{0, 1\}, \quad a_j \in \{0, 1\}, \quad \forall j \in J
    \end{aligned}
    \label{eq:hl-continuous}
\end{equation}
where \(J\) denotes the indices of the hl-continuous actions, with each action represented by a vector \(\mathbf{v}_{j}\) and with a predefined cost, \(\text{cost}_{j} \in \mathbb{R}_{+}\). The boolean variable \( a_{j} \) indicates the inclusion (\( a_j=1 \)) or exclusion (\( a_j=0 \)) of the \( j^\textrm{th} \) hl-continuous action, while \(\epsilon_{j}\) encodes the sign of this action, representing addition (\(\epsilon_j=1\)) or subtraction (\(\epsilon_j=0\)). The coefficients \(\mathbf{c}\) and intercept \(b\) are the parameters of the linear  classifier, and \(\delta\) is a small positive value that ensures strict inequality.

\subsection{The Single-Agent hl-discrete CFE Generation}
\label{subsec:cfe_hld_sa}

Below, we formally define hl-discrete actions and the single-agent hl-discrete CFE generation process based on a threshold classifier that determines feature eligibility.

\begin{definition}[The hl-discrete action]  
An hl-discrete action represents a binary vector that adds capabilities to specific features to meet the eligibility threshold.  For example, consider the agent state \(\mathbf{x} = [0, 0, 0, 0, 1]\) and the hl-discrete action \(\mathbf{v}_{j} = [1, 1, 0, 0, 0]\). When taken, the hl-discrete action adds capabilities to features \(1\) and \(2\) of \(\mathbf{x}\), transforming it to a new state \([1, 1, 0, 0, 1]\). Although we focus on binary actions, the formulation is extensible to more general cases.
\end{definition}

\paragraph{The singe-agent hl-discrete CFE generator.} This generator produces an hl-discrete CFE by solving a weighted set cover problem. Specifically, it identifies the lowest-cost subset of hl-discrete actions, each with a predefined cost, that a negatively classified agent \(\mathbf{x} \in \{0, 1\}^{n}\) (e.g., someone deemed a health risk) can take to achieve a desirable classification (e.g., no longer classified as a health risk). The problem can be formally defined as follows:
\begin{equation}
    \begin{aligned}
    \textrm{minimize}\quad & \sum_{j\in J}{\text{cost}_{j}a_{j}}\\
    \textrm{s.t.} \quad & \sum_{j\in J}{d_{ji}a_{j}} + x_{i} \geq t_{i}, \ \forall i \in [n],\\
                         & a_{j} \in \{0, 1\}, \ d_{ji} \in \{0, 1\},   \\
    \end{aligned}
    \label{eq:hl-discrete}
\end{equation}
where \(J\) are the indices of the hl-discrete actions, each represented by a vector \(\mathbf{v}_{j}\) and with a predefined cost: \(\text{cost}_{j} \in \mathbb{R}_{+}\). The threshold classifier \(\mathbf{t} = \{t_{1}, t_{2},\cdots, t_{n} \}\) over \(n\) features classifies an agent state \(\mathbf{x}\) positive if \(x_{i} \geq t_{i}, \ \forall i \in [n]\), and negative otherwise. The binary variable \(a_{j}\) denotes inclusion (\(a_{j}=1\)) or exclusion (\(a_{j}=0\)) of the \(j^\textrm{th}\) hl-discrete action, while \(d_{ji}\) indicates whether the  \(j^\textrm{th}\)  hl-discrete action transforms (adds capabilities to) the feature \(i\) of the agent state \(\mathbf{x}\), i.e., when performed, the new agent state \(\mathbf{x}+\mathbf{v}_{j} = \mathbf{x}'\) is such that \(x'_{i} > x_{i}\) and \(x'_{i} \geq t_{i}\).

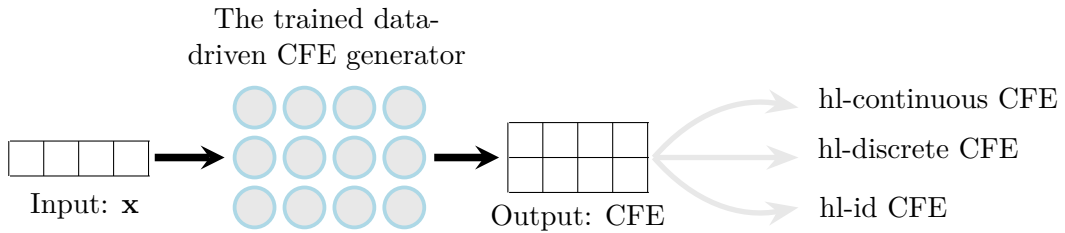
\begin{figure}[b!]
\centering
    \begin{tikzpicture}[scale=0.66]
      \def\radius{0.4}
      \def\vspacing{1}
      \def\vspacingLine{1.0} 
      \def\numRows{1}
      \def\numCols{4}
      
      \def\numColsr{4}
      \def\numRowsr{2}
      \def\cellSize{0.7}

    \begin{scope}[shift={(-3cm,1.33cm)}]
      \foreach \i in {0,...,\numRows} {
        \draw (0, -\i*\cellSize) -- (\numCols*\cellSize, -\i*\cellSize);
      }
    \end{scope}
    
    \begin{scope}[shift={(-3cm,1.33cm)},xshift=-\cellSize]
      \foreach \j in {0,...,\numCols} {
        \draw (\j*\cellSize, 0) -- (\j*\cellSize, -\numRows*\cellSize);
      }
    \end{scope}
    \node[below] at (-1.5cm, 0.5cm) {Input: \(\mathbf{x}\)};

    \begin{scope}[shift={(7cm,1.7cm)}]
      \foreach \i in {0,...,\numRowsr} {
        \draw (0, -\i*\cellSize) -- (\numColsr*\cellSize, -\i*\cellSize);
      }
    \end{scope}
    
    \begin{scope}[shift={(7cm,1.7cm)},xshift=-\cellSize]
      \foreach \j in {0,...,\numColsr} {
        \draw (\j*\cellSize, 0) -- (\j*\cellSize, -\numRowsr*\cellSize);
      }
    \end{scope}
    
    \node[below] at (8.375cm, 0.25cm) {Output: \(\text{CFE}\)};

    \draw[-stealth, line width=0.1cm, black] (-0.1, 1) -- (1.3, 1) node[midway, above, text=black] {};
    \draw[-stealth, line width=0.1cm, black] (5.5, 1) -- (6.8, 1) node[midway, above, text=black] {};

      \foreach \x in {1,2,3}
        \filldraw[fill=grey, draw=paleblue, line width=0.05cm] ({1.9}, {(\x-1)*\vspacing}) circle (\radius);
    
      \foreach \x in {1,2,3}
        \filldraw[fill=grey, draw=paleblue, line width=0.05cm] ({2.9}, {(\x-1)*\vspacing}) circle (\radius);
    
      \foreach \x in {1,2,3}
        \filldraw[fill=grey, draw=paleblue, line width=0.05cm] ({3.9}, {(\x-1)*\vspacing}) circle (\radius);
    
      \foreach \x in {1,2,3}
        \filldraw[fill=grey, draw=paleblue, line width=0.05cm] ({4.9}, {(\x-1)*\vspacing}) circle (\radius);
    
      \node[above, text width=5cm, align=center] at (3.35cm, 2.55cm) {The trained data-driven CFE generator};

      \coordinate (start) at (9.9,1);
      \draw[->, >=Stealth, bend left=30, line width=2pt, grey](start) to node[above] {} (12.8,2);
      \draw[->, >=Stealth, bend left=0, line width=2pt, grey] (start) to node[right] {} (12.8,1);
      \draw[->, >=Stealth, bend right=30, line width=2pt, grey] (start) to node[below] {} (12.8,0);
    
      \node at (15.6,2.2) {hl-continuous CFE} ;
      \node at (15.2,1.2) {hl-discrete CFE};
      \node at (14.5,0.0) {hl-id CFE};
    
    \end{tikzpicture}
\caption[Data-driven CFE generation]{Given an agent state (profile) \(\mathbf{x}\), a data-driven CFE generator trained on instances of agents and their optimal CFEs (agent–CFE dataset) generates a high-level CFE (hl-continuous, hl-discrete or hl-id) for the agent without the need for generator re-optimization or access to the decision-making classifier.  \label{tikz:all_cfe_gens}}
\end{figure}

\section{The Proposed Data-Driven CFE Generators}
\label{sec:cfe_data_driven_gen}

This section, supplemented by Appendix~\ref{sec:cfe_app_archictures}, details the three proposed data-driven CFE generators: hl-continuous, hl-discrete, and hl-id (see Figure~\ref{tikz:all_cfe_gens}). 
Each generator learns from instances of agents and respective optimal CFEs, defined as the least cost CFE that leads to a favorable model outcome. Once trained, these generators can produce optimal CFEs for new agents without re-optimization. Empirical results demonstrate that even shallow deep-learning architectures perform strongly at this task, that is, generate \textit{correct} CFEs, the least cost CFEs that favorably flip the model outcome. 

The data-driven approaches are computationally more efficient than single-agent CFE generation approaches, which require optimization for each new agent. Furthermore, they are particularly favorable when agent–CFE data is available or aggregatable from various sources. They allow recourse generation to operate independently of decision-makers, function without direct access to the classifier, and handle scenarios where action costs and their explicit effects on features are unknown.

\subsection{The Data-Driven hl-continuous CFE Generator}
\label{subsec:cfe_hlc_dd}

We develop a data-driven hl-continuous CFE generator trained on agent–hl-continuous CFE training dataset, consisting of agents and their corresponding optimal hl-continuous CFEs. Each CFE defines a set of hl-continuous actions along with their associated costs. For instance, a CFE might include actions \{action-a, action-b, action-c\} with corresponding costs \{cost-a, cost-b, cost-c\}. The generator learns from this data to produce hl-continuous CFEs for new agents without requiring generator re-optimization. Below are further details of the generator architecture.

Given the agent–hl-continuous CFE training dataset, we train a neural network model to learn to generate hl-continuous CFEs for testing set agents. Specifically, the generator is a neural network model with three hidden layers, each containing \(2,000\) neurons. The model incorporates \(\ell_2\) regularization, dropout, and batch normalization. The training process uses the Adam optimizer~\citep{kingma2014adam} with early stopping, restoring the best weights after a patience level of \(300\). The model trains with a batch size of \(6,000\)  for an average of \(5,000\)  epochs.
To ensure accurate data-driven hl-continuous CFE generation for both training and testing set agents, we optimize the model loss function \(\mathcal{L}_\textrm{HC}\) given by:
\begin{equation}
 \mathcal{L}_\textrm{HC} = -\frac{1}{M} \sum_{m=1}^{M} \sum_{j=1}^{J} \left[ a_{jm} \log(\hat{a}_{jm}) + (1 - a_{jm}) \log(1 - \hat{a}_{jm}) \right]
\label{eq:hl-continuous-loss}
\end{equation}
where \(\hat{a}_{jm}\) is the predicted probability and \(a_{jm}\) is the true indication of a presence (\(1\)) or absence (\(0\)) of the \(j^\textrm{th}\) hl-continuous action in agent \(m\)'s hl-continuous CFE. There are \(J\) possible hl-continuous actions and \(M\) agents in the agent–hl-continuous CFE training dataset.

\subsection{The Data-Driven hl-discrete CFE Generator}
\label{subsec:cfe_hld_dd}

We propose a data-driven hl-discrete CFE generator, trained using the agent–CFE training dataset and evaluated on the agent–CFE testing dataset. Each agent–CFE dataset comprises instances of agents and their optimal hl-discrete CFEs that specify a set of hl-discrete actions and associated costs. 

Given the agent–hl-discrete CFE training dataset, we design a sequential encoder-decoder model to generate hl-discrete CFEs for new agents without generator re-optimization. 
The model configuration was dependent on the experimental setting. On average, we used \(500\) training epochs with a batch size of \(128\), a dropout rate of \(0.4\), a learning rate of \(0.0005\), and either the mean squared error or binary cross-entropy loss as the objective function. The models, on average, consisted of three layers, each using ReLU activation functions.

\subsection{The Data-Driven hl-id CFE Generator}
\label{subsec:cfe_hlid_dd}

The data-driven hl-id CFE generator is a supervised learning model trained on agent–CFE pairs consisting of agents' initial states (profiles) and their corresponding optimal hl-id CFEs (agent–hl-id CFE dataset). Each CFE includes the hl-id CFE itself and its associated cost. 
When trained, the generator learns to generate CFEs for new agents without requiring re-optimization. Below, we provide details of the generator architecture.

Given the agent–hl-id CFEs training dataset, we design the data-driven hl-id CFE generator as a neural network model with an average of two hidden layers, each consisting of \(2000\) neurons, \(\ell_2\) regularization, dropout, and batch normalization. 
We used the Adam optimizer~\citep{kingma2014adam} and implemented early stopping and restoration of the best weights after a patience level of \(360\). On average, we set the batch size to \(2000\) and the number of epochs set to \(3000\). 
To ensure that the data-driven hl-id CFE generator  performs well on the training dataset and accurately generates hl-id CFEs for agents in the testing set, we optimize the model loss function \( \mathcal{L}_\textrm{HiD}\) given by:
\begin{equation}
 \mathcal{L}_\textrm{HiD} = -\frac{1}{M} \sum_{m=1}^{M} \sum_{k=1}^{K} \left[ a_{km} \log(\hat{a}_{km}) \right]
\label{eq:hl-id-loss}
\end{equation}
where \(\hat{a}_{km}\) is the predicted probability and \(a_{km}\) is the true indication of the \(k^\textrm{th}\) CFE being the hl-id CFE (\(1\)) or not (\(0\)) for the \(m^\textrm{th}\) agent. There are \(K\) possible hl-id CFEs and \(M\) agents in the training dataset.

\section{Experimental Setup}
\label{sec:cfe_exp_setup}

This section provides a detailed description of the experimental setup, including the evaluation metrics employed and the methodology for generating the  \(34\) fully-synthetic agent–CFE datasets and the \(30\) semi-synthetic agent–CFE datasets from real-world healthcare data sources: BRFSS, Foods, and NHANES.

\subsection{Real-world Datasets}
\label{subsec:cfe_real-world_datasets}

Below, we outline the extraction and preprocessing of the real-world datasets used in our experiments, including their statistical descriptions. We split all datasets into an \(80/20\) ratio for training and testing. 

\paragraph{The Foods, BMI, and WHR datasets.}
We extracted the Foods dataset from \citet{usda2016,awram_food_nutritional_values} and the BMI (body mass index) and WHR (waist-to-hip ratio) datasets from NHANES body measurement surveys ~\citep{cdc_nhanes,icpsr_series39}, covering the years \(1999\) to pre-pandemic \(2020\).

To ensure commonality in actionability features between the (Foods, BMI) and the (Foods, WHR) dataset pairs, we selected intersectional nutritional intake features: \textit{protein (gm), carbohydrate (gm), dietary fiber (gm), calcium (mg), iron (mg), magnesium (mg), phosphorus (mg), potassium (mg), sodium (mg), zinc (mg), copper (mg), selenium (mcg), vitamin C (mg), niacin (mg), vitamin B6 (mg), total folate (mcg), vitamin B12 (mcg), total saturated fatty acids (gm), total monounsaturated fatty acids (gm)}, and \textit{total polyunsaturated fatty acids (gm)}.

After preprocessing, for example, removing missing data and ensuring that selected nutritional intake features were a subset of the intersectional ones, the Foods dataset contained \(3901\) food items. Each item includes nutritional composition. We added two cost attributes: Monetary cost (in USD, obtained via web scraping) and Caloric cost (reflecting total caloric content, sourced from ~\citep{calories-ledger}).
In our experiments,  Foods+costs serves as the \textbf{hl-continuous action space}, where food items represent actions, and the cost attributes define the cost constraints. Based on cost type, we define two forms of hl-continuous actions: (i) Foods+monetary cost and (ii) Foods+caloric cost.

The BMI dataset after preprocessing contained \(50918\) agents, each with \(3\) demographic features and \(19\) nutrient intake features, classified as either healthy (\(1\)) or unhealthy (\(0\)) BMI. On the other hand, the WHR dataset contained \(9120\) agents, each with \(3\) demographic features and \(20\) nutrient intake features, classified as either healthy (\(1\)) or unhealthy (\(0\)) WHR. For additional preprocessing details, see Appendix~\ref{subsec:cfe_bmi_whr_preproc}.

\paragraph{The BRFSS dataset.} We extracted the Behavioral Risk Factor Surveillance System (BRFSS) dataset from ~\citet{teboul_diabetes_health_indicators_2024,cdc_brfs_2024}. 
After preprocessing, e.g., removing missing data, the dataset was reduced to  \(13,799\) agents, each represented by \(16\) binary health risk features. These include:  \textit{LowBP, LowChol, HealthBMI, NoSmoke, NoStroke, NoCHD, PhysActivity, Fruits, Veggies, LightAlcoholConsump, AnyHealthcare, DocbcCost, GoodGenHlth, GoodMentHlth, GoodPhysHlth} and \textit{NoDiffWalk}. For additional details, see Appendix~\ref{subsec:cfe_brfss_preproc}.

\subsection{Single-Agent CFE Generation}
\label{subsec:cfe_sa_cfe_gen}

Here and in Appendices~\ref{subsec:cfe_app_lowlevel_cfe}, \ref{subsec:cfe_app_hlc_cfe}, and \ref{subsec:cfe_app_hld_cfe}, we describe the generation of low-level, hl-continuous, and hl-discrete CFEs using single-agent CFE generators (i.e., Equations~\ref{eq:act_recourse}, \ref{eq:hl-continuous}, \ref{eq:hl-discrete}). Since the agents and their computed CFEs will also be used to train and evaluate data-driven CFE generators, we generate CFEs for negatively classified agents in the BMI, WHR, and BRFSS training and testing datasets. 

For BMI and WHR datasets, only intersectional nutritional features are considered actionable, whereas all features are actionable for the BRFSS dataset. We trained binary classifiers to identify agents requiring CFEs and identify classifier parameters to use in single-agent CFE generators. 
Fine-tuned logistic regression models for BMI and WHR achieved test accuracies of \(72.78\%\) and \(85.18\%\), respectively. For the BRFSS dataset, which focuses on wellness checks,  a threshold classifier \(\mathbf{t} = \mathbf{1}_{16}\)  achieved \(100\%\) accuracy.

\paragraph{The single-agent low-level CFE generation.} We generated a low-level CFE for each negatively classified agent in the BMI, WHR, and BRFSS training/testing datasets. We accomplished this by using the agent's initial state and the parameters of the trained binary linear decision-making classifiers for each dataset, along with the ILP framework defined in Equation~\ref{eq:act_recourse}.

\paragraph{The single-agent hl-continuous CFE generation.} For the BMI and WHR training/testing datasets, we use the negatively classified agents alongside two types of hl-continuous actions: Foods+monetary costs and Foods+caloric costs to create hl-continuous CFEs. Using the ILP framework defined in Equation~\ref{eq:hl-continuous}, we generate two distinct forms of hl-continuous CFEs for each agent: the optimal set of food items with minimal monetary cost and the optimal set of food items with minimal caloric cost.

\paragraph{The single-agent hl-discrete CFE generation.} Lastly, using the BRFSS training/testing set agents, the threshold classifier (\(\mathbf{t} = \mathbf{1}_{16}\)), and \(100\) synthetically generated hl-discrete actions (each of length \(16\)) with associated costs, we applied Equation~\ref{eq:hl-discrete} to generate a hl-discrete CFE for each agent. Each CFE represents an optimal set of hl-discrete actions with minimal costs for each agent.

\subsection{Data-Driven CFE Generation}
\label{subsec:cfe_dd_cfe_gen}

This section, along with Appendices~\ref{subsec:cfe_app_single_cfe}, \ref{subsec:cfe_app_fullysynthetic_datasets}, and \ref{sec:cfe_app_archictures}, describe the creation of agent–CFE datasets (where the CFE is either hl-continuous, hl-discrete, or hl-id), and their role in data-driven CFE generation.

\paragraph{The semi-synthetic agent–CFE datasets.} Using agent states and their optimal CFEs from Section~\ref{subsec:cfe_sa_cfe_gen}, we construct training and testing agent–CFE datasets. First, we generate: \(2\) agent–hl-continuous CFE train/test datasets for  BMI, \(2\) agent–hl-continuous CFE train/test datasets for WHR, and \(1\) agent–hl-discrete CFE train/test datasets for  BRFSS. 

Then, given the following agent–CFE datasets: the agent–hl-continuous CFE train/test datasets for BMI, created using Foods+monetary costs as hl-continuous actions; the agent–hl-continuous CFE train/test datasets for WHR, created using Foods+caloric costs as hl-continuous actions; and the agent–hl-discrete CFE train/test datasets for BRFSS, we generate \(3\) agent–hl-id CFE datasets. 
Specifically, for each agent–CFE dataset, we create a unique identifier for the CFE that denotes the single overarching action, resulting in an agent–hl-id dataset corresponding to instances of agents and their hl-id CFE. 

Lastly, for each of the agent–CFE datasets described above, we generated three variations based on the frequency of CFEs in the dataset: \texttt{all} (includes all data), \texttt{>10} (CFEs with more than 10 agents), and \texttt{>40}  (CFEs with more than 40 agents) varied frequency of CFEs agent–CFE datasets.

\paragraph{The fully-synthetic agent–CFE datasets.}
We use the ILP defined in Equation~\ref{eq:hl-discrete} to generate five variants of the agent–hl-discrete CFE datasets: varied dimensionality, frequency of CFEs, information access, feature satisifiability, and actions access. Below, we briefly describe the varied dimensionality and frequency of CFEs datasets and include more details about these and other variants in Appendix~\ref{subsec:cfe_app_fullysynthetic_datasets}.

For varied dimensions agent–CFE datasets, we generated datasets with \(20\), \(50\), and \(100\) dimensions (actionable features), where we set the agent's feature to \(1\) with a probability \(p_f\), and each discrete action can add capabilities to a feature with a probability \(p_a\). The cost of each action depends on the features it transforms.
Lastly, we created three varied frequency of CFEs datasets: \texttt{all}, \texttt{>10}, and \texttt{>40}, and agent–hl-id CFE datasets for each varied dimensions agent–CFE dataset, using a similar approach as in the semi-synthetic agent–CFE datasets described above.

\paragraph{Data-driven CFE generators.}
Given the semi-synthetic and fully-synthetic training agent–CFE datasets, we train the corresponding data-driven generators described in Section~\ref{sec:cfe_data_driven_gen} and evaluate their effectiveness on the testing agent–CFE datasets. See Appendix~\ref{sec:cfe_app_archictures} for supplemental details.

\subsection{Evaluation and Comparative Analysis Metrics}
\label{subsec:cfe_eval_metrics}

We compare single-agent generated low-level CFEs to both hl-continuous and hl-discrete CFEs. Additionally, we assess the performance of data-driven CFE generators. The metrics used for comparison and evaluation are detailed below and in Appendix~\ref{subsec:cfe_evaluation_mets}.

\paragraph{Accuracy of data-driven generators.}
To assess the accuracy of the proposed data-driven CFE generators, we use zero-one loss (see Equation~\ref{eq:evaluation}), which checks if the generated CFE \(\hat{I}\) matches the true CFE \(I\), defined as the least cost CFE that favorably flips the model outcome.
\begin{equation}
\mathcal{L}_\textrm{eval}(I, \hat{I}) = \begin{cases}
                                0 & \text{if } I = \hat{I}\\
                                1 & \text{if } I \neq \hat{I}
                                \end{cases}
    \label{eq:evaluation}
\end{equation}

\paragraph{Comparison metrics.} 
We analyze various factors related to the use of CFEs, including the average number of actions taken, the number of modified features, the proportion of agents sharing the same optimal CFE, and the overall improvement measured as the distance between an agent’s initial state and its final state after following a CFE. Assuming CFEs encourage truthful responses, we refer to this as agent \textit{improvement}. 
We compare these factors when agents follow a low-level CFE versus a high-level CFE, either hl-continuous or hl-discrete.
The comparative analysis focuses on CFEs generated by single-agent CFE generators for negatively classified agents in the training sets of three datasets: BMI, WHR, and BRFSS. To ensure a fair comparison, we include only agent–CFE pairs where both low- and high-level CFEs are available, as the low-level CFE generator (cf. Equation~\ref{eq:act_recourse}) occasionally fails to produce a CFE.

To assess how much each variable, e.g., number of modified features varies across groups, for example, between male and female agents, we compute the coefficient of variations (Equation~\ref{eq:cv}), a normalized measure of dispersion calculated as the ratio of the standard deviation to the mean of the variable \(v\). 
\begin{equation}
    \begin{aligned}
    & \textrm{coefficient of variation} (v) = \frac{\textrm{standard deviation}_{v}}{\textrm{mean}_{v}} \times 100
    \end{aligned}
    \label{eq:cv}
\end{equation}

\section{Experimental Results}
\label{sec:cfe_results}

In this section, we provide comprehensive empirical evidence showcasing the strong performance of our data-driven CFE generators and their advantages over single-agent CFE generators. 
Furthermore, we highlight the advantages of hl-continuous and hl-discrete CFEs, which offer actionable insights that closely align with real-world action spaces, over feature-based low-level CFEs.

\subsection{Comparison of Low-level CFEs to the hl-continuous and hl-discrete CFEs}
\label{subsec:cfe_preferable}

Below and in Appendices~\ref{subsec:cfe_app_higher_improv} and \ref{subsec:cfe_app_fair_personalize}, we provide empirical evidence to show that, compared to low-level CFEs, both hl-continuous and hl-discrete CFEs involve fewer actions but lead to more improvement involving more modified features, are easier to personalize, and simplify the design and interrogation of CFE generators for fairness issues.
\begin{figure}[t!]
\centering
    \begin{subfigure}[b]{0.62\textwidth}  
            \includegraphics[width=\textwidth]{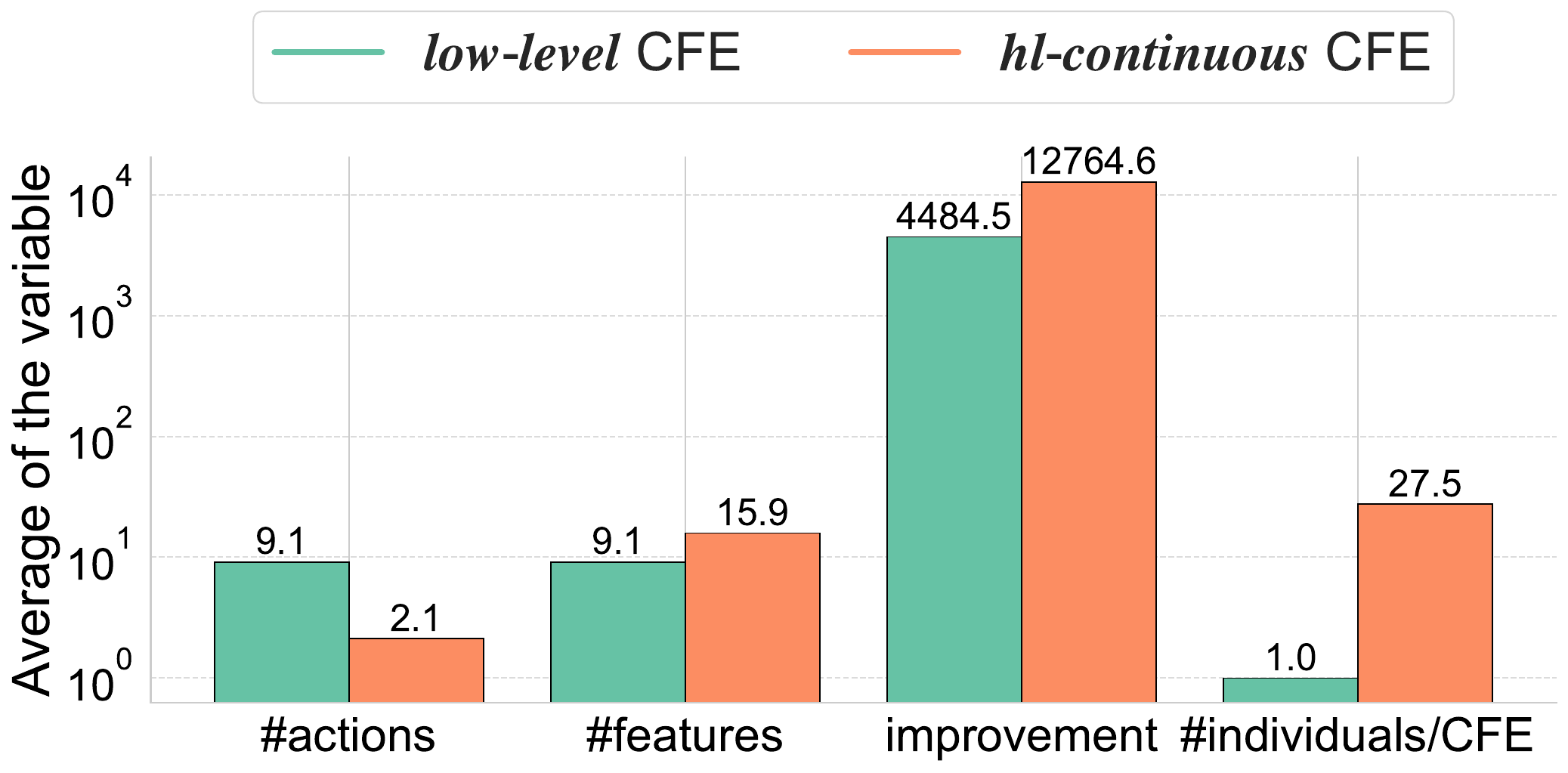}
            \caption{Average of the variables}
            \label{fig:all_whr_calprice}
    \end{subfigure}%
    ~~
    \begin{subfigure}[b]{0.33\textwidth}
            \centering
            \includegraphics[width=0.8\textwidth]{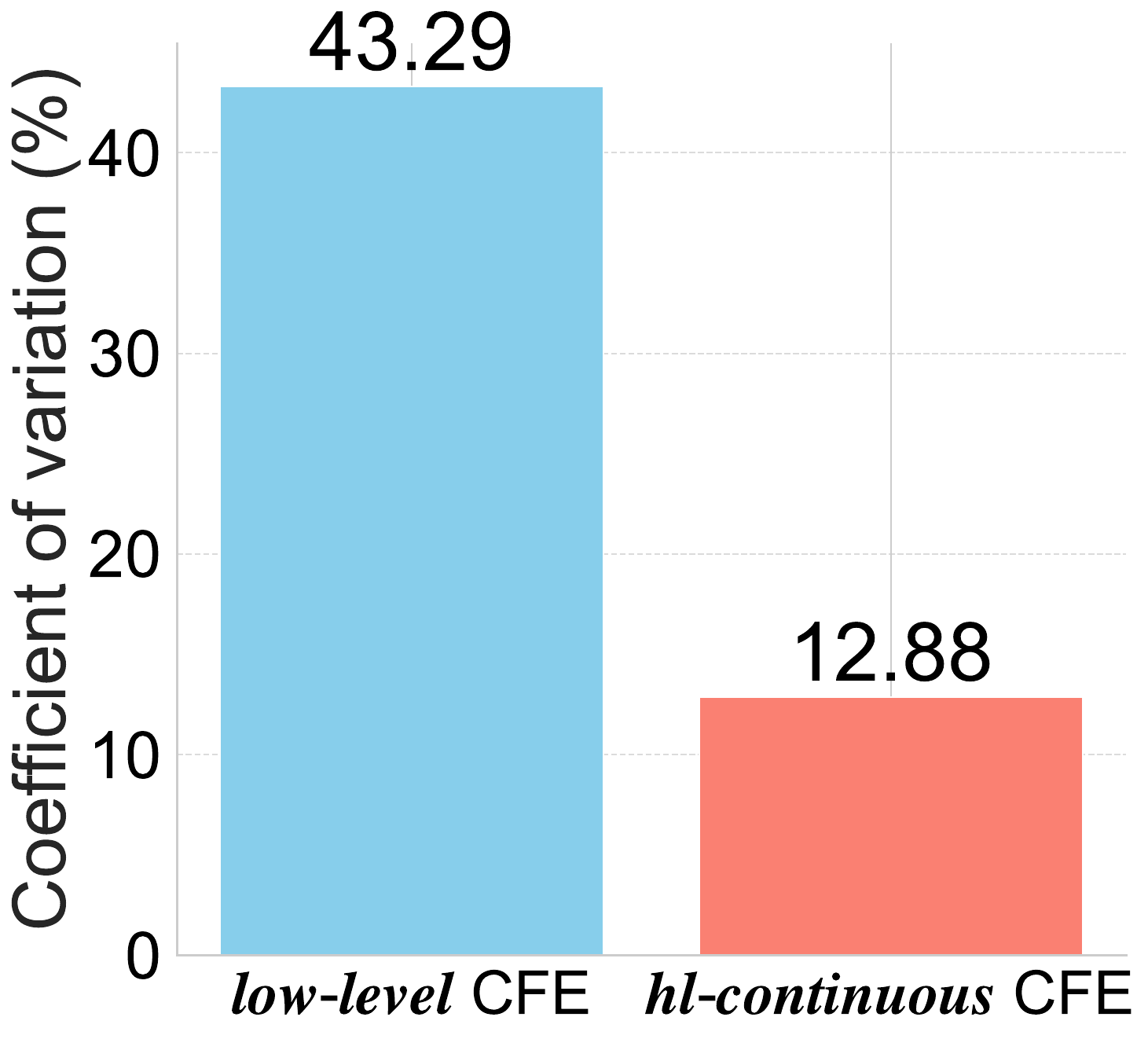}
            \caption{Coefficient of variation in number of modified features across intersectional (race, age, gender) groups}
            \label{fig:whr_var_feats_main}
    \end{subfigure}
   \caption[Comparison of outcomes of taking low-level CFEs vs. high-level CFEs]{On the WHR dataset, \subref{fig:all_whr_calprice} compared to low-level CFEs, to take hl-continuous CFEs require fewer actions but modify a significant number of features and result in higher improvement. On average, each hl-continuous CFE is optimal for multiple agents, with a high frequency of \(27.5\) per CFE, unlike the low-level CFEs with  \(1.0\). Additionally, as shown in \subref{fig:whr_var_feats_main}, there is greater variability in the number of features modified across sensitive groups when using low-level CFEs, suggesting lower fairness than with hl-continuous CFEs. For more details on \subref{fig:all_whr_calprice}, see Appendix Figures~\ref{fig:ap_avg_prox_feats_acts}, \ref{fig:ap_change_feats_improv}, and \ref{fig:ap_change_feats_improv2}; for \subref{fig:whr_var_feats_main}, see Appendix Figures~\ref{fig:whr_real_feats} and \ref{fig:whr_var_feats}.}
    \label{fig:all_whr-calprice_and_fair}
\end{figure}

\paragraph{Fewer actions but higher improvement and more modified features.}
Our results indicate that hl-continuous and hl-discrete CFEs require fewer actions while yielding higher improvements and modifying more features than low-level CFEs. In contrast, low-level CFEs involve more actions but result in lower improvements despite modifying a high number of features. For instance, while on average, on the WHR dataset, the hl-continuous CFEs require only \(2\) actions yet achieve a significantly higher improvement (\(12,765\)), the low-level CFEs involve \(9\) actions but yield a much lower improvement  (\(4,484.5\))  (see Figure~\ref{fig:all_whr_calprice}).

In low-level CFE generation, sparsity (small number of modified features) and proximity (new agent state after taking the CFE close to the initial state) are often a primary goal due to actionable insights being part of the feature space \citep{Ustun19,Verma2020CounterfactualEF}.  We observe a perfect positive correlation between the number of modified features and actions taken (Kendall's \(\tau = 1.0\), \(\textit{p-value} = 0.0\)) and a positive correlation between actions taken and improvement achieved in low-level CFEs (Kendall \(\tau = 0.368, \textit{p-value}=5.41e\)-\(227\)). However, for hl-continuous CFEs, the correlation between the number of modified features and actions taken is positive but weaker (Kendall's \(\tau = 0.722\), \(\textit{p-value} = 0.0\)) and there is almost no relationship between the number of actions taken and improvement achieved (Kendall \(\tau = 0.0625, \textit{p-value}=3.21e\)-\(06\)). Despite requiring fewer actions (\(2\)), hl-continuous CFEs modify significantly more features (\(16\)) compared to low-level CFEs, which involve \(9\) actions and modify \(9\) features (Figures~\ref{fig:main_figure1} and \ref{fig:all_whr_calprice}).

Consequently, unlike taking low-level CFEs, agents using hl-continuous or hl-discrete CFEs tend to become more ``positive'' or ``qualified'' after taking fewer actions. That is, our proposed form of CFEs require fewer actions but result in greater improvement, broader feature modifications, and lower costs in both interpretation and execution (see Figures~\ref{fig:main_figure1} and \ref{fig:all_whr_calprice}, and  Appendix~\ref{subsec:cfe_app_higher_improv} and Figures~\ref{fig:ap_avg_prox_feats_acts}, \ref{fig:ap_change_feats_improv}, \ref{fig:ap_change_feats_improv2} and \ref{fig:ap_kendal_tau}).

\paragraph{Personalization and fairness.}
Since both hl-discrete actions and hl-continuous actions are predefined and real-world-like, it is easier and more transparent to examine the hl-discrete and hl-continuous CFE generators and generated CFEs for potential fairness issues, and to tailor the CFE generation to agents' needs. For example, our data-driven hl-continuous CFE generators can produce CFEs for agents who place greater importance on monetary costs over caloric costs. 

Moreover, agents across the intersectional sensitive (race, age, and gender) groups (e.g., Hispanic, 21-40, Female)  take a comparable number of actions, modifying a closely similar number of features, achieving comparable improvements, and incurring closely similar costs when using hl-continuous CFEs, as indicated by the low coefficient of variation in Figure~\ref{fig:whr_var_feats_main} and Appendix Figures~\ref{fig:whr-calprice_actions_feats_proximity} and \ref{fig:whr-calprice_cost}. In contrast, taking low-level CFEs results in  higher variability across the groups in all these variables (see Figure~\ref{fig:whr_var_feats_main} and Appendices~\ref{sec:cfe_app_var_costs} and \ref{sec:cfe_app_fa_ar_da_comp}).
Thus, high-level CFEs yield fairer outcomes than low-level CFEs.

Lastly, our data-driven hl-discrete CFE generators effectively generate CFEs for all agents, regardless of action restrictions or feature satisfiability variations. This strong performance holds even without explicit knowledge of these variations (see Appendices~\ref{sec:cfe_app_var_thresh_accuracy} and \ref{sec:cfe_app_var_actions_accuracy}).

\subsection{The Data-Driven CFE Generators are Accurate and Resource-efficient}
\label{subsec:cfe_accr_all_cfegen}
\begin{table}[b!]
    \footnotesize
    \renewcommand{\arraystretch}{1.1} 
    \setlength{\tabcolsep}{3pt} 
    \begin{center}
        \begin{minipage}{\textwidth} 
        \begin{subtable}{0.42\textwidth}
            \centering
            \begin{tabular}{lcccc}
                & \multicolumn{3}{c}{\textbf{Accuracy of CFE generators}}\\
                \cmidrule(lr){2-4}
                & hl-continuous & hl-discrete & hl-id & \\
                \midrule
                BMI         & \(0.92\pm0.0053\)    &                        & \(0.94\pm0.0045\)\\
                WHR         & \(0.92\pm0.0176\)    &                        & \(0.97\pm0.0107\)\\
                BRFSS       &                      & \(0.98\pm0.0102\)      & \(0.99\pm0.0050\)\\
                \(20\)-dim  &                      & \(0.94\pm0.0042\)      & \(0.99\pm0.0014\)\\
                \bottomrule
            \end{tabular}
        \end{subtable}
        \hspace{1.1cm} 
        \begin{subtable}{0.42\textwidth}
            \begin{tabular}{llccc}
                & \multicolumn{3}{c}{\textbf{Effect of frequency of CFEs}} \\
                \cmidrule(lr){2-4}
                & $\texttt{all}$ & \texttt{>10} & \texttt{>40}\\
                \midrule
                \(20\)-dim          & \(0.84\pm0.0060\)    & \(0.89\pm0.0052\)    & \(0.94\pm0.0042\)\\
                \(20\)-dim\(\star\) & \(0.97\pm0.0028\)    & \(0.98\pm0.0021\)    & \(0.99\pm0.0014\)\\
                BMI                 & \(0.90\pm0.0057\)    & \(0.91\pm0.0055\)    & \(0.92\pm0.0053\)\\
                BRFSS               & \(0.70\pm0.0182\)    & \(0.86\pm0.0158\)    & \(0.98\pm0.0102\)\\
                \bottomrule
            \end{tabular}
        \end{subtable}
    \end{minipage}
    \end{center}

     \caption[Accuracy of the data-driven CFE generators]{(\textbf{left}) The accuracy of the hl-continuous, hl-discrete, and hl-id data-driven CFE generators on the testing set agents for \texttt{>40}, BMI, BRFSS, WHR, and fully-synthetic (\(20\)-dim): \(20\)-dimensional, datasets. (\textbf{right}) The data-driven CFE generators' accuracy decreases with a decrease in the frequency of CFEs (number of agents for whom a given CFE is optimal) in the agent–CFE training set, regardless of the dataset type. Specifically, training and testing on the (\(20\)-dim): the \(20\)-dimensional agent–hl-discrete CFE dataset, (\(20\)-dim)\(^{\star}\): the \(20\)-dimensional agent–hl-id dataset, (BMI): the BMI agent–hl-continuous CFE dataset, and (BRFSS): the BRFSS agent–hl-discrete CFE dataset all show this trend. The data-driven CFE generators are accurate (\textbf{left}) and their accuracy improves as the frequency of CFEs increases, with those trained highest CFE frequency dataset (\texttt{>40}) performing best (\textbf{right}).}\label{tab:accr_freq_main}
\end{table}
Our results demonstrate that the proposed data-driven CFE generators, operating under various information access constraints, such as no query access to the classifier or, in the case of the data-driven hl-id CFE generator, without knowledge of the cost and impact of actions on agent states; are scalable and accurately and efficiently produce CFEs, without requiring re-optimization of the generator (see Table~\ref{tab:accr_freq_main}(left) and Appendices~\ref{subsec:cfe_app_accurate_approximate}, \ref{sec:cfe_app_scalable_interpretable} and \ref{sec:cfe_app_var_info_accuracy}). 

In contrast to the overly specific low-level CFEs, which are generally unique to each agent, hl-continuous and hl-discrete CFEs are often optimal for a broad range of agents (refer to Figures~\ref{fig:main_figure1} and \ref{fig:all_whr_calprice} and Appendix Figure~\ref{fig:ap_scalability}). 
The removal of the need for re-optimization for each new agent, combined with the general applicability of the actions to agents, enhances the scalability of our proposed CFE generators compared to low-level generators.
Additionally, because the actions in the hl-continuous and hl-discrete CFEs are both general and predefined, they are more transparent and easier to interpret (see Figure~\ref{fig:main_figure1}), making them cheaper and more desirable than the overly specific and unique low-level CFEs.

However, our results show that the accuracy of the proposed data-driven generators declines with the low frequency of CFEs (see Table~\ref{tab:accr_freq_main} (right)) and the generalizability of CFE generation decreases with an increase in the number of actionable features. We observed that this is due to the growing uniqueness of CFEs to agents (see Table~\ref{tab:accr_freq_main} (right)) and Section~\ref{subsec:cfe_potential_challenges}). 
Data augmentation mitigates the negative effects of low CFE frequency. For instance, on the \texttt{all} \(20\)-dimensional dataset, data augmentation improves accuracy from \(0.969\) to \(0.982\).

Lastly, the performance of data-driven CFE generators improves as model complexity increases. For example, on a discrete agent–hl-id dataset, the neural network model outperforms the Hamming distance method (see Appendix Figure~\ref{fig:gens_comp}). Future research could explore more advanced data-driven models for CFE generation and techniques like federated learning to enable CFE generation under limited access to agent–CFE data and privacy constraints.

\section{Limitations and Ethical Considerations}
\label{sec:cfe_limitations}

The decision-maker must have access to data on instances of agents and their corresponding optimal CFEs to train the proposed data-driven CFE generators.  
Although this level of access mitigates some information access challenges, such as needing at least query access to the classifier and representative prediction training data or having an exhaustive list of actions and the associated costs, obtaining a historical agent–CFE dataset may still pose significant challenges. 
Future research could investigate techniques like federated learning and secure multi-party computation to facilitate collaborative training of robust CFE generators under varied privacy and data access constraints. 

While our proposed data-driven CFE generators are agnostic to the underlying classification model, the proposed single-agent hl-continuous CFE generators rely on linear classifiers.  Although the linear models may not always be optimal, they can outperform non-linear models in some contexts \citep{Wainer2016}. When non-linear classifiers are preferred, one practical alternative is to approximate them by linear models (e.g., \citep{Bshouty2012, li2015minmaxkernels, shalizi2020lecture8, Liu2021}), enabling CFE generation with the proposed single-agent CFE generator.  Extending the generator to produce exact CFEs for non-linear models in a scalable and computationally efficient manner remains a challenging but promising avenue for future research.

Additionally, our formulations of hl-continuous and hl-discrete CFEs restrict them to being defined as a set of actions.  More generally, one could consider settings where the order of actions matters, such as where a CFE corresponds to an optimal policy for an agent in a deterministic Markov decision process (MDP).  Further, one could consider actions whose effects are stochastic, and a CFE then corresponds to an optimal policy for the agent in a general MDP.

Since the proposed approaches to data-driven CFE generation are closely related to data-driven algorithm design, ethical concerns related to data-driven algorithms, e.g., potentially propagating and exacerbating biases in historical agent–CFE data and the potential for flawed resource allocation, might apply to our proposed CFE generators. 
Future research should investigate these ethical implications in greater depth.

Although our experiments primarily use healthcare datasets, our data-driven CFE  generation approach generalizes to a broad spectrum of real-world scenarios, such as college admissions, loan applications, judicial systems, and other settings. 
Future works could expand our setup to other data settings and informational access challenges. 
Lastly, we caution readers that the experimentally generated CFEs from our empirical analyses are intended solely for illustrative purposes, and readers should not use them for self-treatment.

\section{Related Work}
\label{sec:cfe_relatedworks}

The proposed single-agent hl-continuous and hl-discrete CFE generation approaches are in principle, similar to search-based optimization CFE generation frameworks \citep{Ramakrishnan19}, single-agent ILP recourse generation approaches \citep{Cui15, Gupta2019EqualizingRA, Ustun19}, and CFE generation methods based on logic and answer-set programming  \citep{Bertossi2020, Liu2023, silva2023}. 
However, unlike these approaches, ours uses predefined real-world-like actions (see Figure~\ref{fig:main_figure1}), resulting in CFEs that involve fewer actions but modify more features and lead to more improvement.

Although most low-level CFE generators operate on a single-agent basis  \citep{Karimi22, Verma2020CounterfactualEF}, recent studies \citep{Pedapati2020, Rawal2020, pmlr-v151-kanamori22a, Ley2023, Carrizosa24}  have introduced approaches that can produce CFEs for multiple agents.
Most closely related to our work is the approach by \citet{pmlr-v151-kanamori22a}, which learns a decision-tree-based global CFE generator that, once learned, can generate CFEs for multiple agents.  
However, unlike \citet{pmlr-v151-kanamori22a}, our data-driven CFE generators generate CFEs with real-world-like actions. 
Moreover, solving the mixed-integer linear programs and the overall Counterfactual Explanation Tree can be computationally prohibitive for scenarios with large action spaces, making our supervised learning approach a more scalable and efficient alternative.

Unlike low-level CFE generators that require, at a minimum, query access to the classifier and knowledge of the cost and impact of each action on state features \citep{Shavit19, Pedapati2020, Rawal2020, Naumann21, Verma22, pmlr-v151-kanamori22a, Toni23, Ley2023, Carrizosa24}, without explicit access to this information, our data-driven CFE generators leverage access to agents and their optimal CFEs to generate CFEs described by real-world-like actions and costs.

While in some ways, the proposed data-driven CFE generators are similar to reinforcement learning-based CFE generation tools \citep{Shavit19, Naumann21, Toni23}, our proposed approach offers 
a more resource-efficient and exact solution alternative to the often high computational and approximate solutions.
Notably, our approach is closest to that of \citet{Verma22}. 
While our method is akin to learning an optimal policy in a large but deterministic family of Markov decision processes (MDPs), \citet{Verma22} focuses on learning optimal policies within smaller, stochastic MDP settings.

Finally, our work also relates to data-driven algorithm design \citep{Gupta16, BalcanDW18, Balcan20}, where models learn from training data instances to generalize to the testing data.  We introduce novel data-driven CFE generators that address the question: \emph{Can we, by learning from training agent–CFE data (i.e., instances of agents and their optimal CFEs), develop a CFE generator that quickly provides optimal CFEs for new agents?}  Our proposed approach excels in generating CFEs for new agents, is computationally efficient and scalable, and functions effectively under varied informational access settings.

\section{Conclusion}
\label{sec:cfe_conclusion}

In this work, we propose three forms of recourse where actionable insights align closely with real-world actions and investigate settings where CFEs can be generated by analyzing the similarities between negatively classified agents using data-driven approaches.
Our findings show that compared to low-level CFEs, both hl-continuous and hl-discrete CFEs require fewer actions, modify more features, and result in higher improvements.
Additionally, the CFEs are fairer across sensitive groups and are easier to examine, compare, and personalize than low-level CFEs.
Lastly, we empirically show that the proposed data-driven CFE generators are accurate, resource-efficient, and perform effectively under various information access constraints, including limited or restricted access to classifier parameters and training data.

\chapter{Selectively Revealing Positive and Negative Role Models to Help People Make Good Decisions}
\label{chap:revealrm}
\section{Introduction}
\label{sec:intro}
Consider a government agency that wants to provide a booklet with information about how to correctly file your taxes.  The booklet will not be able to cover every possible question that any taxpayer may have, but it should cover common tax situations that arise.  For these situations, the booklet may include positive examples showing what to do and/or negative examples showing what not to do.  For example, IRS Publication 970\footnote{``Tax Benefits for Education'' \url{https://www.irs.gov/pub/irs-pdf/p970.pdf}} gives a large number of examples showing what to report for various types of fellowships, scholarships, and grants, as well as warnings about common mistakes.  Given a limited budget of how many examples can be included in the booklet, the government would like to include those that will be most helpful.  In this case, it is natural to model a taxpayer facing a given scenario as following a positive example if a relevant one is present in the booklet, avoiding an incorrect action if it is described as incorrect in the booklet, and otherwise choosing randomly from among the reasonable options they have.

This work formalizes the setting as an unweighted bipartite graph with taxpayers (agents) on the left and strategies (targets) on the right. Edges connect taxpayers to strategies within their collection. Each strategy is classified as \textit{positive} if it reflects desirable decision-making patterns and as \textit{negative} if it reflects those that the agents should avoid. Initially, agents do not know which of their adjacent strategies are positive or negative. In this case, the booklet is the subset of targets whose labels are revealed by the social planner.

Our goal is to study how the government agency (i.e., a social planner) with a limited budget can help agents identify and emulate or imitate positive targets, thereby improving their decision quality, by revealing the labels of a limited number of targets. Revealing positive targets causes agents to emulate their behavior, while revealing negative ones indicates decisions that should be avoided, but does not suggest which choices are good. Consequently, in the standard model, the planner's objective is to reveal labels of a budgeted subset of targets to maximize \textit{social welfare}, defined as the total probability that agents emulate adjacent positive targets, given the revealed subset.  We extend the standard model to consider a setting where the planner identifies agents most likely to emulate negative targets and uses the intervention budget to connect them directly to positive ones, ensuring these agents emulate a positive target. In the tax filing setting, this could involve identifying taxpayers prone to bad strategies and pairing them with positive ones to follow.

\paragraph{Contributions.}
While motivated by a tax filing example, the proposed models are more broadly applicable to other domains in which agents rely on role models, options or exemplars in their neighborhood to make decisions. Appendix~\ref{sec:revealrm_app_prac-examples} provides additional examples. Below are our main contributions.
\begin{enumerate}
    \item In the standard model, we show that when the social planner reveals negative targets, the social welfare function remains monotone, but may become supermodular (Section~\ref{subsec:revealrm_monosub}). Consequently, the approximation guarantees of the classic polynomial-time, budget constrained greedy algorithm can deteriorate to as low as $\frac{2}{\sqrt{n}+2}$, where $n$ denotes the number of agents, when both positive and negative targets can be revealed. To address this limitation, we introduce a proxy welfare function that remains submodular even when revealed targets include negative ones. When all agents have at most $c$ negative target neighbors, this proxy achieves a constant-factor approximation to the true welfare (gain) (Section~\ref{subsec:revealrm_approxs}).

    \item In a setting where agents are divided into $w$ groups (for constant $w$), we show that running the classic greedy algorithm on each group $a$ with budget $\lceil K/w \rceil$ guarantees that $c$-bounded agents in $a$ achieve a welfare gain of $\Omega(\mathrm{OPT}_{a}^{K})$, i.e., within a constant factor of the optimum achievable if the full budget $K$ were allocated to that group. When agents are not $c$-bounded, we show that this guarantee may fail to hold (Section~\ref{sec:revealrm_fairness}). 

    \item To assess the potential for stronger algorithmic results and demonstrate that the existing guarantees are essentially tight, we establish NP-Hardness for the social welfare maximization problem when the planner can reveal only positive targets, only negative targets, or both (Section~\ref{sec:revealrm_np-hardness}). Then, in Section~\ref{sec:revealrm_learn}, we study a learning-theoretic variant of the standard model in which the left-hand side of the bipartite graph is replaced by a probability distribution $D$ over agents. We establish a sample-complexity guarantee for the setting where the social planner observes the neighborhood of each agent sampled from a distribution $D$.
    
    \item In cases where agents either have poor neighborhoods or are unaware of the adjacent positive targets, the standard model fails to help these agents make good decisions. To address these limitations, Section~\ref{sec:revealrm_extensions} introduces two intervention mechanisms: connecting poorly positioned agents to positive targets before or after revealing a set of atmost $K$ welfare-maximizing targets (Section~\ref{sec:revealrm_intervention_model}), and increasing the visibility of positive targets so that neighboring agents can observe them (Section~\ref{sec:revealrm_coveragemodel}).

    \item Finally, we conduct extensive semi-synthetic experiments using bipartite graphs generated from four real-world datasets: Adult, Student Performance (Mathematics and Portuguese), and Garment Workers Productivity. We empirically evaluate and compare several greedy strategies under the standard model, examine gains from targeted intervention, and assess the performance of Algorithm~\ref{alg:greedy_lbreveal} in the learning setting\footnote{Our code is publicly available at \href{https://github.com/knaggita/InformationDisclosure}{https://github.com/knaggita/InformationDisclosure}.} (Section~\ref{sec:revealrm_experiments}).
\end{enumerate}

\subsection{Related Work}
\label{sec:revealrm_related-work}
\paragraph{Personalized recourse.}
The growing reliance on machine learning (ML) models to make high-stakes decisions (e.g., for hiring and loan approvals) raises urgent questions about transparency and the provision of guidance that enables individuals to improve their outcomes. This concern motivated extensive work on personalized recourse, typically operationalized through single-agent~\citep{Karimi22,Verma2020CounterfactualEF} or multi-agent~\citep{pmlr-v151-kanamori22a,learningactionablecounterfactualexplanations,Ley2023,Carrizosa24,Pedapati2020} frameworks.
These approaches assume access to the agents' initial feature states and action spaces and identify the minimum cost set of actions that lead to a desirable prediction. In contrast, the social planner in our setting lacks such information and instead releases limited signals, namely whether adjacent role models are positive or negative influences, enabling agents to improve outcomes by emulating adjacent positive role models.

\paragraph{Strategic learning.}
Our line of inquiry is closely related to strategic learning under manipulation graphs~\citep{lechner2022learning, Zhang_Conitzer_2021, SabaAKH24, Attias2025PACLW, ahmadi2022classificationstrategicagentsgame, cohen2024learnability}, where each agent's reaction set is shaped by its local neighborhood. 
The key distinction is that the social planner in our setting does not control the labeling function and is limited to revealing only a small subset of labels, a setup analogous to strategic learning with restricted label queries~\citep{balcan2025active}.
Our work is also related to research on strategic learning under imitative strategic behavior and partial information release, in which agents strategically modify their features by observing or imitating the strategies of social targets~\citep{HodaHVN19,RaabRLY21,ZhangXKM22,TianXZM24} or of historical feature-prediction pairs~\citep{GaneshVII21,YahavCZ21,Cohen2024}. Unlike these studies, we do not assume agents know which targets to emulate; rather, we study how selectively revealing labels for a subset of targets can guide agents toward better decisions. 
Lastly, our work is tangentially related to prior research that leverages counterfactual explanations to help agents respond optimally in strategic classification settings~\citep{TsirtsisSG20,TianXPX25}; however, unlike these works, ours is a multi-agent approach that steers agents toward positive targets to emulate rather than recommending feature-level actions for individual agents.

\paragraph{Influence maximization.}
Our objective is analogous to influence maximization, which aims to iteratively identify the most influential nodes to shape agents' behaviors~\citep{RichardsonMD02,KempeDKJ03,KamarthiHVP20}. Traditional methods typically assume a monotone, submodular objective and apply greedy strategies to maximize influence across multi-step diffusion processes~\citep{du2017scalable,LiYFJ18,LiYGH23}. The importance of submodularity extends beyond influence maximization; there is a rich literature on learning submodular functions from data~\citep{balcan2018submodular,balcan2011learning}. Although submodularity often enables efficient optimization and learning guarantees, recursive diffusion-based methods remain computationally demanding. By contrast, we study one-step models on bipartite graphs, where the objective may remain monotone but become supermodular. In this setting, greedy strategies can perform poorly, but our approach avoids the complexity of multi-step diffusion.

\section{Problem Formulation}
\label{sec:revealrm_model_prelim}
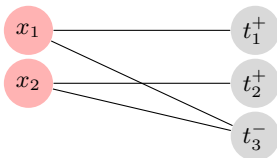
\begin{figure}[ht!]
    \centering
    \begin{minipage}[t]{0.33\textwidth}
        \vspace{0.5cm} 
        \centering
        \begin{tikzpicture}[
            font=\footnotesize, 
            agent/.style={
                circle,
                fill=#1!30, 
                inner sep=1.5pt, 
                minimum size=0.65cm, 
                text centered
            },
            posAgent/.style={agent={blue}},
            negAgent/.style={agent={red}},
            goodTarget/.style={agent={green}},
            badTarget/.style={agent={orange}},
            unknownTarget/.style={agent={gray}},
            edgeLabel/.style={font=\footnotesize, midway, fill=white, inner sep=0.1cm}
        ]
        
            \node[negAgent] (a1) at (0,1.4)  {\(x_{1}\)};
            \node[negAgent] (a2) at (0,0.7)  {\(x_{2}\)};
            \node[unknownTarget] (t1) at (3,1.4) {\(t_{1}^{+}\)};
            \node[unknownTarget] (t2) at (3,0.7) {\(t_{2}^{+}\)};
            \node[unknownTarget] (t3) at (3,0)   {\(t_{3}^{-}\)};
            
            \draw (a1) -- (t1);
            \draw (a1) -- (t3);
            \draw (a2) -- (t2);
            \draw (a2) -- (t3);
        
        \end{tikzpicture}
    \end{minipage}%
    \hspace{0.03\textwidth}
    \vspace{0cm} 
    \begin{minipage}[t]{0.63\textwidth}
        \caption[Structure of the considered social network]{An unweighted bipartite graph in which the LHS nodes \(\{x_1, x_2\}\) represent agents, the RHS nodes \(\{t_1, t_2, t_3\}\) represent targets, and edges connect agents to targets they can emulate. Initially, agents do not know whether a target is positive or negative.}
        \label{fig:intro_bipartite_graph}
    \end{minipage}
\end{figure}

We model the setting as an unweighted bipartite graph $\graph = (\Xs \cup \Ts, E)$ (Figure~\ref{fig:intro_bipartite_graph}), where $\Xs$ is the set of $n$ agents (left-hand nodes), $\Ts$ is the set of $m$ targets (right-hand nodes), and $E$ contains edges between each agent and the targets it can emulate. Let $f: \Ts \to \{-1, +1\}$ be the true target labeling function only known to the social planner. 
For an agent $x \in \Xs$, let $N(x) = \{t \in \Ts : (x,t) \in E\}$ denote its neighborhood, and 
$\delta_x^{+} = |\{t \in N(x): f(t)=+1\}|$ and $\delta_x^{-} = |\{t \in N(x): f(t)=-1\}|$ be the number of positive and negative neighbors, respectively. Of the $m = \mpos + \mneg$ targets, $\mpos$ are positive, representing desirable behaviors agents should emulate, and $\mneg$ are negative, representing behaviors agents should avoid. We next describe how each agent chooses a target to emulate from their neighborhood and how the social planner selects a subset of targets whose information (i.e., labels) is revealed.

\paragraph{Agents' choice of who to emulate.}
Agents do not observe targets' labels and rely only on those revealed in the set $S \subseteq \Ts$. For an agent $x$, let $P_x^{+}(S) = \{t \in N(x) \cap S : f(t)=+1\}$ denote the adjacent targets revealed as positive, and $P_x^{-}(S) = \{t \in N(x) \cap S : f(t)=-1\}$ denote those revealed as negative. If no adjacent targets are revealed, meaning either $S=\emptyset$ or $N(x)\cap S=\emptyset$, the agent emulates a target selected uniformly at random from $N(x)$. Otherwise, the agent assigns zero probability to adjacent targets revealed as negative. If any adjacent targets are revealed as positive, it selects uniformly among them; if none are positive, it selects uniformly among the adjacent unlabeled targets. If all targets in $N(x)$ are revealed as negative, the probability that agent $x$ emulates a positive target is $0$. Accordingly, we define the total probability mass that agent $x$ assigns to positively labeled target neighbors given the revealed set $S$ as follows:
\[
    Q^{S}(x) = 
    \begin{cases}
        1 & \text{if } |P^{+}_{x}(S)|>0,\\
        \dfrac{\delta^{+}_{x}}{\delta^{+}_{x} + 
        (\delta^{-}_{x} - |P^{-}_{x}(S)|)} & \text{if } |P^{+}_{x}(S)|=0 \text{ and } 
        N(x)>0,\\
        0 & \text{otherwise.}
    \end{cases}
\]

\paragraph{The $\splan$ information reveal.}
Assume the $\splan$ has full knowledge of the graph, including all agents and the finite set of targets, and observes each target's true label through the labeling function $f: \Ts \to \{+1, -1\}$. Further, the $\splan$ is aware of the aforementioned process by which agents choose which target in their neighborhood to emulate.  Given a target reveal budget $K \in \mathbb{N}$, the objective of the $\splan$ is to reveal the labels of the subset of targets\footnote{When clear from context, ``reveal a subset of targets'' denotes ``reveals the labels of a subset of targets''} $S \subseteq \Ts$  with $|S| \le K$ that maximizes the probability that the agents choose to emulate positively labeled targets. That is, the social planner aims to find
\begin{align*}
    S^\star = \argmax_{\substack{S:\, |S|\le K}}\;F(S).
\end{align*}
where the social welfare function $F$ is defined as:
\begin{equation}
\label{eq:social_welfare}
    F(S) = \sum_{x \in \Xs} Q^S(x)
\end{equation}
The gain in social welfare from revealing $S$ is defined as the difference between the social welfare under $S$ and the social welfare under the empty set:
\begin{equation}
\label{eq:gain_social_welfare}
    G(S) = F(S) - F(\emptyset)
\end{equation}
The marginal gain of revealing a target $t \in \Ts \setminus S$ given a revealed set $S \subseteq \Ts$ is defined as:
\begin{equation}
\label{eq:marginalgain}
    \D_{t}(S) = F(S \cup \{t\}) - F(S)
\end{equation}

\subsection{Monotonicity and Submodularity of the Social Welfare Function} 
\label{subsec:revealrm_monosub}

In this section, we first demonstrate that the social welfare function is a monotonically increasing function, and then explore the conditions under which it is submodular.
\begin{proposition}
    The social welfare function is a monotonically increasing function. That is, for any set of revealed targets \(A \subseteq \Ts\), $F(A \cup \{t\}) \geq F(A)$ for all $t \in \Ts \setminus A$.
\label{prop:monotone}
\end{proposition}

Intuitively, revealing an additional target cannot reduce an agent's probability of selecting a positive target from its neighborhood. The proof is provided in Appendix~\ref{sec:revealrm_app-prelimmonotonicty} for completeness.

We now analyze the submodularity of the social welfare function when the $\splan$ can reveal only positive targets and when revealed targets include negative ones. 
\begin{definition}[Submodularity]
\label{def:submodularity}
    A function $F: 2^{\Ts} \to \mathbb{R}_{\geq 0}$ is submodular if the marginal gain (Eqn.~\ref{eq:marginalgain}) of adding a revealed target \(t\) to a smaller revealed target set \(A \subseteq B\) is at least as large as the marginal gain of adding it to a larger revealed target set \(B\). That is, for every $A, B \subseteq \Ts$ where $A \subseteq B$, and every $t \in \Ts \setminus B$, we have
    \begin{align*}
        F(A \cup \{t\}) - F(A) \geq F(B \cup \{t\}) - F(B).
    \end{align*}
\end{definition}

Proposition~\ref{prop:sw_posonly} establishes that when the $\splan$ is restricted to revealing only positive targets, the social welfare function $F(S)$ is monotone and submodular.
\begin{proposition}
\label{prop:sw_posonly}
    When the $\splan$ is restricted to only revealing positive targets, then 
    $F: 2^{\Ts^+} \to \mathbb{R}_{\geq 0}$ is submodular. 
\end{proposition}
The function $F$ is monotone (Proposition~\ref{prop:monotone}), and the marginal gain from revealing a positive target is at least as large when the current revealed target set is small as when it is large and more such targets are already known.  Formal proof in Appendix~\ref{sec:revealrm_app-prelimsubmod-posonly}.

Proposition~\ref{prop:sw_negonly} shows that when the revealed targets include negative ones, $F$ remains monotone, but might not necessarily be submodular.
\begin{proposition}
\label{prop:sw_negonly}
    When the targets the $\splan$ reveals include negative ones, then the social welfare function might not necessarily be submodular. 
\end{proposition}
\begin{proof}[Proof sketch]
    Consider an agent adjacent to at least two positive and two negative targets.  Revealing an additional positive target results in $0$ marginal gain, whereas revealing an additional negative target before any adjacent positive target is revealed results in increasing marginal gain because the probability of emulating an adjacent positive target increases with increase in the number of revealed adjacent negative targets. Full proof in Appendix~\ref{sec:revealrm_app-prelimsubmod-negonly}.
\end{proof}

\section{The Standard Model}
\label{sec:revealrm_standardmodel}
We begin with a simple and intuitive greedy algorithm that selects up to $K$ targets to maximize social welfare (Section~\ref{sec:revealrm_standard_greedy}). Appendix~\ref{sec:revealrm_app-classicgreedy} proves polynomial-time complexity.

Its performance, however, depends critically on the structure of the welfare function. Although the function is always monotone, submodularity hinges on the planner’s disclosure policy. When disclosure is restricted to positive targets, submodularity is preserved, and the greedy algorithm achieves the standard $(1 - 1/e)$-approximation guarantee. Once negative targets are allowed, submodularity may fail, and the algorithm can perform arbitrarily poorly. 
To restore the guarantee, we introduce a proxy welfare function that remains submodular even when negative targets can be revealed. Additionally, when all agents have atmost $c$ negative target neighbors, we show that this proxy achieves a constant-factor approximation to the true optimal welfare gain (Section~\ref{subsec:revealrm_approxs}).

Beyond the single-group case, we investigate the algorithm's performance on multiple agent subpopulations, formalizing a fairness guarantee that ensures each group $a$ attains welfare (gain) proportional to the maximum achievable under budget $K$ if $a$ is catered to in isolation (Section~\ref{sec:revealrm_fairness}).

Finally, we assess the potential for stronger algorithmic results and demonstrate that the existing guarantees are essentially tight. When only positive targets can be revealed,  no polynomial-time algorithm can achieve a substantially better worst-case guarantee, and when both positive and negative targets can be revealed, there is no approximation solution (Section~\ref{sec:revealrm_np-hardness}). 
See Appendix~\ref{sec:revealrm_app_standard_algos} for missing proofs and alternative greedy strategies.

\subsection{The Greedy Algorithm} 
\label{sec:revealrm_standard_greedy}
The main result of this section is a greedy algorithm that selects up to $K$ targets to maximize social welfare (Algorithm~\ref{alg:greedy_lbreveal}). Proposition~\ref{prop:opt_runtime_greedy} in Appendix~\ref{sec:revealrm_app-classicgreedy} analyzes its complexity.

\begin{algorithm}[t!]
\caption{Greedy Target Reveal}
\label{alg:greedy_lbreveal}
\SetAlgoLined

\KwIn{Graph $\graph = (\Xs \cup \Ts, E)$, candidate targets $\Ts' \subseteq \Ts$, labels $\{f(t)\}_{t \in \Ts'}$, budget $K$, initial set $S'=\emptyset$}
\KwOut{Solution set $S_{\mathrm g} \subseteq \Ts'$ with $|S_{\mathrm g}| \le K$ and social welfare $F(S_{\mathrm g})$}

$S_{\mathrm g} \gets S'$

\While{$|S_{\mathrm g}| \leq K$}{
    $t^\star \gets \arg\max\limits_{t \in \Ts' \setminus S_{\mathrm g}}
    \big( F(S_{\mathrm g} \cup \{t\}) - F(S_{\mathrm g}) \big)$

    \If{$F(S_{\mathrm g} \cup \{t^\star\}) = F(S_{\mathrm g})$}{
        \textbf{break}
    }

    $S_{\mathrm g} \gets S_{\mathrm g} \cup \{t^\star\}$
}

\Return $(S_{\mathrm g}, F(S_{\mathrm g}))$

\end{algorithm}

\paragraph{Overview of Algorithm~\ref{alg:greedy_lbreveal}.}
At each iteration, Algorithm~\ref{alg:greedy_lbreveal} reveals the target \(t^\star \in \Ts \setminus S_{\mathrm{g}}\) that yields the highest marginal gain \((F(S_{\mathrm{g}} \cup \{t^\star\}) - F(S_{\mathrm{g}}))\). The process repeats until no unrevealed target yields a positive marginal gain or when the budget is exhausted.

Unless otherwise stated, the initial revealed target set is empty \(S'=\emptyset\). Algorithm~\ref{alg:greedy_lbreveal} is run with target set \(\Ts'=\Ts^{+}=\{t\in\Ts \mid f(t)=+1\}\) when target reveal is restricted to only positive targets, with \(\Ts'=\Ts^{-}=\{t\in\Ts \mid f(t)=-1\}\) when restricted to negative targets, and with \(\Ts'=\Ts\) when no restriction is imposed.

\subsection{Approximation Guarantees}
\label{subsec:revealrm_approxs}

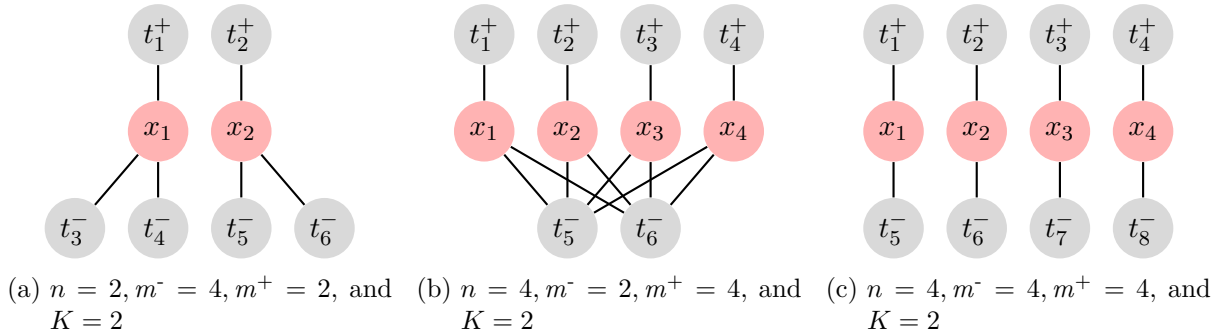
\begin{figure}[ht!]
\centering
    \begin{subfigure}[t]{0.32\textwidth}
    \centering
           \begin{tikzpicture}[
            node distance=1cm and 1cm,
            agent/.style={
                circle,
                fill=#1!30, 
                inner sep=1.5pt, 
                minimum size=0.8cm, 
                text centered
            },
            inputNode/.style={agent={red}},
            posTarget/.style={agent={gray}},
            negTarget/.style={agent={gray}},
            connect/.style={thick},
            scale=0.55
        ]
        
        \foreach \i in {1,2} {
            \node[inputNode] (x\i) at (\i*2,0) {$x_{\i}$};
        }
        \foreach \i in {1,2} {
            \node[posTarget] (tp\i) at (\i*2,2.33) {$t_{\i}^{+}$};
        }
        
        \node[negTarget] (tn3) at (0,-2.33) {$t_{3}^{-}$};
        \node[negTarget] (tn4) at (2,-2.33) {$t_{4}^{-}$};
        \node[negTarget] (tn5) at (4,-2.33) {$t_{5}^{-}$};
        \node[negTarget] (tn6) at (6,-2.33) {$t_{6}^{-}$};
        
        \foreach \i in {1,2} {
            \draw[connect] (x\i) -- (tp\i);
        }

        \draw[connect] (x1) -- (tn3);
        \draw[connect] (x1) -- (tn4);
        \draw[connect] (x2) -- (tn5);
        \draw[connect] (x2) -- (tn6);
    
    \end{tikzpicture}
    \caption{\(n= 2, \mneg = 4, \mpos = 2,\) and $K=2$}
    \label{fig:pos_optimal}
    \end{subfigure}
    \hfill
    \begin{subfigure}[t]{0.32\textwidth}
    \centering
    \begin{tikzpicture}[
            node distance=1cm and 1cm,
            agent/.style={
                circle,
                fill=#1!30, 
                inner sep=1.5pt, 
                minimum size=0.8cm, 
                text centered
            },
            inputNode/.style={agent={red}},
            posTarget/.style={agent={gray}},
            negTarget/.style={agent={gray}},
            connect/.style={thick},
            scale=0.55
        ]
        
        \foreach \i in {1,2,3,4} {
            \node[inputNode] (x\i) at (\i*2,0) {$x_{\i}$};
        }
        \foreach \i in {1,2,3,4} {
            \node[posTarget] (tp\i) at (\i*2,2.33) {$t_{\i}^{+}$};
        }
        
        \node[negTarget] (tn5) at (4,-2.33) {$t_{5}^{-}$};
        \node[negTarget] (tn6) at (6,-2.33) {$t_{6}^{-}$};
        
        \foreach \i in {1,2,3,4} {
            \draw[connect] (x\i) -- (tp\i);
            \draw[connect] (x\i) -- (tn5);
            \draw[connect] (x\i) -- (tn6);
        }
    \end{tikzpicture}
    \caption{\(n= 4, \mneg = 2, \mpos = 4,\) and $K=2$}
    \label{fig:neg_optimal}
    \end{subfigure}
    \hfill
    \begin{subfigure}[t]{0.32\textwidth}
    \centering
    \begin{tikzpicture}[
            node distance=1cm and 1cm,
            agent/.style={
                circle,
                fill=#1!30, 
                inner sep=1.5pt, 
                minimum size=0.8cm, 
                text centered
            },
            inputNode/.style={agent={red}},
            posTarget/.style={agent={gray}},
            negTarget/.style={agent={gray}},
            connect/.style={thick},
            scale=0.55
        ]
        
        \foreach \i in {1,2,3,4} {
            \node[inputNode] (x\i) at (\i*2,0) {$x_{\i}$};
        }
        \foreach \i in {1,2,3,4} {
            \pgfmathtruncatemacro{\j}{\i+4}
            \node[posTarget] (tp\i) at (\i*2,2.33) {$t_{\i}^{+}$};
            \node[negTarget] (tn\j) at (\i*2,-2.33) {$t_{\j}^{-}$};
            
        }
        
        \foreach \i in {1,2,3,4} {
            \pgfmathtruncatemacro{\j}{\i+4}
            \draw[connect] (x\i) -- (tp\i);
            \draw[connect] (x\i) -- (tn\j);
        }
    \end{tikzpicture}
    \caption{\(n= 4, \mneg = 4, \mpos = 4,\) and $K=2$}
    \label{fig:either_optimal}
    \end{subfigure}
    \caption{Even with the same budget, the composition of the optimal target set varies with graph structure.  \label{fig:var_optimal}}
\end{figure}

The composition of the optimal solution set depends on the graph and may include only positive targets (Figure~\ref{fig:pos_optimal}), only negative targets (Figure~\ref{fig:neg_optimal}), or both positive and negative targets (Figure~\ref{fig:either_optimal}).  
Additionally, since the information disclosure policy affects the submodularity of the social welfare function (Section~\ref{subsec:revealrm_monosub}), in this section, we examine the approximation guarantees of the classic greedy algorithm (Algorithm \ref{alg:greedy_lbreveal}) under three disclosure regimes: revealing only positive targets, revealing only negative targets, and unrestricted disclosure.

\subsubsection{The social planner is restricted to revealing only positive targets}

By Proposition~\ref{prop:sw_posonly}, restricting the planner to positive targets makes $F$ monotone and  submodular, implying that Algorithm~\ref{alg:greedy_lbreveal} achieves a $(1 - 1/e)$-approximation (Theorem~\ref{thm:monotonesub}).

\begin{theorem}
\label{thm:monotonesub}
   When the $\splan$ can only reveal positive targets, Algorithm~\ref{alg:greedy_lbreveal} achieves an \((1 - 1/e)\)-approximation for the  \(\displaystyle \max_{\substack{S:\, |S|\le K \\ f(t)=1\ \forall t\in S}}\!\!F(S) \) problem. That is, $F(S_g) \ge (1 - 1/e)\, F(S^\star)$, where $S_g$ is the greedy solution and $S^\star$ is the optimal solution. 
\end{theorem}
This guarantee follows from the classical result of \citet{Nemhauser78}.

Note that when the $\splan$ is restricted to revealing only positive targets, the welfare gain (Eqn.~\ref{eq:gain_social_welfare}) is monotone and submodular because $F(\emptyset)$ is a constant and $F(S)$ is monotone and submodular. Consequently, the approximation guarantee in Theorem~\ref{thm:monotonesub} extends to this setting. That is, $G(S_g) \ge (1 - 1/e)\, G(S^\star)$ where $S_g$ is the greedy solution and $S^\star$ is the optimal solution.

\subsubsection{The social planner can only reveal negative targets}

By Proposition~\ref{prop:sw_negonly}, restricting the social planner to revealing only negative targets preserves monotonicity but not submodularity of $F$. We construct an example where the approximation ratio of Algorithm~\ref{alg:greedy_lbreveal} can be strictly below $3/\sqrt{2n}$, where $n$ is the number of agents (Theorem~\ref{thm:neg-appratio} in Appendix~\ref{sec:revealrm_app_sm_approxs}).

\subsubsection{The social planner can reveal both positive and negative targets}

Assume the social planner may reveal both positive and negative targets. By Proposition~\ref{prop:sw_negonly}, including negative targets in the revealed set preserves monotonicity but not the submodularity of true social welfare function $F$. We construct an example where the approximation ratio of Algorithm~\ref{alg:greedy_lbreveal} can be strictly below 
\(2/(\sqrt{n}+2)\),  where \(n\) is the number of agents (Appendix~Theorem~\ref{thm:negpos-appratio}).

Now assume each agent has at most $c$ negative target neighbors for some constant $c \ge 1$. That is, we assume  $\delta_x^- \le c$ for all $x \in \Xs$. We call such an agent $c$-bounded. Definition~\ref{def:proxy_sw} introduces the proxy social welfare function, with proxy welfare gains defined in Definition~\ref{def:proxywelfare_gain}.
\begin{definition}[Proxy social welfare function]
\label{def:proxy_sw}
\begin{equation}
    F_p(S) = \sum_{x \in \Xs} Q^{S}_{p}(x), \qquad \mathrm{where}
\label{eq:proxy_sw}
\end{equation}
\[
Q^{S}_{p}(x) = 
\begin{cases}
1,
& \mathrm{if} \ |P^{+}_{x}(S)|>0, \\
\dfrac{\delta_x^{+}}{|N(x)|-\mathbf{1}_{\{|P^{-}_{x}(S)|\geq1\}}}
\left(\!1+\dfrac{ \max\{0,|P^{-}_{x}(S)|-1\}}{|N(x)|}\!\right)
& \mathrm{if} \ |P^{+}_{x}(S)|=0 \ \mathrm{and} \ |N(x)|>0, \\
0,
& \mathrm{otherwise.}
\end{cases}
\]
is the proxy total probability mass assigned by agent $x$ to positively labeled targets in its neighborhood given the revealed set $S$. 
In particular, if we reveal an agent's negative neighbor, we only increase $Q^{S}_{p}(x)$ by the amount that revealing the first negative neighbor helped. 
\end{definition}
\begin{definition}[Proxy welfare gain]
\label{def:proxywelfare_gain}
The proxy social welfare gain from revealing $S$ is defined as the difference between the proxy welfare under $S$ and the welfare under the empty set:
\begin{equation}
    G_p(S) = F_p(S) - F(\emptyset)
\label{eq:proxygain_social_welfare}
\end{equation}
\end{definition}

By definition (Eqn.~\ref{eq:social_welfare}~and Defn.~\ref{def:proxywelfare_gain}), the proxy welfare (gain) is less than or equal to the true welfare (gain). What we show is that if agents are $c$-bounded for some constant $c$, then in fact the proxy welfare (gain) is within a constant factor of the true welfare (gain). 
That is, $F_p(S) \leq F(S) \leq cF_p(S)$ and $G_p(S) \leq G(S) \leq cG_p(S)$ (Appendix~Lemma~\ref{lem:proxy}).  Moreover, when the social planner can reveal both positive and negative targets, the proxy social welfare function is submodular, and the proxy is submodular on the gain (Appendix~Lemmas~\ref{lem:proxy-sub} and~\ref{col:proxygain-sub}). Consequently, \textit{proxy-greedy}, defined as running the classical greedy algorithm on the proxy welfare function rather than the true welfare, achieves a constant-factor approximation to the true social welfare gain (Theorem~\ref{thm:negpos-appratio2}). 
Note that for any solution set, the proxy achieves $(1/c)$-factor approximation to the true optimal welfare (Appendix~Remark~\ref{rem:1/capprox}).

\begin{theorem}
\label{thm:negpos-appratio2}
    When revealed targets may include negative ones, and all agents are $c$-bounded for some constant $c \geq 1$, the proxy-greedy algorithm achieves a constant-factor approximation to the true social welfare gain. That is, if $S_p$ denotes the set revealed by proxy-greedy and $S^\star$ an optimal set under the true social welfare function, then $G(S_p) \geq \frac{1-1/e}{c} G(S^\star)$.
\end{theorem}
\begin{proof}
    Proof in Appendix~\ref{subsec:revealrm_app-posneg-constapprox}
\end{proof}

\subsection{Simultaneous Approximate Optimality}
\label{sec:revealrm_fairness}
In this section, we establish fairness guarantees by showing that there exists a solution set that is simultaneously approximately optimal for all groups, and also examine how the inclusion of negative targets in the revealed target set affects this per-group approximation guarantee.
That is, using Algorithm~\ref{alg:greedy_lbreveal}, along with the true and proxy social welfare functions and the $c$-boundedness property, we assess the fairness of the greedy algorithm in a grouped setting.

Suppose agents are divided into $w$ groups $A_1,\ldots,A_w$, and the social planner has a total reveal budget of $K$. 
For each group $a \in [w]$, define
\[
\mathrm{OPT}_{a}^{k}
= \max_{\substack{S \subseteq \Ts \\ |S| \le k}}
\left( \sum_{x \in A_a} Q^{S}(x) - \sum_{x \in A_a} Q^{\emptyset}(x) \right),
\qquad 1 \le k \le K.
\]
as the maximum social welfare gain for group $a$ using a budget of $k$, assuming we focus solely on that group.
Throughout this section, we assume that the number of groups $w$ is a constant and that the total target reveal budget is 
$K \ge w$. A revealed target set is \textit{simultaneously approximately optimal for all groups} (i.e., satisfies all groups), if every group $a \in [w]$ receives social welfare gain proportional to $\mathrm{OPT}_{a}^{K}$. This goal is natural because no group can achieve more than $O(\mathrm{OPT}_{a}^{K})$ social welfare, even if it were allocated the entire budget. We now characterize conditions under which this fairness guarantee can be met.

For arbitrary graphs, if the social planner can only reveal positive targets and allocates a budget of $K_a = \lceil K/w \rceil$ to each group, then all groups can be satisfied (Appendix Theorem~\ref{thm:pos_fairsub}). In contrast, when negative targets may be revealed, there exist graph structures where no solution satisfies all groups (Appendix Remark~\ref{rem:posneg_notfair}). 
On the other hand, when the revealed targets may include negative ones, but all agents are $c$-bounded, then running the proxy-greedy algorithm separately on each group with a budget of $K_a$ guarantees that each group $a$ is helped by $\Omega(\mathrm{OPT}_{a}^{K})$  (Appendix Corollary~\ref{col:proxy-fairness}).

\subsection{NP-Hardness Results}
\label{sec:revealrm_np-hardness}
In this section, we show that when the $\splan$ is restricted to positive targets and the social welfare function is monotone and submodular, hardness follows from a reduction from the max-$K$-cover problem. When the planner is restricted to only revealing negative targets, we prove hardness of the maximization problem via a reduction from the $K$-clique problem in a $\theta$-regular graph.

\begin{theorem}
\label{thm:np-hardness_posonly}
Given a set of agents $x_1,\ldots,x_n$ and a set of targets $\Ts$, when the $\splan$ can only reveal positive targets 
$\Ts^+ = \{t \in \Ts | f(t)=+1\}$, the problem of finding $S_g \subseteq \Ts^+$ of size at most $K \ (|S_g| \le K)$ that maximizes social welfare $F(S_g)$ is NP-hard. Also, unless $\mathrm{P}=\mathrm{NP}$, the problem of finding a positive target subset $S_g \subseteq \Ts^+$ of size $K$ that maximizes social welfare cannot be approximated within a factor better than 
$1-1/e$.
\end{theorem}
\begin{proof}[Proof sketch]
    We prove NP-hardness by a polynomial-time reduction from max-$K$-cover. Given a universe $U=\{e_1,\ldots,e_n\}$ and sets $\mathcal C=\{C_1,\ldots,C_m\}$, we create an instance of our problem by creating one agent for each element and one positive target for each set, connecting an agent to the target if and only if it is in the set. To ensure all agents have the same initial welfare, each agent is also connected to a number of private negative targets equal to its positive degree, ensuring that before any positive target is revealed, every agent contributes $1/2$, so $F(\emptyset)=n/2$. Revealing a positive target raises the contribution of all adjacent agents from $1/2$ to $1$, implying that for any $S\subseteq\Ts^+$ with 
    $|S|\le K$, $F(S)=\frac{n}{2}+\frac12\bigl|\bigcup_{t_j\in S}C_j\bigr|$. Hence, achieving welfare at least 
    $W=\frac{n}{2}+\frac{\mathcal E}{2}$ is equivalent to covering at least $\mathcal E$ elements with at most $K$ sets, establishing NP-hardness. Moreover, since the welfare gain beyond the baseline is exactly proportional to the achieved coverage, any approximation for maximizing social welfare induces an approximation of the same factor for max-$K$-cover. Therefore, unless $\mathrm{P}=\mathrm{NP}$, no polynomial-time algorithm can approximate the problem within a factor better than $1-1/e$. Formal proof is included in Appendix~\ref{sec:revealrm_sm_nphardness-posonly}.
\end{proof}

\begin{remark}
    If the $\splan$ can reveal both positive and negative targets, the problem of revealing a subset $S_g \subseteq \Ts$ with  
    $|S_g| \leq K$ that maximizes social welfare remains NP-hard. This follows directly from 
    Theorem~\ref{thm:np-hardness_posonly} where even though both positive and negative targets can be revealed, revealing negative targets results in less social welfare than revealing positive ones, so the $\splan$'s optimal strategy reduces to the revealing only positive targets. Hence, permitting both types of targets does not change the NP-hardness of the problem.
\end{remark}

\begin{theorem}
\label{thm:np-hardness_negonly}
Given a graph $\graph = (\Xs \cup \Ts, E)$ with $n = |\Xs|$ agents and target set $\Ts$, suppose the $\splan$ can reveal only negative targets $\Ts^{-} = \{t \in \Ts \mid f(t) = -1\}$. Then finding a subset $S_g \subseteq \Ts^{-}$ with $|S_g| \leq K$ that maximizes social welfare is NP-hard.
\end{theorem}
\begin{proof}[Proof sketch]
    We prove NP-hardness by a polynomial-time reduction from the $K$-clique problem in a 
    $\theta$-regular graph. Given an instance $\mathcal G_c=(V,E_c)$, we create an instance of our problem by creating by constructing a bipartite instance with one negative target $t_v^-$ per vertex $v\in V$, one positive target $t_{uv}^+$ per edge $\{u,v\}\in E_c$, and one agent $x_{uv}$ per edge, with neighborhood $N(x_{uv})=\{t_u^-,t_v^-,t_{uv}^+\}$. With no revealed targets, each agent contributes $1/3$ to welfare. Revealing a single adjacent negative target increases that agent's contribution by $1/6$, while revealing both adjacent negatives increases it by $2/3$. Setting the budget to $K$ and the threshold $W$ to $\frac{n}{3}+\frac{K\theta}{6}+\frac{1}{3}\binom{K}{2}$, a set $S\subseteq \Ts^-$ of size $K$ achieves social welfare of at least $W$ if and only if the corresponding vertices form a $K$-clique since non-clique pairs fail to realize the $\binom{K}{2}$ agents that gain the full $2/3$ increase in probability for emulating a positive target. Thus, deciding whether such a set $S$ exists is NP-hard, and because a candidate solution can be verified in polynomial time, the decision problem is NP-complete, which implies NP-hardness of the optimization problem. Formal proof is included in Appendix~\ref{sec:revealrm_sm_nphardness-negonly}.
\end{proof}

\subsection{Learning Setting}
\label{sec:revealrm_learn}
Consider a setting where the left-hand side of the bipartite graph $\mathcal{G} = (\Xs \cup \Ts, E)$ is replaced by a probability distribution $D$ over agents. The social planner draws agents i.i.d. from $D$ and for each agent, the planner observes its neighborhood (adjacent targets) and the probability of emulating a positive target. The planner's goal is to reveal a subset of targets $S\subseteq \Ts$ whose social welfare deviates from the true value (Eq.~\ref{eq:fs_dist}) by at most $\varepsilon$.
\begin{equation}
\label{eq:fs_dist}
F(S) = \mathbb{E}_{x \sim D} \left[ Q^S(x) \right]
\end{equation}
Given a budget $K$ and a sample graph $\mathcal{G} = (\Xs \cup \Ts, E)$ with 
$\mathcal{X}$ agents drawn i.i.d. from $D$, the $\splan$ runs Algorithm~\ref{alg:greedy_lbreveal} on this train graph $\mathcal{G}$ and returns the revealed target set $S_g \subseteq \Ts$ with 
$|S_g| \le K$ as its hypothesis. For a new agent $x_i$, $Q^{S_g}(x_i)$ estimates the agent's probability of emulating a positive target, and the performance of the hypothesis is measured as social welfare per agent. Theorem~\ref{thm:sc_bound} gives a sufficient sample size to ensure, with high probability, that the welfare returned by the hypothesis is within $\varepsilon$ of the true value.
\begin{theorem}
\label{thm:sc_bound}
Let $S_g \subseteq \Ts$ be the target set revealed by Algorithm~\ref{alg:greedy_lbreveal} 
on graph $\mathcal{G} = (\mathcal{X} \cup \mathcal{T},E)$, where
$\mathcal{X}$ is a set of agents sampled independently from ${\cal D},$  $|\Ts|=m$, and the target reveal budget is \(K\). There exists a universal constant $C>0$, such that for any $\varepsilon>0, \delta \leq 1$, if $$|\mathcal{X}| \ge  C((\varepsilon^2)^{-1}(K\log m + \log (1/\delta))),$$
then with probability at least $1-\delta$ the social welfare of the revealed target set $S_g$ differs from its true value by at most $\varepsilon$.
\end{theorem}
\begin{proof}
    See Appendix~\ref{sec:revealrm_app_learningsetting}
\end{proof}

\section{Extensions of the Standard Model}
\label{sec:revealrm_extensions}
The standard model (Section~\ref{sec:revealrm_standardmodel}) operates under the assumption that agents can observe neighboring targets but cannot distinguish positive targets from negative ones. While this captures many relevant settings, it overlooks an important practical constraint: even when a planner reveals a welfare-maximizing set of targets, some agents may still have a low probability of emulating a positive target because their local neighborhoods contain (zero)few positive targets or because no positive targets are directly observable to them. To address this limitation, we extend the standard model to allow budget-constrained interventions that go beyond the passive disclosure of target information. First, the targeted intervention models (Section~\ref{sec:revealrm_intervention_model}) enable the planner to directly connect agents prone to emulating negative targets with positive targets. Second, the coverage radius model (Section~\ref{sec:revealrm_coveragemodel}) increases the visibility of positive targets, expanding the set of agents who can observe and potentially emulate them.

\subsection{The Targeted Interventions Model}
\label{sec:revealrm_intervention_model}
The targeted intervention model focuses on settings in which the $\splan$ identifies high-risk agents, namely those most likely to emulate a negative target, and directly connects them to positive targets. For example, suppose that the school counselor knows that some students don’t have successful STEM mentors, and the counselor intervenes by directly assigning a STEM mentor to some of these students. 

We study two approaches of targeted intervention modeling: one applied before running the standard greedy algorithm with budget \(K\) (pre-reveal), and one applied after it (post-reveal). Below, we provide an overview of both the pre- and post-reveal targeted intervention approaches.  For completeness, Appendix~\ref{sec:revealrm_app_interventionmodel_algos} provides the full algorithmic details of both approaches, and Appendix~\ref{sec:revealrm_app-itmexps} empirically compares their intervention gains while examining how performance varies with the intervention and target-reveal budgets \((B, K)\).

\paragraph{Overview of the pre- and post-reveal intervention algorithms.}
Let $B$ denote the intervention budget, and let $S_o$ be the set of targets revealed either before or after executing the classical greedy algorithm (Algorithm~\ref{alg:greedy_lbreveal}) with a  $K$ target reveal budget.
Let $\Xs_{\mathrm{hr}} \subseteq \Xs$ denote the set of at most $B$ high-risk agents selected for intervention by directly connecting them to a positive target.

Intervening on agent $x \in \Xs_{\text{hr}}$ raises its social welfare from $Q^{S_o}(x)$ to $1$. If $Q^{S_o}(x) = 1$, then the intervention was redundant. The closer $Q^{S_o}(x)$ is to $0$, the larger the intervention gain (i.e., the difference between social welfare from pre- or post-reveal intervention and Algorithm~\ref{alg:greedy_lbreveal}).
In \textit{pre-reveal intervention} (Appendix~Algorithm \ref{alg:TI_greedy_label_reveal}), total social welfare equals the welfare from applying Algorithm \ref{alg:greedy_lbreveal} to the updated graph after removing the intervened-on agents and their edges, plus the welfare from intervening on the high-risk agents $Q' \leq B$.
In \textit{post-reveal intervention} (Appendix~Algorithm \ref{alg:greedy_TI_label_reveal}), total social welfare equals the welfare returned by Algorithm \ref{alg:greedy_lbreveal} on the full graph, plus the welfare from intervening on high-risk agents $Q' = \sum_{x \in \Xs_{\text{hr}}} \big(1 - Q^{S_o}(x)\big)$.

\subsection{The Coverage Radius Model}
\label{sec:revealrm_coveragemodel}
In this section, we study a setting in which some agents are unaware of nearby positive targets, and a social planner intervenes by increasing their visibility so that agents can observe and learn from them. Consider a school counselor who knows which alumni have followed successful career paths. Some of the students the counselor advises can potentially learn from many of these alumni, but they are unaware of most of them. Therefore, the counselor intervenes by highlighting a subset of successful alumni, for example, through newsletters or targeted emails, thus increasing the students’ awareness of these role models.

Formally, consider a geometric bipartite graph $\graph^{+} = (\Xs \cup \Ts^+, E)$ with a set of \textit{d}-featured agents $\Xs=\{x_1,\ldots,x_n\}\subset \mathbb{R}^d$ on the left-hand side and positive targets 
$\Ts^+=\{t_1,\ldots,t_m\}\subset \mathbb{R}^d$ on the right. Each target is labeled positive, and an (unobserved) edge exists between an agent and a target if their Euclidean distance is at most $r$.

To make adjacent positive role models visible to agents so the agents can emulate them, the social planner could either expand agents' visibility or increase the reach of targets. Since the former is trivial, the coverage radius model focuses on interventions from the targets' perspective.  

Each target $t_i$ is assigned a radius $r_i\ge 0$, and an agent $x_j$ is reached if $\|t_i-x_j\|_2 \le r_i$ for some $i \in [m]$. Initially, $r_i=0$ for all $i \in [m]$, and a total radius budget $R$ constrains the intervention. 
The objective of the social planner is to maximize the number of agents reached:
\begin{equation}
\label{eq:coverage-max}
    \begin{aligned}
    \max_{\, r_1,\ldots,r_m \ge 0}\quad
    & \sum_{j=1}^{n} \mathbf{1}\!\left( \exists i \in [m]: \|t_i-x_j\|_2 \le r_i \right) \\
    \text{s.t.}\quad
    & \sum_{i=1}^{m} r_i \le R
    \end{aligned}
\end{equation}

Algorithm~\ref{alg:radius_greedy_coverage} in Appendix~\ref{sec:revealrm_app_coveragemodel_algos} presents a greedy approach to this problem, and Appendix \ref{sec:revealrm_app-cmexps} demonstrates its effectiveness on real-world datasets.

\section{Experiments}
\label{sec:revealrm_experiments}
We conduct extensive experiments to evaluate the performance of greedy strategies in practical settings under the standard model without information disclosure restrictions, and to assess the proposed algorithms under alternative model settings using semi-synthetic geometric bipartite graphs generated from the Adult, Student Performance (Mathematics and Portuguese), and Garment Workers Productivity datasets. Details on the datasets and preprocessing procedures are provided in Appendix~\ref{subsec:revealrm_app-datasets}.

\paragraph{Generation of geometric bipartite graphs.} 
We generate the geometric bipartite graph from two feature sets 
$(\mathcal{X}{\mathrm{LHS}} \in \mathbb{R}^{n \times \rho} \ \text{and} \ \mathcal{X}{\mathrm{RHS}} \in \mathbb{R}^{m^{\star} \times \rho})$ extracted from a given dataset, where $\rho$ denotes the number of features, $n$ the number of agents, and $m^{\star}$ the number of \textit{all} targets.
First, we compute the pairwise distances between the agents ($\mathcal{X}_{\mathrm{LHS}}$) and the targets 
$(\mathcal{X}_{\mathrm{RHS}})$:
\[D_{ij} = \|x_i - x_j\|_2 , \quad i \in [n],\; j \in [m^{\star}]\]
For each agent $i$, its neighborhood $\mathcal{N}(i)$ is defined either by the $k$NN method, where a target $j \in \mathcal{N}(i)$ iff it is among the $k_{\max}\geq 1$ closest targets to $i$ according to $D_{ij}$, or by a distance threshold method, where target $j \in \mathcal{N}(i)$ iff $D_{ij} \le \ell$. The edge set is then given by $E = \{(i,j): j \in \mathcal{N}(i)\}$.\\
Now, together with the \textit{used} targets $\displaystyle \mathcal{T} = \bigcup_{i=1}^n \mathcal{N}(i) \in \mathbb{R}^{m \times \rho}$ and their labels $f(j) = y_{\mathrm{RHS}}[j]$ for $j \in \mathcal{T}$, the bipartite graph is given by $\graph = (\mathcal{X}_{\mathrm{LHS}}\cup \mathcal{T}, E)$. 

See Appendix~\ref{subsec:revealrm_app-graphstats} for more details on the generated graphs.

\paragraph{Algorithms, parameters, and evaluation metrics.}
In the single-group standard model setting, we compare social welfare without budget constraints $F(S_{\mathrm{full}})$ and with zero budget $F(S_{\mathrm{o}})$ to budgeted strategies: random selection, classic greedy, heuristic greedy, and bruteforce search. 
In the grouped setting, we compare average group social welfare (gain) (total group welfare (gain) divided by group size) achieved by the Algorithm \ref{alg:greedy_lbreveal} when applied to (i) the full graph and (ii) male and female bipartite subgraphs constructed from the Adult, Math, and Portuguese datasets.

Under the targeted intervention model, for varying target-reveal $(K)$ and intervention $(B)$ budgets, we evaluate the intervention gains achieved by the pre- and post-reveal intervention. 
These gains are respectively defined as
$\Delta_F(\mathrm{ig}, \mathrm{g}) = F(S_{\mathrm{ig}}) - F(S_g), \ \text{and} \ \Delta_F(\mathrm{gi}, \mathrm{g}) = F(S_{\mathrm{gi}}) - F(S_g), $ where $F(S_g)$, $F(S_{\mathrm{ig}})$, and $F(S_{\mathrm{gi}})$ denote the welfare returned by Algorithms~\ref{alg:greedy_lbreveal}, \ref{alg:TI_greedy_label_reveal}, and \ref{alg:greedy_TI_label_reveal}, respectively.

In the learning setting, we report training $(\mathrm{tr})$ and testing $(\mathrm{ts})$ performance averaged over $100$ independent train-test splits with different random seeds. We report the $\mathrm{tr}$ and $\mathrm{ts}$ performance results using two metrics: $\mathrm{Perf}_{2}$, where $100$ denotes success on all helpable agents (those with both positive and negative target neighbors), including all sampled agents; and $\mathrm{Perf}_{3}$, where 100 denotes success on all helpable agents, excluding unhelpable ones. Full details on algorithms, parameters, and evaluation metrics used are included in Appendices \ref{subsec:revealrm_app-algosexp} and \ref{subsec:revealrm_app-perfeval}.

Empirical results for various settings are reported below and in Appendices~\ref{sec:revealrm_app-smresults}--\ref{sec:revealrm_app-lsexps}.

\begin{figure}[t!]
\captionsetup[subfigure]{justification=Centering}
\begin{subfigure}[t]{0.48\textwidth}
    \includegraphics[width=\textwidth]{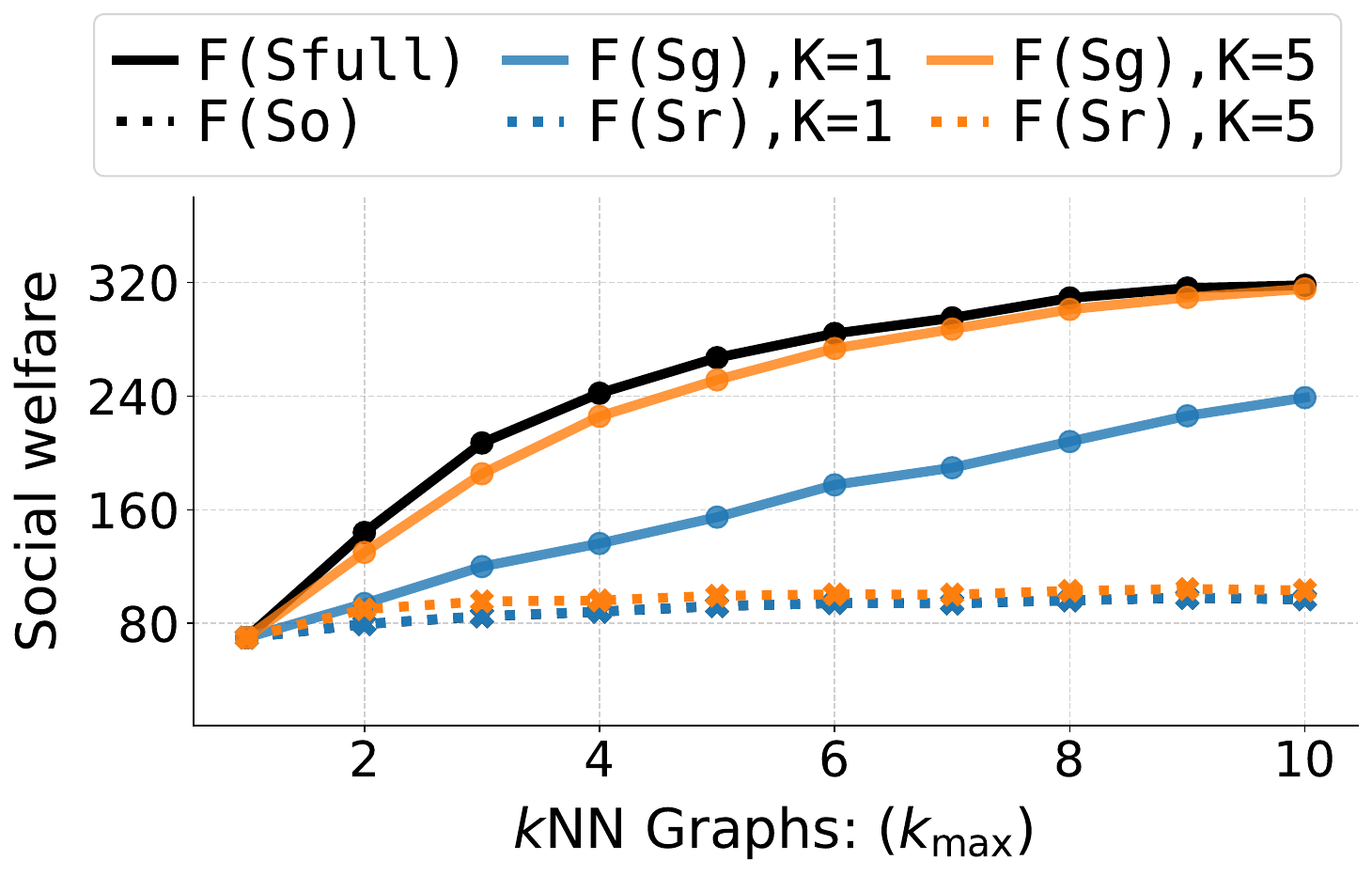}
    \caption{The \(k\)NN graphs: \\$F(S_{\mathrm{g}})$ vs. $F(S_{\mathrm{r}})$}
    \label{fig:adult_sr_sg_knn}
\end{subfigure}
\hfill
\begin{subfigure}[t]{0.48\textwidth}
    \includegraphics[width=\linewidth]{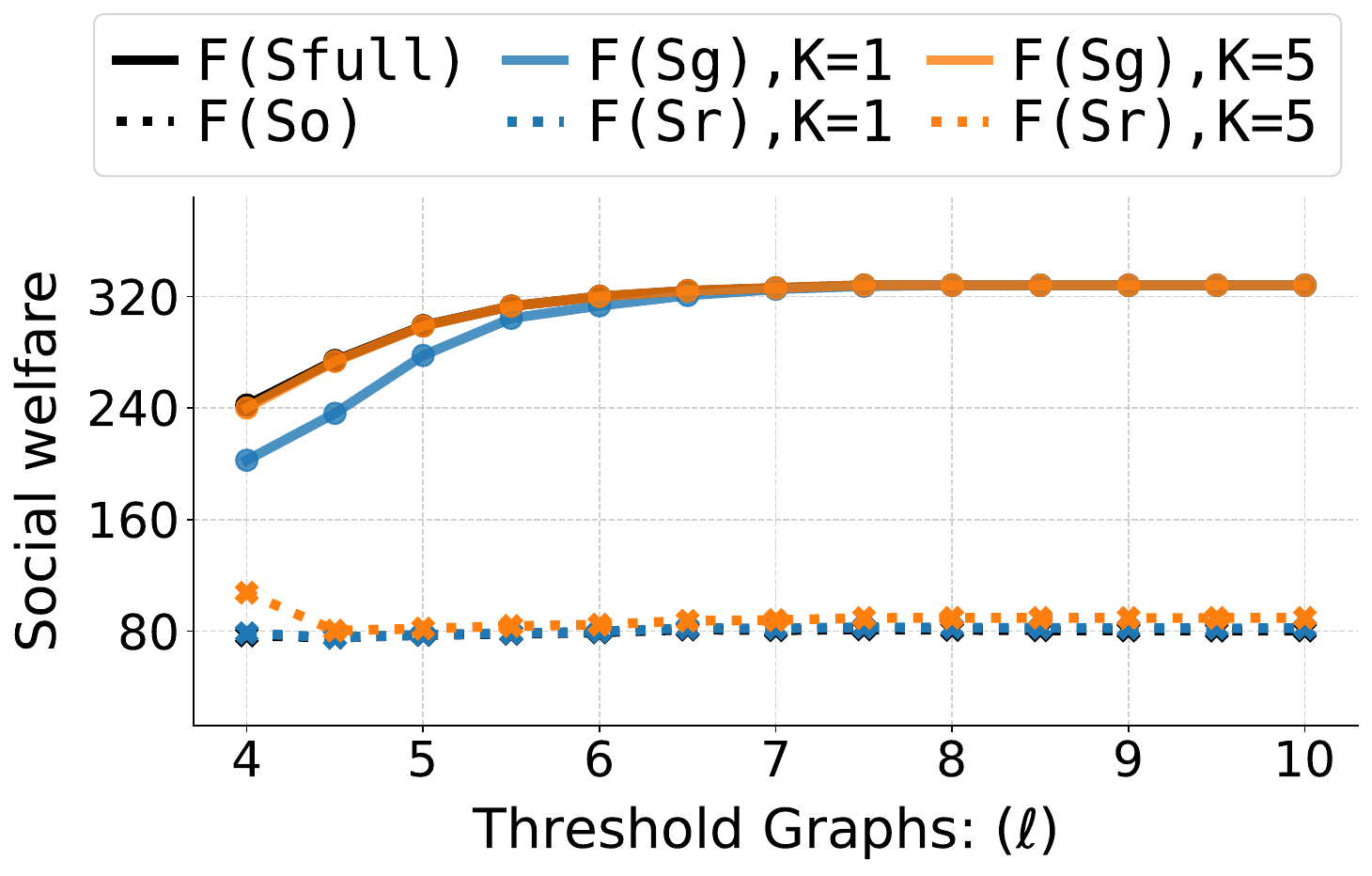}
    \caption{The threshold graphs: \\$F(S_{\mathrm{g}})$ vs. $F(S_{\mathrm{r}})$}
    \label{fig:adult_sr_sg_thresh}
\end{subfigure}
\caption[$F(S_{\mathrm{g}})$ vs. $F(S_{\mathrm{r}})$ on Adult dataset $k$NN and threshold graphs]{Comparative social welfare returned by random selection $F(S_{\mathrm{r}})$ and Algorithm~\ref{alg:greedy_lbreveal} $F(S_{\mathrm{g}})$ on \textbf{Adult} dataset for $K =\{1,5\}$. Black lines mark the maximum $F(S_{\mathrm{full}})$ and minimum $F(S_{\mathrm{o}})$ welfare. Algorithm~\ref{alg:greedy_lbreveal} consistently outperforms random selection when executed at the same target reveal budget $K$.}
\label{fig:sr_sg_shr_shg}
\end{figure}

\subsection{Empirical Results under the Standard Model}
\label{subsec:revealrm_exp_sm}

\paragraph{One group setting.} 
The performance of the budgeted strategies heavily depends on the network structure. When connectivity is low, overall social welfare remains very small, regardless of the algorithm or budget used (see Appendix Tables~\ref{tab:math_kmax_r_stats} and \ref{tab:portuguese_kmax_r_stats} where 
$\ell \le 5.0$, and corresponding results in Figures~\ref{fig:app_sr_sg} and \ref{fig:app_shg_shr}, subfigures (\textbf{f,g}), when $\ell \le 5.0$).
As connectivity increases, particularly in threshold-generated graphs (Appendix Tables~\ref{tab:adult_kmax_r_stats}--\ref{tab:productivity_kmax_r_stats}), the social welfare achieved by classic greedy often matches the maximum achievable ($F(S_{\mathrm{full}})$) (Figure~\ref{fig:adult_sr_sg_thresh}; Appendix Figures~\ref{fig:app_sr_sg}--\ref{fig:adult-port_sstar_sg_shg}, subfigures \textbf{e--h}), because more positive targets are connected to nearly all helpable agents. 
Additionally, when executed at the same $K$, Algorithm~\ref{alg:greedy_lbreveal} consistently outperforms random 
(Figure~\ref{fig:adult_sr_sg_knn}; Appendix Figures~\ref{fig:app_adult_sr_sg_knn}--\subref{fig:app_prod_sr_sg_knn}). Even with high budgets and connectivity, random selection can yield comparably very low social welfare (cf.~Figure~\ref{fig:sr_sg_shr_shg}).
These results and those in Appendix~\ref{sec:revealrm_app-smresults} suggest that although greedy may have weaker theoretical guarantees without information disclosure constraints, it performs well in practice, likely because the graphs are typically well-connected and balanced.

\paragraph{Fairness in a grouped setting.}
With low graph connectivity and before revealing any targets (i.e., $K=0$), the female group generally has lower average social welfare than the male group (Appendix Figures~\ref{fig:app_adult_knn_undivided}--\subref{fig:app_port_knn_undivided} and \ref{fig:app_adult_knn_divided}--\subref{fig:app_port_knn_divided}). 
As connectivity and the budget increase, the average group social welfare (gain) increases and is closely similar across groups (Appendix Figures~\ref{fig:app_adult_thresh_undivided}--\subref{fig:app_port_thresh_undivided} and 
\ref{fig:app_adult_thresh_divided}--\subref{fig:app_port_thresh_divided}), both in the case where the greedy is run exclusively on a specific group at $K/2$ (Appendix Figure~\ref{fig:app_knn_thresh_divided}) and when it's run on the whole graph at a budget of $K$ (Appendix Figure~\ref{fig:app_knn_thresh_undivided}). 
See Appendix~\ref{sec:revealrm_app=fairnessexp} for more empirical results on the fairness setting.

\subsection{Empirical Results under the Targeted Interventions Model}
\label{subsec:revealrm_exp_itm}
\begin{figure}[ht!]
    \centering
    \begin{minipage}[t]{0.45\textwidth}
        \vspace{0.5cm} 
        \centering
         \includegraphics[width=\linewidth]{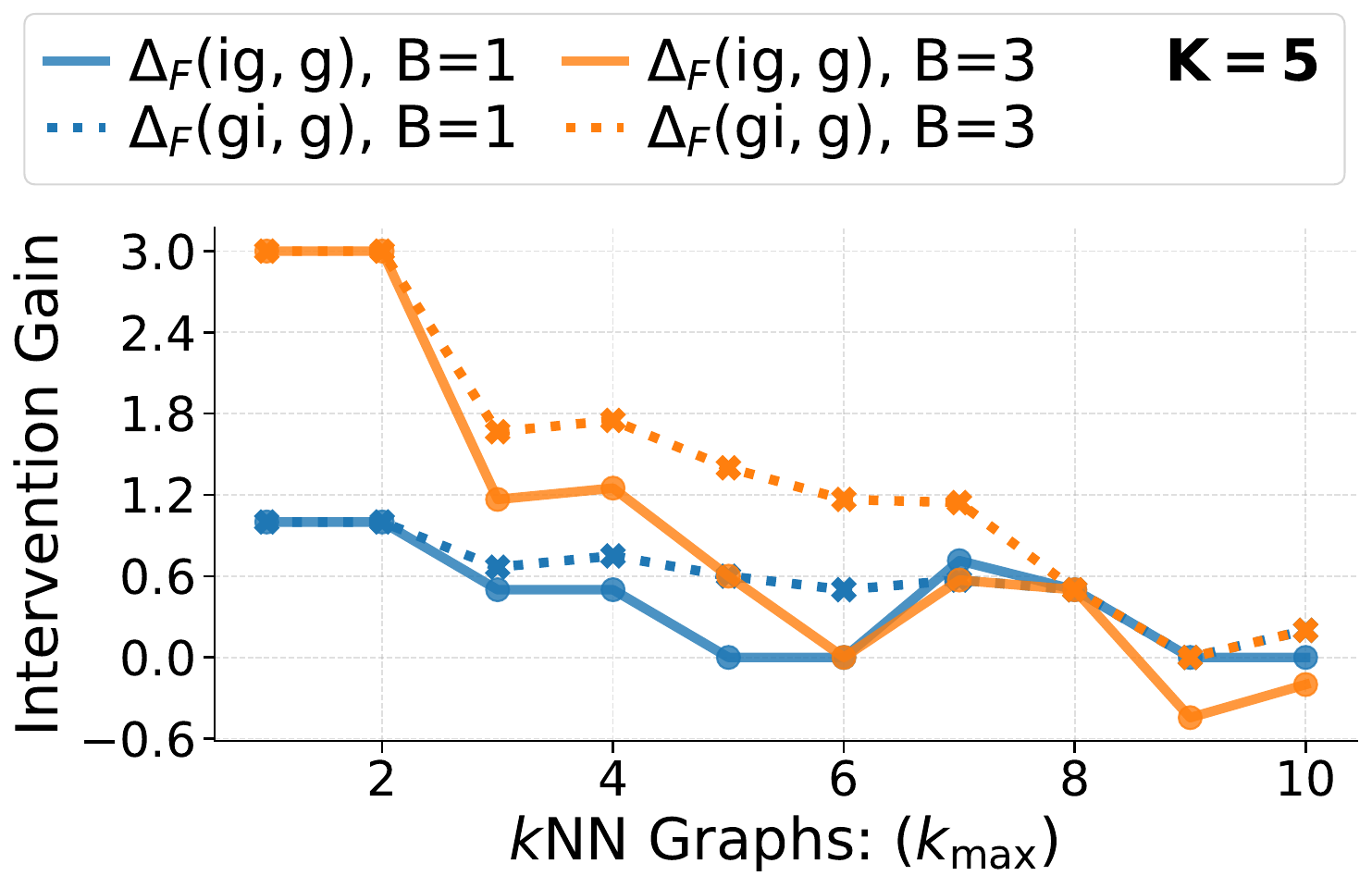}
    \end{minipage}%
    \hspace{0.02\textwidth}
    \vspace{0cm}
    \begin{minipage}[t]{0.52\textwidth} 
        \caption[Comparison of pre-reveal and post-reveal intervention gains]{Comparison of pre- ($\Delta_{F}(ig,g)$) and post-reveal ($\Delta_{F}(gi,g))$) intervention gains, for $K\!=\!5$ with 
        $B\!=\!\{1,3\}$ across $k$NN graphs on the \textbf{Productivity} dataset. Gains increase with decrease in $K$, and pre-reveal gains may be negative.}
        \label{fig:prod_itm_knn_thresh}
    \end{minipage}
\end{figure}

Overall, intervention gains are upper-bounded by the intervention budget $B$ and are larger with a smaller target reveal budget $K$ (Figure~\ref{fig:prod_itm_knn_thresh}; Appendix Figures~\ref{fig:adult_math_itm_knn_thresh} and \ref{fig:port_prod_itm_knn_thresh}) and in graphs where many agents lack positive neighbors (e.g., Appendix Figures~\ref{fig:adult_itm_knn1} and \ref{fig:adult_itm_knn5}). 
When classic greedy is already optimal and most agents have positive neighbors, intervention becomes redundant or underutilized, leading to little or no intervention gains
(Appendix  Figures~\ref{fig:adult_math_itm_knn_thresh} and \ref{fig:port_prod_itm_knn_thresh} (subfigures (\textbf{b,d,f,h}))).
Post-reveal interventions always yield positive intervention gains that are also usually at least as large as those from pre-reveal interventions (Figure~\ref{fig:prod_itm_knn_thresh};  
Appendix  Figures~\ref{fig:adult_math_itm_knn_thresh} and \ref{fig:port_prod_itm_knn_thresh} (subfigures (\textbf{a,c,e,g}))). As shown in Figure~\ref{fig:prod_itm_knn_thresh}, pre-reveal intervention can sometimes yield negative intervention gains because removing high-risk agents and their edges early may distort the graph, causing Algorithm~\ref{alg:greedy_lbreveal} to reveal a target set with lower social welfare than it would have otherwise, especially when high-risk agents already had high probabilities of positive emulation. See Appendix~\ref{sec:revealrm_app-itmexps} for more empirical results under targeted intervention.

\subsection{Empirical Results under the Learning Setting}
\label{subsec:revealrm_exp_ls}
\begin{figure}[ht!]
\centering
\begin{subfigure}[t]{0.48\textwidth}
    \centering
    \includegraphics[width=\linewidth]{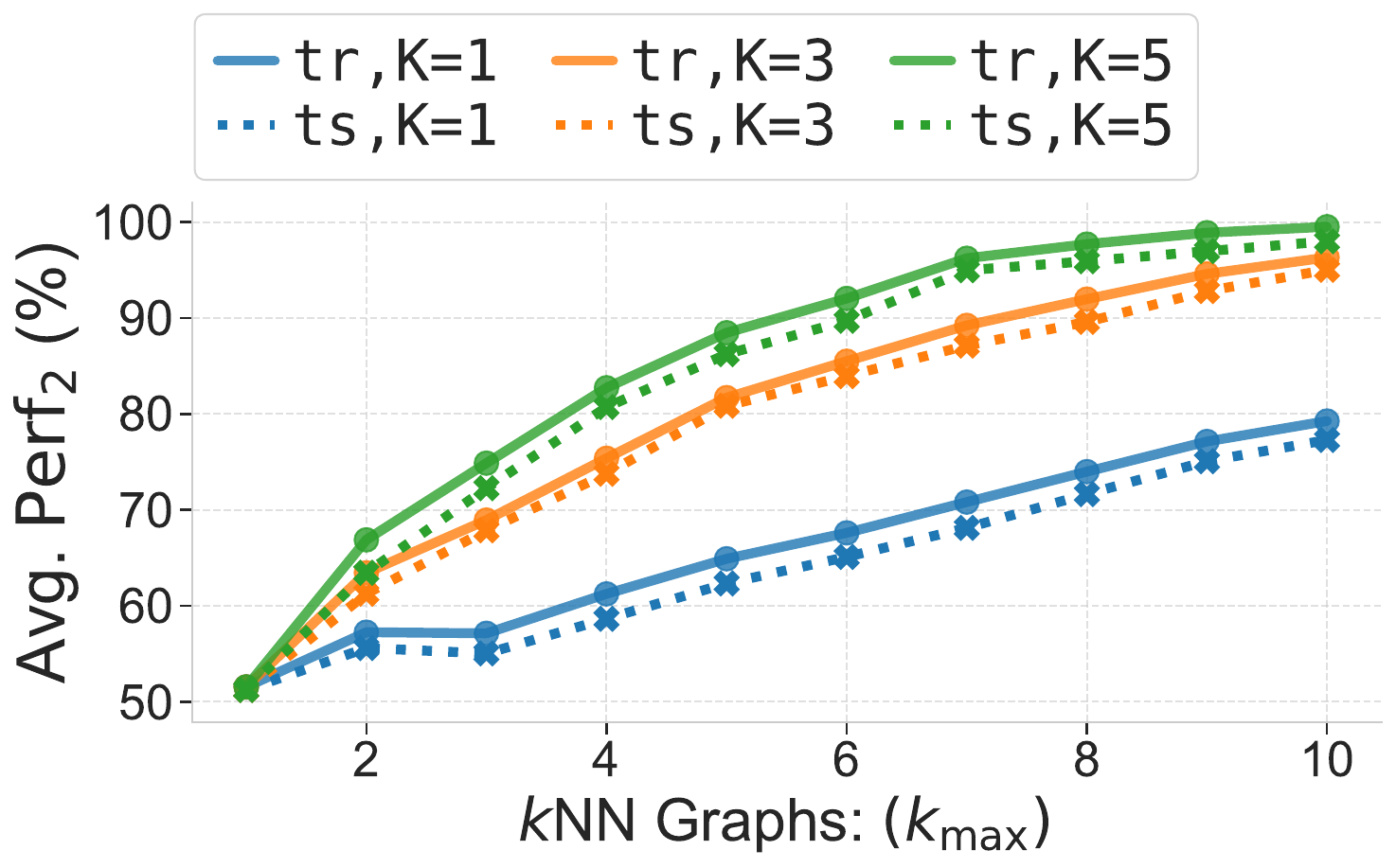}
    \caption{The \(k\)NN generated graphs: \(\mathrm{Perf}_{2}\)}
    \label{fig:math_learn_knn_perf2}
\end{subfigure}
\hfill
\begin{subfigure}[t]{0.48\textwidth}
    \centering
    \includegraphics[width=\linewidth]{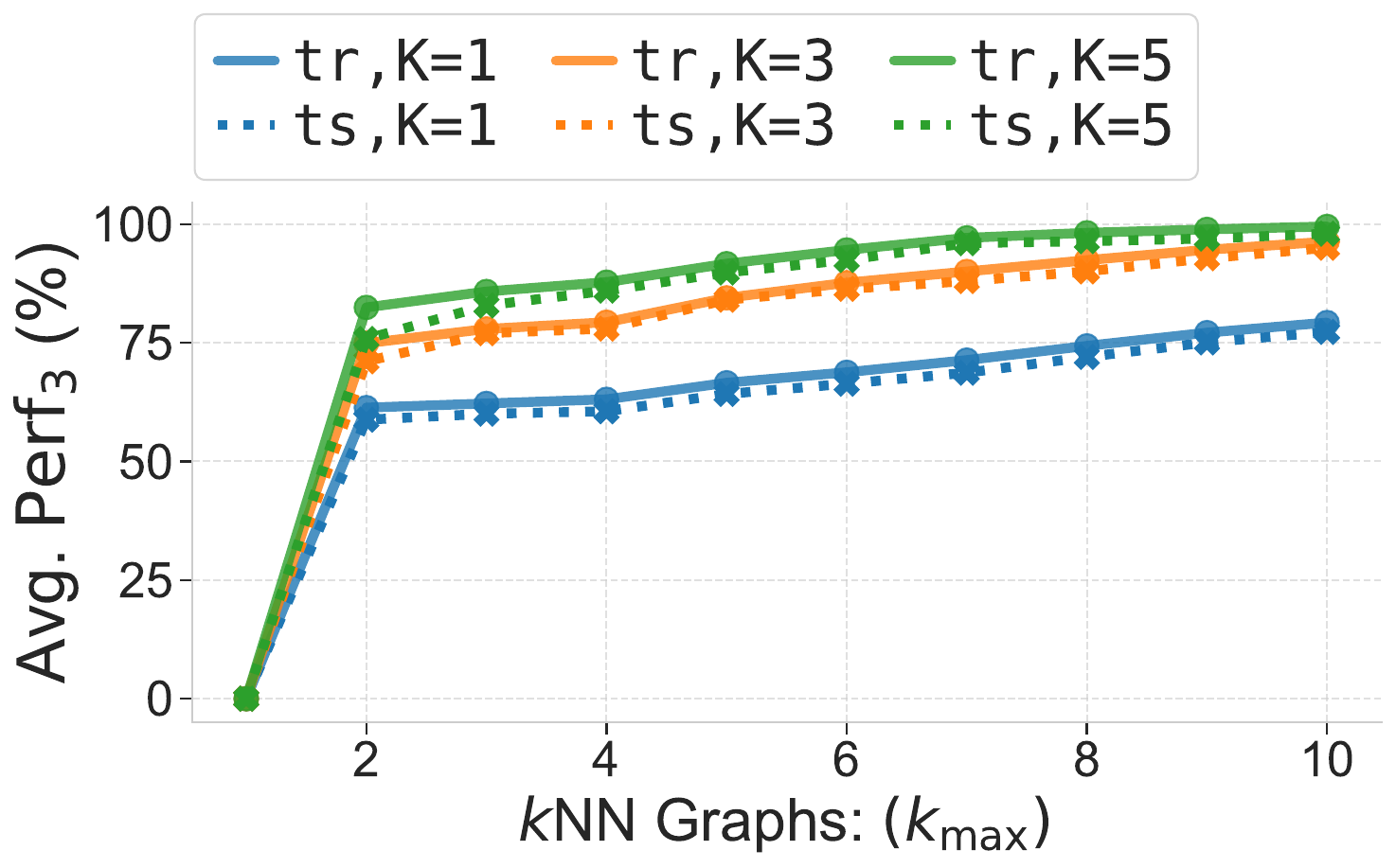}
    \caption{The \(k\)NN generated graphs: \(\mathrm{Perf}_{3}\)}
    \label{fig:math_learn_knn_perf3}
\end{subfigure}
\caption[Analysis of the $\mathrm{tr}$ and $\mathrm{ts}$ performance $(\mathrm{Perf}_2, \mathrm{Perf}_3)$ scores]{Analysis of the training ($\mathrm{tr}$) and testing ($\mathrm{ts}$) performance $(\mathrm{Perf}_2, \mathrm{Perf}_3)$ scores when Algorithm~\ref{alg:greedy_lbreveal} is run under budget $K=\{1,5\}$ on $k$NN graphs (see Appendix Table~\ref{tab:math_kmax_r_stats}) from the \textbf{Math} dataset. When each agent in train/test sets has at most one neighbor, $\mathrm{Perf}_3$ is zero (\subref{fig:math_learn_knn_perf3}). Both 
$\mathrm{Perf}_2$ and $\mathrm{Perf}_3$ increase with $K$.} 
\label{fig:math_learn_knn_thresh}
\end{figure}

The evaluation of Algorithm~\ref{alg:greedy_lbreveal} over $100$ randomized train-test splits shows that the average training performance generally matches or slightly exceeds testing performance across all datasets and metrics (Figure~\ref{fig:math_learn_knn_thresh} and Appendix Figures~\ref{fig:app_adult-math_learn_knn_thresh} and \ref{fig:app_port-prod_learn_knn_thresh}). 
As graph connectivity increases, especially in threshold-generated graphs, training and testing performances converge, and the impact of a higher budget diminishes (Appendix Figures~\ref{fig:app_adult-math_learn_knn_thresh} and \ref{fig:app_port-prod_learn_knn_thresh} 
(subfigures (\textbf{d--f, j--l}))). 
In contrast, lower connectivity, particularly in $k$NN graphs, amplifies budget effects, with higher budgets consistently producing better or equal train/test performance  (Figure~\ref{fig:math_learn_knn_thresh} and Appendix 
Figures~\ref{fig:app_adult-math_learn_knn_thresh} and \ref{fig:app_port-prod_learn_knn_thresh}  (subfigures (\textbf{a--c, g--i}))). 
Lastly, train/test performance depends on graph structure, neighbor positivity, and metric; e.g., when 
$|N(x)|=1$ for all $x \in \Xs$, $\mathrm{Perf}_{2}$ equals the fraction of agents connected to positive targets (Figure~\ref{fig:math_learn_knn_perf2}), and $\mathrm{Perf}_{3}=0$ reflects number of helpable agents  (Figure~\ref{fig:math_learn_knn_perf3}). See Appendix~\ref{sec:revealrm_app-lsexps} for more empirical learning results.

\section{Conclusion}
\label{sec:revealrm_conclusion}
We propose various greedy strategies to help agents make good decisions when they observe targets within their social circles but lack information about the targets' labels. Although theoretical performance guarantees may weaken once negative targets can be in the revealed set because the true social welfare function may become supermodular, empirical evidence suggests that the classic greedy algorithm would likely perform well in practice, as graphs are more likely to be well-connected and balanced. To preserve submodularity, we introduce a proxy welfare function that achieves a constant-factor approximation to the true optimal welfare (gains) when agents are $c$-bounded. When agents are divided into groups, the proxy ensures that each group's welfare gain is within a constant factor of the optimum under full budget allocation. We also study interventions in which a social planner either directly connects high-risk agents to positive targets or increases the visibility of selected targets when agents are otherwise unaware of them. Future work could incorporate weighted edges to model heterogeneous emulation probabilities, extend our modeling setup to strategic classification by treating emulation of positive targets as improvement and negative targets as gaming, and generalize the deterministic graph to a stochastic setting (e.g., bipartite stochastic block models) where targets are positive with probability $p$ and edges from agents to positive and negative targets form with probabilities $\qpos$ and $\qneg$, respectively.
See Appendix~\ref{sec:revealrm_limitations} for additional discussion.

\clearpage
\thispagestyle{empty} 
\begin{center}
    {\Huge\bfseries
    Part~III \\[1.5em]
    Shaping Human-AI Interactions to \\[0.3em]
    Jointly Optimize Competing \\[0.6em]
    Objectives 
    }
\end{center}
\chapter{PAC Learning with Improvements}
\label{chap:paclearn}
\section{Introduction}

There has been growing interest in recent years in machine learning settings where a deployed classifier will influence the behavior of the entities it is aiming to classify.  For example, a classifier that maps loan applicants to credit scores and then uses a particular cutoff $\hat{\theta}$ to determine whether an applicant should receive a loan will induce those below the cutoff value to take actions to improve their score. This setting is called strategic classification \citep{hardt2016strategic} or measure management \citep{bloomfield2016counts} when the actions taken do not truly improve the agent's quality, and performative prediction \citep{perdomo2021performativeprediction} more generally.  
In this chapter, our focus is on the case that the improvements are real e.g., paying off high-interest credit card debt, taking a money management class, etc. for genuinely improving one's loan application.  
That is, the agent  responds to the classifier in order to potentially improve 
their classification~\citep{Kleinberg2018HowDC,Miller2019StrategicCI}, changing its true features in the process. The classifier must take this ``strategic improvement'' response into account.

Unlike previous works on strategic improvement that focus extensively on efficiently incentivizing and maximizing agent improvement (e.g., \citep{Miller2019StrategicCI,Kleinberg2018HowDC,Haghtalab2020MaximizingWW,Shavit2020LearningFS}, among others), we aim to understand how an agent's capacity for improvement impacts learnability, sample complexity, and algorithm design for accurate classification.
One high-level take-away from our theoretical analysis and empirical results is that the ability of agents to improve favors algorithms that are more ‘‘conservative'' in their decisions.  This is both due to the reduced concern over false-negative errors (since agents in those regions may still be able to improve to be classified as positive) and the increased concern over false-positive errors (which may cause individuals to ``improve'' incorrectly).

To illustrate the potential reduction in 
sample complexity that result from agents' ability to improve, one of the most basic lower bounds in machine learning is that in nearly any nontrivial setting, it takes \emph{at least} $1/\epsilon$ samples to learn to error $\epsilon$ (and more, if the classifier being learned is complex).  However, if agents have the ability to improve by a small amount, we may actually be able to achieve \emph{zero} error by just being ``close enough''.   
To the best of our knowledge, this  has not been previously observed in the strategic improvement literature.
Returning to the loan example above, suppose that by putting in effort, agents can improve their credit score by some small amount $r$, and suppose we are in the realizable case that there is some true threshold $\theta$ such that agents with credit score above $\theta$ will be good customers and those who score below $\theta$ will not. In that case, if we learn an approximation $\hat{\theta}$ of $\theta$ such that $\theta \leq \hat{\theta} \leq \theta + r$ and use it as a cutoff to determine who should receive a loan, we can actually achieve zero error in  
that (a) any agent classified as positive is truly qualified, and (b) any agent who truly is qualified can get classified as positive by putting in effort.  Thus, the ability for agents to improve can potentially allow for a goal one could not hope to achieve otherwise.

We also observe fundamental differences in the inherent learnability of concept classes, compared to both standard PAC learning where the agents cannot respond to the classifier, as well as the strategic PAC setting where the agent tries to deceive the classifier to obtain a more favorable classification. Somewhat surprisingly, learning with improvements can sometimes be easier than the standard PAC setting, and it can sometimes be harder than strategic classification. We show that proper learnability with improvements in the realizable setting is closely linked to the concept class being intersection-closed.

Concretely, our contributions are as follows:

\begin{itemize}
    \item In Section~\ref{sec:paclearn_separating-PAC-and-improvements}, we show a separation between the standard PAC model and our model of PAC learning with improvements. Specifically, we show that
    a finite VC dimension is neither necessary nor sufficient for PAC learnability with improvements. We further show a similar separation from the more recently studied  PAC model for strategic classification \citep{hardt2016strategic,strategicPAC}.
    
    \item In Section~\ref{sec:paclearn_geometric-concepts}, we study learnability of geometric concepts in $\mathbb{R}^d$. We show that any intersection-closed concept class is learnable under our model, and show that the generalization error can be smaller than the standard PAC setting for interesting cases including thresholds and high-dimensional rectangles. We also show that the intersection-closed property is essentially necessary for proper learnability in our setting.

    \item In Section~\ref{sec:paclearn_graph-model}, we study a graph model in which each node represents an agent and the improvement set of an agent is the set of its neighbors in the graph. We establish near-tight bounds on the number of labeled points the learner needs to see to learn a hypothesis which achieves zero error with high probability, given the ability to learn the labels of uniformly random nodes. We further show that it is possible to learn a ``fairer'' hypothesis that also enables improvement whenever it leads to a better classification for an agent. We also study a \emph{teaching} setting where the teacher aims to find the smallest set of labels needed to ensure that a risk-averse student achieves zero-error, and show that providing the labels for the dominating set of the positive subgraph (induced by the true positive nodes) is sufficient.

    \item In Section~\ref{sec:paclearn_experiments}, 
    we conduct experiments on three real-world and one fully synthetic binary classification tabular datasets to investigate how the error rate of a model function (\(h\)) decreases when test-set agents that it initially classified as negative improve. Our results indicate that while risk-averse models may start with higher error rates, their errors rapidly drop as the negatively classified test agents improve and the improvement budget (\(r\)) increases. 

    A stricter penalty for false positives typically leads to more accurate positive classifications, resulting in greater gains from agent improvements. In most 
    cases, test errors decline sharply, sometimes reaching zero (e.g., in Figure~\ref{fig:synthetic_0.5}).
\end{itemize}

\paragraph{Related Work.}
Learning in the presence of strategic (``gaming''), utility-maximizing agents has gained increasing attention in recent years (\citep{Hu:2019:,Milli2018TheSC,braverman_et_al,strategicperceptron,Haghtalab2020MaximizingWW}, among others). Early research framed this problem as a Stackelberg competition \citep{hardt2016strategic,adversarial_games_pred}, where negatively classified agents manipulate their features to obtain more favorable outcomes if the benefits outweigh the costs. \citet{Kleinberg2018HowDC} extend this model by considering agents who can both manipulate and genuinely improve their features, proposing a mechanism that incentivizes authentic improvement. {This model has been studied under a causal lens, where the learner may not a priori know which features correspond to manipulation or improvement. Strategic learning from observable data requires solving a causal inference problem in this setting~\citep{Miller2019StrategicCI}, and the ability to test different decision rules can be helpful~\citep{Shavit2020LearningFS}.}
\citet{ahmadi2022classificationstrategicagentsgame} consider a similar setting and propose classification models that balance maximizing true positives with minimizing false positives. 

We extend this line of work by analyzing the inherent learnability of classes, the sample complexity of learning, and the ability to achieve zero-error classification, when agents can truly improve. {Inherent learnability of concepts has been studied in the strategic manipulation setting \citep{strategicPAC,cohen2024learnability,lechner2022learning}, but not in the strategic improvement setting. In Section~\ref{sec:paclearn_comparison-strategic}, we show how our improvement setting differs from  strategic manipulation with respect to learnability.
The sample complexity of learning in the presence of purely improving agents has  been studied by \citet{Haghtalab2020MaximizingWW}, but from a social welfare perspective where the goal is to maximize the true positives after improvement. In contrast, our primary focus is classification error, which is more sensitive to false positives. In Section~\ref{sec:paclearn_enabling-improvement}, we show that these two objectives need not be in conflict and may be simultaneously optimized. Finally, the ability of a learner to achieve zero-error for non-trivial concept classes and distributions has not been previously observed in any strategic or non-strategic setting.}

Our work also relates to research in reliable machine learning \citep{rivest1988learning,yaniv10a}, where a learner may abstain from classification to avoid mistakes, balancing coverage (the proportion of classified points) against error. In contrast, we strive for a zero false positive rate and minimal false negative rate, aligning with learning under one-sided error \citep{natarajan1987learning,kalaiReliable}.
We include a more detailed discussion of the related work in Appendix~\ref{app:related-work}.

\section{Formal Setting: PAC Learning with Improvements}\label{sec:paclearn_setting}
 Let $\mathcal{X}$ denote the instance space consisting of agents with the ability to \emph{improve}, as defined below. We restrict our attention to the case of binary classification, i.e., the label space is $\{0,1\}$. Without loss of generality, we refer to label $0$ as the \emph{negative} class and label $1$ as the \emph{positive} class. Let $\Delta:\mathcal{X}\rightarrow 2^\mathcal{X}$ denote the \emph{improvement function} that maps each point (agent) $x\in \mathcal{X}$ to a set of points $\Delta(x)$ (the \emph{improvement set}) to which $x$ can potentially \emph{improve} itself in order to be classified positively. For example, if $\mathcal{X}$ is a metric space, we can define $\Delta(x)$ as the  $\ell_p$-ball centered at $x$. 
 Let $\mathcal{H}\subseteq \{0,1\}^\mathcal{X}$ denote the concept space, that is, the set of candidate classifiers.
 We will focus on the \emph{realizable} setting, i.e.\ we assume the existence of an unknown (to the learner) target concept $f^{\star}:\mathcal{X}\rightarrow\{0,1\}$ that correctly labels all points in $\mathcal{X}$ and satisfies $f^{\star}\in \mathcal{H}$. 

The intuition behind the model is as follows. The learner first publishes a classifier 
$h:\mathcal{X}\rightarrow\{0,1\}$ (potentially based on some data sample labeled according to $f^{\star}$). Each agent then reacts to 
$h$~\citep{zrnic2022leadsfollowsstrategicclassification, hardt2016strategic}---if it was classified negatively by $h$, the agent attempts to find a point in its improvement set that is positively classified by $h$ and moves to it. Note that the agents do not know the true function $f^{\star}$ and as a result cannot react with respect to the ground truth, only based on $h$.

We formalize this as the \emph{reaction set} with respect to $h$,

\begin{align}
    \Delta_{h}(x) =\!
    \begin{cases}
        \{x\} \text{\ \ \ if } h(x) = 1,\\
        \{x\} \text{\ \ \ if } \{x'\in \Delta(x) \mid h(x')=1\}= \emptyset, \\
        \{x'\in \Delta(x):  h(x')=1\} \hfill \text{ otherwise.}
    \end{cases}%
    \end{align}
In other words, if $h$ classifies $x$ as positive, the agent $x$ stays in place and does not attempt to improve. If $h$ classifies $x$  as negative, there are two types of reactions. Either, there is no point in its improvement set that can improve the agent's classification according to $h$ and the agent again stays put. Otherwise, the agent reacts and moves to be predicted positive by $h$. This corresponds to utility-maximizing agents that have a utility of $1$ for being classified as positive, a utility of $0$ for being classified as negative, and that incur a cost for moving, where $\Delta(x)$ corresponds to the points that $x$ can move to at a cost less than $1$. 

We say that a test point $x$ has been misclassified if there exists a point in the reaction set of agent $x$ where $h$ disagrees with $f^{\star}$, formally, 
\begin{align}
\label{def:loss-function}
    \mathsc{Loss}(x; h, f^{\star})=\max_{x'\in \Delta_{h}(x)} \mathbb{I}\left[h\left(x'\right)\neq f^{\star}\left(x'\right)\right]. 
\end{align}

\begin{remark}
\label{rmk:misleading-improvement}
  The formulation of the loss function in \eqref{def:loss-function} allows for scenarios where an input \( x \) initially satisfies \( f^{\star}(x) = 1\) and \( h(x) = 0 \), but under \( \Delta_h(x) \), it may transition to a setting where \( f^{\star}(x') = 0 \) and \( h(x') = 1 \) for some \( x' \in \Delta_h(x) \). Think of an example where there are two features such that improving one often comes at the expense of the other. For instance, consider the trade-off between strength and endurance in athletics. Let $f^{\star}(x)$ represent a person's endurance (e.g., marathon running capability), and $h(x)$ represent their strength (e.g., sprinting power). Focusing on increasing $h(x)$ through strength training enhances power, but this often comes at the expense of endurance, thus reducing $f^{\star}(x)$. This reflects the natural conflict between optimizing for one feature while sacrificing the other.
\end{remark}

In words, this corresponds to an assumption that agents will improve to a point in their reaction set while breaking ties adversarially, or equivalently, that they will break ties in favor of points $x'$ for which $f^{\star}(x')=0$. 
This assumption is natural if we want our positive results to be robust to unknown tie-breaking mechanisms, and would also hold if improving to points $x' \in \Delta_h(x)$ whose true label according to $f^{\star}$ is negative is less effort than improving such points whose true label is positive.
Note that this loss function favors classifiers that label uncertain points as negative rather than positive.
For example, if $\{x \mid h(x)=1\} \subseteq \{x \mid f^{\star}(x)=1\}$ then $h$ may still have zero loss if all points $x$ in the difference have at least one point $x' \in \Delta(x)$ for which $h(x')=1$.  The fact that true positives might need to put in effort to improve in order to be classified as positive (or that some negative points are not able to improve themselves to be classified as positive by $h$ even if they would have been able to do so with respect to $f^{\star}$) does not count as an error in our setting. 

See Section~\ref{subsec:paclearn_thresholds} for a concrete example.

Analogous to standard PAC learning, we assume the learner has access to a finite set of samples $S\in \mathcal{X}^m$ drawn randomly according to some fixed but unknown distribution $\mathcal{D}$ over $\mathcal{X}$, and labeled by $f^{\star}$. The learner's population loss is given by $\mathsc{Loss}_{\mathcal{D}}(h,f^{\star})=\mathbb{P}_{x\sim \mathcal{D}}\left[\mathsc{Loss}(x; h, f^{\star})\right]$. This is formalized in the following.

\begin{definition}[PAC Learning with improvements]
\label{def:pac-learning-with-improvements}
    Algorithm $\mathcal{A}$ \emph{PAC-learns with improvements} a concept class $\mathcal{H}$ with respect to improvement function $\Delta$ and data distribution $\mathcal{D}$ using sample size $M\coloneqq M(\epsilon,\delta,\Delta,\mathcal{H},\mathcal{D})$\footnote{We say the sample complexity of $\mathcal{A}$ is the smallest such $M$.}, if for any $f^{\star} \in \mathcal{H}$, any $\epsilon > 0$ and $\delta>0$, the following holds. Algorithm \( \mathcal{A} \), with access to a sample $S\overset{\text{i.i.d.}}{\sim}\mathcal{D}^M$ labeled according to $f^{\star}$, produces with probability at least \( 1 - \delta \) a hypothesis 
    $h$ 
    with \( \mathsc{Loss}_{\mathcal{D}}(h,f^{\star})\le \epsilon \). We further say that $\mathcal{A}$ learns  $\mathcal{H}$ w.r.t.\ $\Delta$ and $\mathcal{D}$ with zero-error with sample size $M$ if for any $\delta>0$, given $S\overset{\text{i.i.d.}}{\sim}\mathcal{D}^M$ labeled by $f^{\star}$, it returns $h$
    with \( \mathsc{Loss}_{\mathcal{D}}(h,f^{\star})=0\) with probability at least $1-\delta$. We will also consider distribution-independent learning, where the guarantee should hold for all distributions $\mathcal{D}$ and proper learning where we require $h \in \mathcal{H}$.
\end{definition}
Note that in our learning with improvements setting zero-error can be achieved by learning (with high probability) from a finite sample in several interesting cases, which is impossible to achieve in the standard PAC model.

\section{Separating PAC Learning with Improvements from the Standard and Strategic PAC Models}\label{sec:paclearn_separating-PAC-and-improvements}

In this section, we prove that learning with improvements diverges from the 
behavior of the standard PAC model for binary classification, and also from the more recently studied PAC learning model for strategic classification \citep{hardt2016strategic,strategicPAC}.

\subsection{Comparison with the standard PAC model}

In the standard PAC model, the learnability of a concept class is equivalent to the class having a finite VC dimension. However, in our setting, where agents can improve, this condition is neither necessary nor sufficient for learnability. Concretely, we demonstrate that a class with an infinite VC dimension can still be learnable with improvements. We also provide examples of hypothesis classes with finite VC dimensions and corresponding improvement sets that cannot be learned in our framework. 

\begin{theorem}
    Finite VC dimension is neither necessary nor sufficient for PAC learnability with improvements. 
\end{theorem}

\begin{proof}
    The proof is in Examples \ref{ex:example_notnecessary} and \ref{ex:example_notsufficient} below.

    \begin{example}[Finite VC dimension is not necessary for learnability with improvements]
        Consider any class $\mathcal{H}$ of infinite VC-dimension, and define $\Delta(x)=\mathcal{X}$ for all examples $x\in \mathcal{X}$.  We can learn this class $\mathcal{H}$ with respect to this improvement function $\Delta$ with sample complexity $M(\epsilon,\delta)=\frac{1}{\epsilon}\ln(\frac{1}{\delta})$ for any data distribution $\mathcal{D}$ as follows.  First, draw a sample $S$ of size $M(\epsilon,\delta)$. Next, if all examples in $S$ are negative, then output the ``all-negative'' classifier; otherwise, select any positive example $x^* \in S$ and output the classifier $h(x)=\mathbb{I}[x=x^*]$. Note that in the latter case, the hypothesis $h$ has error zero, because all agents will improve to $x^*$.  Therefore, if $\Pr_{x\sim \mathcal{D}}[f^*(x)=1] > \epsilon$, then $h$ will have zero error with probability at least $1-\delta$, whereas if $\Pr_{x\sim \mathcal{D}}[f^*(x)=1] \leq \epsilon$, then $h$ will have error at most $\epsilon$ with probability $1$.
        
        \label{ex:example_notnecessary}
    \end{example}

    \begin{example}[Finite VC dimension is not sufficient for learnability with improvements]
        Let the instance space $\mathcal{X}$ be $[0,1]$, let ${\mathcal H}=\{h_{abcd} : h_{abcd}(x)=1 \mbox{ iff }x\in [a,b)\cup(c,d]\}$ (i.e., $\mathcal{H}$ is the class of unions of two intervals, where to make the example easier, we define the intervals to be half-open), and let $\mathcal{D}$ be the uniform distribution over $[0,1]$. We define $\Delta$ as follows.  For $x\in [0,1/4) \cup (3/4,1]$ let $\Delta(x)=[0,1]$; for $x\in [1/4,3/4]$, let $\Delta(x)=\{x\}$. 
    
        We claim that no algorithm with finite training data can guarantee an expected error of less than $1/4$, even though the class is easily PAC-learnable without improvements.
    
        Consider a target function defined as the union of two intervals $[1/4, b) \cup (b, 3/4]$ where the number $b$ was randomly chosen in $[1/4, 3/4]$.  With probability $1$, the learner will not see the point $b$ in its training data, so it learns nothing from its training data about the location of $b$.  Finally, if the learner outputs a classifier whose positive region has probability mass $\leq 1/4$, then its error rate is at least $1/4$ because the positive examples cannot move so at least half of their probability mass will get misclassified.  On the other hand, if the learner outputs a classifier whose positive region has probability mass greater than $1/4$, then it has at least a $50 \%$ chance of including a negative point in its positive region (it will surely include a negative point if it is not contained in $[1/4, 3/4]$ and has at least a 50\% chance of doing so otherwise, since $b$ was uniformly chosen from $[1/4,3/4]$).  If the classifier has a negative point in its positive region, then it will have an error rate at least $50\%$, because all the negatives in $[0, 1/4)$ and $(3/4, 1]$ will move to a false positive (here we use that agents break ties adversarially).  So, either way, its expected error is $25\%$.
    \label{ex:example_notsufficient}
    \end{example}

\end{proof}

Union of two intervals is arguably the simplest class that is not intersection-closed (Definition~\ref{def:intersection-closed}). Indeed, we show in Section~\ref{sec:paclearn_intersection-closed} that such an example could not be possible for intersection-closed classes. 

\paragraph{A separation from standard PAC learning model when the learner's hypothesis space $\tilde{\mathcal{H}}$ is a strict subset of the concept space ${\mathcal{H}}$ that contains $f^{\star}$.} 
We show below yet another sense in which there is a separation when learning with improvements compared to the standard PAC setting. Let $\tilde{\mathcal{H}}\subset \mathcal{H}$ denote the set of hypotheses the learner is allowed to output, and  the target concept $f^{\star}\in \mathcal{H}\setminus \tilde{\mathcal{H}}$. 
We construct an example where the error of the best hypothesis in $\tilde{\mathcal{H}}$ is zero in the standard PAC setting, but it is impossible to avoid a constant error rate when learning the same target with improvements using the same hypothesis space $\tilde{\mathcal{H}}$.

\begin{theorem}
    Consider a hypothesis class $\tilde{\mathcal{H}}$ and target function $f^{\star} \not\in \tilde{\mathcal{H}}$.  It is possible to have $d(f^{\star},\tilde{\mathcal{H}})=0$ in the standard PAC setting (there exists $h\in \tilde{\mathcal{H}}$ that achieves error $0$) but $d(f^{\star},\tilde{\mathcal{H}}) = 1/2$ for PAC learning with improvements (i.e., all  $h\in \tilde{\mathcal{H}}$ have error at least $1/2$). Here, $d(f^{\star},\tilde{\mathcal{H}})$ denotes the error of the best classifier in $\tilde{\mathcal{H}}$ w.r.t. the target $f^{\star}.$
\label{thm:non-realizable-targets}
\end{theorem}

\begin{proof}
    See Example~\ref{ex:example4} in Appendix \ref{app:example-error-gap}
\end{proof}

\subsection{Comparison with the  PAC Model for Strategic Classification}\label{sec:paclearn_comparison-strategic}

We first observe that the  strategic classification loss can be obtained by a subtle modification to our loss function (Equation~\ref{def:loss-function}),

\begin{align}
\label{def:loss-strategic}
    \mathsc{Loss}^{\mathsc{Str}}(x; h, f^{\star})=\max_{x'\in \Delta_{h}(x)} \mathbb{I}\left[h\left(x'\right)\neq f^{\star}(x)\right]. 
\end{align}

\noindent Intuitively,  for a negative point with $f^{\star}(x)=0$, $\Delta_h(x)$ here denotes the set of points that the agent $x$ can ``pretend'' to be in order to potentially deceive the classifier $h$ into incorrectly classifying the agent positive. Since the movement within $\Delta_h(x)$ is viewed as a manipulation by the agent $x$, the prediction on the strategically perturbed point is compared with the original label of $x$, i.e.\ $f^{\star}(x)$.

Prior work has shown that learnability in the strategic classification setting is captured by the \emph{strategic VC dimension} (SVC) introduced by \citet{strategicPAC}. We state below the definition of SVC, adapted to our setting above which is a special case of the strategic classification setting studied in \citet{strategicPAC}.

\begin{definition}[Strategic VC dimension \citep{strategicPAC}]
\label{def:strategic-vc}
    Define the $n$-th shattering coefficient of a strategic classification problem as
\[\sigma_n(\mathcal{H},\Delta)=\max_{(x_1,\dots,x_n)\in\mathcal{X}^n}|\{(h(x_1'),\dots,h(x_n')):h\in\mathcal{H},x_i'\in\Delta_h(x_i)\}|.\]
\noindent Then SVC$(\mathcal{H},\Delta)=\sup\{n\ge 0:\sigma_n(\mathcal{H},\Delta)=2^n\}$.
\end{definition}

\noindent A natural question to ask is whether learning with improvements is ``easier'' than strategic classification. That is, if a concept space $\mathcal{H}$ is learnable w.r.t.\ $\Delta$ and $\mathcal{D}$ in the strategic classification setting, then is it also learnable with improvements? Interestingly, we answer this question in the negative. More precisely, we show that finite SVC (which is known from prior work to be a sufficient condition for  strategic PAC learning) is actually not a  sufficient condition for PAC learnability with improvements.

\begin{theorem}
\label{thm:strategic-easier}
    Finite strategic VC dimension \citep{strategicPAC} does not necessarily imply PAC learnability with improvements. 
\end{theorem}

\begin{proof}
    Let the instance space $\mathcal{X}$ be $[0,1]$, let ${\mathcal H}=\{h_{abcd} : h_{abcd}(x)=1 \mbox{ iff }x\in [a,b)\cup(c,d]\}$, and let $\mathcal{D}$ be the uniform distribution over $[0,1]$. We define $\Delta$ as follows.  For $x\in [0,3/4)$ let $\Delta(x)=\mathcal{B}(x,1/4)=(x-1/4,x+1/4)\cap [0,1]$; for $x\in [3/4,1]$, let $\Delta(x)=\{x\}$. 
    
    We claim that no algorithm with finite training data can guarantee an expected error of less than $1/16$ for the above  when learning with improvements, even though the class is  PAC-learnable in the strategic classification setting. To see the latter, note that SVC$(\mathcal{H},\Delta)\le 4$. Indeed, consider the points $(0,1/4,1/2,3/4,1)\in\mathcal{X}^5$. Notice the (strategic) labeling $(1,0,1,0,1)$ cannot be achieved for any $h\in\mathcal{H}$, which establishes the claim.
    
    Now consider a target function defined as the union of two intervals $[1/2, b) \cup (b, 1]$ where the number $b$ was randomly chosen in $[3/4, 1]$.  The learner will not see the point $b$ given a finite training set, so it learns nothing about the location of $b$ (almost surely).  Now, we consider two cases. Either, the learner outputs a classifier whose positive region has probability mass at most $1/16$ over the interval $[3/4, 7/8]$. Then its error rate is at least $1/16$ because the positive examples in $[3/4,7/8]$ cannot move so at least half of their probability mass will get misclassified.  On the other hand, if the learner outputs a classifier whose positive region has probability mass greater than $1/16$ on the interval $[3/4,7/8]$, then it has at least a $50 \%$ chance of including the negative point $b$ in its positive region (over the random choice of the target function).  If the classifier has a negative point in $[3/4,7/8]$ that is incorrectly predicted to be positive, then it will have an error rate at least $1/16$, because all the positives in $[5/8, 3/4)$ will move to a false positive (here we use that agents break ties adversarially, see also Remark \ref{rmk:misleading-improvement}).  So, either way, its expected error is  $\ge 1/16$.
\end{proof}

\paragraph{Learnability with improvements may be easier than strategic classification.} 
It is not too hard to come up with examples where it is easier to learn in the improvements setting when compared to the strategic setting. Example~\ref{ex:example3.1} shows that it is possible to learn perfectly with improvements (with zero error) in a setting where avoiding a large constant error is unavoidable in the strategic classification setting.
In Example~\ref{ex:example3.2}, although the error might not be zero when learning with improvements, it would be significantly lower than in the strategic classification setting. 

\begin{example}[The ``all-negative'', ``all-positive'', and ``singleton-positive'' classifiers] 
    Define $\Delta(x)=\mathcal{X}$ for all agents $x\in \mathcal{X}$. Suppose the ``all-negative'' classifier $h_-(x)=0$,  the ``all-positive'' classifier $h_+(x)=1$, and all ``singleton-positive'' classifiers $h_{x^{\star}}(x)=\mathbb{I}[x=x^{\star}]$ lie in the concept space $\mathcal{H}$. Select any $f^{\star}\in\mathcal{H}$ and any data distribution $\mathcal{D}$ over $\mathcal{X}$ such that $\mathbb{P}_{x\sim \mathcal{D}}[f^{\star}(x)=0]=\mathbb{P}_{x\sim \mathcal{D}}[f^{\star}(x)=1]=\frac{1}{2}$. Now with $O(\log \frac{1}{\delta})$ examples, the learner sees a positive example, say $x^+$, in its training set with probability $\ge\delta$. Outputting $h_{x^+}$ achieves zero-error in the learning with improvements setting, as all negative points can improve to $x^+$. In contrast, a learner in the strategic classification setting must suffer an error of at least $1/2$ here. Indeed, either the learner outputs $h_-$ and suffers an error of $1/2$ on the positive points. Or, the learner selects an $h$ that labels at least one point as positive and incurs an error $1/2$ on the negative points, all of which  successfully deceive the learner.
\label{ex:example3.1}
\end{example}

\begin{example}[The improvement function is consistent with $f^{\star}$]
    Consider an improvement function $\Delta$ that takes into account $f^{\star}$, such that $f^{\star}(x')=1$ for $x'\in\Delta(x)$ for any $x\in\mathcal{X}$. That is the improvement function is in a certain sense consistent with $f^{\star}$, guaranteeing positive classification after any move. In this setting, any classifier $h$ will have lower error in the improvements setting compared to strategic classification. This is because a negative point that moves and becomes positive is an error in  strategic learning but the point would have genuinely improved in this case.
\label{ex:example3.2}
\end{example}

\paragraph{Learnability with improvements may yield more errors than strategic classification.} 
Contrasting Example~\ref{ex:example3.2} with Remark \ref{rmk:misleading-improvement}, when $\Delta$ does not satisfy the property that the improvement function is consistent with $f^{\star}$, then it is possible to do worse in the improvement setting (e.g.\ Theorem \ref{thm:strategic-easier}) because some true positive examples can potentially become negative when moving in response to a false positive for the learner's hypothesis $h$.

\section{PAC Learning of Geometric Concepts}\label{sec:paclearn_geometric-concepts}
In this section, we first demonstrate the gain of the learner when agents can improve for the natural class of thresholds on the real line, where agents can move by a distance of at most $r$. We then study intersection-closed classes. In particular, we derive sample complexity bounds for the class of axis-aligned hyperrectangles, where the improvement sets are the $\ell_\infty$ balls. We further establish negative results for proper learners in the absence of the intersection-closed property. Lastly, we study the class of homogeneous halfspaces under the uniform distribution over the unit ball, where agents can improve by adjusting their angle. Complete proofs for this section are located in Appendix~\ref{app:proofs-geometric}.\looseness-1

We will use $(a)_+$ to denote $\max\{a,0\}$.

\subsection{Warm-up: Zero  Error for Learning Thresholds}\label{subsec:paclearn_thresholds}
Let \(\mathcal{H} = \{ h_t : t \in \mathbb{R} \}\) be the class of one-sided threshold functions, where $h_t(x) = \mathbb{I}\{x \geq t\}$.

The improvement set of $x$ is simply the closed ball centered at $x$ with radius $r$, i.e., $\Delta(x) =\{ x' \mid \lvert x - x'\rvert \leq r\} $.
Suppose the data distribution $\mathcal{D}$ is uniform over $[0,1]$, and labels are generated according to a target threshold $h_{t^{\star}} \in \mathcal{H}$ for some $t^{\star} \in [0,1]$. Let $S = \{(x_i, y_i)\}_{i=1}^m$ be the set of training samples, where $x_i \overset{\text{i.i.d.}}{\sim} \mathcal{D}$ and $y_i = h_{t^{\star}}(x_i)$. 

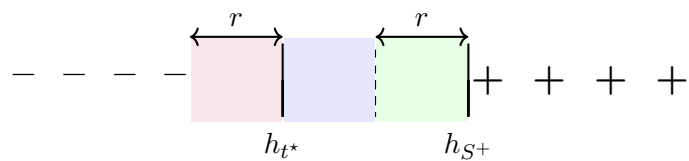
\begin{figure}[!b]
    \centering
    \begin{tikzpicture}[scale=0.88]

        \node (m1) {\Large \textbf{--}};
        \node[right=1mm of m1] (m2) {\Large \textbf{--}};
        \node[right=1mm of m2] (m3) {\Large \textbf{--}};
        \node[right=1mm of m3] (m4) {\Large \textbf{--}};

        \node[rectangle, fill=purple!10, thick, minimum width=1.2cm, minimum height=1.1cm, right=-1.0mm of m4] (r1) {};
        \node[rectangle, fill=blue!10, thick, minimum width=1.2cm, minimum height=1.1cm, right=0mm of r1] (r2) {};
        \node[rectangle, fill=green!10, thick, minimum width=1.2cm, minimum height=1.1cm, right=0mm of r2] (r3) {};

        \coordinate (m1c) at ($(r1)!0.5!(r2)$);
        \draw[thick] (m1c) -- ++(0,-0.55) -- ++(0,1.10);

        \coordinate (m2c) at ($(r2)!0.5!(r3)$);
        \draw[dashed] (m2c) -- ++(0,-0.55) -- ++(0,1.10);

        \coordinate (m3c) at (r3.east);
        \draw[thick] (m3c) -- ++(0,-0.55) -- ++(0,1.10);

        \draw[<->, thick] ([yshift=6.5mm]r1.west) -- ([yshift=6.5mm]r1.east) node[midway, above] {$r$};
        \draw[<->, thick] ([yshift=6.5mm]r3.west) -- ([yshift=6.5mm]r3.east) node[midway, above] {$r$};

        \coordinate (ht) at ($(r1.south)!0.5!(r2.south)$);
        \node[below=0.0mm of ht] {$h_{t^{\star}}$};
        \node[below=0.0mm of m3c] at (m3c |- ht) {$h_{S^+}$};

        \node[right=-1.0mm of r3] (p1) {\Large \textbf{+}};
        \node[right=1mm of p1] (p2) {\Large \textbf{+}};
        \node[right=1mm of p2] (p3) {\Large \textbf{+}};
        \node[right=1mm of p3] (p4) {\Large \textbf{+}};

    \end{tikzpicture}

    \caption{Learning thresholds with improvements.}
    \label{fig:threshold}
\end{figure}

There are several options for choosing a threshold that achieves zero empirical error on $S$, as shown by the shaded area in Figure~\ref{fig:threshold}. Due to the asymmetry of the loss function (Eqn.~\ref{def:loss-function}), we choose the rightmost threshold consistent with $S$. This is the most ``conservative'' option, as any $x$ that improves up to this threshold is guaranteed to be positive with respect to the unknown ground-truth $h_{t^{\star}}$.  This is a property that would not necessarily hold for lower thresholds. We define this threshold with respect to $S$ as follows,

\[t_{S^+} = 
    \begin{cases} 
    \min(S^+), & \text{if } S^+ \neq \emptyset, \\
    1, & \text{if } S^+ = \emptyset,
    \end{cases}
\]
where \(S^+ = \{x_i \in S : y_i = 1\}\) is the set of positive examples in \(S\).
The hypothesis \(h_{S^+}\) is defined as $h_{S^+} = \mathbb{I}\{x \geq t_{S^+}\}$.

Notice that using classifier $h_{S^+}$ will induce agents (at test time) $x\in [t_{S^+}-r,t_{S^+})$ 
to improve to be classified as positive by $h_{S^+}$, which will be a correct classification since $t_{S^+} \geq t^{\star}$.

\begin{theorem}[Thresholds, uniform distribution]
    Let the improvement set $\Delta$ be the closed ball with radius $r$, 
    $\Delta(x) =\{ x'\mid \lvert x - x' \rvert \leq r\} $.
    Let $\mathcal{D}$ be the uniform distribution on $[0,1]$. For any \(\epsilon,\delta \in (0,1/2)\), with probability $1-\delta$,
    \begin{align}
       \mathsc{Loss}_{\mathcal{D}}(h_{S^+},h_{t^{\star}})
       \leq (\epsilon -r)_+,
    \end{align}
    with sample complexity $M = O\left(\frac{1}{\epsilon} \log \frac{1}{\delta}\right).$
\label{thm:thresholds-uniform}
\end{theorem}
\begin{proof}
    See Appendix~\ref{app:thresholds-uniform} 
\end{proof}

Note that the population error is improved from $\epsilon$ (in the standard PAC setting) to $\epsilon-r$ for the same sample size, and we can achieve zero error as long as we set $\epsilon\leq r$.

\paragraph{Learning thresholds with an arbitrary distribution.} We prove a similar result for arbitrary distribution $\mathcal{D}$, where instead of getting $\epsilon-r$ population error, the reduction in the error  
is replaced by the following distribution-dependent quantity
    \begin{align}\label{eq:IR-threshold}
        p(h_{S^+};h_{t^{\star}},\mathcal{D},r)= \mathbb{P}_{x\sim \mathcal{D}}\left[x\in[t_{S^+}-r,t_{S^+}]\right].
    \end{align}
See Theorem~\ref{thm:threshold-arbitrary-D} in Appendix~\ref{app:thresholds-arbitrary}.

\paragraph{Intersection-closed under max.}
Note that the class of thresholds is closed under intersection: 
$
\bigcap_{i=1}^n h_{t_i} = h_{\max\{t_1, t_2, \dots, t_n\}}.
$
In the following, we extend the analysis to such hypothesis classes, more generally in Section~\ref{sec:paclearn_intersection-closed}.

\subsection{Intersection Closed Classes}\label{sec:paclearn_intersection-closed}

The learnability of intersection-closed hypothesis classes in the standard PAC model has been extensively studied \citep{Helmbold1990,auer1997learning,auer1998line,closureAlgo2007,darnstadt2015optimal}. In this section, we study  the learnability with improvements of these classes. 
We start with the following definitions. 

\begin{definition}[Closure operator of a set]
    For any set $S\subseteq \mathcal{X}$ and any hypothesis class $\mathcal{H}\subseteq 2^\mathcal{X}$, the \emph{closure of $S$ with respect to $\mathcal{H}$}, denoted by $\clos_\mathcal{H}(S):2^\mathcal{X}\rightarrow 2^\mathcal{X}$, is defined as the intersection of all hypotheses in $\mathcal{H}$ that contain $S$, that is, $\clos_{\mathcal{H}}(S)=\underset {h\in {\mathcal{H}}, S\subseteq h}{\bigcap} h$. In words, the closure of $S$ is the smallest hypotheses in ${\mathcal{H}}$ which contains $S$. If $\lrset{h:{\mathcal{H}}: S\subseteq h}=\emptyset$, then $\clos_{\mathcal{H}}(S)=\mathcal{X}$.
\end{definition}

\begin{definition}[Intersection-closed classes]
    \label{def:intersection-closed}
    A hypothesis class $\mathcal{H}\subset 2^\mathcal{X}$ is \emph{intersection-closed} if for all finite $S\subseteq \mathcal{X}$, $\clos_{\mathcal{H}}(S)\in \mathcal{H}$. In words, the intersection of all hypotheses in $\mathcal{H}$ containing an arbitrary subset of the domain belongs to $\mathcal{H}$. For finite hypothesis classes, an equivalent definition states that for any $h_1,h_2\in \mathcal{H}$, the intersection $h_1\cap h_2$ is in $\mathcal{H}$ as well \citep{natarajan1987learning}.
\end{definition}
Many natural hypothesis classes are intersection-closed, for example, axis-parallel $d$ dimensional hyperrectangles, intersections of halfspaces, $k$-CNF boolean functions, and subspaces of a linear space.

The \textit{Closure algorithm} is a learning algorithm that generates a hypothesis by taking the closure of the positive examples in a given dataset, and negative examples do not influence the generated hypothesis. The hypothesis returned by this algorithm is always the smallest hypothesis containing all of the positive examples seen so far in the training set.
\begin{definition}[Closure algorithm \citep{natarajan1987learning,Helmbold1990}]
    Let \\$S=\{(x_1,f^{\star}(x_1)), \ldots, (x_m, f^{\star}(x_m))\}$ be a set of labeled examples, where $f^{\star}\in {\mathcal{H}}$,  \(x_i \in \mathcal{X}\) and \(y_i \in \{0, 1\}\). The hypothesis \(h^c_S\) produced by the closure algorithm is defined as:\looseness-1
    \[
    h^c_S(x) =
    \begin{cases}
    1, & \text{if } x \in \clos_{\mathcal{H}}\left(\{x_i \in S : y_i = 1\}\right), \\
    0, & \text{otherwise}.
    \end{cases}
    \]
    Here, $\clos_{\mathcal{H}}\left(\{x_i \in S : y_i = 1\}\right)$ denotes the closure of the set of positive examples in $S$ with respect to ${\mathcal{H}}$.\label{def:closure-algorithm}
\end{definition}

The closure algorithm learns intersection-closed classes with VC dimension $d$ with an optimal sample complexity of $\Theta\left(\frac{1}{\epsilon}(d+\log \frac{1}{\delta})\right)$ \citep{closureAlgo2007,darnstadt2015optimal}.

We apply the closure algorithm for learning with improvements. In order to quantify the improvement gain of the returned hypothesis, we define the \emph{improvement region} of $h$ as the set of points that can improve from a negative to a (correct) positive classification by $h$.

\begin{definition}[Improvement region] 
The improvement region of hypothesis $h\subseteq f^{\star}$, w.r.t. $f^{\star}$ and $\Delta$ is
    \begin{align}\label{def:IR}
    \begin{split}
        \ir(h;f^{\star},\Delta)
        \coloneq
        \lrset{x:h(x)=0, \exists x'\in\Delta(x):h(x')=f^{\star}(x')=1}.
    \end{split}
    \end{align}
    
    The gain from improvements is the probability mass of the improvement region under $\mathcal{D}$: \\ $\mathbb{P}_{x \sim \mathcal{D}}\left[x \in \ir(h;f^{\star},\Delta) \right].$
    
\end{definition}
Note that for the class of thresholds, the closure algorithm returns exactly the hypothesis $h_{S^+}$, 
and the probability mass of the improvement region is $p(h_{S^+};h_{t^{\star}},\mathcal{D},r)$ (cf.\ Eqn.~\ref{eq:IR-threshold}).

\paragraph{Axis-Aligned  Hyperrectangles in $[0,1]^d$.}
An axis-aligned hyperrectangle classifier assigns a value of 1 to a point if and only if the point lies within a specific rectangle. 
Formally, let $a=(a_1,\ldots,a_d),b=(b_1,\ldots,b_d)\in[0,1]^d$ where $a_i\leq b_i$ for $i\in\{1,\ldots,d\}\coloneq[d]$. 
A hyperrectangle 
$R_{(a,b)}=\prod_{i\in [d]}[a_i,b_i]$
classifies a point $x=(x_1,\ldots,x_d)$ as: 
$ R_{(a,b)}(x)=\mathbb{I}\{x_i
\in [a_i,b_i],\; \forall i \in [d]\}$.

In the following, we show that the closure algorithm learns with improvements the hypothesis class $\mathcal{H}_{\text{rec}}=\{R_{(a,b)}: a,b\in[0,1]^d\}.$

\begin{theorem}[Axis-aligned Hyperrectangles]
    Let the improvement set $\Delta$ be the closed $\ell_\infty$ ball with radius $r$, $\Delta(x) =\{ x'\mid \norm{x - x'}_\infty \leq r\} $.
    Let $R_S^c$ be the rectangle returned by the closure algorithm given $S\overset{\text{i.i.d.}}{\sim} \mathcal{D}^m$, and $R^{\star}$ be the target rectangle. For
    any distribution $\mathcal{D}$, for any \(\epsilon,\delta \in (0,1/2)\), with probability $1-\delta$,
    
    \begin{align}
       \mathsc{Loss}_{\mathcal{D}}(R_S^c,R^{\star})
       \leq \left(\epsilon - \mathbb{P}_{x \sim \mathcal{D}}\left[x \in \ir(R_S^c;R^{\star},\Delta) \right]\right)_+,
    \end{align}
    with sample complexity $M = O\left(\frac{1}{\epsilon}\left(d+ \log \frac{1}{\delta}\right)\right)$. 
    
    When $\mathcal{D}$ is the uniform distribution on $[0,1]^2$, we can get the following expression.
    Denote by $l_1$ and $l_2$ the width and height (respectively) of the rectangle $R_S^c$. Then, 
    \begin{align}
        \mathbb{P}_{x \sim \mathcal{D}}\left[x \in \ir(R_S^c;R^{\star},\Delta) \right] = 2r(l_1+l_2)+4r^2.
    \end{align}
\label{thm:rectangles}
\end{theorem}

\begin{proof}
    See Appendix~\ref{app:rectangles}
\end{proof}
Note that, as opposed to the simple case of thresholds,  the improvement region for hyperrectangles depends on the geometry of the target hypothesis.

\paragraph{Arbitrary Intersection-closed Classes.} We will now show that any intersection-closed concept class with a finite VC dimension is PAC learnable with improvements \emph{w.r.t. any} improvement function $\Delta$.

\begin{theorem}
    Let $\mathcal{H}$ be an intersection-closed concept class on instance space $\mathcal{X}$. There is a learner that PAC-learns with improvements $\mathcal{H}$ with respect to any improvement function $\Delta$ and any data distribution $\mathcal{D}$ given a sample of size $O\left(\frac{1}{\epsilon}(d_{\text{VC}}(\mathcal{H})+\log\frac{1}{\delta})\right)$, where $d_{\text{VC}}(\mathcal{H})$ denotes the VC-dimension of $\mathcal{H}$.

\label{thm:intersection-closed}
\end{theorem}

\begin{proof}
    Let $S\sim\mathcal{D}^m$ and $h^c_S$ denote the classifier learned by the closure algorithm (Definition~\ref{def:closure-algorithm}). For some $m=O(\frac{1}{\epsilon}(d_{\text{VC}}(\mathcal{H})+\log\frac{1}{\delta}))$, we know from prior work \citep{closureAlgo2007,darnstadt2015optimal} that $h^c_S$ satisfies, with probability at least $1-\delta$,
    \[\mathbb{P}_{x\sim \mathcal{D}}[h^c_S(x)\ne f^{\star}(x)]\le \epsilon,\]
    for any target concept $f^{\star}\in \mathcal{H}$.

    Now for any $x\in\mathcal{X}$ if $h^c_S(x)=1$, the improvement loss $\mathsc{Loss}(x; h^c_S, f^{\star})=\mathbb{I}[h^c_S(x)\ne f^{\star}(x)]=0$ since $\Delta_{h^c_S}(x)=\{x\}$ and $f^{\star}(x)=1$ since $h^c_S$ is obtained using the closure algorithm. If $h^c_S(x)=0$ and if $\Delta_{h^c_S}(x)\ne\{x\}$, for any point $x'\in\Delta_{h^c_S}(x)$, we have $h^c_S(x')=1=f^{\star}(x')$ and therefore $\mathsc{Loss}(x; h^c_S, f^{\star})=0$. So the only points for which $h^c_S$ can make a mistake are points where $h^c_S(x)=0$ and $\Delta_{h^c_S}(x)=\{x\}$, i.e.\ the points do not move in reaction to ${h^c_S}$. This implies $h^c_S$ must disagree with $f^{\star}$ on these points also in the PAC setting. But the probability mass of these points is at most $\epsilon$ as noted above.
\end{proof}

\paragraph{Hardness of proper learning in the absence of the intersection-closed property.} We can also establish the following negative result which indicates the hardness of proper learning in the absence of the intersection-closed property.

\begin{theorem}
    Let $\mathcal{H}$ be any concept class on a finite instance space $\mathcal{X}$ 
    such that at least one point $x'\in\mathcal{X}$ is classified negative by all $h\in \mathcal{H}$ (i.e.\ $\{x\mid h(x)=0 \text{ for all } h\in \mathcal{H}\}\ne\emptyset$), and suppose $\mathcal{H}\mid_{\mathcal{X}\setminus\{x'\}}$ is not intersection-closed on $\mathcal{X}\setminus\{x'\}$. Then there exists a data distribution $\mathcal{D}$ and an improvement function $\Delta$ such that no proper learner can PAC-learn with improvements $\mathcal{H}$ w.r.t.\ $\Delta$ and $\mathcal{D}$.
    \label{thm:hardness-intersection-closed}
\end{theorem}

\begin{proof}
    See Appendix~\ref{app:hardness-intersection-closed}
\end{proof}

\paragraph{Absence of intersection-closed property and a negative point all \(h\in \mathcal{H}\) agree on.}
Note that in Theorem~\ref{thm:hardness-intersection-closed}, we have an additional requirement that all classifiers in the concept space agree on some negative point (intuitively, agents who should never achieve positive classification). Consider the following simple example, Example~\ref{ex:examplex}, where this condition does not hold, the concept class is not intersection-closed, and learnability is possible in our setting.

\begin{example}[Absence of intersection-closed property and a negative point all \(h\in \mathcal{H}\) agree on]
    Suppose $\mathcal{X}=\{x_1,x_2\}$ and $\mathcal{H}=\{h_1,h_2\}$ with $h_1(x_1)=1,h_1(x_2)=0$ and $h_2(x)=1-h_1(x)$ for either $x\in\mathcal{X}$. Clearly $h_1\cap h_2\notin \mathcal{H}$ and  $\mathcal{H}$ is not intersection-closed, yet knowledge of a single label tells us the target concept.
\label{ex:examplex}
\end{example}

\subsection{Halfspaces on the Unit Ball}

We now consider the problem of learning homogeneous halfspaces with respect to the uniform distribution on the unit ball (or any spherically-symmetric distribution), when agents have the ability to improve by an angle of $r$.
\begin{theorem}\label{thm:halfspaces}
 Consider the class of $d$-dimensional halfspaces passing through the origin, i.e., $\mathcal{H} = \{x \mapsto \operatorname{sign}(w^Tx):w \in \R^d\}$. Suppose \( \mathcal{X} \) is the surface of the origin-centered unit sphere in \( \mathbb{R}^d \) for \( d > 2 \), and \( \mathcal{D} \) is the uniform distribution on \( \mathcal{X} \). 
For each point \( x \in \mathcal{X} \), define its neighborhood \( \Delta(x) = \{ x' \mid \arccos(\langle x, x' \rangle) \leq r \} \). For any \(\epsilon,\delta \in (0,1/2)\), and training sample $S \overset{\text{i.i.d.}}{\sim} \mathcal{D}^m$ of size $\tilde{O}\left(\frac{d + \log\frac{1}{\delta}}{r}\right)$,  with probability $1-\delta$,
$\mathsc{Loss}_{\mathcal{D}}({\textrm POS}_{\mathcal{H}}(S),f^{\star})=0,$ where ${\textrm POS}_{\mathcal{H}}(S)$ is the  intersection of the positive regions of all $h \in \mathcal{H}$ consistent with the training set $S$.

\end{theorem}

\begin{proof}
The algorithm will use ${\textrm POS}(\mathcal{H}_S)$ as its classifier; that is, a point $x'$ is classified as positive if \emph{every} $h\in \mathcal{H}$ consistent with $S$ labels $x'$ as positive.  Note that this classifier will have zero loss under improvement function $\Delta$ if every $h\in \mathcal{H}$ consistent with $S$ has angle at most $r$ with $f^{\star}$; this is because any positive example $x$ can then move into the positive agreement region simply by moving an angular distance $r$ in the direction of the normal vector to $f^{\star}$.  So, all that remains is to show that after $m$ examples, with probability at least $1-\delta$, every $h\in \mathcal{H}$ consistent with $S$ has angle at most $r$ with $f^{\star}$.

Consider some $h \in \mathcal{H}$ given by \( h(x) = \operatorname{sign}(w^Tx) \).
The probability mass lying in the disagreement region of $h$ and $f^{\star}$ is:
\begin{align}
\rho_{\mathcal{D}} (h,f^{\star}) = \mathbb{P}_{x \sim \mathcal{D}}[h(x) \neq f^{\star}(x)] = \frac{\arccos(\langle w, w^{\star} \rangle)}{\pi}.
\label{unitballactive}
\end{align}
Therefore, to have the property that every $h\in \mathcal{H}$ consistent with $S$ has angle at most $r$ with $f^{\star}$, we just need that every $h\in \mathcal{H}$ of error greater than $\frac{r}{\pi}$ should make at least one mistake on $S$.  By standard realizable VC bounds, it suffices to have
\[
m =  \mathcal{\tilde{O}}\left(\frac{d+\log\frac{1}{\delta}}{r}\right)
\]
number of i.i.d. samples for this to hold with probability at least $1-\delta$.
\end{proof}

\begin{remark}
  It is worth noting that while the aforementioned result relies on the classifier ${\textrm POS}(\mathcal{H}_S)$, which is a fairly complex function, a similar guarantee can be achieved using a linear classifier (though non-homogeneous, so it is still not ``proper''). Specifically, by obtaining a sufficiently large sample, one can construct a homogeneous linear classifier whose angle with respect to the target is at most \( \frac{r}{2} \). We can then shift that classifier by $r/2$ (so it is no longer homogeneous) to ensure its positive region is contained inside the positive region for $f^{\star}$.
\end{remark}

\section{Zero-error Learning in the Graph Model}\label{sec:paclearn_graph-model}

In this section, we will consider a general discrete model for studying classification of agents with the ability to improve. The agents are located on the nodes of an undirected graph, and the edges determine the improvement function, i.e.\ the agents can move to neighboring nodes in order to potentially improve their classification. Note that the graph nodes correspond to an arbitrary discrete instance space $\mathcal{X}$. Remarkably, zero error may be attained even in this general setting. All proofs in this Section are deferred to Appendix~\ref{app:graph}.\looseness-1

Formally, let \( G = (V, E) \) denote an undirected graph. The vertex set \(V = \{x_1, x_2, \dots, x_n\}\) represents a fixed collection of \(n\) points corresponding to a finite instance space $\mathcal{X}$. 
The edge set \( E \subseteq V \times V \) captures the adjacency information relevant for defining the improvement function. More precisely, for a given vertex \( x \in V \), the improvement set of \( x \) is given by its neighborhood in the graph, i.e.\
$\Delta(x) = \{x' \in V \ | \ (x, x') \in E\}$\footnote{Our results readily extend to $\Delta(x) = \{x' \in V \ | \ d_G(x, x') \le r\}$, where $d_G$ denotes the shortest path metric on $G$, by applying our arguments to $G^r$, the $r^\textrm{th}$ power of $G$ (see appendix).}. 
Let \( f^{\star}: V \to \{0, +1\} \) represent the target labeling (or partition) of the vertices in the graph \( G \). Assume that 
the hypothesis space \( \mathcal{H} \) is the set of all possible labelings of the graph, which is finite. 
\subsection{Near-tight Sample Complexity for Zero-Error}

Our first result is to show that we can obtain zero-error in the learning with improvements setting, when the data distribution $\mathcal{D}$ is given by a uniform distribution over $V$, and obtain near-tight bounds on the sample complexity. Our learner in this case is the ‘‘conservative'' classifier $h\in\mathcal{H}$ that classifies exactly the positive points seen in the sample as positive, and the remaining points as negative. Even though we allow $f^{\star}$ to be an arbitrary labeling in $\mathcal{H}$, we do not need to see all the labels to learn an $h$ that achieves zero error w.r.t.\ $f^{\star}$. Intuitively, this is because for any positively labeled node $x$ it is sufficient to see the label of $x$ or that of one of its neighbors since the agents can move to a neighbor predicted positive by $h$. We further show that no algorithm can achieve a better sample complexity, up to some logarithmic factors.

\begin{theorem}
Let \( G = (V, E) \) be an undirected graph with \( n = |V| \) vertices, and let \( f^{\star}: V \to \{0, +1\} \) denote the ground truth labeling function. 
Let \( d_{\min}^+ \) denote the minimum degree of the vertices in \( G^+ \), the induced subgraph of $G$ on the vertices  $x\in V$ with $f^{\star}(x) = 1$. Assume that the data distribution \(\mathcal{D}\) is uniform on \( V \). For any $\delta>0$, 
and training sample $S \overset{\text{i.i.d.}}{\sim} \mathcal{D}^m$ of size  $m =  O \left(\frac{n (\log n + \log \frac{1}{\delta})}{d_{\min}^+ + 1}\right)$, there exists a learner that achieves zero generalization error, i.e.\ learns a hypothesis $h$ such that $\mathsc{Loss}_{\mathcal{D}}(h,f^{\star})=0$, with probability at least $1-\delta$ over the draw of $S$. Moreover, there exists a graph $G$ for which any learner that achieves zero generalization error must see at least $\Omega\left(\frac{n}{d_{\min}^+ + 1} \log \frac{n}{d_{\min}^+ + 1}\right)$ labeled points in the training sample, with high constant probability.
\label{DSSamplecomplexity}
\end{theorem}

\begin{proof} Given a sample $S$ labeled by $f^{\star}$, let $S^+=\{x \in S \mid f^{\star}(x) = +1\}$ denote the set of positive points in $S$. To achieve the claimed upper bound on the sample complexity of learning with zero-error, the learner outputs $h_S$ with $h_S(x)=\mathbb{I}\{x\in S^+\}$, that is the classifier which positively classifies exactly the points in $S^+$. We will now show that with a sample of size $m=O \left(\frac{n (\log n + \log \frac{1}{\delta})}{d_{\min}^+ + 1}\right)$, the proposed $h_S$ achieves zero generalization error with probability at least $1-\delta$.

Let \( V^+ = \{x \in V \mid f^{\star}(x) = +1\} \) denote the set of vertices in $G^+$. We say that $x\in V^+$ is \emph{covered} by the sample $S$ if $x\in S$ or there exists $x'\in S$ such that $x'\in V^+$ and $(x,x')\in E$. Note that if every $x\in V^+$ is covered by $S$, then  $\mathsc{Loss}_{\mathcal{D}}(h_S,f^{\star})=0$ (formally established in Theorem~\ref{thm:dominating-set-teaching}). It is therefore sufficiently to bound determine the sample size needed to guarantee that every positive vertex is covered with high probability. 

Let \( x \in V^+ \) be a vertex in the positive subgraph \( G^+ \). For \( x \) to be covered, it must either be included in the sample $S$, or have at least one of its neighbors \( x' \in \Delta(x) \) included in $S$. The probability of sampling \( x \) directly in one draw is \( \frac{1}{n} \). The probability of sampling any of its neighbors is proportional to its degree in \( G^+ \). Thus, the total probability of covering \( x \) in one draw is:
\[
p_{\text{cover}}(x) = \frac{1}{n} \cdot \left(1 + d(x)\right),
\]
where \( d(x) \) is the degree of \( x \) in \( G^+ \). Since \( d(x) \geq d_{\min}^+ \), we have:
\[
p_{\text{cover}}(x) \geq \frac{d_{\min}^+ + 1}{n}.
\]

To ensure the desired coverage holds with probability at least \( 1 - \delta \), we analyze the failure probability for a single vertex. The probability that a given vertex \( x \in V^+ \) is not covered after \( m \) samples is
\[
\mathbb{P}[x \text{ is not covered}] \leq \left(1 - \frac{d_{\min}^+ + 1}{n}\right)^m.
\]

\noindent To ensure that this holds for all \( |V^+| \leq n \) vertices, we apply the union bound
\begin{align}
    \mathbb{P}[\exists x \in V^+ \text{ not covered}] \leq n \left(1 - \frac{d_{\min}^+ + 1}{n}\right)^m.
\end{align}

\noindent Therefore, the sample size \( m \) required to ensure that the probability of the above bad event is at most $\delta$ is given by 
\begin{align}
    m = O\left(\frac{n (\log n + \log \frac{1}{\delta})}{d_{\min}^+ + 1}\right).
\end{align}

To establish the lower bound, consider a graph \( G \) on \( n \) vertices 
consisting of \( k \) disjoint cliques, each of the same size \( \frac{n}{k} = d_{\min}^+ + 1 \), see Figure~\ref{posclique}. Now note that if our sample $S$ does not contain any node from any one of the cliques (say $C$), then zero-error is not possible. This is because, one of two cases occur. If the learner's hypothesis $h$ predicts any point in $C$ as positive then we can select an $f^{\star}$ that predicts $C$ entirely as negative while being consistent with $S$, causing the population loss to be at least $\frac{1}{k}$. On the other hand, if the learned $h$ predicts all points in $C$ as negative, then we can set $f^{\star}$ to label $C$ entirely as positive, again incurring a population loss of at least $\frac{1}{k}$.

\begin{figure}[ht]
    \begin{center}
    \centerline{\includegraphics[width=0.4\columnwidth]{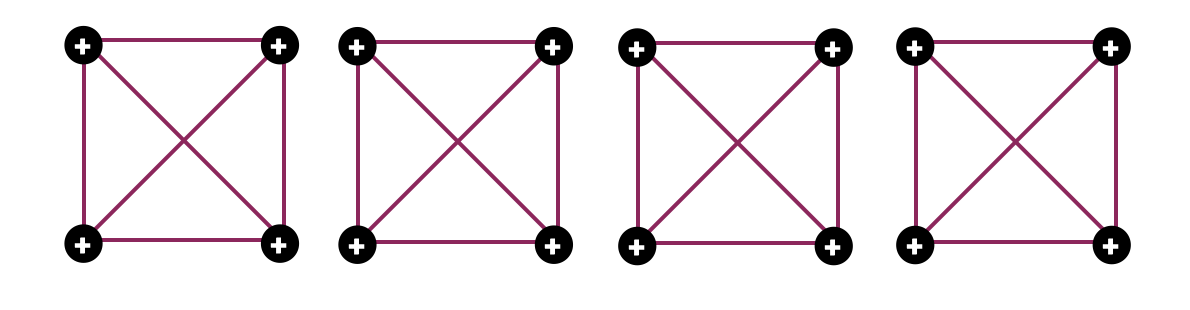}}
    \caption[The graph $G$ used to establish our lower bound]{The graph $G$ used to establish our lower bound on the zero-error sample complexity. The graph consists of $k$ components, each of size $\frac{n}{k}$.}
    \label{posclique}
    \end{center}
\end{figure}

Our goal therefore is to determine a lower bound on the number of points required to ensure that every clique has at least one of its vertices included in the training sample $S$, which ensures that for every positive vertex, either the vertex itself or one of its neighbors is included. Using the standard coupon collector analysis, the number of trials needed to collect $k = \frac{n}{d_{\min}^+ + 1}$ coupons is $\Omega(k\log k)$ with high constant probability.
\end{proof}

We note that the value of \( d_{\min}^+ \) (and, therefore our bound on the sample complexity) is generally not known to the learner in advance, and it depends on the graph structure as well as the (unknown) target labeling  $f^{\star}$. It is an interesting open question to design a learner that can determine whether a sample of sufficient size has been collected to guarantee zero-error. 

For the special case of  the complete graph,  our sample complexity bound becomes $m=\tilde{O}\left(\frac{n}{n^+}\right)$, where $n^+$ is the number of nodes labeled positive by $f^{\star}$.

\subsection{Enabling Improvement Whenever It Helps}\label{sec:paclearn_enabling-improvement}

Note that our loss function $\mathsc{Loss}(x; h, f^{\star})$ 
penalizes the learner for mistakes w.r.t.\ the target $f^{\star}$ after the agents have potentially reacted to $h$. However, we say nothing about whether a negative point $x$ that truly has the ability to improve and get positively classified (i.e.\ $f^{\star}(x')=1$ for some $x'\in\Delta(x)$) will also be able to do so under our published classifier $h$. 
We will now consider an alternative measure of the performance of $h$ (conceptually captures recall in the improvement setting) which measures the probability mass of the points for which we fail to enable improvement even though it is possible under $f^{\star}$. Formally, we define

\begin{eqnarray}
\label{def:loss-function-enabling}
    \mathsc{Loss}^{\mathsc{e}}_{\mathcal{D}}(h, f^{\star}) 
    =
    \mathbb{P}_{x \sim\mathcal{D}}\left[f^{\star}(x)=0 \land  \mathbb{I}[\Delta_{f^{\star}}(x) = \{x\}] = \mathbb{I}[\Delta_h(x) =\{x\}] \right]. \nonumber
\end{eqnarray}

\noindent That is, we wish to ensure that an agent $x$ with $f^{\star}(x)=0$ and an option to improve to a truly positive point in its reaction set w.r.t.\ $f^{\star}$, will also see some option to improve and get positively classified according to $h$.

In Theorem~\ref{thm:enable-improvement-and-zero-loss} in Appendix~\ref{app:improvwhenhelps}, we obtain near-tight sample complexity bounds for learning a concept $h$ that simultaneously guarantees that $\mathsc{Loss}^{\mathsc{e}}_{\mathcal{D}}(h, f^{\star})=0$ and $\mathsc{Loss}_{\mathcal{D}}(h, f^{\star})=0$ when the data distribution  is uniform over $V$.

\subsection{Teaching a Risk-Averse Student}\label{sec:paclearn_teaching-student}

The theory of teaching \citep{GOLDMAN199520} studies the size of the smallest set of labeled examples needed to guarantee that a unique function in the concept space is consistent with the set (for labels according to any target concept in the space). If the teacher (that knows $f^{\star}$) provides this labeled set, then a student that can do consistent learning (find a concept consistent with training data) will learn the target concept $f^{\star}$.
In our learning with improvements over the graph setting, it is natural to consider a simple variant where the student outputs the most risk-averse concept that only labels positive points seen in the labeled set received from the teacher as positive. Here we will consider the question of the minimum number of labeled examples the teacher needs to provide to the risk-averse student to achieve zero-error in our setting.

Let $G^+$ denote the induced subgraph of $G$ on \( V^+ = \{x \in V \mid f^{\star}(x) = +1\} \), the nodes labeled positive by the target concept $f^{\star}$. We show that it is sufficient for the teacher to present the labels of a \emph{dominating set} of $G^+$ (Definition~\ref{defDS}) for the risk-averse student to learn a zero-error classifier $h$. This observation also motivates and helps establish our learning result (Theorem~\ref{DSSamplecomplexity}). 
\begin{theorem}\label{thm:dominating-set-teaching}
    Let \( G = (V, E) \) be an undirected graph, and  \( f^{\star}\) 
    be the target labeling. Let \( G^+ \) denote the induced subgraph on the vertices  $x\in V$ with $f^{\star}(x) = 1$, and $S^+$ denote the dominating set of $G^+$. Then $\mathsc{Loss}(x; h_{S^+},f^{\star})=0$ for any $x\in V$, where $h_{S^+}(x)=\mathbb{I}\{x\in S^+\}$.\looseness-1 
\end{theorem}

\begin{proof}
    See Appendix~\ref{app:teaching-student}
\end{proof}

\section{Evaluation}\label{sec:paclearn_experiments}
Below, and in Appendix~\ref{app:sec_eval}, we empirically examines various improvement-aware algorithms, specifically practical risk-aversion strategies (loss-based and threshold-based) since risk-averse classifiers perform best according to our theory, which consider agents improving within a limited improvement budget $r$. We also explore whether and under what conditions the model error can be reduced to zero when negatively classified agents can improve within an  $\ell_{\infty}$ ball of radius $r$. Recall that zero-error is a remarkable property achievable in the learning with improvements setting, even for fairly general concepts (cf. Section~\ref{sec:paclearn_graph-model} where the instance space is discrete but the concepts can be arbitrary functions).

\paragraph{Datasets.}  We use three real-world tabular datasets: the Adult UCI \citep{BeckerBK}, the OULAD and the Law school datasets \citep{le2022survey}, and a synthetic \(8\)-dimensional binary classification dataset with class separability \(4\) and minimal outliers, generated using Scikit-learn's \texttt{make\_classification} function \citep{make_classification_scikit_learn}. In each case we train a zero-error model \(f^\star\) on the entire dataset, which we treat as the true labeling function for our experiments.
Let \(\mathcal{S}_{T} = \{(x, y) \mid x \in \mathbb{R}^d, \, y \in \{0,1\}\}\)  represent the dataset (e.g., Adult), where \(x\) is the feature vector and \(y = f^\star(x)\) is the label. For all experiments, we split \(\mathcal{S}_{T}\) into training \(\mathcal{S}_\textrm{train}\) (\(70\%\)) and testing \(\mathcal{S}_\textrm{test}\) (\(30\%\)) subsets. Further dataset details, including improvement features and class distributions, are provided in Appendix~\ref{app:sec_datasets}.

\paragraph{Classifiers.} For each full dataset \(\mathcal{S}_{T}\), we trained a zero-error model \(f^\star\) using decision trees. We trained the decision-maker model  \(h: \mathbb{R}^d \to \{0,1\}\), taking the form of a two-layer neural network, on \(\mathcal{S}_\textrm{train}\) with tuned hyperparameters. To assess the loss function's impact on error drop rate when agents improve, we trained both a standard model with binary cross-entropy (\(\mathcal{L}_{\textrm{BCE}}\)) loss and a risk-averse model with weighted-BCE (\(\mathcal{L}_{\textrm{wBCE}}\)) loss (Equation~\ref{eq:losses}).
\begin{equation}\label{eq:losses}
    \begin{aligned}
    \mathcal{L}_{\textrm{wBCE}} &= - \frac{1}{n} \sum_{i=1}^n \Bigg[w_{\textrm{FP}} (1 - y_{i})\log(1 - \hat{y}_{i}) + w_{\textrm{FN}} y_{i}\log(\hat{y}_{i}) \Bigg]
    \end{aligned}
\end{equation}
where, \(n = \lvert\mathcal{S}_{\textrm{train}}\rvert\), \(y \in \{0, 1\}\) is the true label, \(\hat{y} \in (0,1]\) is the model prediction, and \(w_{\textrm{FP}}\) and \(w_{\textrm{FN}}\) are the false positive and false negative weights, respectively. Setting $w_{\textrm{FP}}=w_{\textrm{FN}}=1$ in $\mathcal{L}_{\textrm{wBCE}}$ recovers $\mathcal{L}_{\textrm{BCE}}$. 
Beyond the \(\mathcal{L}_{\textrm{wBCE}}\) loss function, which penalizes false positives more heavily, we explore another form of risk-averse classification by applying a higher threshold of \(0.9\) (instead of the usual \(0.5\)) to the sigmoid output of the final layer. For further details, refer to Appendix~\ref{app:sec_classifiers}.

\paragraph{Improvement.} Given a trained model function \(h\), a data sample \(x \in \mathcal{S}_{\textrm{test}}\)  with an undesirable model outcome \( h(x) = 0 \), and a subset of improvable features along with a predefined improvement budget \( r \), we use Projected Gradient Descent~\citep{AlexAdversarial} to compute the minimal change within the budget \( r \) required to transform \(x\) into a positive outcome \( h(x') = 1 \). Specifically, we aim to find:
\begin{equation} \label{eq:improve_pgd}
x' = \textrm{Proj}_{\Delta(x)}  \left( x_{(t)} + \alpha \cdot  \textrm{sign}(\nabla_{x_{(t)}} \mathcal{L}(h(x_{(t)}), h(x))) \right)
\end{equation}
\(\textrm{such that} \ h(x') = 1\). Here, \( \nabla \mathcal{L}(h(x_{(t)}), h(x)) \) represents the gradient of the loss function (BCE or wBCE), \( t \) the current iteration, and \( \alpha \) the step size. 
\( \textrm{Proj}_{\Delta(x)} \) denotes the projection of \(x_{(t)}\) onto the \(\ell_\infty\) ball of radius \(r\) centered at \(x\), \(\Delta(x) = \{ x_{(t)} \in \mathbb{R}^{d} : \|x_{(t)} - x\|_{\infty} \leq r \} \). This ensures that updates remain within the \( r \)-ball constraint.
A successful improvement occurs when a negatively classified sample \( x \) 
transforms within the specified budget \(r\) into \( x' \) such that \( h(x') = 1 \) and \( f^\star(x') = 1 \) 
(see Appendix~\ref{app:sec_improve}).

\begin{figure}[!htb]
    \centering
    \begin{subfigure}[t]{0.45\linewidth}
        \centering
        \includegraphics[width=0.88\linewidth]{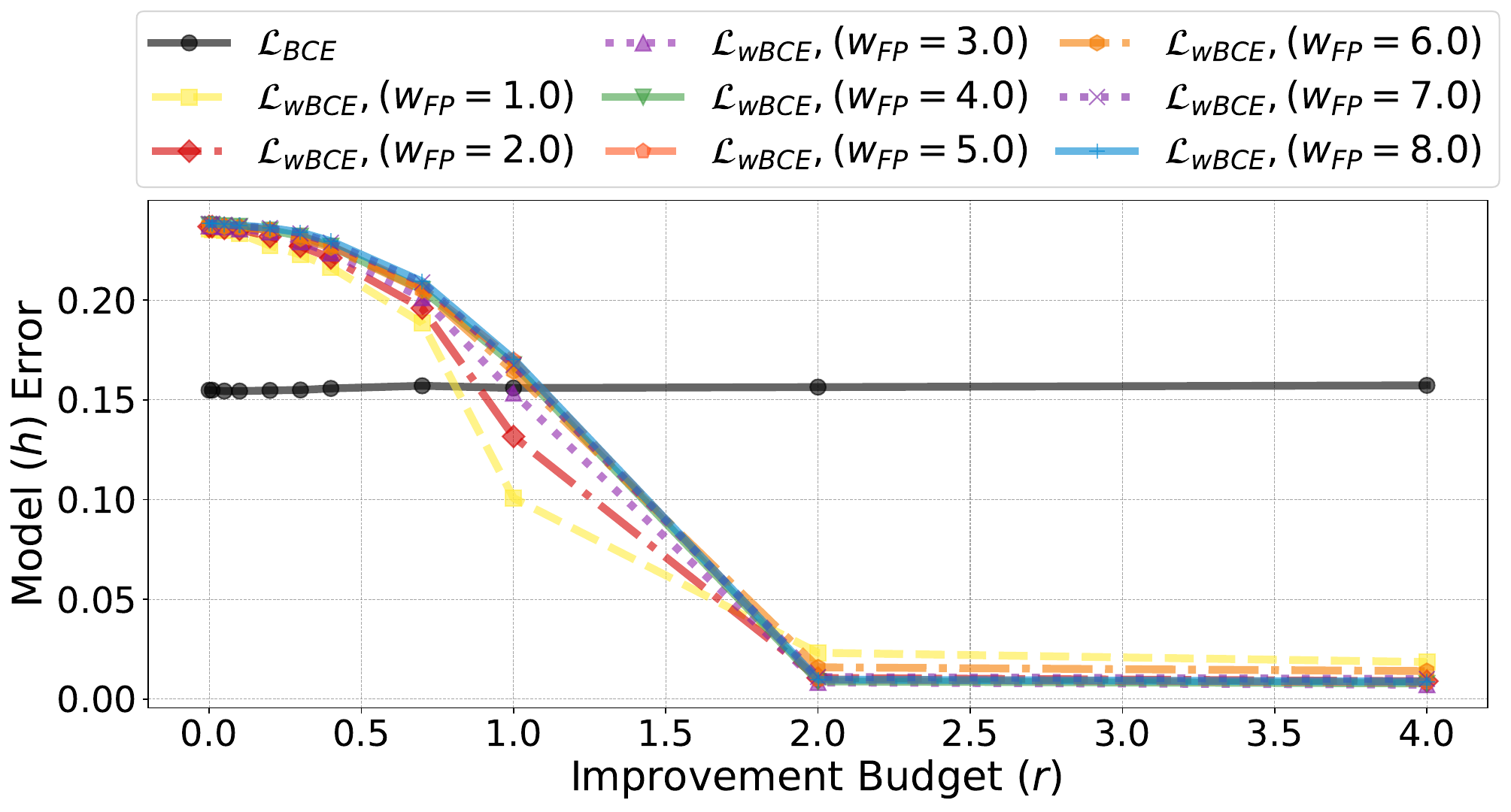}
    \caption{Adult \(\big(w_\textrm{FN}=0.001\big)\)}
    \label{fig:adult_0.5}
    \end{subfigure}
    ~
    \begin{subfigure}[t]{0.45\linewidth}
        \centering
        \includegraphics[width=0.88\linewidth]{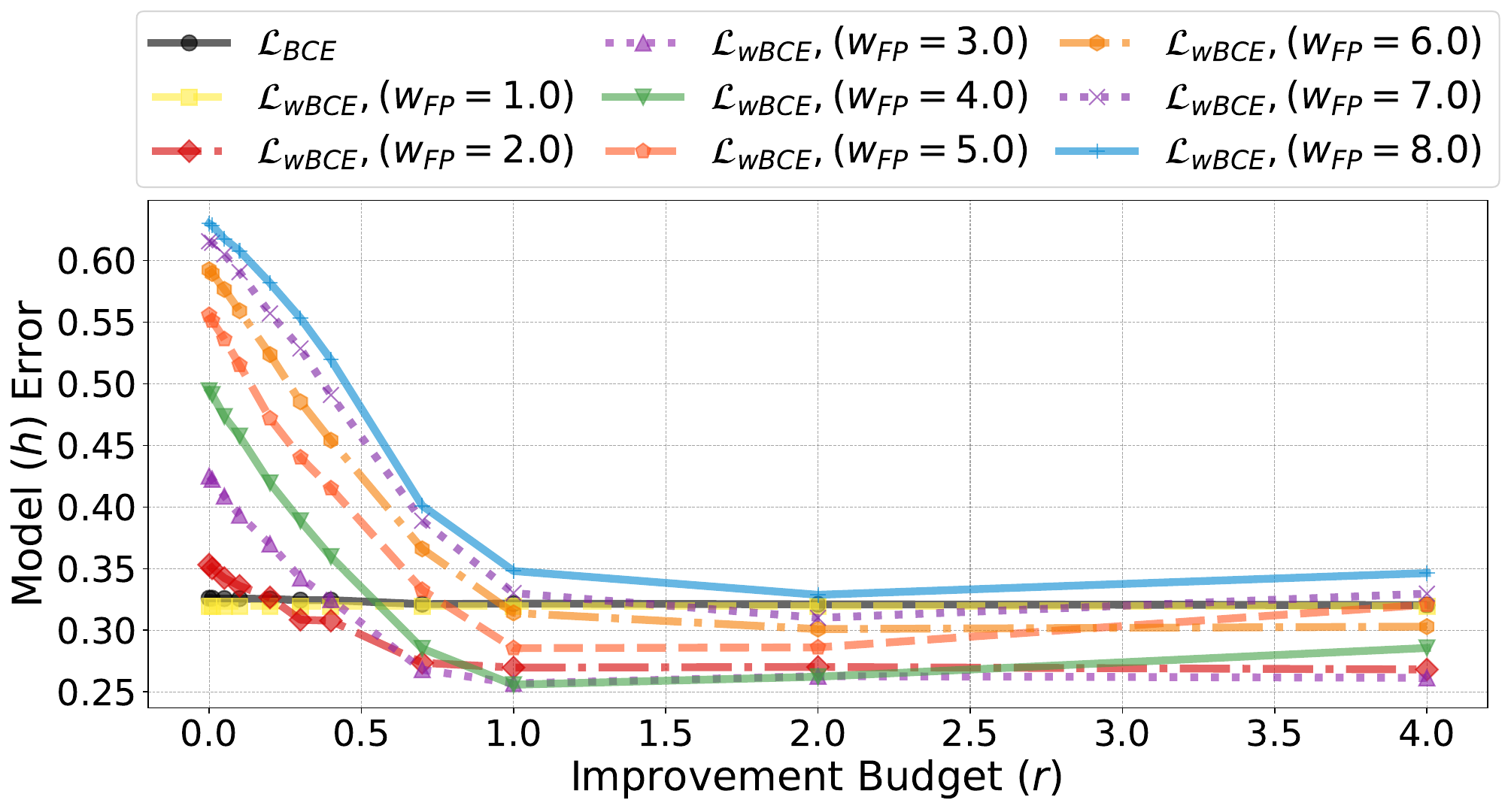}
    \caption{OULAD \(\big(w_\textrm{FN}=1.33\big)\)}
    \label{fig:oulad_0.5}
    \end{subfigure}
    \hfill
    \begin{subfigure}[t]{0.45\linewidth}
        \centering
        \includegraphics[width=0.88\linewidth]{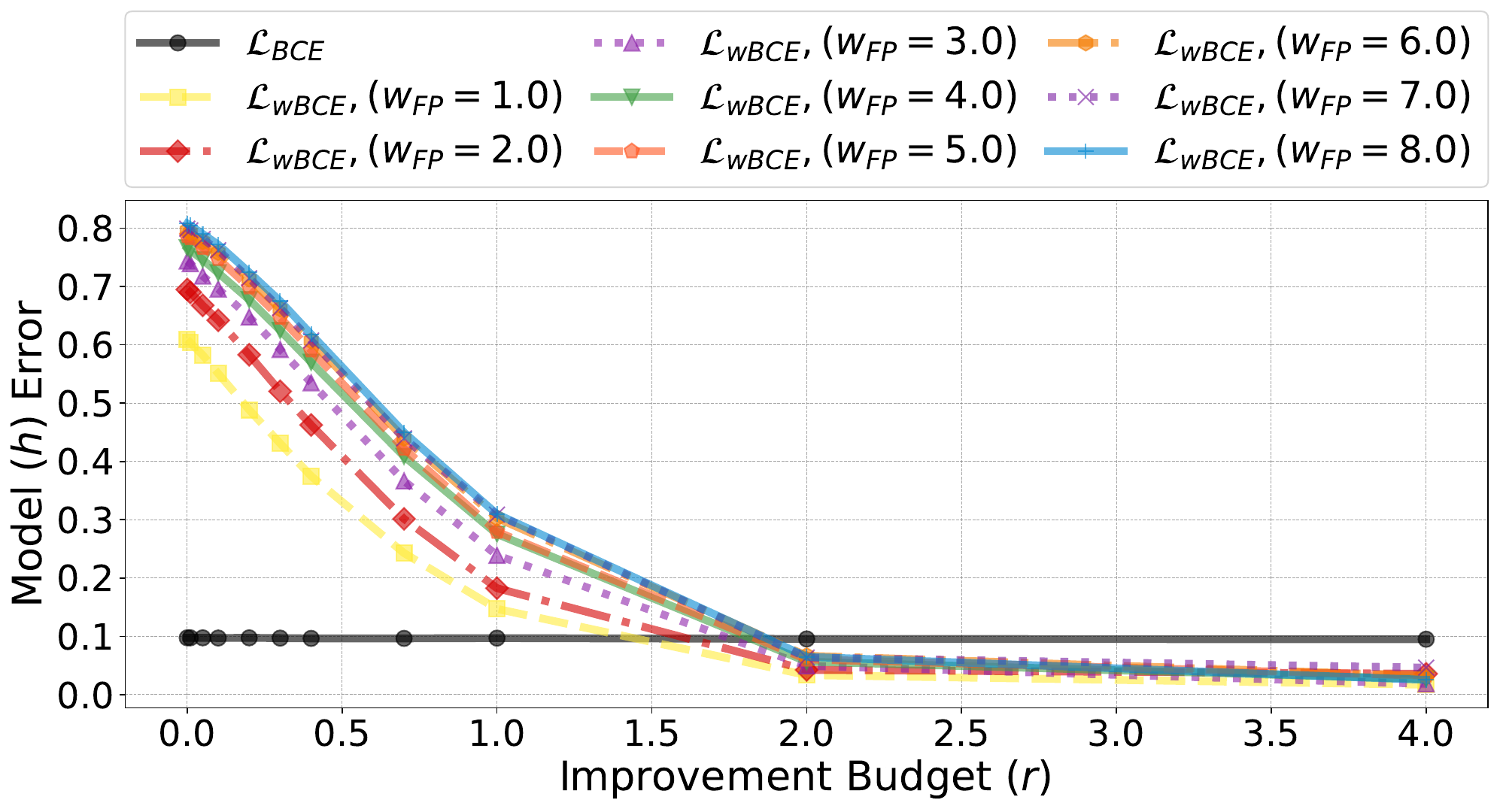}
    \caption{Law school \(\big(\mathcal{L}_\textrm{wBCE} \ \textrm{where} \ w_\textrm{FN}=0.009\big)\)}
    \label{fig:law_0.5}
    \end{subfigure}
    ~
    \begin{subfigure}[t]{0.45\linewidth}
        \centering
        \includegraphics[width=0.88\linewidth]{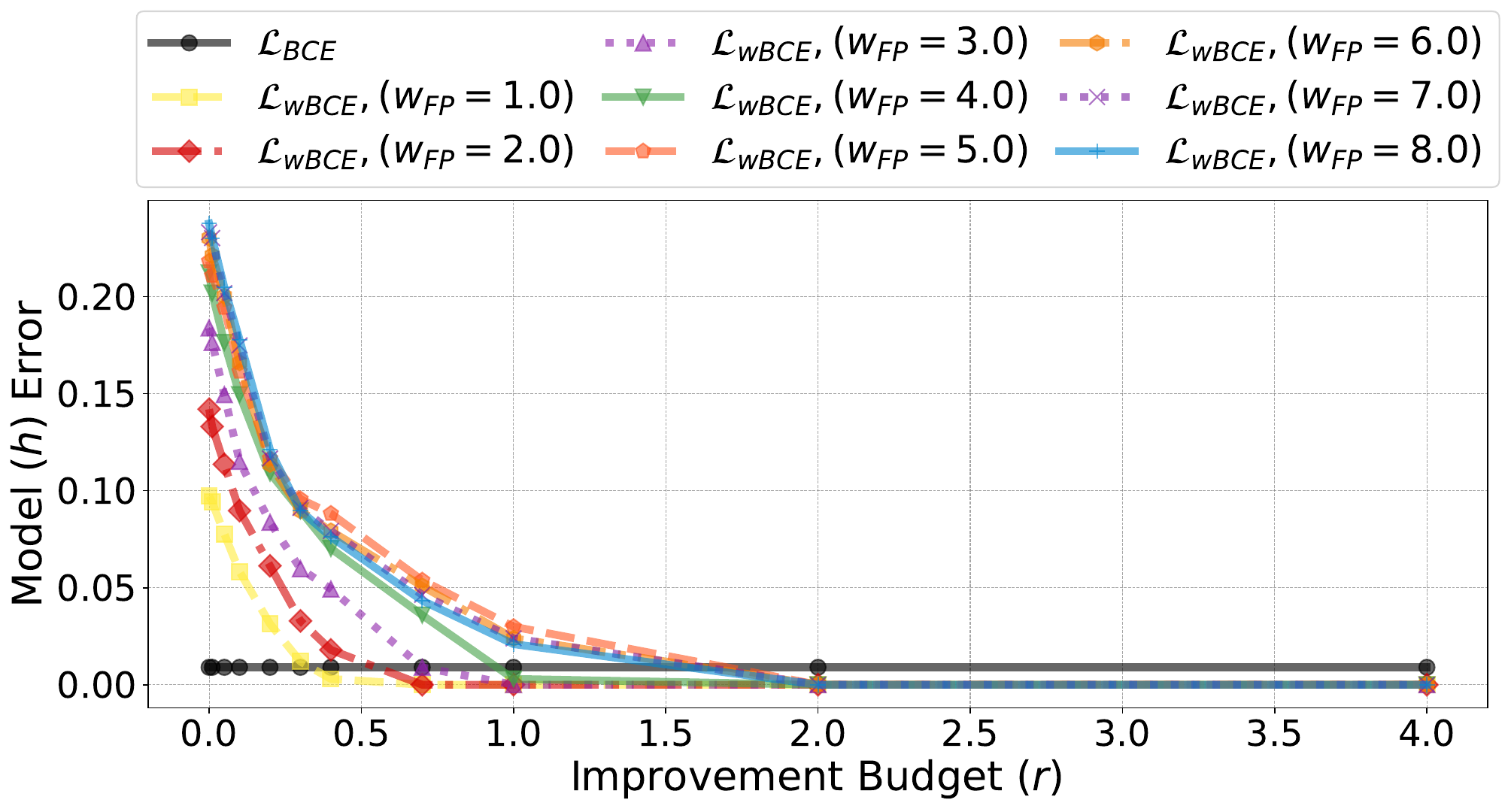}
    \caption{Synthetic \(\big(w_\textrm{FN}=0.009\big)\)}
    \label{fig:synthetic_0.5}
    \end{subfigure}
    
    \caption[We compare the performance gains when agents improve to the risk-averse]{We compare the performance gains when agents improve to the risk-averse \(\big(\mathcal{L}_{\textrm{wBCE}}, \frac{w_\textrm{FP}}{w_\textrm{FN}}>1,  w_{\textrm{FP}}=\{i\}_{i=1}^{8}\big)\) and the standard  (\(\mathcal{L}_\textrm{BCE}, w_{\textrm{FP}}= w_{\textrm{FN}}=1\)) models across four datasets (Adult, OULAD, Law school, and Synthetic) using a fixed classification threshold of \(0.5\). Higher improvement budgets (\(r\)) and greater risk-aversion (high \(\frac{w_\textrm{FP}}{w_\textrm{FN}}\)) accelerate error reduction. See Figure~\ref{fig:app_thresh0.5thresh0.9} (Appendix) for a side-by-side comparison with threshold (\(0.9\)).}
    \label{fig:main_thresh0.5}
\end{figure}
\paragraph{Results.}
Here, we highlight the key insights from our evaluations. A more detailed discussion of the results, along with additional empirical evaluation, can be found in Appendix~\ref{app:sec_results}.

\begin{figure}[!t]
    \centering
    \begin{subfigure}[t]{0.45\linewidth}
        \centering
        \includegraphics[width=\linewidth]{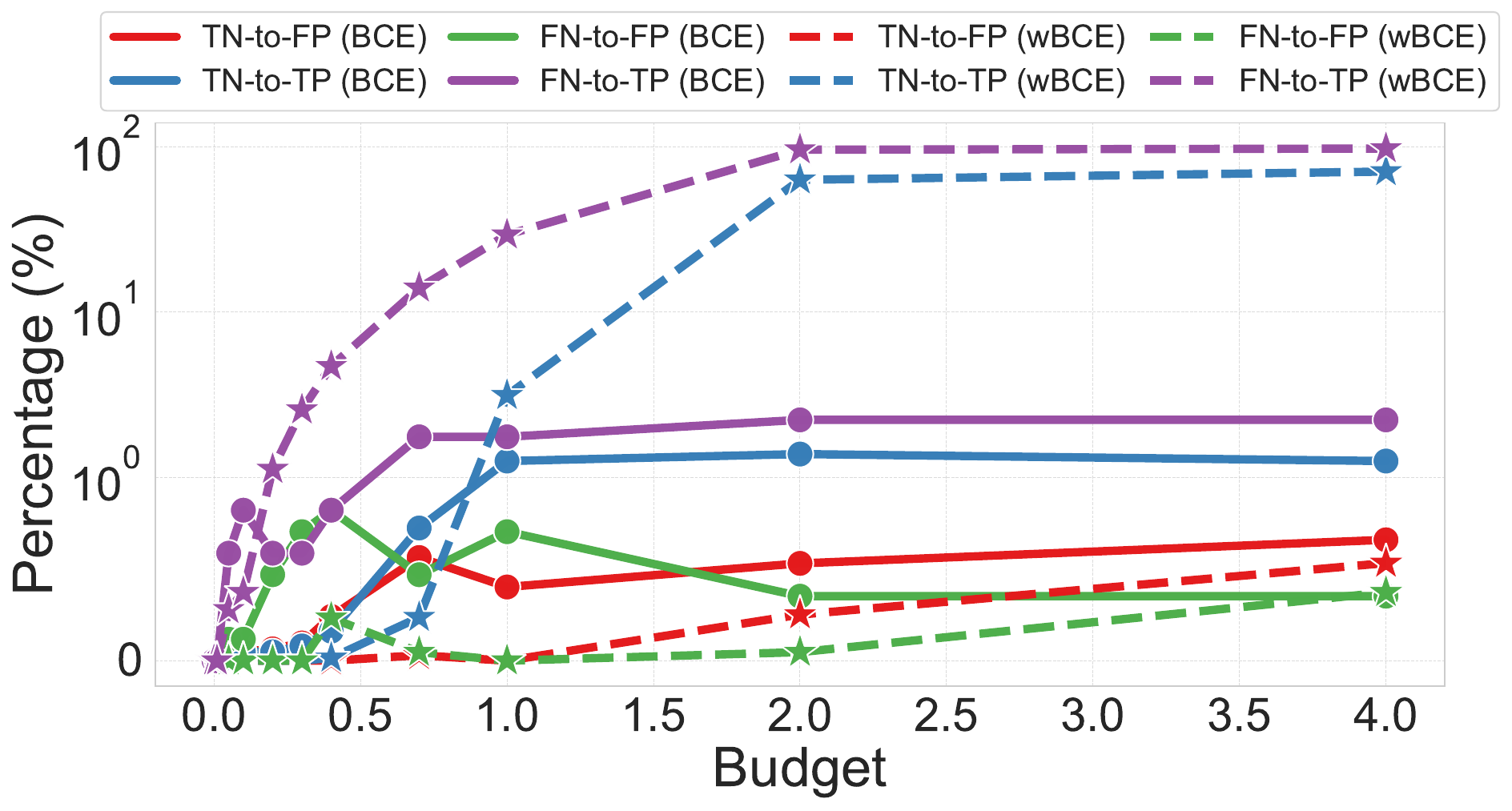}
     \caption{\(\%\) of agents that transition from TN/FN to TP/FP}
     \label{fig:adult_move_0.5}
    \end{subfigure}
    \hfill
    \begin{subfigure}[t]{0.48\linewidth}
        \centering
        \includegraphics[width=\linewidth]{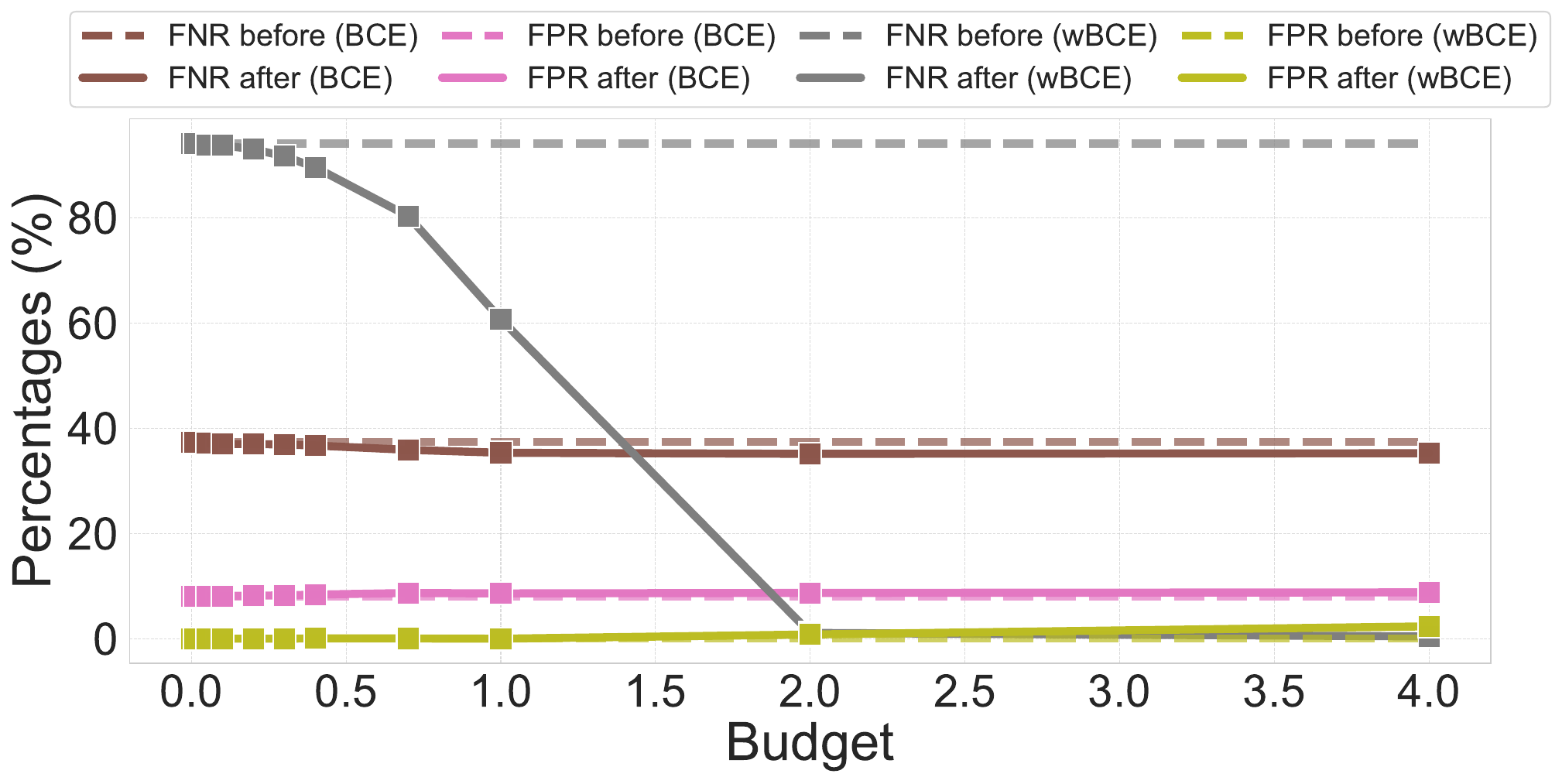}
    \caption{FNR/FPR before and after agents' improvement}
    \label{fig:adult_move_fpr_fnr_0.5}
    \end{subfigure}
    
    \caption[The percentage of negatively classified agents]{The percentage of negatively classified agents (true negatives (TN) and false negatives (FN)) that transition to true positives (TP) and false positives (FP) after responding to the classifier (\(h(x)\)) is shown in Figure~\ref{fig:adult_move_0.5}. 
    Figure~\ref{fig:adult_move_fpr_fnr_0.5} shows the FPR and FNR before and after agents move. While the  wBCE-trained model used \(w_{\textrm{FP}}=0.001\) and \(w_{\textrm{FN}} = 4.4\), the BCE-trained model used \(w_{\textrm{FP}} = w_{\textrm{FN}} = 1\), and an agent is classified as positive if the probability of being positive is above \(0.5\).}
    \label{fig:adult_move_erroreval_0.5}
\end{figure}

First, risk-averse (wBCE-trained) models consistently outperform standard (BCE-trained) models in reducing overall error as the improvement budget increases (Figures~\ref{fig:main_thresh0.5}, \ref{fig:bce_wbce_threshvar}, and \ref{fig:app_thresh0.5thresh0.9}).
While error gains relative to BCE-trained models tend to cancel out, wBCE-trained models retain low false positive rates after agent movement and exhibit a marked decline in false negatives as the improvement budget increases (Figure~\ref{fig:adult_move_erroreval_0.5} and Appendix Figure~\ref{fig:oulad_synthetic_move_erroreval_0.5}). 
Among risk-averse strategies, loss-based risk aversion, 
where \(\mathcal{L}_{\textrm{wBCE}}\)-trained models use \(\frac{w_\textrm{FP}}{w_\textrm{FN}} > 1\) outperforms threshold-based approaches that classify agents as positive only if predicted probability exceeds \(0.9\) (Figure~\ref{fig:main_thresh0.5}).
Second, modest improvement budgets (\(r \leq 2.0\)) lead to substantial error reduction, but returns diminish for (\(r > 2.0\)), especially with a decision threshold of \(0.5\).
Lastly, dataset class separability (Appendix, Figures~\ref{fig:jumbleness} and \ref{fig:orig_knn}) significantly affects the optimal level of risk aversion and the relationship between improvement budget and error reduction.

In summary, risk-averse models initially incur higher errors but achieve rapid error reduction as agents improve and \(r\) increases. 
A stricter false-positive penalty improves the positive agreement region, reducing test error, sometimes to zero (e.g., Appendix, Figure~\ref{fig:app_synthetic_0.5}).

\section{Discussion} 
We propose a novel model for learning with strategic agents where the agents are allowed to improve. Surprisingly, we are able to achieve zero error (with high probability) by designing appropriate risk-averse learners for several well-studied concept classes, including a fairly general discrete graph-based model. We show that the VC dimension of the concept class is not the correct combinatorial dimension to capture learnability in the context of improvements. We further show that the intersection-closed property is sufficient, and in a certain sense necessary for proper learning with respect to any improvement set. We leave open the question of characterizing improper PAC learnability with improvements in terms of the concept class and the improvement sets available to the agents. 

\chapter{Theory of Classification of Improving-and-Gaming Agents}
\label{chap:ocagi_theory}
\section{Introduction}
\label{sec:ocagi_intro}

Consider a bank offering loans.  Based on observable information about applicants, it must decide which of them are loan-worthy and which are not.  For example, it might compute a credit score based on some (perhaps linear) function of observable features and then compare the result to a cutoff value.  So far, this looks like a standard binary classification problem. However, there is an additional wrinkle: individuals have agency and may be able to modify their observable features somewhat if it will help them get approved for a loan.  This wrinkle brings both challenges and opportunities.  A challenge is that some of these actions may involve ``gaming'' the system: performing activities that do not affect their true loan-worthiness such as changing how they spend on different credit cards.  An opportunity is that other actions, such as taking a money-management course, may truly help them become more loan-worthy, increasing the number of good loans the bank can give out. How can the bank best set its loan criteria in such settings to maximize the number of loans given out subject to not giving loans to unqualified applicants?  

Alternatively, consider a school that would like to prepare students for the workforce.  There are many different career paths a student might take, so the school would like to have multiple different criteria for graduation (multiple tracks or majors) such that satisfying any one of them will earn the student a diploma.  Imagine there is a limited set of options the school can choose from, and once the school chooses some subset of them as criteria, every student selects the easiest of those criteria to fulfill (or none, if all are too hard) and then may or may not become truly qualified for the workforce, depending perhaps on the extent to which satisfying that criterion involved gaming versus true improvement.  How can the school best select criteria to maximize the number of students who become truly qualified for the workforce while minimizing the number of diplomas given to unqualified students? 

In this work we consider algorithmic and learning-theoretic formulations of such scenarios, where a binary classification must be made in the presence of both gaming and improvement actions with a goal of maximizing true-positive predictions while keeping false-positives to a minimum.  Specifically, we consider the following two formulations (given in more detail in Section~\ref{sec:ocagi_model}). 

\begin{description}
\item[General Discrete Model:] In this formulation, we are given a weighted, colored bipartite graph with $n$ nodes on the left representing agents, and $m$ nodes on the right representing distinct possible ways agents could be considered {\em qualified} for the prize at hand (the loan, the diploma, etc.).  For example, the nodes on the right could represent different possible definitions of ``credit-worthy'' or could represent different bundles of activities sufficient to receive a diploma.  Each edge has both a {\em weight} representing the amount of effort the agent would need to achieve the given qualification and a {\em color} blue or red indicating whether the agent would indeed be truly qualified or not (respectively) if it did so.  The goal of the classifier is to select a subset $\pfinal$ of points on the right such that if each agent in the neighborhood of $\pfinal$  takes its least-cost edge into $\pfinal$, then a large number of blue edges and very few red edges are taken (many good loans and few bad loans are given out); more specific objectives will be detailed in Section~\ref{sec:ocagi_general}.

In the learning-theoretic version of this problem, the left-hand-side of the graph is replaced with a probability distribution ${\cal D}$ over nodes (where a node is given by its neighborhood and the weights and colors of its edges).  We have sampling access to ${\cal D}$ and our goal is to find a subset $\pfinal$ of points on the right-hand-side with good performance under ${\cal D}$.  In a partial-information version, when we sample a point from ${\cal D}$ we do not get to observe its edges, only where the agent goes to and whether it was qualified.  That is, learning proceeds in rounds, where in each round we choose a subset $\points'$ of points on the right, and then for a random draw $x \sim {\cal D}$ we observe what point $p \in \points'$ (if any) was selected and the color of the edge taken.

\item[Linear Model:] In this formulation, we assume agents are points $\vec{x}\in\mathbb{R}^d$ (they have $d$ real-valued features) and there is a linear separator $f^{\star}: \vec{a}^{\star}\vec{x} \geq b^{\star}$ with non-negative weights that separates the truly qualified individuals from the unqualified ones.  Agents have the ability to increase their $j$th feature at cost $\vecj{c}$ (decreasing is free) and receive value 1 for being classified as positive.  However, only some features correspond to true improvement and others involve just gaming. That is, if an agent begins at $\init{x}$ and moves to a point $\perc{x}$, their true qualification is not $f^{\star}(\perc{x})$ but rather $f^{\star}(\true{x})$, where $\true{x}$ agrees with $\init{x}$ in the gaming directions and with $\perc{x}$ in the improvement directions.  Movement costs and which features are improvement versus gaming  are assumed to be the same for all agents. The goal is to find a classifier that produces a large number of true positives and few false positives. Note that using $f^{\star}$ itself will be optimal if the coordinate $j$ maximizing $\vecstarj{a}/\vecj{c}$ (having the most ``bang per buck'') is an improvement direction, so the interesting case is when this is a gaming direction.  Also note that shifting $f^{\star}$ in this direction (adding  $\vecstarj{a}/\vecj{c}$ to $b^{\star}$) will be a perfect classifier but may not be optimal because it does not take advantage of the ability to encourage agents to improve.   We consider settings where (a) the mechanism designer must use a linear classifier, (b) arbitrary classifiers are allowed, and (c) a polynomial-sized set $\points$ of ``target points'' is given and the mechanism designer must select some subset $\pfinal\subseteq \points$ as its classifier --- this is a special case of our General Discrete Model. 
\end{description}

In this chapter, we consider both models.  We give an efficient algorithm for the general discrete model for the problem of maximizing the number of blue edges taken subject to no red edges taken (maximizing the number of good loans given out subject to no bad loans) and show how to extend this to the partial-information learning setting.  We also show hardness for the problem of maximizing the number of blue edges subject to a nonzero bound on the number of red edges, and show that this hardness holds even for the simplest finite-point linear model. Furthermore, we show the problem of maximizing the number of true positives subject to no false positives is NP-hard in the linear model when we are not given a polynomial-sized set of target points.  We additionally give algorithms for the linear model.
We provide an algorithm that determines whether there exists a linear classifier which classifies all agents accurately and causes all improvable agents to become qualified. In the special two-dimensional case, we design a linear classifier maximizing the number of true positives minus false positives; and a general (not necessarily linear) classifier that maximizes true positives subject to no false positives.

\subsection{Related Work}

There is an exciting and growing literature on decision-making in the presence of strategic agents. Much of this work considers agents whose actions are only gaming and do not change their true label (see \citep{hardt2016strategic,revealed_preferences,Hu:2019:,Milli2018TheSC,strategicperceptron,adversarial_games_pred,Frankel2019ImprovingIF,braverman_et_al} among others)  but researchers have also been investigating mechanism design in the presence of agents who can both game and improve {\citep{Kleinberg2018HowDC,harris2021stateful,Alon2020MultiagentEM,xiao2020optimal,Miller2019StrategicCI,Haghtalab2020MaximizingWW,Bechavod2020CausalFD,Shavit2020LearningFS}}.  

\citet{Kleinberg2018HowDC} consider a single agent with a variety of gaming and improvement actions available, that are then converted into observable features through an effort-conversion matrix.  They then examine mechanisms for incentivizing desired action vectors, showing among other things that any vector that can be incentivized by a monotone mechanism can also be incentivized by a linear mechanism.  
\citet{harris2021stateful} consider a multi-round version of the \citet{Kleinberg2018HowDC} model in which true improvements carry over to future rounds whereas gaming effort do not; they show that in this model, the principal (the decision-maker) can incentivize the agent to produce a greater range of desirable behaviors. 

\citet{Alon2020MultiagentEM} consider a multi-agent extension of the \citet{Kleinberg2018HowDC} model, where agents all begin at the same place (the origin) but each have their own effort-conversion matrix. The goal of the designer is to choose an evaluation mechanism---mapping observable features to payoffs---that encourages all agents to take {\em admissible} actions, assuming that agents will maximize payoff subject to budget constraints.  They specifically consider the case (1) that there is a single admissible action vector, and (2) that individual actions are either improvement or gaming actions and no agent should take a gaming action. Among other results they show that unlike in \citep{Kleinberg2018HowDC}, nonlinear evaluation mechanisms can now be more powerful than linear ones; they also analyze the complexity of a variety of associated optimization problems.  We can think of our setting to some extent in this language by viewing any action that makes an agent truly qualified as ``admissible'' (and specifically the blue edges in our general discrete model).  However, two key distinctions are (1) in our setting we can only give the loan/diploma or not---we do not have the flexibility to choose arbitrary payoffs, and (2) we assume agents may begin at different starting locations (but have the same costs for movement in our linear model). 

\citet{xiao2020optimal} define a problem they call the {\em Multiple Agents Contract Problem} which is very similar to our General Discrete Model, except instead of binary (red/blue) colors, the edges have different values to the principal, and instead of producing a classification, the principal can assign an arbitrary payment profile to the right-hand-side nodes.  They prove that maximizing payoff to the principal is NP-hard, and give an algorithm for a case of related agents in which there is a certain strict ordering among agents and costs. 

\citet{Shavit2020LearningFS}, building on \citet{Miller2019StrategicCI}, consider the goal of getting agents to improve without loss of predictive accuracy.  As in our setting, they assume agents begin a different starting locations, and then modify their profiles from there, and they also consider a learning formulation.  However, their focus is on a regression model in which agents' payoffs are an inner product of their observable features with a decision vector; this means that the incentives are basically the same no matter what the initial location of an agent is.  In contrast, in our binary classification setting, even in the linear model the effect of a proposed classifier on an agent may depend greatly (and in a non-convex manner) on the initial location of the agent.  \citet{Bechavod2020CausalFD} also consider a linear regression learning setting: agents arrive one at a time iid from a fixed distribution and then modify their state by changing a single variable based on the current regression vector.  As in our linear model, some directions are improvement and some are gaming.  They consider a limited feedback setting where the learner sees only the dot-product of the agent's true position with the true regression function, plus noise, and the learner's goal is to recover the true regression function.

\citet{Haghtalab2020MaximizingWW} consider a similar setting to ours in which there are improvement and gaming actions, and the designer is limited to binary classification, where agents receive value 1 for being classified as positive. Among other results, they give approximation algorithms for the goal of maximizing the total amount of true improvement that occurs when the allowed mechanisms are linear separators and agents have $\ell_2$ movement costs.  In contrast, our goal is to maximize true positive classifications while minimizing false positives, and in the linear case our movement cost assumptions are {somewhat different}.

\paragraph{Organization of this Chapter.}
{Section~\ref{sec:ocagi_model} introduces the general discrete model and linear model more formally.
In Section~\ref{sec:ocagi_general}, we give an efficient algorithm for the problem of maximizing the number of true positives subject to no false positives in the general discrete model, and provide hardness results for the problem of maximizing the number of true positives subject to a nonzero bound on false positives (in either the general discrete model or the linear model when arbitrary classifiers are allowed) and hardness for the problem of maximizing the number of true positives subject to no false positives in the linear model when arbitrary classifiers are allowed. In Section~\ref{sec:ocagi_learning}, we consider a learning-theoretic version of the problem of maximizing true positives subject to no false positives, and provide efficient learning algorithms as well as upper and lower bounds on the number of samples needed. In Section~\ref{sec:ocagi_linear}, we focus on the linear model and provide algorithms specific to this setting. We provide an algorithm that determines whether there exists a linear classifier which classifies all agents accurately and causes all improvable agents to become qualified. In the special two-dimensional case, we design a linear classifier maximizing the number of true positives minus false positives; and a general (not necessarily linear) classifier that maximizes true positives subject to no false positives.}

\section{Model}
\label{sec:ocagi_model}

We study a binary classification problem. As the mechanism designer or classifier, we would like to maximize the number of agents we correctly classify as positive (true positives), and minimize the number of unqualified agents we misclassify as positive (false positives). 

Agents are assumed to be utility maximizers and wish to be classified as positive.  Each agent $i \in \{1, \ldots, n\}$ has a set of actions it can perform, and it will choose the cheapest of these that causes it to be classified as positive if that cost is not greater than its value of receiving a positive classification. We use $\q$ to denote the set of truly qualified agents.  If an agent is initially not qualified (not in $\q$), some of its actions may cause it to become truly qualified, whereas others may not.  However, the classifier cannot see which action was taken, only the observable result of that action.  Therefore, the challenge of the mechanism designer is to determine which observable results to classify as positive to maximize  correct positive classifications while minimizing false positives.

\subsection{General Discrete Model}\label{sec:ocagi_gen_model}
In this model, we assume that as a mechanism designer we are given a polynomial-sized set $\points$ of criteria we may select from (e.g., graduation criteria or criteria for being approved for a loan), and are limited to choosing some subset $\pfinal \subseteq \points$ as the criteria we will use.  We then will classify as positive any agent that meets any one of these criteria, {and as negative any agent who does not}.  Specifically, we are given a weighted, colored bipartite graph with the $n$ agents on the left and the set $\points$ of criteria on the right.  Edge $(i,j)$ corresponds to agent $i$ taking an action to satisfy criteria $j$ and is colored blue or red depending on whether that action would make the agent truly qualified or not, respectively.  Each edge also has a weight representing its cost to that agent, and only actions whose costs are less than the value to the agent of being classified as positive are shown.   Given a set  $\pfinal \subseteq \points$ chosen by the mechanism designer, each agent in the neighborhood of $\pfinal$ will choose its cheapest edge into $\pfinal$ as the action it will take, and will be classified as positive by the mechanism; agents not in the neighborhood of $\pfinal$ will be classified as negative.

We also consider a learning-theoretic version of this problem, where the left-hand-side of the graph is replaced with a probability distribution ${\cal D}$ over nodes.  We have sampling access to ${\cal D}$ and our goal is to find a subset $\pfinal$ of points on the right-hand-side with good performance under ${\cal D}$.  In a partial-information (bandit-style) version, when we sample a point from ${\cal D}$ we do not get to observe its edges, only where it goes to and whether it was qualified.  That is, learning proceeds in rounds, where in each round we choose a subset $\points'$ of points on the right, and then for a random draw $x \sim {\cal D}$ we observe what point $p \in \points'$ (if any) was selected and the color of the edge taken.

\begin{figure}[ht!]
\centering
\includegraphics[width=10cm]{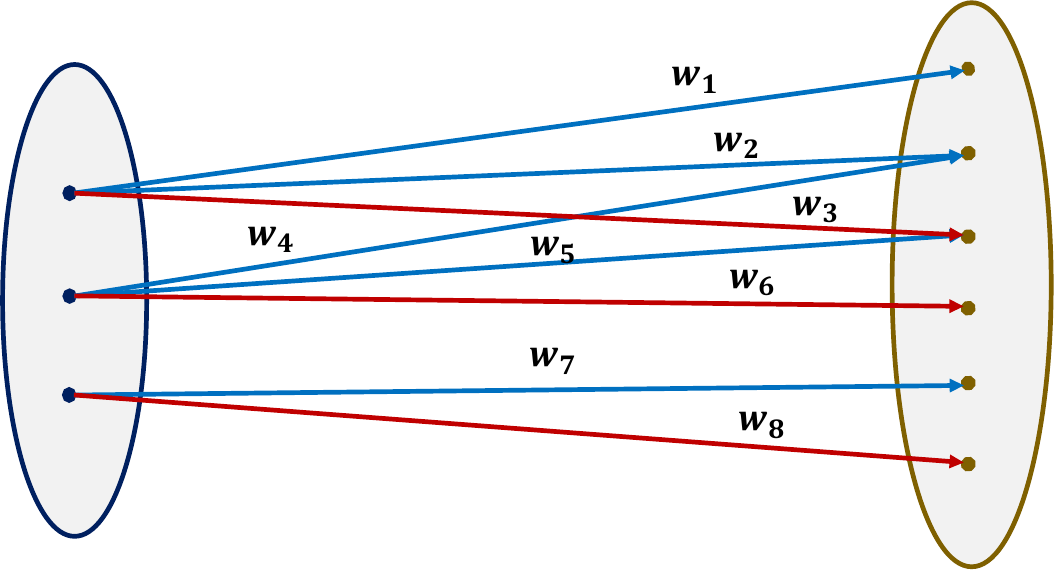}
\caption[General discrete model]
{Points on the left are the agents, and those on the right are the set $\points$ of possible criteria; $w_i$ is the cost of satisfying the criterion. A red edge means the agent taking that action would not truly be qualified. A blue edge means that the agent taking that action would be qualified.
}\label{fig:matching}
\end{figure}

\subsection{Linear Model}
\label{sec:ocagi_linear_model}

In the linear model, agents have $d$ real-valued features.  Each agent $i$ begins at an initial point $\init{x}_i \in \mathbb{R}^d$, and there is assumed to be a linear threshold function $f^{\star}: \vec{a}^{\star}\vec{x} \geq b^{\star}$ with non-negative weights that separates the truly qualified individuals from the unqualified ones.  Agents have the ability to increase their $j$th feature at cost $\vecj{c}$ (decreasing is free) and receive value 1 for being classified as positive.  However, only some features correspond to true improvement and others involve just gaming. That is, if an agent begins at $\init{x}$ and moves to a point $\perc{x}$, their true qualification is not $f^{\star}(\perc{x})$ but rather $f^{\star}(\true{x})$, where $\true{x}$ agrees with $\init{x}$ in the gaming directions and with $\perc{x}$ in the improvement directions.   On the other hand, the classification rule can only be based only on $\perc{x}$ and not $\true{x}$ (or $\init{x}$).  Movement costs and which features are improvement versus gaming  are assumed to be the same for all agents.  So, for any agent $i$, $cost(\init{x}_i,\perc{x}_i)=\sum_{j=1}^d \vecj{c} \left(\percij{x}-\initij{x}\right)^+$, where $x^+=\max\{x,0\}$ and $\vecj{c}$ is the cost per unit of movement in the positive direction of dimension $j$. 

We consider settings where (a) the mechanism designer must use a linear classifier (a linear threshold function), (b) arbitrary classifiers are allowed, and (c) a polynomial-sized set $\points$ of ``target points'' is given and the mechanism designer must select some subset $\pfinal\subseteq \points$ as its classifier.  Notice that this last case is a special case of the general discrete model because given each initial state $\init{x}_i$, we can compute the costs to move to each $p \in \points$ and whether doing so will make the agent truly qualified, to produce the desired weighted, colored bipartite graph. 

\begin{figure}[ht!]
\centering
\includegraphics[width=11cm]{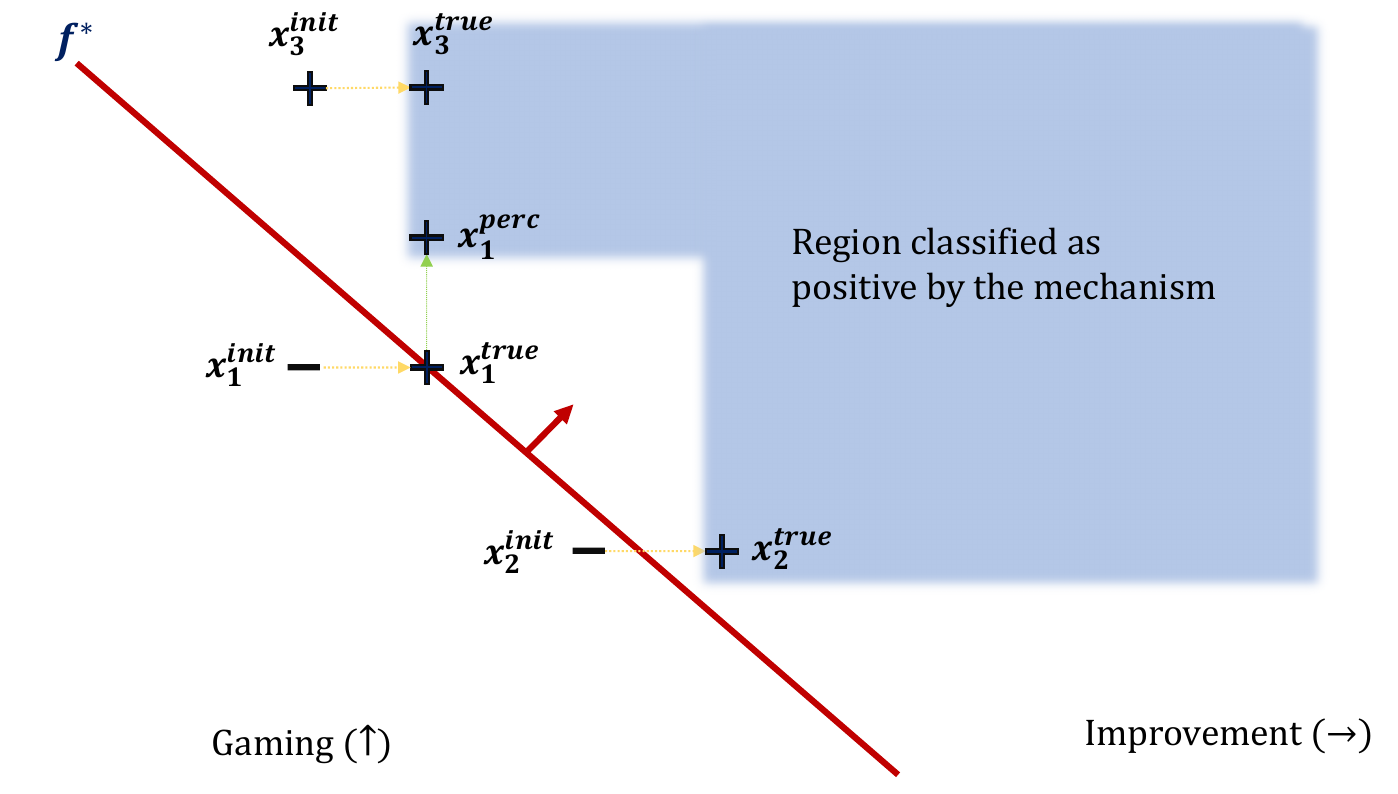}
\caption[An example of the linear model]
{An example of the linear model (the horizontal axis is an improvement direction and the vertical axis is a gaming direction) with a mechanism using a non-linear classifier.  There are three agents, two of whom are initially not qualified.  All three become qualified and are correctly classified as positive by the mechanism.
}\label{fig:linear_case}
\end{figure}

\section{Algorithmic and Hardness Results}
\label{sec:ocagi_general}

In this section we first provide an algorithm for the problem of maximizing the number of true positives subject to no false positives in the general discrete model. Then, we provide hardness results for the problem of maximizing the number of true positives subject to a nonzero bound on false positives (in either the general discrete model or the linear model when arbitrary classifiers are allowed) and hardness for the problem of maximizing the number of true positives subject to no false positives in the linear model when arbitrary classifiers are allowed. Later in Section~\ref{sec:ocagi_learning} we extend our algorithmic results to the learning model and in Section~\ref{sec:ocagi_linear} we give algorithms for learning linear classifiers in the linear model.

\subsection{Maximize True Positives Subject to No False Positives}

The main result of this section is an algorithm that given a weighted, colored bipartite graph $\graph$ with agents, $\agents$, on the left and potential criteria, $\points$, on the right, finds $\pfinal \subseteq \points$ such that using $\pfinal$ as the criteria  maximizes the number of agents taking a blue edge (true positive) subject to no agent taking a red edge (false positive). We call the agents that take a blue edge \emph{improving agents} and the agents taking a red edge \emph{gaming agents}. The algorithm, although simple in structure, satisfies strong properties noted afterwards; and serves as the building block of the learning algorithms in Section~\ref{sec:ocagi_learning}. Furthermore, as shown in the following subsection, natural generalizations of the objective function make the problem computationally hard. Therefore, the algorithm together with the hardness results tightly characterize the settings for which there is an efficient algorithm, or the problem is NP-hard.

\paragraph{Overview of Algorithm~\ref{alg:maximal}.} 
The algorithm takes in a weighted, colored bipartite graph $\graph = (\agents \cup \points,E)$ and outputs  $\pfinal$, a subset of $\points$ that specifies the final criteria. Initially, $\pfinal$ is set to $\points$. The algorithm proceeds in rounds. In each round, it visits all the nodes (agents) in $\agents$ to determine whether there is an agent who takes a red edge to its lowest cost neighbor $p \in \pfinal$. If there is such a gaming agent, its corresponding criteria, $p$, is removed from $\pfinal$. These rounds continue until there is no gaming agent and therefore no removal of criteria in a single round, or the current set of criteria is empty. 

\begin{algorithm}
    \SetNoFillComment
    \SetAlgoLined
    \SetKwInOut{Input}{Input}
    \SetKwInOut{Output}{Output}
    \SetKw{Return}{return}
     \DontPrintSemicolon
    \Input{A bipartite graph  $\graph = (\agents \cup \points,            E)$ with edge weights $w_{e}$. Outgoing edges assumed sorted by weight.
           Red edges  $E_{R} \subseteq E$.
           Blue edges $E_{B} \subseteq E$.
        }
        
    \Output{$\pfinal$ }
    $\pfinal \leftarrow \points$\ \tcp{Initialization of the set}

    \While{$\pfinal \neq \emptyset$}{
        $flag = 0$\;
        \tcc{Loop through all $x_i \in \agents$}
        \For{$i=1, 2, \cdots$}  { 
            Let $e = (x_i, p \in \pfinal)$ be the outgoing edge from $x_i$ with lowest weight\;
            \If{$e\in E_R$}{
                $flag = 1$\ \tcp{at least one agent is gaming}
                $\pfinal \leftarrow \pfinal \setminus \{p\}$\
            }
        }
        \If{flag is $0$}{
            \Return $\pfinal$\
        }
    }
    \Return $\emptyset$\ \tcp{When $0$ false positive is not possible}
    \caption{Maximize true positives subject to no false positives.}
    \label{alg:maximal}
\end{algorithm}

\proposition{\label{prop:running-time-greedy-alg}Algorithm~\ref{alg:maximal} has running time of $O(|\points|n)$}. 
\proof Proof in Appendix~\ref{app:ocagi_hardness}.

\begin{theorem}
\label{proof:greedy-correctness}
Algorithm~\ref{alg:maximal} finds the set of criteria, $\pfinal$, that maximizes {the number of true positives subject to no false positive.} \end{theorem}
\begin{proof}
Let A be the improving agents (agents taking blue edges) associated with the set of criteria $\pfinal$. We show that having any other set $Q \subseteq \points$ as the criteria, either causes an agent to take a red edge, or no more than $|A|$ agents to take blue edges. To do so, consider partitioning $Q$ into two subsets $Q^F$ and $Q^{\bar{F}}$, where $Q^F \subseteq \pfinal$ and $Q^{\bar{F}} \subseteq \points \setminus \pfinal$.

First, we show that if $Q^{\bar{F}} \neq \emptyset$, an agent takes a red edge. To prove this claim, suppose by  contradiction that $Q^{\bar{F}}$ is nonempty and consider the first time the algorithm deletes an element $p \in Q^{\bar{F}}$. At this stage, the set of criteria in the algorithm $\points'$ is a superset of $Q^{\bar{F}} \cup \pfinal$. By definition, $p$ is the lowest-weight neighbor of a gaming agent, $a$, in $\points'$. This implies that $p$ is also the lowest-weight neighbor of $a$ in $Q \subseteq Q^{\bar{F}} \cup \pfinal \subseteq \points'$, and $a$ is a gaming agent given the criteria set $Q$. This implies the claim. 

Secondly, we show that among the sets of criteria with no gaming agent, $\pfinal$ has the highest number of improving agents. The previous claim implies that any set of criteria with no gaming agent is a subset of $\pfinal$. Now, we need to show that among $Q \subseteq \pfinal$, $\pfinal$ has the largest set of improving agents. This is trivial, since by considering a subset we may only lose on agents in $A$ that do not have a neighbor in $Q$ or their lowest-weight edge is red. Therefore, any $Q \subseteq \pfinal$  has at most $|A|$ improving agents.
\end{proof}

Algorithm~\ref{alg:maximal} satisfies the following strong properties.
\begin{enumerate}[label=(\alph*)]
    \item \label{pr:point-wise-optimal} {\em point-wise optimality}:
    For any agent $i$, if there exists a solution in which $i$ takes a blue edge and no agent takes a red edge, then the algorithm finds such a solution. 
    \item {\em general for weighted setting}: The algorithm works optimally in the more general setting that each agent has a weight and the objective is to maximize the sum of weights of improving agents subject to the constraint of no gaming agent. This is a direct implication of property \ref{pr:point-wise-optimal}.
    \item {\em max-min fairness:} Suppose the agents are from different populations and the objective is to maximize the minimum number of agents improving from each population subject to no gaming. By property \ref{pr:point-wise-optimal}, the algorithm satisfies this max-min fairness notion.
    \item {\em heterogeneous utilities}: The algorithm works optimally in the more general setting that agents have different values for being classified positive.
    \item {\em minimizing the total cost of improvement}: Since the algorithm only removes $p \in \points$ that causes an agent to game, with $\pfinal$ each agent incurs the minimal cost subject to no agent gaming.
\end{enumerate}

\begin{remark} 
The sets of criteria satisfying the no false positive constraint is not downward closed. In other words, a subset of a set of criteria that satisfies the no false positives property does not necessarily satisfy this property.
\end{remark}

\subsection{Hardness Results}

In this part, we prove hardness results for maximizing the number of true positives when the constraints in the previous subsection are relaxed. First, we show that if we relax the no false positives constraint to a bounded number of false positives, the problem becomes NP-hard; moreover, this holds even for the simpler linear model. Then, for the linear model, we show if we are not given a finite set of potential criteria $\points$, it is NP-hard to find criteria that maximize true positives subject to no false positives.

\begin{theorem}
\label{thm:atmost_k_game}
Given the initial feature vectors of agents $\init{x}_1, \init{x}_2, \ldots, \init{x}_n \in \mathbb{R}^d$ and a set $\points$ of potential criteria, the problem of finding a subset $\pfinal \subseteq \points$ that maximizes the number of true positives subject to at most $k$ false positives is NP-hard. 
\end{theorem}

\begin{proofsketch} The proof is done by a reduction from the Max-$k$-Cover problem with $n$ elements where the goal is to choose $k$ sets covering the most elements. For every element $e_i$ in the Max-$k$-Cover, we consider agent $i$, and for every set $S_j$ in the Max-$k$-Cover problem we consider agent $n+j$ and a target point $\vec{p}_j$. The coordinates of the initial points and the target points are set such that agent $i$ corresponding to element $e_i$ can only move to target point $\vec{p}_j$ such that $e_i \in S_j$ and become a true positive; moreover, agent $n+j$ corresponding to set $S_j$ can only move to target point $\vec{p}_j$ and become a false positive. On the one hand, since including each $\vec{p}_j$ in the final set of criteria, $\pfinal$, causes exactly one agent to be a false positive, $\pfinal$ must contain at most $k$ target points. On the other hand, to maximize the number of true positives a set of $k$ target points that the maximum number of agents can reach to it must be selected. This is equivalent to the Max-$k$-Cover solution.
A formal proof is included in Appendix~\ref{app:ocagi_hardness}.
\end{proofsketch}

\begin{theorem}
\label{thm:relaxed-no-destination-pts}
Suppose we are given a set of $n$ agents where $\init{x}_1, \init{x}_2, \ldots, \init{x}_n$ denote their initial feature vectors. Deciding whether there exists a set of target points $\pfinal\subseteq \mathbb{R}^d$ for which all the agents become true positives is NP-hard. 
\end{theorem}

\begin{proofsketch}

The proof is done by a reduction from the approximate version of the hitting set problem where given a set  of elements, $\mathcal{E}=\{e_1, \ldots, e_n\}$ and  a family of sets of elements, $\mathcal{F}=\{S_1, S_2, \ldots, S_m\}$, the goal is to find a minimum size set $S^{\star}$ that intersects all $S_i$. We construct an $n+1$-dimensional space, where the first $n$ dimensions are improvement dimensions and correspond to the $n$ elements, and the last dimension is gaming. We consider two sets of agents. For each $S_i$, we consider a corresponding agent $i$; these are the \emph{usual} agents. We also consider agent $m+1$, a \emph{special} agent that does not correspond to any particular set. The construction is such that each agent needs to move $2k$ units along the improvement dimensions to become truly qualified. Further details of the construction can be found in the full proof.
The proof includes two directions. (1) If all the agents can become true positives by reaching to a set of target points $\pfinal\subseteq \mathbb{R}^d$, then we can construct a hitting set of size at most $2k$; and (2) if it is not possible, then there does not exist a hitting set of size $k$.

We briefly cover the key ideas in each direction. To show the first direction, suppose all the agents can become true positives when presented with target points $\pfinal\subseteq \mathbb{R}^d$. Consider the target point that each agent selects. Using our construction, we show the special agent does not afford to reach to the target points of the usual agents. Also, for each usual agent $i$, there exists element $e_j$ in their corresponding set such that the target point of the special agent has value more than $1$ in coordinate $j$. In order for the special agent to afford to reach to its target point, the number of improvement coordinates with value at least $1$ must be at most $2k$. The elements corresponding to these coordinates constitute a hitting set of size at most $2k$.
To prove the reverse direction we argue: if there exists a hitting set $S^{\star}$ of size $k$, there is a set of target points that encourages all the agents to become true positives. To do so, we construct a set of target points $\pfinal = \{\vec{p}_1, \ldots, \vec{p}_{m+1}\}$, using the elements in the hitting set, that when the size of the hitting set is $k$ makes every agent become true positive. A formal proof is included in Appendix~\ref{app:ocagi_hardness}.
\end{proofsketch}

The following is a direct corollary of Theorem~\ref{thm:relaxed-no-destination-pts}.

\begin{corollary}
Given the initial feature vectors of agents, $\init{x}_1, \init{x}_2, \ldots, \init{x}_n \in \mathbb{R}^d$, finding a set of target points $\pfinal \subseteq \mathbb{R}^d$ that maximizes the number of true positives subject to no false positives is NP-hard.
\label{cor:d-dim-max-TP-no-FP}
\end{corollary}

\section{Learning Results}
\label{sec:ocagi_learning}

In this section we consider a learning-theoretic version of our problem, where the left-hand-side of the graph is replaced with a probability distribution ${\cal D}$ over nodes.  We have sampling access to ${\cal D}$ and our goal is to find a subset $\pfinal$ of points on the right-hand-side with good performance under ${\cal D}$. 
We provide two different algorithmic results and upper bounds on the number of samples for producing a good solution, depending on the information each sample reveals. The first upper bound works for the case where by sampling an agent, its neighborhood (neighboring edges, their colors and weights) is revealed. The second upper bound works in a partial-information (bandit-style) setting, where when we sample a point from ${\cal D}$ we do not get to observe its edges, only where it goes to and whether it was qualified. Finally, we provide a lower bound on the necessary number of samples for any algorithm. The lower bound holds even for the simpler linear model. On the technical side, the algorithms in Sections~\ref{sec:ocagi_learning_full} and \ref{sec:ocagi_learning_partial} use Algorithm~\ref{alg:maximal} as a subroutine and generalize it to a broader setting. The lower bound in Section~\ref{sec:ocagi_learning_lower}, however, holds for \emph{any} PAC learning algorithm, and requires substantially different ideas.

The following definition is crucial in this section.
\begin{definition}[$\opt$, performance, and error]
Let $\opt$ be the maximum probability mass of true positives achievable subject to zero false positives. We denote the probability mass of true positives of an algorithm as its \emph{performance} and the probability mass of false positives as its \emph{error}. A hypothesis is desired if it has comparable performance to $\opt$ and small error.
\end{definition}

\subsection{Sufficient Number of Samples in the Full Information Setting}
\label{sec:ocagi_learning_full}

The main result of this section is that a number of samples linear in $|\points|$ and $1/\eps$ is sufficient for Algorithm~\ref{alg:maximal} to learn a desired hypothesis with high probability.
Specifically, suppose the learner has access to a weighted, colored bipartite graph $\graph = (\agents \cup \points,E)$, where $\agents$ are sampled from ${\cal D}$, and $\points$ is the set of the potential criteria. The learner runs Algorithm~\ref{alg:maximal} with the graph as the input and uses the algorithm output, $\pfinal \subseteq \points$, as its hypothesis, 
i.e., after the training phase it classifies any agent with an edge to $\pfinal$ as positive and any other agent as negative.  We show that a linear number of samples is sufficient so that with high probability, the probability mass of true positives classified by $\pfinal$ is close to $\opt$ and the probability mass of false positives is small.

\begin{theorem}
\label{thm:distributional-upper-bound}
Consider $\pfinal$ as the outcome of Algorithm~\ref{alg:maximal} on $\graph = (\agents \cup \points,E)$, where $\agents$ contains samples from ${\cal D}$. For any $0<\eps, \delta \leq 1$, if $|\agents| \geq \eps^{-1} (\ln(2)|\points|+\ln(1/\delta))$, then with probability at least $1-\delta$ 
the set $\pfinal$ achieves performance at least 
$\opt-\eps$ (i.e., at least $\opt-\eps$ probability mass of true positives)
subject to at most $\eps$ error ($\eps$ probability mass of false positives).
\end{theorem}

\begin{proof}
First, we prove the error bound. By Algorithm~\ref{alg:maximal}, $\pfinal$ has error $0$ for $\agents$. Now, consider a set of criteria $\points' \subseteq \points$ with error at least $\eps$ for distribution $D$. The probability of the error being equal to $0$ over $\agents$ is at most $(1-\eps)^{|\agents|}$. The number of subsets of $\points$ is $2^{|\points|}$. So, by union bound the probability that there exists a set of criteria $\points'$ with error greater than $\eps$ over $D$ and error $0$ over $\agents$ is at most $2^{|\points|}(1-\eps)^{|\agents|}\leq 2^{|\points|}e^{-\eps|\agents|}$. This probability is at most $\delta$ for $|\agents| = \eps^{-1}\Big(\ln(2)|\points| + \ln(1/\delta)\Big)$. Therefore, with  $\eps^{-1}\Big(\ln(2)|\points| + \ln(1/\delta)\Big)$ number of samples, with probability at least $1-\delta$, $\pfinal$ has at most $\eps$ error.

Next, we show the performance guarantee. Recall that $\opt$ is the maximum achievable probability mass of true positives subject to no false positive for distribution $D$. Set $\agents$ contains a subset of the points in the distribution. Algorithm~\ref{alg:maximal} only deletes points from $\points$ if they cause a false positive in $\agents$. Therefore, the output of the algorithm on agents $\agents$, $\pfinal$, is a superset of the criteria set of the optimal zero-error solution. 
This means that the probability mass of examples predicted positive is at least $\opt$, and since at most an $\eps$ probability mass is false-positives, the performance of $\pfinal$ is at least $\opt-\eps$.
\end{proof}

\subsection{Sufficient Number of Samples in the Partial Information Setting}
\label{sec:ocagi_learning_partial}

In this section we consider a partial information (bandit-style) setting. Similar to before, the learner has access to a sample set $\agents$ drawn from $D$ and a set of potential criteria $\points$. However, observing a sample in $\agents$ does not reveal its edges, and the learner can only observe the criterion that the sample selects and whether it becomes truly qualified. The main result of this section is an algorithm,
Algorithm~\ref{alg:upper-bound-no-input-graph}, for this setting and a guarantee on the number of samples sufficient for it to achieve performance at least $\opt-\eps$ and error at most $\eps$ with high probability.

\begin{algorithm}
    \SetNoFillComment
    \SetAlgoLined
    \SetKwInOut{Input}{Input}
    \SetKwInOut{Output}{Output}
    \SetKw{Continue}{continue}
    \SetKw{Return}{return}
    \Input{$\mathcal{P}$}
    \Output{$\mathcal{P}^{final}$}
    $\points^\text{final} \leftarrow \points$\;
    \While{$\points^\text{final}\neq \emptyset$}{
        Sample $\mathcal{X}\sim\mathcal{D}$ of size $\frac{1}{\eps}\ln {\frac{|\points|}{\delta}}$ \;
        \If{$\exists x \in \mathcal{X}$ such that $x$ takes a red edge to $p\in \points^\text{final}$}{
            $\points^\text{final} \leftarrow \points^\text{final} \setminus \{p\}$\;
            \Continue\;
        }
        \tcc{if no one from $\mathcal{X}$ takes a red edge:}
        \Return $\points^\text{final}$\;
    }
    \Return $\emptyset$\;
\caption{Learn a low error $\pfinal$ in partial-information setting}
\label{alg:upper-bound-no-input-graph}
\end{algorithm}

\paragraph{Overview of Algorithm~\ref{alg:upper-bound-no-input-graph}.} 
In each iteration, a set of examples of size $\eps^{-1}\ln (|\points|/\delta)$ is sampled. After agents select points in $\points$ (if any), we observe the points selected and whether they became truly qualified (in a real-world application, one can think of performing a test to check if each agent is truly qualified). If some agent does not become truly qualified (fails the test), the algorithm deletes the point they have selected. If a set $\points^\text{final}$, survives for $\eps^{-1}\ln ({|\points|}/{\delta})$ subsequent examples, the algorithm terminates and returns $\pfinal$ as the the final set of criteria of the algorithm. Since the number of false positives (agents taking red edges) is bounded by $|\points|$, the algorithm will terminate after at most $\eps^{-1}|\points|\ln (|\points|/\delta)$ samples.

The following theorem proves that with a high probability, Algorithm~\ref{alg:upper-bound-no-input-graph} outputs $\pfinal$ with a high performance and a low error.

\begin{theorem}
For any $0<\eps, \delta \leq 1$, Algorithm~\ref{alg:upper-bound-no-input-graph} by using at most $\eps^{-1}|\points|\ln (|\points|/\delta)$ total samples outputs a set of criteria $\pfinal$ that with probability at least $1-\delta$ achieves performance at least $\opt-\eps$ (i.e., at least $\opt-\eps$ probability mass of true positives) subject to at most $\eps$ error ($\eps$ probability mass of false positives).
\end{theorem}

\begin{proof}

First, we prove the error bound. Consider the sequence of $\pfinal$ at the beginning of each iteration of the while loop in Algorithm~\ref{alg:upper-bound-no-input-graph}. Let these sets be $\pfinal_1, \pfinal_2, \ldots$. The probability that $\pfinal_i$ with error greater than $\eps$ over $D$ does not produce any false positives in the following $\eps^{-1}\ln (|\points|/\delta)$ samples is at most $(1-\eps)^{\eps^{-1}\ln (|\points|/\delta)}$. Note that the while loop in Algorithm~\ref{alg:upper-bound-no-input-graph} runs at most $|\points|$ times. Therefore, the number of distinct sets of $\pfinal_i$ considered in the algorithm is at most $|\points|$. By a union bound, the probability that there exists $\pfinal_i$ with error greater than $\eps$ over $D$ and error $0$ over the samples in its iteration is at most $|\points|(1-\eps)^{\eps^{-1}\ln (|\points|/\delta)}$, which using $1-x \leq e^{-x}$ is at most $\delta$.

Proving the performance guarantee is identical to that of Theorem~\ref{thm:distributional-upper-bound}.
\end{proof}

\subsection{Necessary Number of Samples}
\label{sec:ocagi_learning_lower}

The main result of this section is a lower bound on the necessary number of samples for learning a desired hypothesis. The lower bound provided holds even for the simpler linear model. To restate the setup, suppose  the learner has access to a set of initial positions of agents $\agents$ and a set of potential criteria (also called target points in the linear model) $\points$ where $\agents$ are sampled from distribution ${\cal D}$. We lower bound the required number of samples for any learning algorithm that with probability at least $1/2$ achieves high performance and low error.

\begin{theorem} 
Any algorithm for PAC learning a set $\pfinal$ that with probability at least $1/2$ achieves performance at least $(3/4)\cdot \opt$ (i.e., at least $(3/4)\cdot \opt$ probability mass of true positives) subject to at most $\eps$ error ($\eps$ probability mass of false positives) must use $\Omega(|\points|/\eps)$ examples in the worst case.
\end{theorem}

For ease of notation, in the proof we use $m := |\points|$. 

\begin{proof}
We construct a concept class, a distribution of agents and a set of potential criteria (target points) that forces any PAC learning algorithm to take many samples. First, consider a concept class $C$ that includes solutions whose final set of target points $\pfinal$ has size exactly $3m/4$. Therefore,  $|C| = \binom{m}{3m/4}$. Target concept $c$ is chosen randomly from $C$. Secondly, we construct a linear setting. We consider a two-dimensional space where $\f^{\star}: \vec{x}[1]+\vec{x}[2] \geq 2m$. Let $\points=\{\vec{p}_1, \ldots, \vec{p}_m\}$. All $\vec{p}_i$ satisfy $\vec{p}_i[1]+\vec{p}_i[2] = 2m$ and $\vec{p}_i[1] = 2i$. The costs of moving in either dimension is $1$ per unit of movement, i.e., $\vec{c}[1]=\vec{c}[2]=1$. Dimension $1$ is an improvement dimension and dimension $2$ is a gaming dimension. Let the set of examples $S=\{\vec{x}_1, \vec{x}_2, \ldots, \vec{x}_{2m}\}$ denote the distinct potential initial positions of any agents. We construct the examples in proximity of the target points such that each example can afford to move to exactly one target point, called its \emph{designated} target point. More formally, for $i \leq m$, let $\vec{x}_i[1] = \vec{p}_i[1]-1, \vec{x}_i[2] = \vec{p}_i[2]$ and $\vec{x}_{m+i}[1] = \vec{p}_i[1], \vec{x}_{m+i}[2] = \vec{p}_i[2]-1$. With this setup, examples $i$ and $m+i$ are in proximity of their designated target point $\vec{p}_i$. Examples $\vec{x}_i$ such that $i \leq m$ are \emph{improving examples} since any agent with initial position $\vec{x}_i$ becomes truly qualified by moving to their designated target points and examples $\vec{x}_i$ such that $i > m$ are \emph{gaming examples} since any agent with initial position $\vec{x}_i$ does not become truly qualified. Finally, we consider a distribution $D_c$ over the examples. Let $P_G$ be the target points not included in the concept $c$. For each $i$ such that $\vec{p}_i \in P_G$, $\Pr[\vec{x}_{m+i}] = 128\eps/m$;  for each $i$ such that $\vec{p}_i \notin P_G$, $\Pr[\vec{x}_{m+i}] = 0$; and for each $i$, $\Pr[\vec{x}_{i}] = (1-32\eps)/m$.  With this probability distribution, there is a $0$ probability mass over gaming examples with designated target point $\notin P_G$, total probability mass of $32\eps$ distributed uniformly over gaming examples with designated target points $\in P_G$, and total probability mass of $1-32\eps$ distributed uniformly over improving examples. Note that with this construction, the target concept $c$ uses $\pfinal = \points \setminus P_G$ which achieves performance $\opt = 3/4$ and error equal to $0$.

Now, let $L$ be any PAC learning algorithm for $C$. Consider running $L$ when the target concept $c \in C$ is chosen randomly and the input distribution is $D_c$. Recall that $P_G$ is the set of target points not included in $c$ which is also the set of target points that positive mass of gaming examples can reach to. The purpose of the algorithm is to learn $P_G$. For this purpose the algorithm only benefits from sampling gaming examples. This is because each gaming example $\vec{x}_{m+i}$ reveals $\vec{p}_i \in P_G$; and in contrast, since improving examples are distributed uniformly across \emph{all} target points $\points$, sampling an improving example does not provide any information about $P_G$. We denote observing a gaming example with designated target point $\vec{p} \in  P_G$ as ``observing a point in $P_G$''. We consider set $O_G$ which includes any observed point in $P_G$. In what follows we assume $L$ never includes any point in $O_G$ in its set of final target points $\pfinal$. This is because any point in $O_G$ causes an error of $128\eps/m$, while replacing it with any $\points \setminus O_G$ causes less error while the probability mass of true positives does not decrease.

The proof consists of two parts. In the first part, we argue that the number of $\vec{p} \in  P_G$ that the algorithm has observed after $n \leq m/(256 \eps)$ samples is limited and a considerable number are yet unobserved. More formally, in the first part we argue after drawing $n$ samples, with high probability $L$ has observed at most $3/4$ fraction of distinct points in $\points_G$. The proof goes as follows. Consider $O_G$ after drawing $n$ samples and let $B$ be the event that $|O_G| \geq 3|P_G|/4 = 3m/16$. Since each example is a gaming example with probability $32\eps$, the expectation of $|O_G|$ is at most $m/(256\eps) \times 32\eps = m/8$. Using Chernoff-Hoeffding bounds, $\Pr[|O_G| \geq (m/8)(1+1/2)]\leq e^{-m/96}\leq 10^{-4}$, where the last inequality holds if $m \geq 1000$.\footnote{We use the assumption of $m \geq 1000$ for the concentration bounds. For $m < 1000$, $L$ still needs to observe $\Omega(1/\eps)$ to sample a gaming example.} Therefore, $B$ happens with probability at least $0.99$.

Then in the second part, we argue if the algorithm does not include many of the unobserved points from $P_G$ in $\pfinal$ it has low performance; and if it does it has high error. Let $E$ be the event that $L$ includes at least $1/16$ fraction of the points from $P_G$ in $\pfinal$. As discussed previously $L$ does not include any points from $O_G$. Therefore, $E$ is also the event that $L$ includes at least $1/16$ fraction of $P_G \setminus O_G$ in $\pfinal$. Note that even after observing the samples, the gaming points corresponding to $P_G \setminus O_G$, are still uniformly distributed among $\points \setminus O_G$. From the setup, it is clear that the problem of predicting whether the designated target point of an example belongs to $P_G \setminus O_G$ is equivalent to predicting the outcome of a random process. To have a performance of $3/4\ \opt$, $L$ needs to include at least $3/4 \times 3/4 > 1/2$ fraction of $\points$; and since it does not include any $O_G$ it needs to include at least $1/2$ fraction of $\points \setminus O_G$, which is what we assume in the remaining part of the proof. We argue that conditioned on $B$, event $E$ happens with a high probability. Let $Z_i$ be an indicator random variable indicating whether $L$ includes $\vec{p}_i \in P_G \setminus O_G$  in $\pfinal$. Including at least $1/2$ fraction of $\points \setminus O_G$ implies $\E[Z_i] \geq 1/2$. Let $Z$ be the sum of $Z_i$ for the unobserved points in $P_G$, i.e., $Z = \sum Z_i \cdot \mathbbm{1}{[\vec{p}_i \in P_G \setminus O_G]}$. Conditioned on $B$, the event that $|O_G| \geq 3|P_G|/4$, we find $\E[Z\mid B] \geq |P_G|/8$. We have
\begin{align} 
&\Pr[(Z<\frac{|P_G|}{16})\mid B] \leq e^{-\frac{|P_G|}{64}}\leq e^{-\frac{250}{64}} < 0.03,\\
&\Pr[E\mid B] > 0.97,\\
&\Pr[E] \geq \Pr[E\cap B] = \Pr[E\mid B]\cdot \Pr[B] \geq 0.96;
\end{align}
where the first line follows from Chernoff-Hoeffding bounds and using $m \geq 1000$; the second line is a direct implication of the first line; and the last line uses properties of conditional probabilities and a lower bound on the probability of event $B$ that we found in the first part of the proof.

Finally, we argue that if $E$ happens, in expectation more than an $\eps$ fraction of the $m$ examples would game. Since there is a total probability mass of $32\eps$ distributed uniformly over gaming examples with designated points $\in P_G$, and $L$ includes at least $1/16$ fraction of the points from $P_G$, in expectation an $(32\eps)(1/16)=2\eps$ fraction of the examples would game.

\end{proof}

\section{Algorithmic Results Specific to the Linear Model}
\label{sec:ocagi_linear}

The algorithmic results provided so far work in both the general discrete and the linear discrete models. In this section we focus on the linear model and provide algorithmic results for various problems. These algorithms do not follow the greedy structure of the previous algorithms, and use novel technical ideas. First, we consider the problem of designing \emph{linear classifiers}. Section~\ref{sec:ocagi_linear_classifier_properties} provides introductory observations and definitions about linear classifiers. Section~\ref{sec:ocagi_linear_classifier_result} presents the main result of this section which determines whether there exists a linear classifier that classifies all agents accurately and causes all improvable agents to become qualified. Section~\ref{sec:ocagi_linear_classifier_2dim} provides a linear classifier maximizing the number of true positives minus false positives in the two-dimensional case. Then, we shift focus to general (not necessarily linear) classifiers in a two-dimensional space and in Section~\ref{sec:ocagi_two-dimension} provide an algorithm for maximizing true positives subject to no false positives.

\subsection{Properties of Linear Classifiers}
\label{sec:ocagi_linear_classifier_properties}

Before diving into discussion of the algorithmic results, we provide observations about linear classifiers to set the context. We also provide optimal classifiers in special cases. 

For the following discussion, consider linear classifier $\f^{\star}: \vec{a}^{\star}\vec{x} \geq b^{\star}$ that separates the truly qualified agents from unqualified agents.

\begin{observation}
\label{obs:move} 
With linear classifier $\f:\vec{a}\vec{x} \geq b$, any utility maximizing agent that achieves non-negative utility by changing their features moves in dimension $\argmax_j \vecj{a}/\vecj{c}$. \end{observation}

\begin{definition}[Movement dimension]
\label{def:move_dim}
The movement dimension of linear classifier $\f:\vec{a}\vec{x} \geq b$ is the utility maximizing dimension $\argmax_j \vecj{a}/\vecj{c}$ discussed in Observation~\ref{obs:move}. If there are multiple such dimensions the ties are broken in favor of improvement dimensions and then lexicographically.
\end{definition}

\begin{definition}[Encourage improvement/gaming]
\label{def:encourage}
A classifier encourages improvement if its movement dimension is an improvement dimension. It encourages gaming otherwise.
\end{definition}

\begin{definition}[The dim-$j$ improving] 
A linear classifier is dim-$j$ improving if it encourages improvement  and its movement dimension is along dimension $j$.
\end{definition}

The following definition captures the set of agents that potentially can improve to become truly qualified.

\begin{definition}[Improvement margin, improvable agents] 
The improvement margin includes all the agents that can afford (do not have to incur a cost of more than $1$) to move in an improvement dimension and become truly qualified. Formally, any initially unqualified agent $i$, i.e., $\vec{a}^{\star}\init{x}_i < b^{\star}$, that has distance $\leq 1/\vecj{c}$ along an improvement dimension $j$ to $\f^{\star}$ is in the improvement margin.
\end{definition}

\begin{lemma}
\label{lem:all-improve}
If $\f^{\star}:\vec{a}^{\star} x \geq b^{\star}$ encourages improvement, the optimal classifier is $\f^{\star}$---among all linear or nonlinear classifiers. 
\end{lemma}

\begin{proof} 
$\f^{\star}$ classifies initially qualified agents and unqualified unimprovable agents accurately. Also, all the agents in the improvement margin improve, become qualified, and are accurately classified as positive. 
\end{proof}

\begin{lemma}
\label{lem:shifted_opt}
Let $j$ be the movement dimension of classifier $\f^{\star}$. The classifier $\g: \vec{a}^{\star}\vec{x} \geq b^{\star} + \vecstarj{a}/\vecj{c}$ classifies all the initially qualified agents as positive and the rest as negative.
\end{lemma}
\begin{proof}
Initially unqualified agents, $\vec{a}^{\star}\init{x}_i < b^{\star}$, can move at most $1/\vecj{c}$ in dimension $j$ which is not enough to reach to $\g$. Therefore, these agents are classified as negative by $\g$. On the other hand, initially qualified agents, $\vec{a}^{\star}\init{x}_i \geq b^{\star}$, afford to reach to $\g$ and receive nonnegative utility. Therefore, they will be classified as positive.
\end{proof}

\begin{corollary}
\label{cor:all-game}
If all the dimensions are gaming dimensions, $\g: \vec{a}^{\star}\vec{x} \geq b^{\star} + \vecstarj{a}/\vecj{c}$ is the optimal classifier, where $j$ is the movement dimension of $\f^{\star}$.
\end{corollary}
\begin{proof}
If all dimensions are gaming dimensions, there are no improvable agents. Therefore, all agents are either initially qualified or unimprovable and unqualified. By Lemma~\ref{lem:shifted_opt}, $\g$ classifies all such agents accurately.
\end{proof}

By Lemma~\ref{lem:shifted_opt}, $\g: \vec{a}^{\star}\vec{x} \geq b^{\star} + \vecstarj{a}/\vecj{c}$ may be a ``reasonable" solution because it classifies all the initially qualified as positive and does not result in any false positive classifications. However, it misses out on any new true positives resulting from encouraging agents to become qualified. From this point on, we aim to study other classifiers (not necessarily parallel to $\f^{\star}$) with the hope of encouraging other agents to become qualified.

\subsection{Linear Classifier for Improvable Agents}\label{sec:ocagi_linear_classifier_result}

In this subsection, we study a problem that takes as input three disjoint subsets of the agents, $\mathcal{S}^\text{yes}$, $\mathcal{S}^\text{no}$, and $\mathcal{S}^\text{imp}$, and outputs a linear classifier (if one exists) that satisfies the following properties.

\begin{enumerate}[label=\roman*]
\item \label{prop:yes} Classifies agent $i$ such that $\init{x}_i \in \mathcal{S}^\text{yes}$ as positive.
\item \label{prop:no} Classifies agent $i$ such that $\init{x}_i \in \mathcal{S}^\text{no}$ as negative.
\item \label{prop:imp} Encourages agent $i$ such that $\init{x}_i \in \mathcal{S}^\text{imp}$ to improve and become truly qualified, i.e., $\true{x}_i \in \q$, and classifies $i$ as positive.
\end{enumerate}

The main result of the section is solving this problem in polynomial time. When $\mathcal{S}^\text{yes}$ is the set of initially qualified agents, $\mathcal{S}^\text{no}$ is the set of unqualified and unimprovable, and $\mathcal{S}^\text{imp}$ is the set of improvable agents, this problem determines whether there exists a linear classifier that classifies $\mathcal{S}^\text{yes}$ and $\mathcal{S}^\text{no}$ accurately and makes all the improvable agents qualified.

To solve this problem, we divide it into subproblems as following: Does there exist a linear classifier with movement direction in \emph{dimension $j$} that satisfies properties \ref{prop:yes}, \ref{prop:no}, and \ref{prop:imp}? If the answer is ``yes" for some dimension $j$, then the answer to the main problem is ``yes". If the answer is ``no" for all $1\leq j \leq d$, no linear classifier satisfying the three properties exists. 

Note that if $\mathcal{S}^\text{imp}$ is nonempty, in order to satisfy property \ref{prop:imp}, dimension $j$ must be an improvement dimension. Therefore, we study the following problem. 

\begin{problem}
\label{prob:dimj}
Does there exist a dim-$j$ improving classifier (a linear classifier encouraging improvement \emph{in dimension $j$}) that satisfies properties \ref{prop:yes}, \ref{prop:no}, and \ref{prop:imp}?
\end{problem}
We propose a linear program that solves Problem~\ref{prob:dimj}.
The following definition and observations illustrate the conditions under which a dim-$j$ improving classifier satisfies each property for agent $i$.

\begin{figure}[ht]
\includegraphics[width=7cm]{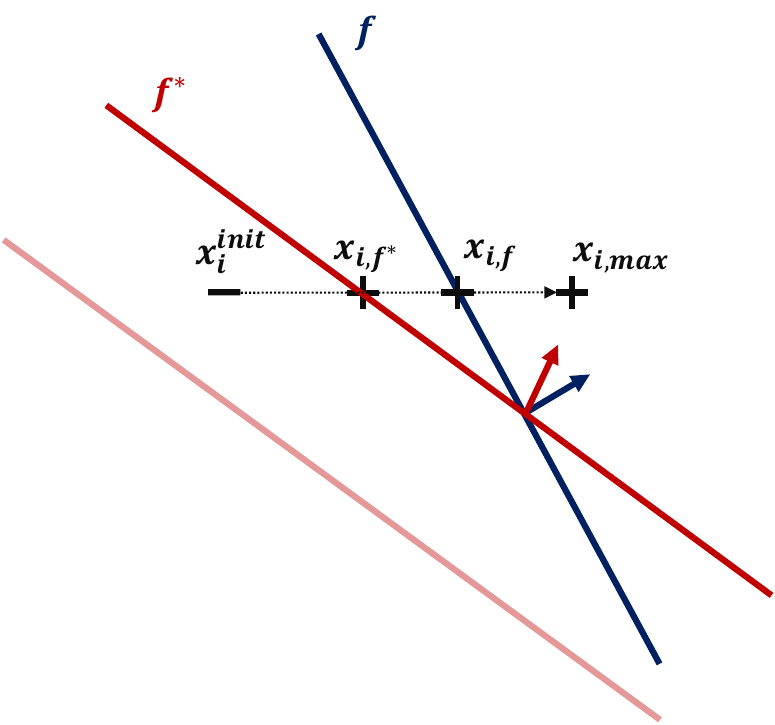}
\centering
\caption[Depiction of a dim-$j$ improving classifier]
{Depicting $\init{x}_i, \vec{x}_{i,\f^{\star}}, \vec{x}_{i,\f}, \vec{x}_{i,\text{max}}$ in Definition~\ref{def:xif} and Observation~\ref{obs:xif}. The horizontal axis shows dimension $j$ in the definition.
}\label{fig:improve_vs_game2}
\end{figure}

\begin{definition}
\label{def:xif}
For a fixed improvement dimension $j$ and classifiers $\f^{\star}:\vec{a}^{\star}\vec{x} \geq b^{\star} $ and $\f:\vec{a}\vec{x} \geq b $, the points $\vec{x}_{i,\f^{\star}}$, $\vec{x}_{i,\f}$, $\vec{x}_{i,\text{max}}$ are defined as follows (depicted in Figure~\ref{fig:improve_vs_game2}.):
\begin{itemize}
\item $\vec{x}_{i,\f^{\star}}$ is the projection of $\init{x}_i$ on the separating hyperplane of classifier $\f^{\star}$ along dimension $j$.
\item $\vec{x}_{i,\f}$ is the projection of $\init{x}_i$ on the separating hyperplane of classifier $\f$ along dimension $j$.
\item $\vec{x}_{i,\text{max}}$ is the shifted $\init{x}_i$ along dimension $j$ by $1/\vecj{c}$.
\end{itemize}
More formally, for all coordinates $ k\neq j$, we have $ \vec{x}_{i,\f^{\star}}[k] = \vec{x}_{i,\f}[k] = \vec{x}_{i,\text{max}}[k] = \init{x}_i[k]$. Also, since $\vec{a}^{\star} \vec{x}_{i,\f^{\star}} = b^{\star}$, we have
$\vec{x}_{i,\f^{\star}}[j] = {\left(b^{\star}-\sum_{k\neq j}\vecstark{a} \init{x}_i[k]\right)}/{\vecstarj{a}}$. Similarly, since $\vec{a} \vec{x}_{i,\f} = b $, we have
$\vec{x}_{i,\f}[j] = {\left(b-\sum_{k\neq j}\vecstark{a} \init{x}_i[k]\right)}/\vecj{a}$. Finally, $\vec{x}_{i,\text{max}}[j] = \init{x}_i[j] + 1/\vecj{c}$. 
\end{definition}

\begin{observation}
\label{obs:all_positive}
A dim-$j$ improving classifier $\f:\vec{a}\vec{x} \geq b$ classifies agent $i$ as positive (property \ref{prop:yes}) if $\vec{a}\vec{x}_{i,\text{max}} \geq b$. It classifies agent $i$ as negative (property \ref{prop:no}) if $\vec{a}\vec{x}_{i,\text{max}} < b$.
\end{observation}
\begin{observation}\label{obs:xif}
Using a dim-$j$ improving classifier $\f$, agent $i$ becomes qualified and is classified as positive (property \ref{prop:imp}) if and only if $\vec{x}_{i,\f^{\star}}[j] \leq \vec{x}_{i,\f}[j] \leq \vec{x}_{i,\text{max}}[j]$. See Figure~\ref{fig:improve_vs_game2}.
\end{observation}

\begin{proposition}\label{pr:three_set_LP}
The following LP captures Problem~\ref{prob:dimj}, where the variables are $\vec{a}$ and $b$.
\begin{align}
\frac{\veck{a}}{\veck{c}} &\leq
\frac{\vecj{a}}{\vecj{c}} && \forall k\neq j \label{LP:game-vs-improve-cond1}\\
b &\leq \vec{a} \vec{x}_{i,\text{max}}   && \forall \init{x}_i \in \mathcal{S}^\text{yes}\label{LP:game-vs-improve-cond2}\\
\vec{a} \vec{x}_{i,\text{max}}  &< b && \forall \init{x}_i \in \mathcal{S}^\text{no}\label{LP:game-vs-improve-cond3}\\
\vec{x}_{i,\f^{\star}}[j] &\leq \vec{x}_{i,\f}[j] && \forall \init{x}_i \in \mathcal{S}^\text{imp}\label{LP:game-vs-improve-cond4}\\
\vec{x}_{i,\f}[j] &\leq \vec{x}_{i,\text{max}}[j] && \forall \init{x}_i \in \mathcal{S}^\text{imp}\label{LP:game-vs-improve-cond5}
\end{align}
\end{proposition}
Constraint \ref{LP:game-vs-improve-cond1} asserts that the movement direction of the classifier is along dimension $j$. Constraint \ref{LP:game-vs-improve-cond2} asserts property \ref{prop:yes}. Constraint \ref{LP:game-vs-improve-cond3} asserts property \ref{prop:no}. Finally, constraints \ref{LP:game-vs-improve-cond4} and \ref{LP:game-vs-improve-cond5} assert property \ref{prop:imp}.

\begin{theorem}\label{thm:lp}
Given the sets $\mathcal{S}^\text{yes}$, $\mathcal{S}^\text{no}$, and $\mathcal{S}^\text{imp}$, there is a polynomial-time algorithm that outputs a linear classifier (if one exists) that satisfies Properties \ref{prop:yes}, \ref{prop:no},\ref{prop:imp}, or declares non-existence of such a classifier.
\end{theorem}
\begin{proof}
If $\mathcal{S}^\text{imp} \neq \emptyset$, run LP \ref{LP:game-vs-improve-cond1}-\ref{LP:game-vs-improve-cond5} for all improvement dimensions $j$. If $\mathcal{S}^\text{imp} = \emptyset$, run the LP for $1 \leq j \leq n$. By Problem~\ref{pr:three_set_LP}, if there exist feasible solution $\vec{a}$ and $b$ for one of these LPs, $\f: \vec{a}\vec{x} \geq b$ is a classifier satisfying properties \ref{prop:yes}, \ref{prop:no}, and \ref{prop:imp}. 
\end{proof}
\begin{corollary}
There is a polynomial-time algorithm that determines whether there exists a linear classifier that classifies the initially qualified as positive,  unqualified unimprovable agents as negative, encourages the agents in the improvement margin to improve to become qualified, and classifies them as positive. If such a classifier exists, it maximizes true positives subject to no false positives.
\end{corollary}

\begin{remark}
Theorem~\ref{thm:relaxed-no-destination-pts} asserts that
given the initial feature vectors of agents, $\init{x}_1, \init{x}_2, \ldots,$ $\init{x}_n\in \mathbb{R}^d$, deciding whether there exists a classifier for which all the agents become true positives is NP-hard. However, when limiting to linear classifiers this problem is no longer NP-Hard. Using Theorem~\ref{thm:lp}, by setting $\mathcal{S}^\text{yes}$ to the set of initially qualified agents, and $\mathcal{S}^\text{imp}$ to the rest of the agents, this problem is solvable in polynomial time. 
\end{remark}

\subsection{Optimal Linear Classifier in Two-Dimensional Space}
\label{sec:ocagi_linear_classifier_2dim}

In this subsection we continue considering linear classifiers but focus on the two-dimensional case. The main result is an algorithm (Algorithm~\ref{alg:g-i-classifiers}) for finding a linear classifier that maximizes the number of true positives minus the number of false  positives.

First, note that if $\f^{\star}: \vec{a}^{\star}\vec{x}\geq b^{\star}$ encourages improvement, then the optimal linear classifier is $\f^{\star}$ (Lemma~\ref{lem:all-improve}). Also, if 
both dimensions are gaming dimensions, then a shifted $\f^{\star}$ is optimal (Corollary~\ref{cor:all-game}). Therefore, for the rest of this subsection we focus on the case where (1) dimension one (horizontal axis) is an improvement dimension, (2) dimension two (vertical axis) is a gaming dimension, and (3) $\f^{\star}$ encourages movement along the gaming dimension.

The following observation determines the agents that contribute to the true positives or false positives of a linear classifier.

\begin{figure}[ht!]
\includegraphics[width=6.6cm]{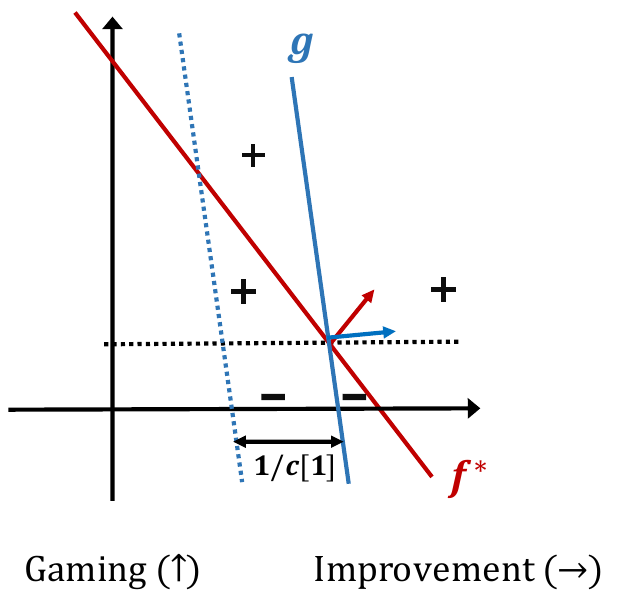}
\centering
\caption[Areas of initial position of agents that contribute to true and false positives]
{Related to Observation~\ref{obs:true_false_positive}. This figure identifies the areas of initial position of agents that contribute to true positives, depicted by ``$+$", and false positives, depicted by ``$-$", when using classifier $\g$ that encourages improvement along dimension $1$ (horizontal axis).
}\label{fig:true_false_positives}
\end{figure}

\begin{observation}[true positives and false positives]
\label{obs:true_false_positive}
Consider classifier $\g: \vec{a}\vec{x} \geq b$ that encourages improvement along dimension 1 (horizontal axis).
In Figure~\ref{fig:true_false_positives}, we divide $\mathbb{R}^2$ into sub-areas and identify the areas of initial positions of agents that contribute to true positives and false positives. By Observation~\ref{obs:all_positive}, each agent $i$ classified positive by $\g$ satisfies $\vec{a}\init{x}_i + \vec{a}[1]/\vec{c}[1] \geq b$. This is the set of agents with initial position at most at distance $1/\vec{c}[1]$ horizontally from $\g$ (the area on the right side of the dotted blue line in Figure~\ref{fig:true_false_positives}).
Each agent classified as positive by $\g$, is either a true positive or a false positive. Using Observation~\ref{obs:xif}---which identifies the initial positions of agents that improve, become qualified, and get classified as positive---any part above (with higher value in dimension $2$ than) the intersection of $\g$ and $\f^{\star}$, on the right of the dotted blue line contributes to true positives (depicted by ``$+$" in the figure). On the other hand, any part below (with less value in dimension $2$ than) the intersection of $\g$ and $\f^{\star}$, on right of the dotted blue line, and the left side of $\f^{\star}$ contributes to false positives (depicted by ``$-$" in the figure).
\end{observation}

In the following lemma, we find a dominating set of linear classifiers.

\begin{lemma}
\label{lem:best_slope}
There exists a linear classifier  $\g:\vec{a}\vec{x} \geq b$ satisfying $\vecone{a}/\vecone{c} = \vectwo{a}/\vectwo{c}$ that  maximizes the number of true positives minus false positives in $\mathbb{R}^2$ among improvement-encouraging linear classifiers.
\end{lemma}

\begin{figure}[ht]
\includegraphics[width=6.6cm]{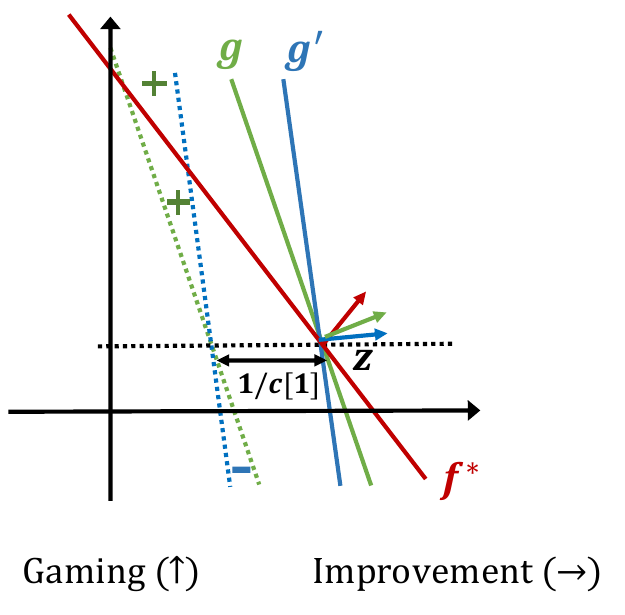}
\centering
\caption[Classifiers that encourage improvement along the horizontal axis]
{Related to Lemma~\ref{lem:best_slope}. $\g$ and $\g'$ are classifiers that encourage improvement along dimension $1$ (horizontal axis). The area between the dotted blue and green lines above the intersection shows the area of initial positions of agents that contribute to true positives by $\g$ and not by $\g'$---it is denoted by ``$+$". The area between the two lines below the intersection shows the area of initial positions of agents that contribute to false positives by $\g'$ and not by $\g$---it is denoted by ``$-$".
}\label{fig:optimal_slope}
\end{figure}

\begin{proof}
Consider an arbitrary linear classifier $\g':\vec{a}'\vec{x} \geq b'$ that encourages improvement. By Observation~\ref{obs:move}, $\vec{a}'[1]/\vec{c}[1]\geq \vec{a}'[2]/\vec{c}[2]$. Consider point $\vec{z}$ where $\g'$ intersects with $\f^{\star}$. Let $\g:\vec{a}\vec{x} \geq b$ be the classifier satisfying $\vecone{a}/\vecone{c} = \vectwo{a}/\vectwo{c}$ that passes through $\vec{z}$. We show $\g$ has objective value at least as that of $\g'$. 
Note that, by construction, $\g$ lies between $\g'$ and $\f^{\star}$ in $\mathbb{R}^2$; in Figure~\ref{fig:optimal_slope}, the blue classifier, the green classifier, and the red classifier illustrate $\g'$, $g$, and $\f^{\star}$, respectively. Observation~\ref{obs:true_false_positive} determines the areas of true positives and false positives based on the initial positions of the agents and the classifier in use. By Observation~\ref{obs:true_false_positive}, $\g$ includes all the true positives of $\g'$ as well as the area between the dotted blue and green lines above their intersection; this part is illustrated by ``$+$" in Figure~\ref{fig:optimal_slope}. Also, $\g'$ includes all the false positives of $\g$ as well as the area between the blue and green dotted lines below their intersection; this part is illustrated by ``$-$" in Figure~\ref{fig:optimal_slope}. Since the true positive area of $\g$ is a superset and its false positive area is a subset compared to $\g'$, it has weakly higher objective value.
\end{proof}

\paragraph{Overview of Algorithm~\ref{alg:g-i-classifiers}.}  
Among potential linear classifiers, the algorithm finds the best of gaming-encouraging and improvement-encouraging linear classifiers, and outputs the better of the two in terms of the objective function. The best gaming-encouraging classifier is given in Corollary~\ref{cor:all-game}. The best improvement-encouraging classifier is found as follows: Using Lemma~\ref{lem:best_slope}, the optimal slope of the classifier is known. Therefore, we only need to determine a crossing point to determine the classifier. For all agents $i$, consider linear classifier $g:\vec{c}^{\star}\vec{x} \geq \vec{c}\init{x}_i + 1$. This classifier satisfies the optimal slope from Lemma~\ref{lem:best_slope} and is at distance $1/\vecone{c}$ horizontally from $\init{x}_i$. Therefore, $i$ is the farthest agent to reach the classifier and be classified positive. Among these classifier, find one that maximizes the objective value---this is the optimum improvement classifier.

\begin{algorithm}[!ht]
    \SetNoFillComment
    \SetAlgoLined
    \SetKwInOut{Input}{Input}
    \SetKwInOut{Output}{Output}
    \SetKw{Return}{return}
     \DontPrintSemicolon
    \Input{Initial positions of agents:  $\mathcal{X} \subseteq \mathbb{R}^2$ \tcp{dim1 is improvement. dim2 is gaming.}}
    \Input{Linear model: $\f^{\star}:\vec{a}^{\star} \vec{x} \geq b^{\star}$ \tcp{$\f^{\star}$ encourages movement in dim2.}}
            \Output{$\f$ \tcp{best linear classifier}}
    $\text{objective} = 0$ \tcp{maximum objective value observed}
    $f:\vec{a}^{\star}\vec{x} \geq b^{\star} +  \vec{a}^{\star}[2]/\vectwo{c}$\ \tcp{best gaming-encouraging linear classifier}
    $\text{objective} = \text{count\_true\_positives}(f) - \text{count\_false\_positives}(f)$\;

    \For{$i=1, 2, \cdots, |\mathcal{X}|$}{ 
        $g:\vec{c}\vec{x} \geq \vec{c}\init{x}_i + 1$ \tcp{ classifier with optimal slope corresponding to agent $i$}
        $\text{diff} = \text{count\_true\_positives}(g) - \text{count\_false\_positives}(g)$\;
        \If{$\text{diff} > \text{objective}$}{
            $\text{objective} = \text{diff}$\;
            $f = g$\;
        }
    }
    \Return $f$\;
    \caption{Find a linear classifier that maximizes true positives minus false positives}
    \label{alg:g-i-classifiers}
\end{algorithm}

\begin{theorem}
Algorithm~\ref{alg:g-i-classifiers} finds the linear classifier maximizing the number of true positives minus false positives in $\mathbb{R}^2$, when dimension 1 is improvement and dimension 2 is gaming. When both dimensions are improvement or gaming, the optimal classifiers are given in Lemma~\ref{lem:all-improve} and Corollary~\ref{cor:all-game}.
\end{theorem}
\begin{proof}
The algorithms considers the best of two groups of linear classifiers---gaming-encouraging and improvement-encouraging. The best gaming-encouraging classifier is given in Corollary~\ref{cor:all-game}. Among the improvement-encouraging classifiers, Lemma~\ref{lem:best_slope} gives the optimal slope of the classifier. We claim we only need to consider classifiers with the optimal slope such that some agent $i$ is exactly at distance $1/\vecone{c}$ horizontally from the classifier -- a classifier where $i$ is the farthest agent that will be classified positive. Consider the set of such parallel classifiers $\f_1, \f_2, \ldots, \f_m$, where they are sorted based on their $y$-intercept such that $\f_1$ is the classifier corresponding to agent $i = \argmax \vec{c}\init{x}_i$, and $\f_m$ is the classifier corresponding to agent $i = \argmin \vec{c}\init{x}_i$. To prove the claim, we show any other classifier with the optimal slope is dominated by the classifiers we consider. First, note that since any potential parallel classifier strictly between $\f_j$ and $\f_{j+1}$ classifies the agents exactly as in $\f_j$, these classifiers are dominated and we do not need to consider them. Secondly, we do not need to consider any other parallel classifier with $y$-intercept less than $\f_m$ or higher than $\f_1$. Any parallel classifier with smaller intercept than $\f_m$ has the same classification as $\f_m$ (classifying all agents as positive). Also, any parallel classifier with larger intercept than $\f_1$ does not classify any agents as positive; therefore, has objective value $0$, and is dominated in terms of objective value by the best gaming-encouraging classifier.
\end{proof}

\begin{remark}
Theorem~\ref{thm:atmost_k_game} implies maximizing the objective function considered in this subsection --maximizing the number of true positives minus false positives-- is NP-hard in $\mathbb{R}^d$ when $d$ is not a constant and we are not limited to linear classifiers.
\end{remark}

\subsection{Optimal General Classifier in Two-Dimensional Space}
\label{sec:ocagi_two-dimension}

In this subsection, we consider the problem of maximizing true positives subject to no false positives in a $2$-dimensional space, where the horizontal dimension is improvement, and the vertical dimension is gaming.
We provide an algorithm in the linear model that given a set of agents, returns a set of target points $\pfinal\subset \mathbb{R}^2$ that maximizes true positives subject to no false positives. 
Note that unlike Algorithm~\ref{alg:maximal}, our algorithm in this subsection does not take a finite set of target points $\points$ as input.
For simplicity, by scaling we may assume wlog that $c = \vec{c}[1]= \vec{c}[2]$.

\begin{figure}[ht!]
    \centering
    \includegraphics[width=9cm]{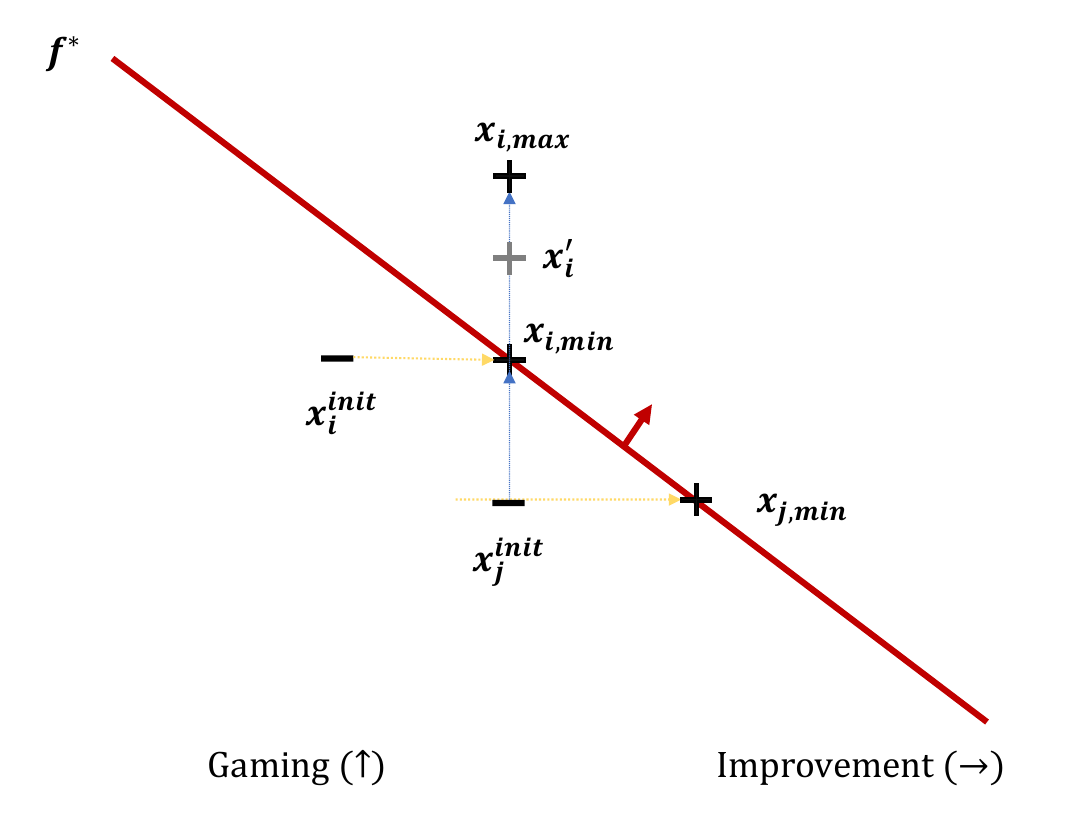}
    \caption[$\vec{x'}_i$ is pushed along the gaming dimension so $\init{x}_j$ no longer moves to it]
    {In Algorithm~\ref{alg:2-dimension}, $\vec{x'}_i$ is pushed along the gaming dimension so $\init{x}_j$ no longer moves to it.
    }\label{fig:2-dim-alg}
\end{figure}

\paragraph{Overview of Algorithm~\ref{alg:2-dimension}.} 
First, all the points $\init{x}_{i}$ for $1\leq i \leq m$ are sorted along the gaming dimension in a descending order, such that $\init{x}_{n}$ has the smallest value in the gaming dimension. Our goal is to find designated points, $\vec{x'}_i$, for each $\init{x}_i$. Starting with $\init{x}_n$, for each point $\init{x}_i$, move $\init{x}_i$ along the improvement dimension until it crosses the line $\vec{a^{\star}}\vec{x}=b^{\star}$ at $\vec{x}_{i,min}$ (See Figure~\ref{fig:2-dim-alg}). Let $\vec{x'}_i$, the designated point of $\init{x}_i$, be initially $\vec{x'}_i = \vec{x}_{i,min}$. If given the current set of designated points for agents $n, n-1, \ldots, i$, another point $\init{x}_j$ for $j>i$ maximizes utility by moving to $\vec{x'}_i$ and becomes false positive, push $\vec{x'}_i$ upward along the gaming dimension, until $\init{x}_j$ no longer picks $\vec{x'}_i$. When pushing $\vec{x'}_i$ along the gaming dimension, let $\vec{x}_{i,max}$ denote the furthest point that $\init{x}_i$ can afford to reach to it. If the final point $\vec{x'}_i$ is such that $\init{x}_i$ cannot afford to move to it, i.e. $\vec{x'}_i[2]>\vec{x}_{i,max}[2]$, discard $\vec{x'}_i$. Otherwise, $\vec{x'}_i$ is added to $\pfinal$.

Note that we assume that if a point $\init{x}_j$ can improve to $\vec{x'}_j$ and game to $\vec{x'}_i$ with the same cost, it would pick the improvement option.

\begin{algorithm}
    \SetNoFillComment
    \SetAlgoLined
    \SetKwInOut{Input}{Input}
    \SetKwInOut{Output}{Output}
    \SetKw{Continue}{continue}
    \SetKw{Return}{return}
    \Input{$\agents$, $f^{\star}: \vec{a}^{\star}\vec{x}\geq b^{\star}$}
    \Output{$\mathcal{P}^{\text{final}}$}
    Sort $\vec{x}_i\in \agents$ in a descending order of $\vec{x}_i[2]$\;
    \For{$i=n,\cdots,1$}{
        \tcc{Let $\vec{x}_{i,min}$ be the projection of $\vec{x}_i$ on $\vec{a}^{\star}\vec{x}=b^{\star}$ along the improvement dimension}
        $\vec{x}_{i,min} = \Big(\frac{b^{\star}-\vec{a}^{\star}[2]\vec{x}_i[2]}{\vec{a^{\star}}[1]},\vec{x}_i[2]\Big)$\;
        \If{$\vec{x}_{i,min}[1]-\vec{x}_i[1]>1/c$}{
        \tcc{$\vec{x}_i$ cannot become true positive.}
            \Continue\;
        }
        $\vec{x'}_i \leftarrow \vec{x}_{i,min}$\;
        \For{$j=n,\cdots,i+1$}{
            \If{$cost(\vec{x}_j,\vec{x}'_j)>cost(\vec{x}_j,\vec{x}'_i)$}{
                $\vec{x}'_i \leftarrow (\vec{x}'_i[1], \vec{x}'_i[2]+cost(\vec{x}_j,\vec{x}'_j)-cost(\vec{x}_j,\vec{x}'_i))$\;
            }
        }
        \If{$\vec{x'}_i[2] > \vec{x}_{i,max}[2]$}{
            \tcc{$\vec{x}_i$ cannot become true positive without another point becoming false positive.}
           $\vec{x'}_i = (\vec{x'}_i[1], \infty)$\;
        }
        $\mathcal{P}^{\text{final}}\leftarrow \mathcal{P}^{\text{final}}\cup \vec{x'}_{i}$\;
        \Return $\mathcal{P}^{\text{final}}$\;
    }

\caption{Maximizing the number of true positives in $2$-dimensions.}   
\label{alg:2-dimension}
\end{algorithm}
In order to show that Algorithm~\ref{alg:2-dimension} maximizes true positives subject to no false positives, we need the following observation and lemma.

\begin{observation}
Line $\vec{a}^{\star}\vec{x}=b^{\star}$ has a negative slope, i.e., each feature is defined so that larger is better. Therefore, after the points in $\agents$ are sorted, if an agent $\init{x}_j$ where $j<i$ reaches to any point $\vec{x'}_i\in [\vec{x}_{i,min}, \vec{x}_{i,max}]$, then $\init{x}_j$ becomes true positive. On the other hand, for $j>i$, if $\init{x}_j$ moves to any point $\vec{x'}_i\in[\vec{x}_{i,min}, \vec{x}_{i,max}]$, then $\vec{x}_j$ becomes false positive.
\label{observation:2-dimension-negative-slope}
\end{observation}

\begin{lemma}
Consider a point $\vec{p}$ such that $\vec{p}[1]\geq \vec{x}_{i,min}[1]$, and another point $\vec{q}\in [\vec{x}_{i,min},\vec{x}_{i,max}]$. Suppose $cost(\init{x}_i, \vec{p}) = cost(\init{x}_i, \vec{q})$. Then, for any $j>i$, it is the case that $cost(\init{x}_j, \vec{p})\leq cost(\init{x}_j, \vec{q})$.
\label{lemma:2-dim-triangle-inequalities}
\end{lemma}

\begin{proof}
Proof is deferred to Appendix~\ref{appendix:proof-lemma-2-dim-triangle-inequalities}.
\end{proof}

\begin{theorem}
Given initial feature vectors of agents, $\init{x}_1, \init{x}_2, \ldots, \init{x}_n \in \mathbb{R}^2$, Algorithm~\ref{alg:2-dimension} maximizes the number of true positives subject to no false positives.
\end{theorem}

\begin{proof}
Suppose not.  Let $\vec{x}_1^{\opt}, \ldots,\vec{x}_n^{\opt}$ be an optimal solution that agrees with $\vec{x'}_1, \ldots, \vec{x'}_n$ on as large a suffix as possible, and let $i$ be the largest index such that $\vec{x}_i^{\opt}\neq \vec{x}_i'$ (so $\vec{x}_j^{\opt}= \vec{x}_j'$ for all $j>i$).  

First, note that $i\neq n$.  This is because $\vec{x'}_n = \vec{x}_{n,min}$, which is the cheapest point that agent $n$ can reach to become a true positive; moreover, any other point moving to $\vec{x}_n'$ is a true improvement.  So, replacing $\vec{x}_n^{\opt}$ with $\vec{x'}_n$ only helps.

Next, we claim that even if $i<n$, replacing $\vec{x}_i^{\opt}$ with $\vec{x'}_i$ can only improve the optimal solution.  First, if $cost(\init{x}_i, \vec{x}_i^{\opt}) \geq cost(\init{x}_i, \vec{x'}_i)$ then replacing $\vec{x}_i^{\opt}$ with $\vec{x'}_i$ only helps by the same argument as above and the fact that $\vec{x'}_i$ was chosen so that no agent $j>i$ manipulates to it; here we are using the fact that the suffixes of the two solutions agree.  On the other hand, suppose that $cost(\init{x}_i, \vec{x}_i^{\opt}) <  cost(\init{x}_i, \vec{x'}_i)$ and $cost(\init{x}_i, \vec{x}_i^{\opt}) \leq 1/c$.  Since $\init{x}_i$ cannot become a false positive by moving to $\vec{x}_i^{\opt}$, this means that  $\vec{x}_i^{\opt}[1]\geq \vec{x}_{i,min}[1]$. There exists a point $\vec{q}\in [\vec{x}_{i,min}, \vec{x}_{i,max}]$ such that $cost(\init{x}_{i}, \vec{x}_i^{\opt}) = cost(\init{x}_{i}, \vec{q})$, which implies that $cost(\init{x}_{i}, \vec{q}) < cost(\init{x}_{i}, \vec{x'}_{i})$. The reason that $\vec{q}$ was not selected as $\vec{x'}_i$ is that there exists an agent $\init{x}_j$ where $\init{x}_j$ moves to $\vec{q}$ and becomes false positive. By Observation~\ref{observation:2-dimension-negative-slope},  $j>i$. Hence, $cost(\init{x}_j, \vec{q})<cost(\init{x}_j, \vec{x'}_j)$ and $cost(\init{x}_j, \vec{q}) \leq 1/c$.
By Lemma~\ref{lemma:2-dim-triangle-inequalities}, $cost(\init{x}_j, \vec{x}_i^{\opt})\leq cost(\init{x}_j, \vec{q})$, so
$cost(\init{x}_j, \vec{x}_i^{\opt}) < cost(\init{x}_j, \vec{x'}_j)$ and $cost(\init{x}_j, \vec{x}_i^{\opt}) \leq 1/c$.
Hence, $\init{x}_j$  is closer to $\vec{x}_i^{\opt}$ compared to $\vec{x'}_j=\vec{x}_j^{\opt}$ and so agent $j$ would become a false positive under $\opt$, which contradicts the definition of $\opt$. 
So, this second case cannot occur.  

Therefore, Algorithm~\ref{alg:2-dimension} maximizes the number of true positives subject to having no false positives.
\end{proof}

\begin{remark}
By Corollary~\ref{cor:d-dim-max-TP-no-FP}, this problem is NP-hard when $\agents \subset \mathbb{R}^d$ for general (not constant) $d$.
\end{remark}

\chapter{Empirical Study on the Classification of Improving-and-Gaming Agents}
\label{chap:ocagi_exp1}

\section{Introduction}
\label{sec:ocagiexp_introduction}

In this Chapter, we conduct an empirical study on the classification of data points that act as agents that can both strategically game and improve their features within an $\ell_{\infty}$ or $\ell_{2}$ ball of radius $r$.  In particular, we evaluate the loss and threshold strategic classifier designs that incorporate knowledge of agents' behavior in classification. These approaches are broadly aligned with the framework of \citet{ahmadi2022classificationstrategicagentsgame}, which establishes a theoretical foundation for the accurate classification of all agents, penalizing gaming and incentivizing all improvable agents to become qualified. Consistent with this line of work, the decision-maker's objective in the empirical studies is to maximize true positives while minimizing false positives. That is, the decision-maker's utility is defined as follows. 
\begin{definition}[Utility]
\label{def:ocagi_utility}
\begin{equation}
\mathrm{utility} = \frac{\mathrm{TP}-\mathrm{FP}}
{n}, \qquad \mathrm{where}
\label{eq:ocagi_utility}
\end{equation}
the variables $\mathrm{TP}$ and $\mathrm{FP}$ denote the number of true positives and false positives after the agents that changed themselves are classified, while $n$ is the total number of classified agents.
\end{definition}
A utility value of $1$ corresponds to the ideal setting in which all initially negative agents become genuinely qualified through improvement rather than gaming, and all the initially positive agents are correctly classified as positive. 

Although this utility metric is suitable for many classification settings, its limitations become apparent in high-stakes applications where the costs and benefits of prediction outcomes differ substantially. In domains such as marketing and lending, true positive classifications generate significant gains, whereas false positive ones result in substantial losses \citep{Westland2018PrivateIC,marketing17,XU2024103004,VOROBYEV2022102786}. For example, correctly identifying interested customers increases revenue, while targeting disinterested ones wastes marketing resources. Similarly, approving loans for creditworthy applicants yields profit, whereas lending to defaulters leads to financial loss. To account for such asymmetries, we also consider a cost-sensitive utility metric:
\begin{definition}[Weighted utility]
\label{def:ocagi_wutility}
\begin{equation}
\mathrm{utility}(1,8) = \mathrm{TP}\times C_{\mathrm{TP}} - 
\mathrm{FP}\times C_{\mathrm{FP}}, \qquad \mathrm{where}
\label{eq:ocagi_wutility}
\end{equation}
$C_{\mathrm{TP}}=1$ represents the gain associated with a correct positive prediction and $C_{\mathrm{FP}}=8$ represents the penalty associated with a false positive. Note that the selected costs are used for experimental illustration and are not intended to reflect any particular real-world application.
\end{definition}

\section{Experimental Setup}
\label{sec:ocagiexp_setup}

Below, we describe the experimental setup used to examine several factors affecting model performance. Specifically, we investigate whether the model's level of risk aversion influences the rate of increase in the $\textrm{utility}$ score when agents can both game and improve. We also examine whether achieving zero error on classification implies $100\%$ $\textrm{utility}$ score, how the choice of movement constraint norm ($\ell_{\infty}$) versus ($\ell_{2}$) affects error reduction and $\textrm{utility}$ score improvement, and whether restricting agents to improvement-only behavior, as opposed to considering both gaming and improvement, changes the resulting $\textrm{utility}$ score.

All experiments were conducted on a laptop computer equipped with a \(2.6\)-GHz 6-Core Intel Core \(i7\) processor, \(16\) GB of \(2400\)-MHz DDR4 RAM, and an Intel UHD Graphics \(630\) graphics card with \(1536\) MB of memory.

\subsection{Datasets} 
The empirical study is conducted on four datasets: three real-world tabular benchmarks (i.e., the Adult, OULAD, and Law School datasets)  and one synthetic \(8\)-dimensional binary classification dataset. Table~\ref{tab:ocagi_datasets} summarizes the main characteristics of the datasets used in our evaluation. Additional dataset characteristics, including class distributions and separability, are provided in~\cite{Attias2025PACLW}.
Formally, let
\[
\mathcal{X} = \{(x, y) \mid x \in \mathbb{R}^d,\; y \in \{0,1\}\}
\]
denote a dataset (e.g., Adult), where \(x\) is the agent's initial feature vector and \(y = f^\star(x)\) is its associated label. Here, \(f^\star\) denotes a zero-error model trained on the full dataset and used as the ground-truth labeling function.
In all experiments, we partition \(\mathcal{X}\) into training and test subsets, denoted by \(\mathcal{X}_{\mathrm{tr}}\) and \(\mathcal{X}_{\mathrm{ts}}\), comprising \(70\%\) and \(30\%\) of the data, respectively.

\begin{table*}[b!]
\footnotesize
\renewcommand{\arraystretch}{1.08}

\centering
\begin{tabularx}{\textwidth}{
    p{1.5cm}
    p{2.5cm}
    c
    c
    >{\raggedright\arraybackslash}X
}
\toprule
\textbf{Dataset} &
\textbf{Label} &
\(d\) &
\textbf{Train/Test} &
\textbf{Improvable and gameable features} \\
\midrule

Adult &
\makecell[l]{\(1\!:income\!>50K\)\\\(0\!:income\!\leq50K\)} &
14 &
\(21113/9049\) &
\textbf{Imp.:} hours-per-week, capital-gain, capital-loss,
fnlwgt, educational-num, workclass, education,
occupation;
\newline
\textbf{Game:} age, workclass, marital-status,
relationship, race, gender, native-country.
\\

OULAD &
\makecell[l]{\(1\!:score=\!\text{pass}\)\\\(0\!:score=\!\text{fail}\)} &
11 &
\(15093/6469\) &
\textbf{Imp.:} code\_module, code\_presentation,
imd\_band, highest\_education,
num\_of\_prev\_attempts, studied\_credits;
\newline
\textbf{Game:} id\_student, gender, region,
age\_band, num\_of\_prev\_attempts, disability.
\\

Law School &
\makecell[l]{\(1\!:score=\!\text{pass}\)\\\(0\!:score=\!\text{fail}\)} &
11 &
\(14558/6240\) &
\textbf{Imp.:} decile1b, decile3, lsat, ugpa,
zfygpa, zgpa, fulltime, fam\_inc, tier;
\newline
\textbf{Game:} fam\_inc, male, tier, race.
\\

Synthetic &
\makecell[l]{\(1\!:\!\text{positive}\)\\\(0\!:\!\text{negative}\)} &
8 &
\(1561/669\) &
\textbf{Imp.:} feature\_0, feature\_2,
feature\_4, feature\_6;
\newline
\textbf{Game:} feature\_1, feature\_3,
feature\_5, feature\_7.
\\

\bottomrule
\end{tabularx}

\caption{Details of the tabular datasets used in the experiments.}
\label{tab:ocagi_datasets}
\end{table*}

\subsection{\texorpdfstring{The $f^\star$ Model and the Decision-maker's Model ($h$)}{The \textit{f*} Model and the Decision-maker's Model (\textit{h})} }
\label{sec:ocagiexp_classifiers}

For each full dataset \(\mathcal{X}\), except OULAD, we trained a zero-error model \(f^\star\) using a decision tree classifier with the following hyperparameters: criterion = ``gini'', min\_samples\_split = \(2\), min\_samples\_leaf = \(1\), and random\_state = \(42\). For the OULAD dataset, \(f^\star\) was instead implemented as a random forest classifier with default hyperparameters and random\_state = \(42\).

For each dataset, we then trained a decision-maker model \(h: \mathbb{R}^d \to \{0,1\}\), implemented as a two-layer neural network, on \(\mathcal{X}_{\textrm{tr}}\) using dataset-specific tuned hyperparameters. We considered two variants of the decision-maker model: (i) a non-strategic model trained with the standard binary cross-entropy loss (\(\mathcal{L}_{\textrm{BCE}}\)) and (ii) a strategic risk-averse model trained with weighted binary cross-entropy loss (\(\mathcal{L}_{\textrm{wBCE}}\)) (Equation~\ref{eq:ocagi_losses}).
\begin{equation}
\label{eq:ocagi_losses}
\begin{aligned}
\mathcal{L}_{\textrm{wBCE}}
=
- \frac{1}{n} \sum_{i=1}^{n}
\Big[
w_{\textrm{FP}} (1-y_i)\log(1-\hat{y}_i)
+
w_{\textrm{FN}} y_i \log(\hat{y}_i)
\Big]
\end{aligned}
\end{equation}
where \(n = |\mathcal{X}_{\textrm{tr}}|\), \(y_i \in \{0,1\}\) denotes the ground-truth label, \(\hat{y}_i \in (0,1]\) denotes the predicted probability, and \(w_{\textrm{FP}}\) and \(w_{\textrm{FN}}\) are the weights assigned to false positives and false negatives, respectively. Setting \(w_{\textrm{FP}} = w_{\textrm{FN}} = 1\) reduces \(\mathcal{L}_{\textrm{wBCE}}\) to \(\mathcal{L}_{\textrm{BCE}}\).

Another form of strategic model design we consider is threshold-based risk aversion. That is, we classify an agent as positive or negative by applying either the standard non-strategic decision threshold of \(0.5\) or the more risk-averse strategic thresholds of \(0.9\) and \(0.99\) to the sigmoid output of the neural network's final layer.

\subsection{Agents' Improvement and or Gaming}
\label{subsec:ocagiexp_improvegame}

Following \citet{ahmadi2022classificationstrategicagentsgame}, we consider a setting in which features are partitioned into gaming and improvement features.  Consider, for example, a grade-2 enrollment AI model that uses \textit{age}, \textit{household income}, \textit{area code}, and \textit{grade-1 report card} to determine admission. Although all features are important in determining the child's admission, a subset is susceptible to gaming. For instance, a parent may misreport a child's age to appear more qualified, or truthfully report higher household income if doing so improves admission prospects.

Formally, let \(\mathcal{X}_{\mathrm{ts}} = \{(x, y) \mid x \in \mathbb{R}^d, \, y \in \{0,1\}\}\) denote the test-time population, where \(x\) is a feature vector and \(y = f^\star(x)\) is the corresponding ground-truth label. Let \(I\) denote the set of indices corresponding to improvement features and \(G\) the set of indices corresponding to gameable features.
Given a trained model \(h\), an agent \(x \in \mathcal{X}_{\textrm{ts}}\) with an undesirable model outcome \(h(x) = 0\) along with a predefined movement budget \(r\), we use Projected Gradient Descent~\citep{AlexAdversarial} to compute feasible perturbation that produces a modified feature vector \(x' \in \mathbb{R}^d\) satisfying \(h(x') = 1\), while remaining within the budget constraint \(r\). That is, we aim to find
\begin{equation} \label{eq:improve_pgd2}
x' = \mathrm{Proj}_{\Delta(x)} \left( x_{(t)} + \alpha \cdot \mathrm{sign}\!\left(\nabla_{x_{(t)}} \mathcal{L}\big(h(x_{(t)}), h(x)\big)\right) \right),
\end{equation}
where \(t\) indexes the current iteration, \(\alpha\) is the step size, and \(\nabla \mathcal{L}(h(x_{(t)}), h(x))\) denotes the gradient of the loss function. The projection operator \(\mathrm{Proj}_{\Delta(x)}\) enforces the constraint set
\[
\Delta(x) = \{ x_{(t)} \in \mathbb{R}^d : \|x_{(t)} - x\|_{\infty} \leq r \}
\quad \text{or} \quad
\Delta(x) = \{ x_{(t)} \in \mathbb{R}^d : \|x_{(t)} - x\|_{2} \leq r \},
\]
ensuring that updates remain within a ball of radius \(r\) centered at $x$.

To evaluate whether \(f^\star(x') = 1\), we do not use the full vector \(x'\) directly. Instead, we construct a partially immutable representation in which gaming features are held fixed:
\[
[x'_i]_{i \in I} \cup [x'_j]_{j \in G} \quad \text{with} \quad x'_j = x_j \ \forall j \in G.
\]
This ensures that only changes in improvement features contribute to the evaluation of the change in the agent's true label.

Algorithm~\ref{alg:ocagi_improv_gaming} details the full procedure for modeling improvement and gaming within an \(\ell_{\infty}\) or \(\ell_{2}\) ball of radius \(r\) centered at \(x\). We tune the step size \(\alpha\), the number of iterations \(T\), and the budget \(r \in [0,4]\). The loss function is either standard non-strategic binary cross-entropy \(\mathcal{L}_{\mathrm{BCE}}\) or its strategic risk-averse variant \(\mathcal{L}_{\mathrm{wBCE}}\).
When restricting agents to improvement-only behavior, we modify line \(5\) of Algorithm~\ref{alg:ocagi_improv_gaming} as follows:
\[
\rho_{(t)}[i] =
\begin{cases} 
\alpha \cdot \mathrm{sign}(\mathbf{g}_{(t)}[i]), & \text{if } i \in I, \\
0, & \text{otherwise},
\end{cases}
\]

\begin{algorithm}
    \SetAlgoLined
    \SetKwInOut{Input}{Input}
    \SetKwInOut{Output}{Output}
    \SetKw{Continue}{continue}
    \SetKw{Return}{return}
    \Input{initial state $x_{\textrm{orig}}$, step size $\alpha$, movement budget $r$, number of iterations $T$, norm $p$}
    \Output{final state $x'$}
    $x'_{(0)} \gets x_{\textrm{orig}}$\;
    \For{$t = 0$ \KwTo $T-1$}{
        \tcc{compute gradient} 
        $\mathbf{g}_{(t)} \gets \nabla_{x'_{(t)}} \mathcal{L}(h(x'_{(t)}), h(x_{\textrm{orig}}))$
    
        \tcc{update features by taking a step in direction of sign of gradient}
        \For{$i = 1$ \KwTo $d$}{
            $\rho_{(t)}[i] \gets \alpha \cdot \text{sign}(\mathbf{g}_{(t)}[i])$
        }
    
        $x'_{(t+1)} \gets x'_{(t)} + \rho_{(t)}$
    
        \tcc{project onto $r$-ball around $x_{\textrm{orig}}$}
        \eIf{$p = \infty$}{
           $
             x'_{(t+1)} \gets \Big(x_{\textrm{orig}} + \textrm{clip}_{[-r, r]}(x'_{(t+1)} - x_{\textrm{orig}})\Big)
          $
        }{
           $
            x'_{(t+1)} \gets \Bigg(x_{\textrm{orig}} + r \cdot \frac{x'_{(t+1)} - x_{\textrm{orig}}}{\max(\|x'_{(t+1)} - x_{\textrm{orig}}\|_2, r)}\Bigg)
           $
        }
    }
    \Return $x' = x'_{(T)}$\;
\caption{Agent improvement and gaming}
\label{alg:ocagi_improv_gaming}
\end{algorithm}

\section{Experimental Results}
\label{sec:ocagiexp_results}

Although the primary objective of this work is to investigate practical strategic classification methods that maximize a decision-maker’s (weighted) utility in settings where agents can both game and improve, we also present results from exploratory empirical studies examining: (1) how the choice and degree of strategic behavior incorporated into algorithm design affect model error reduction and utility gain; (2) whether zero model error necessarily implies a perfect \(100\%\) utility score; (3) how optimizing weighted utility differs from optimizing unweighted utility; (4) how restricting agents to improvement-only behavior, rather than allowing both gaming and improvement, affects error reduction and utility gain; and (5) how the choice of movement constraint, \(\ell_{\infty}\) versus \(\ell_{2}\), influences error reduction and utility gain.

\subsection{Effect of Level of Risk-aversion}
\label{subsec:ocagiexp_results_riskaversion} 

\begin{figure*}[t!]
    \centering

    \begin{subfigure}[t]{0.47\linewidth}
        \centering
        \includegraphics[width=\linewidth]{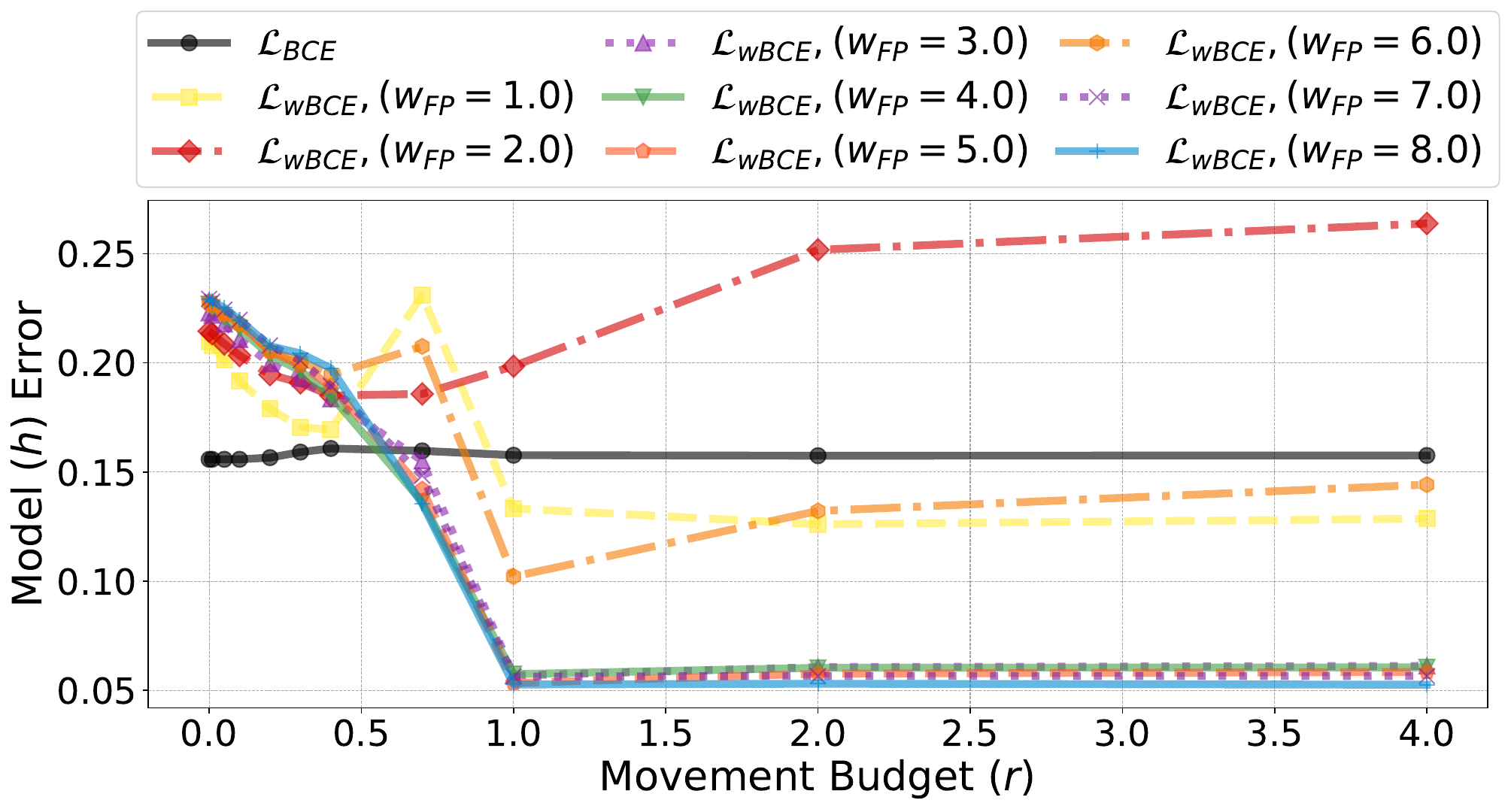}
        \caption{The \textbf{error drop} rate variations when agents respond to a BCE-trained model and wBCE-trained models with \textbf{lower false penalty} where \(\big(\mathbf{w_\textrm{FN}=0.1}, w_\textrm{FP}=\{i\}_{i=1}^{8}\big)\).}
        \label{fig:ocagi_adult_error_0.5linfno_0.1_wfpnvar}
    \end{subfigure}
    \hfill
    \begin{subfigure}[t]{0.47\linewidth}
        \centering
        \includegraphics[width=\linewidth]{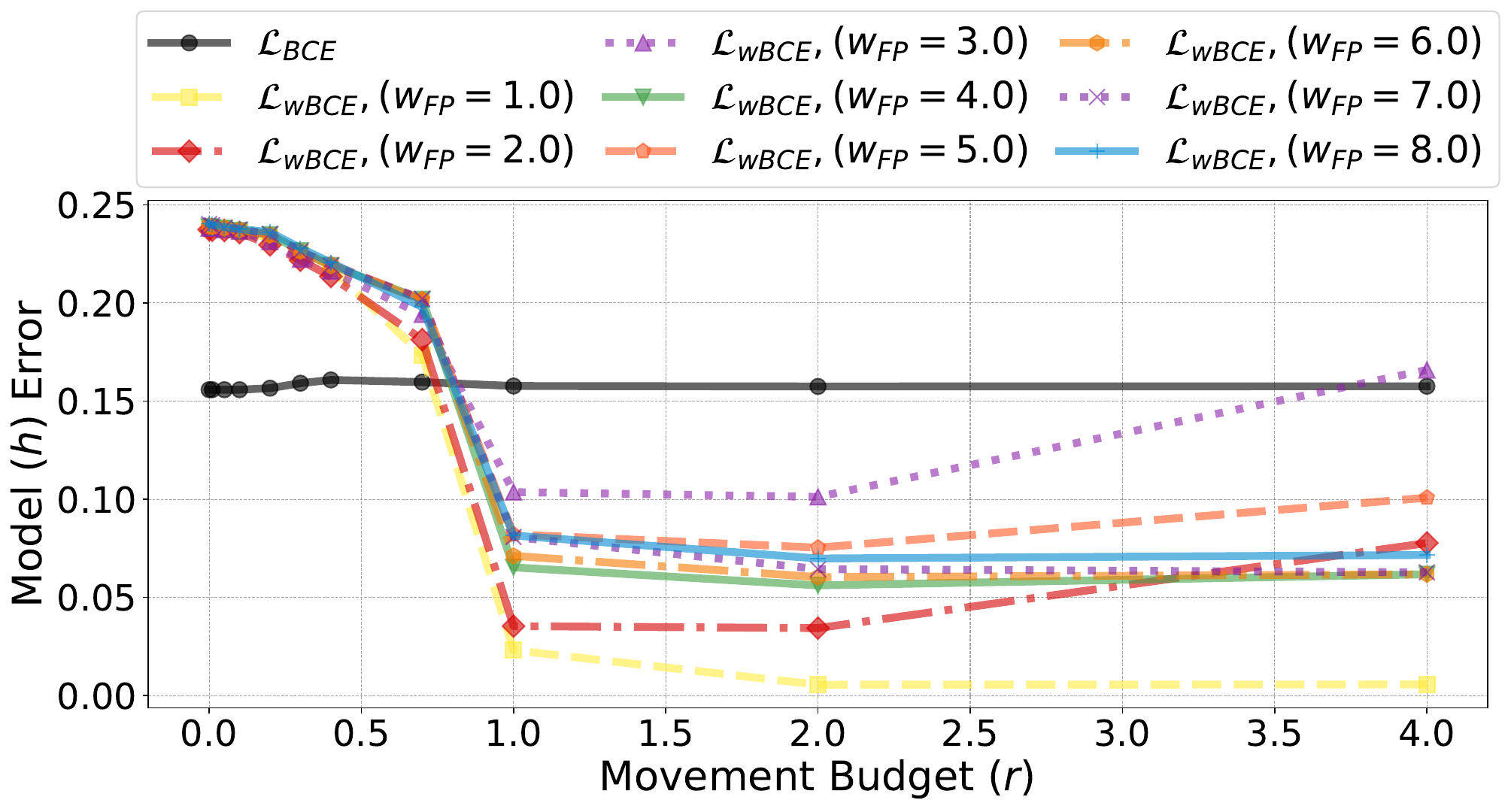}
        \caption{The \textbf{error drop} rate variations when agents respond to a BCE-trained model and wBCE-trained models with \textbf{higher false penalty} where \(\big(\mathbf{w_\textrm{FN}=0.0001}, w_\textrm{FP}=\{i\}_{i=1}^{8}\big)\).}
        \label{fig:ocagi_adult_error_0.5linfno_0.0001_wfpnvar}
    \end{subfigure}

    \vspace{1.2em}

    \begin{subfigure}[t]{0.47\linewidth}
        \centering
        \includegraphics[width=\linewidth]{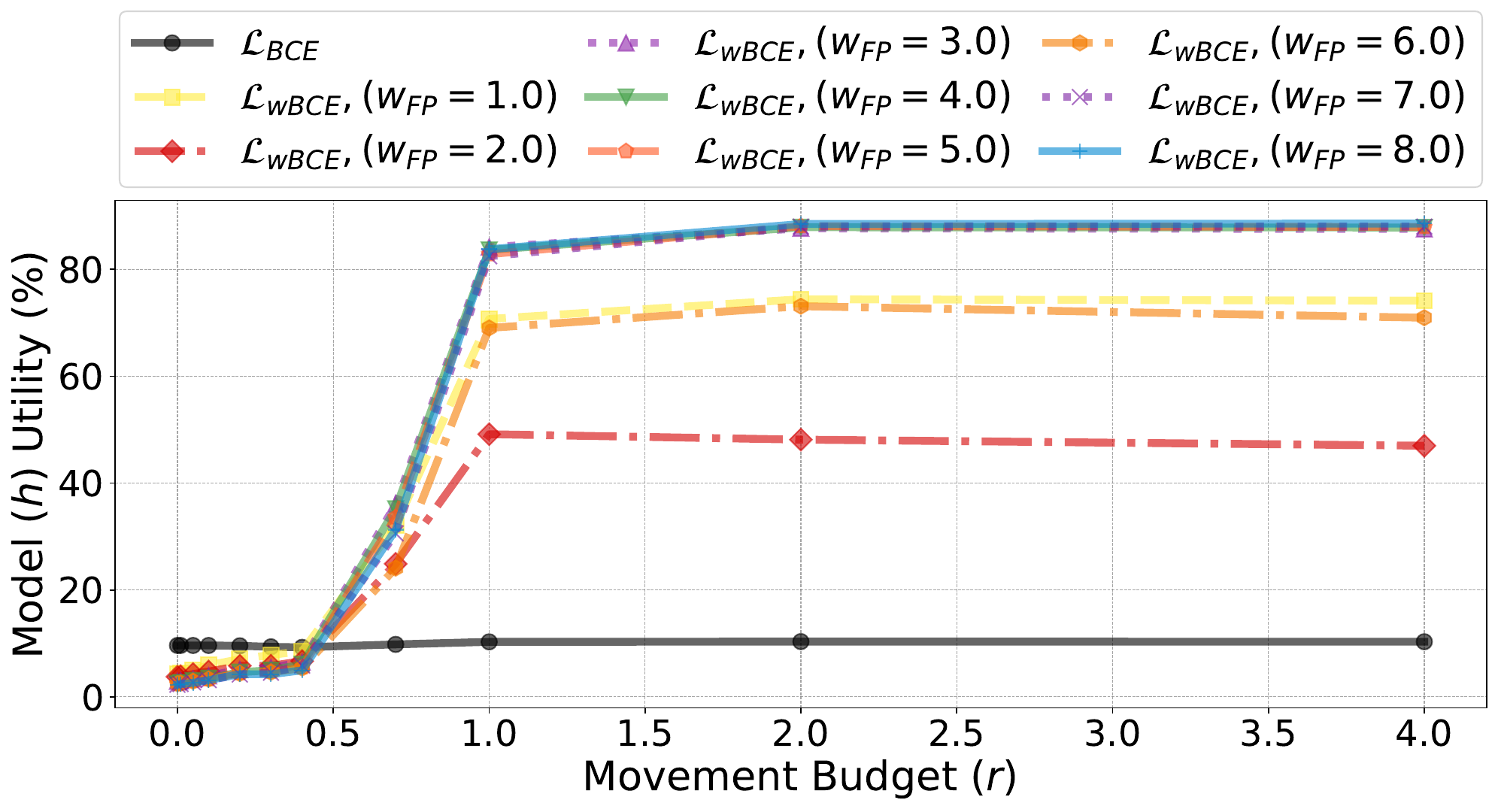}
        \caption{The \textbf{utility score} variations when agents respond to a BCE-trained model and wBCE-trained models with \textbf{lower false penalty} where \(\big(\mathbf{w_\textrm{FN}=0.1}, w_\textrm{FP}=\{i\}_{i=1}^{8}\big)\).}
        \label{fig:ocagi_adult_tpfp_0.5linfno_0.1_wfpnvar}
    \end{subfigure}
    \hfill
    \begin{subfigure}[t]{0.47\linewidth}
        \centering
        \includegraphics[width=\linewidth]{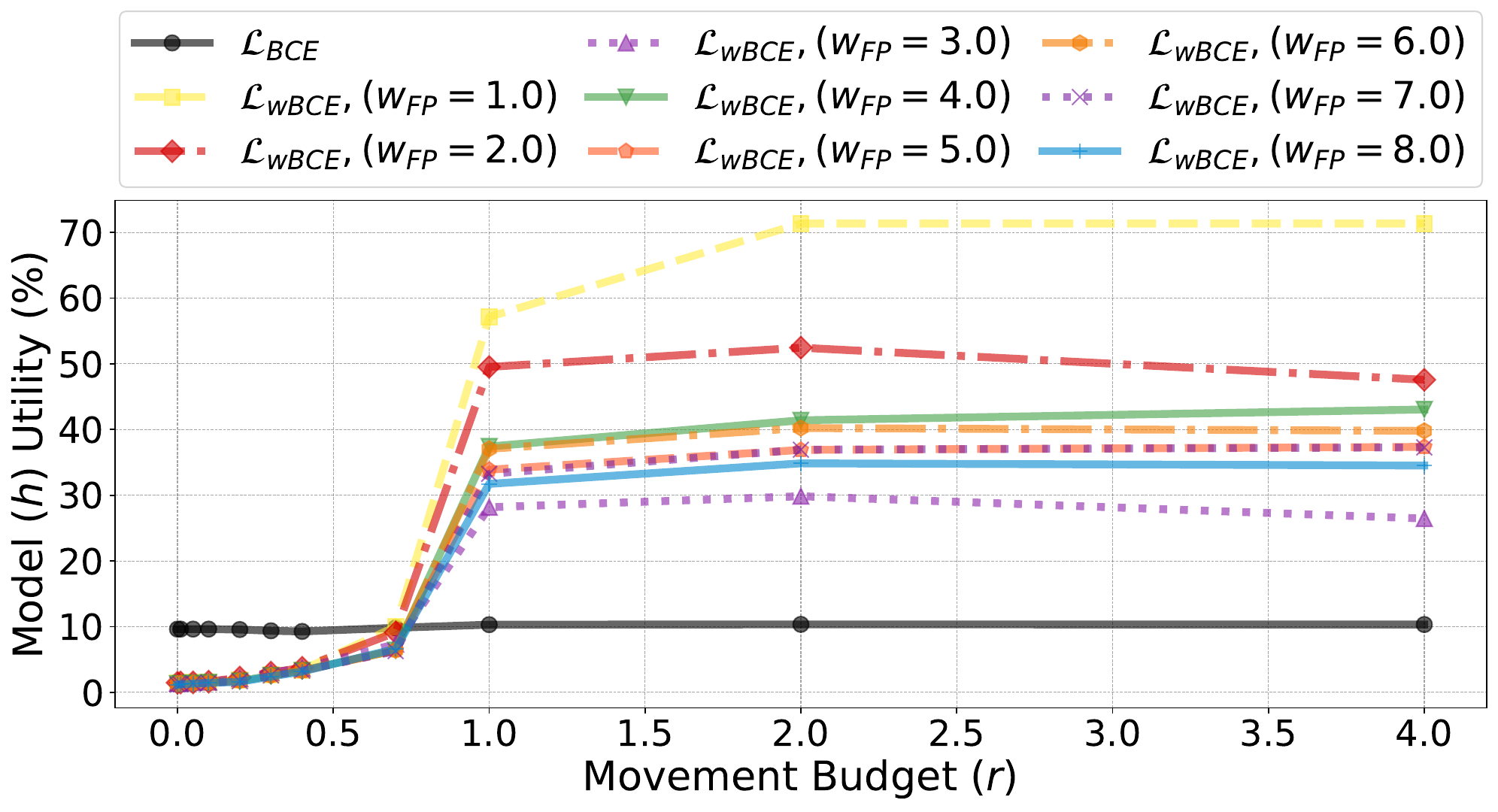}
        \caption{The \textbf{utility score} variations when agents respond to a BCE-trained model and wBCE-trained models with \textbf{higher false penalty} where \(\big(\mathbf{w_\textrm{FN}=0.0001}, w_\textrm{FP}=\{i\}_{i=1}^{8}\big)\).}
        \label{fig:ocagi_adult_tpfp_0.5linfno_0.0001_wfpnvar}
    \end{subfigure}

    \caption[On the Adult dataset, a comparative analysis to study the effect of level of risk-aversion]{On the \textbf{Adult} dataset, the error drop rates and \(\textrm{utility}\)  score increments when agents respond to BCE- and wBCE-trained models with varying risk-aversion levels. Figures~\subref{fig:ocagi_adult_error_0.5linfno_0.1_wfpnvar} and \subref{fig:ocagi_adult_error_0.5linfno_0.0001_wfpnvar} show the model error reduction as movement budget increases under different false positive weight penalties, while Figures~\subref{fig:ocagi_adult_tpfp_0.5linfno_0.1_wfpnvar} and \subref{fig:ocagi_adult_tpfp_0.5linfno_0.0001_wfpnvar} show the corresponding \(\textrm{utility}\) score increments. Figures~\subref{fig:ocagi_adult_error_0.5linfno_0.1_wfpnvar} and \subref{fig:ocagi_adult_tpfp_0.5linfno_0.1_wfpnvar} correspond to a lower risk-aversion level than Figures~\subref{fig:ocagi_adult_error_0.5linfno_0.0001_wfpnvar} and \subref{fig:ocagi_adult_tpfp_0.5linfno_0.0001_wfpnvar}.
    In all cases, agents move within an \(\ell_{\infty}\) ball, and they are classified as positive if the probability is higher than \(0.5\). High risk-aversion negatively affects the \(\textrm{utility}\)  score but leads to a faster error drop rate.}
    \label{fig:adult_errors_tpfp_th0.5linf_0.1and0.0001}
\end{figure*}

When agents adjust their features through gaming or genuine improvement in response to \(\mathcal{L}_{BCE}\)-trained models with  (\(w_{FN} = 0.1, w_{FP} = \{i\}_{i=1}^{8} \)) and (\(w_{FN} = 0.0001, w_{FP} = \{i\}_{i=1}^{8}\)), the resulting error drop rates and the change in \(\textrm{utility}\) scores vary based on the false positive mistakes penalty, that is, the \(\frac{w_{FN}}{w_{FP}}\) ratio.
Generally, the models with lower false positive mistakes penalty (see Figures~\ref{fig:ocagi_adult_error_0.5linfno_0.1_wfpnvar} and \ref{fig:ocagi_adult_tpfp_0.5linfno_0.1_wfpnvar}) had higher errors and higher \(\textrm{utility}\) scores after agents move, than the models with higher false positive mistakes penalty (see Figures~\ref{fig:ocagi_adult_error_0.5linfno_0.0001_wfpnvar} and \ref{fig:ocagi_adult_tpfp_0.5linfno_0.0001_wfpnvar}).

However, in both cases, a consistent trend emerges: models with lower errors, such as the (\(w_{FN} = 0.0001, w_{FP} =1 \)) trained model in Figure~\ref{fig:ocagi_adult_error_0.5linfno_0.0001_wfpnvar}  with error close to zero, achieved the highest \(\textrm{utility}\) scores (see Figure~\ref{fig:ocagi_adult_tpfp_0.5linfno_0.0001_wfpnvar}) than those with higher errors (see Figure~\ref{fig:ocagi_adult_error_0.5linfno_0.0001_wfpnvar}). 

The observations, together, suggest that while increasing risk aversion improves error reduction, excessive risk aversion may reduce the \(\textrm{utility}\) scores. The reason is that fewer negatively classified agents attempt feature modifications to be classified as positive when the movement (gaming/improving) cost is prohibitively high under a risk-averse classifier.

\subsection{Effect of Choice of Risk-aversion}
\label{subsec:ocagiexp_results_choice_riskaversion} 

We comparatively analyze how the choice of risk-aversion: loss-based or threshold-based, affects the error drop rate and the utility score increment when agents modify their features through gaming or genuine improvement in response to \(\mathcal{L}_{BCE}\)-trained models with  (\(w_{FN} = 0.1, w_{FP} = \{i\}_{i=1}^{8} \)) and (\(w_{FN} = 0.0001, w_{FP} = \{i\}_{i=1}^{8}\)).

\begin{figure*}[t!]
    \centering
    \begin{subfigure}[t]{0.47\linewidth}
            \centering
            \includegraphics[width=\linewidth]{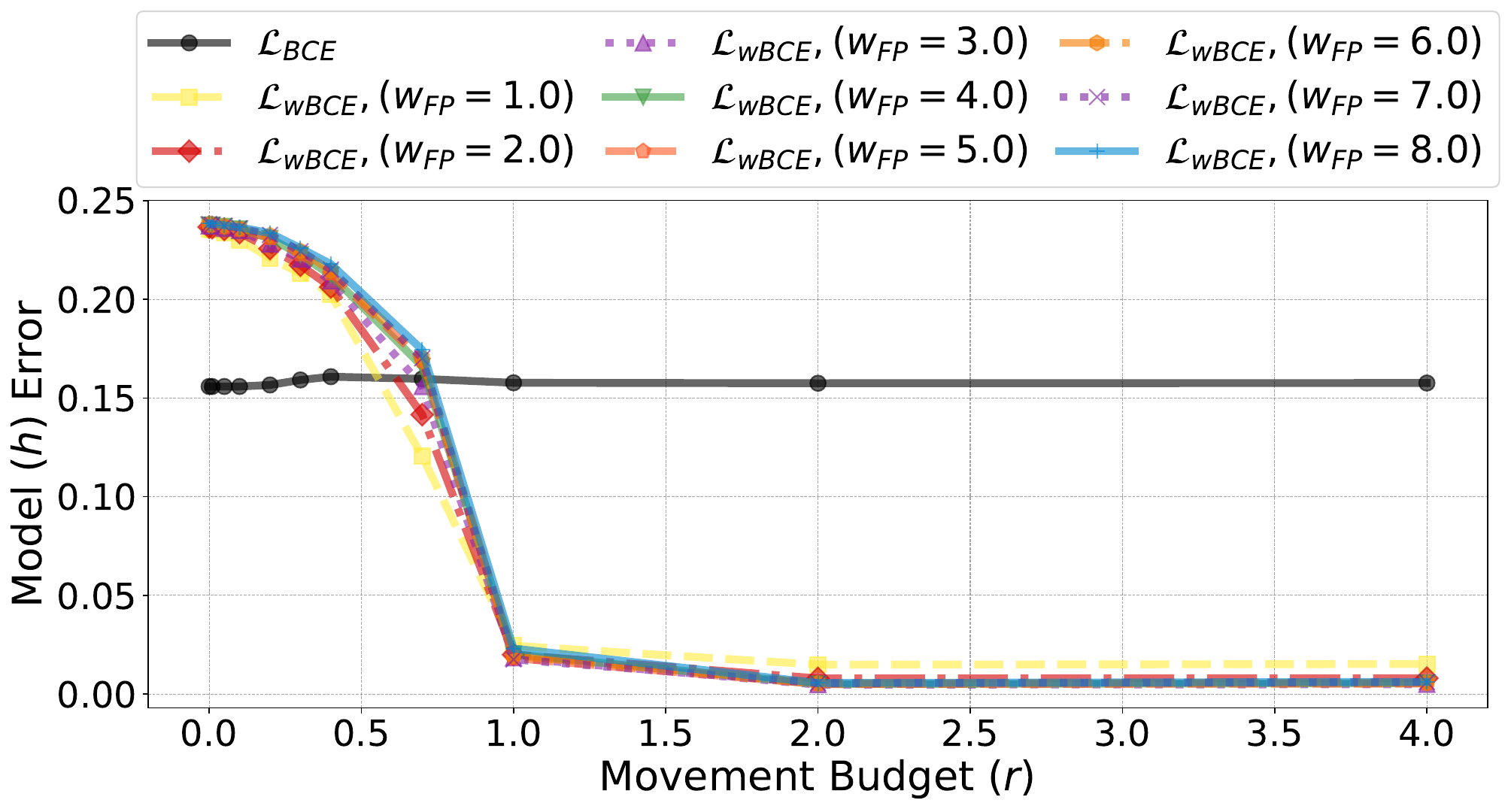}
    \caption{The \textbf{\(\textrm{error drop}\)} rates when agents can both game and improve in response to BCE- and wBCE-trained models and a classification threshold of \(\mathbf{0.5}\)}
    \label{fig:ocagi_adult_error_0.5linfno1}
    \end{subfigure}
    \hfill
    \begin{subfigure}[t]{0.47\linewidth}
        \centering
        \includegraphics[width=\linewidth]{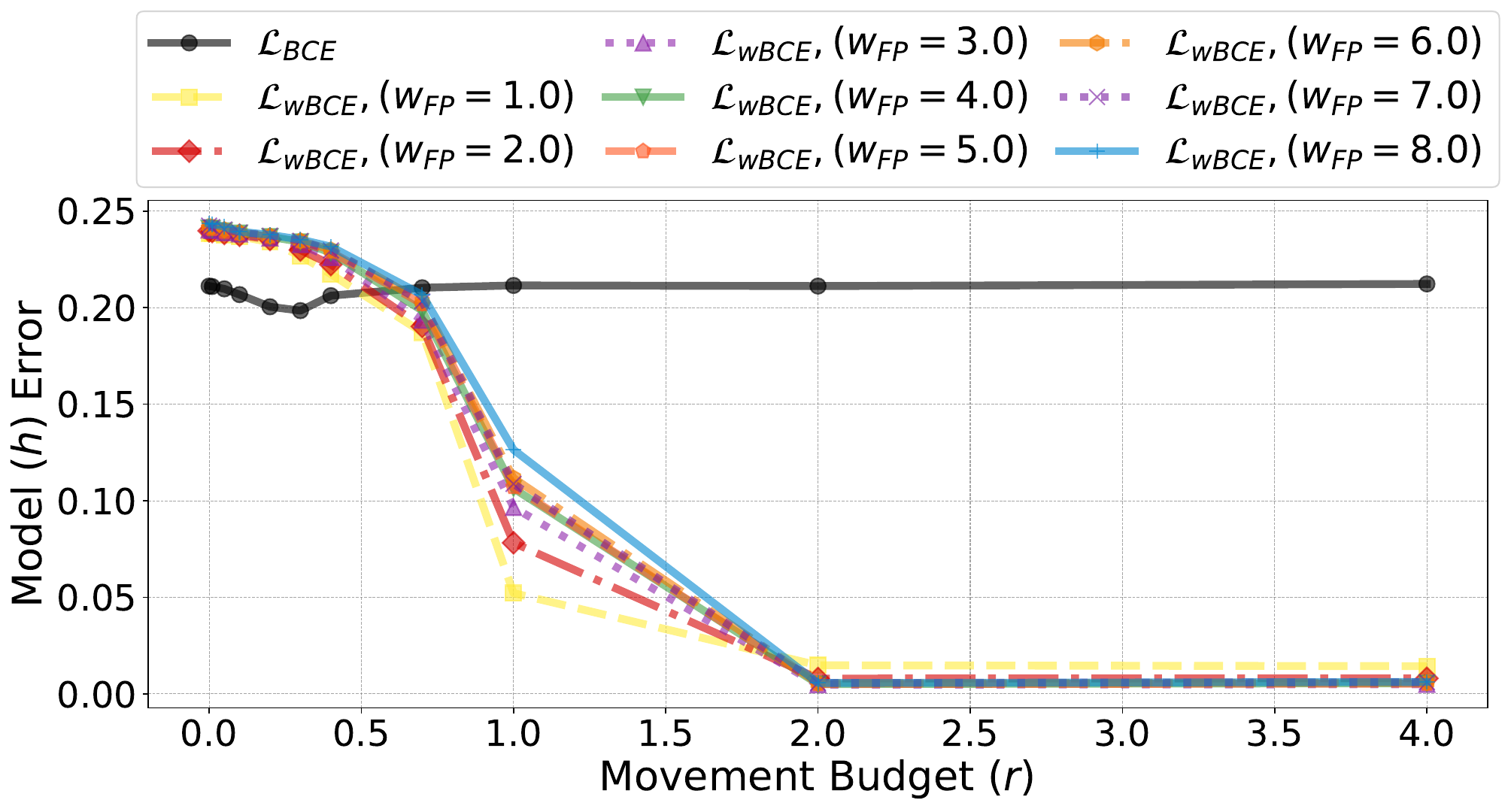}
    \caption{The \textbf{\(\textrm{error drop}\)} rates when agents can both game and improve in response to BCE- and wBCE-trained models and a classification threshold of \(\mathbf{0.9}\)}
    \label{fig:ocagi_adult_error_0.9linfno1}
    \end{subfigure}

    \vspace{1.2em}

    \begin{subfigure}[t]{0.47\linewidth}
        \centering
        \includegraphics[width=\linewidth]{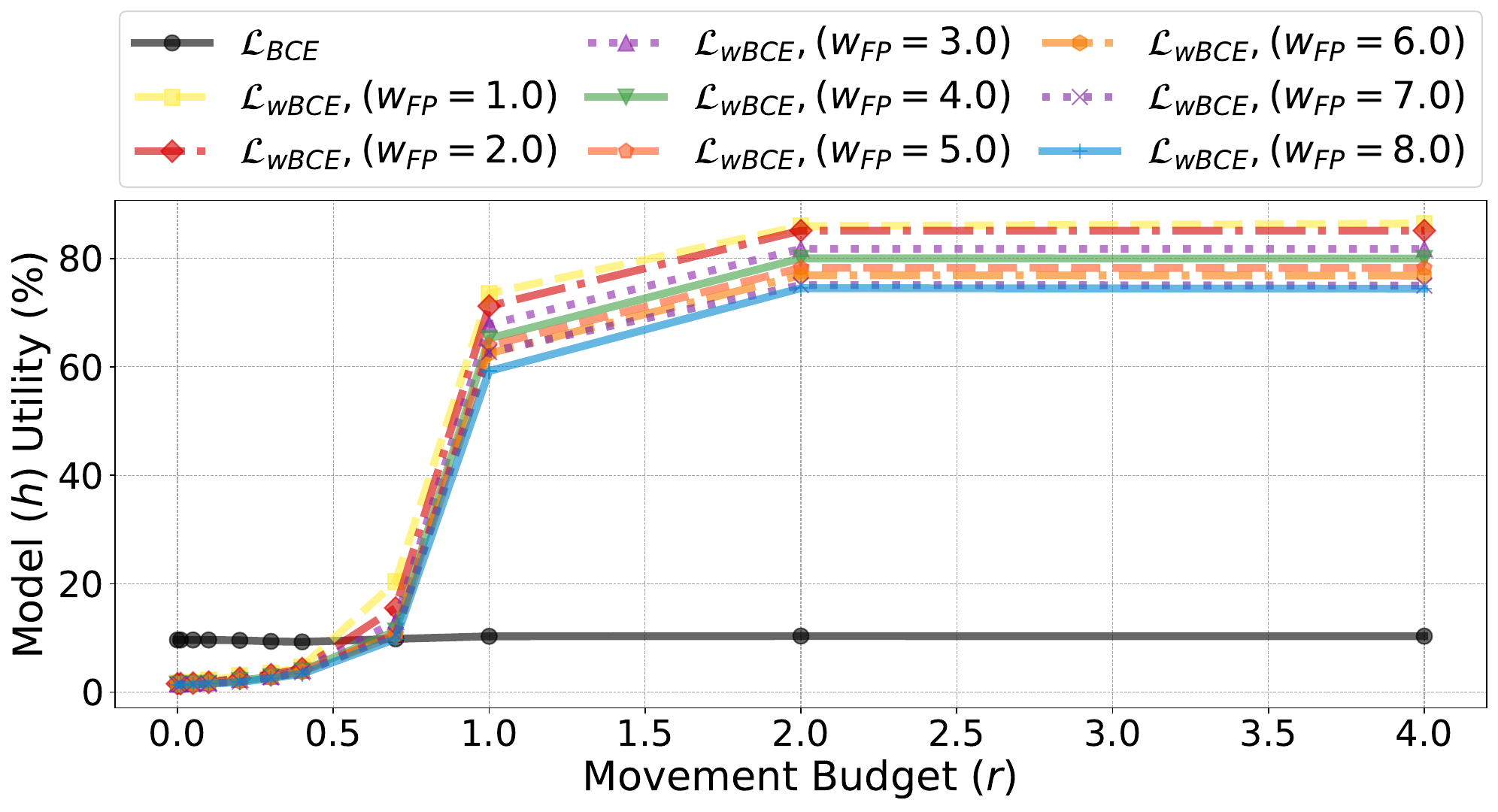}
    \caption{The \textbf{\(\textrm{utility}\) scores} when agents can both game and improve in response to BCE- and wBCE-trained models and a classification threshold of \(\mathbf{0.5}\)}
    \label{fig:ocagi_adult_tpfp_0.5linfno_0.001_wfpnvar1}
    \end{subfigure}
    \hfill
    \begin{subfigure}[t]{0.47\linewidth}
        \centering
        \includegraphics[width=\linewidth]{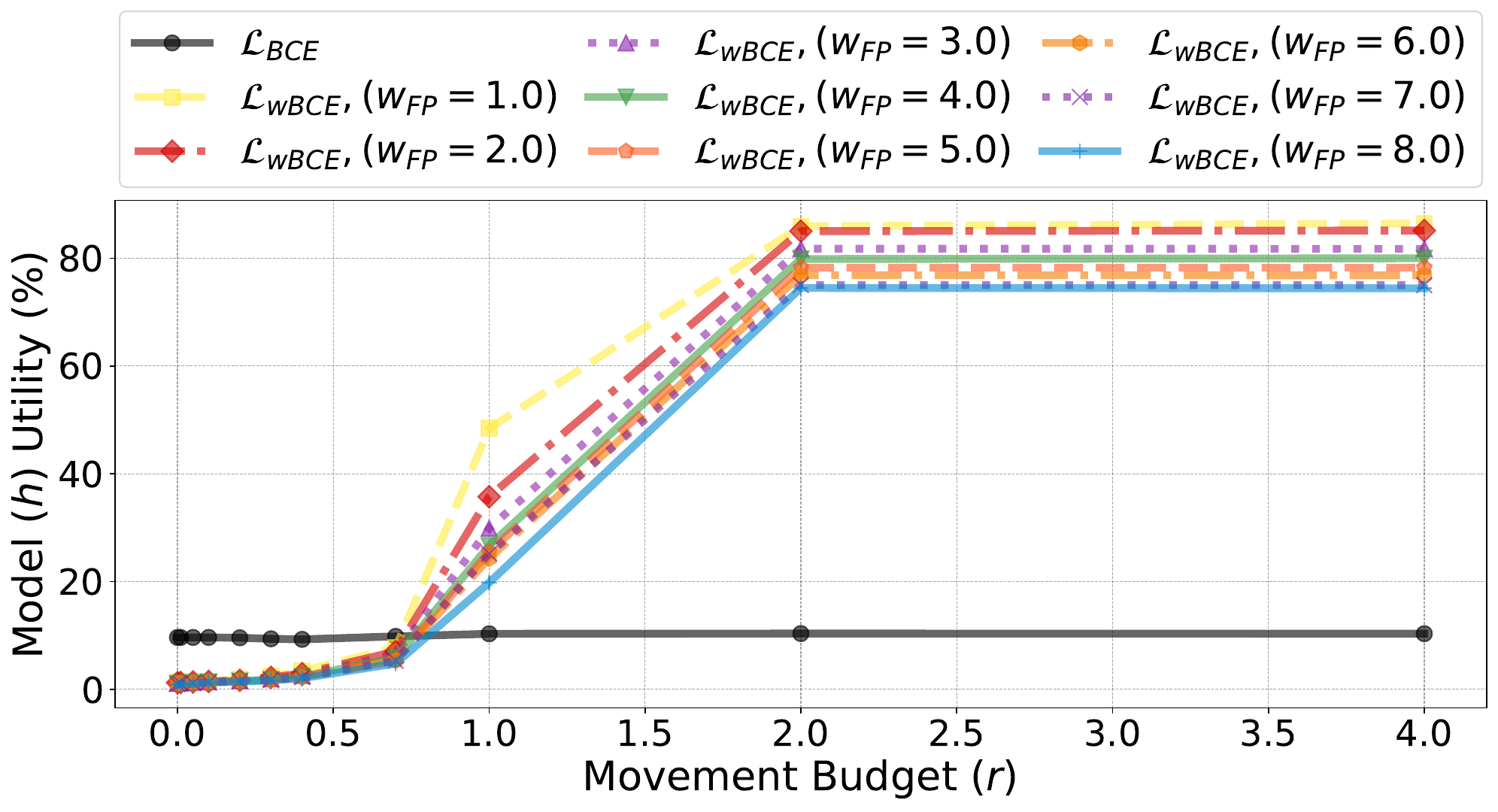}
    \caption{The \textbf{\(\textrm{utility}\) scores}  when agents can both game and improve in response to BCE- and wBCE-trained models and a classification threshold of \(\mathbf{0.9}\)}
    \label{fig:ocagi_adult_tpfp_0.9linfno_0.001_wfpnvar1}
    \end{subfigure}
    
    \caption[On the Adult dataset, a comparative analysis to study the effect of the choice of a risk-aversion strategy]{On the \textbf{Adult} dataset, we conduct a comparative analysis of the effect of choice of risk aversion on the error reduction rate (\subref{fig:ocagi_adult_error_0.5linfno1} and \subref{fig:ocagi_adult_error_0.9linfno1}) and utility score increment (\subref{fig:ocagi_adult_tpfp_0.5linfno_0.001_wfpnvar1} and \subref{fig:ocagi_adult_tpfp_0.9linfno_0.001_wfpnvar1}) when agents modify their features under different movement budgets \(r\) in response to a BCE-trained model and wBCE-trained models with weight configurations \(\big(\mathbf{w_\textrm{FN}=0.001}, w_\textrm{FP}=\{i\}_{i=1}^{8}\big)\).  In all cases, agents move within an \(\ell_{\infty}\) ball.}
    \label{fig:adult_errors_tpfp_th0.5_0.9linf_no}
\end{figure*}

In general, the loss-based risk-aversion is more effective than the threshold-based risk-aversion (See Figure~\ref{fig:adult_errors_tpfp_th0.5_0.9linf_no}). 
Specifically, we observe that the higher the false positive penalty (higher \(\frac{w_{FN}}{w_{FP}}\) ratio), the faster the error reduction rate and utility score increment as the budget increases. On the other hand, the higher the threshold (\(0.9\) instead of \(0.5\)), the slower the error reduction rate and the utility score increment as the budget increases.

\subsection{Does Zero Error Imply Perfect Utility?}
\label{subsec:ocagiexp_results_errorvutility}

\begin{figure*}[ht!]
    \centering
    \begin{subfigure}[t]{0.47\linewidth}
        \centering
        \includegraphics[width=\linewidth]{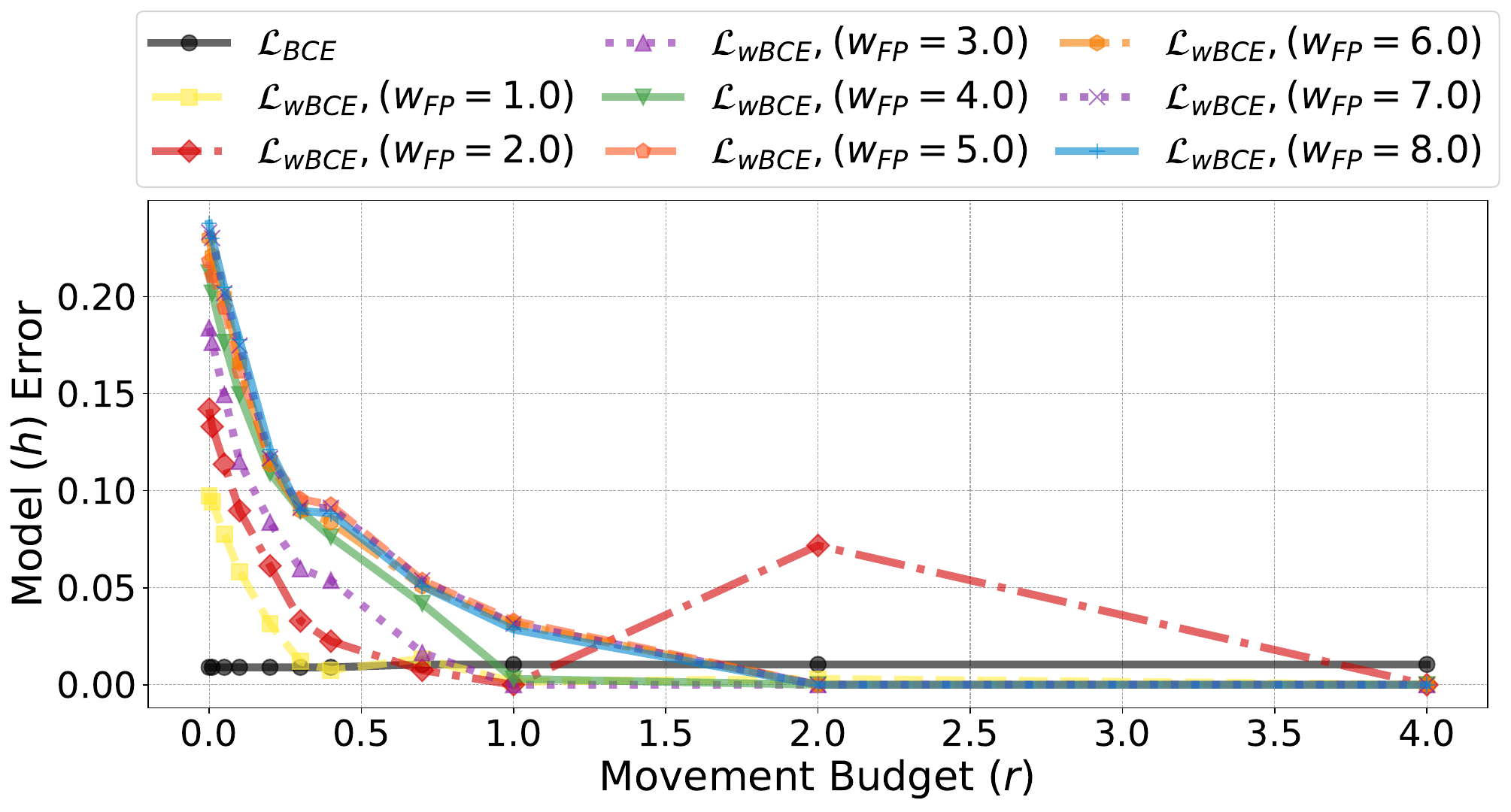}
    \caption{The \textbf{error drop} rate variations when agents respond to a BCE-trained model and wBCE-trained models with weight configurations \(\big({w_\textrm{FN}=0.009}, w_\textrm{FP}=\{i\}_{i=1}^{8}\big)\)}
    \label{fig:ocagi_synthetic_error_0.5linfno}
    \end{subfigure}
    \hfill
    \begin{subfigure}[t]{0.47\linewidth}
        \centering
        \includegraphics[width=\linewidth]{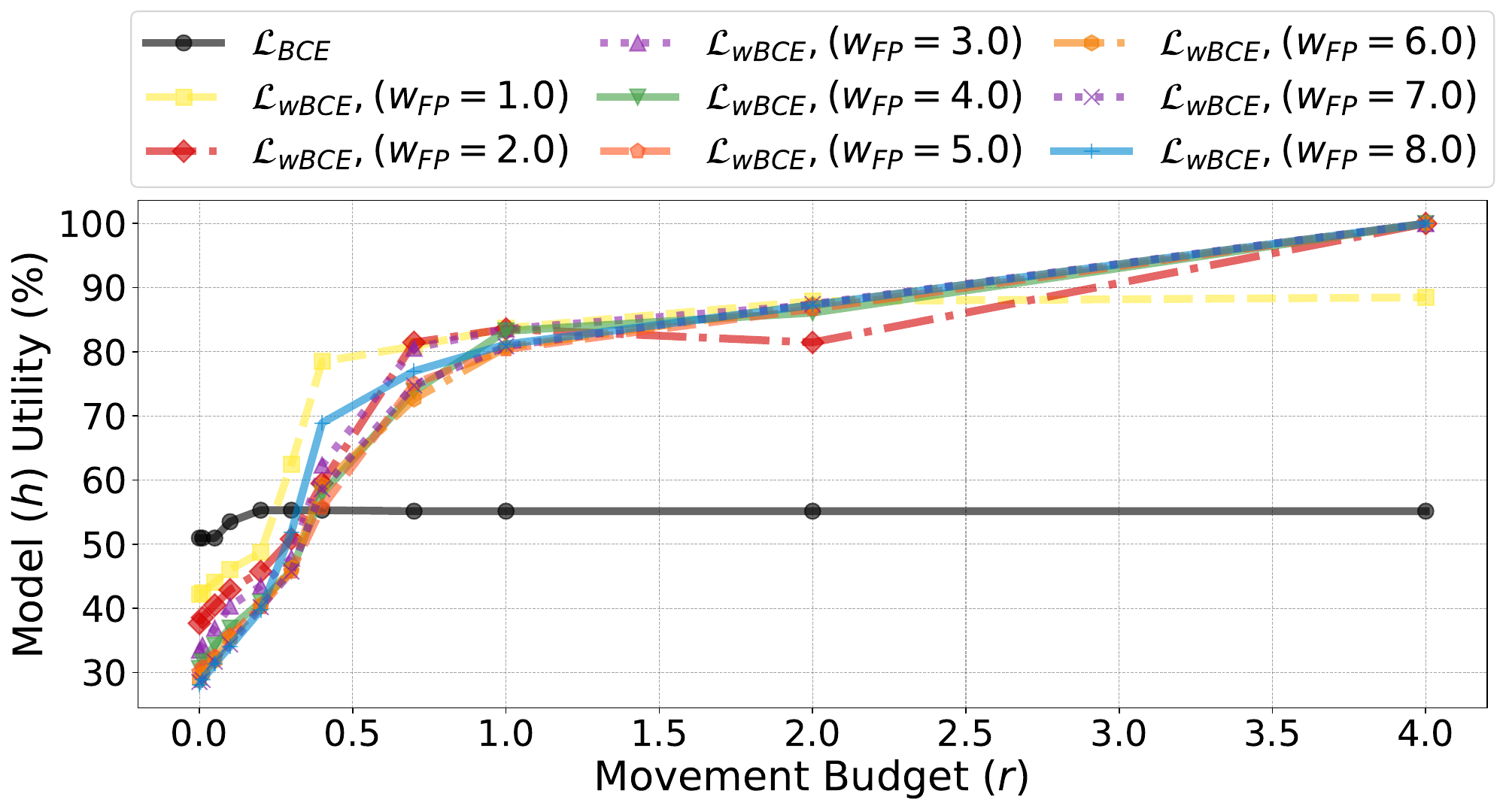}
    \caption{The \textbf{utility score} variations when agents respond to a BCE-trained model and wBCE-trained models with weight configurations \(\big({w_\textrm{FN}=0.009}, w_\textrm{FP}=\{i\}_{i=1}^{8}\big)\)}
    \label{fig:ocagi_synthetic_tpfp_0.5linfno}
    \end{subfigure}

    \vspace{1.2em}

    \begin{subfigure}[t]{0.47\linewidth}
        \centering
        \includegraphics[width=\linewidth]{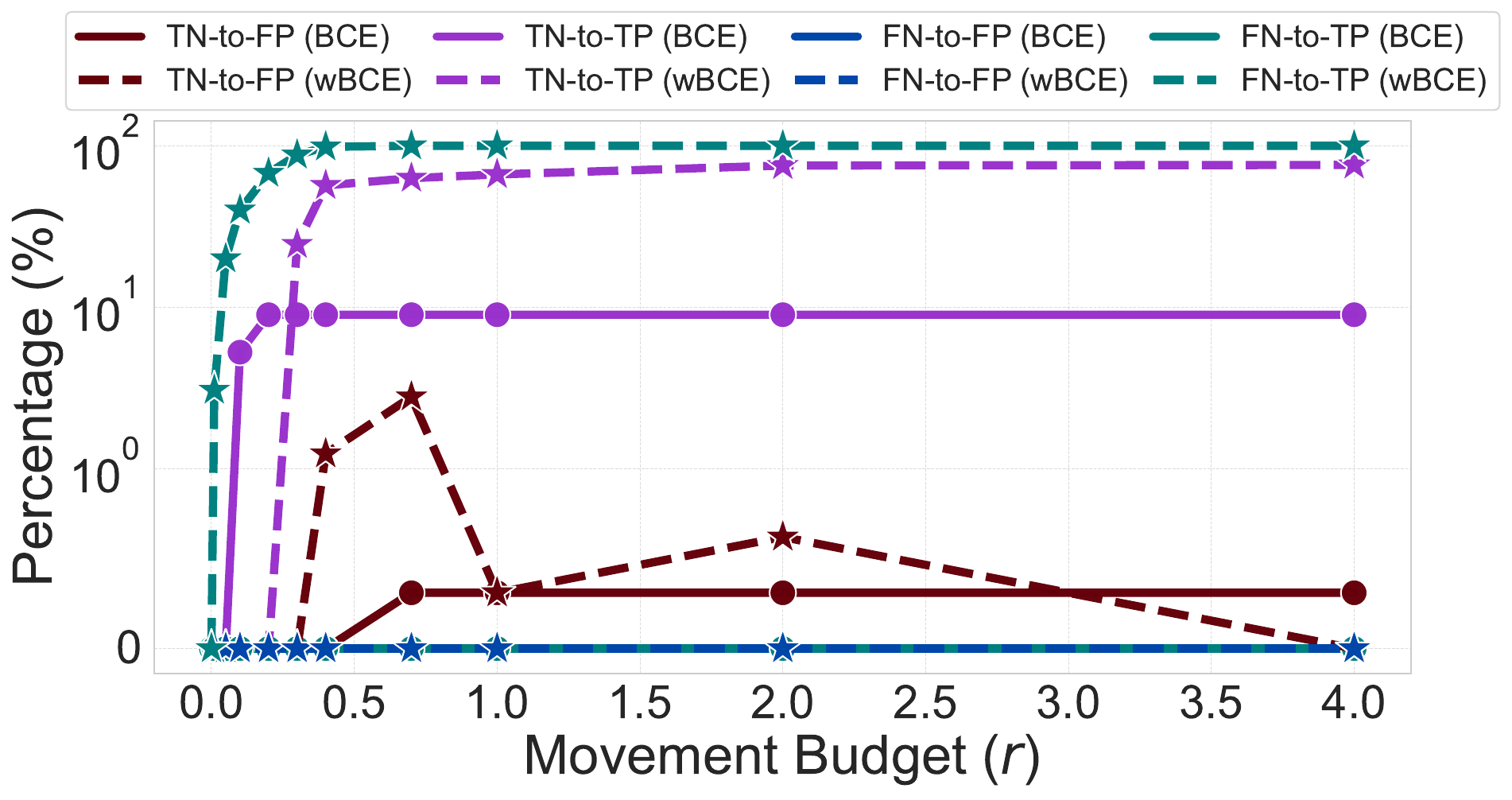}
    \caption{Percentage of agents that transition \textbf{from TN/FN to TP/FP} in response to a BCE-trained model and wBCE-trained models with weight configurations \(\big(w_\textrm{FN}=0.009, \mathbf{w_\textrm{FP}=1.0}\big)\)}
    \label{fig:ocagi_synthetic_move_0.5linf1.0no}
    \end{subfigure}
    \hfill
    \begin{subfigure}[t]{0.47\linewidth}
        \centering
        \includegraphics[width=\linewidth]{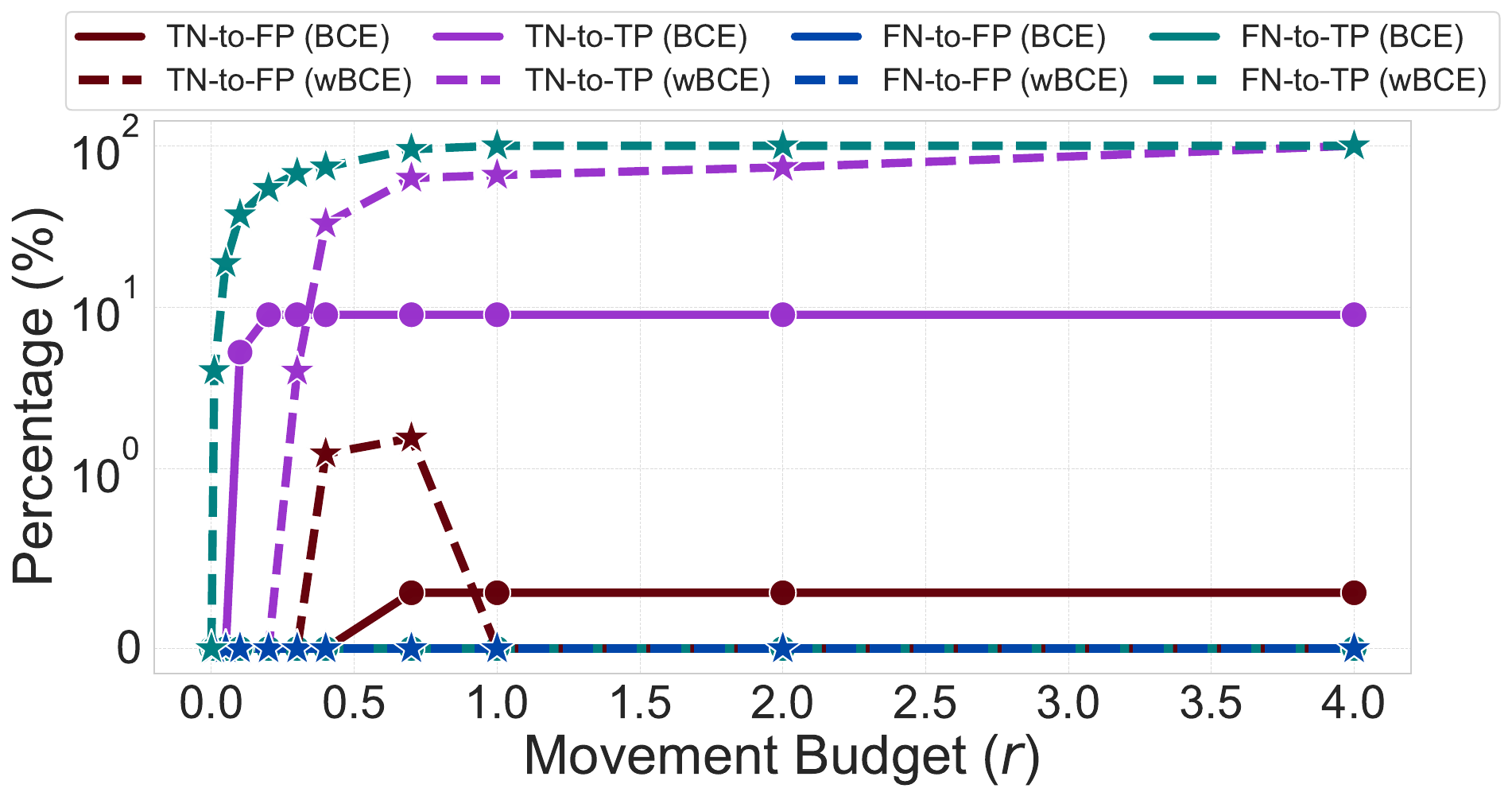}
    \caption{Percentage of agents that transition \textbf{from TN/FN to TP/FP} in response to a BCE-trained model and wBCE-trained models with weight configurations \(\big(w_\textrm{FN}=0.009,\mathbf{ w_\textrm{FP}=3.0}\big)\)}
    \label{fig:ocagi_synthetic_move_0.5linf3.0no}
    \end{subfigure}
        
    \caption[A study to investigate if perfect model accuracy implies perfect utility]{On the \textbf{Synthetic} dataset, we compare the performance of a BCE-trained model with wBCE-trained models under different weights \(\big(w_\textrm{FN}=0.009, w_\textrm{FP}=\{i\}_{i=1}^{8}\big)\), focusing on the error drop rate and the \(\textrm{utility}\) score increment rate as agents modify their features (either by gaming or improving) in response to the models. Figures~\subref{fig:ocagi_synthetic_error_0.5linfno} and \subref{fig:ocagi_synthetic_tpfp_0.5linfno} illustrate the model's error reduction and the corresponding \(\textrm{utility}\) score increment as a function of the agents' movement budget \(r\). Figures~\subref{fig:ocagi_synthetic_move_0.5linf1.0no} and \subref{fig:ocagi_synthetic_move_0.5linf3.0no} show the percentage of agents transitioning between states (e.g., from true negative to false positive) after modifying their features in response to a BCE-trained model and wBCE-trained models with weight settings of \(\big(w_\textrm{FN}=0.009, w_\textrm{FP}=1.0\big)\) and \(\big(w_\textrm{FN}=0.009, w_\textrm{FP}=3.0\big)\), respectively. Zero error doesn't imply \(100\%\) \(\textrm{utility}\) score. In all cases agents move within an \(\ell_{\infty}\) ball and they are classified as positive if the probability is higher than \(0.5\).}
    \label{fig:synthetic_errors_tpfp_th0.5linf}
\end{figure*}

When a model achieves zero error after agents move (either through gaming or improvement), it does not necessarily imply a \(100\%\) score on the \(\textrm{utility}\) metric. For instance, in Figure~\ref{fig:ocagi_synthetic_error_0.5linfno}, models trained with \(\mathcal{L}_{BCE}\) using (\(w_{FN} = 0.009, w_\textrm{FP}=1.0\)) and (\(w_{FN} = 0.009, w_\textrm{FP}=3.0\)) both exhibit error reduction to zero when agents move within a budget of \(r \geq 2.0\). 
However, with (\(w_{FN} = 0.009, w_\textrm{FP}=1.0\)), the model only approaches \(90\%\) on the \(\textrm{utility}\) metric. 
In contrast, with (\(w_{FN} = 0.009, w_\textrm{FP}=3.0\)), the model reaches \(100\%\) on this metric when agents move within a budget of \(r = 4.0\) (see Figure~\ref{fig:ocagi_synthetic_tpfp_0.5linfno}).

To better understand this behavior, we investigate agents' movement in detail. That is, with the (\(w_{FN} = 0.009, w_\textrm{FP}=1.0\)) trained model (Figures~\ref{fig:ocagi_synthetic_move_0.5linf1.0no}), at the \(r=4.0\) movement budget, only some true negatives move to become true positive (purple dotted line), while in the  model trained with
(\(w_{FN} = 0.009, w_\textrm{FP}=3.0\)), all negatively classified agents (both true and false negatives), move to become true positives (Figure~\ref{fig:ocagi_synthetic_move_0.5linf3.0no}). 
As shown in the Appendix, Figures~\ref{fig:adult_errors_tpfp_th0.5linf}, \ref{fig:law_errors_tpfp_th0.5linf}, and \ref{fig:oulad_errors_tpfp_th0.5linf}, we observe a similar pattern on the Adult, Law School, and OULAD datasets.

\subsection{Effect  of Geometric Constraints}
\label{subsec:ocagiexp_results_movnorm}

When agents operate within an \(\ell_{\infty}\) norm ball, each feature modification is bounded, whereas within an \(\ell_{2}\) norm ball, the cumulative magnitude of feature changes is constrained. These differing geometric constraints shape the agents' behavior by limiting either the per-feature adjustment (\(\ell_{\infty}\)) or the overall change vector magnitude (\(\ell_{2}\)).

\begin{figure*}[ht!]
    \centering

    \begin{subfigure}[t]{0.47\linewidth}
        \centering
        \includegraphics[width=\linewidth]{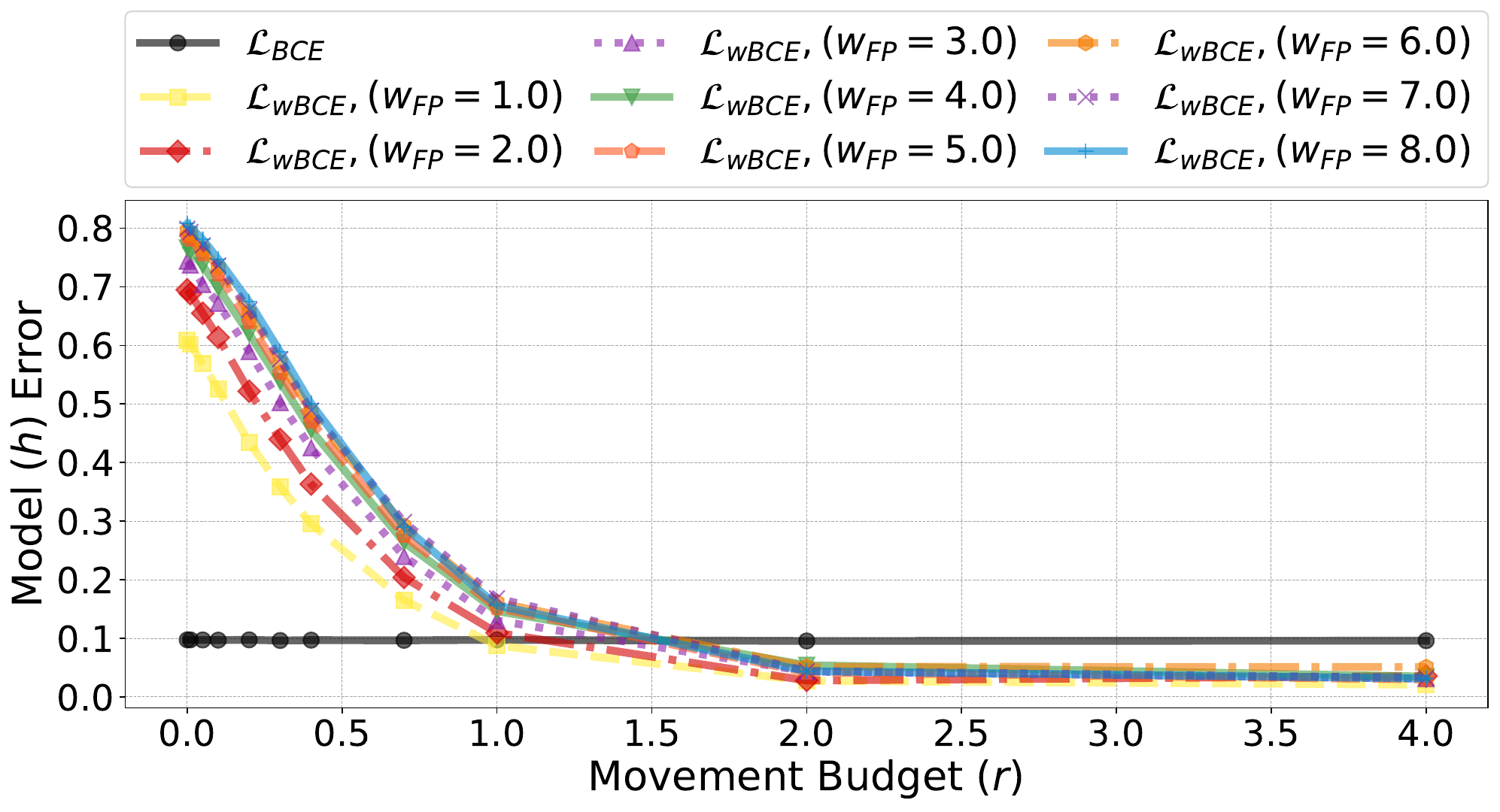}
    \caption{The \textbf{error drop} rate when agents  can both game and improve within an \(\boldsymbol{\ell}_\infty\)-ball}
    \label{fig:ocagi_law_error_0.5linfno0.009wfpvaried}
    \end{subfigure}
    \hfill
    \begin{subfigure}[t]{0.47\linewidth}
        \centering
        \includegraphics[width=\linewidth]{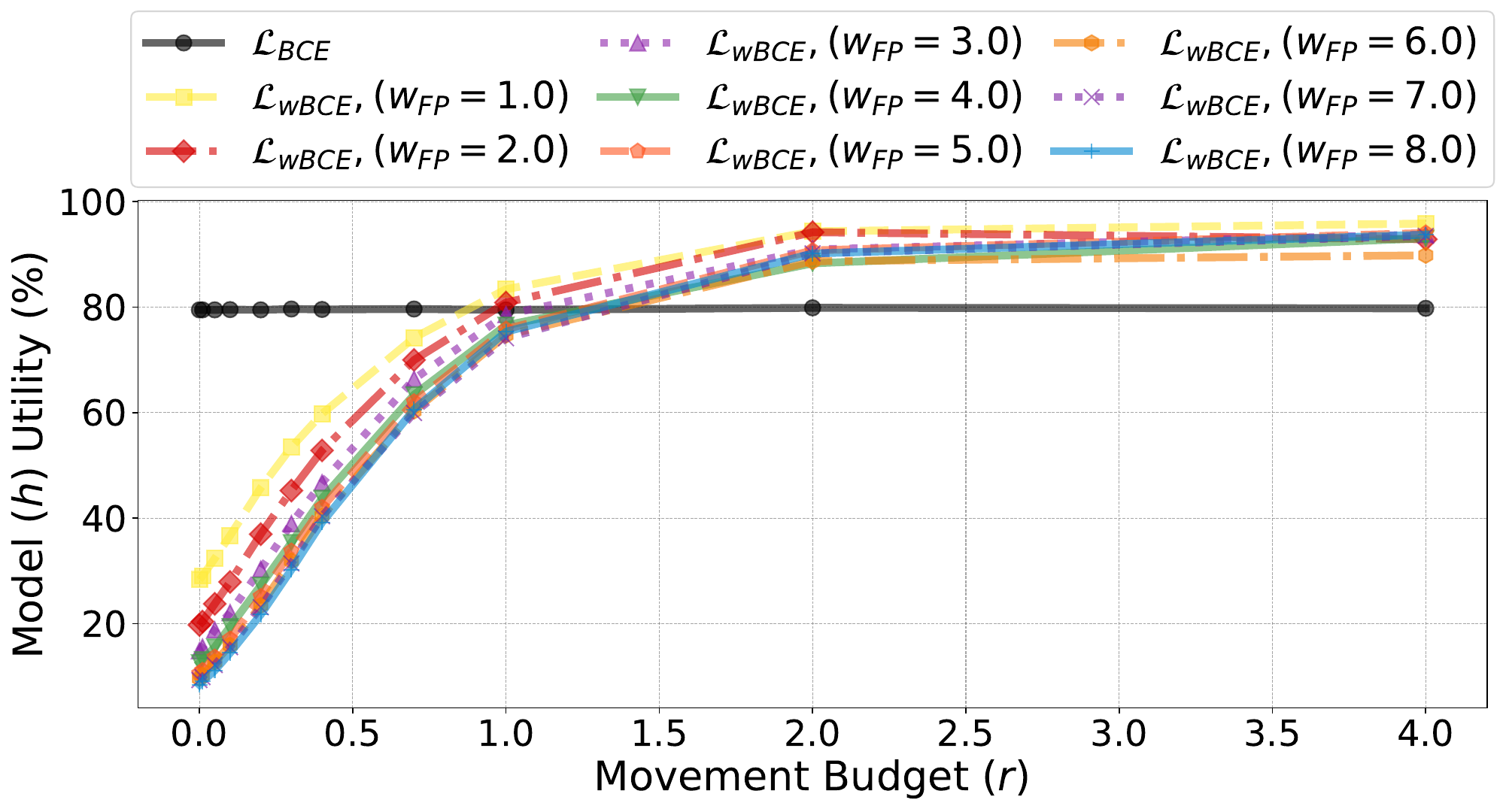}
   \caption{The \textbf{utility score} when agents can both game and improve within an \(\boldsymbol{\ell}_\infty\)-ball}
    \label{fig:ocagi_law_tpfp_0.5linfno0.009wfpvaried}
    \end{subfigure}

    \vspace{1.2em}

    \begin{subfigure}[t]{0.47\linewidth}
        \centering
        \includegraphics[width=\linewidth]{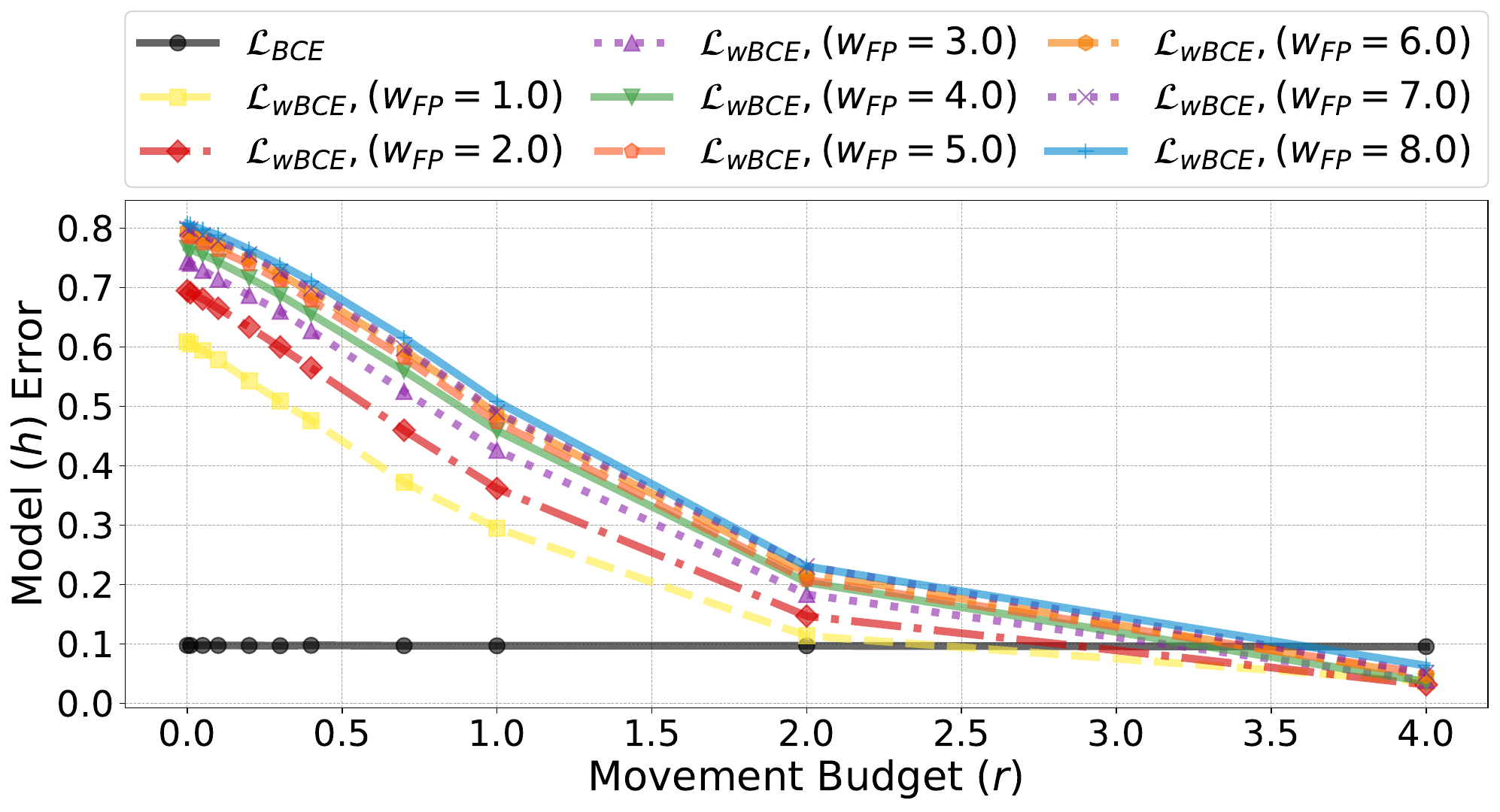}
    \caption{The \textbf{error drop} rate when agents can both game and improve within an \(\boldsymbol{\ell}_2\)-ball}
    \label{fig:ocagi_law_error_0.5l2no0.009wfpvaried}
    \end{subfigure}
    \hfill
    \begin{subfigure}[t]{0.47\linewidth}
        \centering
        \includegraphics[width=\linewidth]{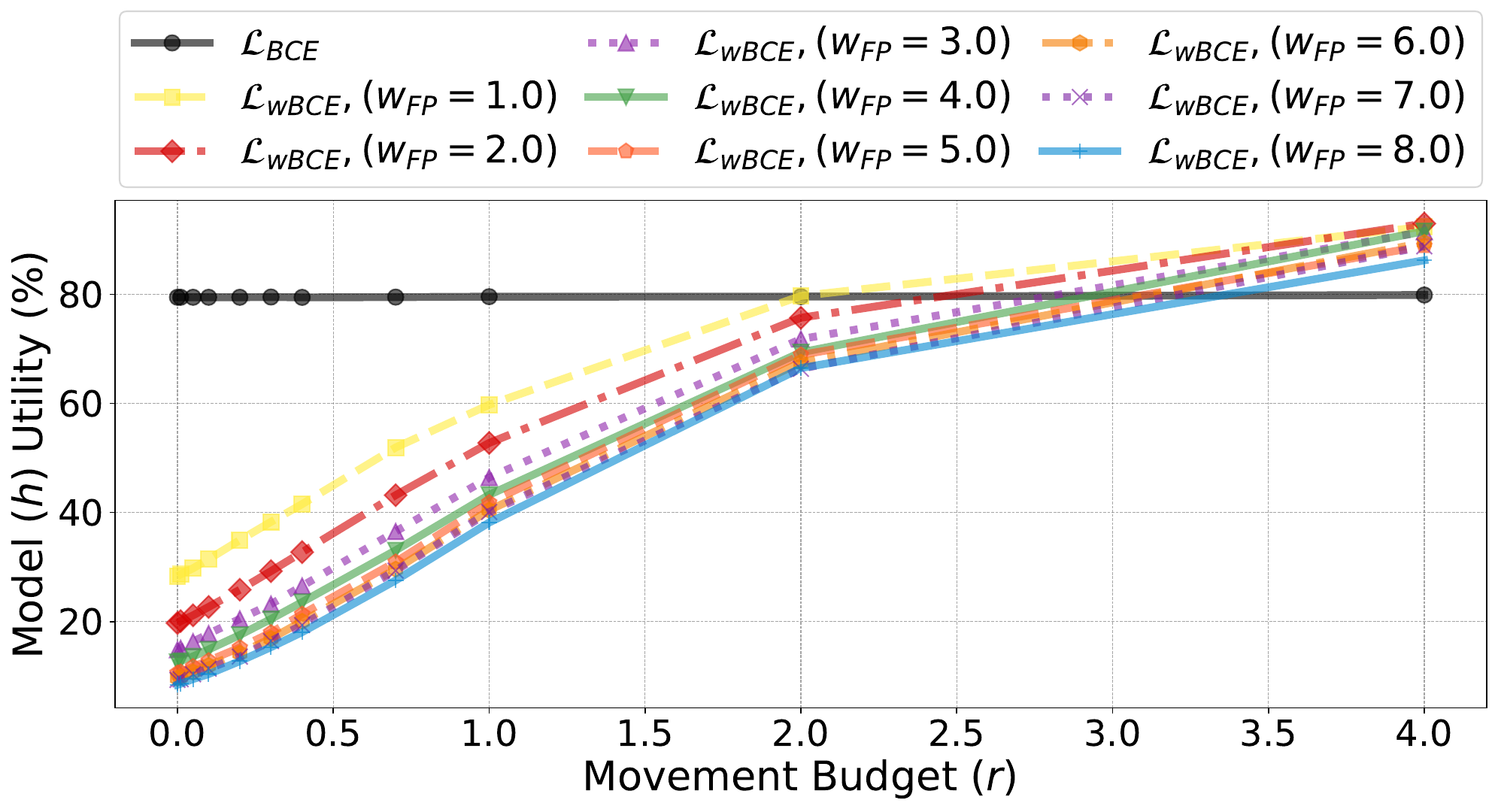}
    \caption{The \textbf{utility score} when agents can both game and improve within an \(\boldsymbol{\ell}_2\)-ball}
    \label{fig:ocagi_law_tpfp_0.5l20.009wfpvaried}
    \end{subfigure}
        
    \caption[On the Law School dataset, a comparative analysis to study the effect  of geometric constraints ]{On the \textbf{Law School} dataset, a comparative analysis of the variation  in error drop rate and \(\textrm{utility}\) score increment when agents can both game and improve under different movement budgets \(r\) and within \(\ell_{\infty}\) and \(\ell_{2}\) norm balls, in response to BCE- and wBCE-trained models. Figures~\subref{fig:ocagi_law_error_0.5linfno0.009wfpvaried} and \subref{fig:ocagi_law_tpfp_0.5linfno0.009wfpvaried} show results for the when agents move within \(\ell_{\infty}\) ball, while Figures~\subref{fig:ocagi_law_error_0.5l2no0.009wfpvaried} and \subref{fig:ocagi_law_tpfp_0.5l20.009wfpvaried} show results for the when agents move within \(\ell_{2}\) ball. In all cases, agents are classified as positive if the probability is higher than \(0.5\).}
    \label{law_error_0.5l2linfno0.009wfpvaried}
\end{figure*}

Empirically, we find that enforcing the \(\ell_{\infty}\) norm constraint when agents modify their features through gaming and improvement consistently results in faster increases in utility scores and reductions in model error, compared to enforcing the \(\ell_{2}\) constraint (see Figure~\ref{law_error_0.5l2linfno0.009wfpvaried}). 
Moreover,  confining agents' movement to an \(\ell_{\infty}\)-ball leads to higher utility scores than when restricted to an \(\ell_{2}\)-ball.

\subsection{The Weighted Utility Metric: Utility(1,8)}
\label{subsec:ocagiexp_results_profits}

\begin{figure*}[ht!]
    \centering

    \begin{subfigure}[t]{0.47\linewidth}
        \centering
        \includegraphics[width=\linewidth]{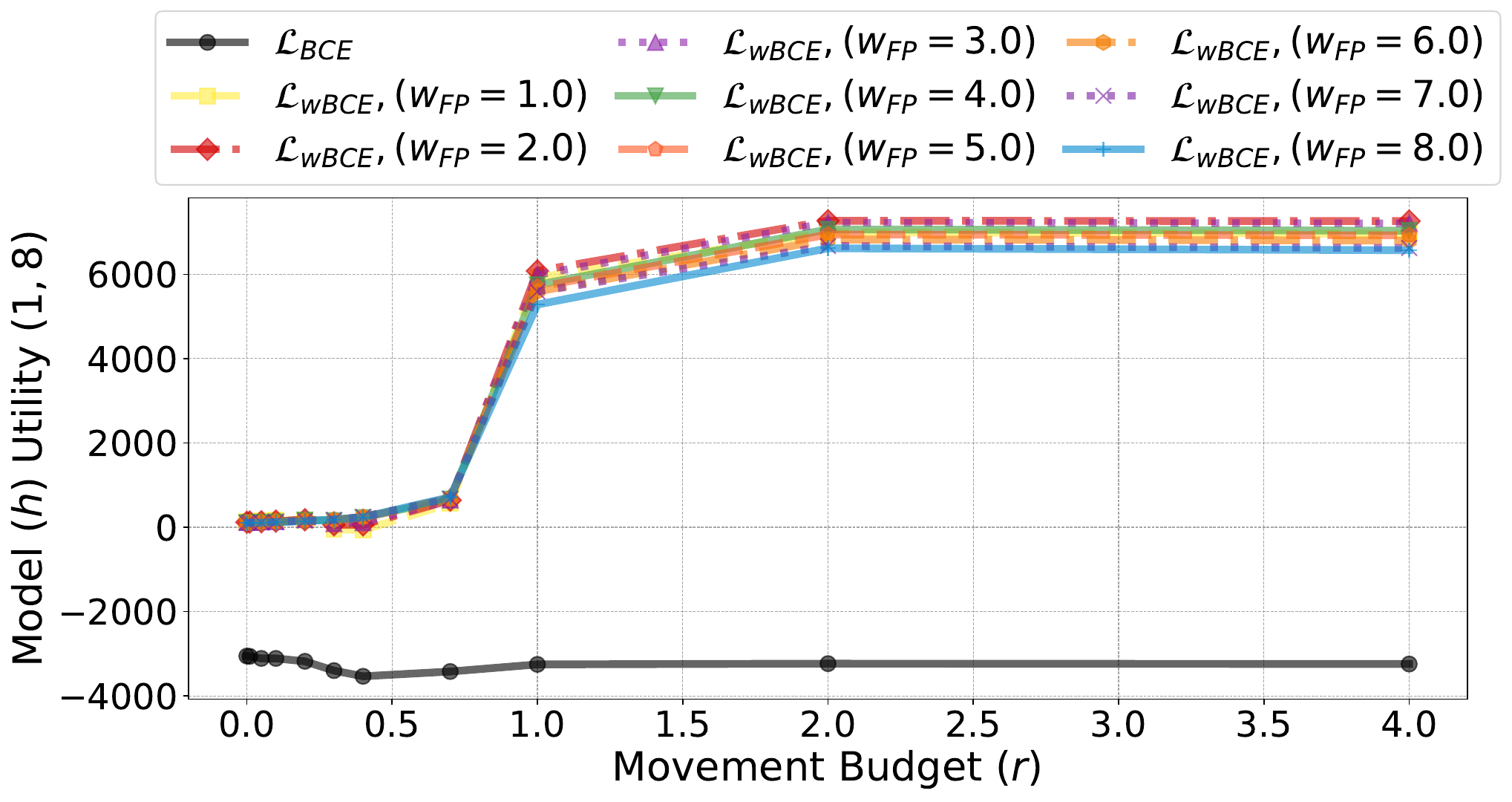}
    \caption{On the \textbf{Adult} dataset, the \textbf{utility(\(1,8\)) score} variations when agents respond to a BCE-trained model and wBCE-trained models with weight configurations \(\big(\mathbf{w_\textrm{FN}=0.001}, w_\textrm{FP}=\{i\}_{i=1}^{8}\big)\)}
    \label{fig:ocagi_adult_tpfp_0.5linfno_profit}
    \end{subfigure}
    \hfill
    \begin{subfigure}[t]{0.47\linewidth}
        \centering
        \includegraphics[width=\linewidth]{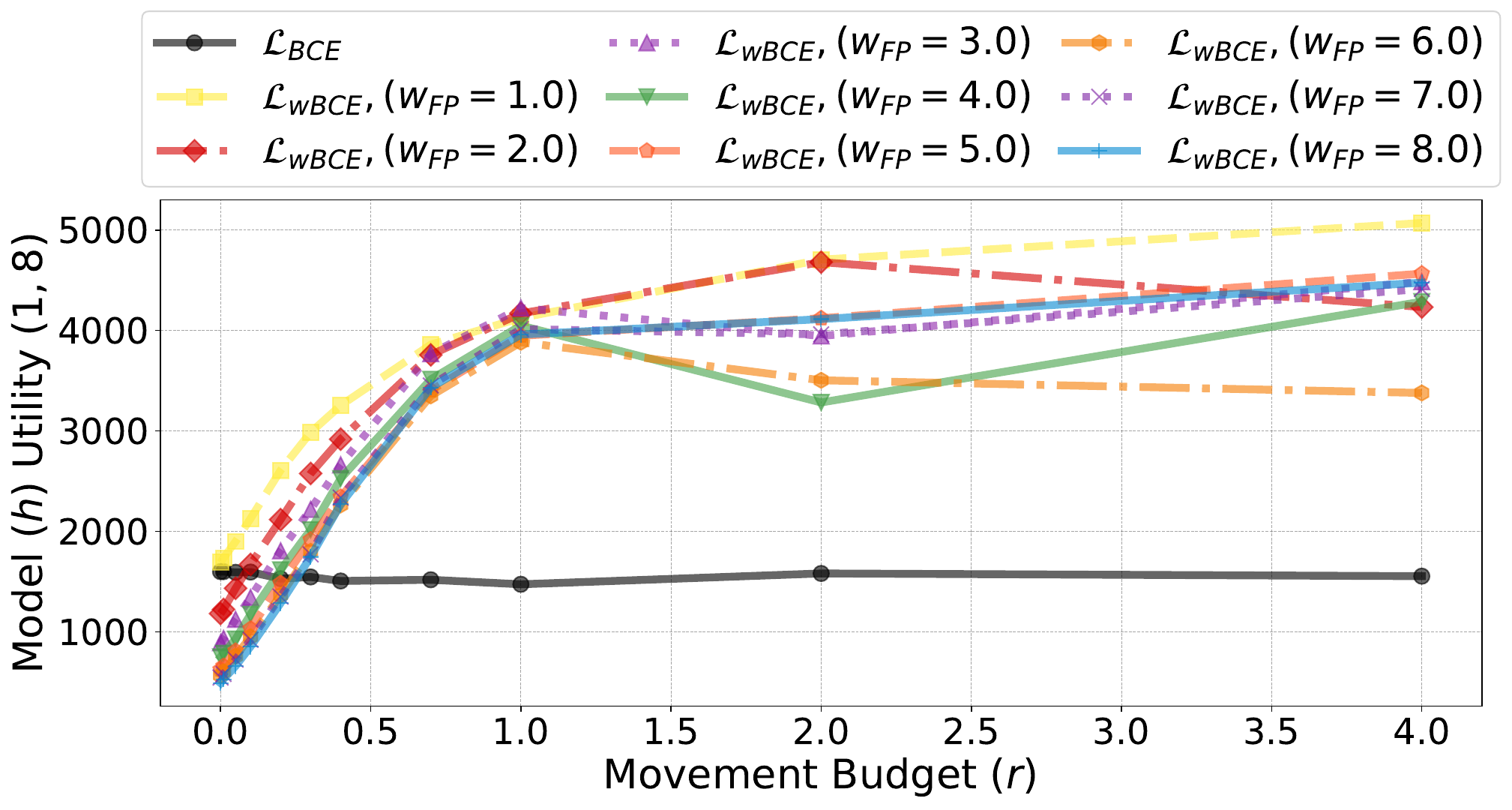}
    \caption{On the \textbf{Law School} dataset, the \textbf{utility(\(1,8\)) score} variations when agents respond to a BCE-trained model and wBCE-trained models with weight configurations \(\big(\mathbf{w_\textrm{FN}=0.009}, w_\textrm{FP}=\{i\}_{i=1}^{8}\big)\)}
    \label{fig:ocagi_law_tpfp_0.5linfno_profit}
    \end{subfigure}

    \vspace{1.2em}

    \begin{subfigure}[t]{0.47\linewidth}
        \centering
        \includegraphics[width=\linewidth]{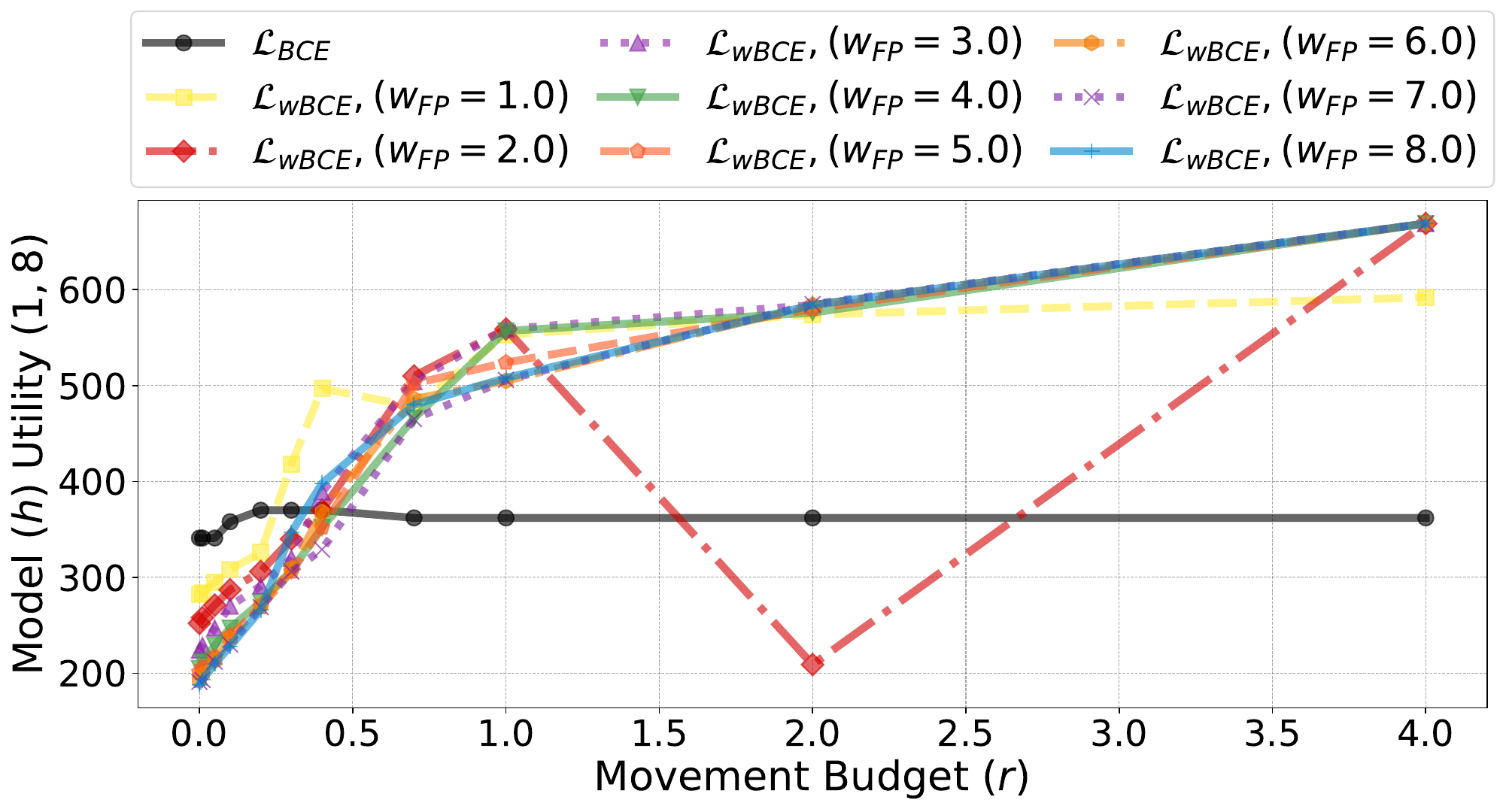}
    \caption{On the \textbf{Synthetic} dataset, the \textbf{utility(\(1,8\)) score} variations when agents respond to a BCE-trained model and wBCE-trained models with weight configurations \(\big(\mathbf{w_\textrm{FN}=0.009}, w_\textrm{FP}=\{i\}_{i=1}^{8}\big)\)}
    \label{fig:ocagi_synthetic_tpfp_0.5linfno_profit}
    \end{subfigure}
    \hfill
    \begin{subfigure}[t]{0.47\linewidth}
        \centering
        \includegraphics[width=\linewidth]{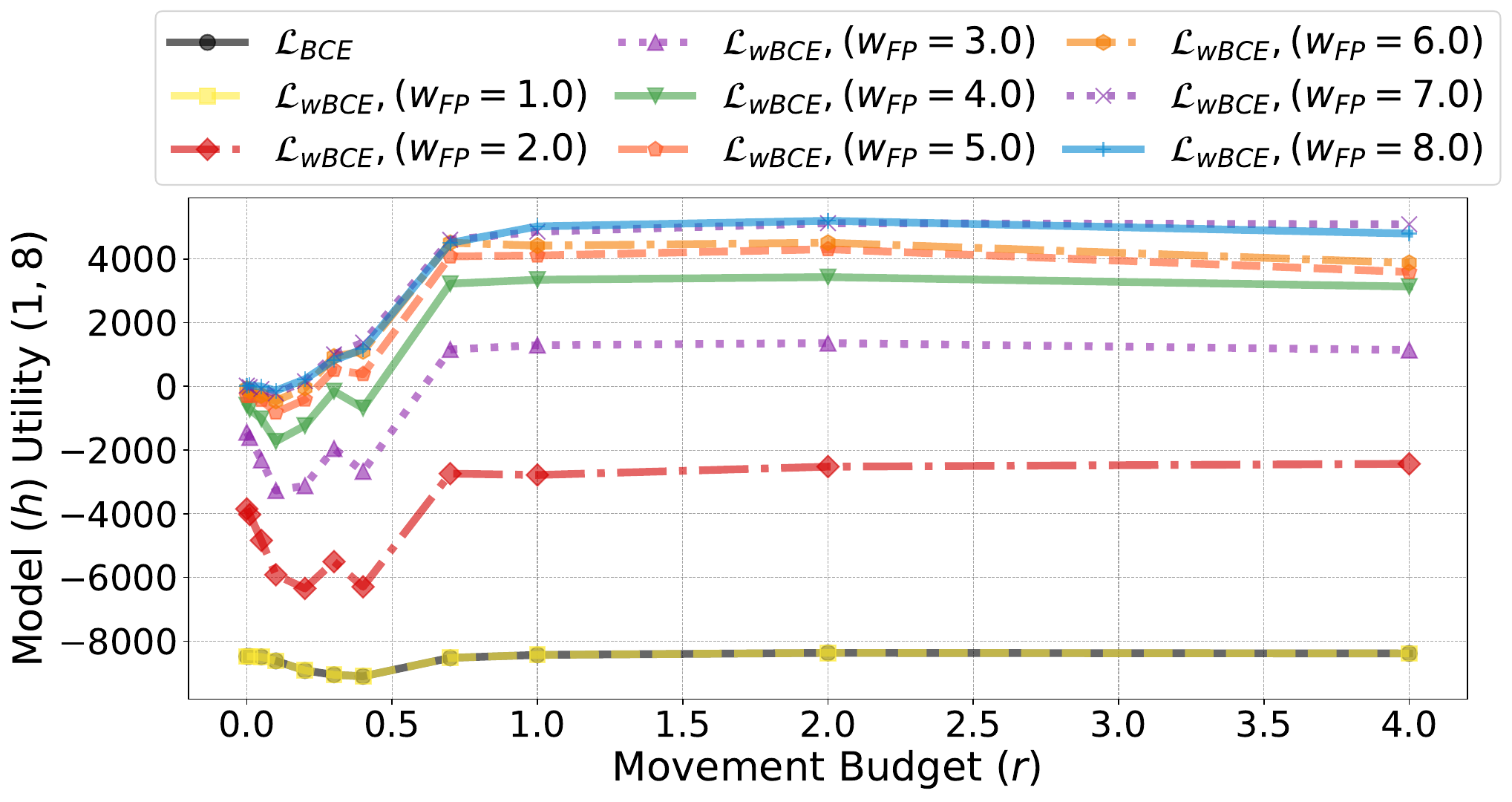}
    \caption{On the \textbf{OULAD} dataset, the \textbf{utility(\(1,8\)) score} variations when agents respond to a BCE-trained model and wBCE-trained models with weight configurations \(\big(\mathbf{w_\textrm{FN}=1.0}, w_\textrm{FP}=\{i\}_{i=1}^{8}\big)\)}
    \label{fig:ocagi_oulad_tpfp_0.5linfno_profit}
    \end{subfigure}

    \caption[A comparative analysis to study the weighted utility]{Comparative analysis of the variation in \(\textrm{utility}(1,8)\)  (see Equation~\ref{eq:ocagi_wutility}) scores when agents modify their features under different movement budgets \(r\) in response to a BCE-trained model and wBCE-trained models with varying weights across four datasets: Adult \subref{fig:ocagi_adult_tpfp_0.5linfno_profit}, Law School \subref{fig:ocagi_law_tpfp_0.5linfno_profit}, Synthetic \subref{fig:ocagi_synthetic_tpfp_0.5linfno_profit}, and OULAD \subref{fig:ocagi_oulad_tpfp_0.5linfno_profit}. 
    Agents game and improve in response to a BCE-trained model and wBCE-trained models with varying weighted false positive weights \(\big(w_\textrm{FP}=\{i\}_{i=1}^{8}\big)\) and specific false negative weights: \(\big(w_\textrm{FN}=0.001\big)\) on Adult \subref{fig:ocagi_adult_tpfp_0.5linfno_profit},  \(\big(w_\textrm{FN}=0.009\big)\) on Law School \subref{fig:ocagi_law_tpfp_0.5linfno_profit} and Synthetic \subref{fig:ocagi_synthetic_tpfp_0.5linfno_profit}, and \(\big(w_\textrm{FN}=1.0\big)\) on OULAD \subref{fig:ocagi_oulad_tpfp_0.5linfno_profit}.
    In all cases, all agents move within an \(\ell_{\infty}\) ball and are classified as positive if the probability is higher than \(0.5\).}
    \label{fig:allds_tpfp_th0.5linf_profits}
\end{figure*}

Empirical results show that risk-averse models trained with wBCE loss functions consistently outperform standard BCE-trained models in terms of \(\textrm{utility}(1,8)\) score as the improvement budget increases (Figure~\ref{fig:allds_tpfp_th0.5linf_profits}). However, the level of risk aversion required to achieve high \(\textrm{utility}(1,8)\) scores varies with dataset class separability. Datasets with high class separability, such as Law School and Synthetic, achieve high \(\textrm{utility}(1,8)\) with relatively low risk aversion. In contrast, datasets with low class separability, such as OULAD, require stronger risk aversion to attain high \(\textrm{utility}(1,8)\) gains.

To better understand the variation in \(\textrm{utility}(1,8)\) scores, we analyze the ratio of true-to-false positives after agents move. 
As shown in Appendix, Figure~\ref{fig:oulad_adult_tpfp_th0.5linfno}, when true-to-false positives ratio falls below the critical threshold of \(8\), corresponding to the cost ratio defined in Equation~\ref{eq:ocagi_wutility}, the \(\textrm{utility}(1,8)\) score becomes negative. For example, on the Adult dataset, the BCE-trained model (\textcolor{black}{\textbf{black}}) yields a true-to-false positive ratio of approximately \(2.5\) and consequently a negative \(\textrm{utility}(1,8)\) score. In contrast, the wBCE-trained model with \colorbox{yellow}{\(\{w_{FP}=1.0, \ w_{FN}=0.001\}\)}
achieves a ratio of approximately \(80\) and a positive utility score. 
Similarly, on the OULAD dataset, a wBCE-trained model with \colorbox{LightRed}{\(\{w_{FP}=2.0, \ w_{FN}=1.0\}\)} yields a ratio of 5.5 and negative \(\textrm{utility}(1,8)\), whereas increasing the false positive weight to \(8.0\) \colorbox{skyblue}{\(\{w_{FP}=8.0, \ w_{FN}=1.0\}\)} improves the ratio to approximately \(31\) and results in positive \(\textrm{utility}(1,8)\) score.

\subsection{Improvement Only vs Gaming+Improvement}
\label{subsec:ocagiexp_results_gameimprov}

Under comparable conditions, identical models, datasets, level and choice of risk-aversion, and classification thresholds, the \(\textrm{utility}\) score differs depending on whether agents can only improve or can both improve and game. This variation arises despite holding all other parameters constant, indicating that the nature of agents' strategic behavior significantly influences \(\textrm{utility}\) scores, most especially since the movement space slightly varies as agents who can game and improve access the whole feature space and those restricted to improvement operate within a constrained subspace.

\begin{figure*}[ht!]
    \centering

    \begin{subfigure}[t]{0.47\linewidth}
        \centering
        \includegraphics[width=\linewidth]{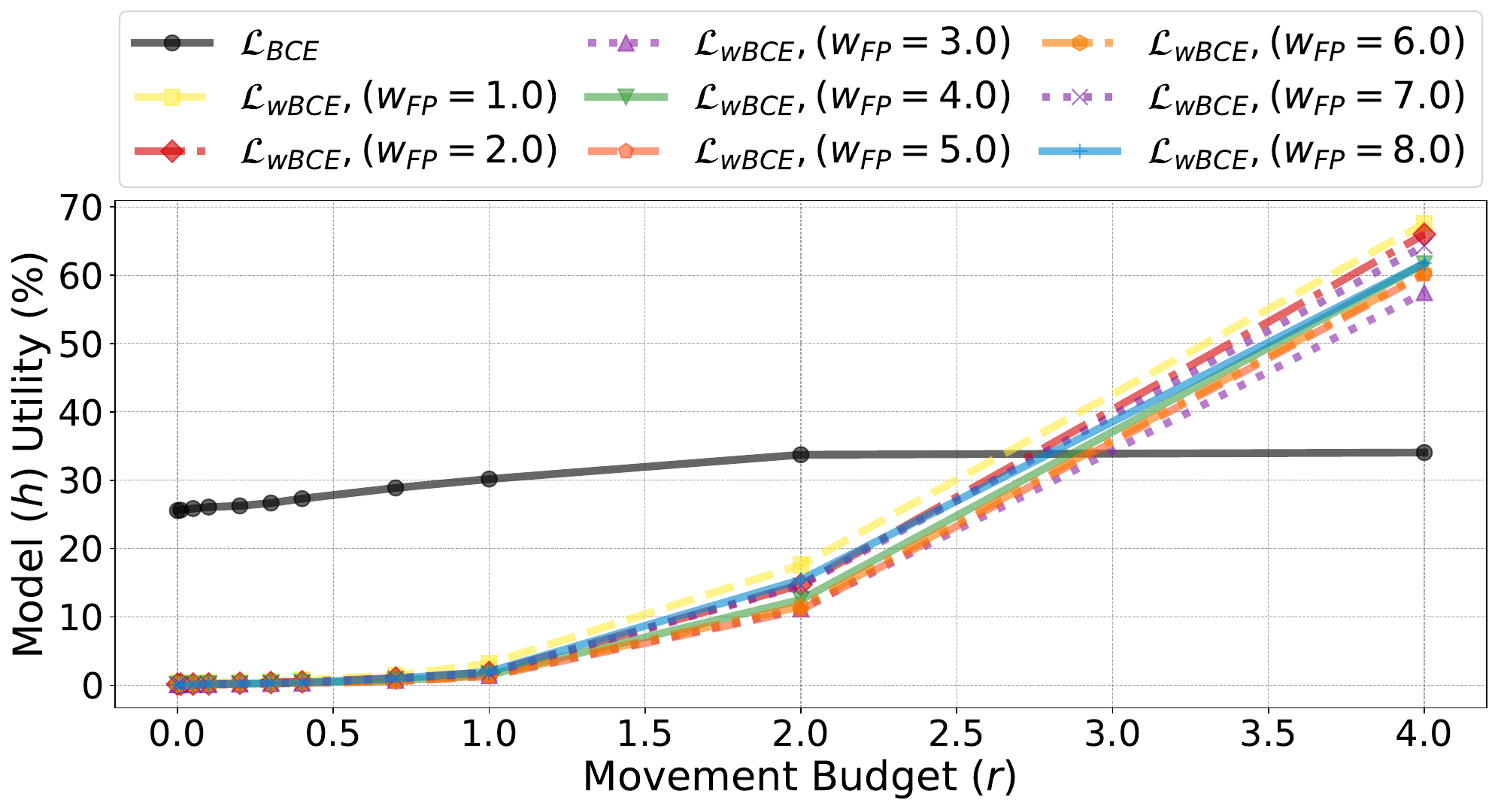}
    \caption{On the \textbf{Law School} dataset, the \(\textrm{utility}\) scores when agents can \textbf{both game and improve} in response to a BCE-trained model and wBCE-trained models with weights \(\big(w_\textrm{FN}=0.009, w_\textrm{FP}=\{i\}_{i=1}^{8}\big)\) and a classification threshold of \(\mathbf{0.99}\)}
    \label{fig:ocagi_law_tpfp_0.99l2no_0.009_wfpnvar}
    \end{subfigure}
    \hfill
    \begin{subfigure}[t]{0.47\linewidth}
        \centering
        \includegraphics[width=\linewidth]{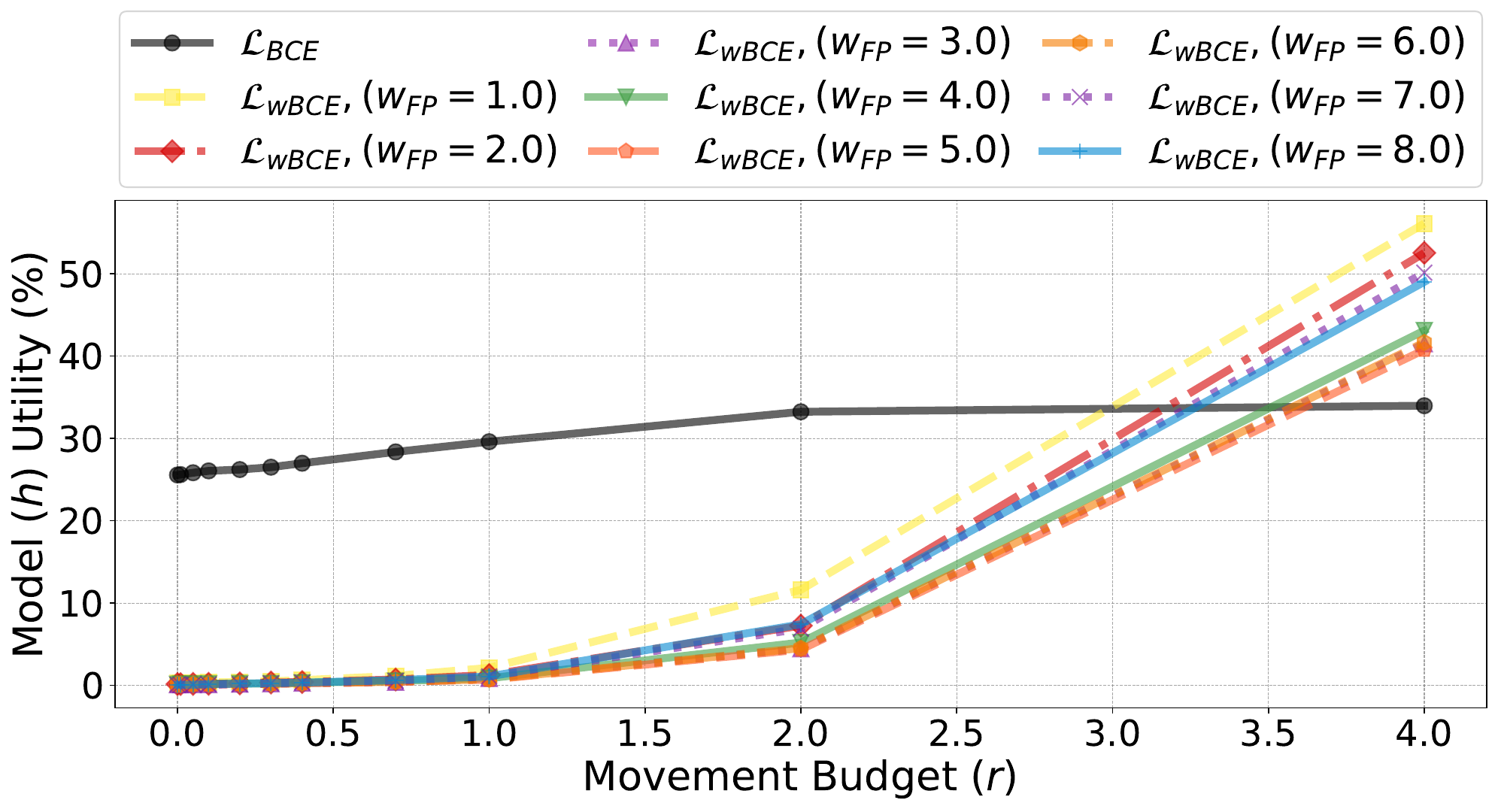}
    \caption{On the \textbf{Law School} dataset, the \(\textrm{utility}\) score variation when agents can \textbf{only improve} in response to a BCE-trained model and wBCE-trained models with weights \(\big(w_\textrm{FN}=0.009, w_\textrm{FP}=\{i\}_{i=1}^{8}\big)\) and a classification threshold of \(\mathbf{0.99}\)}
    \label{fig:ocagi_law_tpfp_0.99l2yes_0.009_wfpnvar}
    \end{subfigure}

    \vspace{1.2em}

    \begin{subfigure}[t]{0.47\linewidth}
        \centering
        \includegraphics[width=\linewidth]{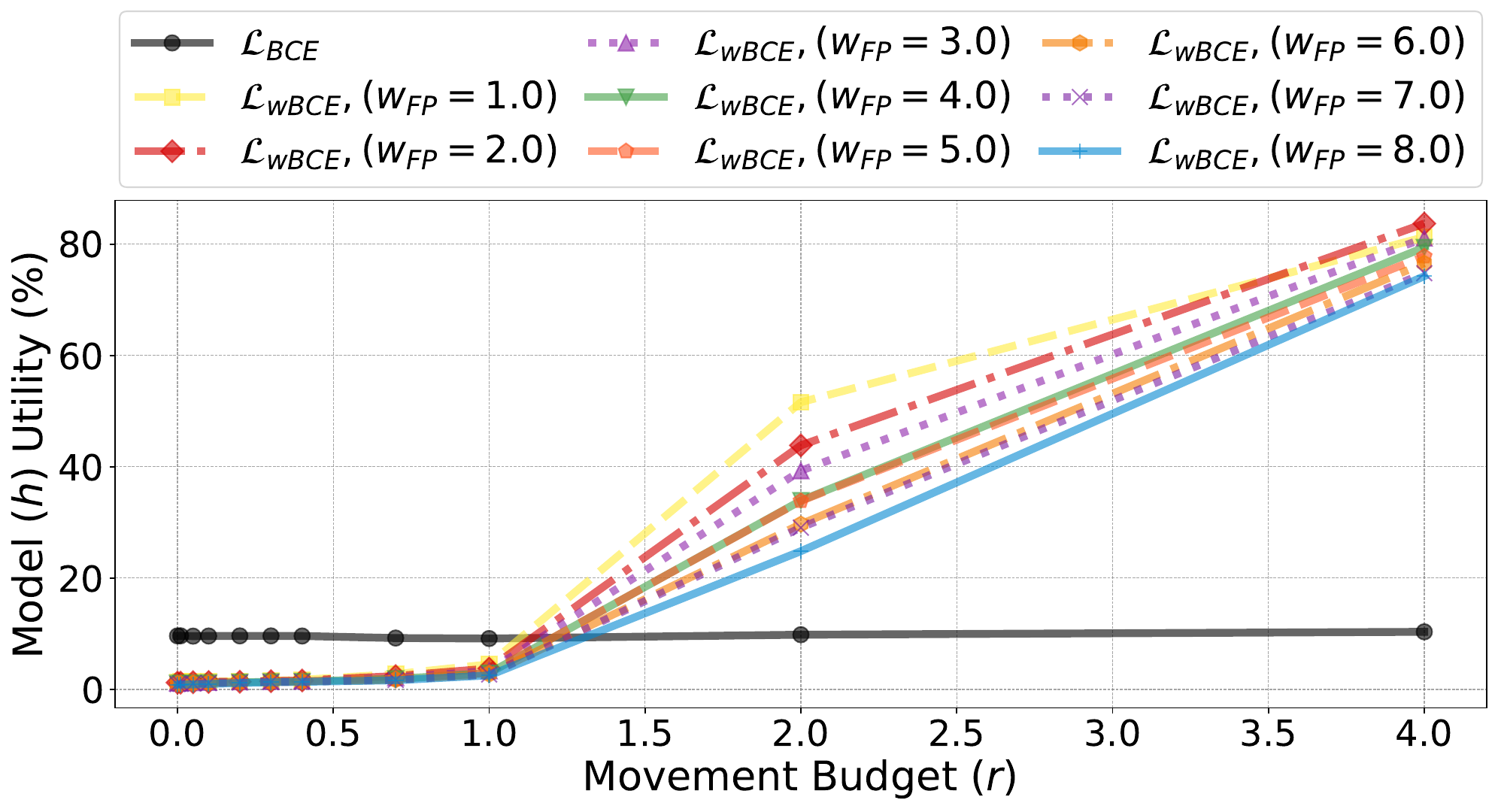}
    \caption{On the \textbf{Adult} dataset, the \(\textrm{utility}\) score variation when agents can \textbf{both game and improve} in response to a BCE-trained model and wBCE-trained models with weights \(\big(w_\textrm{FN}=0.001, w_\textrm{FP}=\{i\}_{i=1}^{8}\big)\) and a classification threshold of \(\mathbf{0.9}\)}
    \label{fig:ocagi_adult_tpfp_0.9l2no_0.001_wfpnvar}
    \end{subfigure}
    \hfill
    \begin{subfigure}[t]{0.47\linewidth}
        \centering
        \includegraphics[width=\linewidth]{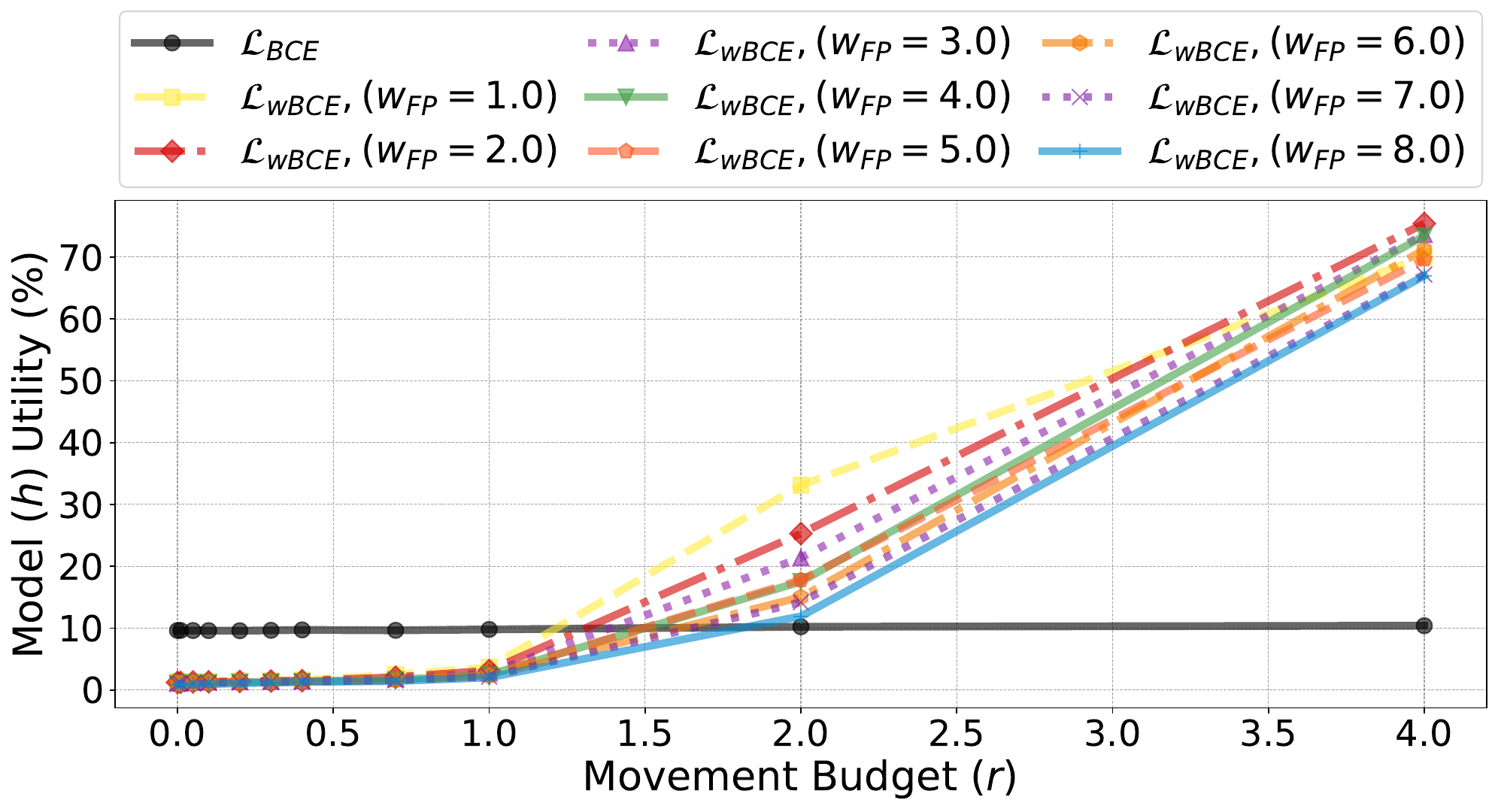}
    \caption{On the \textbf{Adult} dataset, the \(\textrm{utility}\) score variation when agents can \textbf{only improve} in response to a BCE-trained model and wBCE-trained models with weights \(\big(w_\textrm{FN}=0.001, w_\textrm{FP}=\{i\}_{i=1}^{8}\big)\) and a classification threshold of \(\mathbf{0.9}\)}
    \label{fig:ocagi_adult_tpfp_0.9l2yes_0.001_wfpnvar}
    \end{subfigure}

    \caption[A comparative analysis to study the effect of improvement only vs. ]{Comparative analysis of the variation in \(\textrm{utility}\) scores when agents modify their features within an \(\ell_{2}\) ball under different movement budgets \(r\) in response to BCE- and wBCE-trained models. Figures~\subref{fig:ocagi_law_tpfp_0.99l2no_0.009_wfpnvar} and \subref{fig:ocagi_law_tpfp_0.99l2yes_0.009_wfpnvar} show results for the Law School dataset, while \subref{fig:ocagi_adult_tpfp_0.9l2no_0.001_wfpnvar} and \subref{fig:ocagi_adult_tpfp_0.9l2yes_0.001_wfpnvar} show results for the Adult dataset. The analysis considers scenarios where agents can both game and improve (\subref{fig:ocagi_law_tpfp_0.99l2no_0.009_wfpnvar} and \subref{fig:ocagi_adult_tpfp_0.9l2no_0.001_wfpnvar}) versus scenarios where agents can only improve (\subref{fig:ocagi_law_tpfp_0.99l2yes_0.009_wfpnvar} and \subref{fig:ocagi_adult_tpfp_0.9l2yes_0.001_wfpnvar}).}
    \label{fig:adult_errors_tpfp_th0.5linf_yes/no}
\end{figure*}

Figure~\ref{fig:adult_errors_tpfp_th0.5linf_yes/no} shows that, on average, when agents can both improve and game, the models achieve a higher \(\textrm{utility}\) score than when they can only improve. This effect is particularly evident when the model has high penalties for false positives, suggesting that gaming can offset such penalties when agents can both game and improve.

\section{Conclusion}
\label{sec:ocagiexp_discussion}

Empirical results show that the loss-based risk aversion strategic classifier designs get better utility scores and error reductions than threshold-based methods when classifying agents who can both game and improve.  Although increasing risk aversion generally leads to a drop in classification error (sometimes to zero), excessive risk aversion might reduce the utility scores due to costly feature changes that discourage negatively classified agents from moving. Both unweighted and weighted utility scores increase with a larger movement budget under risk-averse models; however, the weighted utility score becomes negative when the true-to-false positive ratio drops below a critical threshold of \(8\). Movement constraint geometry also matters: \(\ell_{\infty}\) norms, which clip individual feature changes, allow more movements than \(\ell_{2}\) norms, which limit total vector magnitude. Finally, utility scores are lower when agents can only improve than when they can both improve and game.

\chapter{Directions for Future Research}
\label{chap:conclusion}
This thesis presents various approaches to understanding interactions among agents and between agents and AI-driven systems, and how to shape these interactions to incentivize agents' improvement while enabling both agents and systems to achieve their respective objectives.
Each thesis chapter concludes with a discussion of potential future directions, and in some cases, these are expanded upon in the chapter's appendix. In this overview, however, we focus on broader research directions that apply to the thesis as a whole. These include the following.

\paragraph{Question 1.} 
\textit{How does the inclusion or exclusion of a particular agent reshape agent-AI interactions and affect the objectives of both the participating agents and the AI system?}

The work presented in this thesis focuses on shaping agent-AI interactions to create pathways for agent improvement and to jointly optimize the objectives of both agents and the AI system. However, in many practical settings, such as labor markets, agents' behavior is often influenced by the actions of other agents. For example, \citet{PascualEzama2015} shows that individuals increase their level of cheating when they are aware of others doing so, and \citet{Carrell173} finds that higher levels of peer cheating significantly raises the likelihood that an individual will cheat. These findings suggest that when designing AI-driven decision systems (e.g., for admissions and hiring), it is crucial to account for both agent-agent and agent-AI interactions, a perspective briefly explored by \citet{hossain2025strategic}.

Future work will adopt a broader perspective by investigating how in-group and out-group peer effects, along with social incentives, shape both positive and negative externalities and influence shifts in agents' improvement and gaming behaviors. It will also investigate if predictions for a given agent could change when another agent is added to or removed from the model's training data, especially in settings where the data may be strategically manipulated (e.g., in online strategic classification).

\paragraph{Question 2.} 
\textit{How will generative AI affect predictive decision-making and strategic learning?}

The strategic classification literature, including the work presented Chapters~\ref{chap:setfair} \ref{chap:paclearn}, and \ref{chap:ocagi_theory} typically assumes that agents share a common belief and or understanding of the AI-driven decision system and can allocate effort across features in a fully rational and comparable way to achieve favorable outcomes. In practice, these assumptions rarely hold.
Generative AI tools, particularly large language models (LLMs), may fundamentally reshape this landscape by lowering informational barriers and redistributing strategic capabilities. New research suggests that agents can use LLMs to infer which features to modify even without direct knowledge of the decision rule \citep{TianXPX25}, while unequal access to such tools may produce new forms of disparity \citep{LeeCJK25}.

Future work will investigate the psychological and social consequences of these shifts, including how reliance on AI-generated guidance alters the agents' and decision-makers' sense of agency, responsibility, and risk-taking. At the same time, the growing integration of LLMs into various high-stakes domains (e.g., hiring, admissions, and lending) raises questions about labor displacement, skill transformation, and resource allocation, as both agents and decision-makers increasingly depend on automated assistance. Future work will also examine the legal and economic ramifications of AI-mediated strategies\footnote{For instance,  
\href{https://www.nytimes.com/2026/01/21/business/ai-hiring-tools-lawsuit-eightfold-fcra.html?unlocked_article_code=1.GFA.9XQK.n_nH_2Z3omQR&smid=url-share}{the class action lawsuit filed against Eightfold AI for violating the Fair Credit Reporting Act}}, 
such as fairness, transparency and accountability for AI-assisted strategic manipulation, and the regulatory and system updates challenges posed by rapidly adaptive decision environments. Together, these directions situate predictive systems within the broader social infrastructure shaped by emerging technologies.

\paragraph{Question 3.} 
\textit{How does the medium through which a decision-maker communicates with agents affect agents' best responses?}

Beyond the prediction itself, the channel through which a decision-making system communicates with individuals being evaluated can shape psychological responses, including risk aversion, perceived legitimacy, and willingness to comply with or contest decision outcomes. My prior work on child-device and parent-child interactions shows that perceptions of animacy and anthropomorphism influence how people engage with AI systems and other individuals (Chapter~\ref{chap:parental_concern}). This suggests that communication modalities are not merely technical design choices but carry important social consequences. Decision-making systems both elicit individuals' revealed preferences (e.g., input features) and communicate decision outcomes through a range of channels, including in-person interactions, online interfaces, automated notifications, human intermediaries, and embodied technologies such as robots.

Future research will examine how these communication modes affect trust, perceived fairness, accountability, and strategic behavior across diverse populations. This agenda could, for example, be useful for AI ethics education as it raises broader questions about the responsible deployment of emerging technologies in institutional settings, where interface design choices may carry legal and organizational implications. Addressing these issues will require an interdisciplinary approach that integrates algorithm design, human-subject experimentation, large-scale data analysis,  and socially informed development of AI systems.

\begin{appendices}

%
\chapter{Setting Reachable Targets to Maximize People's Improvement}
\label{app:setfair}
\section{Missing Proofs of Section~\ref{sec:setfair_max_total_improvement}}
\label{app:setfair_missing_max_tot}

\subsection{Proof of Theorem~\ref{thm:total-improvement}}

\begin{numberedtheorem}{\ref{thm:total-improvement}}
Algorithm~\ref{alg:recurrence-dp-one-group} finds a set of targets that achieves the optimal social welfare (maximum total improvement) that is feasible using at most $k$ targets given $n$ agents. The algorithm runs in $\mathcal{O}(n^3)$.
\end{numberedtheorem}

\begin{proof}
Proof of correctness follows by induction. Suppose that the value computed for all $T(\tau',\kappa')$ where $(\tau',\kappa') < (\tau,\kappa)$ is correct. Here ``$<$'' means 
$(\tau',\kappa')$ is computed before $(\tau,\kappa)$ which is when $\kappa'<\kappa$ and $\tau'\geq \tau$.  First, if either $\tau= \tau_{\text{max}}$ or $\kappa= 0$, the induction hypothesis holds since $T(\tau_{\max},\kappa)=0$ for all $1\leq \kappa\leq k$, and $T(\tau,0) = 0$, for all $\tau\in\T_p$. To show the inductive step holds note that the algorithm considers the optimal value for $T(\tau,\kappa)$ as the maximum of the $\displaystyle T(\tau',\kappa-1) \ + \textstyle\sum_{\tau\leq p_i <\tau'\text{ s.t. } \tau'-p_i \leq\Delta_i}(\tau'-p_i)$ over all the possible placement of the leftmost target $\tau'$. Since $T(\tau',\kappa-1)$ is computed correctly by the induction hypothesis and all the possible placements of the leftmost target are considered, the value obtained at $T(\tau,\kappa)$ is optimal and correct. 

Now we proceed to bounding the time-complexity. There are $\mathcal{O}(nk)$ subproblems to be computed. Consider a 
pre-computation stage where  $\sum_{\tau\leq p_i <\tau'\text{ s.t. } \tau'-p_i \leq\Delta_i}(\tau'-p_i)$ is computed for all pairs of $\tau, \tau'\in \T_p$. This stage takes $\mathcal{O}(n^3)$ time. Computation of each subproblem $T(\tau,\kappa)$ for all $\tau\in \T_p$ and $1\leq \kappa \leq k$ requires $\mathcal{O}(n)$ operations. This is because to compute $\max$ in property 3), we compute $\displaystyle T(\tau',\kappa-1) \ + \textstyle\sum_{\tau\leq p_i <\tau'\text{ s.t. } \tau'-p_i \leq\Delta_i}(\tau'-p_i)$ for $\mathcal{O}(n)$ potential target levels greater than $\tau$, for which each takes $\mathcal{O}(1)$ time. Since there are $\mathcal{O}(nk)$ subproblems, the running time of the algorithm is $\mathcal{O}(n^2k +n^3) = \mathcal{O}(n^3)$. 
\end{proof}

\section{Missing Proofs of Section~\ref{sec:setfair_max_min}}
\label{app:setfair_missing_max_min}

\subsection{Proof of Proposition~\ref{prop:running-time-fairness-exact}}

\begin{numberedtheorem}{\ref{prop:running-time-fairness-exact}}
Algorithm~\ref{alg:recurrence-exact-fairness-objective} constructs the Pareto frontier for groups' social welfare using at most $k$ targets given $n$ agents in $g$ groups, and has a running time of $\mathcal{O}(n^{g+2}kg\Delta_{\max}^g)$, where $\Delta_{\max}$ is the maximum improvement capacity.
\end{numberedtheorem}

\begin{proof}
Proof of correctness follows by induction and it is along the same lines as proof of Algorithm~\ref{alg:recurrence-dp-one-group}. 
Suppose that Pareto-frontiers constructed for all $T(\tau',\kappa')$ where $(\tau',\kappa') < (\tau,\kappa)$ is correct. Here ``$<$'' means 
$(\tau',\kappa')$ is computed before $(\tau,\kappa)$ which is when $\kappa'<\kappa$ and $\tau'\geq \tau$.  First, if either $\tau= \tau_{\text{max}}$ or $\kappa= 0$, the induction hypothesis holds since $T(\tau_{\max},\kappa)=\emptyset$ for all $1\leq \kappa\leq k$, and $T(\tau,0) = \emptyset$, for all $\tau\in\T_p$. The inductive step holds since the algorithm considers all the possible placement of the leftmost target $\tau'$. Since $T(\tau',\kappa-1)$ is computed correctly by the induction hypothesis and all the possible placements of the leftmost target are considered, the Pareto-frontier constructed at $T(\tau,\kappa)$ is correct.

Now we proceed to bounding the time complexity. Initially, in a pre-computation stage, for each pair of targets $\tau, \tau'\in \mathcal{T}_p$,  $\sum_{\tau\leq p_i <\tau'\text{ s.t. } \tau'-p_i \leq\Delta_{\ell}} \mathbbm{1}\big\{i\in G_{\ell}\big\}(\tau'-p_i)$ is pre-computed for all groups and is stored in a tuple of size $g$. This stage can be done in $\mathcal{O}(n^3)$.
Each set $T(\tau,\kappa)$ has size at most $(n\Delta_{\max}+1)^g$, since each individual can move for one of the values $\{0,\cdots,\Delta_{\max}\}$ and therefore, the total improvement in each group is one of the values $\{0,\cdots,n\Delta_{\max}\}$.
At each step of the recurrence, given the information stored in the pre-computation stage, the summation can be computed in $\mathcal{O}(g)$. 
When computing a subproblem $T(\tau,\kappa)$, the recurrence searches over $\mathcal{O}(n)$ targets $\tau'\in \T_p$, and at most $(n\Delta_{\max}+1)^g$ tuples of group improvement in $T(\tau',\kappa-1)$. As a result, solving each subproblem takes $\mathcal{O}(ng(n\Delta_{\max})^g)$.
The total number of subproblems that need to get solved is $\mathcal{O}(nk)$.
Therefore, the total running time of the algorithm is $\mathcal{O}(n^{g+2}kg\Delta_{\max}^g+n^3)$ = $\mathcal{O}(n^{g+2}kg\Delta_{\max}^g)$.
\end{proof}

\subsection{Proof of Corollary~\ref{cor:exact-fairness}}

\begin{numberedcorollary}{\ref{cor:exact-fairness}}
There is an efficient algorithm that finds a set of at most $k$ targets that maximizes minimum improvement across all groups, i.e., maximizing $\min_{1 \leq \ell \leq g} \sw_\ell$.
\end{numberedcorollary}
\begin{proof}
Algorithm~\ref{alg:recurrence-exact-fairness-objective} constructs the Pareto frontier for groups' social welfare. By iterating through all Pareto-optimal solutions, we can find the solution that maximizes the minimum improvement across all groups. There are at most $(n\Delta_{\max}+1)^g$ Pareto-optimal solutions. Finding the minimum improvement in each solution takes $\mathcal{O}(g)$. Therefore, in total, finding the solution that maximizes the minimum improvement across all groups takes $\mathcal{O}(g(n\Delta_{\max})^g)$.
\end{proof}

\section{An FPTAS for Maximizing Minimum Group Improvement}
\label{sec:setfair_appendix-FPTAS}

In this section, we present a Fully Polynomial Time Approximation Scheme (FPTAS) to maximize minimum improvement across all groups. Here, we assume that each group $\ell$ has its own improvement capacity $\Delta_{\ell}$.

\begin{lemma}
\label{thm:approx-guarantee} 
Algorithm~\ref{algo:Max-min} gives a ($1-\eps$)-approximation for the max-min objective when $k\geq g$.
\end{lemma}

\begin{proof}
The proof is by induction. Consider an improvement tuple $(I_1,\cdots,I_g)$ corresponding to an arbitrary set of $k-1$ targets, 
and let $(I'_1,\cdots,I'_g)$ denote the rounded down values where $I'_{\ell} = \mu_{\ell}\floor{\frac{I_{\ell}}{\mu_{\ell}}}$ for all $1\leq \ell \leq g$. Suppose that for all $1\leq \ell \leq g$,
$I'_{\ell} \geq I_{\ell}-(k-1)\mu_{\ell}$.

Now consider an improvement tuple $(J_1,\cdots,J_g)$ corresponding to an
arbitrary set of $k$ targets. For each $1\leq \ell \leq g$, let $J'_{\ell} = \mu_{\ell}\floor{\frac{J_{\ell}}{\mu_{\ell}}}$. We show that for each $1\leq \ell\leq g$, $J'_{\ell} \geq J_{\ell} - k\mu_{\ell}$.
For all $1\leq \ell \leq g$, let $J_{\ell} = L_{\ell} + I_{\ell}$, where $L_{\ell}$ is the improvement of group $\ell$ that the leftmost target provides, and $I_{\ell}$ captures the true improvement of group $\ell$ that the remaining $k-1$ targets provide. Let $I'_{\ell} = \mu_{\ell}\floor{\frac{I_{\ell}}{\mu_{\ell}}}$. 
Then $J'_{\ell} = \mu_{\ell}\floor{\frac{L_{\ell}+I'_{\ell}}{\mu_{\ell}}}$ implying that $J'_{\ell}\geq L_{\ell} + I'_{\ell}-\mu_{\ell}$. By the induction hypothesis, $I'_{\ell} \geq I_{\ell}-(k-1)\mu_{\ell}$. Therefore,

\[
J'_{\ell} \geq L_{\ell} + I'_{\ell}-\mu_{\ell}
\geq L_{\ell}+I_{\ell}-(k-1)\mu_{\ell}-\mu_{\ell}
= L_{\ell}+I_{\ell}-k\mu_{\ell}
\]
Therefore, for each set of $k$ targets, the rounded improvement of each group ${\ell}$ stored in the table is within an additive factor of $k\mu_{\ell} = \eps\Delta_{\ell}/(16g^3)$ of its true improvement. We argue that in the solution returned by the algorithm, improvement of each group is at least $(1-\eps)OPT$. First, when $k\geq 1$, each group can improve for at least $\Delta_{\ell}$ by setting a target within a distance of $\Delta_{\ell}$ from its rightmost agent. Now, using Theorem~\ref{thm:approx} when $k\geq g$, there exists a solution that is simultaneously $1/(16g^3)$-optimal for all groups. Therefore, the optimum value of the max-min objective is at least $OPT\geq \Delta_{\ell}/(16g^3)$ for all $1\leq \ell \leq g$. Therefore, for each solution consisting of $k$ targets, the rounded improvement of each group is within an additive factor of $\eps OPT$ of its true improvement.
As a result, the minimum group improvement in the returned solution is at least $(1-\eps)OPT$. \end{proof}

\paragraph{}
\makebox[\textwidth][c]{%
\begin{minipage}{1.06\textwidth} 
\setlength{\algoheightrule}{0pt}
\setlength{\algotitleheightrule}{0pt}
\begin{algorithm}[H]
  \caption[Find an optimal solution for the max-min objective]
  {The algorithm considers two separate cases of $k< g$, and $k\geq g$. For the $k\geq g$ case, the algorithm finds a set of $k$ targets that approximates the max-min objective within a factor of $1-\eps$ for any arbitrary value of $\eps>0$. For the $k<g$ case, it finds an optimal solution for the max-min objective.}
  \SetAlgoLined
    \label{algo:Max-min}
    \paragraph{}
    \begin{enumerate}[left=0pt..1em, rightmargin=2cm]
    \justifying
    \item[] For the $k\geq g$ case, there exists an FPTAS for the max-min objective as follows. First, run a dynamic program using the following recursive function to get a set of Pareto-optimal solutions. In this Pareto-frontier, we show the solution that maximizes minimum improvement across all groups, gives a ($1-\eps$)-approximation for the max-min objective. In the recurrence, $\mu_{\ell} = \varepsilon\Delta_{\ell}/(16kg^3)$ for $1\leq \ell\leq g$, and $\Delta_{\ell}$ is the improvement capacity of agents in group $\ell$.

    \item[] {\footnotesize
            \begin{gather*} 
            \mathcal{F}(\tau',k') = \Bigg\{\Bigg(\mu_{\ell} \Bigg\lfloor\frac{I'_{\ell}+\sum_{\substack{\tau'\leq p_i <\tau\\ \text{s.t} \ \tau-p_i \leq\Delta_{\ell}}} \mathbbm{1}\{i\in G_{\ell}\} (\tau-p_i)}{\mu_{\ell}} \Bigg\rfloor \Bigg)_{\ell=1}^{g} \Bigg| (I'_{\ell})_{\ell=1}^{g} \in \mathcal{F}(\tau,k'-1), \tau\in \mathcal{T}_p, \tau\geq \tau' \Bigg\}
            \end{gather*}
        }

    \item[] Intuitively, $\mathcal{F}(\tau',k')$ stores the rounded down values of the feasible tuples of group improvements when all agents on or to the right of $\tau'$ are available and $k'$ targets are used. The corresponding set of targets used to construct the improvement tuples in $\mathcal{F}(\tau',k')$ is kept in a hash table $\mathcal{S}(\tau',k')$, whose keys are the improvement tuples in $\mathcal{F}(\tau',k')$. The dynamic program ends after computing $\mathcal{F}(\tau_{\min},k)$ and $\mathcal{S}(\tau_{\min},k)$. At the end, we output the set of targets in $\mathcal{S}(\tau_{\min},k)$  corresponding to the improvement tuple that maximizes the improvement of the worst-off group.
    
    \item[] Lemma~\ref{thm:approx-guarantee} shows that this algorithm gives a ($1-\eps$)-approximation for the max-min objective when $k\geq g$.

    \item[] When $k<g$, for each subset of $\T_p$ of size at most $k$ that is corresponding to a placement of targets, we store its corresponding improvement tuple. Next, we iterate through all improvement tuples and output the one that maximizes minimum improvement.
    \end{enumerate}
\end{algorithm}
\setlength{\algoheightrule}{1pt} 
\setlength{\algotitleheightrule}{1pt} 
\end{minipage}
}

In the following, we bound the approximation factor of our algorithm in both cases of $k\geq g$ and $k<g$.

\begin{corollary}
Algorithm~\ref{algo:Max-min} described above gives a ($1-\varepsilon$)-approximation for the max-min objective.
\end{corollary}

\begin{proof}
For the case of $k\geq g$, by Theorem~\ref{thm:approx-guarantee} the algorithm outputs a ($1-\varepsilon$)-approximation. For $k<g$, it outputs an optimum solution. Therefore, in total, it gives a ($1-\varepsilon$)-approximation for the max-min objective.
\end{proof}

In the following, we bound the time-complexity of the algorithm.

\begin{theorem}
\label{thm:running-time}
Algorithm~\ref{algo:Max-min} has a running time of
$\mathcal{O}(n^{g+2}k^{g+1}g^{3g+1}/\eps^g)$.
\end{theorem}

\begin{proof}
Initially, in a pre-computation stage, for each pair of targets \\
$\tau, \tau'\in \mathcal{T}_p$,  $\sum_{\tau'\leq p_i <\tau\text{ s.t. } \tau-p_i \leq\Delta_{\ell}} \mathbbm{1}\Big\{i\in G_{\ell}\Big\}(\tau-p_i)$ is pre-computed for all groups and is stored in a tuple of size $g$.
This stage can be done in $\mathcal{O}(n^3)$.
Now, first consider the case where $k\geq g$. We show the dynamic programming algorithm using recurrence $\mathcal{F}(\tau',k')$ has a running time of 
$\mathcal{O}(n^{g+2}k^{g+1}g^{3g+1}/\eps^g)$.
 Each set $\mathcal{F}(\tau',k')$ and $\mathcal{S}(\tau',k')$ has size at most $\prod_{\ell=1}^{g}(n\Delta_{\ell}/\mu_{\ell})^{g} = (16nkg^3/\eps)^{g}$.
At each step of the recurrence, given the information stored in the pre-computation stage, the summation can be computed in $\mathcal{O}(g)$.
When computing $\mathcal{F}(\tau',k')$, the recurrence searches over $\mathcal{O}(n)$ targets $\tau\in \mathcal{T}_p$, and at most  $\prod_{\ell=1}^{g}(n\Delta_{\ell}/\mu_{\ell})^{g} = (16nkg^3/\eps)^{g}$ tuples of group improvement in $\mathcal{F}(\tau,k'-1)$. As a result, solving each subproblem takes $\mathcal{O}(ng(nkg^3/\eps)^{g})$. The total number of subproblems that need to get solved is $\mathcal{O}(nk)$. Therefore, the total running time of computing $\mathcal{F}(\tau_{\min},k)$ is 
$\mathcal{O}(n^{g+2}k^{g+1}g^{3g+1}/\eps^g)$.

Next, consider the case where $k<g$. The algorithm considers $\mathcal{O}(n^g)$ placements of targets. Given the pre-computation stage,
computing the improvement tuple corresponding to each placement of targets   takes $\mathcal{O}(kg)$. As a result, this case takes $\mathcal{O}(kgn^g)$. 

Therefore, the total running time of algorithm is \\
$\mathcal{O}(n^3+n^{g+2}k^{g+1}g^{3g+1}/\eps^g+kgn^g) = \mathcal{O}(n^{g+2}k^{g+1}g^{3g+1}/\eps^g)$.
\end{proof}

\section{Missing Proofs of Section~\ref{sec:setfair_approx_optimality}}
\label{app:setfair_approximation}

\begin{numberedlemma}{\ref{lm:make_distant}}
Consider solution $\T: \tau_1< \tau_2< \ldots$ with total improvement $I$ such that for all $j$,  $\tau_{j+2}-\tau_j \geq \Delta$. Consider the procedure in Definition~\ref{def:distant_procedure}. This procedure results in a solution $\T': \tau'_1< \tau'_2< \ldots$ where $\forall j \; \tau'_{j+1}-\tau'_j \geq 2\Delta$, has total improvement at least $ I/4$, and $|\T'| \leq \lceil |\T|/4 \rceil$. Particularly, for $|\T|\leq \lceil k/g \rceil$ where $k \geq g$, the number of final targets, $|\T'|$, is at most $\lfloor k/g \rfloor$.
\end{numberedlemma}

\begin{proof}[Proof of Lemma~\ref{lm:make_distant}]
Since the best out of $4$ parts have been selected, the total improvement at the end of the procedure is at least $1/4$ fraction of $I$.
In addition, in the final set, every pair of consecutive targets are indexed $\tau_{j}$ and $\tau_{j+4}$. Therefore, since originally for all $j$, $\tau_{j+2}-\tau_j \geq \Delta$, we have $\tau_{j+4}-\tau_j \geq 2\Delta$. Finally, since in each set of $\tau_j,\ldots, \tau_{j+4}$ exactly one target is selected, the final number of targets is at most $\lceil |\T|/4 \rceil$.
\end{proof}

\begin{numberedlemma}{\ref{lm:step_three}}
At the end of step $3$ in Algorithm~\ref{alg:approx}, (i) the distance between every two targets in $\T_\ell$ is at least $\Delta$; (ii) each target $\tau \in \T_\ell$ is optimal, i.e., maximizes total improvement for agents in $G_\ell \cap [\tau-\Delta, \tau)$; and (iii) the total amount of improvement of $G_\ell$ using solution $\T_\ell$ does not decrease compared to the previous step.
\end{numberedlemma}

\begin{proof}[Proof of Lemma~\ref{lm:step_three}]
Let $\tau$ be a target at the beginning of step $3$ and $\tau'$ be its replacement at the end of this step. 

We first prove statement (i). First, we argue for agents in $[\tau-\Delta, \tau)$, the optimal target $\tau'$  belongs to $[\tau, \tau+\Delta]$. Intuitively, the reason is that all these agents afford to improve to $\tau$; therefore, a target smaller than $\tau$ is suboptimal. Also, none of the agents affords to improve beyond $\tau+\Delta$. More formally, if $\tau'<\tau$, agents in $[\tau-\Delta, \tau')$ improve less compared to a target at $\tau$ and agents in $[\tau',\tau)$ do not improve. On the other hand, if $\tau'>\tau+\Delta$, none of the agents can reach $\tau'$ and the total improvement for these agents will be $0$. Therefore, at the end of this step, every target $\tau$ is replaced with $\tau' \in [\tau, \tau+\Delta]$. Now, by Observation~\ref{obs:delta_apart} and Lemma~\ref{lm:make_distant}, the distance between consecutive targets at the end of step $2$ is at least $2\Delta$. Therefore, after the modification explained (shifting each target to the right by less than $\Delta$) this distance decreases by at most $\Delta$ and becomes at least $\Delta$. 

Now, we move on to statement (ii). We need to argue if $\tau'$ is optimal for agents in $[\tau-\Delta, \tau)$, it is also optimal for agents in $[\tau'-\Delta, \tau')$. By Lemma~\ref{lm:make_distant}, at the beginning of step $3$, there are no targets in $(\tau, \tau+2\Delta)$; more specifically, there are no targets for agents in $[\tau,\tau+\Delta)$ and these agents get eliminated in this step. Therefore, since $\tau'$ belongs to $[\tau, \tau+\Delta]$, as shown in the proof of statement (i), we only need to argue that if $\tau'$ is optimal for $[\tau-\Delta, \tau)$, it is also optimal for $[\tau'-\Delta, \tau)$. Suppose this was not the case, and there was another target $\tau''$ which was optimal for this set. Since the agents in $[\tau'-\Delta, \tau)$ are the only agents with positive amount of improvement for target $\tau'$, replacing $\tau'$ with $\tau''$ would result in higher improvement for the whole set of agents in $[\tau-\Delta, \tau)$ which is in contradiction with definition of $\tau'$.  

Finally, we argue statement (iii). In step $3$, the agents not improving in step $2$ have been eliminated and the new targets only (weakly) increased the total improvement of the remaining agents. Therefore, the total amount of improvement does not decrease in this step.
\end{proof}

\begin{numberedlemma}{\ref{lm:step_four}}
Consider $\T$ as the union of all solutions at the end of step $3$. For all $\tau \in \T$, consider the interval $[\tau-\Delta, \tau)$ which consists of agents that improve to target $\tau$ if it were the only target available. At the end of step $4$, (i) there will be a target in $[\tau-\Delta+\Delta/g, \tau]$, and (ii) there will be no targets in $(\tau-\Delta, \tau-\Delta+\Delta/g)$.
\end{numberedlemma}
\begin{proof}[Proof of Lemma~\ref{lm:step_four}] Statement (i) is equivalent to (i') for any $s \in S$, there will be a target in $[s+\Delta/g, s+\Delta)$; and statement (ii) is equivalent to (ii') for any $s \in S$, there will be no targets in $(s, s+ \Delta/g)$. We prove (i') and (ii').

We first show the size of each part is at most $g$; i.e. $\forall i, |S_i| \geq g$. The proof is by contradiction. Suppose there exists $|S_i| \geq g+1$. Therefore, there exist $s_j < s_{j'} \in S_i$ and group index $\ell$, such that $s_j+\Delta, s_{j'}+\Delta \in \T_\ell$, and all $s$ satisfying $s_j< s< s_{j'} \in S_i$ corresponding to targets in distinct groups other than $\ell$. Therefore, there are at most $g-1$ such $s$. Hence, $s_{j'}-s_j < g\times \Delta/g = \Delta$, implying there are two targets in $\T_\ell$ at distance strictly less than $\Delta$ which is in contradiction with Lemma~\ref{lm:step_three}.

Now, we prove statement (i''). In step $4$, the final target corresponding to part $S_i : s_u\leq s_{u+1}\leq \ldots \leq s_v$ is defined as $\tau^{\star}_i = \min\{\tau_v, s_{v+1}\}$. By definition, $\tau^{\star}_i \leq s_{v+1}$; therefore, it is (weakly) to the left of any $s_j$ for $j \geq v+1$. Also,  using $|S_i| \leq g$, $s_v < s_u+ (g-1)\Delta/g$, which implies $\tau_u - s_v > \Delta/g$,
and since by definition, $s_{v+1}-s_v \geq \Delta/g$, both $s_{v+1}$ and $\tau_u$ are at least at distance $\Delta/g$ to the right of $s_v$ and any $s_j$ such that $j \leq v$. This proves statement (i'').  

Finally, we prove (i'). In the proof of (ii'), we showed that $\tau^{\star}_i \geq s_v + \Delta/g$ which implies $\tau^{\star}_i \geq s + \Delta/g, \; \forall s \in S_i$. Therefore, it suffices to show $\tau^{\star}_i \leq s_u + \Delta$, which then implies $\tau^{\star}_i \leq s + \Delta, \; \forall s \in S_i$. The definition of $\tau^{\star}_i$ directly implies $\tau^{\star}_i \leq s_u + \Delta$.
\end{proof}

\section{Distance Between Consecutive Target Levels}

Observation~\ref{obs:delta_apart} shows it is without loss of optimality to assume the distance between every other targets is at least $\Delta$ in the common improvement capacity model. The following example investigates this property for \emph{consecutive} targets, and shows an instance where the distance between two consecutive targets is arbitrarily small compared to $\Delta$ in the optimal solution.

\begin{example}\label{ex:consecutive_less_than_delta}
Suppose $\Delta=1$ and there is no limit on the number of targets.  Suppose there is an agent at position $0$, an agent at position $1$, and $m$ agents at position $1+1/m$. The optimal solution is $\T = \{\tau_1=1, \tau_2=1+1/m, \tau_3=2+1/m\}$. As $m \rightarrow \infty$, the distance between $\tau_1$ and $\tau_2$ approaches $0$.
\end{example}

\chapter{Generating Actionable Insights in Large State Spaces to Help People Reverse Unfavorable Decision Outcomes}
\label{app:cfes}
\section{The hl-discrete CFEs vs. hl-continuous CFEs}
\label{sec:cfe_app_hld-hlc-diff}

We further elaborate on the distinction between hl-continuous and hl-discrete CFEs and a potentially interesting future direction.

\paragraph{Differentiation between hl-discrete and hl-continuous CFEs.} 
The key distinction between hl-discrete and hl-continuous CFEs is how they affect features, where the former caters to feature eligibility and the latter numerically modifies feature values.
For instance, consider the feature \textit{healthBMI} in App2, Figure~\ref{fig:main_figure1}.
While hl-discrete CFE would ensure BMI is past a desired threshold, thus ensuring \textit{healthBMI} = 1, the hl-continuous CFE would affect the actual feature values, such as reducing BMI from \(30\) to \(22.3\).

It is important to note that a threshold defining feature eligibility can stem from any classifier type. Continuing with the \textit{healthBMI} example, the binary eligibility (\(\{0,1\}\)) is determined based on whether the feature crosses a predefined threshold of what qualifies as a healthy BMI, as dictated by the classification model’s set threshold.

These two forms of CFEs are suited for different contexts. The hl-discrete CFEs are particularly effective in rule-based systems, such as eligibility checks, wellness evaluations, or quality and safety assessments. In contrast, hl-continuous CFEs are better suited for scenarios where fine-grained numerical adjustments are meaningful, typically in settings where low-level CFEs are effective.

While it might be easier to go from hl-continuous CFEs to hl-discrete CFEs, the reverse is less trivial and may require additional considerations.

\paragraph{Integration of high-level actions into existing CFE generation methods.} 
While it may be feasible to incorporate the hl-continuous or hl-discrete actions into existing CFE generation methods workflows, it's currently unclear which methods are best suited, how readily they can be adapted, or what challenges might emerge. For example, in evolutionary algorithm-based approaches, generating exact CFEs under this new paradigm could be computationally intensive, particularly in large action spaces, potentially leading only to approximate or Pareto-optimal solutions. We hope future research explores how to effectively adapt existing CFE generation methods to this new paradigm and uncover any associated interesting challenges.

\section{Supplemental Datasets Details}
\label{sec:cfe_app_all-datasets}

This section describes the supplemental details about the datasets used in the experiments. We conducted all experiments on a laptop with a CPU featuring the following hardware specifications: a 2.6 GHz 6-Core Intel Core i7 processor, 16 GB of 2400 MHz DDR4 RAM, and an Intel UHD Graphics 630 with 1536 MB of video memory. In all cases where we implement Equations~\ref{eq:act_recourse}, \ref{eq:hl-continuous} and \ref{eq:hl-discrete}, we use the \texttt{CVXPY} Python package \citep{diamond2016cvxpy,agrawal2018rewriting}.

\subsection{Datasets Extraction and Preprocessing}
\label{subsec:cfe_app_realworld_datasets}

First, we describe the extraction and preprocessing of real-world datasets: Foods, BMI, WHR, and BRFSS. Then, we describe the creation of semi-synthetic agent–hl-continuous CFE, agent–hl-discrete CFE, and agent–hl-id CFE datasets.

\paragraph{Foods, Body Mass Index (BMI), and Waist-to-Hip Ratio (WHR) Datasets.} 
\label{subsec:cfe_bmi_whr_preproc}
The extraction and preprocessing of the actions and agent information from the Foods, Body Mass Index (BMI), and Waist-to-Hip Ratio (WHR) Datasets.

\begin{enumerate}[label=\arabic*)]
    \item \textbf{Intersectional nutritional features.} 
    After extracting the datasets for Foods, BMI, and WHR and removing features with missing values in the Foods dataset, we selected an intersectional subset of nutritional value features in the Foods and BMI datasets and the Foods and WHR datasets. This subset consisted of \(20\) features, including:  \textit{`protein (gm)', `carbohydrate (gm)', `dietary fiber (gm)', `calcium (mg)', `iron (mg)', `magnesium (mg)', `phosphorus (mg)', `potassium (mg)', `sodium (mg)', `zinc (mg)', `copper (mg)', `selenium (mcg)', `vitamin C (mg)', `niacin (mg)', `vitamin B6 (mg)', `total folate (mcg)', `vitamin B12 (mcg)', `total saturated fatty acids (gm)', `total monounsaturated fatty acids (gm)'},  and \textit{`total polyunsaturated fatty acids (gm)'}.
    
    \item \textbf{Foods dataset preprocessing.} 
    The Foods dataset from \citet{awram_food_nutritional_values} initially contained \(53\) features. After finding the intersectional subset of nutritional value features and removing datapoints with missing values, the dataset had \(27\) features. These included the following: \textit{`NDB\_No', `Shrt\_Desc',  `GmWt\_1', `GmWt\_Desc1', `GmWt\_2', `GmWt\_Desc2'},  and \textit{`Refuse\_Pct'}, along with the \(20\) nutritional features described above. 
    To add costs to the dataset, we web-scraped the average USD prices and extracted caloric prices for each food item given their name specified in the \textit{`Shrt\_Desc'} feature. Out of \(3901\) food items, we successfully extracted USD prices for \(3871\) food items and caloric prices for \(3125\) food items. Therefore, when using USD prices as costs, there were \(3871\) possible hl-continuous actions, while using caloric prices meant \(3125\) possible hl-continuous actions.
    
    \item \textbf{BMI dataset preprocessing.} 
    The body mass index (BMI) dataset originally had \(57\) features. After removal of features with at least \(20\%\) null values and selecting the above nutritional features, except the feature \textit{`total folate (mcg)'}, we had \(23\) features including: \textit{`gender', `age', `race'}, and \textit{ `body mass index (kg/m**2)'}.
    We selected agents whose age was greater than or equal to \(20\) at the time of surveys. Using the features \textit{`body mass index (kg/m**2)'} and \textit{`age'}, we computed the target label (binary class variable) for each agent as either healthy (\(1\)) BMI or unhealthy (\(0\)) ~\citep{webmd_bmi_calculator}.
    We then removed the feature \textit{`body mass index (kg/m**2)'} and all the duplicates datapoints. 
    At the end of data preprocessing, we did the \(80/20\) train/test data split resulting in \(40734\) data points in the predictive training set and \(10184\) in the predictive testing set.
    
    \item \textbf{WHR dataset preprocessing.} 
    Unlike the BMI dataset, there were fewer datapoints with `waist-to-hip ratio' (WHR) information among the NHANES body measurement surveys (for years 1999 to prepandemic 2020) we scraped. 
    First, we removed all features with at least \(20\%\) null values. 
    Then using the features \textit{`waist circumference (cm)',  `hip circumference (cm)'} and \textit{`gender`}, we created the binary class variable \textit{whr-class} \citep{wikipedia_waist_hip_ratio}, indicating healthy (\(1\)) or unhealthy (\(0\)) WHR.
    After preprocessing, we had \(23\) features, including the \(20\) nutritional features described above and the  demographic features: \textit{`gender', `age'}, and  \textit{`race'}.  
    Lastly, we removed the duplicates and split the dataset \(80/20\), creating \(7296\) data points in the predictive training set and \(1824\) in the predictive testing set.

\end{enumerate}

\paragraph{Behavioral Risk Factor Surveillance System (BRFSS) Dataset.}
\label{subsec:cfe_brfss_preproc}
The initial BRFSS dataset comprised \(253680\)  rows and \(22\) features, each detailing various health and demographic attributes of agents \citep{teboul_diabetes_health_indicators_2024}. 

First, we removed all data points where `\textit{Age}' \(= 1\) denoting an age range of \(18\)-\(24\) because computation a new variable which relied on age being equal to or above \(20\) years, which reduced the dataset to \(247,980\) rows. 
The new variable was called `\textit{HealthBMI},' an adult health BMI classification value  ~\citep{webmd_bmi_calculator} from the feature `\textit{BMI}.'
Next, we transformed the existing features, which were predominantly binary, into new features where the \(1\) represents a desirable condition and \(0\) otherwise. We focused particularly on features we deemed actionable and renamed them to enhance their intuitiveness, specific to satisfiability. For instance, we renamed the feature `\textit{HighBP}', which indicated high blood pressure (0 = no, 1 = yes), to `\textit{LowBP}': \{1 = yes (lowBP), 0 = no (highBP)\}.
Additionally, we removed six features `\textit{CholCheck},' \textit{Diabetes\_012},'  `\textit{Sex},' `\textit{Age},'  `\textit{Education},' and '\textit{Income},'  and remained with \(16\) features. 

These final \(16\) binary features included the following: `\textit{LowBP}':  \{1 = yes (lowBP), 0 = no (highBP)\}, `\textit{LowChol}': \{1 = yes (lowChol), 0 = no (highChol)\}.  The feature `\textit{HealthBMI}': \{1 =yes (healthy), 0 = no (unhealthy),  `\textit{NoSmoke}':  \{1 = yes, 0 = no\},  `\textit{NoStroke}': \{1 = yes, 0 = no\},  `\textit{NoCHD}':  \{1 = yes, 0 = no\}, `\textit{PhysActivity}':   \{1 = yes, 0 = no\}, `\textit{Fruits}':   \{1 = yes, 0 = no\},  `\textit{Veggies}':  \{1 = yes, 0 = no\}, `\textit{LightAlcoholConsump}':  \{1 = yes, 0 = no\}, `\textit{AnyHealthcare}':  \{1 = yes, 0 = no\}, `\textit{DocbcCost}':  \{1 = yes, 0 = no\},  `\textit{GoodGenHlth}': \{1 = excellent (1,2,3), 0 = bad (4,5)\}, `\textit{GoodMentHlth}': \{1 = \{1 = good (\(<2\)), 0 = bad (\(\geq2\))\}, `\textit{GoodPhysHlth}': \{1 = good (\(<2\)), 0 = bad (\(\geq2\))\}, and `\textit{NoDiffWalk}': \{1 = yes, 0 = no\}.

Since we consider the setting where \(\mathbf{t} = \mathbf{1}_{16}\), of the remaining data points, \(8392\) were considered to have a desirable outcome (no health risk) because all their features met the respective feature thresholds. 
Lastly, after removing the duplicate health risk agents and splitting the whole dataset \(80/20\), we had \(11039\) data points in the predictive training set and \(2760\) in the predictive testing set.

\subsection{Single-Agent CFE Generation and Semi-Synthetic Agent–CFE Datasets}
\label{subsec:cfe_app_single_cfe}

For all datasets, to determine which agents require CFEs (negatively classified agents), we use the classification models. Specifically, we trained logistic regression models on the BMI and WHR training sets, tuning the \textit{solver} and \textit{max\_iter} hyperparameters using \emph{GridSearchCV}. The best-performing models achieved test accuracies of \(72.78\%\) on the BMI dataset and \(85.18\%\) on the WHR dataset. For the BFRSS dataset, the threshold classifier \(\mathbf{t} = \mathbf{1}_{16}\) achieved \(100\%\) test accuracy. We then used these models to determine the classifier parameters needed for single-agent CFE generation (see Equations~\ref{eq:act_recourse}, \ref{eq:hl-continuous}, and \ref{eq:hl-discrete}) and the specific agents requiring CFEs in the BMI, WHR, and BRFSS train/test datasets.
Below are details on the actionable features for each of the datasets. Refer to Appendix~\ref{subsec:cfe_bmi_whr_preproc} and Appendix~\ref{subsec:cfe_brfss_preproc} for a detailed description of the meaning of the features. 

\begin{enumerate}[label=\arabic*)]

    \item \textbf{BMI actionable features.} 
    For BMI agents states, we considered the following \(\mathbf{19}\) actionable features: `\textit{protein (gm)}', `\textit{carbohydrate (gm)}', `\textit{dietary fiber (gm)}', `\textit{calcium (mg)}', `\textit{iron (mg)}', `\textit{magnesium (mg)}', `\textit{phosphorus (mg)}', `\textit{potassium (mg)}', `\textit{sodium (mg)}', `\textit{zinc (mg)}', `\textit{copper (mg)}', `\textit{selenium (mcg)}', `\textit{vitamin C (mg)}', `\textit{niacin (mg)}', `\textit{vitamin B6 (mg)}', `\textit{vitamin B12 (mcg)}',  `\textit{total saturated fatty acids (gm)}', `\textit{total monounsaturated fatty acids (gm)}', and '\textit{total polyunsaturated fatty acids (gm)}'.
    
    \item \textbf{WHR actionable features.} 
    For the generation of recourse for WHR agents, we use the following \(\mathbf{20}\) actionable features: `\textit{protein (gm)}', `\textit{carbohydrate (gm)}', `\textit{dietary fiber (gm)}', `\textit{calcium (mg)}', `\textit{iron (mg)}', `\textit{magnesium (mg)}', `\textit{phosphorus (mg)}', `\textit{potassium (mg)}', `\textit{sodium (mg)}', `\textit{zinc (mg)}', `\textit{copper (mg)}', `\textit{selenium (mcg)}', `\textit{vitamin C (mg)}', `\textit{niacin (mg)}', `\textit{vitamin B6 (mg)}', `\textit{total folate (mcg)}', `\textit{vitamin B12 (mcg)}',  `\textit{total saturated fatty acids (gm)}', `\textit{total monounsaturated fatty acids (gm)}', and '\textit{total polyunsaturated fatty acids (gm)}'.
    
    \item \textbf{BRFSS actionable features.} 
    Lastly, for the BRFSS agent states, we considered the following \(\mathbf{16}\)  actionable features: `\textit{PhysActivity}', `\textit{Fruits}', `\textit{Veggies}', `\textit{AnyHealthcare}', `\textit{LowBP}', `\textit{NoSmoke}', `\textit{LowChol}', `\textit{HealthBMI}', `\textit{NoStroke}', `\textit{NoCHD}', `\textit{LightAlcoholConsump}', `\textit{DocbcCost}', `\textit{GoodGenHlth}', `\textit{GoodMentHlth}', `\textit{GoodPhysHlth'}, and `\textit{NoDiffWalk}'.

\end{enumerate}

\begin{figure}[b!]
\begin{center}
    \begin{subfigure}[b]{0.66\textwidth}  
    \centering
            \includegraphics[width=0.66\textwidth]{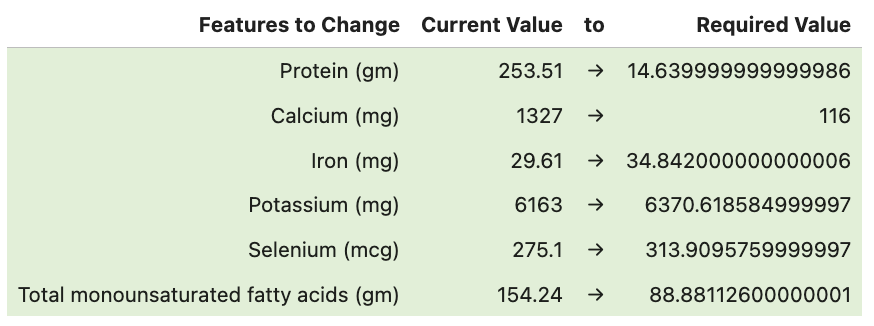}
            \caption{for a BMI agent state}
            \label{fig:bmi_ar}
    \end{subfigure}%
    \vskip\baselineskip
    \begin{subfigure}[b]{0.58\textwidth}            
            \includegraphics[width=0.97\textwidth]{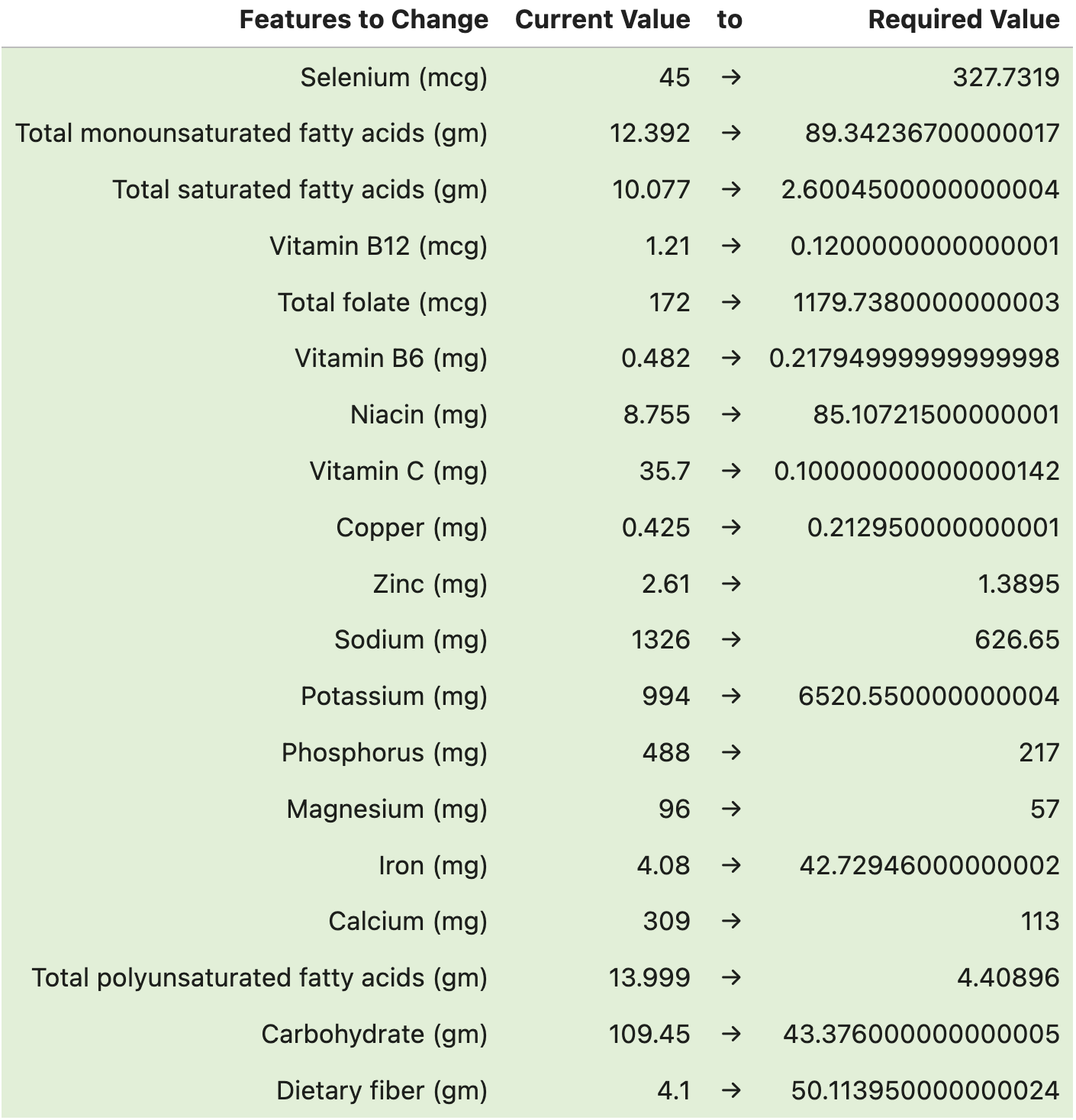}
            \caption{for a WHR agent state}
            \label{fig:whr_ar}
    \end{subfigure}
    \hfill
    \begin{subfigure}[b]{0.4\textwidth}
            \centering
            \includegraphics[width=0.97\textwidth]{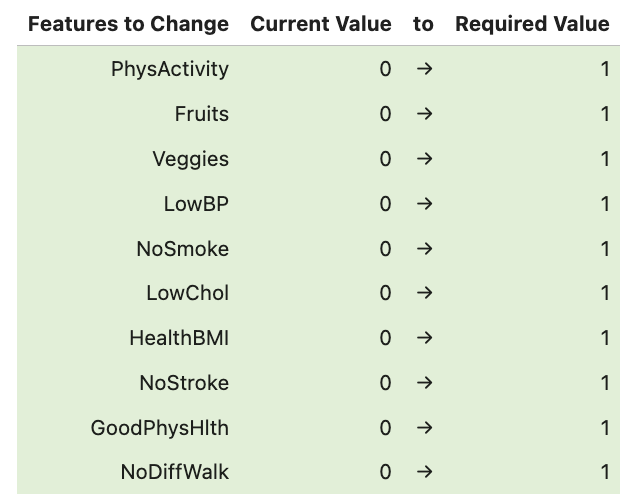}
            \caption{ for a BRFSS agent state}
            \label{fig:brfss_ar}
    \end{subfigure}
\end{center}
\caption[Examples of low-level CFEs on BMI, WHR and BRFSS datasets]
{Given a negatively classified BMI agent with actionable features in the order specified in Appendix~\ref{subsec:cfe_app_single_cfe} and values \([253.51, 352.76, 48.2, 1327., 29.61, 1204., 3966.,\)
\(6163.,5890.0, 44.19,7.903, 275.1, 30.,109.198, 3.492,2.3, 59.686, 154.24, 113.429]\), the low-level CFE generator (cf. Equation~\ref{eq:act_recourse}) generates CFE \subref{fig:bmi_ar} to help them become positive. On the other hand, given a negatively classified WHR agent with actionable features  \([29.03, 109.45, 4.1, 309., 4.08, 96.,488., 994., 1326., 2.61, 0.425,45.,\) \(35.7, 8.755, 0.482, 172.,1.21, 10.077, 12.392, 13.999]\)  in the order as described in Appendix~\ref{subsec:cfe_app_single_cfe}, the low-level CFE generator generates CFE \subref{fig:whr_ar}. Lastly, the low-level CFE generator generates CFE \subref{fig:brfss_ar} or an agent negatively classified based on their BRFSS features, with values \([0, 0, 0, 1, 0, 0, 0, 0, 0, 1, 1, 1, 1, 1, 0, 0]\). All the low-level CFEs (\subref{fig:bmi_ar}, \subref{fig:whr_ar} and \subref{fig:brfss_ar}) are feature-based and precisely describe which features to change and by how much.
}\label{fig:ex_recourse} 
\end{figure}

\paragraph{Low-level CFE data generation.}
\label{subsec:cfe_app_lowlevel_cfe}
Given negatively classified agents in the training and testing BMI, WHR and BRFSS datasets, and the actionable features for the corresponding datasets, we generate low-level CFEs using the low-level CFE generator (actionable recourse) \citep{Ustun19} described in Equation~\ref{eq:act_recourse}. Figure~\ref{fig:ex_recourse} illustrates examples of the generated low-level CFEs for the BMI, WHR and BRFSS datasets.

\begin{figure}[ht!]
\begin{center}
    \begin{subfigure}[]{0.77\textwidth}  
    \centering
            \includegraphics[width=0.44\textwidth]{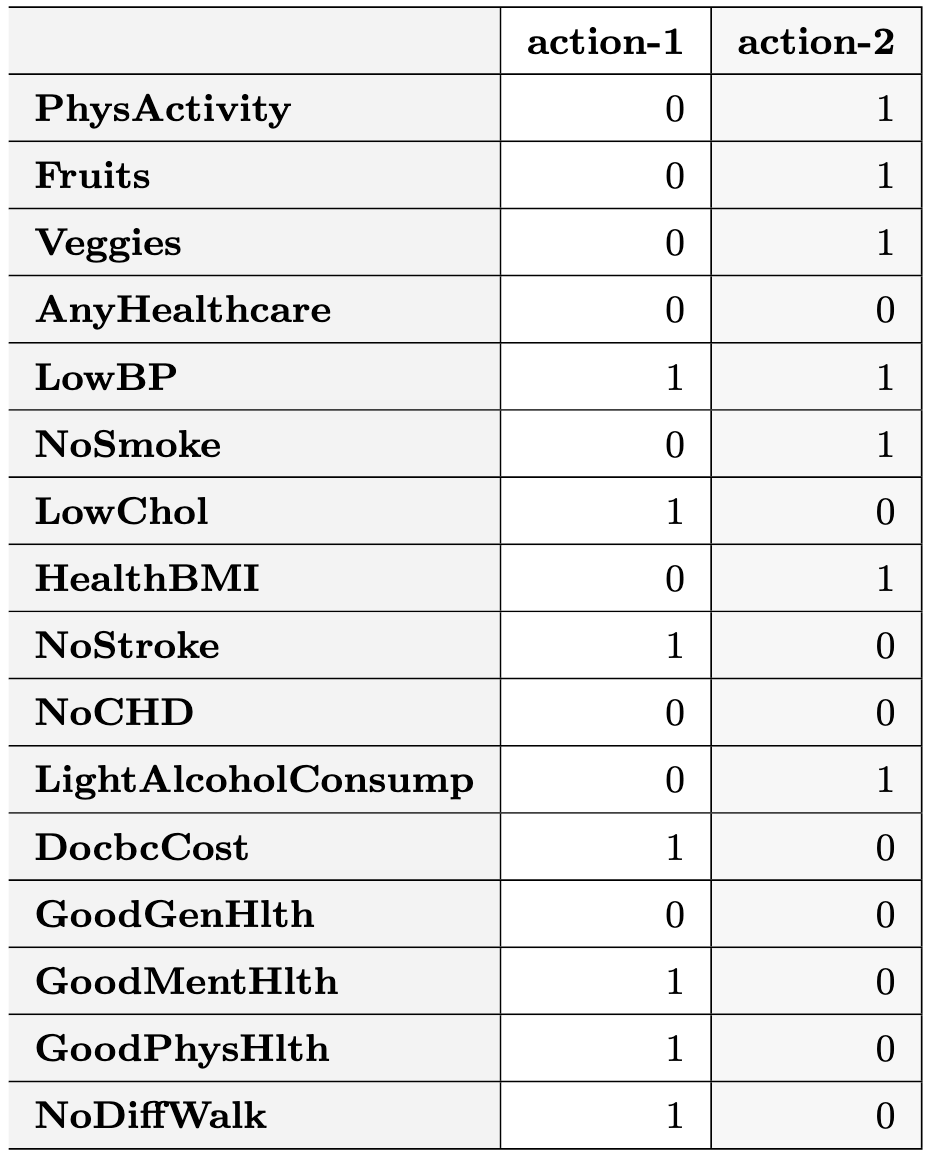}
            \caption{for a BRFSS agent state}
            \label{fig:app_brfss_da}
    \end{subfigure}%
    \vskip\baselineskip
    \begin{subfigure}[]{0.5\textwidth}            
            \includegraphics[width=0.94\textwidth]{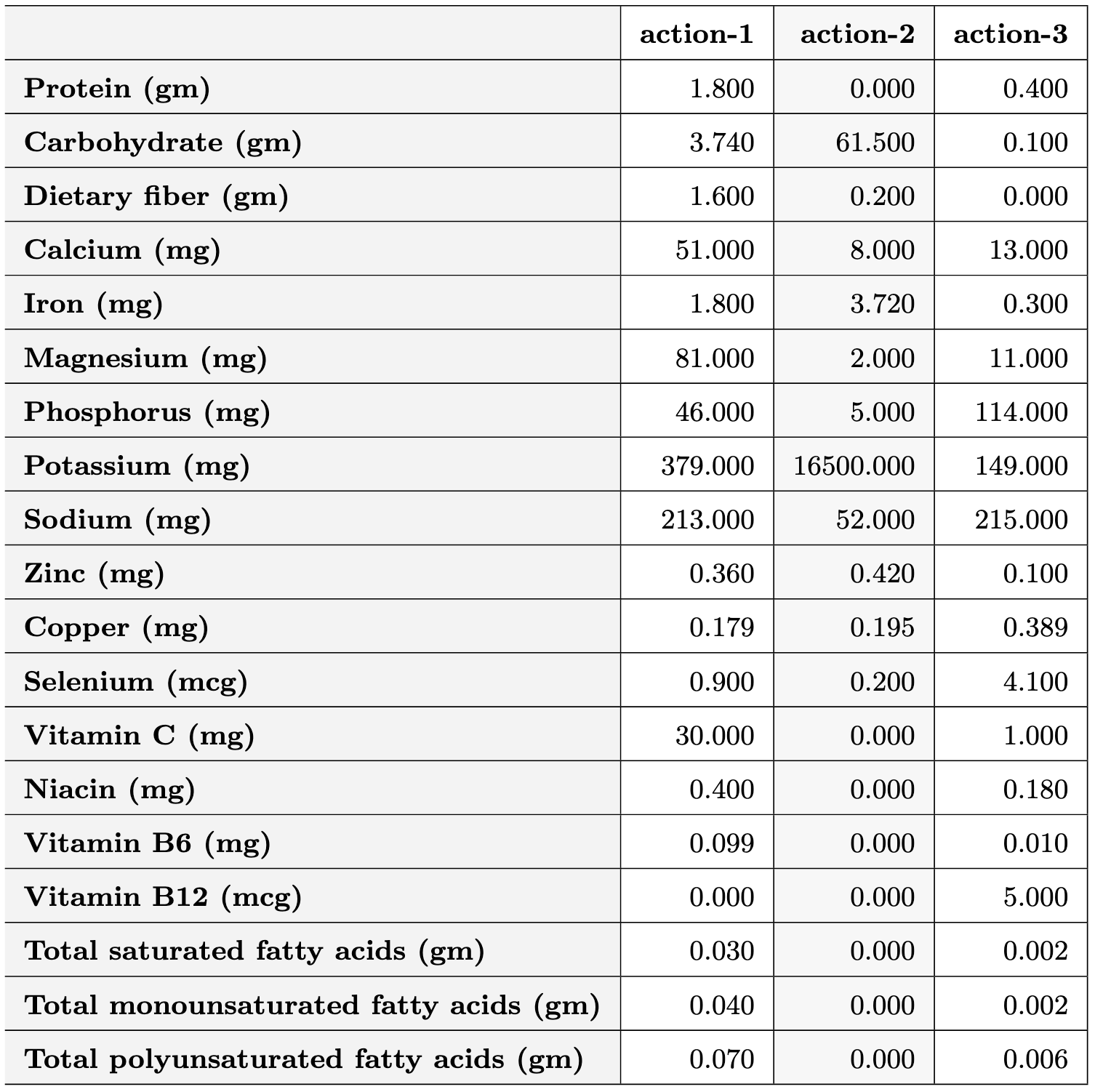}
            \caption{for a BMI agent state}
            \label{fig:app_bmi_fa}
    \end{subfigure}
    \hfill
    \begin{subfigure}[]{0.42\textwidth}
            \centering
            \includegraphics[width=.915\textwidth]{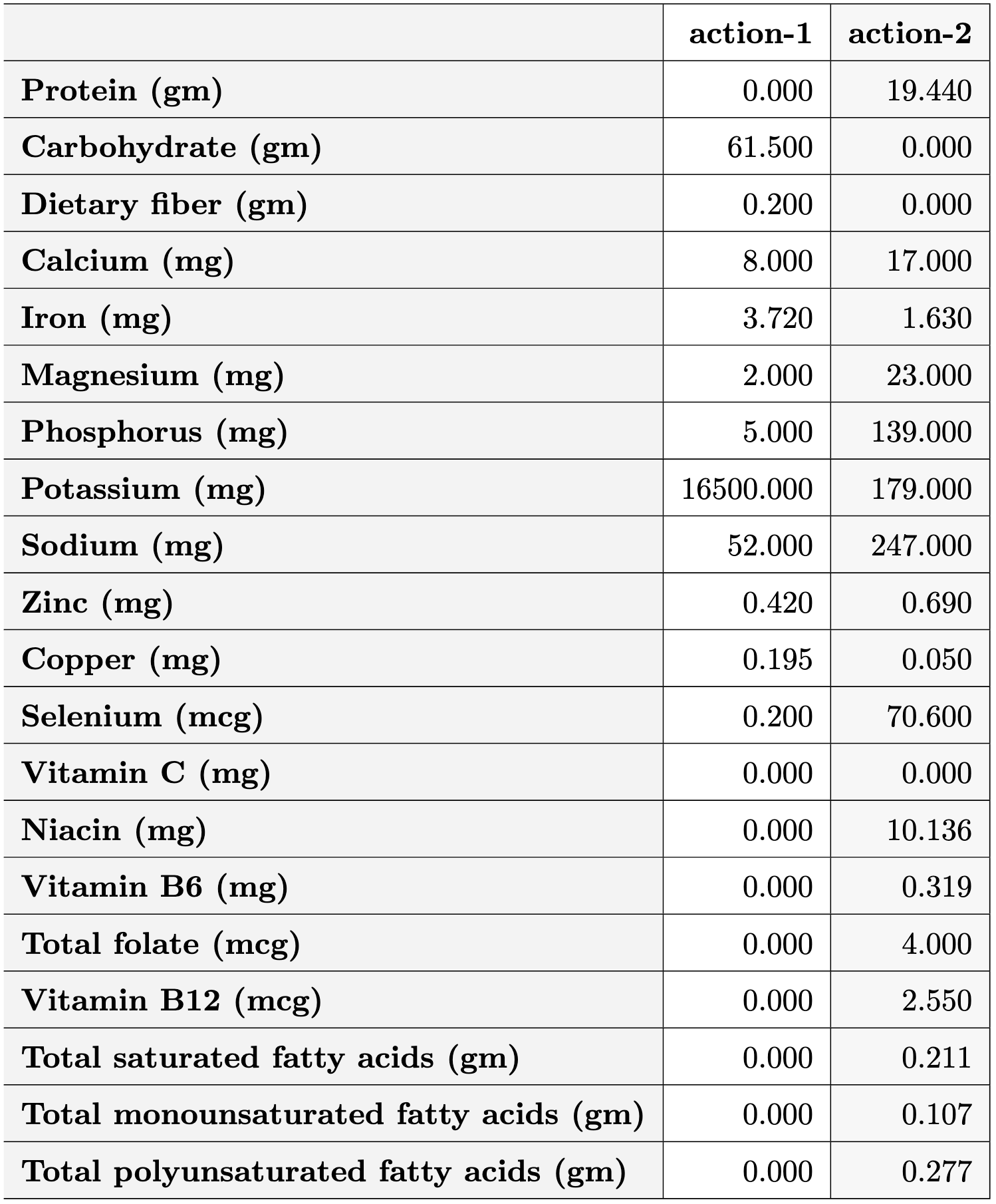}
            \caption{ for a WHR agent state}
            \label{fig:app_whr_fa}
    \end{subfigure}
\end{center}
\caption[Examples of hl-continuous and hl-discrete CFEs]
{For an agent negatively classified based on their BRFSS features, with values \([0, 0, 0, 1, 0, 0, 0, 0, 0, 1, 1, 1, 1, 1, 0, 0]\) in order similar to \subref{fig:app_brfss_da}, the hl-discrete CFE generator recommends hl-discrete CFE \subref{fig:app_brfss_da} with hl-discrete actions, \textbf{action-1} and \textbf{action-2}. 
Additionally, for a negatively classified BMI agent, given their actionable features with values
\([253.51, 352.76, 48.2, 1327., 29.61, 1204., 3966., 6163., 5890.0,\) \(44.19,7.903, 275.1, 30., 109.198, 3.492, 2.3, 59.686, 154.24, 113.429]\), arranged in the same order as features shown in \subref{fig:app_bmi_fa}, the hl-continuous CFE generator recommends CFE \subref{fig:app_bmi_fa} containing the following hl-continuous actions: \textbf{action-1}: \textit{take Swiss chard, raw}, \textbf{action-2}: \textit{take leavening agents: cream of tartar}), and \textbf{action-3}: \textit{take clams, mixed species, canned, in liquid}. Similarly, for a negatively classified WHR agent with actionable feature values ordered as features in \subref{fig:app_whr_fa} \([29.03, 109.45, 4.1,\) \(309., 4.08, 96., 488., 994., 1326., 2.61, 0.425, 45., 35.7, 8.755, 0.482, 172., 1.21, 10.077,\) \(12.392, 13.999]\) the hl-continuous CFE generator recommends CFE \subref{fig:app_whr_fa} with the following hl-continuous actions: \textbf{action-1}: \textit{take leavening agents: cream of tartar}  and \textbf{action-2}: \textit{take fish, tuna, light, canned in water, drained solids}.  
}\label{fig:ex_fa_da}
\end{figure}

\paragraph{Generation of agent–hl-continuous CFE datasets.}
\label{subsec:cfe_app_hlc_cfe}

Here, we describe the single-agent hl-continuous CFE generation and the creation of the four semi-synthetic, agent–hl-continuous CFE datasets from the train/test BMI and WHR datasets and the \(2\) forms of hl-continuous actions: Food+monetary costs and Food+caloric costs actions. Figure~\ref{fig:ex_fa_da} shows examples of the generated hl-continuous CFEs for BMI and WHR agents.

For each negatively classified BMI and WHR train/test set agent, we generated hl-continuous CFEs using two types of actions: Foods+monetary costs and Foods+caloric costs. Leveraging the respective classifier parameters and the ILP formulation (Equation~\ref{eq:hl-continuous}), we computed two distinct CFEs for each agent, one optimized for monetary cost and the other for caloric cost, each specifying an optimal set of food items.

As a result, we produced four unique agent–hl-continuous CFE datasets. We generated two CFEs for each training/testing BMI agent, one for each form of hl-continuous actions, yielding \(40692\) agent–CFE pairs for training and \(10167\) pairs for testing in each case. Similarly, for the WHR dataset, we generated \(6387\) training and \(1603\) testing agent–CFE pairs for both Foods+monetary and Foods+caloric costs.

\paragraph{Generation of agent–hl-discrete CFE datasets.}
\label{subsec:cfe_app_hld_cfe}

We outline the single-agent hl-discrete CFE generation and the generation of agent–hl-discrete CFE dataset, which consists of negatively classified BRFSS train/test agents and their corresponding optimal synthetic hl-discrete CFEs. Figure~\ref{fig:ex_fa_da} shows an example of the generated hl-discrete CFE for a BRFSS agent.

First, we generated \(100\) synthetic \(16\)-dimensional, binary hl-discrete actions. The probability \(p_{a}\) of an action satisfying feature eligibility was set to  \(0.5\). The cost of satisfying feature eligibility was randomly predefined and remained uniform across all actions and agents. The total cost of an action was the sum of the costs associated with satisfying each feature’s eligibility.

Next, we employed a threshold classifier with \(\mathbf{t} = \mathbf{1}_{16}\) where an agent \(\mathbf{x}\) is classified as not at health risk if \(x_{i} \geq t_{i}, \ \forall i \in [n]\), and as a health risk otherwise. This classifier achieved perfect test accuracy \(100.00\%\) on the BRFSS dataset, allowing us to accurately identify agents requiring CFEs in the training and testing datasets.

Using the identified agents, the synthetic hl-discrete actions, and the threshold classifier \(\mathbf{t} = \mathbf{1}_n\), we applied Equation~\ref{eq:hl-discrete} to generate the agent–hl-discrete CFE dataset. As a result, we obtained \(11,039\) agent–CFE pairs in the training dataset and \(2,760\) in the testing agent–hl-discrete CFE dataset, where each CFE comprised optimal hl-discrete synthetic actions.

\paragraph{Generation of agent–hl-id CFE datasets and other variants.}
\label{subsec:cfe_app_hlid_var_cfe} 

After creating the agent–hl-continuous CFE, agent–hl-discrete CFE, training/testing datasets, we generate other agent–CFE dataset variants from them.

\begin{enumerate}[label=\arabic*)]

    \item \textbf{The agent–hl-id CFE datasets.} 
    Given the agent–hl-continuous CFE and agent–hl-discrete CFE datasets described in Sections~\ref{subsec:cfe_app_hlc_cfe} and \ref{subsec:cfe_app_hld_cfe}, we created corresponding agent–hl-id CFE datasets.  This process involves encoding each CFE in the agent–CFE dataset with a unique identifier that distinguishes it from all other possible CFEs in that dataset. 
    For example, given instances of agents--hl-discrete CFEs, we generate unique identifiers for all the hl-discrete CFEs to generate corresponding hl-id CFEs. At the end, we had 5 agent–hl-id CFE training/testing datasets.

    \item \textbf{The semi-synthetic varied frequency of CFEs agent–CFE datasets.} 
    For each of the generated agent–hl-continuous CFE, agent–hl-discrete CFE, and the agent–hl-id CFE training/testing datasets described above, we generate three frequency of CFE dataset variants:  \texttt{all} (including all data), \texttt{>10} (more than \(10\) agents per CFE), and \texttt{>40} (more than \(40\) agents per CFE).
    
\end{enumerate}

\subsection{The Fully-synthetic Agent–CFE Datasets}
\label{subsec:cfe_app_fullysynthetic_datasets}

\begin{figure}[b!]
    \centering
    \begin{subfigure}[t]{0.48\textwidth}
        \centering
        \includegraphics[width=\textwidth]{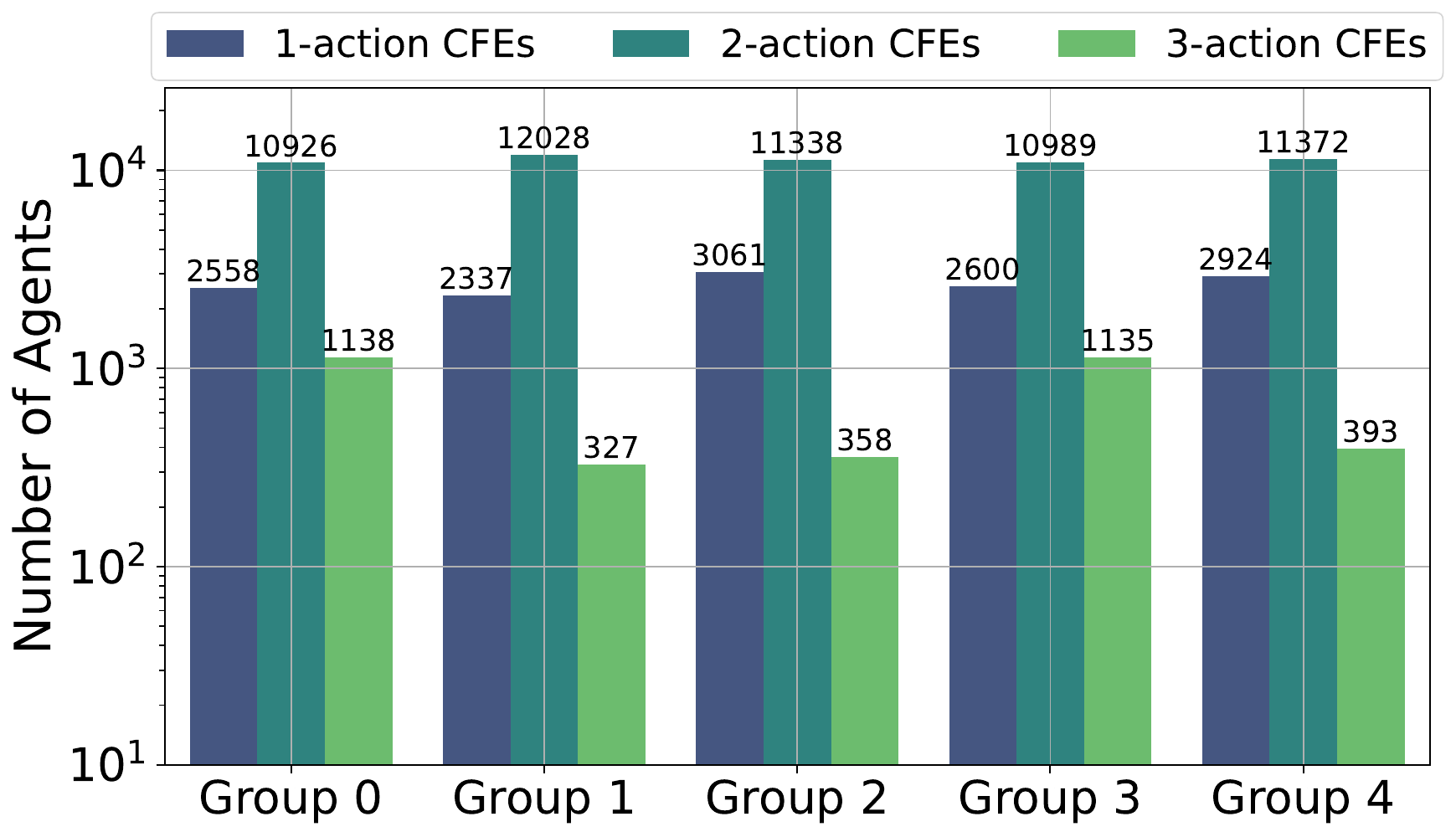}
        \caption{\texttt{manual groups} \label{fig:grp5_20}}
    \end{subfigure}%
    \hfill
    \begin{subfigure}[t]{0.48\textwidth}
        \centering
        \includegraphics[width=\textwidth]{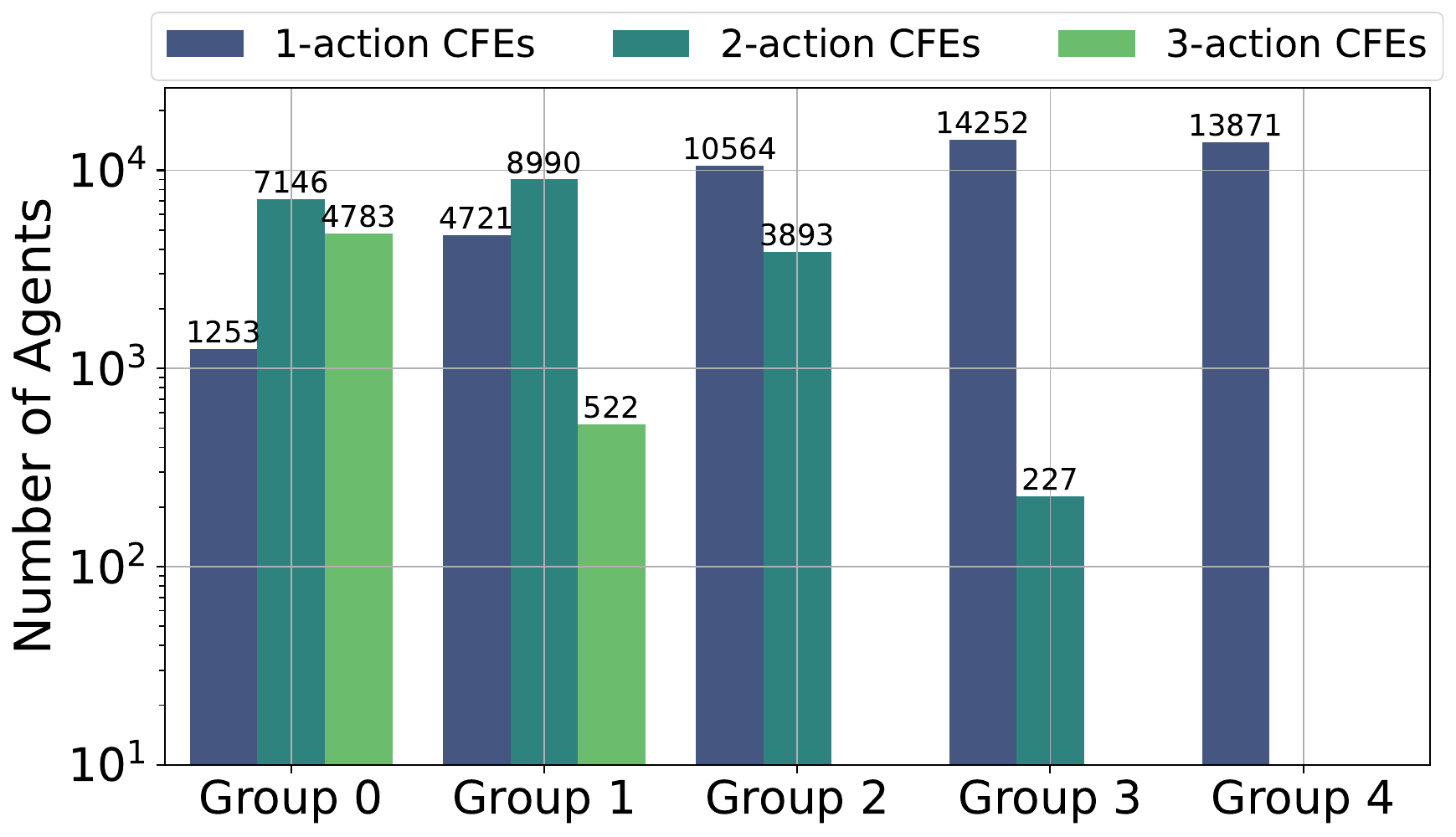}
        \caption{\texttt{probabilistic groups} \label{fig:grp5_acts20}}
    \end{subfigure}
    \caption[Statistics on the varied access to actions agent–hl-discrete CFE datasets]
    {Statistics on the varied access to actions agent–hl-discrete CFE datasets for \texttt{manual groups} and \texttt{probabilistic groups}. In the \texttt{probabilistic groups}, the action probability \(p_a\) varies as follows:  Group \(0\) (\(p_a = 0.4\)), Group \(1\) (\(p_a = 0.5\)), Group \(2\) (\(p_a =0.6\)), Group \(3\) (\(p_a = 0.7\)), and Group \(4\) (\(p_a = 0.8\)).  While the \texttt{manual groups} exhibit a more balanced distribution in terms of the number of actions taken by agents, the \texttt{probabilistic groups} introduce disparities where agents in certain groups have access only to more expensive and limited hl-discrete actions compared to others. 
    }\label{fig:5grps20}
\end{figure}

We created five kinds of fully-synthetic agent–hl-discrete CFE datasets: varied dimension, frequency of CFEs, information access, feature satisfiability, and actions access. 
We provide statistical detailed information about the five variations of the agent–hl-discrete CFE datasets in Table~\ref{table:datasets_statistics} and Figure~\ref{fig:5grps20}.
\begin{table}[b!]
\centering
\footnotesize
\renewcommand{\arraystretch}{1.0} 
\setlength{\tabcolsep}{3pt} 
\begin{tabular}{lccccc}
    \toprule
    \textbf{Dataset name} & \textbf{Dataset size} & \textbf{One-action CFEs} & \textbf{Two-action CFEs} & \textbf{Three-action CFEs}\\
    \midrule
    \(20\)-dimensional dataset & \(71125\) & \(23687\) & \(44858\) & \(\hphantom{0}2576\) \\
    \(50\)-dimensional dataset & \(98966\) & \(\hphantom{0}1262\) & \(96770\) & \(\hphantom{00}934\) \\
    \(100\)-dimensional dataset & \(99728\) & \(\hphantom{0000}0\) & \(45515\) & \(54213\)\\
    \texttt{manual groups} & \(73484\) & \(13480\) & \(56653\) & \(\hphantom{0}3351\)\\
    \texttt{probabilistic groups} & \(70226\) & \(44661\) & \(20258\) & \(\hphantom{0}5307\)\\
    \texttt{First10} & \(74524\) & \(61794\) & \(12046\) & \(\hphantom{000}39\) \\
    \texttt{First5} & \(74594\) & \(60656\) & \(\hphantom{0}6005\) & \(\hphantom{0000}0\)\\
    \texttt{Last10} & \(74401\) & \(53822\) & \(19952\) & \(\hphantom{0000}1\)\\
    \texttt{Last5} & \(74565\) & \(66068\) & \(\hphantom{00}644\) & \(\hphantom{0000}0\)\\
    \texttt{Mid5} & \(74594\) & \(63530\) & \(\hphantom{0}3010\) & \(\hphantom{0000}0\)\\
    \bottomrule
\end{tabular}
\caption[Statistics of some of the fully-synthetic agent–hl-discrete CFE datasets]
{Statistics of the fully-synthetic agent–hl-discrete CFE datasets used in the experiments. Each hl-discrete CFE for each agent in all datasets has atmost \(3\) hl-discrete actions. 
}\label{table:datasets_statistics}
\end{table}

\paragraph{Varied dimensions agent–CFE datasets.} 
\label{subsubsec:cfe_app_var_dimensions_ds} 

We created \(20\)-, \(50\)- and \(100\)-dimensional agent states datasets by varying the number of actionable features (\(n = 20, 50, 100\)) and keeping \(p_{f} = 0.68\) the same for all datasets. We consider a unit vector threshold of length \(n\). The cost associated with satisfying a feature's eligibility was predefined randomly and the same across all actions and agents. Each action was of length \(n\), \(p_{a}\) was \(0.5\), and action cost was the sum of the cost for each features the action fulfills. 
To create the \(20\)-, \(50\)- and \(100\)-dimensional agent–hl-discrete CFE datasets, we computed the hl-discrete CFEs for each varied dimensional agent states datasets using the information above and the ILP defined in Equation~\ref{eq:hl-discrete}.

\paragraph{Varied frequency of CFEs agent–CFE datasets.}
\label{subsubsec:cfe_app_var_fre_ds} 

To investigate the effect of frequency of CFEs in the agent–CFE training set on the performance of the data-driven CFE generator, we create three varied frequency of CFEs agent–CFE datasets. 
For each of the varied dimensions agent–hl-discrete CFE datasets described in Appendix~\ref{subsubsec:cfe_app_var_dimensions_ds}, before the train/test split, we created three frequency-based agent–CFE datasets: \texttt{all}, where all data is included, \texttt{>10}, where we ensure a frequency of more than \(10\) agents per hl-discrete CFE, and \texttt{>40} with insurance of a frequency of more than \(40\) agents per hl-discrete CFE.

\paragraph{Varied information access agent–CFE datasets.} 
\label{subsubsec:cfe_app_var_info_ds} 

We construct varied information access agent–CFE datasets for synthetically generated agents and their corresponding fully-synthetic hl-discrete CFEs. That is, for each of the \(20\)-, \(50\)- and \(100\)-dimensional agent–hl-discrete CFE datasets and their corresponding frequency-based datasets (\texttt{all}, \texttt{>10}, and \texttt{>40}), we created three varied information access datasets: agent–hl-discrete CFE dataset where the original hl-discrete CFE remains unchanged, the agent–hl-discrete-named CFE dataset where a unique name encodes each hl-discrete action in the hl-discrete CFE, and the agent–hl-discrete-id CFE dataset where a unique identifier denotes the entire hl-discrete CFE. 
For example, consider an agent \(\mathbf{x} = [0, 0, 0, 0, 1]\) and their corresponding hl-discrete CFE given by \(\{[0, 0, 1, 1, 0], [0, 1, 0, 0, 0], [1, 0, 0, 0, 0] \}\).  The hl-discrete-named CFE \(\{a, b, c \}\) where each hl-discrete action has a name (e.g., \(a\)) that uniquely identifies a specific hl-discrete action (e.g., \([0, 0, 1, 1, 0]\)) among all hl-discrete actions. On the other hand, a unique name, say \(z\), denotes the hl-discrete-id CFE, where \(z\) uniquely represents this specific hl-discrete CFE among all the hl-discrete CFEs.

This setting aims to study the effectiveness of the data-driven CFE generators under various information access constraints within an agent–CFE training set, for example, (1) full access to hl-discrete actions and their effects on features (hl-discrete CFE), (2) access only to the names of hl-discrete actions without any information on how each action affects features (hl-discrete-named CFE), and (3) minimal information access, where only hl-discrete-id CFEs are known, with no explicit knowledge of the corresponding hl-discrete actions or their impact on features.

Given the agent–hl-discrete CFE varied information access datasets, we use the data-driven hl-continuous CFE generator to generate hl-discrete CFEs, data-driven hl-continuous CFE generator for hl-discrete-named CFEs, and data-driven hl-id CFE generators for hl-discrete-id CFEs. 

\paragraph{Varied feature satisfiability agent–CFE datasets.} 
\label{subsubsec:cfe_app_var_classifier_ds} 

Using the ILP formulation defined in Equation~\ref{eq:hl-discrete} with \(n=20\), and following the same agent and hl-discrete generation approach as in Appendix~\ref{subsubsec:cfe_app_var_dimensions_ds} while varying the feature satisfiability for the threshold-based binary classifier (differing in which features are classifier-active (non-zero)), we generated five agent–hl-discrete CFE datasets. 
For the dataset \texttt{Last5}, the threshold vector is set as \(\mathbf{t} = [\text{15 zeros}, \text{5 ones}]\), while for the dataset \texttt{First5}, it is set as \(\mathbf{t} = [\text{5 ones}, \text{15 zeros}]\). The third dataset, \texttt{First10}, has a threshold vector of \(\mathbf{t} = [\text{10 ones}, \text{5 zeros}]\), and the dataset \texttt{Last10} has \(\mathbf{t} = [\text{10 zeros}, \text{10 ones}]\). Finally, the dataset \texttt{Mid5} has all features set to $0$ except for the $5$ middle features set to $1$.

These varied feature satisfiability  agent–hl-discrete CFE datasets are specifically created to investigate the effect of feature satisfiability on the nature of the hl-discrete CFEs and the effectiveness of the data-driven hl-continuous CFE generator at generating CFEs for new agents.

\paragraph{Varied access to actions agent–CFE datasets.} 
\label{subsubsec:cfe_app_var_actions_ds} 

Lastly, we consider two settings where grouped agents have restricted access to a set of actions: 1) \texttt{manual groups} where actions generated with the same probability \(p_{a} = 0.5\) and agents are randomly assigned a restricted subset of actions; and 2) \texttt{probabilistic groups} where agents are assigned to groups and each group has its actions generated by different probabilities \(p_{a} = [0.4, 0.5, 0.6, 0.7, 0.8].\) See Figure~\ref{fig:5grps20} for the statistics of the datasets.

We designed the varied access to actions agent–hl-discrete CFE datasets to empirically investigate fairness in CFE generation. Specifically, we examine the impact of restricting access of a group of agents to some actions on the nature of hl-discrete CFEs, such as CFE costs and the variations in accuracy of data-driven hl-continuous CFE generators across different groups.

\section{Supplemental Details on Data-Driven CFE Generators }
\label{sec:cfe_app_archictures}

This section includes supplemental details about the architectures of the data-driven CFE generators, details about other baseline models, and future works. 
Although we do not explicitly create a separate validation set during the initial \(80/20\) data split for training and testing, we use ``\emph{validation\_split}'' when training all the generator models.

\subsection{The Data-Driven hl-continuous CFE Generator}

The neural-network hl-continuous CFE generator we use in these experiments is susceptible to imbalance and overfitting. Therefore, we weight and regularize the loss function \(\mathcal{L}_\textrm{HC}\) in Equation~\ref{eq:hl-continuous-loss} as follows:
\begin{equation}
    \mathcal{L}_\textrm{HC}^{w} =   p_w \mathcal{L}_\textrm{HC}  + \alpha \frac{1}{M} \sum_{m=1}^{M} || \hat{a}_m - a_m||_{1}
    \label{eq:app_regularize_fa}
\end{equation}

The weighting factor \(p_w\) weights \(\mathcal{L}_\textrm{HC}\) by scaling the contribution of each agent to the loss function. 
The term \(\alpha \frac{1}{M} \sum_{m=1}^{M} || \hat{a}_m - a_m||_{1}\) regularizes the model, thus preventing overfitting by nudging the model towards producing hl-continuous CFEs closer to \(a_m\)'s distribution. 
We, on average chose the values of \(\alpha\) from the set \(\{0.05, 0.1, 0.07\}\) and \(p_w\) from \(\{0.05, 0.1, 0.07\}\).

\begin{figure}[b!]
\centering
\begin{tikzpicture}[scale=0.55]
  \def\radius{0.4}
  \def\vspacing{1}
  \def\vspacingLine{1.0} 
  \def\numRows{1}
  \def\numCols{5}
  
  \def\numColsr{5}
  \def\numRowsr{2}
  \def\cellSize{0.7}

\begin{scope}[shift={(0cm,-4cm)}]
  \foreach \i in {0,...,\numRows} {
    \draw (0, -\i*\cellSize) -- (\numCols*\cellSize, -\i*\cellSize);
  }
\end{scope}

\begin{scope}[shift={(0cm,-4cm)},xshift=-\cellSize]
  \foreach \j in {0,...,\numCols} {
    \draw (\j*\cellSize, 0) -- (\j*\cellSize, -\numRows*\cellSize);
  }
\end{scope}
\node[below] at (1.8cm, -5.2cm) {Input: \(\mathbf{x} \in \{0,1\}^{n}\)};

\begin{scope}[shift={(8cm,-3.5cm)}]
  \foreach \i in {0,...,\numRowsr} {
    \draw (0, -\i*\cellSize) -- (\numColsr*\cellSize, -\i*\cellSize);
  }
\end{scope}

\begin{scope}[shift={(8cm,-3.5cm)},xshift=-\cellSize]
  \foreach \j in {0,...,\numColsr} {
    \draw (\j*\cellSize, 0) -- (\j*\cellSize, -\numRowsr*\cellSize);
  }
\end{scope}
\node[below] at (9.8cm, -5.2cm) {Output: \(\hat{I} \in \{0,1\}^{s\times n}\)};

\draw[-stealth, line width=0.1cm, black] (6.2, 1) -- (7.2, 1) node[midway, above, text=black] {};
\draw[-stealth, line width=0.1cm, black] (3.7, 1) -- (4.7, 1) node[midway, above, text=black] {};

\draw[-stealth, line width=0.1cm, black] (9.7, -1.6) -- (9.7, -3.4) node[midway, right, text=black] {};
\draw[-stealth, line width=0.1cm, black] (1.7, -3.8) -- (1.7, -1.6) node[midway, right, text=black] {};

  \foreach \x in {1,2,3,4,5,6}
    \filldraw[fill=grey, draw=darkblue, line width=0.03cm] ({0}, {-3*\radius + (\x-1)*\vspacing}) circle (\radius);
    
  \foreach \x in {1,2,3,4,5}
    \filldraw[fill=grey, draw=darkblue, line width=0.03cm] ({1.5}, {-2*\radius + (\x-1)*\vspacing}) circle (\radius);

  \foreach \x in {1,2,3,4}
    \filldraw[fill=grey, draw=darkblue, line width=0.03cm] ({3}, {-1*\radius + (\x-1)*\vspacing}) circle (\radius);

\node[above] at (1cm, 4.8cm) {Encoder};

  \foreach \x in {1,2,3}
    \filldraw[fill=grey, draw=black, line width=0.03cm] ({5.5}, {(\x-1)*\vspacing}) circle (\radius);

  \node[above] at (5.5cm, 3cm) {Internal state};

  \foreach \x in {1,2,3,4}
    \filldraw[fill=grey, draw=darkgreen, line width=0.03cm] ({8}, {-1*\radius + (\x-1)*\vspacing}) circle (\radius);

  \foreach \x in {1,2,3,4,5}
    \filldraw[fill=grey, draw=darkgreen, line width=0.03cm] ({9.5}, {-2*\radius + (\x-1)*\vspacing}) circle (\radius);
    
  \foreach \x in {1,2,3,4,5,6}
    \filldraw[fill=grey, draw=darkgreen, line width=0.03cm] ({11}, {-3*\radius + (\x-1)*\vspacing}) circle (\radius);

\node[above] at (10cm, 4.8cm) {Decoder};

\end{tikzpicture}
\caption[An encoder-decoder data-driven hl-discrete CFE generator]
{An encoder-decoder data-driven hl-discrete CFE generator, where \(n\) is the data dimension and \(s\) is the number of hl-discrete actions in the generated hl-discrete CFE \(\hat{I}\). 
}\label{tikz:fig_encdec}
\end{figure}
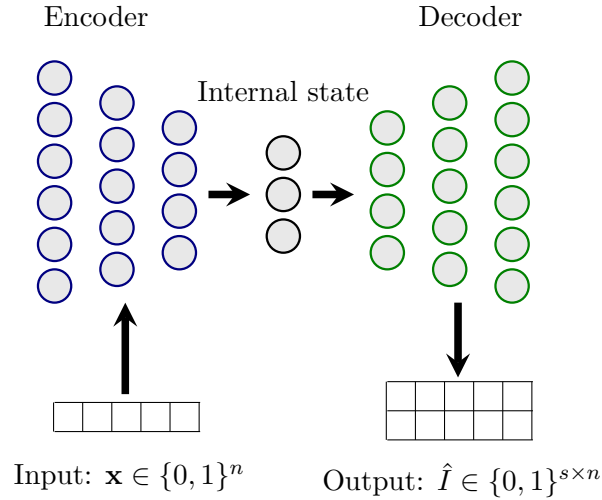

\subsection{The Data-Driven hl-discrete CFE Generator}

Figure~\ref{tikz:fig_encdec} shows the architecture of the neural-network based data-driven hl-discrete CFE generator.

\subsection{The Hamming Distance Data-Driven CFE Generator}

To produce hl-discrete-id CFEs (refer to Appendix~\ref{subsubsec:cfe_app_var_info_ds}) for new agents, we mainly used the data-driven hl-id CFE generator. However, we wanted to investigate the effect of model complexity on the accuracy of CFE generation. Therefore, we compare the more complex data-driven hl-id CFE generator (refer to Section~\ref{subsec:cfe_hlid_dd}) with a basic model, e.g., Hamming distance-based CFE generator, whose choice is due to the agent features being binary for this setting. Below is a description of the Hamming distance hl-discrete-id CFE generator.

Given a negatively classified new agent \(\mathbf{x}_{ts}\), we compute the Hamming distance (see Figure~\ref{tikz:ham_dist}) between them and each of the agents \(\mathbf{x}_{tr}\) in the agent–hl-discrete-id CFE training set. 
Then, based on these distances, we choose the \(k\) nearest training set agents and their associated  hl-discrete-id CFEs. We then use the most common hl-discrete-id CFE as the hl-discrete-id CFE for the new agent \(\mathbf{x}_{ts}\). We experimented with varied number of nearest neighbors: \(5, 10\) and \(15\), for the \(20\)-, \(50\)- and \(100\)-dimensional agent–hl-discrete-id CFE datasets, respectively. 
\begin{figure}[ht!]
    \centering
    \begin{tikzpicture}[scale=0.66]

        \draw[thick] (0.5,0.5) rectangle (20.5,-0.5);
        \draw[thick] (0.5,-1.5) rectangle (20.5,-2.5);

        \node[anchor=east] at (0,0) {\large{\(\mathbf{x}_{tr}\)}};
        \node[anchor=east] at (0,-2) {\large{\(\mathbf{x}_{ts}\)}};
        
        \tikzset{every node/.style={inner sep=5pt, minimum size=8pt}}
        
        \foreach \x/\bit in {1/1,2/0,3/1,4/0,5/1,6/1,7/0,8/1,9/0,10/1,11/0,12/1,13/1,14/0,15/1,16/0,17/1,18/0,19/1,20/0} {
            \node at (\x,0) {\bit};
        }
        
        \foreach \x/\bit in {1/1,2/0,3/1,4/0,5/0,6/0,7/1,8/1,9/0,10/1,11/0,12/1,13/0,14/1,15/1,16/0,17/0,18/0,19/1,20/0} {
            \node at (\x,-2) {\bit};
        }

        \foreach \x/\bit in {4,5,6, 12,13, 16} {
            \draw[red, fill=red, opacity=0.3] (\x+0.5, -2.5) rectangle (\x+1.5, 0.5);
        }

        \node at (10,-4) {Hamming Distance: (\(\mathbf{x}_{tr}, \ \mathbf{x}_{ts}\)) \(=  6\)};
    \end{tikzpicture}
    \caption[Hamming distance-based CFE generator]
    {Hamming distance between the agent–CFE training set agent \(\mathbf{x}_{tr}\) and a testing set agent \(\mathbf{x}_{ts}\). 
    }\label{tikz:ham_dist}
\end{figure}

\subsection{Extensions of Data-Driven CFE Generation}

\paragraph{Robust data-driven CFE generators.} CFE generators, whether single-agent, global, or data-driven, can be explicitly or implicitly compromised by changes in the underlying classifier, especially if that classifier is inaccurate or evolves with time. Low-level, feature-based CFEs, such as those proposed by \citet{Ustun19}, are particularly vulnerable: their specificity makes them highly sensitive to even minor modifications in the classification model. Data-driven CFE generators are also at risk of model drift and classification errors, primarily due to their reliance on agent–CFE datasets, implicitly or explicitly shaped by aggregation methods that depend on the current classifier. 

To address issues related to classification model errors and changes, recent robust CFE generation approaches \citep{Jiang2024} account for model and distribution drift during the CFE generation process, which would improve the overall performance of all CFE generators, whether single-agent, global, or data-driven.

A key direction for future research is to investigate the robustness of data-driven CFE generators under model drift. In particular, it would be valuable to assess whether high-level CFEs offer improved resilience to model changes or errors compared to feature-based, low-level CFEs. Furthermore, future work could focus on valuing agent–CFE data instances and developing robust aggregation and data-sharing strategies (e.g., federated learning) to enhance the overall performance of data-driven CFE generators.

\paragraph{Extension to multiple valid CFEs generation.} 
In the agent–CFE dataset, each agent is associated with a single valid CFE of minimal cost. The probability of agents possessing multiple unique valid CFEs with identical minimal costs was negligible because we assigned unique feature eligibility costs, resulting in varied overall hl-discrete action costs and uniquely optimal CFEs. In the rare cases where such ties occurred, we excluded the corresponding agents from the dataset.
Although we ensure each agent has a unique optimal CFE in the agent–CFE dataset, other valid CFEs that flip the prediction may exist but incur marginally higher costs. This scenario is more prevalent when generating sets of actions, whether hl-discrete or hl-continuous, as different combinations may achieve the same classification outcome at a higher cost.

Our current evaluation function (Equation~\ref{eq:evaluation}) is overly strict, disproportionately penalizing valid CFEs that incur higher costs. 
A promising avenue for future work involves extending data-driven CFE generators to produce sets of valid CFEs rather than single optimal ones. Instead of training the generators on agent–CFE pairs (each agent mapped to a unique minimal-cost CFE), train the generators on agent–validCFE sets, where each agent is associated with multiple valid CFEs. 
During inference, evaluate each generated CFE for prediction flip and relative cost, potentially enhancing robustness of the CFE generators.

\section{Supplemental Evaluation and Comparative Analysis Metrics}
\label{subsec:cfe_evaluation_mets}

Here, we provide additional details on the evaluation metrics used to compare low-level CFEs with both hl-continuous and hl-discrete CFEs, as well as to assess the effectiveness of data-driven CFE generators.

\subsection{Comparison Metrics}
We evaluate each key variable (\(v\)), such as the number of actions taken, features modified, agents using the same CFE, and agent improvement, when agents follow a low-level CFE vs. a high-level CFE (hl-continuous, or hl-discrete). Specifically, using the general Equation~\ref{eq:variable_change}, we define comparison metric \(\delta_{v}\) that compares each variables when an agent follows a low-level CFE versus an hl-continuous or hl-discrete CFE. 
\begin{equation}
    \begin{aligned}
    & \delta_{v} (P, Q)  =  P_{v} - Q_{v}
    \end{aligned}
    \label{eq:variable_change}
\end{equation}
Where \(P\) and \(Q\) denote two CFEs under consideration, e.g., \(P\) may correspond to a low-level CFE and \(Q\) to an hl-discrete CFE. The terms \(P_{v}\) and \(Q_{v}\) denote the variable value, such as the number of actions taken, following the execution of each CFE.

For all variables, a positive \(\delta_{v}\) indicates that the low-level CFE variable value is higher than the compared CFE, while a negative \(\delta_{v}\) indicates the opposite. The magnitude of \(\delta_{v}\) reflects the extent of this difference. Below are the specific \(\delta_{v}\) metrics.

\paragraph{Difference in number of actions taken.} 
The metric \(\delta_{\textrm{actions}}(\cdot,\cdot)\) (Equation~\ref{eq:actions_change})  quantifies the difference in the number of actions taken when an agent executes a low-level CFE versus an hl-continuous or hl-discrete CFE.
\begin{equation}
    \begin{aligned}
    & \delta_{\textrm{actions}} (P, Q)  = P_{\textrm{actions}} - Q_{\textrm{actions}}
    \end{aligned}
    \label{eq:actions_change}
\end{equation}
Here, \(P\) and \(Q\) represent the low-level CFE and the hl-continuous or hl-discrete CFE, respectively, while \(P_{\textrm{actions}}\) and \(Q_{\textrm{actions}}\) denote the number of actions taken when executing each. 

\paragraph{Difference in agent improvement.} 
The metric \(\delta_{\textrm{improvement}}(\cdot,\cdot)\) (Equation~\ref{eq:improvement_change})  quantifies the difference in the agent improvement earned when an agent executes a low-level CFE versus an hl-continuous or hl-discrete CFE.
\begin{equation}
    \begin{aligned}
    & \delta_{\textrm{improvement}} (P, Q)  = P_{\textrm{improvement}} - Q_{\textrm{improvement}}
    \end{aligned}
    \label{eq:improvement_change}
\end{equation}
Here, \(P\) and \(Q\) represent the low-level CFE and the hl-continuous or hl-discrete CFE, respectively, while \(P_{\textrm{improvement}}\) and \(Q_{\textrm{improvement}}\) denote the agent improvement earned when executing each. Specifically, 
\begin{equation}
    \begin{aligned}
    & P_{\textrm{improvement}} = \|\mathbf{x}' - \mathbf{x}\|
    \end{aligned}
    \label{eq:proximity_eq}
\end{equation}
where \(P\) is the CFE taken and \(\mathbf{x}'\) is the resultant agent state after taking the CFE from \(\mathbf{x}\), which is the initial agent state. Ideally high improvement (less proximate), that is, \(\mathbf{x}'\) more distant from \(\mathbf{x}\) is preferred.

\paragraph{Difference in number of features modified.} 
The metric \(\delta_{\textrm{features}}(\cdot,\cdot)\) (Equation~\ref{eq:features_change})  quantifies the difference in the number features modified when an agent executes a low-level CFE versus an hl-continuous or hl-discrete CFE.
\begin{equation}
    \begin{aligned}
    & \delta_{\textrm{features}} (P, Q)  = P_{\textrm{features}} - Q_{\textrm{features}}
    \end{aligned}
    \label{eq:features_change}
\end{equation}
Here, \(P\) and \(Q\) represent the low-level CFE and the hl-continuous or hl-discrete CFE, respectively, while \(P_{\textrm{features}}\) and \(Q_{\textrm{features}}\) denote the number of features modified when executing each.

\subsection{Statistical Significance between Variables}
Given the different variables, e.g., list of the number of actions taken, number of modified features, and improvement achieved with each CFE: hl-continuous, hl-discrete, and low-level, we compute the statistical significance of the differences.
We use the Scipy stats tool \citep{SciPy-Kendalltau2023} to compute the Kendall tau and \(p\)-value to assess the statistical significance of the relationship between the two variables at a time.

\section{Supplemental Experimental Results}
\label{sec:cfe_app_exper_results}

In this section, we provide additional and thorough empirical evidence demonstrating the strong performance of the proposed data-driven CFE generators in producing optimal CFEs for new agents. We also highlight the strong and desirable characteristics of the hl-continuous and hl-discrete CFEs over the low-level CFEs. Lastly, we analyze how various constraints, such as varied data dimensions, the frequency of CFEs, decision-makers information access, feature satisfiability, and restrictions on agents' access to actions, affect the agent–CFE data distribution and the effectiveness of data-driven CFE generators.

\subsection{High-level CFEs Result in More Improvement and Feature Modifications}
\label{subsec:cfe_app_higher_improv}

\begin{figure}[b!]
\captionsetup[subfigure]{position=top, margin=-0.5cm, labelfont=bf,textfont=normalfont,singlelinecheck=off,justification=raggedright}
\begin{center}
    \begin{subfigure}[b]{0.99\textwidth}
        \centering
        \caption{}
        \label{fig:ap_avg_actions}
        \includegraphics[width=\textwidth]{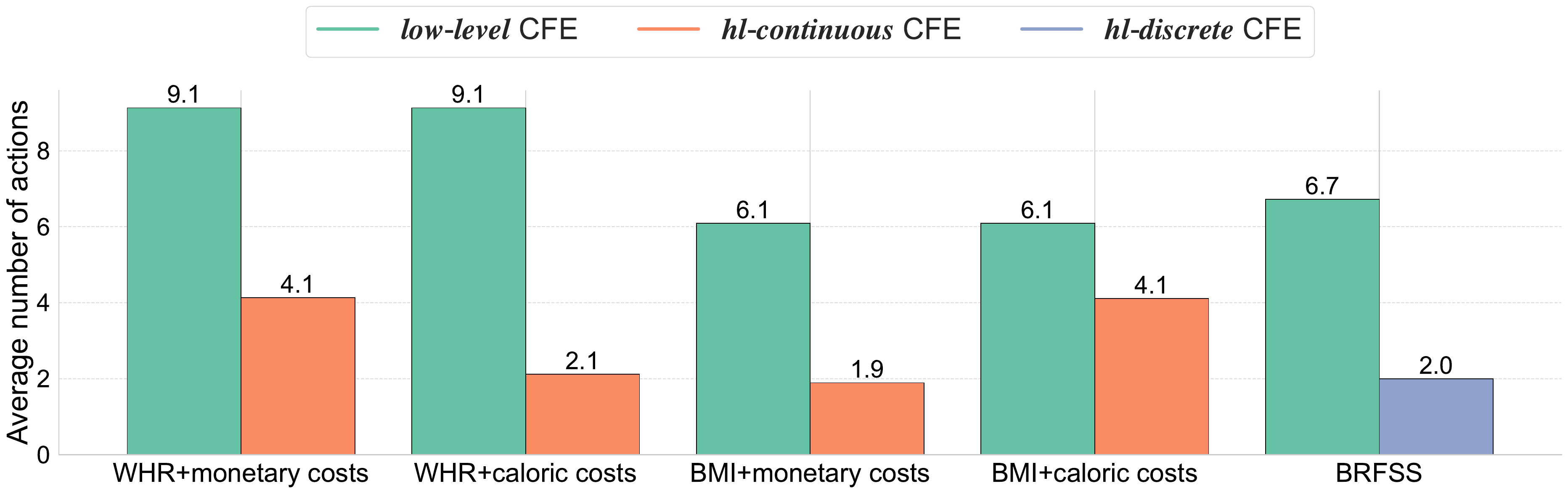}
    \end{subfigure}%
    \vskip\baselineskip
    \begin{subfigure}[b]{0.99\textwidth}
        \centering
        \caption{}
        \label{fig:ap_avg_feats}
        \includegraphics[width=\textwidth]{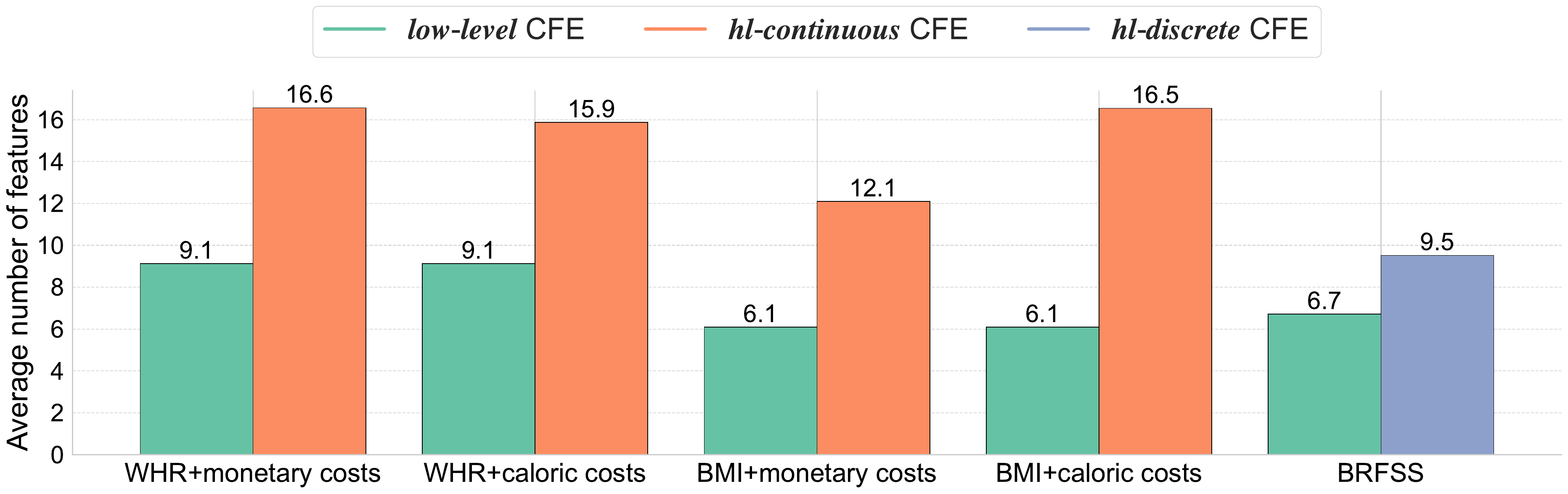}
    \end{subfigure}
    \vskip\baselineskip
    \begin{subfigure}[b]{0.99\textwidth}
        \centering
        \caption{}
        \label{fig:ap_avg_proximity}
        \includegraphics[width=\textwidth]{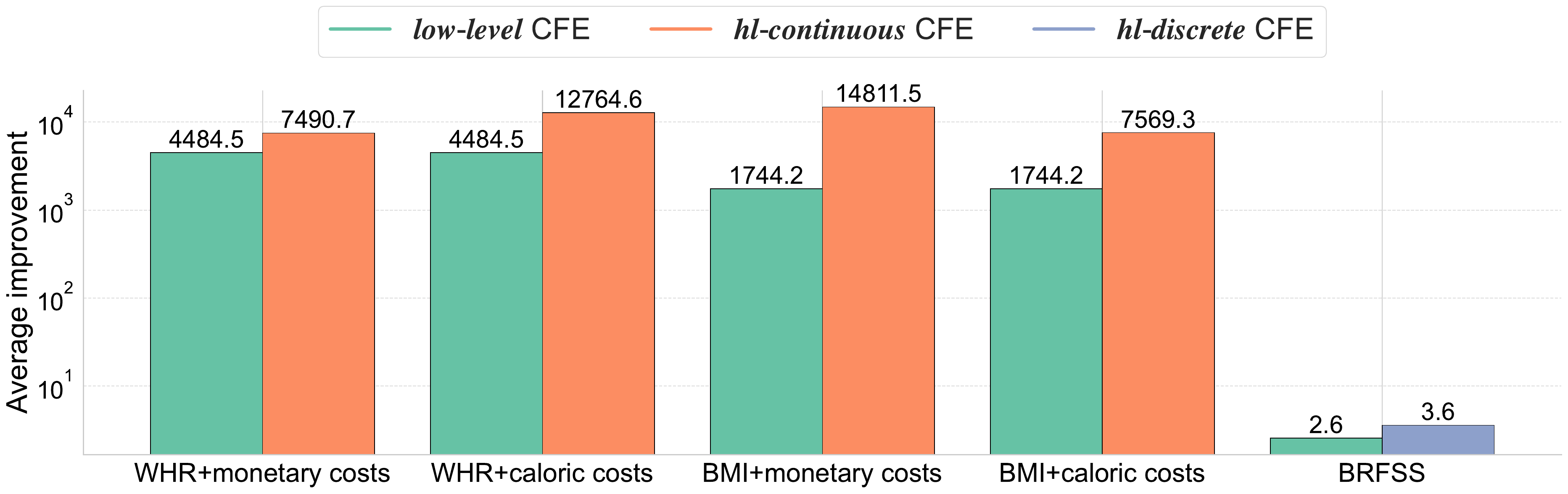}
    \end{subfigure}
\end{center}
\caption[Comparison of hl-continuous and hl-discrete CFEs to low-level CFEs]
{A comparison of hl-continuous CFEs consisting of a set of Food+monetary or Food+caloric cost hl-continuous actions for WHR and BMI datasets, alongside hl-discrete CFEs on the BRFSS dataset, evaluated against low-level CFEs for their respective datasets. All annotations up to one decimal place, low-level CFEs require \subref{fig:ap_avg_actions} more actions but lead to  \subref{fig:ap_avg_feats} fewer feature modifications and \subref{fig:ap_avg_proximity} result in less improvement (i.e., closer resultant agent states) compared to hl-discrete and hl-continuous CFEs.
}\label{fig:ap_avg_prox_feats_acts}
\end{figure}

\begin{figure}[t!]
\begin{center}
\begin{subfigure}[b]{0.49\textwidth}            
            \includegraphics[width=0.9\textwidth]{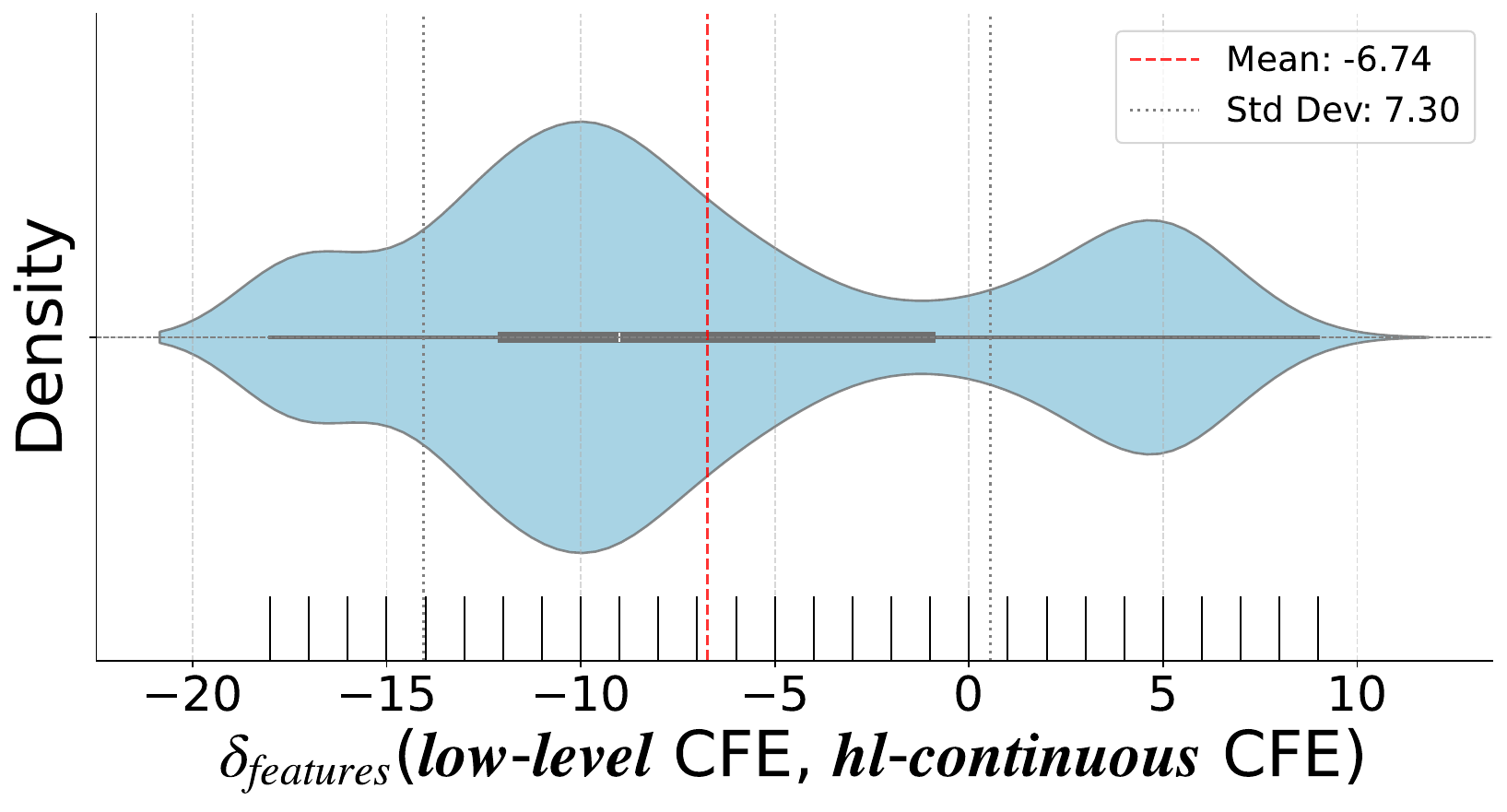}
            \caption{Difference in number of modified features}
            \label{fig:ap_change_feats}
    \end{subfigure}%
    \hfill
    \begin{subfigure}[b]{0.49\textwidth}
            \centering
            \includegraphics[width=0.9\textwidth]{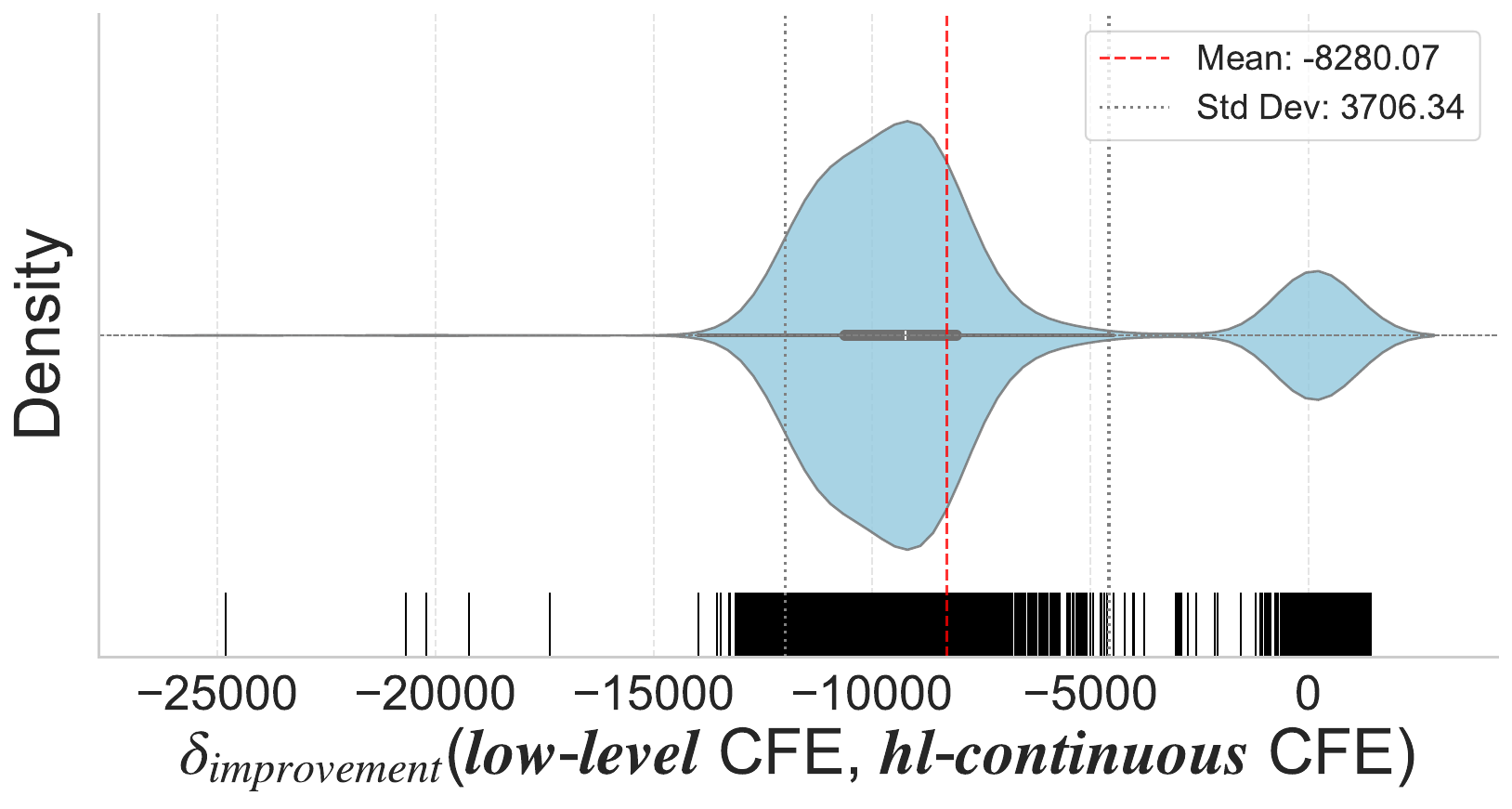}
            \caption{Difference in agent improvement}
            \label{fig:ap_change_immprove}
    \end{subfigure}
\end{center}
\caption[Distribution of difference in number of modified features and agent improvement]{Given WHR negatively classified agents and the low-level and hl-continuous CFEs they took, a computation of \(\delta_{\textrm{improvement}}(P, Q)\) (Equation~\ref{eq:improvement_change})\}  and \(\delta_{\textrm{features}}(P, Q)\) (Equation~\ref{eq:features_change}) where \(P\) denotes taking a low-level CFE and \(Q\) denotes taking an hl-continuous CFE, shows that most of the density is negative implying that 
when agents take hl-continuous CFEs, a higher number of their features is modified  \subref{fig:ap_change_feats} and resultant improvement is significantly higher \subref{fig:ap_change_immprove} than if they took low-level CFEs.} 
\label{fig:ap_change_feats_improv}
\end{figure}

Unlike low-level CFEs, high-level CFEs (hl-continuous and hl-discrete CFEs) involve fewer actions on average (see Figure~\ref{fig:ap_avg_actions}) and results in higher improvements (Figures~\ref{fig:ap_avg_proximity} and ~\ref{fig:ap_change_immprove}) and simultaneously modify multiple features  (see Figures~\ref{fig:ap_avg_feats} and ~\ref{fig:ap_change_feats}).

\begin{figure}[ht!]
\begin{center}
    \begin{subfigure}[b]{0.363\textwidth}
        \centering
        \includegraphics[width=1\textwidth]{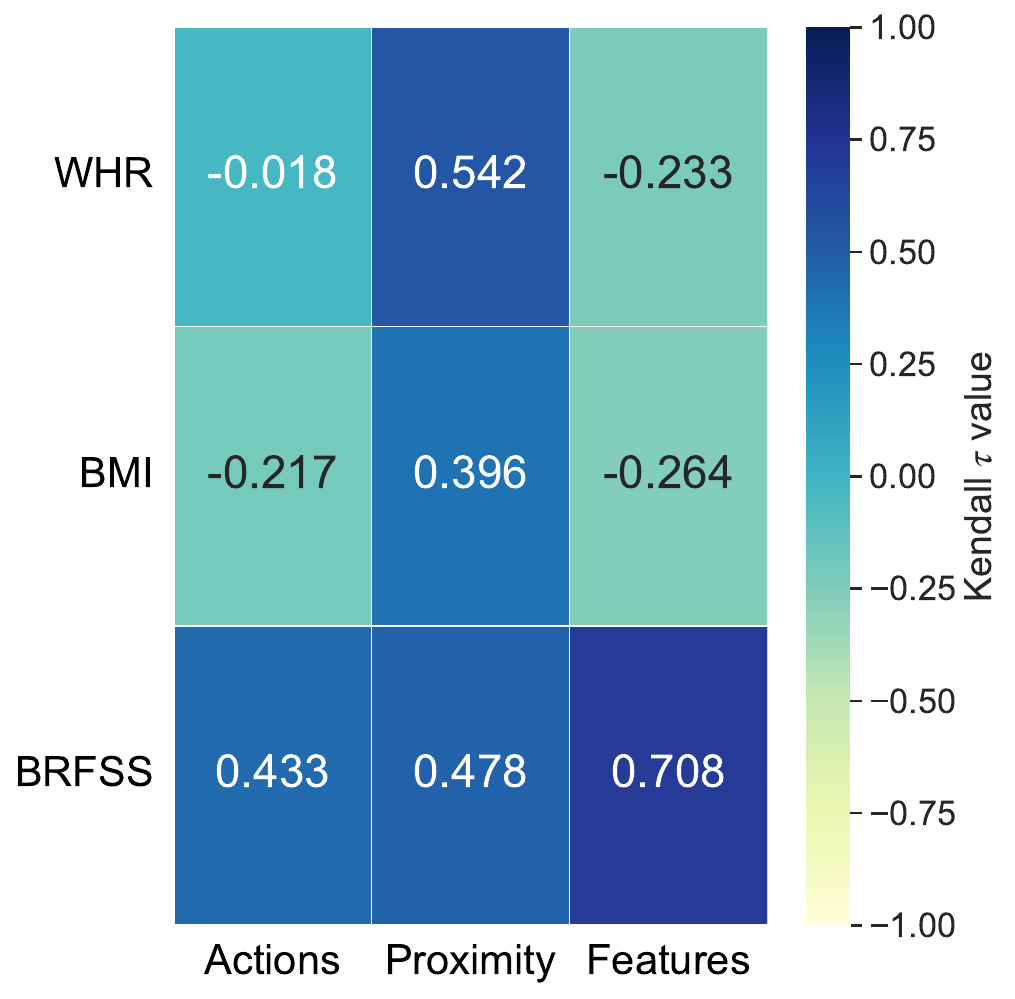}
        \caption{}
        \label{fig:ap_tau_ar_vs_fa}
    \end{subfigure}%
    \hfill
    \begin{subfigure}[b]{0.63\textwidth}
        \centering
        \includegraphics[width=1\textwidth]{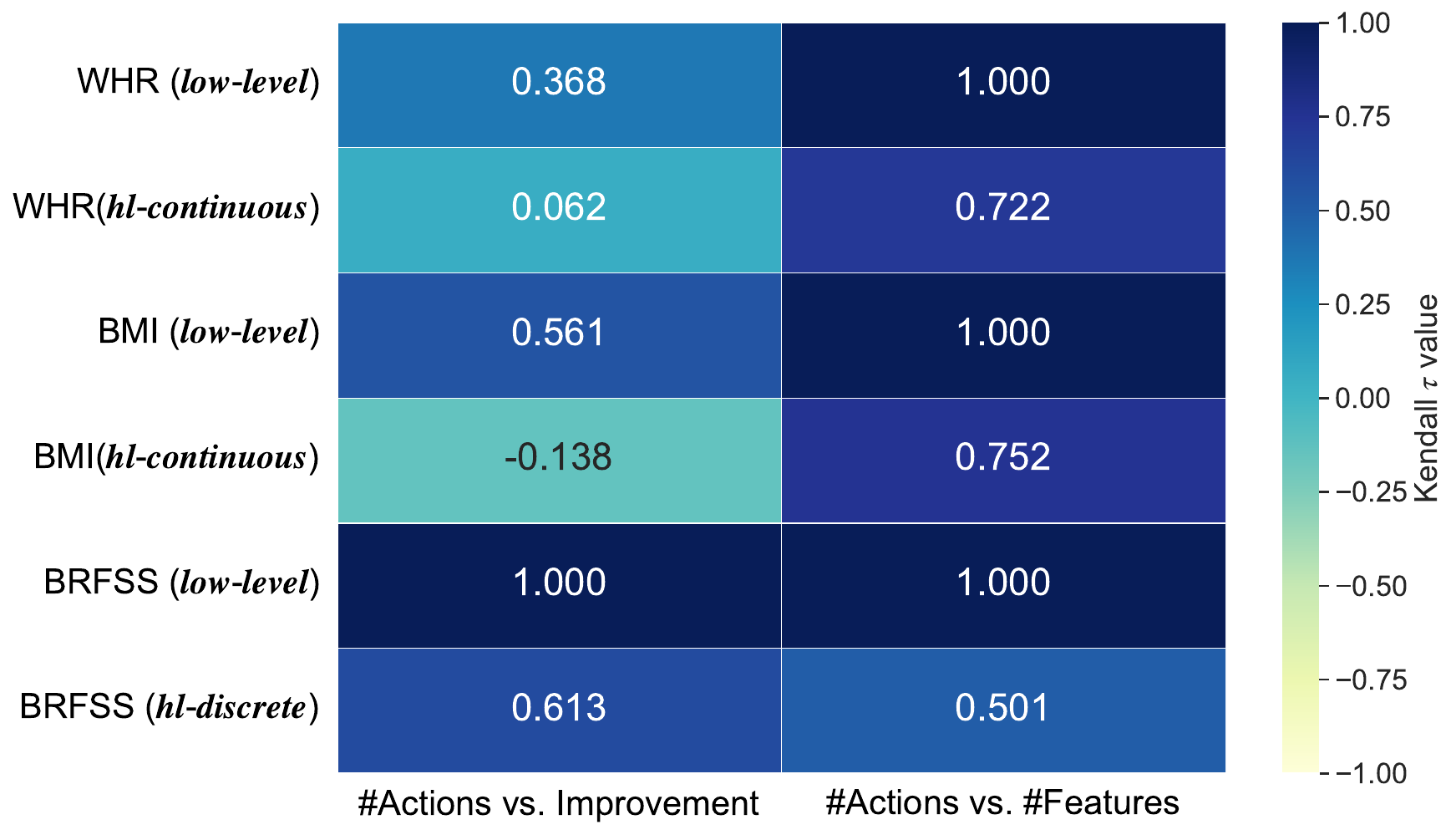}
        \caption{}
        \label{fig:ap_tau_acts_vs_pro_feat}
    \end{subfigure}
\end{center}
\caption[The low-level CFEs have a perfect positive relationship]
{In \subref{fig:ap_tau_ar_vs_fa}, we illustrate the correlations for three different aspects: (1) between the number of actions taken with CFEs P and Q, (2) between the number of features modified with CFEs P and Q, and (3) between the improvement achieved after taking CFEs P and Q. For the BMI and WHR datasets, P and Q represent low-level and hl-continuous CFEs, respectively. For the BRFSS dataset, P and Q denote low-level and hl-discrete CFEs, respectively. 
On the other hand, \subref{fig:ap_tau_acts_vs_pro_feat} shows the correlation between the number of actions taken and the number of modified features and between the number of actions taken and improvement achieved for each CFE and dataset. 
In general, low-level CFEs have a perfect positive relationship between the number of actions and modified features
}\label{fig:ap_kendal_tau}
\end{figure}

While low-level CFEs exhibit a perfect correlation between the number of actions taken and the number of features modified, hl-continuous and hl-discrete CFEs show a positive but weaker relationship (Figure~\ref{fig:ap_tau_acts_vs_pro_feat}). Additionally, hl-discrete and low-level CFEs have a strong positive correlation (\(\tau = 0.708\)) in the number of modified features (see BRFSS dataset in Figure~\ref{fig:ap_tau_ar_vs_fa}).  In contrast, hl-continuous CFEs show a weak negative correlation with low-level CFEs in the number of modified features and actions taken (see BMI and WHR datasets in Figure~\ref{fig:ap_tau_ar_vs_fa}, \(\tau = -0.2684\) and \(\tau = -0.233\) respectively).

\begin{figure}[ht!]
\centering
    \begin{subfigure}[b]{0.49\textwidth}
        \centering
        \includegraphics[width=0.99\textwidth]{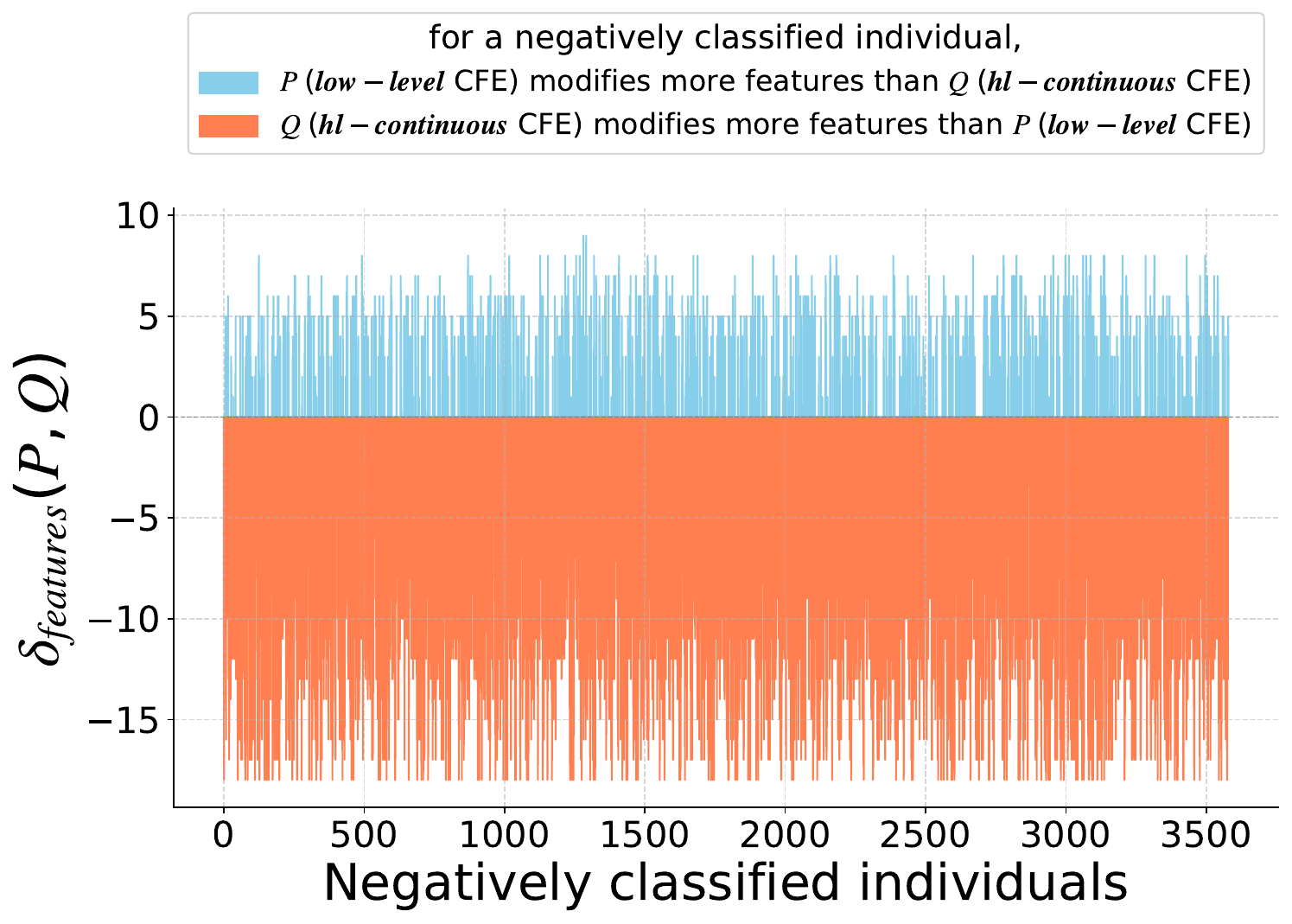}
        \caption{Difference in number of modified features}
        \label{fig:sparsity_app}
    \end{subfigure}%
    \hfill
    \begin{subfigure}[b]{0.49\textwidth}
        \centering
        \includegraphics[width=0.99\textwidth]{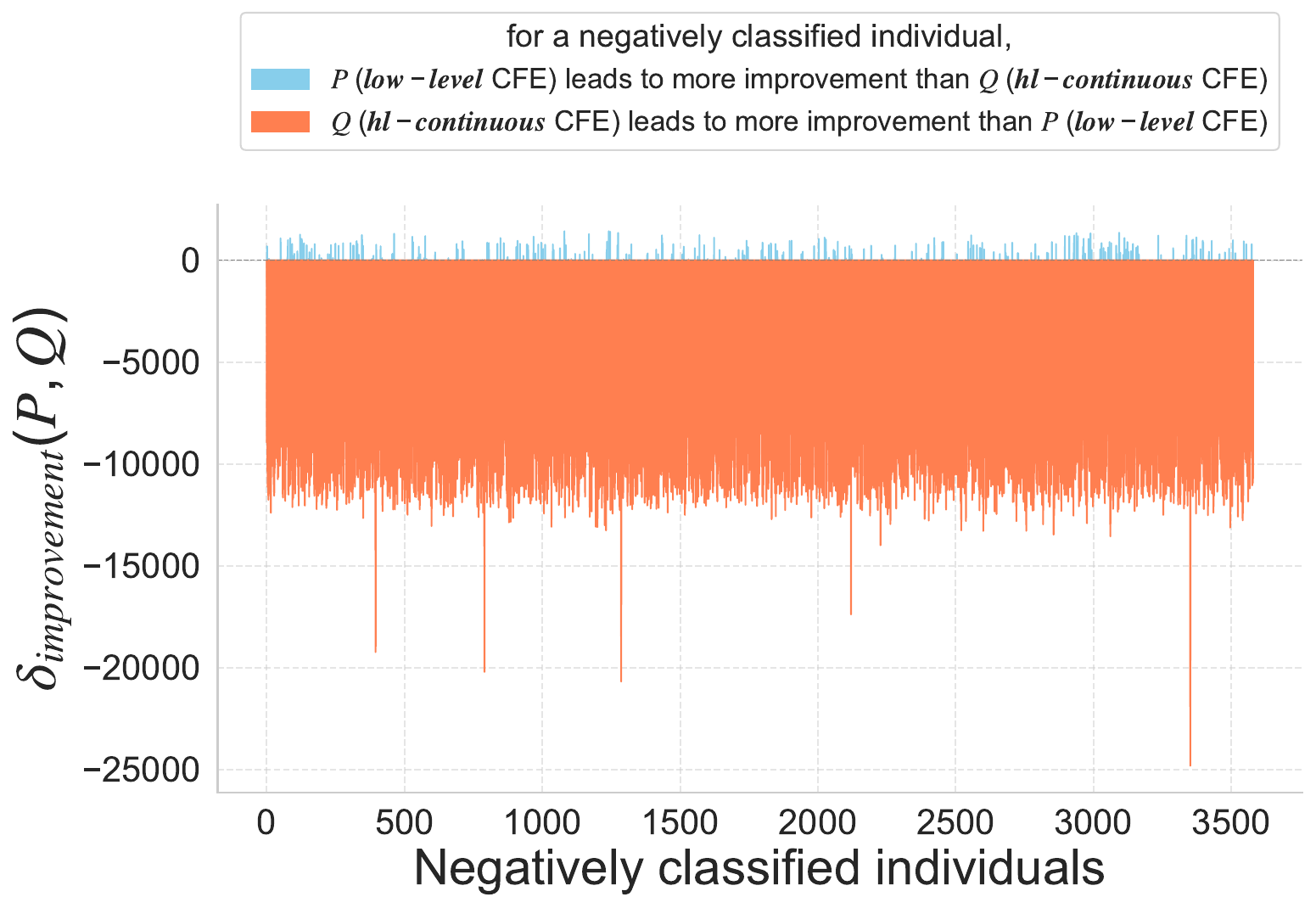}
        \caption{Difference in improvement achieved}
        \label{fig:proximity_app}
    \end{subfigure}
    \caption[Difference in number of modified features and agent improvement]{A comparative analysis of the \subref{fig:sparsity_app} difference in number of modified features (\(\delta_{\textrm{features}}(P, Q)\) (Equation~\ref{eq:features_change})) and \subref{fig:proximity_app} difference in agent improvement (\(\delta_{\textrm{improvement}}(P, Q)\) (Equation~\ref{eq:improvement_change})) when each negatively classified WHR agent takes a low-level CFE (\textbf{P}) versus an hl-continuous CFE (\textbf{Q}), shows that hl-continuous CFEs modify more features \subref{fig:sparsity_app} and lead to significantly higher improvement \subref{fig:proximity_app} than low-level CFEs.}
    \label{fig:ap_change_feats_improv2}
\end{figure}

\subsection{High-level CFEs are Easier to Personalize and Lead to Fairer Outcomes}
\label{subsec:cfe_app_fair_personalize}

Fairness in CFE generation has primarily been studied along the dimension of equalizing the recourse costs across different groups (e.g., \citep{Gupta2019EqualizingRA}). In this work, we extend the analysis by exploring several dimensions of fairness in CFE generation. 

First, we investigate how agents across sensitive groups using the same CFE generator (same kind of CFEs) experience differences in how much they improve, the number of actions taken, the number of modified features, and the costs incurred.
Second, we explore the effects of limiting agents to a subset of actions (varied access to actions) on the distribution of agent–CFE datasets and the accuracy of data-driven CFE generators across groups.
Lastly, we examine variations in feature satisfiability (differences in what features need to be satisfied) across agent groups, influences the distribution of the agent–CFE dataset, and the performance of data-driven CFE generators in generating CFEs for different agent groups. 

In addition to fairness, we also investigate the personalization of CFE generation along two dimensions. 1) Agents may be interested in a subset of actions (varied access to actions) and thus restricted to CFEs that involve only specific actions. 2) Agents might prioritize different costs in the CFE generation process  (varied cost preferences) and thus prefer CFE generators that optimize those specific costs in CFE generation, e.g., caloric costs over monetary ones.

\paragraph{Fairness based on variability of CFEs execution outcome.}
\label{sec:cfe_app_fa_ar_da_comp}

We analyze variations in costs incurred, actions taken, features modified, and agent improvement across sensitive groups to assess the fairness of low-level CFEs compared to hl-continuous and hl-discrete CFEs.

\begin{figure}[b!]
\centering
    \begin{subfigure}[b]{0.71\textwidth}            
            \includegraphics[width=1\textwidth]{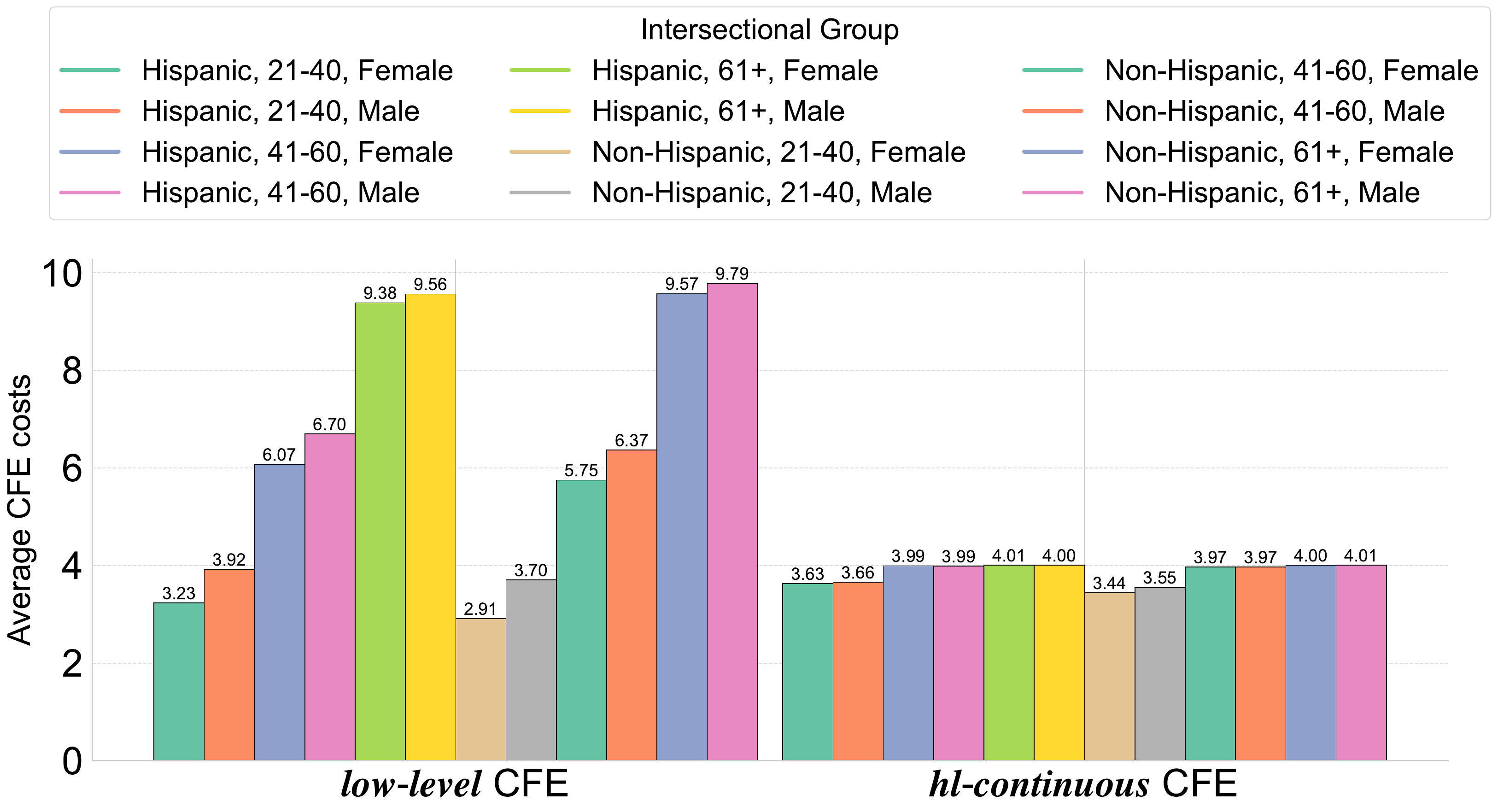}
            \caption{BMI: average costs incurred across groups}
            \label{fig:bmi_real_costs}
    \end{subfigure}%
    \hfill
    \begin{subfigure}[b]{0.29\textwidth}
            \centering
            \includegraphics[width=1\textwidth]{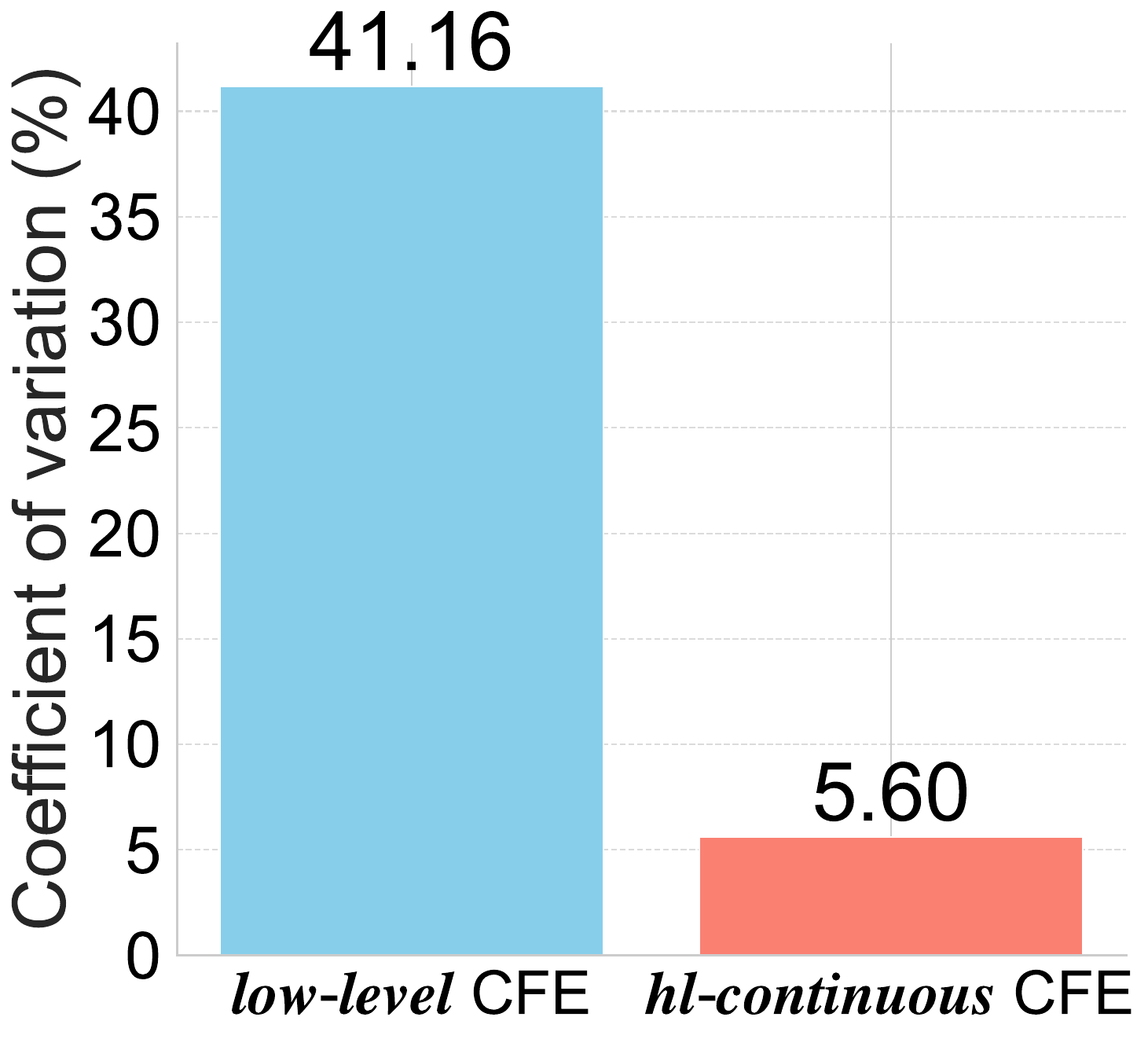}
            \caption{BMI: variability in costs}
            \label{fig:bmi_var_costs}
    \end{subfigure}
    \vskip\baselineskip
    \begin{subfigure}[b]{0.71\textwidth}            
            \includegraphics[width=1\textwidth]{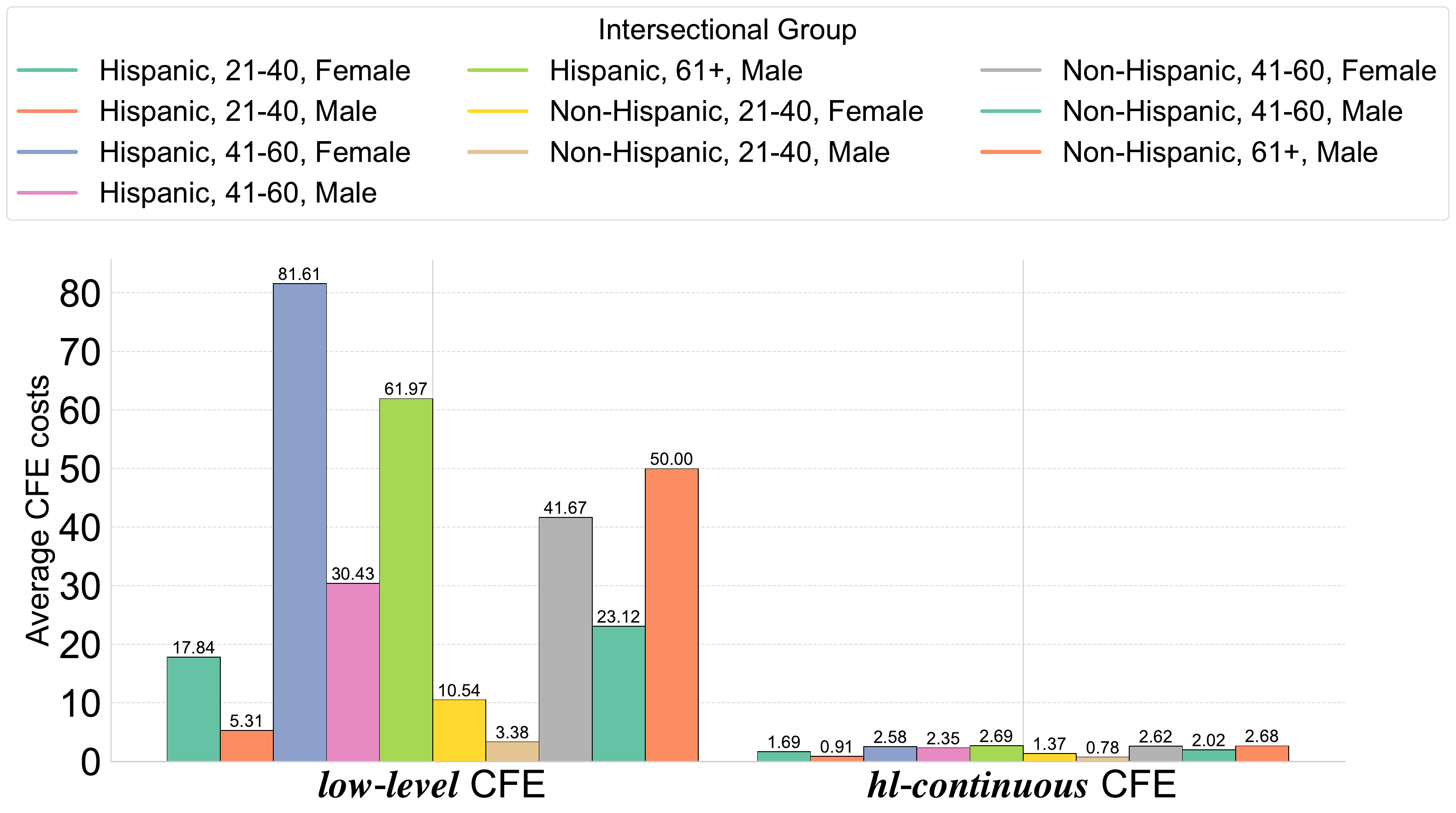}
            \caption{WHR: average cost  incurred across groups}
            \label{fig:whr-calprice_real_costs}
    \end{subfigure}%
    \hfill
    \begin{subfigure}[b]{0.29\textwidth}
            \centering
            \includegraphics[width=1\textwidth]{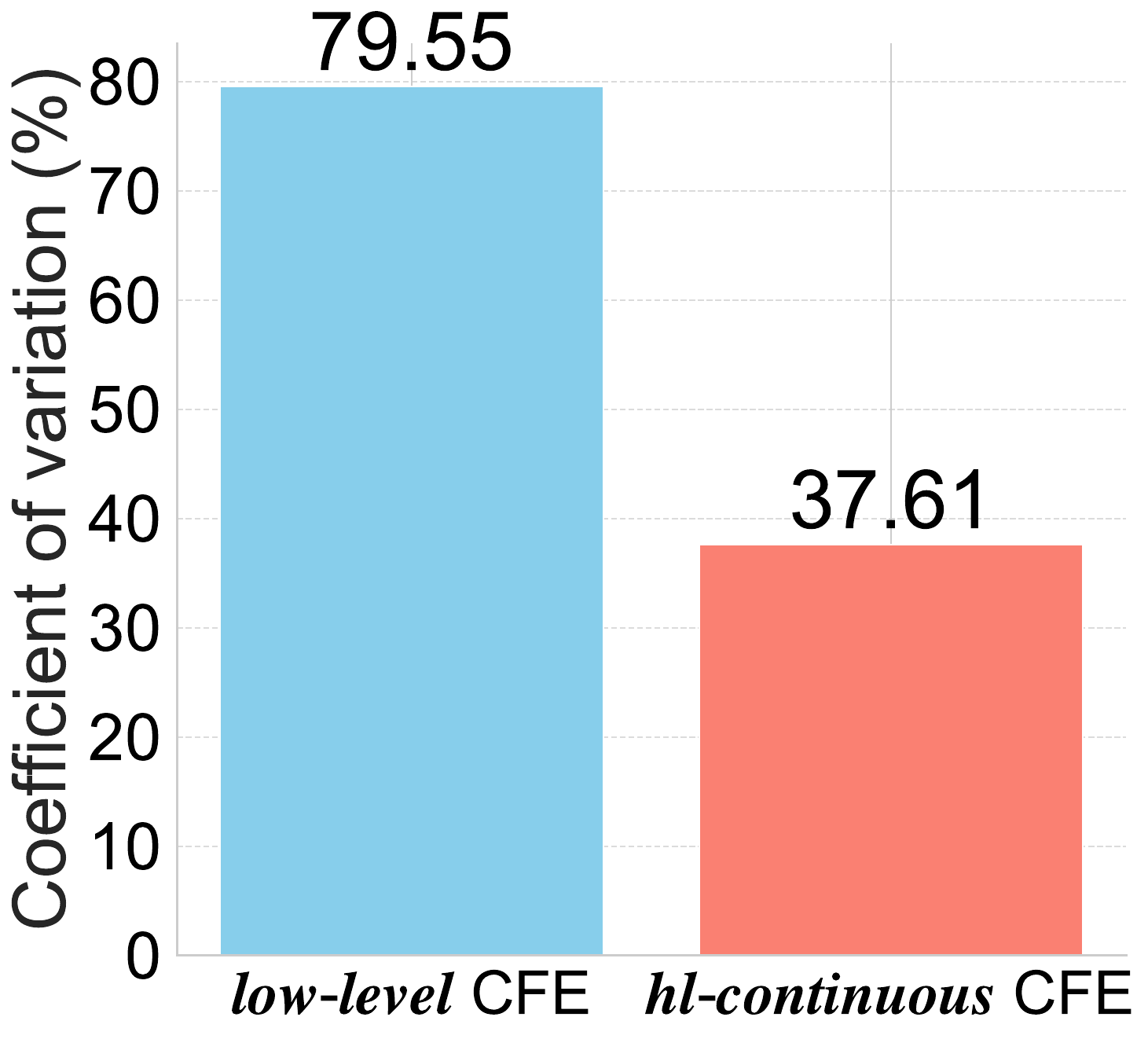}
            \caption{WHR: variability in costs}
            \label{fig:whr-calprice_var_costs}
    \end{subfigure}
    \caption[A comparative analysis of costs for executing CFEs]
    {A comparative analysis of average cost variations among agents in sensitive groups when taking low-level CFEs versus high-level continuous CFEs. Although not comparable across CFEs, \subref{fig:bmi_real_costs} and \subref{fig:whr-calprice_real_costs} show the distribution of costs between groups for each CFE, and \subref{fig:bmi_var_costs} and \subref{fig:whr-calprice_var_costs} show the coefficient of variations; indicating how variable around mean the average costs in groups are. Costs across sensitive groups vary more when agents take low-level CFEs than when they take hl-continuous CFEs.
    }\label{fig:whr-calprice_cost}
\end{figure}

\begin{figure}[htpb!]
\centering
    \begin{subfigure}[b]{0.71\textwidth}            
            \includegraphics[width=0.88\textwidth]{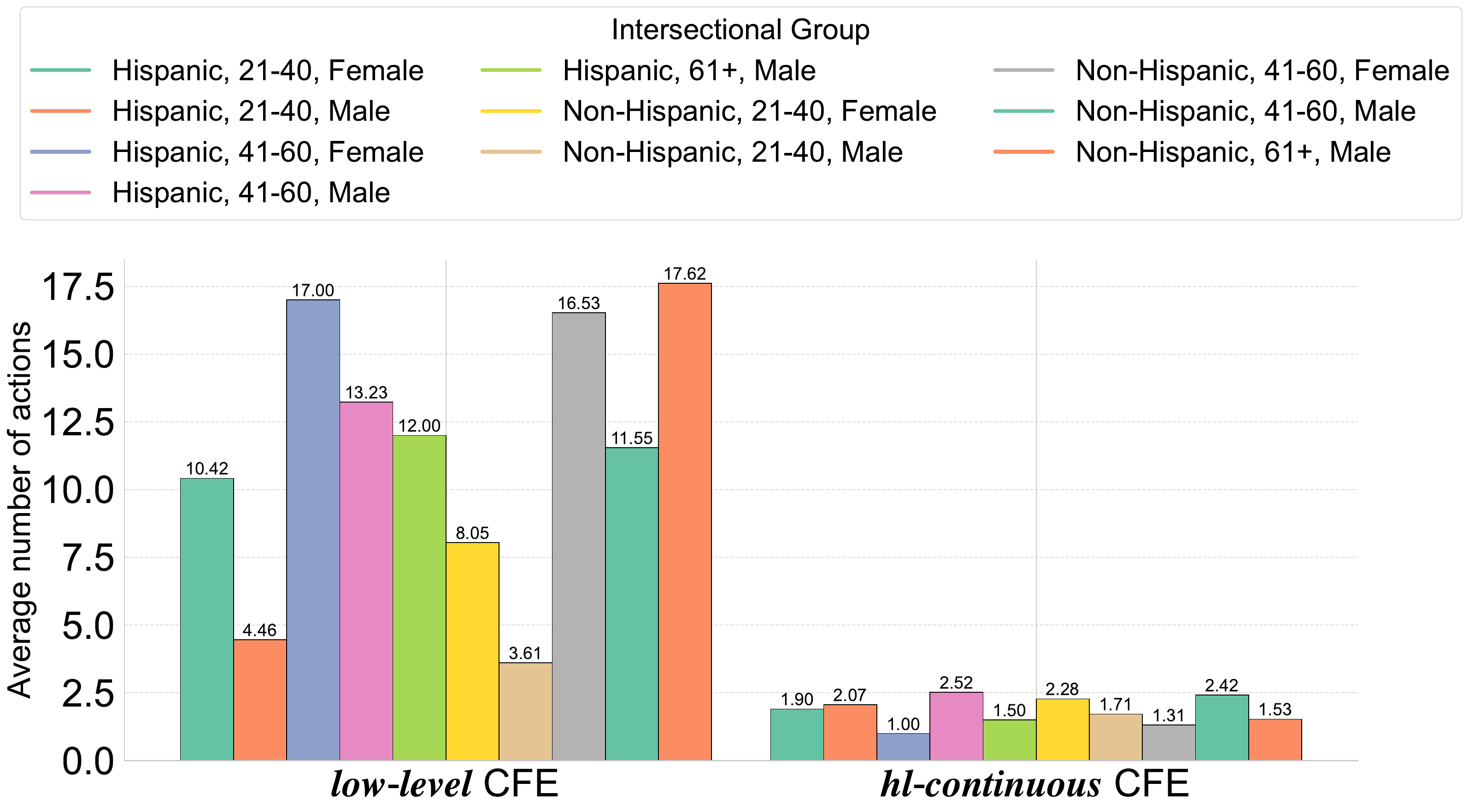}
            \caption{Average number of actions taken across groups}
            \label{fig:whr_real_acts}
    \end{subfigure}%
    \hfill
    \begin{subfigure}[b]{0.28\textwidth}
            \centering
            \includegraphics[width=0.88\textwidth]{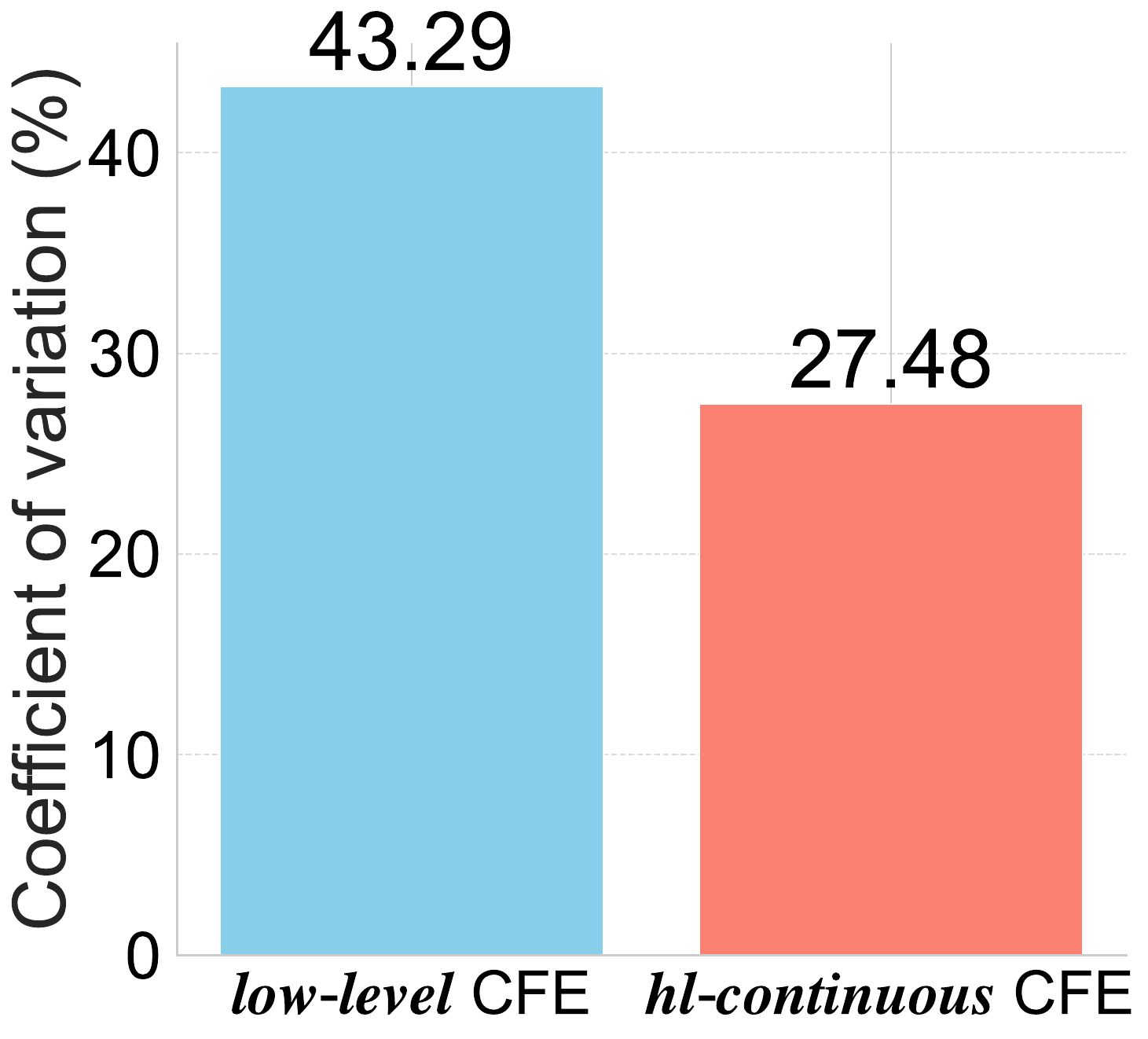}
            \caption{Variability in \#actions}
            \label{fig:whr_var_acts}
    \end{subfigure}
    \vskip\baselineskip
    \begin{subfigure}[b]{0.71\textwidth}            
            \includegraphics[width=0.88\textwidth]{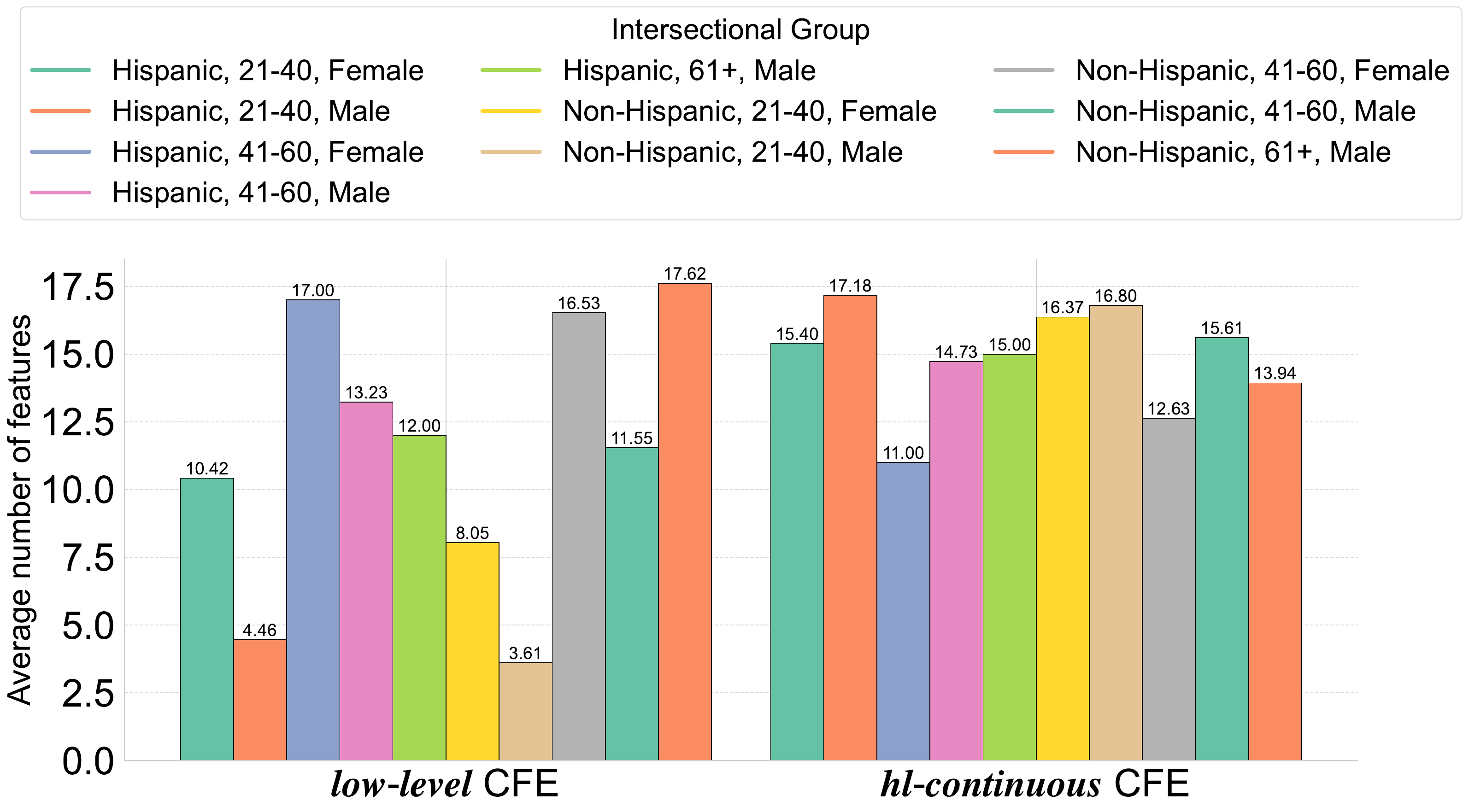}
            \caption{Average number of features modified across groups}
            \label{fig:whr_real_feats}
    \end{subfigure}%
    \begin{subfigure}[b]{0.28\textwidth}
            \centering
            \includegraphics[width=0.88\textwidth]{cfes/cfe_figures/whr_calprice_intersect_group_num_feats_cv.pdf}
            \caption{Variability in \#features}
            \label{fig:whr_var_feats}
    \end{subfigure}
    \vskip\baselineskip
    \begin{subfigure}[b]{0.71\textwidth}            
            \includegraphics[width=0.88\textwidth]{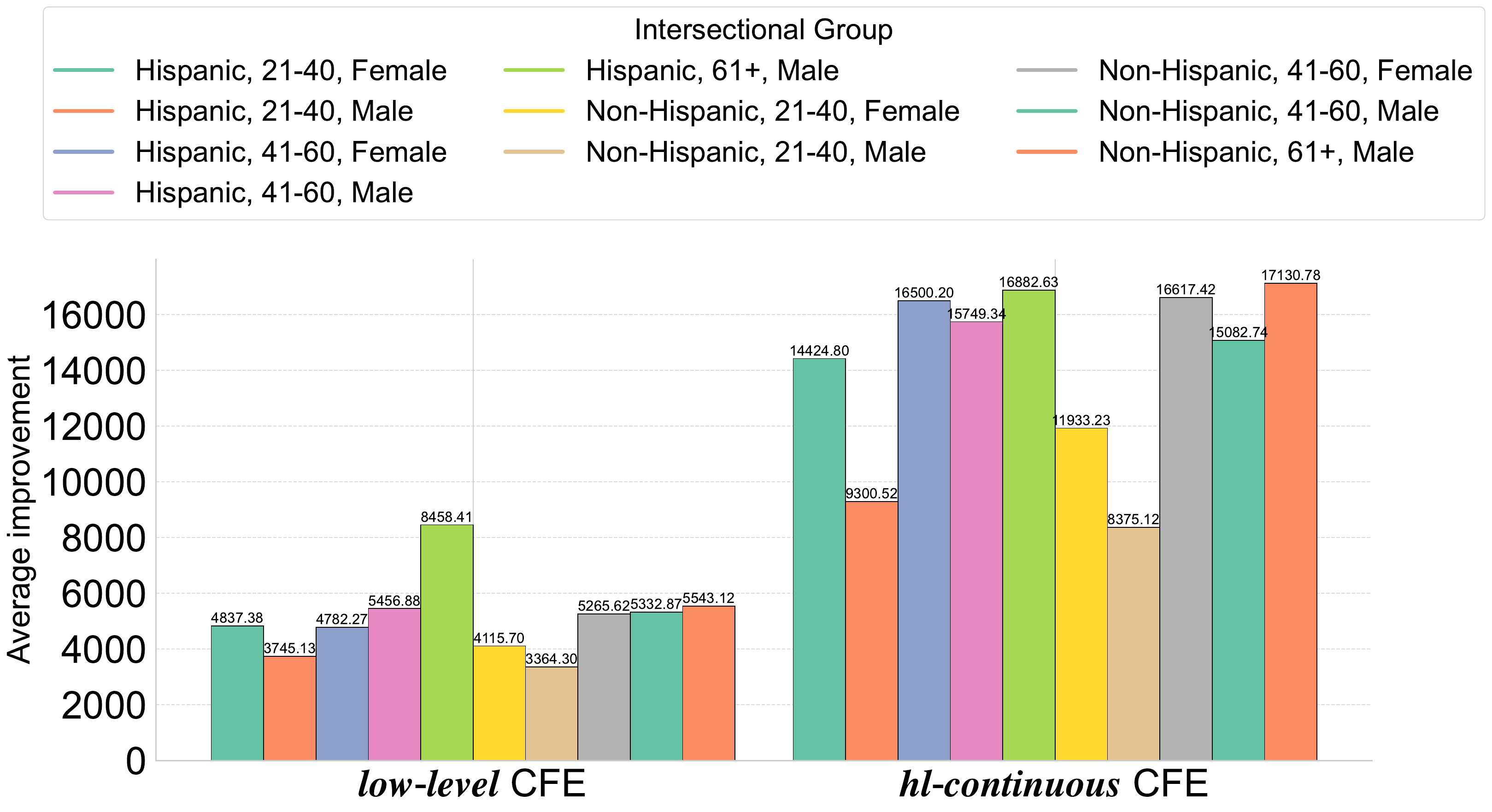}
            \caption{Average improvement achieved across groups}
            \label{fig:whr_real_proximity}
    \end{subfigure}%
    \hfill
    \begin{subfigure}[b]{0.28\textwidth}
            \centering
            \includegraphics[width=0.88\textwidth]{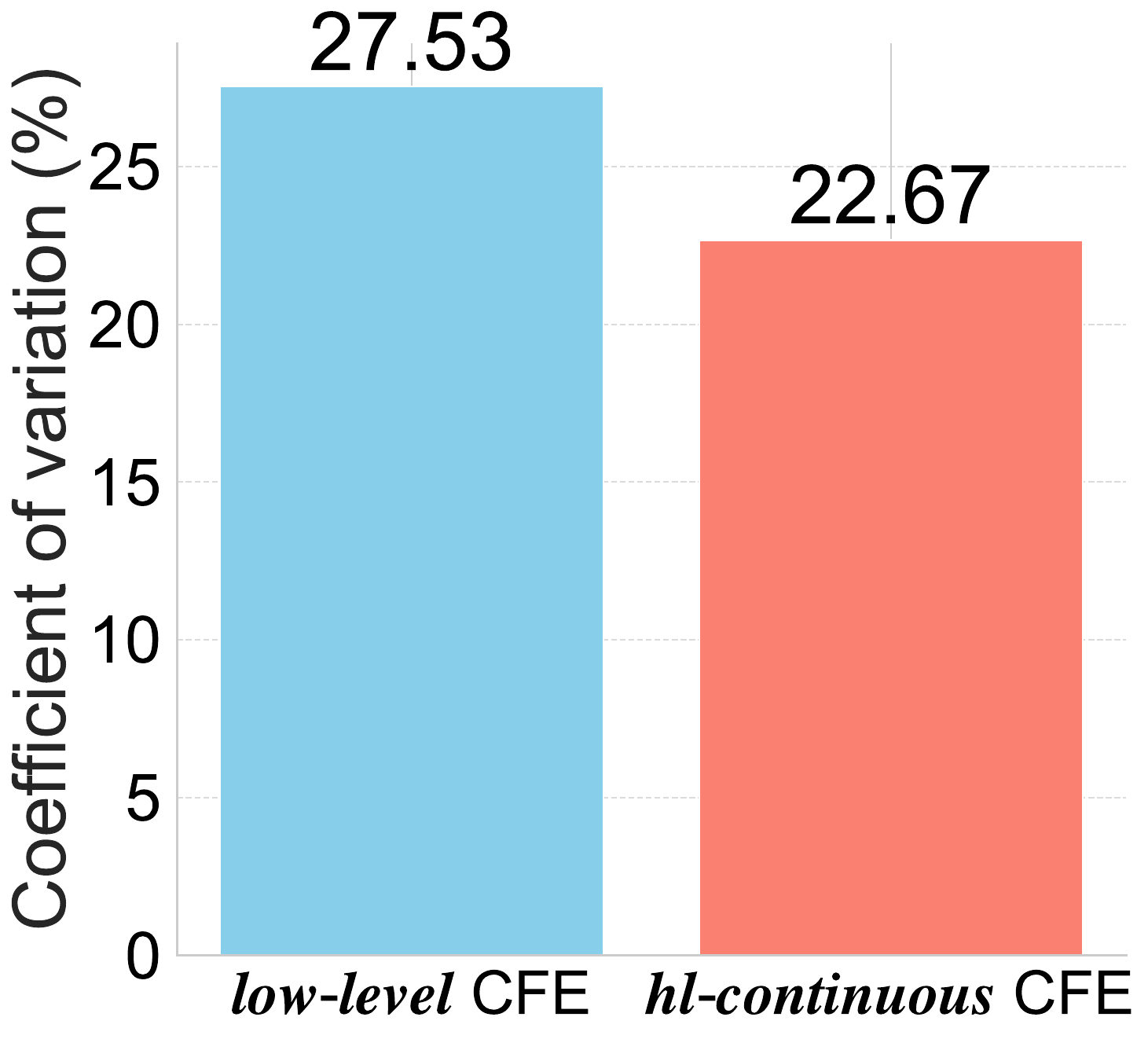}
            \caption{Variability in improvement}
            \label{fig:wh_var_proximity}
    \end{subfigure}
    \caption[A comparative analysis of variables resulting from execution of CFEs]
    {A comparative analysis of the average number of actions taken, the number of features modified, and the improvement achieved by agents across sensitive groups when they take low-level CFEs versus hl-continuous CFEs. \subref{fig:whr_real_acts}, \subref{fig:whr_real_feats}, and \subref{fig:whr_real_proximity} show the raw distributions for these variables across sensitive groups while \subref{fig:whr_var_acts}, \subref{fig:whr_var_feats}, and \subref{fig:wh_var_proximity} present the coefficients of variation that concisely illustrate the extent of dispersion around the mean for each variable. In summary, low-level CFEs are less fair than hl-continuous CFEs, which exhibit lower coefficients of variation, ensuring more comparable outcomes for agents from different groups.
    }\label{fig:whr-calprice_actions_feats_proximity}
\end{figure}

\begin{enumerate}[label=\arabic*)]
    \item \textbf{Variability in improvement and number of modified features and actions taken.}
    To quantify differences in agent experiences across sensitive groups, we compute the coefficient of variation for three key variables: improvement, number of modified features, and number of actions taken.
    Figure~\ref{fig:whr-calprice_actions_feats_proximity} illustrates that in the WHR dataset, low-level CFEs exhibit substantial variation across sensitive groups regarding agent improvement, actions taken, and features modified. Specifically, the coefficient of variation for agent improvement was \(27.53\%\)  with low-level CFEs versus \(22.67\%\) otherwise. The variation in the number of actions taken was \(43.29\%\) compared to \(27.48\%\), and for modified features, \(43.29\%\) compared to \(12.88\%\). These findings indicate that the benefits of low-level CFEs are not distributed equally across sensitive groups, potentially favoring some over others, thus raising fairness concerns in CFE generation.
    
    \item \textbf{Variability in costs incurred across sensitive groups.} Although the costs agents incur by taking low-level CFEs cannot be directly compared with taking hl-continuous CFEs because they are contextually different, we study how the costs of executing the same kind of CFEs varies across agents in different sensitive groups. 
    
    Our results show that the costs incurred in taking low-level CFEs vary more widely across various sensitive groups than in taking hl-continuous CFEs. For example, in Figure~\ref{fig:whr-calprice_cost}, the coefficient of variation for taking low-level CFEs is \(41.16\%\) and \(79.55\%\) versus \(5.60\%\) and \(37.61\%\) with taking hl-continuous CFEs, on BMI and WHR datasets, respectively. Therefore, compared to taking hl-continuous CFEs, taking low-level CFEs is more biased and more likely to favor some sensitive groups.
\end{enumerate}

\begin{figure}[t!]
\centering
    \begin{subfigure}[b]{0.47\textwidth}            
            \includegraphics[width=0.88\textwidth]{cfes/cfe_figures/whr_recourse.png}
            \caption{low-level CFE}
            \label{fig:whr_ar_ap}
    \end{subfigure}%
    \vskip\baselineskip
    \begin{subfigure}[b]{0.42\textwidth}            
            \includegraphics[width=0.88\textwidth]{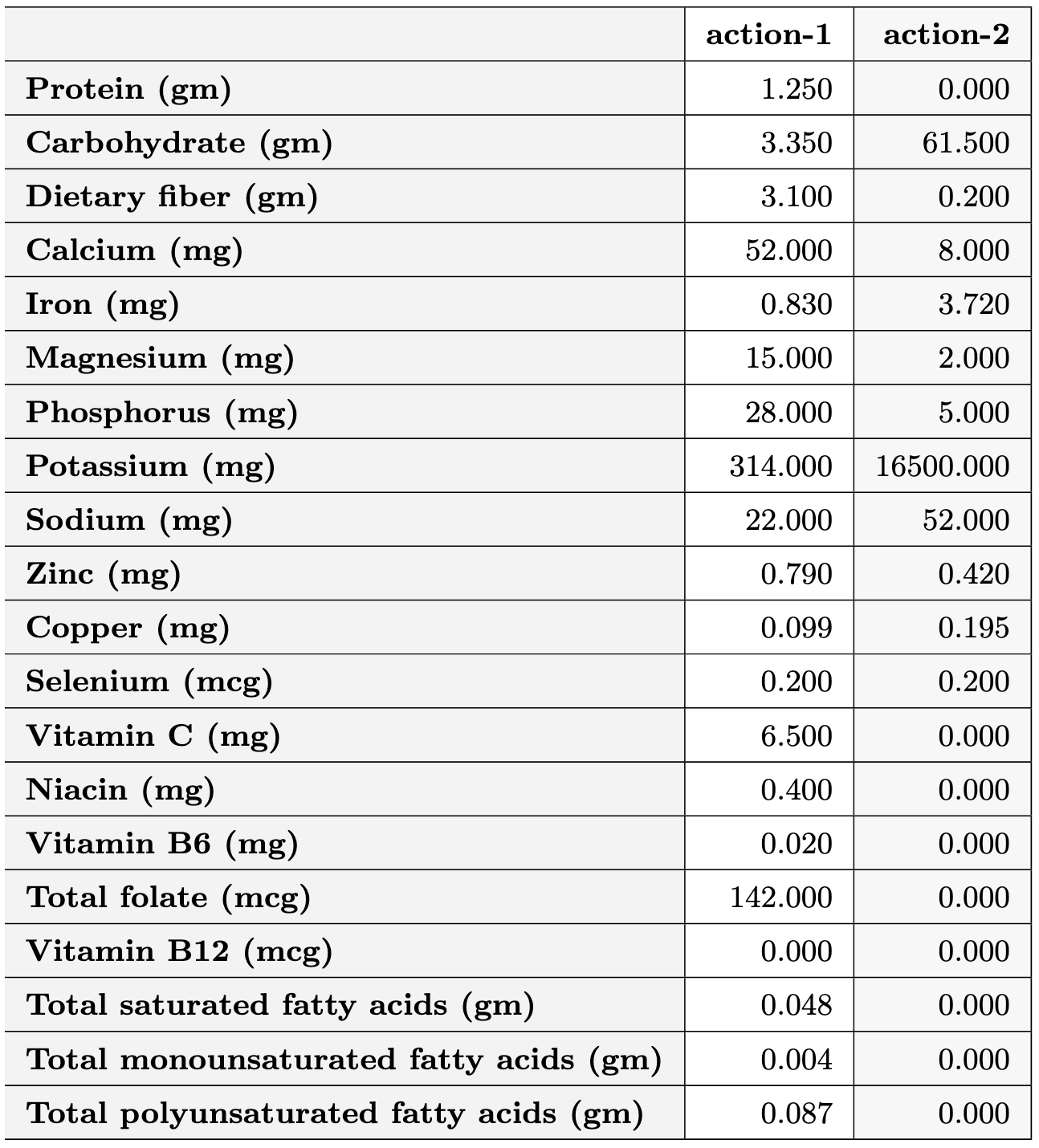}
            \caption{hl-continuous CFE with caloric costs}
            \label{fig:whr_caloric_cost}
    \end{subfigure}%
    ~
    \begin{subfigure}[b]{0.42\textwidth}
            \centering
            \includegraphics[width=0.88\textwidth]{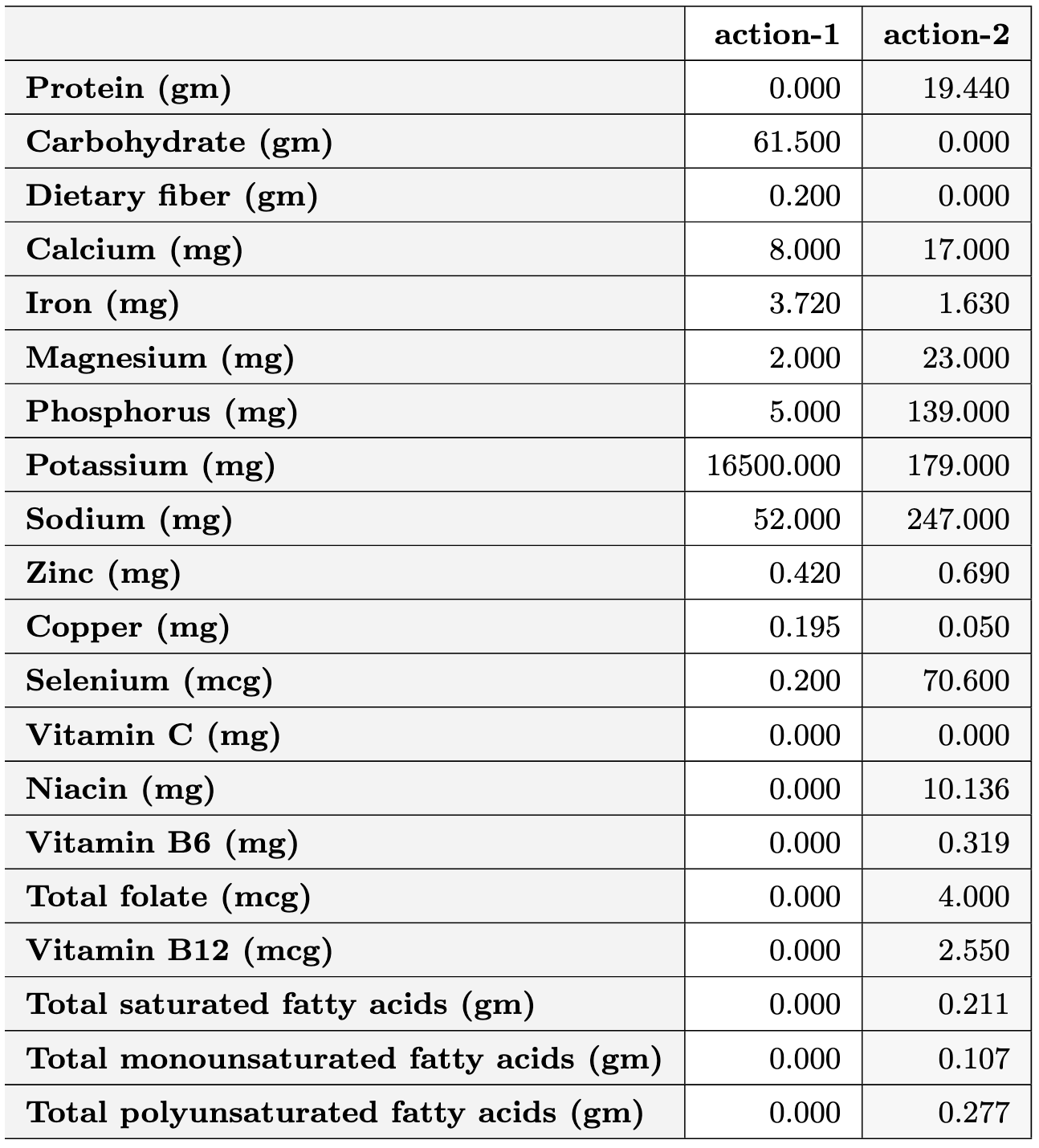}
            \caption{hl-continuous CFE with monetary costs}
            \label{fig:whr_monetary_cost}
    \end{subfigure}
    \caption[Low-level CFEs involve more actions but results low improvement]
    {When given actionable features values \([29.03, 109.45, 4.1, 309., 4.08, 96.,488., 994.,\) \( 1326., 2.61, 0.425, 45., 35.7, 8.755, 0.482,172.,1.21, 10.077, 12.392, 13.999]\), in the same order as shown in \subref{fig:whr_caloric_cost} and \subref{fig:whr_monetary_cost}, for a negatively classified WHR agent, the low-level CFE generator recommends a CFE \subref{fig:whr_ar_ap} with a cost of \(56.588\). This CFE was unique to the agent. In contrast, the hl-continuous CFE generator generates two CFEs optimized for different agent's preferences. When optimizing for caloric cost, the CFE generator generates CFE \subref{fig:whr_ar_ap} with a cost of \(2.750\). This CFE, which was also optimal for other \(25\) negatively classified agents, includes \textbf{action-1} (\textit{consume endive, raw}) and \textbf{action-2} (\textit{consume leavening agents: cream of tartar}). When optimizing for monetary cost, the CFE generator produces a CFE \subref{fig:whr_caloric_cost} of cost \(4.010\). This CFE, also optimal for other \(105\) agents, consists of \textbf{action-1} (\textit{consume leavening agents: cream of tartar}) and \textbf{action-2} (\textit{consume fish, tuna, light, canned in water, drained solids}).  Lastly, while the low-level CFE \subref{fig:whr_ar_ap} involves \(19\) actions but modifies \(19\) features and improve by \(5679.95\), the hl-continuous CFEs both involve \(2\) actions but modify \(19\) features and lead to an improvement of \(16815.04\) \subref{fig:whr_caloric_cost} and \(16682.62\) \subref{fig:whr_monetary_cost}. 
    }\label{fig:caloric_monetary}
\end{figure}

\paragraph{Varied preferences in nature of costs.}
\label{sec:cfe_app_var_costs}

We model two types of hl-continuous CFEs: a set of Foods+monetary costs hl-continuous actions and a set of Foods+caloric costs hl-continuous actions (see Appendix~\ref{subsec:cfe_bmi_whr_preproc} and \ref{subsec:cfe_app_hlc_cfe}). 
In a setting where negatively classified agents care more about monetary costs over caloric costs, and vice versa, the CFE generator adapts to these preferences and recommends the corresponding optimal CFE, as demonstrated in Figure~\ref{fig:caloric_monetary}.

Additionally, regardless of whether monetary or caloric costs were the desired costs by the agent, we consistently observed that taking hl-continuous CFEs involved fewer actions, resulted in more feature modifications and higher improvement when compared to low-level CFEs (Figures~\ref{fig:whr-calprice_actions_feats_proximity} and \ref{fig:caloric_monetary} ).
Future research could investigate the data-driven CFE generation at the intersection of various competing agent objectives, e.g., monetary and caloric costs.

\paragraph{Varied feature satisfiability.}
\label{sec:cfe_app_var_thresh_accuracy}

In general, as shown in Table~\ref{table:datasets_statistics}, compared to the unit threshold datasets: \(20\)- \(50\)- and \(100\)-dimensional agent–CFE datasets, agents in the varied binary feature satisfiability datasets described in Appendix~\ref{subsubsec:cfe_app_var_classifier_ds} required fewer action due to the fewer features to satisfy to get a desirable classification. 

Our results show that without explicit knowledge of the varied feature satisfiability when given testing set agents, the data-driven hl-discrete CFE generator trained on instances of a mixture of varied feature satisfiability agent–hl-discrete CFE  datasets successfully generates the right hl-discrete CFEs for the new agents. 
The data-driven hl-discrete CFE generator achieves an accuracy of \(99.683\%\) on \texttt{First10}, \(99.496\%\) on \texttt{Last10}, \(100\%\) on \texttt{First5}, \(100\%\) on \texttt{Mid5}, and \(100\%\) on \texttt{Last5}, dataset variants.

\begin{table}[b!]
    \begin{center}
    \footnotesize
    \renewcommand{\arraystretch}{1.1} 
    \setlength{\tabcolsep}{3pt} 
        \begin{tabular}{llllll}
            \toprule
            \multicolumn{2}{c}{Manual Groups} & & \multicolumn{2}{c}{Probabilistic Groups}\\
            \cmidrule(lr){1-2} \cmidrule(lr){3-5}
            Group & Accuracy  & & Group  &  Accuracy\\
            \midrule
            Group \(0\) & \(0.881\pm0.01200\)  & & Group \(0\) (\(0.4\)) & \(0.880\pm0.04400\)\\
            Group \(1\) & \(0.871\pm0.01260\)  & & Group \(1\) (\(0.5\)) & \(0.771\pm0.02081\)\\
            Group \(2\) & \(0.875\pm0.01249\)  & & Group \(2\) (\(0.6\)) & \(0.802\pm0.01571\)\\
            Group \(3\) & \(0.847\pm0.01359\)  & & Group \(3\) (\(0.7\)) & \(0.873\pm0.01241\)\\
            Group \(4\) & \(0.886\pm0.01212\)  & & Group \(4\) (\(0.8\)) & \(0.931\pm0.00947\)\\
            \bottomrule
        \end{tabular}
    \end{center}
    \caption[Group-wise accuracy of the data-driven hl-discrete CFE generator]
    {Group-wise accuracy of the data-driven hl-discrete CFE generator on \texttt{manual groups} \& \texttt{probabilistic groups} (see Appendix~\ref{subsubsec:cfe_app_var_actions_ds}). While the accuracy on the \texttt{manual groups} was within the same range (\(87\%\)), it greatly varied across the \texttt{probabilistic groups}.
    }\label{tab:varied_actions_access}
\end{table}

\paragraph{Varied access to actions.}
\label{sec:cfe_app_var_actions_accuracy}

The \texttt{manual groups} agent–hl-discrete CFE datasets (described in Appendix~\ref{subsubsec:cfe_app_var_actions_ds}) are more balanced in terms of the number of actions agents take (see Figure~\ref{fig:grp5_20}). The reason is agents have access to the same distribution of hl-discrete actions, i.e., although agents in each group have access to only a selected group of hl-discrete actions, all the hl-discrete actions for all groups were generated with the same probability, \(p_a = 0.5\).
However, for the \texttt{probabilistic groups} agent–hl-discrete CFEs datasets (described in Appendix~\ref{subsubsec:cfe_app_var_actions_ds}), Figure~\ref{fig:grp5_acts20} shows that as the probability of hl-discrete capabilities \(p_a\) decreases, the number of actions required to get all the necessary capabilities to transform their states to get a positive model outcome increases. In other words, agents in certain groups only have access to more expensive and limited hl-discrete actions compared to others. For instance, agents in the \texttt{probabilistic groups} Group \(0\) face more difficulty in achieving positive classification outcomes than those in the Group \(4\).

Since the agents in the \texttt{manual groups} agent–hl-discrete CFE datasets had more balanced access to hl-discrete actions as depicted in Figure~\ref{fig:grp5_20}, the data-driven hl-discrete CFE generators had almost similar accuracy (\({\sim}87\%\)) in the generation of CFEs across all agents in different \texttt{manual groups}, as shown in Table~\ref{tab:varied_actions_access} (\textbf{left}). 
On the other hand, since the agents in the \texttt{probabilistic groups} had access to varied hl-discrete actions, the accuracy of the data-driven hl-discrete CFE generator varied greatly across the groups, as shown in Table~\ref{tab:varied_actions_access} (\textbf{right}). For instance, as expected, the CFEs for \texttt{probabilistic groups} Group \(4\) agents with one-action hl-discrete CFEs were more accurately generated with an accuracy of \(93.06\%\) as compared to Group \(0\) and Group \(1\) agents, generated at an accuracy of \(88.04\%\) and \(77.09\%\), respectively.

\subsection{Accuracy of the Data-Driven CFE Generators}
\label{subsec:cfe_app_accurate_approximate}

Our results show that the data-driven CFE generators are accurate and confident information-specific CFE generators. Additionally, unlike low-level CFE generators that sometimes fail to produce a CFE entirely for an agent, our data-driven CFE generators generate approximately good CFEs instead of no CFEs at all. The supplemental results in this appendix subsection are mainly for the fully-synthetic datasets. 

\paragraph{Accuracy and confidence.} 
\label{sec:cfe_app_accuracy}

The proposed data-driven CFE generators are evidenced to perform strongly on the varied datasets. 
As shown in Figure~\ref{fig:20dim_var_info_access_accr}, on the \(20\)-dimensional \texttt{all} agent–CFE dataset variants, the CFE generators achieved high accuracy at generating hl-discrete CFEs, hl-discrete-id CFEs, and hl-discrete-id CFEs. All the generators perform best on the single-action CFE agents.
Furthermore, with strong confidence, i.e., low margin error rates (see Table~\ref{tab:var_dim_accuracy}), the proposed data-driven CFE generators performed well on all datasets regardless of the data dimension or frequency of CFEs. Notably, they excelled on high-frequency datasets, that is to say, \texttt{>40} datasets regardless of the data dimensions, as seen in Table~\ref{tab:var_dim_accuracy}.

\begin{figure}[b!]
    \centering
    \begin{tikzpicture}[scale=1.0]
    \pgfplotstableread[col sep=comma]{
    X, Y, perc
    0, 4691, 99.4
    1, 28, 0.6
    }\tAx 
    \pgfplotstableread[col sep=comma]{
    X, Y, perc
    0, 8645, 96.2
    1, 343, 3.8
    }\tBx 
    \pgfplotstableread[col sep=comma]{
    X, Y, perc
    0, 451, 87.1
    1, 67, 12.9
    }\tCx

    \pgfplotstableread[col sep=comma]{

    X, Y, perc
    0, 4510, 93.1
    1, 360 , 6.9
    }\tAxb 
    \pgfplotstableread[col sep=comma]{
    X, Y, perc
    0, 7435, 83.1
    1, 1066 , 16.9
    }\tBxb 
    \pgfplotstableread[col sep=comma]{
    X, Y, perc
    0, 261, 53.5
    1, 145, 46.5
    }\tCxb

    \pgfplotstableread[col sep=comma]{
    X, Y, perc
    0, 4465, 94.6
    1,254, 5.4
    }\tAxc 
    \pgfplotstableread[col sep=comma]{
    X, Y, perc
    0, 7153, 79.6
    1,1835,  20.4
    }\tBxc
    \pgfplotstableread[col sep=comma]{
    X, Y, perc
    0, 311, 60.0
    1,207, 40.0
    }\tCxc
    
    \begin{groupplot}[
        group style={
            group size=3 by 1,
            horizontal sep=0.2cm,
            x descriptions at=edge bottom,
            y descriptions at=edge left,
        },
        ybar,
        axis on top,
        height=5.3cm,
        width=6cm, 
        ybar=2pt,   
        enlarge y limits={value=.1,upper},
        ymin=1,    
        ymax=100000, 
        log origin=infty, 
        ymode=log, 
        axis x line*=bottom,
        axis y line*=left,
        y axis line style={opacity=0},
        tickwidth=1pt,
        enlarge x limits=0.5,
        ymajorgrids=true,
        major grid style={lightgray},
        legend style={
            at={(-0.5,1.2)},
            anchor=north,
            legend columns=3, 
            legend cell align=left,
            font=\footnotesize,
            fill opacity=2, 
            draw opacity=0.5,
        },
        ylabel=\textbf{Number of agents},
        xtick=data, 
        ticklabel style={/pgf/number format/.cd, use comma, 1000 sep = {}},
        nodes near coords,
        nodes near coords style={
                font=\scriptsize,
                rotate=90,
                anchor=west,
        },
        point meta=explicit symbolic,
        nodes near coords={\pgfmathprintnumber{\pgfplotspointmeta}\,\%},
        every axis legend/.append style={font=\footnotesize}, 
    ]
    
    \nextgroupplot[xlabel=(a) hl-discrete-id CFEs]
        \addplot[draw=none, postaction={pattern=north east lines}, fill={rgb,255:red,17;green,119;blue,51}]
            table[x=X,y=Y,meta=perc] {\tAx};
        \addplot[draw=none, postaction={pattern=dots}, fill={rgb,255:red,68;green,170;blue,153}]
            table[x=X,y=Y,meta=perc] {\tBx};
        \addplot[draw=none, fill={rgb,255:red,136;green,204;blue,238}]
            table[x=X,y=Y,meta=perc] {\tCx};
    
    \nextgroupplot[xlabel=(b) hl-discrete-named CFEs]
        \addplot[draw=none, postaction={pattern=north east lines}, fill={rgb,255:red,17;green,119;blue,51}]
            table[x=X,y=Y,meta=perc] {\tAxb};
        \addplot[draw=none, postaction={pattern=dots}, fill={rgb,255:red,68;green,170;blue,153}]
            table[x=X,y=Y,meta=perc] {\tBxb};
        \addplot[draw=none, fill={rgb,255:red,136;green,204;blue,238}]
            table[x=X,y=Y,meta=perc] {\tCxb};
            
    \nextgroupplot[xlabel=(c) hl-discrete CFEs]
        \addplot[draw=none, postaction={pattern=north east lines}, fill={rgb,255:red,17;green,119;blue,51}]
            table[x=X,y=Y,meta=perc] {\tAxc};
        \addplot[draw=none, postaction={pattern=dots}, fill={rgb,255:red,68;green,170;blue,153}]
            table[x=X,y=Y,meta=perc] {\tBxc};
        \addplot[draw=none, fill={rgb,255:red,136;green,204;blue,238}]
            table[x=X,y=Y,meta=perc] {\tCxc};
        \legend{1-action, 2-action, 3-action}
    
    \end{groupplot}
    \end{tikzpicture}

    \caption[Accuracy of the data-driven hl-id CFE generators]
    {The data-driven hl-id CFE generator for the (\textbf{a}) hl-discrete-id CFEs, the data-driven hl-continuous CFE generator for the (\textbf{b}) hl-discrete-named CFEs, and the data-driven hl-discrete CFE generator for the (\textbf{c}) hl-discrete CFEs, achieved strong performance on the \(20\)-dimensional \texttt{all} agent–hl-discrete CFE, varied information access, test datasets (new agents for the respective variants).
    }\label{fig:20dim_var_info_access_accr}
\end{figure}
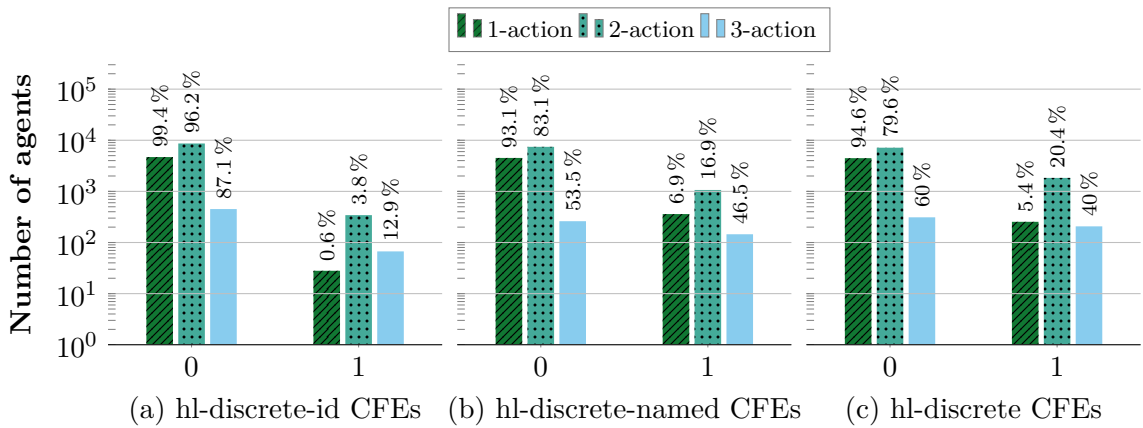
\begin{table}[ht!]
    \renewcommand{\arraystretch}{1}
    \begin{center}
        \begin{tabular}{lllll}
            & \multicolumn{4}{c}{\textbf{Performance of the data-driven CFE generators}} \\
            \cmidrule(lr){2-5}
            & & \texttt{all} & \texttt{>10}& \texttt{>40} \\
            \midrule
            & \(\hphantom{1}20\)-dimensional& \({0.969}\pm0.00284\) & \(0.984\pm0.00208\) & \(0.993\pm0.00141\) \\
            & \(\hphantom{1}50\)-dimensional & \({0.744}\pm0.00608\) & \(0.838\pm0.00534\) & \(0.915\pm0.00458\) \\
            & \(100\)-dimensional & \(0.354\pm0.00664\) & \(0.630\pm0.00778\) & \(0.856\pm0.00772\)\\
            \bottomrule
        \end{tabular}
    \end{center}
    \caption[Accuracy of the data-driven hl-discrete CFE generators]
    {A comparative analysis of the performance of the data-driven hl-discrete CFE generator across agent–hl-discrete CFE datasets with varied dimensions (\(20\)-, \(50\)- and \(100\)-dimensional) and varied frequency of the CFEs (\texttt{all}, \texttt{>10}, and \texttt{>40}). Results indicate that accuracy declines as dimensionality increases and CFE frequency decreases. Notably, the \(20\)-dimensional \texttt{>40} dataset, which has the lowest dimensionality and highest CFE frequency, achieved the highest accuracy.
    }\label{tab:var_dim_accuracy}
\end{table}

\begin{figure}[t!]
    \begin{center}
        \begin{subfigure}[t]{0.48\textwidth}
            \centering
            \includegraphics[width=0.77\textwidth]{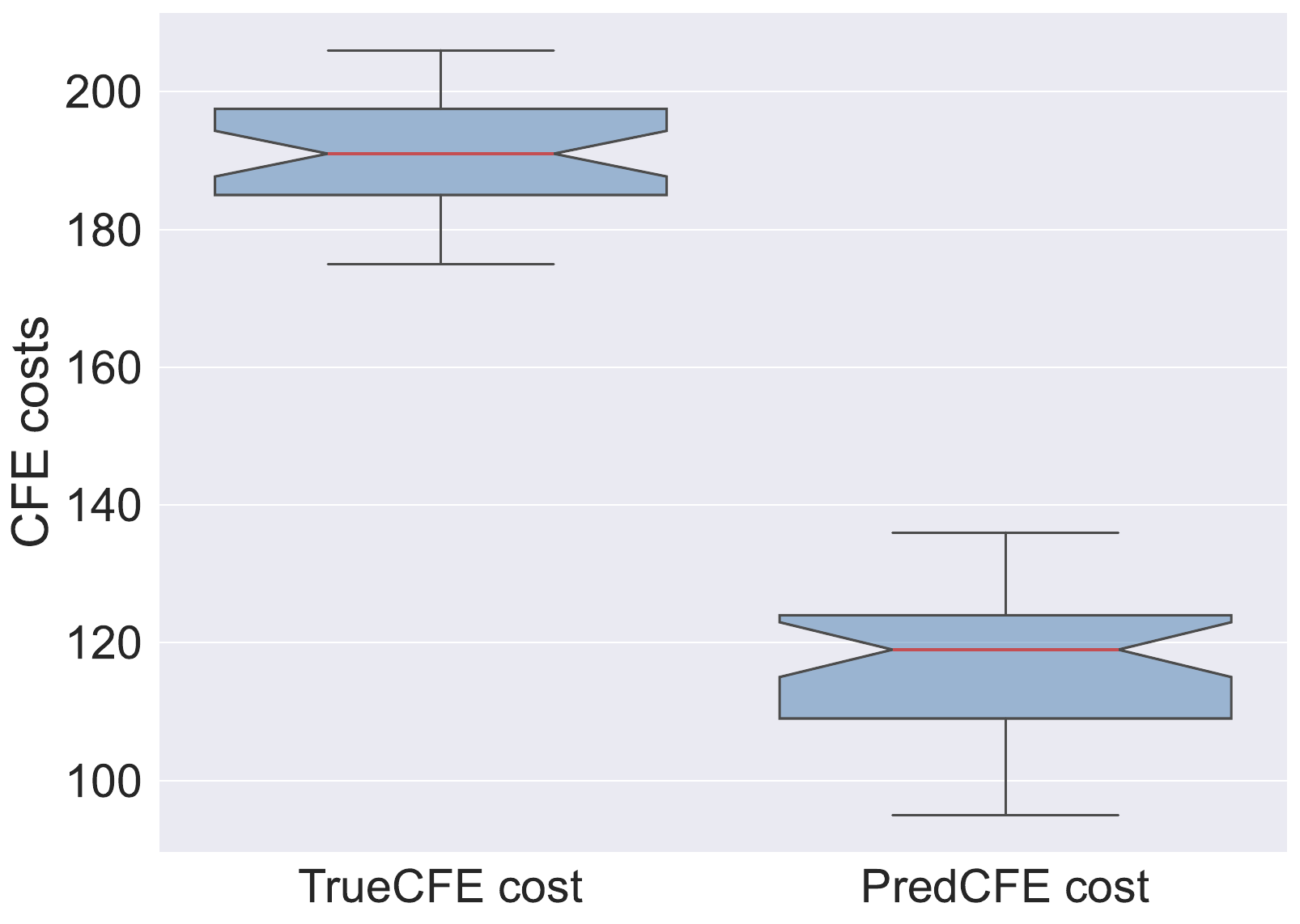}
            \caption{Invalid CFE mistakes}
            \label{fig:inval_mistakes}
        \end{subfigure}%
        \hfill
        \begin{subfigure}[t]{0.48\textwidth}
            \centering
            \includegraphics[width=0.77\textwidth]{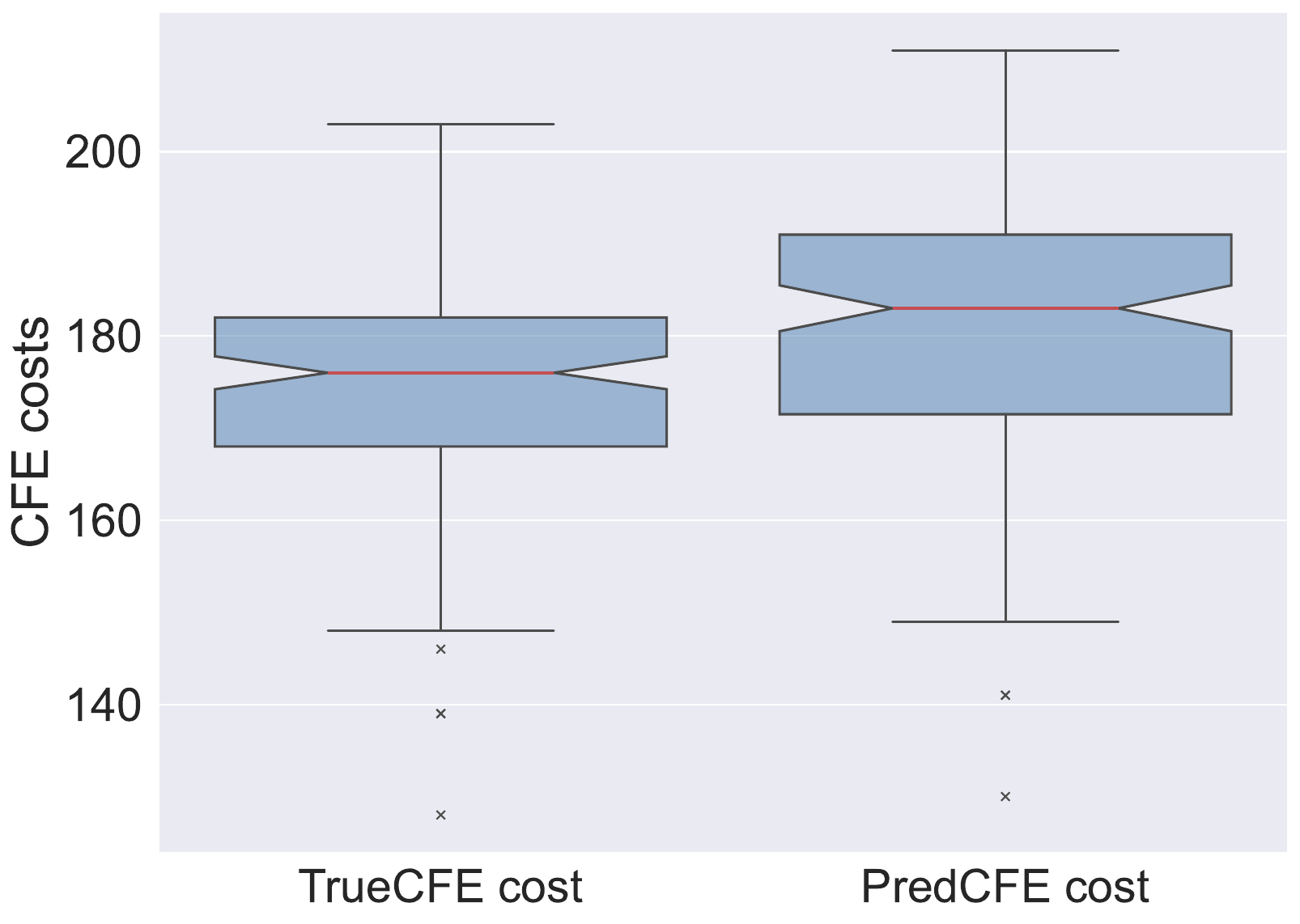}
            \caption{Valid CFE mistakes}  
            \label{fig:val_mistakes}
        \end{subfigure}
        \caption[Data-driven CFE generators produce valid CFE mistakes when suboptimal]
        {A generated CFE is a mistake if the CFE doesn't match the true CFE. A valid CFE mistake transforms the agent's initial state to get a desirable model outcome. An invalid CFE mistake does not favorably transform the agent state.
        Distribution of costs of generated and true CFEs for \subref{fig:inval_mistakes} invalid and \subref{fig:val_mistakes} valid CFE mistakes the data-driven hl-id CFE generator makes on \(20\)-dimensional \texttt{all} agent–hl-discrete-id dataset.  Valid CFE mistakes are, by definition, more expensive than the true CFEs, while invalid CFE mistakes are cheaper than the true CFEs. 
        }\label{fig:val_inval_mistakes}
    \end{center}
\end{figure}

\paragraph{Approximation.}
Unlike ILP-based low-level CFE generators, which do not generate CFEs for agents when the ILP solution is sub-optimal or infeasible, our data-driven CFE generators alternatively produce valid CFE mistakes when suboptimal (see Figure~\ref{fig:val_inval_mistakes}). 
For example, of the \(1.58\%,16.23\%\) and \(37.00\%\) mistakes the hl-id generator makes on the \(20\)-, \(50\)-, and \(100\)-dimensional \texttt{>10} agent–hl-discrete-id CFE datasets,  \(100\%, 99.23\%,\) and \(87.29\%\), respectively, were valid CFE mistakes. Similarly, the majority of the mistakes of the hl-discrete CFE generators were valid, e.g., on the \(20\)-dimensional \texttt{>10} agent–hl-discrete CFE dataset, of the \(10.8\%\) mistakes the generator makes, \(63.10\%\) were valid. 

Additionally, the likelihood of the ILP-based low-level CFE generator's failure at generating CFEs (i.e., returns no CFEs) increases with the number of actionable features (data dimensions). 
On the hand, the percentage of valid mistakes from our proposed CFE generators decreases with the frequency of CFEs in the agent–CFE training set, e.g., the percentage of valid mistakes is \(87.29\%\) on the \texttt{>10} dataset and \(57.83\%\) on the \(100\)-dimensional \texttt{all} dataset.

\subsection{Scalability and Interpretability of the Data-Driven CFE Generators}
\label{sec:cfe_app_scalable_interpretable}

Our results show that data-driven CFE generators, including hl-continuous, hl-discrete, and hl-id, are more scalable than low-level CFE generators. In addition, the costs and actions of hl-continuous and hl-discrete CFEs are interpretable and transparent, simplifying validation and comparison.

\begin{figure}[t!]
    \centering
    \includegraphics[width=\textwidth]{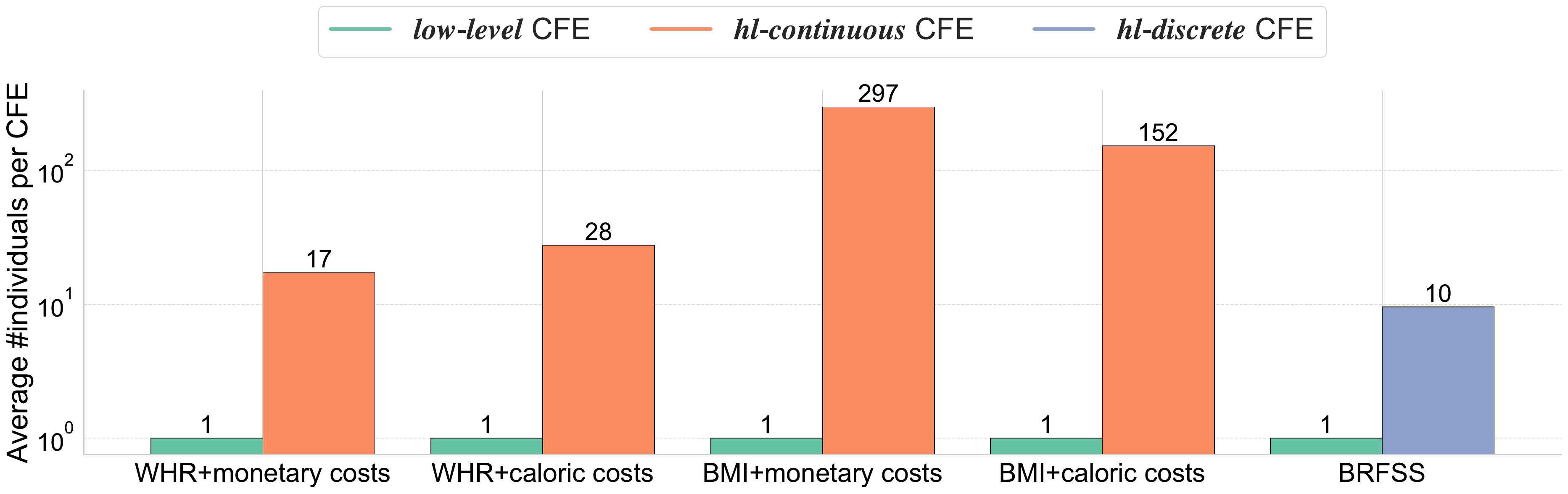}
    \caption[The average number of agents with the same optimal hl-continuous CFE]
    {The average number of agents with the same CFE when agents take hl-continuous CFEs (as a set of Food+monetary and Foods+caloric costs hl-continuous actions) for WHR and BMI datasets, as well as hl-discrete CFEs on the BRFSS dataset, compared to when they take low-level CFEs for the respective datasets. Regardless of the dataset considered, on average, while low-level CFEs were unique to a given agent (individual), hl-continuous and hl-discrete CFEs were simultaneously optimal to multiple agents.
    }\label{fig:ap_scalability}
\end{figure}

\paragraph{Scalability.}
\label{sec:cfe_actual_features_changed}
Unlike the overly specific feature-based actions in low-level CFEs (see Figure~\ref{fig:whr_ar_ap}), actions in hl-continuous and hl-discrete CFEs are more general, increasing the likelihood of their optimality for agents with closely similar profiles. As shown in Figure~\ref{fig:ap_scalability}, while low-level CFEs were typically unique to each agent, hl-continuous and hl-discrete CFEs were often simultaneously optimal for multiple agents (see also Figure~\ref{fig:caloric_monetary}).
Furthermore, while our data-driven CFE generators efficiently produce CFEs for new agents without requiring re-optimization, low-level CFE generators operate on a single-agent basis, making them significantly more resource-intensive.

\paragraph{Interpretability.}
\label{sec:cfe_actual_interpret}
The hl-continuous and hl-discrete CFEs comprise real-world-like actions (see Figures~\ref{fig:whr_caloric_cost} and \ref{fig:whr_monetary_cost}), in contrast to low-level CFEs, which rely on overly specific, feature-based actions (see Figure~\ref{fig:whr_ar_ap}).
Consequently, hl-continuous and hl-discrete CFEs are more intuitive, interpretable, and easier to execute and compare since they align more closely with executable actions.
Moreover, the costs associated with these actions are more transparent and easier to comprehend, given a general understanding of how they were derived, an essential factor in ensuring clarity and trust in following the recommended CFE.

Looking ahead, conducting a comprehensive user study would be a valuable avenue for future work. The human-subject study could empirically validate the transparency and interpretability of the proposed high-level CFEs, serving as a test of the underlying hypotheses.

\subsection{Performance under various Information Access Constraints}
\label{sec:cfe_app_var_info_accuracy}

In addition to other information access constraints, we investigate the effectiveness of the data-driven CFE generators under two more information access constraints. From the original agent–hl-discrete CFE datasets, we created two more information access variants, the agent–hl-discrete-named CFE dataset, and agent–hl-discrete-id CFE dataset as described in Appendix~\ref{subsubsec:cfe_app_var_info_ds}. Given the agent–hl-discrete CFE information access datasets, we use the data-driven hl-discrete CFE generators for the hl-discrete CFEs, hl-continuous CFE generators for hl-discrete-named CFEs, and data-driven hl-id CFE generators for hl-discrete-id CFEs. 

In general, all the data-driven CFE generators, regardless of information access constraints described in Appendix~\ref{subsubsec:cfe_app_var_info_ds}, generate single-action CFEs more accurately than multi-action CFEs. 
For example, the hl-discrete CFE generator, as seen in Figure~\ref{fig:20dim_var_info_access_accr}(\textbf{c}), generates one-action CFEs at an accuracy of \(94.6\%\), two-action CFEs at an accuracy of \(79.6\%\), and three-action CFEs at an accuracy of \(60.0\%\).

However, in general, data-driven hl-id CFE generators were shown in  Figure~\ref{fig:20dim_var_info_access_accr}(\textbf{a}) to \ref{fig:20dim_var_info_access_accr}(\textbf{c}) and Table~\ref{tab:all_results_20} to be more accurate and need less CFE frequency in the training set than the hl-continuous and hl-discrete CFE generators. 
For example, on the \(20\)-dimensional \texttt{all} dataset, the data-driven hl-id CFE generator had an accuracy of \(96.9\%\), compared to \(85.4\%\) with hl-continuous CFE generator and \(83.9\%\) with hl-discrete CFE generator.
\begin{table}[t!]
    \renewcommand{\arraystretch}{1.0} 
    \setlength{\tabcolsep}{3pt} 
    \begin{center}
    \begin{tabular}{lllll}
        & \multicolumn{4}{c}{\textbf{Performance on \(20\)-dimensional datasets}} \\
        \cmidrule(lr){2-5}
        & \texttt{all} & \texttt{>10}& \texttt{>40} & \\
        \midrule
        Data-driven hl-id CFE generator & \(0.969\pm0.00284\) & \(0.984\pm0.00208\) & \(0.993\pm0.00141\)\\
        Data-driven hl-continuous CFE generator & \(0.854\pm0.00581\) & \(0.886\pm0.00531\) & \(0.940\pm0.00411\)\\
        Data-driven hl-discrete CFE generator & \(0.839\pm0.00605\) & \(0.892\pm0.00518\) & \(0.937\pm0.00420\)\\
        \bottomrule
    \end{tabular}
    \end{center}
    \caption[Accuracy of the CFE generators on varied frequency CFEs agent–CFE datasets]
    {Accuracy of the CFE generators on \(20\)-dimensional datasets: \texttt{all}, \texttt{>10}, and \texttt{>40}. All the data-driven CFE generators demonstrate consistently high accuracy, with particularly strongest performance on high CFE frequency datasets (\texttt{>40}).
    }\label{tab:all_results_20}
\end{table}

\subsection{Challenges and Proposed Solutions}
\label{subsec:cfe_potential_challenges}

We identify several challenges in designing data-driven CFE generators: the infrequent occurrence of CFEs, the large number of actionable features, and the significant dependence on the complexity of the CFE generator model.
In this work, we thoroughly examine these challenges, propose plausible solutions, and suggest avenues for future research.

\paragraph{Negatively affected by high number of actionable features.}
\label{sec:cfe_app_var_dim_accuracy}

As the number of actionable features increases, agents CFEs have more actions (see Table~\ref{table:datasets_statistics}). For instance, in the \(100\)-dimensional agent–hl-discrete CFE dataset, \(54.4\%\) of agents' hl-discrete CFEs had three actions, and in the \(20\)-dimensional dataset, \(33.3\%\) of agents needed only one action in their CFE and only \(3.6\%\) had three.

Beyond an increase in the number of actions in the CFEs, the uniqueness of CFEs also rises as the number of actionable features grows. The average frequency of CFEs in the \texttt{all} agent–hl-discrete CFE training set dropped from \(46.64\%\) in the \(20\)-dimensional dataset to \(21.75\%\) in the \(50\)-dimensional dataset and further to \(8.09\%\) in the \(100\)-dimensional dataset. Additionally, \(18.115\%\), \(20.797\%\), and \(31.072\%\) of CFEs in the \(20\)-, \(50\)-, and \(100\)-dimensional \texttt{all} datasets, respectively, were unique (i.e., optimal for only one agent).
This low frequency of CFEs in the agent–CFE datasets led to discrepancies after train/test splits, where some CFEs appeared in one split but not the other. Specifically, for the \(20\)-, \(50\)-, and \(100\)-dimensional datasets, there were \(52\), \(154\), and \(708\) unique CFEs in the testing set that were absent from the training set.

As a result, data-driven CFE generators become less accurate as the number of actionable features increases. As shown in Table~\ref{tab:var_dim_accuracy}, the data-driven hl-id CFE generator consistently performed worse on higher-dimensional datasets. For example, while it achieved \(96.9\%\) accuracy on the \(20\)-dimensional \texttt{all} dataset, its accuracy dropped to \(74.4\%\) on the \(50\)-dimensional \texttt{all} dataset and was lowest on the \(100\)-dimensional \texttt{all} dataset.

\paragraph{Negatively affected by low frequency of CFEs.}
\label{sec:cfe_app_var_freq_accuracy}

We created the varied frequency of CFEs agent–CFE datasets: \texttt{all}, \texttt{>10}, and \texttt{>40} (see Appendix~\ref{subsubsec:cfe_app_var_fre_ds}), to examine how CFE frequency in the agent–hl-discrete CFE dataset impacts the robustness of data-driven CFE generators. After the train/test split, the \texttt{>40} dataset ensured that at least 20 agents shared the same CFE in the training set.
By definition, the \texttt{>40} dataset had the highest CFE frequency, while \texttt{all} had the lowest. This frequency also varied with data dimensionality, as illustrated in Appendix~\ref{sec:cfe_app_var_dim_accuracy}.
\begin{figure}[t!]
\definecolor{c1}{HTML}{FFC20A}
\definecolor{c2}{HTML}{0C7BDC}
    \centering
    \begin{tikzpicture}[scale=0.4]
      \begin{axis}[ 
        axis on top,
        height=10cm, 
        width=10cm, 
        axis x line*=bottom,
        axis y line*=left,
        y axis line style={opacity=0.9},
        x axis line style={opacity=0.9},
        width=\linewidth,
        line width=1,
        tickwidth=1pt,
        font=\Large,
        enlarge x limits=0.2,
        ymajorgrids=true,
        major grid style={lightgray},
        grid style={white},
        xlabel={{Varied frequency of CFEs in the training agent–CFE dataset}},
        ylabel={{Accuracy}},
        y tick label style={
            /pgf/number format/.cd,
            fixed,
            fixed zerofill,
            precision=0
        },
        tickwidth=1pt,
        legend style={
            at={(0.5,1.2)},
            anchor=north,
            legend columns=2, 
            legend cell align=left,
            font=\Large,
            fill opacity=1, 
            draw opacity=0.5,
          },
        ymin = 30,
        ymax = 100,
        xtick=data,
        symbolic x coords={\texttt{>40}, \texttt{>100}, \texttt{>150}, \texttt{>200}}, 
        every axis legend/.append style={font=\Large}
      ]
        \addplot[c1, mark=square*] table
          {%
            X Y
            \texttt{>40}  44.429
            \texttt{>100} 76.550
            \texttt{>150} 87.226
            \texttt{>200} 94.350
          };
          \addlegendentry{\(100\)-dimensional}
    
        \addplot[c2, mark=square*] table
         {%
            X Y
            \texttt{>40}  67.206
            \texttt{>100} 91.545
            \texttt{>150} 95.631
            \texttt{>200} 97.579	
          };
        \addlegendentry{\Large\(50\)-dimensional}   
      \end{axis}
    \end{tikzpicture}
    \caption[Impact of CFE frequency in agent–CFE training datasets]
    {Impact of CFE frequency in agent–CFE training datasets for \(50\) and \(100\)-dimensional datasets on the accuracy of the data-driven CFE generators. As the frequency of CFEs (the number of agents sharing the same optimal CFE) increases, the accuracy of data-driven CFE generators improves. This improvement occurs more rapidly in the lower-dimensional (\(50\)-dimensional) dataset.
    }\label{fig:na_100_50_freq}
\end{figure}
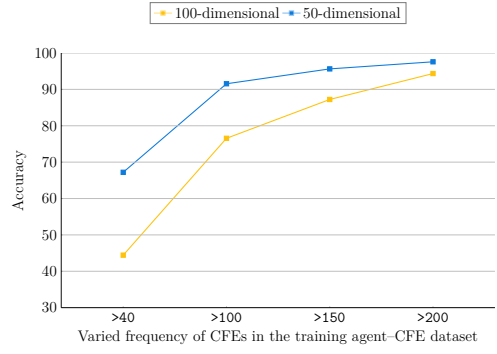

The low frequency of CFEs in the agent–hl-discrete CFE training sets negatively impacted CFE generation across all datasets, regardless of data dimensionality. However, this effect became more pronounced as data dimensions increased. For instance, as shown in Table~\ref{tab:var_dim_accuracy}, the accuracy of CFE generators on the \(20\)-dimensional dataset was highest when CFEs had a frequency of at least \(20\) in the training set (\texttt{>40}) and lowest on the \texttt{all} dataset, where some CFEs appeared in the test set but not in the training set. Specifically, the data-driven hl-id CFE generator achieved an accuracy of \(99.3\%\) on the \(20\)-dimensional \texttt{>40} dataset, compared to \(96.9\%\) on the \(20\)-dimensional \texttt{all} dataset.
In contrast, CFE generation accuracy on the \(20\)-dimensional dataset was significantly higher than on the \(100\)-dimensional dataset. This difference highlights that the negative impact of low CFE frequency in the training set becomes more severe as data dimensionality increases.

Additionally, the minimum frequency of CFEs required for a strong CFE generator increases with the number of actionable features. While the frequency of at least \(20\) in the training set ensured an accuracy of \(99.3\%\) of the CFE generator on the \(20\)-dimensional dataset (see Table~\ref{tab:var_dim_accuracy}), a higher frequency is needed for the \(50\)- and \(100\)-dimensional datasets (see Table~\ref{tab:var_dim_accuracy} and Figure~\ref{fig:na_100_50_freq}). Below is a proposed solution.

\paragraph{}
\makebox[\textwidth][c]{%
\begin{minipage}{0.9\textwidth} 
\begin{algorithm}[H]
    \SetAlgoNoLine
    \caption{The agent–hl-discrete CFE dataset augmentation}
    \label{alg:data_aug}
    \SetAlgoNlRelativeSize{0}
    \SetNlSty{}{}{:}
    \KwInput{an agent \(\mathbf{x}\) and their hl-discrete CFE \(I\), and the threshold classifier \(\mathbf{t}\)}
    \KwOutput{valid derived augmentations of agent \(\mathbf{x}\), \(\mathbf{x}_\textrm{augs}\) with the same CFE}
    \KwData{indices of features \(ids\) where the hl-discrete CFE when taken, adds more than needed capabilities to \(\mathbf{x}\)}

    \(\textrm{augs} \leftarrow 2^{|ids|}\) possible worse-off agents\;
    
    \ForEach{\(\mathbf{aug}\) \textbf{in} \(\textrm{augs}\)}{
        \If{\(\mathbf{aug}\) \textbf{is valid}}{
            \(\mathbf{x}_\textrm{augs} \leftarrow \mathbf{x}_\textrm{augs} \cup \{\mathbf{aug}\}\)\;
        }     
    } 

\end{algorithm}
\end{minipage}
}
\begin{table}[htb!]
    \begin{center}
    \begin{tabular}{lllll}
            & \multicolumn{3}{c}{\textbf{Effect of data augmentation}} \\
            \cmidrule(lr){2-4}
            & \(20\)-dimensional & \(50\)-dimensional & \(100\)-dimensional & \\
            \midrule
            Before data augmentation  & \(0.969\pm0.00284\) & \(0.744\pm0.00608\)   & \(0.354\pm0.00664\) \\
            After \texttt{AG1} & \(0.965\pm0.00303\) & \(0.760\pm0.00595\)  & \(0.505\pm0.00694\) \\
            After \texttt{AG2} & \(0.982\pm0.00218\) & \(0.845\pm0.00504\)   & \(0.790\pm0.00565\) \\
            \bottomrule
    \end{tabular}
    \end{center}
    \caption[Effects of \texttt{AG1} and \texttt{AG2} augmentation on the accuracy]
    {Effects of \texttt{AG1} and \texttt{AG2} augmentation on the accuracy of the data-driven CFE Generator. Data augmentation mitigates the negative impact of low frequency of CFEs and improves the accuracy of data-driven CFE generators on the \(20\)-, \(50\)-, and \(100\)-dimensional: \texttt{all} datasets. 
    }\label{tab:before_after_aug}
\end{table}

\begin{itemize}
    \item We examine the impact of increasing the frequency of CFEs through data augmentation (Algorithm~\ref{alg:data_aug}) on the performance of the data-driven hl-discrete CFE generator. 
    Algorithm~\ref{alg:data_aug} is specific for the agent–hl-discrete CFE datasets with all kinds of threshold classifiers. To generate new agents for which a given hl-discrete CFE is the most optimal, we ensure that no other hl-discrete CFE within the complete set of CFEs can achieve the transformation at a lower cost.
    
    Therefore, given an agent state, we find all possible worse-off agent states, such that the current optimal hl-discrete CFE is still the best CFE for the worse-off agent states.  Worse-off agent states are those such that the features where the hl-discrete CFE adds more capabilities than required to transform the agent state favorably are made worse, i.e., for \(i\) such that \mbox{\({x}^{\star}_{i} > t_{i}, {aug}_{i} < x_{i}\)}.
    Specific to the threshold classifier we use in the experiments, a hl-discrete CFE is adding more capabilities than required to feature \(i\) of \(\mathbf{x}\), if by after the action, the transformed feature \({x}^{\star}_{i}\) is such that \({x}^{\star}_{i} > t_{i}\).  
    The derived worse-off agent state \(\mathbf{aug}\) is valid if \(\mathbf{x}\)'s hl-discrete CFE is also the optimal CFE.

    \item Using Algorithm~\ref{alg:data_aug}, we augment the agent–hl-discrete CFE training set to ensure that each CFE appears at least twice (\texttt{AG1}) and to increase the frequency of CFEs with fewer than 20 occurrences (\texttt{AG2}). As a result, the number of hl-discrete CFEs with fewer than 20 agents significantly decreased from \(813\) to \(638\), \(2676\) to \(2005\), and \(9043\) to \(7144\) for the \(20\)- \(50\)- and \(100\)-dimensional datasets, respectively.

    Experimental results show an improvement in the accuracy of the CFE generators on the test samples after data augmentation, e.g., on the \(100\)-dimensional dataset, the accuracy of the data-driven CFE generator increases from \(35.37\%\) before data augmentation to \(50.54\) after \texttt{AG1}, and \(78.99\%\) after \texttt{AG2} (see Table~\ref{tab:before_after_aug}).  The findings show that data augmentation can improve the robustness of CFE generators in cases where the frequency of CFEs in the agent–CFE training datasets is low.

\end{itemize}

\paragraph{Performance heavily depends on complexity of the CFE generator model.}
\label{sec:cfe_app_sophi_cfegen_accuracy}

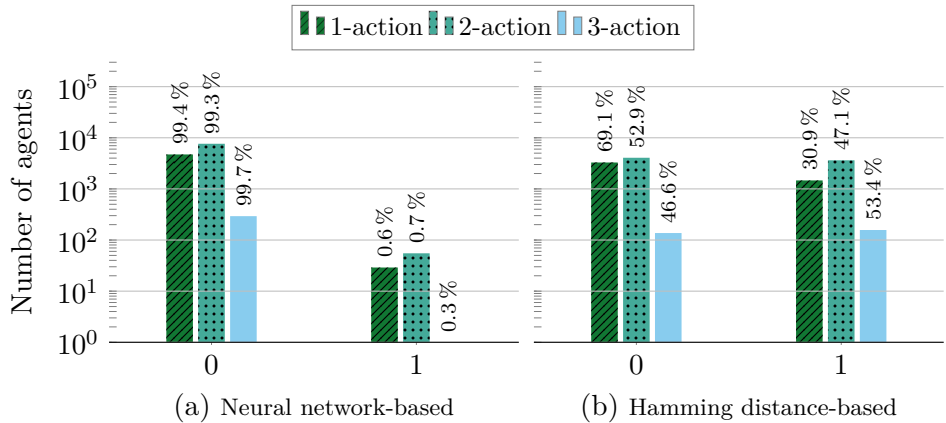
\begin{figure}[ht!]
    \centering
 
    \begin{tikzpicture}[scale=1]
    \pgfplotstableread[col sep=comma]{
    X, Y, perc
    0, 4719, 99.4
    1,29, 0.6
    }\tAyb
    \pgfplotstableread[col sep=comma]{
    X, Y, perc
    0, 7637, 99.3
    1,55, 0.7
    }\tByb
    \pgfplotstableread[col sep=comma]{
    X, Y, perc
    0, 293, 99.7
    1,1, 0.3
    }\tCyb

    \pgfplotstableread[col sep=comma]{
    X, Y, perc
    0, 3283, 69.1
    1,1465, 30.9
    }\tAyc
    \pgfplotstableread[col sep=comma]{
    X, Y, perc
    0, 4066, 52.9
    1,3626,  47.1
    }\tByc
    \pgfplotstableread[col sep=comma]{
    X, Y, perc
    0, 137, 46.6
    1,157, 53.4
    }\tCyc
    
    \begin{groupplot}[
        group style={
            group size=3 by 1,
            horizontal sep=0.2cm,
            x descriptions at=edge bottom,
            y descriptions at=edge left,
        },
        ybar,
        axis on top,
        height=5.3cm, 
        width=7cm, 
        ybar=2pt,  
        enlarge y limits={value=.1,upper},
        ymin=1,    
        ymax=100000,
        log origin=infty,
        ymode=log,
        axis x line*=bottom,
        axis y line*=left,
        y axis line style={opacity=0},
        tickwidth=1pt,
        enlarge x limits=0.5,
        ymajorgrids=true,
        major grid style={lightgray},
        legend style={
            at={(-0.1,1.2)},
            anchor=north,
            legend columns=3, 
            legend cell align=left,
            font=\tiny,
            fill opacity=1, 
            draw opacity=0.5,
        },
        ylabel=Number of agents,
        xtick=data, 
        ticklabel style={/pgf/number format/.cd, use comma, 1000 sep = {}}, 
        nodes near coords,
        nodes near coords style={
                font=\scriptsize,
                rotate=90,
                anchor=west,
        },
        point meta=explicit symbolic,
        nodes near coords={\pgfmathprintnumber{\pgfplotspointmeta}\,\%},
        every axis legend/.append style={font=\small}, 
    ]

    \nextgroupplot[xlabel=(a) \footnotesize{Neural network-based}]
        \addplot[draw=none, postaction={pattern=north east lines}, fill={rgb,255:red,17;green,119;blue,51}]
            table[x=X,y=Y,meta=perc] {\tAyb};
        \addplot[draw=none, postaction={pattern=dots}, fill={rgb,255:red,68;green,170;blue,153}]
            table[x=X,y=Y,meta=perc] {\tByb};
        \addplot[draw=none, fill={rgb,255:red,136;green,204;blue,238}]
            table[x=X,y=Y,meta=perc] {\tCyb};
            
    \nextgroupplot[xlabel=(b) \footnotesize{Hamming distance-based}]
        \addplot[draw=none, postaction={pattern=north east lines}, fill={rgb,255:red,17;green,119;blue,51}]
            table[x=X,y=Y,meta=perc] {\tAyc};
        \addplot[draw=none, postaction={pattern=dots}, fill={rgb,255:red,68;green,170;blue,153}]
            table[x=X,y=Y,meta=perc] {\tByc};
        \addplot[draw=none, fill={rgb,255:red,136;green,204;blue,238}]
            table[x=X,y=Y,meta=perc] {\tCyc};
    \legend{1-action, 2-action, 3-action}
    
    \end{groupplot}
    \end{tikzpicture}

    \caption[Comparison of accuracy of two data-driven hl-discrete CFE generators]
    {A comparison of accuracy of two data-driven hl-discrete CFE generators on the \(20\)-dimensional \texttt{>40} dataset. The neural network model consistently performs better than the hamming distance model.
    }\label{fig:gens_comp}
\end{figure}

Given the agent–hl-discrete-id CFE \(20\)-dimensional, \texttt{>40} dataset variant, we compare the effectiveness of the neural network-based CFE generator against the Hamming distance-based CFE generator. 
As shown in Figure~\ref{fig:gens_comp}, the neural network-based CFE generator demonstrates greater accuracy in generating CFEs for new agents. Interesting for future works is an exploration of the effectiveness of CFE generators based on more advanced and alternative methods, e.g.,  multi-chain neural networks, reinforcement learning, and transformer models.
\chapter{Selectively Revealing Positive and Negative Role Models to Help People Make Good Decisions}
\label{app:revealrm}
\section{Additional Motivating Scenarios}
\label{sec:revealrm_app_prac-examples}

\paragraph{Training videos.} 
Consider a company responsible for creating workplace training videos for client firms with the aim of promoting appropriate professional conduct, such as handling sensitive or confidential information.\footnote{For example, Vector Solutions: \url{https://www.vectorsolutions.com}} In this setting, the video producer acts as the social planner, and the agents are the employees who watch the training videos and subsequently make decisions in workplace situations. For simplicity, assume that the social planner distributes the same standardized training module to all agents, even though their roles and day-to-day environments may differ slightly. For example, a company's level 4 and 5 employees might receive the same training video on workplace professional conduct. Employees operate in diverse roles and environments, and rely on exemplars such as colleagues when deciding how to act in various unfamiliar workplace situations. 
Because the social planner can produce only a limited number of dramatized scenarios, they must select those that improve decision-making across a large and heterogeneous workforce. Positive scenarios, such as reporting a suspicious email to technical support, demonstrate desirable conduct. Negative scenarios, such as the consequences of sharing confidential information with a fraudulent sender, discourage similar agents from making similar mistakes. When employees encounter a situation that closely matches one portrayed positively in the training, they follow the demonstrated action. When a similar situation is portrayed negatively, they avoid the depicted choice and instead look to alternative exemplars for guidance. If employees encounter a situation not covered in the training, they select a local exemplar uniformly at random, for instance by imitating how a colleague handled a comparable circumstance.

\paragraph{Misinformation.} 
Consider a setting in which each social media user has a fixed set of news outlets they follow, but they don’t know which are legitimate (``good'')  and which are  fake (``bad''). The social media manager seeks to reduce misinformation by increasing the expected number of users who ultimately rely on legitimate outlets, subject to a limited budget for verification labels. To achieve this, the manager assigns visible markers to a selected subset of outlets, indicating whether they are legitimate or fake. If a user observes that at least one of the outlets they follow is labeled as legitimate, they will choose to rely on it for news. However, if the only labeled outlet(s) in their collection are marked as fake, the user remains uncertain about the remaining outlets and chooses one at random from the remaining ones to rely on.

\paragraph{High school career day.} 
Consider a high school that runs a career day program to help senior students make informed decisions about their future careers. In practice, however, students often rely on role models from their own social circles (e.g., family members or community figures) without knowing whether emulating them will lead to positive or negative outcomes. The school aims to steer students toward desirable careers, but can only feature a limited number of role models, which makes the choice of whom to highlight especially important. Featured speakers might include positive examples, such as a physician who motivates students to pursue a career in medicine, as well as cautionary ones, like a former gang member whose story illustrates the long-term negative consequences of criminal involvement. After the event, students emulate a role model from their social circle: avoiding those identified as negative, following a role model identified as positive if any were revealed, and otherwise choosing randomly.

\section{Missing Proofs for Section~\ref{sec:revealrm_model_prelim}}
\label{sec:revealrm_app_preliminaries}

\subsection{Proof of Proposition~\ref{prop:monotone}}
\label{sec:revealrm_app-prelimmonotonicty}

\begin{proof}[Proof of Proposition~\ref{prop:monotone}]
    Fix an initial revealed target set $A \subseteq \Ts$ and let $t \in \Ts \setminus A$. 
    Compare the social welfare under $A$ with that under $A \cup \{t\}$. Any agent that already selects a positive target with probability $1$ under $A$ continues to do so when $t$ is revealed. Agents for whom $t$ provides the first revealed positive target neighbor gain probability of $1$ for choosing a positive target to emulate, and no agent's probability decreases. 
    
    Therefore, the total probability of choosing a target node under $A \cup \{t\}$ is always at least as large as the total probability under $A$ in every realization of the random process. The same conclusion holds in expectation $F(A \cup \{t\}) \ge F(A)$. This proves that the social welfare function is monotonically increasing since expanding the revealed target set never lowers the probability that agents select positive targets to emulate.
\end{proof}

\subsection{Proof of Proposition~\ref{prop:sw_posonly}}
\label{sec:revealrm_app-prelimsubmod-posonly}

\begin{proof}[Proof of Proposition~\ref{prop:sw_posonly}]
    Let $F(A)$ denote the social welfare generated by the revealed positive target set $A$, and let $A \subseteq B \subseteq \Ts^+$ where $\Ts^{+}=\{t\in\Ts \mid f(t)=+1\}$. Submodularity requires that for any $A, B \subseteq \Ts^+$ and any 
    $t \in \Ts^+ \setminus B$,
    \[
    F(A \cup \{t\}) - F(A) \geq F(B \cup \{t\}) - F(B).
    \]
    By Proposition~\ref{prop:monotone}, $F$ is monotone: $A \subseteq B$ implies $F(A) \le F(B)$. Revealing a positive target $t$ sets each adjacent agent's probability of emulating a positive target to $1$, unless it is already equal to $1$ under the current revealed set.
    For any set $S \subseteq \Ts^+$, let $\gamma(S)$ be the set of agents whose probability of emulating a positive target equals $1$ under $S$. Monotonicity implies that 
    $\gamma(A) \subseteq \gamma(B)$.
    
    Adding a revealed positive target $t$ to the revealed positive target sets  $A$ and $B$ yields
    \[
    F(A \cup \{t\}) - F(A) = |\gamma(A \cup \{t\})| - |\gamma(A)| = |\gamma(t) \setminus \gamma(A)|.
    \]
    \[
    F(B \cup \{t\}) - F(B) = |\gamma(B \cup \{t\})| - |\gamma(B)|  = |\gamma(t) \setminus \gamma(B)|.
    \]
    
    Since $\gamma(A) \subseteq \gamma(B)$,
    \[
    \gamma(t) \setminus \gamma(A) \supseteq \gamma(t) \setminus \gamma(B),
    \]
    because any agent that already had a probability of $1$ for emulating a positive target under the larger revealed set $B$ is excluded on the right-hand side. That is, adding a revealed positive target $t$ to a large set $B$ yields lower marginal gain because most agents are already committed to positive targets in $B$, so $t$ might be redundant. Adding $t$ to a small set $A$ instead makes it more likely that additional agents now have a probability of $1$ for emulating a positive target. Therefore
    \[
    |\gamma(t) \setminus \gamma(A)| \ge |\gamma(t) \setminus \gamma(B)|.
    \]
    Thus, revealing a positive target $t$ when the revealed positive target set is smaller affects weakly more agents than when the known set is larger. The marginal contribution of revealing a new positive target $t$ declines as the revealed positive target set grows, which establishes submodularity: $ F(A \cup \{t\}) - F(A) \geq F(B \cup \{t\}) - F(B)$.
\end{proof}

\subsection{Proof of Proposition~\ref{prop:sw_negonly}}
\label{sec:revealrm_app-prelimsubmod-negonly}

\begin{example}[Proof of Proposition~\ref{prop:sw_negonly}]
\label{ex:mod_submod_proof}
    Consider the bipartite graph in Figure~\ref{fig:not_submod_proof}, with $n=4$ agents, $\mpos=4$ positive targets, and $\mneg=2$ negative targets. Each agent is adjacent to a unique positive target and to both the negative targets. Example \ref{ex:mod_submod_proof} illustrates that when the $\splan$ is restricted to revealing negative targets in this setting, the social welfare function is not submodular. In particular, it shows that there exists $A, B \subseteq \Ts^-$ with 
    $\Ts^{-}=\{t\in\Ts \mid f(t)=-1\}$ and $A \subseteq B$, and for some $t \in \Ts^- \setminus B$ such that
    \[
    F(A \cup {t}) - F(A) < F(B \cup {t}) - F(B).
    \]
    Given bipartite graph in Figure~\ref{fig:not_submod_proof} and restriction to only revealing negative targets. Let the initially revealed negative target set be $A=\emptyset$, resulting in a social welfare of $F(A)=4/3$.
    Next, let revealed negative target set $B=\{t_5^{-}\}$, and corresponding resulting social welfare $F(B)=2$.
    Now consider revealing another negative target $t_6^{-}\in \Ts^{-} \setminus B$. The marginal gains become
    \[
    F(A \cup \{t_6^{-}\}) - F(A) = 2 - \tfrac{4}{3} = 0.6667, \qquad
    F(B \cup \{t_6^{-}\}) - F(B) = 4 - 2 = 2.
    \]
    Thus the marginal gain is larger when starting from a larger revealed negative target set $B$ than from a smaller one $A$, contradicting the diminishing returns condition required for submodularity.
\end{example}

\begin{figure}[ht!]
\centering
    \begin{tikzpicture}[
        node distance=1cm and 1cm,
        agent/.style={
            circle,
            fill=#1!30, 
            inner sep=1.5pt, 
            minimum size=0.8cm, 
            text centered
        },
        inputNode/.style={agent={red}},
        posTarget/.style={agent={gray}},
        negTarget/.style={agent={gray}},
        connect/.style={thick},
        scale=0.44
    ]
    
    \foreach \i in {1,2,3,4} {
        \node[inputNode] (x\i) at (\i*2,0) {$x_{\i}$};
    }
    \foreach \i in {1,2,3,4} {
        \node[posTarget] (tp\i) at (\i*2,3) {$t_{\i}^{+}$};
    }
    
    \node[negTarget] (tn5) at (4,-3) {$t_{5}^{-}$};
    \node[negTarget] (tn6) at (6,-3) {$t_{6}^{-}$};
    
    \foreach \i in {1,2,3,4} {
        \draw[connect] (x\i) -- (tp\i);
        \draw[connect] (x\i) -- (tn5);
        \draw[connect] (x\i) -- (tn6);
    }
    \end{tikzpicture}

    \caption{A bipartite graph where \(n= 4, \mneg = 2\) and \(\mpos = 4\).}
    \label{fig:not_submod_proof}
\end{figure}

\section{Supplementary Material for Section~\ref{sec:revealrm_standardmodel}}
\label{sec:revealrm_app_standard_algos}

\subsection{Proof of Proposition~\ref{prop:opt_runtime_greedy}}
\label{sec:revealrm_app-classicgreedy}

\begin{proposition}
Algorithm~\ref{alg:greedy_lbreveal} runs in \(O(Kmn\delta )\) time.
\label{prop:opt_runtime_greedy}
\end{proposition}

\begin{proof}
At each iteration, to determine whether to add a target \(t \in \Ts\) to revealed target set, Algorithm~\ref{alg:greedy_lbreveal} computes the resultant marginal gain, which first, involves computing \(F(S_{\mathrm{g}}\cup \{t\})\), a sum over \(Q^{S_{\mathrm{g}}\cup \{t\}}(x)\) for all \(n\) agents \(x \in \Xs\). If each agent has degree of atmost \(\delta = |N(x)|\), then the time complexity of computing the social welfare \(F(S_{\mathrm{g}}\cup \{t\})\) is \(O(n\delta )\). Computing the marginal gain given the previous and the new social welfare is \(O(1)\). Repeating this process for \(m\) targets across atmost \(K\) iterations results in a total time complexity of \(O(Kmn\delta )\).
\end{proof}

\subsection{Bruteforce Algorithm}
\label{sec:revealrm_app-bruteforce}

\begin{algorithm}[ht!]
\caption{Bruteforce Target Reveal}
\label{alg:bruteforce_lbreveal}
\SetAlgoLined
    \KwIn{Bipartite graph $\graph = (\Xs \cup \Ts, E)$, labels $\{f(t)\}_{t \in \Ts}$, budget \(K\)}
    
    \KwOut{Optimal solution set $S^\star \subseteq \Ts \ \text{with} \ |S^\star| \leq K$ and social welfare $F(S^\star)$}
    
    Initialize $S^\star \gets \emptyset$
    
    \ForAll{$S \subseteq \Ts$ with $|S| \leq K$}{
        \If{$\big(F(S) > F(S^\star\big)$ or $\big(F(S) = F(S^\star)$ and $|S| < |S^\star|\big)$}{ 
            $S^\star \gets S$
            
            $F(S^\star) \gets F(S)$
        }
    }
    
    \Return $(S^\star, F(S^\star))$

\end{algorithm}

\begin{proposition}
The bruteforce algorithm (Algorithm~\ref{alg:bruteforce_lbreveal}) runs in 
\(O(m^{K} n \delta)\) time.
\label{prop:opt_runtime_bruteforce}
\end{proposition}

\begin{proof}
Algorithm \ref{alg:bruteforce_lbreveal} enumerates all \(m^{K}\) candidate subsets of possible target reveals and selects the one that yields the highest social welfare. Evaluating social welfare for a single subset takes \(O(n\delta)\) time, so the overall running time of Algorithm \ref{alg:bruteforce_lbreveal} is \(O(m^{K} n \delta)\).
\end{proof}

\subsection{Proofs for Section~\ref{subsec:revealrm_approxs}}
\label{sec:revealrm_app_sm_approxs}

\begin{theorem}
\label{thm:neg-appratio}
    When the $\splan$ is restricted to only revealing negative targets, there exists a graph and a budget $K$ for which Algorithm~\ref{alg:greedy_lbreveal} attains an approximation ratio strictly less than $\frac{3}{\sqrt{2n}}$.
\end{theorem}

\begin{proof}
    The proof is in Example~\ref{ex:revealonlynegs_notsub} below.
\end{proof}

\begin{figure}[b!]
\centering
    \begin{tikzpicture}[
            node distance=1cm and 1cm,
            agent/.style={
                circle,
                fill=#1!30, 
                inner sep=1.5pt, 
                minimum size=1.1cm, 
                text centered
            },
            inputNode/.style={agent={red}},
            posTarget/.style={agent={gray}},
            negTarget/.style={agent={gray}},
            connect/.style={thick},
            scale=0.82
        ]

        \foreach \i in {1,2,3} {
            \node[inputNode] (x\i) at (\i*2.4,0) {$x_{\i}$};
        }
        \node at (8.3,0) {$\ldots$};
        \node[inputNode] (xn) at (9.4,0) {$x_{\lfloor \frac{\kappa^2}{2} \rfloor}$};
        
        \foreach \i in {1,2,3} {
            \node[posTarget] (tp\i) at (\i*2.4,2.4) {$t_{\i}^{+}$};
        }
        \node at (8.3,2.4) {$\ldots$};
        \node[posTarget] (tpn) at (9.4,2.4) {$t_{\lfloor \frac{\kappa^2}{2} \rfloor}^{+}$};
        
        \foreach \i in {1,2,3} {
            \node[negTarget] (tn\i) at (\i*2.4,-2.4) {$t_{\i}^{-}$};
        }
        \node at (8.3,-2.4) {$\ldots$};
        \node[negTarget] (tnn) at (9.4,-2.4) {$t_{\kappa+1}^{-}$};
        
        \foreach \i in {1,2,3,n} {
            \draw[connect] (x\i) -- (tp\i);
            \foreach \j in {1,2,3,n} {
                \draw[connect] (x\i) -- (tn\j);
            }
        }

        \foreach \i in {1,2,3} {
            \node[inputNode] (xp\i) at (9.5+\i*2.0,0) {$x_{\i}^{'}$};
        }
        \node at (16.8,0) {$\ldots$};
        \node[inputNode] (xpn) at (18.0,0) {$x_{\kappa+1}^{'}$};
        
        \foreach \i in {1,2,3} {
            \node[posTarget] (tpp\i) at (9.5+\i*2.0,2.4) {$t_{\i}^{'+}$};
        }
        \node at (16.8,2.4) {$\ldots$};
        \node[posTarget] (tppn) at (18.0,2.4) {$t_{\kappa+1}^{'+}$};
        
        \foreach \i in {1,2,3} {
            \node[negTarget] (tnp\i) at (9.5+\i*2.0,-2.4) {$t_{\i}^{'-}$};
        }
        \node at (16.8,-2.4) {$\ldots$};
        \node[negTarget] (tnpn) at (18.0,-2.4) {$t_{\kappa+1}^{'-}$};
        
        \foreach \i in {1,2,3,n} {
            \draw[connect] (xp\i.north) -- (tpp\i.south);
            \draw[connect] (xp\i.south) -- (tnp\i.north);
        }
    
    \end{tikzpicture}

    \caption[A bipartite graph with \(n= \lfloor \frac{\kappa^2}{2} \rfloor + \kappa+1, \mneg = 2\kappa+2,\) and \(\mpos = n\)]{A bipartite graph example with statistics: \(n= \lfloor \frac{\kappa^2}{2} \rfloor + \kappa+1, \mneg = 2\kappa+2,\) and \(\mpos = n\).  Algorithm~\ref{alg:greedy_lbreveal} achieves an approximation ratio of \(\frac{3}{\sqrt{2n}}\)}
    \label{fig:notsubmodx}
\end{figure}
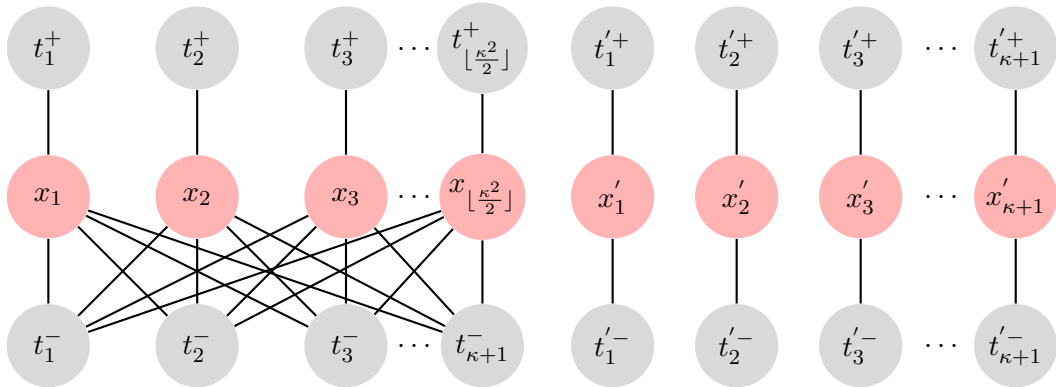

\begin{example}
\label{ex:revealonlynegs_notsub}
    Let $\kappa \in \mathbb{N}$ with $\kappa > 3$, and set the reveal budget to $K \coloneqq \kappa + 1$.
    Consider the bipartite graph shown in Figure~\ref{fig:notsubmodx}, with $n = \lfloor \frac{\kappa^2}{2} \rfloor + \kappa + 1$ agents, $\mneg = 2\kappa + 2$ negative targets, and $\mpos = n$ positive targets. There are two types of agents, referred to as group 1 and group 2. Group 1 consists of $\lfloor \frac{\kappa^2}{2} \rfloor$ agents, each denoted $x_{\star}$. Each $x_{\star}$ is connected to a distinct positive target $t_{\star}^{+}$ and to $\kappa + 1$ of the negative targets $t_{\star}^{-}$. 
    Group 2 consists of $\kappa + 1$ agents, each denoted $x_{\star}^{'}$, and each is connected to a unique pair consisting of one positive and one negative target $(t_{\star}^{'+}, t_{\star}^{'-})$. 
    
    Initially, the social welfare is \(F(\emptyset) =  \frac{\lfloor \frac{\kappa^2}{2} \rfloor}{\kappa+2} + \frac{\kappa+1}{2}\).  
    At first iteration, we analyze which negative target the greedy algorithm picks:
    (1) If any one of the negative targets connected to group 1 agents
    (i.e, any of \(t_{*}^{-}\)), then it achieves a social welfare of
    \(\frac{\lfloor \frac{\kappa^2}{2} \rfloor}{\kappa+1} + \frac{\kappa+1}{2}\);
    (2) If any one of the negative targets connected to group 2 agents (i.e, any of \(t_{*}^{'-}\)), then it achieves a social welfare of
    \(\frac{\lfloor \frac{\kappa^2}{2} \rfloor}{\kappa+2} + \frac{\kappa}{2} + 1\). 
    The resulting social welfare of case 2 is deceptively higher than that of case 1 because, although it initially appears higher, it can mislead the algorithm towards a suboptimal path.

    It can be verified that Algorithm~\ref{alg:greedy_lbreveal} reveals, at each iteration, one of the $t_{*}^{'-}$ targets, until the budget \(K=\kappa+1\) is fully used.
    As a result, at \(K=\kappa+1\) budget, \(\kappa+1\) agents can each emulate a positive target with probability \(1\), and the rest of the 
    \(\lfloor \frac{\kappa^2}{2} \rfloor\) agents can each do this at probability \(\frac{1}{\kappa+2}\). 
    Therefore the approximation ratio is  
    \(
    \frac{\frac{\lfloor \frac{\kappa^2}{2} \rfloor}{\kappa+2} + \kappa+1}{\lfloor \frac{\kappa^2}{2} \rfloor + \frac{\kappa+1}{2}} = 
    \frac{ \frac{\lfloor \frac{\kappa^2}{2} \rfloor + \kappa^2 + 3\kappa + 2 }{\kappa+2} }{ \lfloor \frac{\kappa^2}{2} \rfloor + \frac{\kappa+1}{2} } <
    \frac{\kappa^2 \left( 3 + \frac{6}{\kappa} + \frac{4}{\kappa^2} \right)}{\kappa^3 \left( 1 + \frac{3}{\kappa} + \frac{3}{\kappa^2} + \frac{2}{\kappa^3}\right)} <
    \frac{3}{\kappa} < {\frac{3}{\sqrt{2n}}}
    \)
    which is significantly much worse than \(1-\frac{1}{e}\) for \(K=\kappa+1, \kappa>3\). 

\end{example}

\begin{theorem}
    When the $\splan$ can reveal both positive and negative targets, there exists a graph and a budget $K$ for which Algorithm~\ref{alg:greedy_lbreveal} attains an approximation ratio strictly less than $2/\sqrt{n}+2$.
\label{thm:negpos-appratio}
\end{theorem}

\begin{proof}
    The proof is in Example~\ref{ex:revealposneg_notsub} below.
\end{proof}

\begin{example}
\label{ex:revealposneg_notsub}
    Let $\kappa \in \mathbb{N}$ with $\kappa \ge 3$, and set the reveal budget to $K \coloneqq \kappa + 1$.
    Consider the bipartite graph in Figure \ref{fig:notsubmod}. 
    There are \(n = \kappa^{2}\) agents, each connected to a unique positive target and to all the \(\mneg = \kappa + 1 = \sqrt{n} + 1\) negative targets. 
        
    Before any target is revealed, the social welfare is \(F(\emptyset) =  \frac{\kappa^2}{\kappa+2}\). Revealing any negative target in the first iteration increases the welfare to \(F(\{t^{-}_{*}\}) =  \frac{\kappa^2}{\kappa+1}\), giving a marginal gain of 
    \(\Delta_{t^{-}_{*}}(\emptyset) = \frac{\kappa^2}{\kappa+1} - \frac{\kappa^2}{\kappa+2} = \frac{\kappa^2}{(\kappa+1)(\kappa+2)}\). In contrast, revealing any positive target yields \(F(\{t^{+}_{*}\}) =  \frac{\kappa^2 + \kappa + 1}{\kappa+2}\), with marginal gain  \(\Delta_{t^{+}_{*}}(\emptyset) = \frac{\kappa^2 + k + 1}{\kappa+2} - \frac{\kappa^2}{\kappa+2} = \frac{\kappa+1}{\kappa+2} > \frac{\kappa^2}{(\kappa+1)(\kappa+2)}\). 
        
    Consequently, Algorithm \ref{alg:greedy_lbreveal} reveals a positive target in the first iteration and continues to do so in subsequent steps. The problem is that, under the budget \(K= \kappa+1\), revealing the negative targets would achieve the optimal social welfare \(F(S^\star)=\kappa^2\). However, the marginal gain from releasing negative targets is not only initially much smaller than that of positive targets but also increases only when additional negative targets are revealed. As a result, the Algorithm~\ref{alg:greedy_lbreveal}  never reveals them.
    
    At \(K=\kappa+1\) budget, \(\kappa+1\) agents can each emulate a positive target with probability \(1\), and the rest of the agents can each do this at probability \(\frac{1}{\kappa+2}\).
    Therefore the approximation ratio is  
    \(\frac{\kappa+1 + \frac{\kappa^2-(\kappa+1)}{\kappa+2}}{\kappa^2} = 
    \frac{2 + \frac{2}{k} + \frac{1}{k^2}}{k+2} = \frac{2}{k+2} + \frac{2}{k(k+2)} + \frac{1}{k^2(k+2)} 
     < \frac{2}{\kappa+2} =  {\frac{2}{\sqrt{n}+2}}\)  which is significantly much worse than \(1-\frac{1}{e}\) for \(K=\kappa+1, \kappa>2\). 
\end{example}
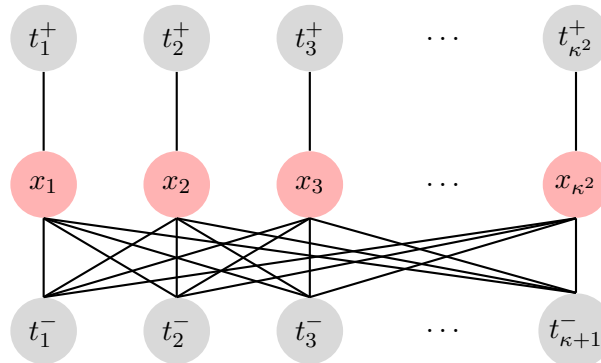
\begin{figure}[ht!]
\centering
    \begin{tikzpicture}[
        node distance=1cm and 1cm,
        agent/.style={
            circle,
            fill=#1!30, 
            inner sep=1.5pt, 
            minimum size=0.88cm, 
            text centered
        },
        inputNode/.style={agent={red}},
        posTarget/.style={agent={gray}},
        negTarget/.style={agent={gray}},
        connect/.style={thick},
        scale=0.88
    ]
    
    \foreach \i in {1,2,3} {
        \node[inputNode] (x\i) at (\i*2,0) {$x_{\i}$};
    }
    \node at (8,0) {$\ldots$};
    \node[inputNode] (xn) at (10,0) {$x_{\kappa^{2}}$};
    
    \foreach \i in {1,2,3} {
        \node[posTarget] (tp\i) at (\i*2,2.2) {$t_{\i}^{+}$};
    }
    \node at (8,2.2) {$\ldots$};
    \node[posTarget] (tpn) at (10,2.2) {$t_{\kappa^{2}}^{+}$};

    \foreach \i in {1,2,3} {
        \node[negTarget] (tn\i) at (\i*2,-2.2) {$t_{\i}^{-}$};
    }
    \node at (8,-2.2) {$\ldots$};
    \node[negTarget] (tnn) at (10,-2.2) {$t_{\kappa+1}^{-}$};
    
    \foreach \i in {1,2,3,n} {
        \draw[connect] (x\i.north) -- (tp\i.south);
        \foreach \j in {1,2,3,n} {
            \draw[connect] (x\i.south) -- (tn\j.north);
        }
    }
    
    \end{tikzpicture}

    \caption[A bipartite graph example with statistics: \(n= \kappa^2, \mneg = \kappa+1,\) and \(\mpos = n\)]{A bipartite graph example with statistics: \(n= \kappa^2, \mneg = \kappa+1,\) and \(\mpos = n\).  Algorithm~\ref{alg:greedy_lbreveal} achieves an approximation ratio of \(\frac{2}{\sqrt{n}+2}\).}
    \label{fig:notsubmod}
\end{figure}

\subsubsection{Proof of Theorem~\ref{thm:negpos-appratio2}}
\label{subsec:revealrm_app-posneg-constapprox}

\begin{lemma}
\label{lem:proxy}
    If all agents are $c$-bounded, then for any target set $S$, both the true social welfare and gain are approximated by their proxy counterparts within a factor of $c$. That is, $F_p(S) \leq F(S) \leq cF_p(S)$ and $G_{p}(S) \leq G(S) \leq c G_{p}(S)$.
\end{lemma}
\begin{proof}
    First, we show that the proxy social welfare is always less than or equal to the true social welfare.
    For a given agent $x$, consider three cases. One, if the agent has no neighbors $N(x)=\emptyset$, then 
    $Q^{S}_{p}(x)=Q^{S}(x) = 0$. 
    Second, if there is at least one revealed positive target neighbor of $x$ in $S$, then 
    $Q^{S}_{p}(x) = Q^{S}(x) = 1$. 
    Otherwise, the proxy probability mass is less than or equal to the true one, since 
    $Q^{S}_{p}(x)$ only increases by the amount that revealing the first negative neighbor helped which is atmost the increase in $Q^{S}(x)$.
    This then implies that $Q^{S}_{p}(x) \leq Q^{S}(x)$. Summing over all agents $x \in \Xs$ proves that for all $S \subseteq \Ts$, $F_{p}(S) \leq F(S)$.

    Next, we show that $F_p(S) \geq \frac{1}{c}F(S)$.     
    Let $t \in \Ts \setminus A$ be a target revealed by the social planner such that $S=A\cup\{t\}$.  Assume that each agent has at most $c \geq 1$ negative neighbors, $\delta^{-}_{x} \leq c$ for all $x \in \Xs$. 
    Then if $t$ is revealed to be positive $|P^{+}_{x}(S)|\geq 1$, both the proxy and true social welfare go up by atmost $1$, that is,  $Q^{S}_{p}(x)-Q^{\emptyset}_{p}(x) = Q^{S}(x)-Q^{\emptyset)}(x) = 1-\frac{\delta^{+}_{x}}{|N(x)|}$.
    If $t$ is negative and $|P^{+}_{x}(S)|>0$, then there will be no effect on both the proxy and true social welfare. 
    Otherwise, if $t$ is negative and $|P^{+}_{x}(S)|=0$ and $|P^{-}_{x}(S)|=\delta^{-}_{x}$, 
    the true social welfare increases by 
    \begin{align*}
        Q^{S}(x)-Q^{\emptyset}(x) 
        & =    1 - \frac{\delta^{+}_{x}}{\delta^{+}_{x}+\delta^{-}_{x}}\\
        & \geq 1 - \frac{1}{1+\delta^{-}_{x}}  \quad (\text{if } \delta^{+}_{x} =1) \\
        & \geq 1 - \frac{1}{1+c}  \quad  (\text{if } \delta^{-}_{x} \leq c)\\
        & =    \frac{c}{1+c}
    \end{align*}
    and the proxy social welfare increases by 
    \begin{align*}
        Q^{S}_{p}(x)-Q^{\emptyset}_{p}(x) 
        & =   \frac{\delta_x^{+}}{\delta^{+}_{x}+\delta^{-}_{x}-1}
        \left(1+\frac{\delta^{-}_{x}-1}{\delta^{+}_{x}+\delta^{-}_{x}}\right) - 
        \frac{\delta^{+}_{x}}{\delta^{+}_{x}+\delta^{-}_{x}}\\
        & \geq \frac{1}{\delta^{-}_{x}}
        \left(\frac{2\delta^{-}_{x}}{1+\delta^{-}_{x}}\right) - 
        \frac{1}{1+\delta^{-}_{x}}   \quad (\text{if } \delta^{+}_{x} =1) \\   
        & \geq \frac{2}{(1+c)} - \frac{1}{1+c}   \quad  (\text{if } \delta^{-}_{x} \leq c)\\  
        & =    \frac{1}{1+c} 
    \end{align*}
    In this case $\frac{Q^{S}_{p}(x)-Q^{\emptyset}_{p}(x)}{Q^{S}(x)-Q^{\emptyset}(x)} \geq \frac{1}{c}.$ Therefore, 
    $\left(Q^{S}_{p}(x)-Q^{\emptyset}_{p}(x)\right) \geq \frac{1}{c}\left(Q^{S}(x)-Q^{\emptyset)}(x)\right)$.
    Lastly, since $Q^{\emptyset}_{p}(x)=Q^{\emptyset}(x)$, and if $Q^{S}(x)>Q^{\emptyset}(x)$, then summing over all agents $x \in \Xs$, for all $S$, $F_p(S) \geq \frac{1}{c}F(S)$.
    Put together, $F_{p}(S) \leq F(S) \leq c F_{p}(S)$, and $G_{p}(S) \leq G(S) \leq c G_{p}(S)$.
\end{proof}

\begin{lemma}
\label{lem:proxy-sub}
    When the social planner can reveal positive and negative targets, the proxy social welfare function is submodular. That is, for every $A, B \subseteq \Ts$ where $A \subseteq B$, every $t \in \Ts \setminus B$, 
    $F_{p}(A \cup \{t\}) - F_{p}(A) \geq F_{p}(B \cup \{t\}) - F_{p}(B)$
\end{lemma}
\begin{proof}
    Consider an agent $x$, and two cases where $t$ is adjacent to $x$ and is either positive or negative. 
    If the adjacent target $t$ is positive, then 
    $Q^{B \cup \{t\}}_{p}(x)-Q^{B}_{p}(x) = 1-Q^{B}_{p}(x)$ if there was previously no positive target neighbors of $x$ in $B$, and $0$ if other positive target neighbors were already revealed 
    $|P^{+}_{x}(B)|>0$.

    In the second case, if target $t$ is adjacent to $x$ and is revealed as negative, then the marginal gain is a constant $Q^{B \cup \{t\}}_{p}(x)-Q^{B}_{p}(x) = \frac{\delta^{+}_{x}}{|N(x)| (|N(x)|-1)}$ defined by the gain from revealing the first negative target.

    In both cases, the marginal is non-increasing as the revealed target set grows, because revealing an additional adjacent target of the same label yields zero gain if it's positive and a constant gain if it's negative. Therefore, 
    $Q^{A \cup \{t\}}_{p}(x)-Q^{A}_{p}(x) \geq Q^{B \cup \{t\}}_{p}(x)-Q^{B}_{p}(x)$ and summing over $x \in \Xs$ and given the sum rule for submodular functions, 
    $F_{p}(A \cup \{t\}) - F_{p}(A) \geq F_{p}(B \cup \{t\}) - F_{p}(B).$ 
\end{proof}

\begin{corollary}
\label{col:proxygain-sub}
    When the social planner can reveal positive and negative targets, the proxy is submodular on the gain. That is, for every 
    $A, B \subseteq \Ts$ where $A \subseteq B$, every $t \in \Ts \setminus B$, 
    $G_{p}(A \cup \{t\}) - G_{p}(A) \geq G_{p}(B \cup \{t\}) - G_{p}(B)$
\end{corollary}
\begin{proof}
    This follows directly from Lemma~\ref{lem:proxy-sub}. Since 
    $G_{p}(A \cup \{t\}) - G_{p}(A) = F_{p}(A \cup \{t\}) - F(\emptyset) - F_{p}(A) + F(\emptyset)$ and 
    $G_{p}(B \cup \{t\}) - G_{p}(B) = F_{p}(B \cup \{t\}) - F(\emptyset) - F_{p}(B) + F(\emptyset)$, 
    then the proxy is submodular on the gain.
\end{proof}

\begin{remark}
\label{rem:1/capprox}
    Assume revealed targets may include negative ones. 
    If all agents are $c$-bounded, then with respect to the true social welfare, $F(\emptyset) \geq \frac{1}{c}F(S^\star)$ where
    $S^\star$ denotes the optimal revealed set when using the true social welfare function (Eqn.~\ref{eq:social_welfare}). 
    Let $S$ be either $\emptyset$ or any arbitrary solution set. Ignore any agent $x \in \Xs$ where 
    $\delta^{+}_{x} = 0$ since $Q^{S}(x) = Q^{S^\star}(x) = 0$. For agents with at least one positive target,  
    $Q^{S}(x) \geq \frac{1}{c+1}Q^{S^\star}(x)$ since $Q^{S^\star}(x) \le 1$ and $Q^{S}(x) \geq \frac{1}{c+1}$. 
    Summing over all such agents yields $F(S) \geq \frac{1}{c+1}F(S^\star)$. For a large $c$, 
    $F(S) \geq \frac{1}{c}F(S^\star)$. Thus, any solution set (including empty set) achieves a 
    $(1/c)$-factor approximation to the true optimal welfare.
\end{remark}

\begin{proof}[Proof of Theorem~\ref{thm:negpos-appratio2}]
    For a target reveal budget of $K$, let $S$ denote the solution returned by Algorithm~\ref{alg:greedy_lbreveal}, and let $S^\star$ denote the optimal solution set under the true social welfare function $F$ (Eqn.~\ref{eq:social_welfare}). 
    Let $S_p$ denote the solution returned by proxy-greedy, and $S_p^\star$ the optimal solution set under the proxy social welfare function $F_p$ (Defn.~\ref{def:proxy_sw}).
    Since $S_p^\star$ maximizes $F_p$, then $G_p(S^{\star}_{p}) \geq  G_p(S^{\star})$. 
    Since by Lemma~\ref{lem:proxy} $G_p(S^\star) \geq \frac{1}{c}G(S^\star)$, then 
    $G_p(S^{\star}_{p}) \geq G_p(S^\star) \geq \frac{1}{c}G(S^\star)$.
    By Lemma~\ref{lem:proxy-sub}, the proxy welfare function is submodular and therefore proxy-greedy achieves a $(1-1/e)$-approximation to the optimal proxy welfare, i.e.,
    $G_{p}(S_p) \ge (1-1/e) G_{p}(S^{\star}_{p}).$ 
    Since by Lemma~\ref{lem:proxy}, $G(S_p) \geq G_p(S_p)$, then, 
    $G(S_p) \geq G_p(S_p) \geq (1-1/e)F_p(S^{\star}_{p}) \geq \frac{1-1/e}{c} G(S^\star)$.
    Thus, $G(S_p) \geq \frac{1-1/e}{c} G(S^\star)$.
\end{proof}

\subsection{Proofs for Section~\ref{sec:revealrm_fairness}}
\label{sec:revealrm_sm_fairnessproofs}

\begin{lemma}
\label{lem:fairsub-whole}
    If the social welfare function is submodular, then $\mathrm{OPT}^{K_a} \geq \mathrm{OPT}^{K}/w$.
\end{lemma}

\begin{proof}
    Let $S_K^{\star}=\{t_1,\ldots,t_K\}$ be the optimal revealed target set for the whole graph under budget $K$, so that $\mathrm{OPT}^K = F(S_K^{\star}) - F(\emptyset)$, and define $S_K^{\star}$ prefix sets as $S_i=\{t_1,\ldots,t_i\}$ for $i\in[K]$, with $S_0=\emptyset$. 
    Thus $\mathrm{OPT}^K$ as a sum of successive marginal gains is 
    $\sum_{i=1}^K \bigl(F(S_i)-F(S_{i-1})\bigr) - F(\emptyset)$.
    Similarly, for $K_a=\lceil K/w\rceil$, optimal welfare gain is 
    $\mathrm{OPT}^{K_a}=\sum_{i=1}^{K_a} \bigl(F(S_i)-F(S_{i-1})\bigr) - F(\emptyset)$. 
    By submodularity of the social welfare function, these marginal gains form a non-increasing sequence, that is,
    $F(S_j)-F(S_{j-1}) \ge F(S_i)-F(S_{i-1})$ for all $j<i$. 
    Therefore, the average marginal gain over the first $K_a$ targets is at least the average over all $K$ targets. That is, $\frac{1}{K_a}\mathrm{OPT}^{K_a} \geq \frac{1}{K}\mathrm{OPT}^{K}$. 
    Since $K_a = \lceil K/w \rceil$ then $\mathrm{OPT}^{K_a} \geq \mathrm{OPT}^{K}/w$.
\end{proof}

\begin{corollary}
\label{col:fairsub}
    If the social welfare function is submodular, then $\mathrm{OPT}_{a}^{K_a} \geq \mathrm{OPT}_{a}^{K} / w$.
\end{corollary}

\begin{proof}
    Let $\mathrm{OPT}_{a}^{K}$ denote the optimal social welfare gain obtained when we focus exclusively on group $a$ with budget $K$, and let $\mathrm{OPT}_{a}^{K_a}$ denote the optimal welfare gain under a smaller budget $K_a = \lceil K/w \rceil$. Since both welfare gains depend only on the nodes and edges within group $a$, then that graph can itself be viewed as the entire graph.
    Thus, by Lemma~\ref{lem:fairsub-whole},
    $\mathrm{OPT}_a^{K_a} \geq \mathrm{OPT}_a^{K}/w$.
\end{proof}

\begin{theorem}
\label{thm:pos_fairsub}
For any arbitrary graph, given a limit $K \geq w$ on the reveal budget and restriction to revealing positive targets, Algorithm~\ref{alg:greedy_lbreveal} outputs a solution set that is simultaneously ($(1-1/e)/w$)-approximately optimal for each group. That is, for any group $a \in [w]$, the algorithm reveals target set $S_{K_a} \subseteq \Ts^{+}$ with 
$|S_{K_a}|\leq K_a$ such that $G(S_{K_a}) \geq \frac{(1-1/e)}{w}\mathrm{OPT}_{a}^{K}$, where $\mathrm{OPT}_{a}^{K}$ is the maximum social welfare gain for group $a$ with budget $K$, restricting candidate targets to $\Ts^{+}$ and optimizing only over that group.
\end{theorem}

\begin{proof}
    Let each group $a$ be assigned a budget $K_a = \lceil K/w \rceil$, such that Algorithm~\ref{alg:greedy_lbreveal} run exclusively on each group $a$ returns a target set $S_{K_a} \subseteq \Ts^{+}$ with $|S_{K_a}|\leq K_a$ such that the social welfare gain of the group is given by $G(S_{K_a}) = F(S_{K_a}) - \sum_{x \in A_a} Q^{\emptyset}(x)$. 
    Since the social planner is restricted to only revealing positive targets and the social welfare gain function is monotone and submodular, then for each group $a \in [w]$, $G(S_{K_a}) \geq (1-1/e)\mathrm{OPT}_{a}^{K_a}$. By Corollary~\ref{col:fairsub}, 
    $(1-1/e)\mathrm{OPT}_{a}^{K_a} \geq \frac{(1-1/e)}{w}\mathrm{OPT}_{a}^{K}$.  Combining everything, it then follows that for any group $a \in [w]$, $G(S_{K_a}) \geq \frac{(1-1/e)}{w}\mathrm{OPT}_{a}^{K}$. 
\end{proof}

\begin{remark}
\label{rem:posneg_notfair}
  When the revealed target set includes negative targets and the social welfare function is monotone but not necessarily submodular (cf. Proposition~\ref{prop:sw_negonly}), there may be no solution that substantially benefits multiple groups at once. For example, consider a case with two groups, each defined by a bipartite graph illustrated in Figure~\ref{fig:notsubmodx} in Appendix~\ref{sec:revealrm_app_sm_approxs}. In this case, achieving high social welfare may require allocating all or nearly all of the target reveal budget $K$ to a single group.
\end{remark}

\begin{corollary}
\label{col:proxy-fairness}
    If the revealed targets include negative ones, and all agents are  $c$-bounded, then under the proxy-greedy algorithm, each group $a$ receives welfare gain at least $\Omega(\mathrm{OPT}_{a}^{K})$.
\end{corollary}

\begin{proof}
    For a given target reveal budget $K$, let $G(S_a)$ denote the welfare gain obtained by running Algorithm~\ref{alg:greedy_lbreveal} exclusively on group $a \in [w]$, and let $\mathrm{OPT}^K_a$ be the corresponding optimal social welfare gain under the true social welfare function $F$ (Eqn.~\ref{eq:social_welfare}). 
    Let $G_{p}(S_{p,a})$ denote the proxy social welfare gain obtained by running proxy-greedy on that group, and let 
    $\mathrm{OPT}^{K}_{p,a}$ be the optimal gain under the proxy social welfare function $F_p$ (Defn.~\ref{def:proxy_sw}). 
    By Lemma~\ref{lem:proxy}, $G(S_{p,a}) \geq G_{p}(S_{p,a})$, and by Theorem~\ref{thm:negpos-appratio2}, 
    $G_{p}(S_{p,a}) \geq (1 - 1/e) \mathrm{OPT}^K_{p,a} \geq \frac{1 - 1/e}{c}\mathrm{OPT}^{K}_{a}$, for any group $a \in [w]$.
    Put together, $G(S_{p,a}) \geq \frac{1 - 1/e}{c} \mathrm{OPT}^{K}_{a}.$ 
\end{proof}

\paragraph{Equal proxy social welfare doesn't imply similar emulation choices.}
Consider the example shown in Figure~\ref{fig:not_submod_proof}, where there are $4$ agents, each connected to a unique positive target and $2$ common negatives. 
Here, revealing $2$ negative or $2$ positive targets results in equal proxy social welfare $(8/3)$, but the two reveals have different implications on agents' emulation choices. 
Revealing two positive targets ensures that $2/4$ agents can emulate a positive target with certainty and the other two can emulate a positive target with probability $1/3$ and a negative target with probability $1/3$. On the other hand, revealing two negative targets ensures that all the agents can emulate a positive target with probability $2/3$ and a negative target with probability $0$ since revealing a negative target means agents avoid it or their probability of emulating it is $0$. In this case, revealing $2$ negative targets results in better emulation choices for all agents, but proxy-greedy might choose to reveal $2$ positive targets instead.

\subsection{Proofs for Section~\ref{sec:revealrm_np-hardness}}
\label{sec:revealrm_sm_nphardness}

\subsubsection{Proof of Theorem~\ref{thm:np-hardness_posonly}}
\label{sec:revealrm_sm_nphardness-posonly}

\begin{proof}[Proof of Theorem~\ref{thm:np-hardness_posonly}]
We prove NP-hardness by reducing the max-$K$-cover problem to the problem below.

\begin{problem}
\label{pr:posonly_problem}
Consider a bipartite graph $\mathcal{G} = (\mathcal{X} \cup \Ts, E)$ with agents 
$\mathcal{X}$ and targets $\Ts$, where each target $t \in \Ts$ has a label $f(t) \in \{+1,-1\}$. 
For each agent $x \in \mathcal{X}$, let their neighborhood be $N(x) = \{ t \in \Ts \mid (x,t) \in E \}$. The goal of the $\splan$ is to find a subset $S \subseteq \Ts^{+}$ of at most $K$ positively labeled targets, where $\Ts^{+} = \{t \in \Ts \mid f(t)=+1 \}$ that, when revealed, maximizes social welfare $F(S) = \sum_{x \in \mathcal{X}} Q^{S}(x)$.  The decision problem asks whether there exists such a positive target subset $S$ with $F(S) \ge W$ where $W$ is the welfare threshold.
\end{problem}

We prove the NP-hardness by reducing the max-$K$-cover problem with varied sized sets to Problem~\ref{pr:posonly_problem}.
In the max-$K$-cover problem, we are given a universe of elements $U = \{e_1,\dots,e_n\}$, a family of sets $\mathcal{C} = \{C_1,\dots,C_m\}$ with $C_j \subseteq U$, and a budget of $K$. The goal is to select at most $K$ sets out of $\mathcal{C}$ whose union covers at least $\mathcal{E}$ elements in the universe.

First, we reduce max-$K$-cover to Problem~\ref{pr:posonly_problem} by constructing, in polynomial time, an instance in which selecting a positive target set $S$ with $|S|\le K$ corresponds exactly to choosing up to $K$ sets in the max-$K$-cover instance, and the resulting social welfare reflects the achieved coverage. 
For each element $e_i \in U$ we create an agent $x_i$, and for each set $C_j \in \mathcal{C},$ a positive target $t_j$, with an edge $(x_i,t_j)$ if and only if $e_i \in C_j$. To ensure that all agents have a similar initial contribution to social welfare, each agent $x_i$ with neighborhood size $|N(x_i)|$ is additionally connected to $|N(x_i)|$ unique negative targets, distinct across agents. 
In this construction, before any positive targets are revealed, each agent contributes $\frac{1}{2}$ to social welfare, so $F(\emptyset)=\frac{n}{2}$. Revealing a positive target $t_j$ increases the contribution of every adjacent agent from $\frac{1}{2}$ to $1$. 

Then, for any revealed set $S\subseteq \Ts^+$ where $|S_g| \leq K$,
\[
F(S) = \frac{n}{2} + \frac{1}{2}\left|\bigcup_{t_j \in S} C_j\right|.
\]
Setting the welfare threshold to $W = \frac{n}{2} + \frac{\mathcal{E}}{2}$, the condition $F(S) \geq W$ is equivalent to
$\left|\bigcup_{t_j \in S} C_j\right| \geq \mathcal{E}.$ Thus, there exists a set of at most $K$ revealed positive targets achieving welfare at least $W$ if and only if there exist at most $K$ sets covering at least 
$\mathcal{E}$ elements in the original instance. Since $F(S)$ is computable in polynomial time, Problem~\ref{pr:posonly_problem} lies in NP, its decision version is NP-complete, and the corresponding optimization problem of selecting $S_g\subseteq \Ts^+$ with $|S_g|\le K$ to maximize social welfare is NP-hard.

Next, we show that the reduction preserves approximation hardness. Since the max-$K$-cover problem is NP-hard to approximate within any factor strictly greater than $1-\frac{1}{e}$ unless $\mathrm{P}=\mathrm{NP}$, any approximation for Problem~\ref{pr:posonly_problem} would yield an approximation of the same quality for max-$K$-cover. 
In Problem~\ref{pr:posonly_problem}, for any positive target solution set $S$ and an optimal solution set $S^\star$, the construction ensures that
$\frac{F(S)-\frac{n}{2}}{F(S^\star)-\frac{n}{2}} = \frac{\big|\bigcup_{t_j \in S} C_j\big|}{\big|\bigcup_{t_j \in S^\star} C_j\big|}.$
Therefore, a polynomial time $\alpha$-approximation for maximizing social welfare in Problem~\ref{pr:posonly_problem} induces a polynomial time $\alpha$-approximation for maximizing coverage in the max-$K$-cover problem. Consequently, if Problem~\ref{pr:posonly_problem} admitted a polynomial-time approximation factor strictly better than $1-\frac{1}{e}$, then max-$K$-cover would also admit such an approximation, contradicting known hardness results. Hence, unless $\mathrm{P}=\mathrm{NP}$, selecting at most $K$ positive targets to maximize social welfare cannot be approximated within a factor better than $1-\frac{1}{e}$.
\end{proof}

\subsubsection{Proof of Theorem~\ref{thm:np-hardness_negonly}}
\label{sec:revealrm_sm_nphardness-negonly}

\begin{proof}[Proof of Theorem~\ref{thm:np-hardness_negonly}]
We prove NP-hardness by reducing the $K$-clique problem in a graph where all vertices have the same degree $\theta$, to the problem below.

\begin{problem}
\label{pr:negonly_problem}
Consider a bipartite graph $\mathcal{G} = (\mathcal{X} \cup \Ts, E)$ with agents 
$\mathcal{X}$ and targets $\Ts$, where each target $t \in \Ts$ has a label $f(t) \in \{+1,-1\}$. 
For each agent $x \in \mathcal{X}$, let their neighborhood be $N(x) = \{ t \in \Ts \mid (x,t) \in E \}$. The goal of the $\splan$ is to find a subset $S \subseteq \Ts^{-}$ of at most $K$ negatively labeled targets, where $\Ts^{-} = \{t \in \Ts \mid f(t)=-1 \}$ that, when revealed, maximizes social welfare $F(S) = \sum_{x \in \mathcal{X}} Q^{S}(x)$.  The decision problem asks whether there exists such a negative target subset $S$ with $F(S) \ge W$ where $W$ is the welfare threshold?
\end{problem}

We prove the NP-hardness by reducing the $K$-clique problem where all vertices have the same degree $\theta$, to Problem~\ref{pr:negonly_problem}.
In the $K$-clique problem, we are given a graph $\mathcal{G}_c = (V,E_c)$ where $|V|=m$ and $|E_c|=n$. The goal is to find if a clique of size $K$ exists in graph $\mathcal{G}_c$. That is $C\subseteq V, \  |C| \leq K$ such that every pair in $C$ is an edge.

First, we show how to construct an instance of Problem~\ref{pr:negonly_problem} from any instance of the $K$-clique problem in polynomial time, such that revealing a negative target set $S$ with $|S| \leq K$ corresponds exactly to finding a clique of size $K$ in the $K$-clique instance, and the resulting social welfare $F(S)$ is at least the number of edges in the clique.
To do so, we  create one negative target $t_v^-$ for each vertex $v\in V$, one positive target $t_{uv}^+$ for each edge $\{u,v\}\in E_c$, and one agent $x_{uv}$ for each edge $\{u,v\}\in E_c$. Therefore each agent's neighborhood is defined as $N(x_{uv})=\{t_u^-,\, t_v^-,\, t_{uv}^+\}$, and a negative target with no edge has no agent, because if an agent were connected to it, it would contribute $0$ to social welfare.  
The target reveal budget is set to $K$ and the welfare threshold to $W.$

In this construction, each agent initially, before revealing any negative target (i.e., $S=\emptyset$), contributes $\frac{1}{3}$ to the social welfare, that is, $F(\emptyset)=\frac{n}{3}$. Revealing a negative target $t_u^-$ increases the contribution of each agent connected to it by $\frac{1}{6}$ because their probability of emulating a positive target goes from $\frac{1}{3}$ to $\frac{1}{2}$. Revealing two negative targets $\{t_u^-,\, t_v^-\}$ increases the contribution of each agent $x_{uv}$ connected to them by $\frac{2}{3}$ since their probability of emulating a positive target in their neighborhood goes from $\frac{1}{3}$ to $1$. 
Every negative target in the revealed target set $S\subseteq\Ts^-, \ |S| \leq K$ is connected to $K-1$ other negative targets in $S$ and 
$\theta-(K-1)$ negative targets outside the target set $S$. For $K$ negative targets inside $S$, there are $K(\theta-(K-1)) =  K\theta-2\binom{K}{2}$ agents with one endpoint in the $S$, and within $S$, there are $\binom{K}{2} = K(K-1)/2$ agents with $2$ endpoints in $S$.
Hence, the social welfare satisfies
\[
F(S) = \frac{n}{3} + \frac{1}{6} \Bigg(K\theta-2\binom{K}{2}\Bigg) + \frac{2}{3}\binom{K}{2} =  \frac{n}{3} + \frac{K\theta}{6} + \frac{1}{3}\binom{K}{2} ,
\]
so that there exists a $K$-clique \textit{iff} there exists $S$ with $|S|=K$ such that $F(S) \ge W.$
Consequently, Problem~\ref{pr:negonly_problem} is NP-hard because it can be reduced from the $K$-clique problem. Since any candidate positive target set $S$ can be verified in polynomial time, the decision problem is in NP, and therefore NP-complete. Therefore, the problem of finding a positive target subset $S_g \subseteq \Ts$ with $|S_g| \leq K$ that maximizes social welfare is NP-hard.
\end{proof}

\subsection{Alternative Greedy Strategies}\label{sec:revealrm_alternative_greedy_strategies}
Due to the performance limitations of the classic greedy approach for budgeted target selection aimed at maximizing social welfare (as discussed in Section~\ref{subsec:revealrm_approxs}), this section proposes alternative greedy strategies, examines them, and compares their effectiveness with that of the classic method.

\subsubsection{The \textit{d}-step Lookahead Greedy Approach}
\label{sec:revealrm_app-lookahead}

The \(d\)-step lookahead greedy algorithm (Algorithm~\ref{alg:dstep_greedy_lbreveal}) generalizes Algorithm~\ref{alg:greedy_lbreveal} by revealing up to \(d \in \mathbb{Z}_{\geq 1}\) targets per iteration, chosen to maximize the marginal gain in social welfare.

When $d$ in Algorithm~\ref{alg:dstep_greedy_lbreveal} is equivalent to the target reveal budget (\(d = K\)), Algorithm~\ref{alg:dstep_greedy_lbreveal} reduces to bruteforce search (Appendix~\ref{sec:revealrm_app-bruteforce}), which evaluates all \(m^{K}\) possible $K$ sized subsets of targets \(\Ts\) to identify the one that maximizes social welfare.

\begin{proposition}
  \label{prop:opt_runtime_dstep_lookahead_greedy} 
  The \(d\)-step lookahead greedy (Algorithm~\ref{alg:dstep_greedy_lbreveal}) runs in \(O\Big(\frac{K}{d}m^{d}n\delta \Big)\) time.
\end{proposition}

\begin{proof}
At each iteration, Algorithm~\ref{alg:dstep_greedy_lbreveal} reveals a target set of size at most \(d\) with the maximum marginal gain, that is 
$\displaystyle S^{iter} = \arg\!\!\max_{\substack{S \subseteq (\Ts \setminus S_{\mathrm{lg}}) \\ |S| \le b}} \big(F(S_{\mathrm{lg}} \cup S) - F(S_{\mathrm{lg}})\big)$.
To find this subset, among the possible number of subsets \(O(m^d)\), the algorithm evaluates the new social welfare \(F(S_{\text{lg}} \cup S) \) for each candidate subset $S$, which takes \(O(m^d n\delta)\) time in total. 
Since each iteration reveals at most \(d\) targets, then there are at most \(O(\frac{K}{d})\) iterations. Therefore, the total running time is \(O(\frac{K}{d} m^d n\delta )\).
\end{proof}

\begin{algorithm}[htp!]
\caption{The \(d\)-step Lookahead Greedy Target Reveal}
\label{alg:dstep_greedy_lbreveal}
\SetAlgoLined
    \KwIn{Graph $\graph = (\Xs \cup \Ts, E)$, labels $\{f(t)\}_{t \in \Ts}$, budget \(K\), depth \(d\)}
    
    \KwOut{Solution set $S_{\mathrm{lg}} \subseteq \Ts \ \text{with} \ |S_{\mathrm{lg}}| \leq K$, and social welfare $F(S_{\mathrm{lg}})$}
    
    Initialize $S_{\mathrm{lg}} \gets \emptyset$
    
    \While{$|S_{\mathrm{lg}}| \leq K$}{
        $b \gets \min(d, \ K - |S_{\mathrm{lg}}|)$
        
        $\displaystyle S^{iter} \gets \arg\!\max_{\substack{S \subseteq (\Ts \setminus S_{\mathrm{lg}}) \\ |S| \le b}} \big(F(S_{\mathrm{lg}} \cup S) - F(S_{\mathrm{lg}})\big)$
        
        \If{$F(S_{\mathrm{lg}} \cup S^{iter}) > F(S_{\mathrm{lg}})$}{
            $S_{\mathrm{lg}} \gets S_{\mathrm{lg}} \cup S^{iter}$
        }\Else{
            \textbf{break}
        }
    }
    
    \Return $(S_{\mathrm{lg}}, \ F(S_{\mathrm{lg}}))$

\end{algorithm}

\paragraph{Comparison of Classic (Algorithm \ref{alg:greedy_lbreveal}) to Lookahead (Algorithm \ref{alg:dstep_greedy_lbreveal}).}
While classic greedy algorithm runs in polynomial time $(O(Kmn\delta))$ (Proposition~\ref{prop:opt_runtime_greedy}), the $d$-step lookahead greedy algorithm incurs a higher computational cost of $O\Big(\frac{K}{d} m^{d} n \delta \Big)$.

There exist cases (Proposition~\ref{prop:2lookaheadvgreedy}) where at a relatively low computational cost, the $2$-step lookahead significantly outperforms the classic greedy approach. However, as shown in Proposition~\ref{prop:klookaheadvgreedy}, there exist cases where $d$-step lookahead surpasses Algorithm~\ref{alg:greedy_lbreveal} only when the lookahead depth ($d$) is the equal to the target reveal budget ($K$). Therefore, to try and balance good performance with computational efficiency, we propose the heuristic greedy approach (Appendix~\ref{sec:revealrm_app_heuristicgreedy}).

\begin{proposition}
\label{prop:2lookaheadvgreedy}
    There exists a graph and a budget $K$ for which a $2$-step lookahead algorithm finds the exact optimal solution with low computational overhead, while the classic greedy algorithm attains an approximation ratio strictly less than $\frac{2}{\sqrt{n}+1}$.
\end{proposition}
\begin{proof}
    Proof is shown in Example~\ref{ex:lookaheadVclassic} below.
\end{proof}
\begin{example}
\label{ex:lookaheadVclassic}
    Let $\kappa \in \mathbb{N}$ with $\kappa \ge 3$, and set the reveal budget to $K \coloneqq \kappa$.
    Consider the bipartite graph with \(n=\kappa^{2}\) agents, where each agent is adjacent to a unique positive target and to all \(\mneg=\kappa\) negative targets. 
    
    Initially, the social welfare is $F(\emptyset)=\frac{\kappa^{2}}{\kappa+1}.$
    Algorithm~\ref{alg:greedy_lbreveal} is indifferent in the first step: revealing either a positive or a negative target yields the same marginal gain. That is, \(\Big(\frac{\kappa^2}{k}=\frac{\kappa^2 + \kappa}{\kappa+1} = \kappa\Big) -  \frac{\kappa^2}{\kappa+1}.\)
    Once this tie appears, the initial choice dictates the entire trajectory. An initial positive target reveal leads Algorithm~\ref{alg:greedy_lbreveal} to keep selecting positive targets and would require \(\kappa^{2}\) iterations for Algorithm~\ref{alg:greedy_lbreveal} to achieve the optimal social welfare. On the other hand, an initial negative target reveal commits it to negative targets and reaches the optimum in only \(\kappa\) iterations.
    
    In contrast, Algorithm \ref{alg:dstep_greedy_lbreveal} with \(d=2\) foresees these outcomes and consistently selects negative targets. As a result, it attains the optimal solution for every \(\kappa>1\) consistent with this bipartite graph structure. This shows that a two-step lookahead greedy approach, while remaining comparatively inexpensive to compute, can guarantee an exact solution. Algorithm~\ref{alg:greedy_lbreveal} doesn't. When it commits to positive targets, its approximation ratio can be less than \(\frac{2}{\sqrt{n}+1}\).
\end{example}

\begin{proposition}
\label{prop:klookaheadvgreedy}
    There exists a graph and a budget $K$ for which a $K$-step lookahead algorithm finds the exact optimal solution with high computational overhead, while the classic greedy algorithm attains an approximation ratio strictly less than 
    $1/\sqrt{0.5 n}$.
\end{proposition}
\begin{proof}
    Proof is shown in Example~\ref{ex:exp_lookaheadVclassic} below.
\end{proof}

\begin{example}
\label{ex:exp_lookaheadVclassic}
    Let $\kappa \in \mathbb{N}$ with $\kappa \geq 7$, and set the reveal budget to $K \coloneqq \kappa + 1$.
    Consider the bipartite graph with \(n = \lfloor \kappa^2 / (\kappa-2) \rfloor + \kappa\) agents, each connected to a unique positive target and to all \(\kappa + 1\) negative targets. 
    
    Initially, the social welfare is \((\lfloor \kappa^2 / (\kappa-2) \rfloor + \kappa)/(\kappa+2)\). 
    Selecting any negative target increases welfare to \((\lfloor \kappa^2 / (\kappa-2) \rfloor + \kappa)/(\kappa+1)\), whereas selecting a positive target results in larger increase 
    \((2\kappa + 1 + \lfloor \kappa^2 / (\kappa-2) \rfloor)/(\kappa+2)\). 
    As a result, Algorithm~\ref{alg:greedy_lbreveal} only reveals positive targets, despite the negative ones being better in the long run, capable of achieving the optimal welfare of \(\lfloor \kappa^2 / (\kappa-2) \rfloor + \kappa\) at budget \(K\). 
    
    With only positive targets revealed, \(\kappa + 1\) agents emulate a positive target with probability one, while the remaining agents
    do so with probability \(1/(\kappa+2)\). 
    This yields an approximation ratio of \(\frac{\kappa+1 + \frac{\lfloor \frac{\kappa^2}{\kappa-2} \rfloor + \kappa -(\kappa+1)}{\kappa+2}}{\kappa^2} < \frac{1}{\kappa} < 1/\sqrt{0.5 n}\).
    
    In contrast, Algorithm~\ref{alg:dstep_greedy_lbreveal} achieves an exact solution for \(\kappa \ge 7\) when its depth 
    \(d = K\). While it performs significantly better than the classic greedy algorithm, it incurs substantially higher computational cost.
\end{example}

\begin{algorithm}[ht!]
\caption{Heuristic Greedy Target Reveal}
\label{alg:heuristic_greedy_lbreveal}
\SetAlgoLined

    \KwIn{Graph $\graph = (\Xs \cup \Ts, E)$,  labels $\{f(t)\}_{t \in \Ts}$, budget \(K\)}
    \KwOut{Solution set $S_{\mathrm{hg}}  \subseteq \Ts  \ \text{with} \ |S_{\mathrm{hg}} | \leq K$, and social welfare $F(S_{\mathrm{hg}} )$}
    
    $\mathcal{R} \leftarrow \emptyset$
    
    \For{$\kappa = 0$ to $K$}{
        
        $(S_+^{(\kappa)}, F_+^{(\kappa)}) \gets \textsc{GreedyLabelReveal}(\graph, \Ts^+, f, \kappa)$
        \Comment{Algorithm~\ref{alg:greedy_lbreveal} with \(\Ts'\!=\!\Ts^{+}\)}
        
        $(S_-^{(\kappa)}, F_-^{(\kappa)}) \gets \textsc{GreedyLabelReveal}(\graph, \Ts^-, f, K-\kappa)$
        \Comment{Algorithm~\ref{alg:greedy_lbreveal} with \(\Ts'\!=\!\Ts^{-}\)}
    
        $S^{(\kappa)} \gets S_+^{(\kappa)} \cup S_-^{(\kappa)}$
        
        $F^{(\kappa)} \gets F(S^{(\kappa)})$
        
        $\mathcal{R} \gets \mathcal{R} \cup \{(\kappa, S^{(\kappa)}, F^{(\kappa)})\}$
    }
    
    $(\kappa_{\mathrm{hg}}, S_{\mathrm{hg}}, F_{\mathrm{hg}}) \leftarrow \displaystyle \arg\max_{(\kappa, S^{(\kappa)}, F^{(\kappa)}) \in \mathcal{R}} F^{(\kappa)}$
    
    \Return $(S_{\mathrm{hg}}, F(S_{\mathrm{hg}}))$

\end{algorithm}

\subsubsection{The Heuristic Greedy Approach}\label{sec:revealrm_app_heuristicgreedy}
In this section, we present the heuristic greedy approach. Although the proposed heuristic algorithm is flexible and can incorporate various algorithms, here we focus on the setting in which the inserted algorithm is the classic greedy method (Algorithm~\ref{alg:greedy_lbreveal}).

\paragraph{Overview of heuristic greedy algorithm (Algorithm~\ref{alg:heuristic_greedy_lbreveal}).} 
The heuristic greedy algorithm proceeds as follows. Given a budget \(K\), the heuristic greedy algorithm considers all budget splits \(\kappa \in \{0,\ldots,K\}\). For each \(\kappa\), Algorithm~\ref{alg:greedy_lbreveal} is run in parallel with budget \(\kappa\) and restriction to positive targets \(\Ts'=\Ts^{+}=\{t\in\Ts \mid f(t)=+1\}\), producing \(S_{+}^{\kappa}\), and with budget \(K-\kappa\) and restriction to negative targets \(\Ts'=\Ts^{-}=\{t\in\Ts \mid f(t)=-1\}\), producing \(S_{-}^{K-\kappa}\). The algorithm returns the \(S_{\mathrm{hg}}\) that maximizes the social welfare, i.e.,
\[S_{\mathrm{hg}}  = \arg\!\max_{\kappa \in [0,K]} F\Big(S_+^{(\kappa)} \cup S_-^{(K-\kappa)}\Big)\]

\paragraph{The heuristic greedy algorithm runs in polynomial time.} 
Proposition~\ref{prop:opt_runtime_heuristic_greedy_lbreveal} shows that 
Algorithm~\ref{alg:heuristic_greedy_lbreveal} runs in \(O(K^2mn\delta )\) time, where  $\delta$ is the maximum agent degree, $m$ the number of targets, and $n$ the number of agents.

\begin{proposition}
The heuristic greedy algorithm (Algorithm~\ref{alg:heuristic_greedy_lbreveal}) runs in \(O\Big(K^{2}mn\delta \Big)\) time.
\label{prop:opt_runtime_heuristic_greedy_lbreveal}
\end{proposition}

\begin{proof}
At each iteration, Algorithm \ref{alg:greedy_lbreveal} is executed separately on the positive and negative targets.
If each of the positive and negative target sets has size of at most \(m\), then for any fixed iteration and reveal budget of
$\kappa \in \{0,\dots, K\}$, the separate Algorithm \ref{alg:greedy_lbreveal} calls require 
$O(\kappa m n \delta) + O((K-\kappa) m n \delta)$ time, equal to $O(K m n \delta)$.
All remaining operations within the loop take constant time.
After the $K$ iterations, the total cost becomes $O(K^{2} m n \delta).$
To compute among the solutions, the one with the optimal social welfare would take $O(K)$ time.
Thus, Algorithm \ref{alg:heuristic_greedy_lbreveal} runs in $O(K^{2} m n \delta)$ time.
\end{proof}

\subsubsection{The Interactive Heuristic Greedy Approach}
\label{sec:revealrm_app_int_heuristicgreedy}
Unlike the heuristic greedy approach which runs Algorithm \ref{alg:greedy_lbreveal} independently on the positive and negative targets, we analyze the interactive variant that couples the two phases. 

The interactive heuristic greedy algorithm first applies Algorithm \ref{alg:greedy_lbreveal} to either the positive or negative targets for a budget of \(\kappa \in \{0,\ldots,K\}\). Then the revealed target set produced in this step becomes the initial set \(S'\) for a second run of Algorithm \ref{alg:greedy_lbreveal} on the opposite target set, using the remaining budget \(K-\kappa\). Below is the outline of the procedure.

\begin{algorithm}[ht!]
\caption{Interactive Heuristic Greedy Target Reveal}
\label{alg:inter_heuristic_greedy_lbreveal}
\SetAlgoLined

\KwIn{Graph $\graph = (\Xs \cup \Ts, E)$, labels $\{f(t)\}_{t \in \Ts}$, budget \(K\)}
\KwOut{Solution set $S_{\mathrm{ihg}}  \subseteq \Ts \ \text{with} \ |S_{\mathrm{ihg}} | \leq K$ and social welfare $F(S_{\mathrm{ihg}} )$}

$\mathcal{R} \leftarrow \emptyset$

\For{$\kappa = 0,1,\ldots,K$}{
  $(S_{+}^{(\kappa)}, F_{+}^{(\kappa)}) \leftarrow \textsc{GreedyLabelReveal}(\graph, \Ts^+, f, \kappa)$
  
  $(S_{-}^{(\kappa)}, F_{-}^{(\kappa)}) \leftarrow \textsc{GreedyLabelReveal}(\graph, \Ts^-, f, \kappa)$

  $(S_{+}^{i(\kappa)}, F_{+}^{i(\kappa)}) \leftarrow \textsc{GreedyLabelReveal}(\graph, \Ts^+, f, K-\kappa, S'=S_{-}^{(\kappa)})$

  $(S_{-}^{i(\kappa)}, F_{-}^{i(\kappa)}) \leftarrow \textsc{GreedyLabelReveal}(\graph, \Ts^-, f, K-\kappa, S'=S_{+}^{(\kappa)})$  

  $\mathcal{R} \leftarrow \mathcal{R} \cup \{(\kappa, S_{+}^{i(\kappa)}, F_{+}^{i(\kappa)}, S_{-}^{i(\kappa)}, F_{-}^{i(\kappa)})\}$
}

$(\kappa^{\star}, v^{\star}) \leftarrow \displaystyle \arg\max_{\kappa \in \{0,\ldots,K\}, \; v \in \{+, -\}} F_{v}^{i(\kappa)}$

$(S_{\mathrm{ihg}}, F_{\mathrm{ihg}})  \leftarrow (S_{v^{\star}}^{i(\kappa^{\star})}, F_{v^{\star}}^{i(\kappa^{\star})})$

\Return $(S_{\mathrm{ihg}}, F(S_{\mathrm{ihg}}))$

\end{algorithm}

\paragraph{Overview of Algorithm~\ref{alg:inter_heuristic_greedy_lbreveal}.} 
There are two settings we consider. In the first, Algorithm \ref{alg:greedy_lbreveal} is run on the positive targets before the negative targets. In the second case, the algorithm is run on the negative targets first, followed by a run on the positive targets for the remaining budget.

\textbf{Case 1:} Let \(S_+^{(\kappa)}\) be the solution of Algorithm~\ref{alg:greedy_lbreveal} when \(\Ts'=\Ts^{+}\) at a given budget \(\kappa\). Then, given the initialization \(S'=S_+^{(\kappa)}\), let \(S_{-}^{i(\kappa)}\) be the remaining part of solution set of Algorithm~\ref{alg:greedy_lbreveal} when \(\Ts'=\Ts^{-}\) at a given budget \(K-\kappa\) such that \(| S_{-}^{i(\kappa)} | \leq K\). 
That is, 
\[
S_{-}^{i(\kappa)} =   \textsc{GreedyLabelReveal}(\graph, \Ts^-, f, K-\kappa, S'=S_{+}^{(\kappa)})
\]

\textbf{Case 2:} 
Let \(S_-^{(\kappa)}\) be the solution of Algorithm~\ref{alg:greedy_lbreveal} when \(\Ts'=\Ts^{-}\) at a given budget \(\kappa\). Then, given the initialization \(S'=S_-^{(\kappa)}\), 
let \(S_{+}^{i(\kappa)}\) be the remaining part of solution set of Algorithm~\ref{alg:greedy_lbreveal} when \(\Ts'=\Ts^{+}\) at a given budget \(K-\kappa\) such that \(| S_{+}^{i(\kappa)} | \leq K\). 
That is, 
\[
S_{+}^{i(\kappa)} =   \textsc{GreedyLabelReveal}(\graph, \Ts^+, f, K-\kappa, S'=S_{-}^{(\kappa)})
\]
Run cases 1 and 2 until \(\kappa=K\). Then, for each case, find the best solution, that is, 
\[
\kappa_{+} = \arg\max_{\kappa \in \{0,\ldots, K\}}  F\bigl(S_{+}^{i(\kappa)}\bigr), \quad
\kappa_{-} = \arg\max_{\kappa \in \{0,\ldots, K\}}  F\bigl(S_{-}^{i(\kappa)}\bigr),
\]
Finally
\[
S_{\mathrm{ihg}}, F(S_{\mathrm{ihg}}) =
\begin{cases}
S_{-}^{i(\kappa_{-})}, F(S_{-}^{i(\kappa_{-})}), & \mathrm{if} \ F\bigl(S_{-}^{i(\kappa_{-})}\bigr) > F\bigl(S_{+}^{i(\kappa_{+})}\bigr),\\[4pt]
S_{+}^{i(\kappa_{+})}, F(S_{+}^{i(\kappa_{+})}), & \mathrm{otherwise.}
\end{cases}
\]

\subsection{Comparison of the Greedy Strategies}
\label{sec:revealrm_app-comp-gstrategies}
Similar to Algorithm~\ref{alg:greedy_lbreveal}, Algorithms~\ref{alg:heuristic_greedy_lbreveal}~and~\ref{alg:inter_heuristic_greedy_lbreveal} have polynomial-time complexity, making them significantly more efficient than the \(d\)-step lookahead greedy algorithm.

As shown in Proposition~\ref{prop:handLBc}, the heuristic greedy algorithm can outperform the classic greedy approach and achieve performance comparable to that of the 
\(d\)-step lookahead algorithm. Nevertheless, there are instances where it performs no better than the classic greedy method, while the \(d\)-step lookahead algorithm with 
\(d = K\) produces the optimal solution (Proposition~\ref{prop:LBhandc}), albeit at a substantially higher computational cost.

Notably, when revealed targets include negative ones and the social welfare function $F$ maybe supermodular, greedy methods face a trade-off between performance and computational cost. Therefore to asses practical performance, we empirically evaluate the classic, heuristic, and $K$-step lookahead greedy algorithms in a semi-synthetic setting (Section~\ref{sec:revealrm_experiments}).

\begin{proposition}
\label{prop:handLBc}
There exists graphs and target reveal budget values for which the heuristic greedy algorithm strictly outperforms the classic greedy approach and achieves performance comparable to that of the \(d\)-step lookahead algorithm.
\end{proposition}

\begin{proof}
In Examples~\ref{ex:revealposneg_notsub}, \ref{ex:lookaheadVclassic}, and \ref{ex:exp_lookaheadVclassic}, both the heuristic greedy algorithm and the \(d\)-step lookahead greedy algorithm recover the optimal solution exactly. In contrast, Algorithm~\ref{alg:greedy_lbreveal} achieves approximation ratios strictly smaller than
\(\frac{2}{\sqrt{n}+2}, \ \frac{2}{\sqrt{n}+1}\), and \(\frac{1}{\sqrt{0.5 n}}\), respectively.
\end{proof}

\begin{proposition}
\label{prop:LBhandc}
There exists a graph and a budget $K$ for which the \(K\)-step lookahead greedy algorithm strictly outperforms the classic and heuristic greedy approaches.
\end{proposition}

\begin{proof}
    In Example~\ref{ex:revealonlynegs_notsub}, both the heuristic and the classic greedy algorithms are misled into selecting a suboptimal sequence of targets, yielding a solution with approximation ratio strictly smaller than \(\frac{3}{\sqrt{2n}}\). In contrast, the \(d\)-step lookahead greedy algorithm with \(d = K\) anticipates these unfavorable choices and avoids suboptimal trajectories, thereby recovering the optimal solution, though at a substantially higher computational cost.
\end{proof}

\begin{proposition}
\label{prop:hBgi}
There exists a graph and budget $K$ for which both the classic greedy and interactive heuristic greedy algorithms find an exact solution, whereas heuristic greedy doesn't.
\end{proposition}
\begin{proof}
    Proof is in Example~\ref{ex:comp-it-hg-g} below.
    \begin{example}
    \label{ex:comp-it-hg-g}
        Consider the bipartite graph in Table~\ref{tab:itgraph}, with ten agents and nine targets. Five targets are positive, \(\{t_0^{+}, t_1^{+}, t_2^{+}, t_3^{+}, t_5^{+}\}\), and four are negative, \(\{t_6^{-}, t_7^{-}, t_8^{-}, t_9^{-}\}\). Each row lists an agent along with its neighborhood.
            
        For a budget of $K=3$, the classic greedy and interactive heuristic greedy algorithms both yield an optimal target set \(\{t_0^{+},t_6^{-},t_9^{-}\}\), where \(\{t_0^{+}\}\) is positive and \(\{t_6^{-}, t_9^{-}\}\) are negative. In contrast, the heuristic greedy algorithm selects only negative targets \(\{t_6^{-},t_7^{-},t_9^{-}\}\), achieving an approximation ratio of $0.96$.
    \end{example}
\end{proof}

\begin{table}[ht!]
    \centering
    \begin{tabular}{c|l}
        \hline
        agent & neighborhood \\
        \hline
        \(x_0\) & \(\{ t_8^{-}, t_9^{-} \}\) \\
        \(x_1\) & \(\{ t_3^{+}, t_6^{-}, t_9^{-} \}\) \\
        \(x_2\) & \(\{ t_0^{+}, t_6^{-}, t_7^{-}, t_8^{-}, t_9^{-} \}\) \\
        \(x_3\) & \(\{ t_5^{+}, t_9^{-} \}\) \\
        \(x_4\) & \(\{ t_7^{-} \}\) \\
        \(x_5\) & \(\{ t_6^{-}, t_7^{-}, t_9^{-} \}\) \\
        \(x_6\) & \(\{ t_9^{-} \}\) \\
        \(x_7\) & \(\{ t_1^{+}, t_2^{+}, t_6^{-}, t_7^{-} \}\) \\
        \(x_8\) & \(\{ t_7^{-}, t_8^{-} \}\) \\
        \(x_9\) & \(\{ t_1^{+}, t_9^{-} \}\) \\
        \hline
    \end{tabular}
    \caption{A bipartite graph where each agent (first column) is connected to its neighborhood of targets (second column) with labels initially unknown to the agents.}
    \label{tab:itgraph}
\end{table}

\section{Supplementary Material for Section~\ref{sec:revealrm_intervention_model}}
\label{sec:revealrm_app_interventionmodel_algos}

In this section, we present pre- and post-reveal intervention algorithms (i.e., Algorithms~\ref{alg:TI_greedy_label_reveal} and \ref{alg:greedy_TI_label_reveal}, respectively) along with their runtime analysis.

\begin{algorithm}[ht!]
\caption{Pre-reveal Targeted Intervention}
\label{alg:TI_greedy_label_reveal}
\SetAlgoLined

    \KwIn{Graph $\graph = (\Xs \cup \Ts, E)$, labels $\{f(t)\}_{t \in \Ts}$, reveal budget $K$, intervene budget $B$}
    \KwOut{Solution set $S_{\mathrm{ig}}$, and pre-reveal intervention social welfare $F(S_{\mathrm{ig}})$}
    
    $S_o \gets \emptyset$
    
    \For{each agent $x \in \mathcal{X}$}{
        $ Q^{S_o}(x) \gets |\{t\in N(x): f(t)=1\}| / |N(x)|$
    }
    
    $B'\gets \mathbf{1}[\exists\,t\in\Ts:f(t)=1]\cdot\min \bigl(B, \ |\{x\in \Xs: \ Q^{S_{o}}(x)<1\}|\bigr)$
    
    $\displaystyle \mathcal{X}_{\text{hr}}\gets \operatorname*{arg\,min}_{\mathcal{X}' \subseteq \mathcal{X}\ | \ |\mathcal{X}'| = B'}  \sum_{x \in \mathcal{X}'} Q^{S_{o}}(x)$
    
    $\graph' = \graph[\mathcal{X} \setminus \mathcal{X}_{\text{hr}}]$
    
    $(S_{\mathrm{ig}}, F(S)) \gets \textsc{GreedyLabelReveal}(\graph', \Ts, f, K)$
    \Comment{run Algorithm~\ref{alg:greedy_lbreveal}}
    
    $\displaystyle F(S_{\mathrm{ig}}) \gets F(S) + B'$ 
    
    \Return $(S_{\mathrm{ig}}, F(S_{\mathrm{ig}}))$

\end{algorithm}

\begin{algorithm}[ht!]
\caption{Post-reveal Targeted  Intervention}
\label{alg:greedy_TI_label_reveal}
\SetAlgoLined
    \KwIn{Graph $\graph = (\Xs \cup \Ts, E)$, labels $\{f(t)\}_{t \in \Ts}$, reveal budget $K$, intervene budget $B$}
    \KwOut{Solution set $S_{\mathrm{gi}}$, and post-reveal intervention social welfare $F(S_{\mathrm{gi}})$}
    
    $(S_o, F(S_o)) \gets \textsc{GreedyLabelReveal}(\graph, \Ts, f, K)$ \Comment{run Algorithm~\ref{alg:greedy_lbreveal}}
    
    \For{each agent $x \in \mathcal{X}$}{
        $Q^{S_o}(x)\gets
        \begin{cases}
        1, & \text{if }|\{t\in N(x)\cap S_o:f(t)=1\}|>0,\\
        \dfrac{|\{t\in N(x)\setminus S_o:f(t)=1\}|}{|\{t\in N(x)\setminus S_o\}|}, & \text{if }|N(x)\setminus S_o|>0,\\
        0, & \text{otherwise.}
        \end{cases}$
    }
    $B'\gets \mathbf{1}[\exists\,t\in\Ts:f(t)=1]\cdot\min \bigl(B, \ |\{x\in \Xs: \ Q^{S_{o}}(x)<1\}|\bigr)$
    
    $\displaystyle\mathcal{X}_{\text{hr}}\gets \operatorname*{arg\,min}_{\mathcal{X}' \subseteq \mathcal{X}\ | \ |\mathcal{X}'| = B'}  \sum_{x \in \mathcal{X}'}Q^{S_{o}}(x)$
    
    $F(S_{\mathrm{gi}}) \gets F(S_o) + \sum_{x \in  \mathcal{X}_{\text{hr}}} (1 -Q^{S_{o}}(x))$
    
    $S_{\mathrm{gi}} \gets S_o$
    
    \Return $(S_{\mathrm{gi}}, F(S_{\mathrm{gi}}))$

\end{algorithm}

\paragraph{Runtime analysis for the targeted intervention model algorithms.}
Algorithms~\ref{alg:TI_greedy_label_reveal} and \ref{alg:greedy_TI_label_reveal} run in \(O(Kmn\delta)\) time, where \(\delta\) is the maximum agent degree, \(m\) the number of targets, and \(n\) the number of agents.

\begin{proposition}
Algorithms~\ref{alg:TI_greedy_label_reveal} and \ref{alg:greedy_TI_label_reveal} each runs in \(O(Kmn\delta)\) time.
\label{prop:opt_runtime_TI_greedy}
\end{proposition}

\begin{proof}
Focusing on the costly procedures in Algorithms~\ref{alg:TI_greedy_label_reveal} and \ref{alg:greedy_TI_label_reveal}, each of them requires running the classic greedy algorithm sub module for at most \(n\) agents, taking \(O(Kmn\delta)\) time. To identify the high-risk agents, both algorithms first compute  \(Q^{S_o}(x)\) for every agent, which adds \(O(n\delta)\) to the computation time. Then ranking agents by \(Q^{S_o}(x)\) value, would take \(O(n \log n)\) time. Combining these terms results in a total running time of $O(Kmn\delta + n\delta + n\log n) = O(Kmn\delta).$
\end{proof}

\section{Supplementary Material for Section~\ref{sec:revealrm_coveragemodel}}
\label{sec:revealrm_app_coveragemodel_algos}

\paragraph{Overview of Algorithm~\ref{alg:radius_greedy_coverage}.}  
The coverage radius algorithm (Algorithm~\ref{alg:radius_greedy_coverage}) proceeds as follows. Let $c \in \{0,1\}^{n}$ denote coverage of $n$ agents where $1$ means that the agent is covered or reached, and \(0\) otherwise. 
For each target $t_{i}$, compute the distance to the nearest uncovered agent relative to the target's current radius $r_{i}$ that is, 
$\displaystyle g_{i} = \min_{j: c_{j} = 0} \|t_i-x_j\|_2 - r_i.$ 
Then, increase the radius $r_k$ of the target  $t_k$ with the smallest cost by that cost $r_k = r_k + g_k,$ update the remaining radius budget by subtracting the cost of coverage, and mark corresponding agent(s) as covered. Continue until the radius budget is exhausted.

\begin{proposition}
Algorithm~\ref{alg:radius_greedy_coverage} runs in \(O(mn(\log n + d))\) time.
\label{prop:opt_runtime_covmodel}
\end{proposition}

\begin{proof}
Algorithm~\ref{alg:radius_greedy_coverage} begins by computing all pairwise distances between the $m$ positive targets and $n$ agents, forming the matrix $D \in \mathbb{R}^{m\times n}$. This step requires $O(mnd)$ time. Given these distances, it then sorts, for each positive target, the distances to all $n$ agents, which costs $O(mn\log n)$ overall. The while loop contributes at most $O(mn)$. Combining all parts, the total running time is
$O(mnd) + O(mn\log n) + O(mn) = O\bigl(mn(d+\log n)\bigr).$
\end{proof}

\begin{algorithm}[ht!]
\caption{Greedy Coverage Radius}
\label{alg:radius_greedy_coverage}
\SetAlgoLined
    \KwIn{Agents $\mathcal{X}\in\mathbb{R}^{n \times \rho}$, targets $\Ts^{+}\in\mathbb{R}^{m \times \rho}$, radius budget $R\in\mathbb{R}_{\geq 0}$}
    \KwOut{Radii $r$, Covered agents $c$}
    
    $D \gets [\,d_{ij}\,]_{i,j} \in \mathbb{R}^{m\times n}$
    
    $r_i \gets 0, \  \forall i \in [m]$, \quad $c_j \gets 0, \ \forall j \in [n]$
    
    $R' \gets R$ \Comment{remaining R budget}
    
    $(\mathit{sorted\_dist}, \mathit{sorted\_idx}) \gets \textit{sort\_along\_rows}(D)$
    
    $\mathit{ptr} \gets \mathbf{0}_m$
    
    \While{$R' > 0$}{
    
        $\mathit{cost}_{i} \gets +\infty, \ \forall i \in [m]$
    
        \For{$i = 1$ to $m$}{
            \While{$\mathit{ptr}_{i} < n$ and $c_{\mathit{sorted\_idx}_i[\mathit{ptr}_i]}$}{
                $\mathit{ptr}_{i} \gets \mathit{ptr}_{i} + 1$
            }
            \If{$\mathit{ptr}_{i} < n$}{
                $\mathit{cost}_{i} \gets \mathit{sorted\_dist}_{i}[\mathit{ptr}_{i}] - r_{i}$
            }
        }
        $\displaystyle t \gets \arg\min_{i}\mathit{cost}_{i}$
    
        \If{$\mathit{cost}_{t} = +\infty$ or $\mathit{cost}_{t} > R'$}{
            \textbf{break}
        }
        $r_{t}\gets r_{t} + \mathit{cost}_{t}$
        
        $R' \gets R' - \mathit{cost}_{t}$
        
        $a \gets \mathit{sorted\_idx}_{t}[\mathit{ptr}_{t}]$
        
        $c_a \gets 1$
        
        $\mathit{ptr}_{t} \gets \mathit{ptr}_{t} + 1$
    }
    
    \Return $(r, c)$

\end{algorithm}

\section{Proof of Theorem~\ref{thm:sc_bound}}
\label{sec:revealrm_app_learningsetting}

\begin{proof}
Let $x_1,\dots,x_n\sim D$ be set of agents drawn i.i.d from \(D\). For any fixed revealed target set 
$S \subseteq \Ts,$ we can compute $Q^S(x_i) \in [0,1]$ for each agent. 
As a result, the empirical social welfare is given as $\hat F(S)=\frac{1}{n}\sum_{i=1}^n Q^S(x_i).$
Consider that the revealed target set $S \subseteq \Ts$ has error at least $\varepsilon$ for distribution $D$. That is, for a fixed revealed target set $S$, $\big|\hat F(S)-F(S)\big|\le\varepsilon$ with probability at least 
$1-\delta$. With a target reveal budget of $K$, each of the subsets $S$ will be of size at most $K$, and therefore, there is atmost $m^K$ potential subsets.

By Hoeffding's inequality, the probability that the revealed target set will have social welfare off by more than $\varepsilon$ can be bounded as follows, $\Pr\bigl(|\hat F(S)-F(S)|\ge\varepsilon\bigr)\le 2\exp(-2n\varepsilon^2).$ By union bound over all the possible subsets $m^K,$ then $\Pr\bigl(\exists S\in\mathcal \Ts:\ |\hat F(S)-F(S)|\ge\varepsilon\bigr) \le 2m^{K}\exp(-2n\varepsilon^2).$ 
To ensure this probability is at most $\delta,$ it suffices that $2m^{K}\exp(-2n\varepsilon^2) \le \delta,$ which holds whenever $n \ge C\Bigl(\frac{1}{\varepsilon^2}\bigl(K\log m + \log\tfrac{1}{\delta}\bigr)\Bigr)$ for a suitable universal constant $C>0$.  
Under this condition, all revealed target sets $S$ of size at most $K$, including $S_g$, have empirical social welfare within $\varepsilon$ of their true value with probability at least $1-\delta$.
\end{proof}

\section{Supplementary Material for Section~\ref{sec:revealrm_experiments}}
\label{sec:revealrm_app-exps}

All experimentation, including bipartite graph generation, algorithmic computations and comparative analytics were performed on a CPU-based system with the following specifications: a 2.6-GHz 6-Core Intel Core i7 processor, 16 GB of 2400-MHz DDR4 RAM, and an Intel UHD Graphics 630 GPU with 1536 MB of memory. 

\subsection{Experimental Setup}
\label{sec:revealrm_app-setupexps}

\paragraph{Datasets.}\label{subsec:revealrm_app-datasets}
We utilized four datasets obtained from the UCI Machine Learning Repository. The first was the \textbf{Adult} (Adult Income) dataset \citep{BeckerBK}. From this, we selected the following features: \textit{age, workclass, fnlwgt, education, education-num, marital-status, occupation, relationship, race, sex, capital-gain, capital-loss, hours-per-week, native-country,} and \textit{income}. The target variable, ``\textit{target}'', was defined as \(1\) if the ``\textit{income}'' value was was greater than \(50k\); otherwise, it was \(0\). Afterward, we removed the ``\textit{income}'' variable from the data features. 

The second dataset was \textbf{Productivity} (Garment Worker Productivity) \citep{UCIImranAAR21,ImranAAR21}. During preprocessing, we first removed the variables ``\textit{date}'' and ``\textit{day}''. Next, missing values in the ``\textit{wip}'' column, the only feature with missing data, were imputed with zeros. Outliers in the incentive column were then eliminated. The target variable, ``\textit{target}'', was defined as a binary indicator: if the difference between ``\textit{actual\_productivity}'' and ``\textit{targeted\_productivity}'' was greater than or equal to zero, the target was set to \(1\); otherwise to \(0\). Finally, we excluded ``\textit{actual\_productivity}'' and ``\textit{targeted\_productivity}'' from data features.

The third and fourth datasets were derived from the Student Performance dataset \citep{CortezP08}, specifically the \textbf{Portuguese} (Student-por) and \textbf{Math} (Student-mat) performance subsets. For both datasets, we defined the target variable, ``\textit{pass}'', as \(1\) if the sum of the three grade variables (\textit{G1, G2, G3}) was greater than or equal to \(35\), and \(0\) otherwise. After defining the target, we removed the grade variables from the feature set.

\paragraph{Preparation of datasets for graph generation.}
For all datasets, we label-encoded categorical variables, removed duplicate rows, and, when necessary, applied subsampling to ensure a maximum of $500$ rows. Each dataset was then divided into data features \(\mathcal{X}_{\mathrm{orig}}\) and labels \(y\), after which the features were standardized and transformed. 

To prepare a given dataset for bipartite graph generation, the feature data was randomly partitioned, with  \(90\%\) of the samples assigned to the left-hand side (LHS), $\mathcal{X}_{\mathrm{LHS}}$  and \(10\%\) to the right-hand side (RHS) $\mathcal{X}_{\mathrm{RHS}}.$ 
Labels of the LHS and RHS samples were directly retrieved from \(y\). We then remove all positively labeled samples from the LHS and disregard labels for the remaining samples. We retain all the RHS samples and their labels in the experiments.

\paragraph{Statistics of the Generated Geometric Bipartite Graphs.}
\label{subsec:revealrm_app-graphstats}
For each generated bipartite graph, we report the following statistics: the dataset name (name), number of data features ($\rho$), maximum number of nearest targets in an agent's neighborhood ($k_{\max}$) or threshold for distance between targets and agents in an agents' neighborhood ($\ell$), number of agents ($n$), number of targets positive ($m^+$) and negative ($m^-$) targets, average agent degree (avg.LHS), number of agents with all-positive neighborhoods (only+Ns), all-negative neighborhoods (only-Ns), and empty-neighborhoods (emptyNs), and lastly, the number of positive targets connected to all helpable agents ($\textit{uni}^{+}$). 

For all experiments under the standard and targeted intervention models, the statistical properties of the bipartite graphs remain as described above. In the learning setting, each bipartite graph is split into training and testing sets. The training set contains \(70\%\) of the agents along with their associated edges, while the remaining \(30\%\) form the testing set. The targets and their labels are kept constant across both sets.  For experiments under the coverage radius model, as described above, each dataset was first split into \(\mathcal{X}_{\mathrm{LHS}}\), \(\mathcal{X}_{\mathrm{RHS}}\), and \(y_{\mathrm{RHS}}\). Then, only positive tar from the RHS were selected and initially assigned a radius of zero, so that no edges exist at the start.

\begin{table}[b!]
\centering
\footnotesize
\setlength{\tabcolsep}{3pt}
\renewcommand{\arraystretch}{0.75}
\setlength{\aboverulesep}{0pt}
\setlength{\belowrulesep}{0pt}
\setlength{\textfloatsep}{6pt}
\caption{Statistics of bipartite graphs generated from the \textbf{Adult} ($\rho=14$) dataset. For all graphs, $n=328$.}
\label{tab:adult_kmax_r_stats}
\begin{tabular}{lrrrrrrr}
\addlinespace[0.1cm]
\toprule
\addlinespace[0.1cm]
Param & value & $(m^-, m^+)$ & avg.LHS & only+Ns & only-Ns & emptyNs & $\textit{uni}^{+}$ \\
\addlinespace[0.1cm]
\midrule
\addlinespace[0.1cm]
$k_{\max}$ & 1  & (36,10) & 1.0  & 70 & 258 & 0 & 0 \\
 & 2  & (37,12) & 2.0  & 15 & 184 & 0 & 0 \\
 & 3  & (37,12) & 3.0  & 4  & 121 & 0 & 0 \\
 & 4  & (37,12) & 4.0  & 2  & 86  & 0 & 0 \\
 & 5  & (37,12) & 5.0  & 1  & 61  & 0 & 0 \\
 & 6  & (37,12) & 6.0  & 0  & 44  & 0 & 0 \\
 & 7  & (37,12) & 7.0  & 0  & 33  & 0 & 0 \\
 & 8  & (37,12) & 8.0  & 0  & 19  & 0 & 0 \\
 & 9  & (37,12) & 9.0  & 0  & 12  & 0 & 0 \\
 & 10 & (37,12) & 10.0 & 0  & 10  & 0 & 0 \\
\addlinespace[0.1cm]
\midrule
\addlinespace[0.1cm]
$\ell$ & 4.0  & (36,12) & 12.70 & 6 & 51 & 35 & 0 \\
 & 4.5  & (37,12) & 19.55 & 2 & 33 & 21 & 0 \\
 & 5.0  & (37,12) & 27.00 & 2 & 20 & 9  & 0 \\
 & 5.5  & (37,12) & 33.47 & 0 & 9  & 6  & 0 \\
 & 6.0  & (37,12) & 38.95 & 0 & 6  & 2  & 0 \\
 & 6.5  & (37,12) & 42.83 & 1 & 4  & 0  & 0 \\
 & 7.0  & (37,12) & 45.70 & 0 & 2  & 0  & 0 \\
 & 7.5  & (37,12) & 47.35 & 0 & 0  & 0  & 0 \\
 & 8.0  & (37,12) & 48.26 & 0 & 0  & 0  & 2 \\
 & 8.5  & (37,12) & 48.70 & 0 & 0  & 0  & 6 \\
 & 9.0  & (37,12) & 48.87 & 0 & 0  & 0  & 8 \\
 & 9.5  & (37,12) & 48.96 & 0 & 0  & 0  & 10 \\
 & 10.0 & (37,12) & 48.99 & 0 & 0  & 0  & 12 \\
\bottomrule
\end{tabular}
\end{table}

\begin{table}[b!]
\centering
\footnotesize
\setlength{\tabcolsep}{3pt}
\renewcommand{\arraystretch}{0.75}
\setlength{\aboverulesep}{0pt}
\setlength{\belowrulesep}{0pt}
\setlength{\textfloatsep}{6pt}
\caption{Statistics of bipartite graphs generated from the \textbf{Math} ($\rho=30$) dataset. For all graphs, $n=206$.}
\label{tab:math_kmax_r_stats}
\begin{tabular}{lrrrrrrr}
\addlinespace[0.22cm]
\toprule
\addlinespace[0.22cm]
Param & value & $(m^-, m^+)$ & avg.LHS & only+Ns & only-Ns & emptyNs & $\textit{uni}^{+}$ \\
\addlinespace[0.22cm]
\midrule
\addlinespace[0.22cm]
$k_{\max}$ & 1  & (19, 16) & 1.0  & 106 & 100 & 0 & 0 \\
 & 2  & (21, 17) & 2.0  & 60  & 52  & 0 & 0 \\
 & 3  & (22, 17) & 3.0  & 21  & 30  & 0 & 0 \\
 & 4  & (22, 17) & 4.0  & 13  & 14  & 0 & 0 \\
 & 5  & (22, 17) & 5.0  & 5   & 8   & 0 & 0 \\
 & 6  & (22, 17) & 6.0  & 5   & 6   & 0 & 0 \\
 & 7  & (22, 17) & 7.0  & 1   & 2   & 0 & 0 \\
 & 8  & (22, 17) & 8.0  & 0   & 1   & 0 & 0 \\
 & 9  & (22, 17) & 9.0  & 0   & 0   & 0 & 0 \\
 & 10 & (22, 17) & 10.0 & 0   & 0   & 0 & 0 \\
\addlinespace[0.22cm]
\midrule
\addlinespace[0.22cm]
$\ell$ & 4.0  & (4, 5)   & 0.09  & 10  & 5   & 190 & 1 \\
 & 4.5  & (9, 7)   & 0.21  & 14  & 13  & 174 & 0 \\
 & 5.0  & (12, 12) & 0.58  & 25  & 19  & 145 & 0 \\
 & 5.5  & (17, 15) & 1.71  & 24  & 34  & 97  & 0 \\
 & 6.0  & (21, 17) & 3.63  & 21  & 27  & 61  & 0 \\
 & 6.5  & (22, 17) & 7.09  & 8   & 22  & 33  & 0 \\
 & 7.0  & (22, 17) & 11.81 & 8   & 23  & 9   & 0 \\
 & 7.5  & (22, 17) & 17.55 & 2   & 12  & 3   & 0 \\
 & 8.0  & (22, 17) & 23.30 & 0   & 8   & 1   & 0 \\
 & 8.5  & (22, 17) & 28.29 & 1   & 3   & 0   & 0 \\
 & 9.0  & (22, 17) & 32.38 & 1   & 2   & 0   & 0 \\
 & 9.5  & (22, 17) & 35.23 & 0   & 1   & 0   & 0 \\
 & 10.0 & (23, 17) & 36.98 & 0   & 1   & 0   & 2 \\
\bottomrule
\end{tabular}
\end{table}

\begin{table}[b!]
\centering
\footnotesize
\setlength{\tabcolsep}{3pt}
\renewcommand{\arraystretch}{0.75}
\setlength{\aboverulesep}{0pt}
\setlength{\belowrulesep}{0pt}
\setlength{\textfloatsep}{6pt}
\caption{Statistics of bipartite graphs generated from the \textbf{Portuguese} ($\rho=30$) dataset. For all graphs, $n=224$.}
\label{tab:portuguese_kmax_r_stats}
\begin{tabular}{lrrrrrrr}
\addlinespace[0.22cm]
\toprule
\addlinespace[0.22cm]
Param & value & $(m^-, m^+)$ & avg.LHS & only+Ns & only-Ns & emptyNs & $\textit{uni}^{+}$ \\
\addlinespace[0.22cm]
\midrule
\addlinespace[0.22cm]
$k_{\max}$ & 1 & (26, 20) & 1.0 & 109 & 115 & 0 & 0 \\
 & 2 & (28, 20) & 2.0 & 70 & 74 & 0 & 0 \\
 & 3 & (28, 20) & 3.0 & 38 & 36 & 0 & 0 \\
 & 4 & (28, 21) & 4.0 & 26 & 22 & 0 & 0 \\
 & 5 & (28, 22) & 5.0 & 17 & 13 & 0 & 0 \\
 & 6 & (28, 22) & 6.0 & 12 & 8  & 0 & 0 \\
 & 7 & (28, 22) & 7.0 & 3  & 6  & 0 & 0 \\
 & 8 & (28, 22) & 8.0 & 3  & 2  & 0 & 0 \\
 & 9 & (28, 22) & 9.0 & 1  & 1  & 0 & 0 \\
 & 10 & (28, 22) & 10.0 & 1 & 0 & 0 & 0 \\
\addlinespace[0.22cm]
\midrule
\addlinespace[0.22cm]
$\ell$ & 4.0 & (1, 5)   & 0.04  & 4   & 0   & 219 & 2 \\
 & 4.5 & (9, 13)  & 0.21  & 18  & 6   & 196 & 0 \\
 & 5.0 & (14, 18) & 0.67  & 32  & 8   & 169 & 0 \\
 & 5.5 & (19, 20) & 1.50  & 30  & 28  & 129 & 0 \\
 & 6.0 & (24, 20) & 3.43  & 26  & 40  & 77  & 0 \\
 & 6.5 & (27, 21) & 7.12  & 15  & 29  & 41  & 0 \\
 & 7.0 & (28, 21) & 12.69 & 6   & 22  & 20  & 0 \\
 & 7.5 & (28, 22) & 19.76 & 4   & 16  & 7   & 0 \\
 & 8.0 & (28, 22) & 27.27 & 1   & 7   & 3   & 0 \\
 & 8.5 & (28, 22) & 34.06 & 0   & 3   & 1   & 0 \\
 & 9.0 & (28, 22) & 39.95 & 0   & 2   & 0   & 0 \\
 & 9.5 & (28, 22) & 44.29 & 0   & 0   & 0   & 0 \\
 & 10.0 & (28, 22) & 47.06 & 0  & 0   & 0   & 1 \\
\bottomrule
\end{tabular}
\end{table}

\begin{table}[t!]
\centering
\footnotesize
\setlength{\tabcolsep}{3pt}
\renewcommand{\arraystretch}{0.75}
\setlength{\aboverulesep}{0pt}
\setlength{\belowrulesep}{0pt}
\setlength{\textfloatsep}{6pt}
\caption{Statistics of bipartite graphs generated from the \textbf{Productivity} ($\rho=11$) dataset. For all graphs, $n=112$.}
\label{tab:productivity_kmax_r_stats}
\begin{tabular}{lrrrrrrr}
\addlinespace[0.2cm]
\toprule
\addlinespace[0.2cm]
Param & value & $(m^-, m^+)$ & avg.LHS & only+Ns & only-Ns & emptyNs & $\textit{uni}^{+}$ \\
\addlinespace[0.2cm]
\midrule
\addlinespace[0.2cm]
$k_{\max}$ & 1  & (8, 24)  & 1.0  & 77 & 35 & 0 & 0 \\
 & 2  & (11, 33) & 2.0  & 62 & 11 & 0 & 0 \\
 & 3  & (11, 34) & 3.0  & 52 & 0  & 0 & 0 \\
 & 4  & (11, 35) & 4.0  & 28 & 0  & 0 & 0 \\
 & 5  & (11, 37) & 5.0  & 16 & 0  & 0 & 0 \\
 & 6  & (11, 37) & 6.0  & 9  & 0  & 0 & 0 \\
 & 7  & (11, 37) & 7.0  & 4  & 0  & 0 & 0 \\
 & 8  & (11, 37) & 8.0  & 3  & 0  & 0 & 0 \\
 & 9  & (11, 37) & 9.0  & 3  & 0  & 0 & 0 \\
 & 10 & (11, 37) & 10.0 & 2  & 0  & 0 & 0 \\
\addlinespace[0.2cm]
\midrule
\addlinespace[0.2cm]
$\ell$ & 4.0  & (11, 39) & 20.33 & 0 & 1 & 7 & 0 \\
 & 4.5  & (11, 39) & 25.60 & 0 & 0 & 6 & 0 \\
 & 5.0  & (11, 39) & 32.12 & 0 & 0 & 6 & 0 \\
 & 5.5  & (11, 39) & 38.80 & 0 & 0 & 6 & 0 \\
 & 6.0  & (11, 39) & 44.24 & 0 & 0 & 5 & 1 \\
 & 6.5  & (11, 39) & 46.70 & 0 & 0 & 5 & 13 \\
 & 7.0  & (11, 39) & 47.49 & 0 & 0 & 5 & 25 \\
 & 7.5  & (11, 39) & 47.73 & 0 & 0 & 5 & 35 \\
 & 8.0  & (11, 39) & 47.77 & 0 & 0 & 5 & 39 \\
 & 8.5  & (11, 39) & 47.84 & 0 & 0 & 4 & 5 \\
 & 9.0  & (11, 39) & 48.05 & 0 & 0 & 4 & 25 \\
 & 9.5  & (11, 39) & 48.26 & 0 & 0 & 2 & 0 \\
 & 10.0 & (11, 39) & 48.71 & 0 & 0 & 2 & 16 \\
\bottomrule
\end{tabular}
\end{table}

\paragraph{The Algorithms and Parameters Used.}
\label{subsec:revealrm_app-algosexp}
Specific to each of the settings, below is information on the algorithms and parameters used.

\begin{enumerate}[label=\arabic*)]
    \item \textbf{The standard model.} For each bipartite graph and every budget \(K \in \{1,5\}\), we computed the social welfare returned by classic greedy \(F(S_{\mathrm{g}})\), the heuristic greedy \(F(S_{\mathrm{hg}})\), random \(F(S_{\mathrm{r}})\) (i.e., \(K\) targets are chosen uniformly at random), and random heuristic \(F(S_{\mathrm{hr}})\)  (i.e., modifies lines 4 and 5 of Algorithm~\ref{alg:heuristic_greedy_lbreveal} to use random algorithm instead).
    For reference, we also compute social welfare with no budget constraints \(F(S_{\mathrm{full}})\), zero budget \(F(S_{\mathrm{o}})\), and the welfare returned by bruteforce search (or \(K\)-step lookahead greedy algorithm) \(F(S^{\star})\).
    
    Note that for experiments under the standard model, Algorithm~\ref{alg:greedy_lbreveal} is executed without restriction on information disclosure.

    \item \textbf{The target interventions model.}
    For each bipartite graph, and for every target reveal budget \(K \in \{1, 5\}\) and targeted intervention budget 
    \(B \in \{1, 3\}\), we compare the pre-reveal and post-reveal (Algorithms~\ref{alg:TI_greedy_label_reveal} and \ref{alg:greedy_TI_label_reveal}) intervention gains (Eqns.~\ref{eq:prereval-gains} and \ref{eq:postreval-gains}).
   
    \item \textbf{The coverage radius model.}
    We computed the number of agents covered using Algorithm~\ref{alg:radius_greedy_coverage}, for each of the $4$ graphs and radius budget \(R \in \{4, 4.5, 5, 5.5, 6, 6.5, 7, 7.5, 8, 8.5,\) \(9, 9.5, 10\}\).
    
    \item \textbf{The learning setting.} 
    The learning algorithm proceeds in two main stages. First, we apply the budget ($K$) constrained greedy algorithm (Algorithm~\ref{alg:greedy_lbreveal}) to the training set graph, where the revealed target set is the learned hypothesis. Next, we assess the performance of this hypothesis on the testing set graph.
\end{enumerate}

\paragraph{Performance and Evaluation.}  
\label{subsec:revealrm_app-perfeval}
Below are the details of the performance evaluation for the different algorithms across a range of model settings.

\begin{enumerate}[label=\arabic*)]

    \item \textbf{Under the standard, targeted intervention, and coverage models.}
    In both the standard and targeted intervention models, we compare algorithms by the social welfare they produce. The coverage model, in contrast, measures the number of agents that fall within reach after expanding the coverage radius of a selected set of positive targets.
    
    In the standard model, each algorithm is evaluated by the social welfare \(F(S)\) it achieves under different target reveal budgets \(K\).
    In the targeted intervention model, we analyze the difference in pre- and post-reveal intervention gains at different target  reveal budgets \(K\) and intervention budgets \(B\). Below is a definition of the intervention gains.

    \item \textbf{Pre-reveal intervention gains:} These are computed as the difference in social welfare returned by the Algorithm~\ref{alg:TI_greedy_label_reveal} ($F(S_{\mathrm{ig}}$) and that from Algorithm~\ref{alg:greedy_lbreveal} 
    ($ F(S_{\mathrm{g}}$):
    \begin{equation}
    \label{eq:prereval-gains}
        \Delta_F(\mathrm{ig}, \mathrm{g}) = F(S_{\mathrm{ig}}) - F(S_{\mathrm{g}}).
    \end{equation}
    \paragraph{Post-reveal intervention gains:} These are computed as the difference in social welfare returned by the Algorithm \ref{alg:greedy_TI_label_reveal} ($F(S_{\mathrm{gi}}$) and that from Algorithm~\ref{alg:greedy_lbreveal} 
    ($ F(S_{\mathrm{g}}$):
    \begin{equation}
    \label{eq:postreval-gains}
        \Delta_F(\mathrm{gi}, \mathrm{g}) = F(S_{\mathrm{gi}}) - F(S_{\mathrm{g}}).
    \end{equation}

    \item \textbf{Under the learning setting.}
    Let $\mathcal{X}_{tr}$ and $\mathcal{X}_{ts}$ denote the training and testing agent sets. Due to notation simplicity, all performance metrics introduced below are defined over the training set $\mathcal{X}_{tr}$, but the same apply to the testing set $\mathcal{X}_{ts}$.
    For each agent $x \in \Xs_{tr}$, let $N(x) \subseteq \Ts$ denote the set of targets in its neighborhood and define 
    $\delta_x^{+} = | \{t \in N(x) : f(t)=+1 \}|$ and  $\delta_x^{-} = |\{t \in N(x) : f(t)=-1\}|$ as the number of positive and negative targets in that neighborhood, respectively.
    To evaluate performance, we consider three kinds of agent subsets.
    \[
    \mathcal{X}_{tr}^{(1)}
    = \{x \in \mathcal{X}_{tr} : \delta_x^{+} > 0 \},
    \quad
    \mathcal{X}_{tr}^{(2)}
    = \mathcal{X}_{tr},
    \quad
    \mathcal{X}_{tr}^{(3)}
    = \{x \in \mathcal{X}_{tr} : \delta_x^{+} > 0 \ \text{and} \ \delta_x^{-} > 0 \}.
    \]
    The first agent set $\mathcal{X}_{tr}^{(1)}$ consists of agents with at least one positive target in their neighborhood, the second  $\mathcal{X}_{tr}^{(2)}$ includes all the agents $\mathcal{X}_{tr}$, and the third  $\mathcal{X}_{tr}^{(3)}$ includes only \textit{helpable agents} (those with both positive and negative target neighbors).
    Let $S_{\mathrm{tr}} \subseteq \Ts$ with $|S_{\mathrm{tr}}| \le K$ be the target set revealed by the classic greedy algorithm when run on the train graph $\graph_{tr} = (\mathcal{X}_{tr} \cup \Ts, E)$ at a budget of $K$, and let $F(S_{\mathrm{tr}})$ denote the resulting social welfare. The performance measures are defined as
    \[
    \mathrm{Perf}_{i}
    = \frac{F(S_{\mathrm{tr}})}{|\mathcal{X}_{tr}^{(i)}|} \times 100\%,
    \ i \in \{1,2\},
    \
    \mathrm{Perf}_{3}
    = \frac{F(S_{\mathrm{tr}}) - |\{x \in \mathcal{X}_{tr} : \delta_x^{+}\geq 1 \, \ \delta_x^{-}=0\}|}{|\mathcal{X}_{tr}^{(3)}|} \times 100\%.
    \]
    A score of $100$ with respect to $\mathrm{Perf}_{1}$ indicates success on agents with at least one positive target neighbor, excluding agents with empty or all-negative neighborhoods. For $\mathrm{Perf}_{2}$, a score of $100$ indicates success on all helpable agents, including all sampled agents, and a score of $100$ in $\mathrm{Perf}_{3}$ indicates success on all helpable agents, excluding unhelpable ones. 
    Theoretical results use $\mathrm{Perf}_{2}$, while empirical analysis considers them all.

\end{enumerate}

\subsection{Empirical Results under the Standard Model}
\label{sec:revealrm_app-smresults}

For all single-group results under the standard model,  \(F(S_{\mathrm{full}})\) represents the social welfare without any budget constraints, while \(F(S_{\mathrm{o}})\) corresponds to the social welfare when the budget is zero, and no targets are revealed. 
Under budget constraints, \(F(S_{\mathrm{r}})\) denotes the social welfare obtained by randomly revealing targets, \(F(S_{\mathrm{g}})\) corresponds that of Algorithm~\ref{alg:greedy_lbreveal}, \(F(S_{\mathrm{hr}})\) and \(F(S_{\mathrm{hg}})\) represent the social welfare achieved by the heuristic random and heuristic greedy strategies, respectively, and \(F(S^{\mathrm{*}})\) denotes the optimal social welfare computed via bruteforce search.

\begin{figure}[b!]
\vspace{0.66cm}
\captionsetup[subfigure]{justification=Centering}
\begin{subfigure}[t]{0.24\textwidth}
\centering
    \includegraphics[width=\textwidth]{reveal_rolemodels/revealrm_figures/aplot_Adult_sr_sg_knn.pdf}
    \caption{\(k\)NN graphs: $F(S_{\mathrm{g}})$ vs. $F(S_{\mathrm{r}})$}
    \label{fig:app_adult_sr_sg_knn}
\end{subfigure}
\begin{subfigure}[t]{0.24\textwidth}
\centering
    \includegraphics[width=\linewidth]{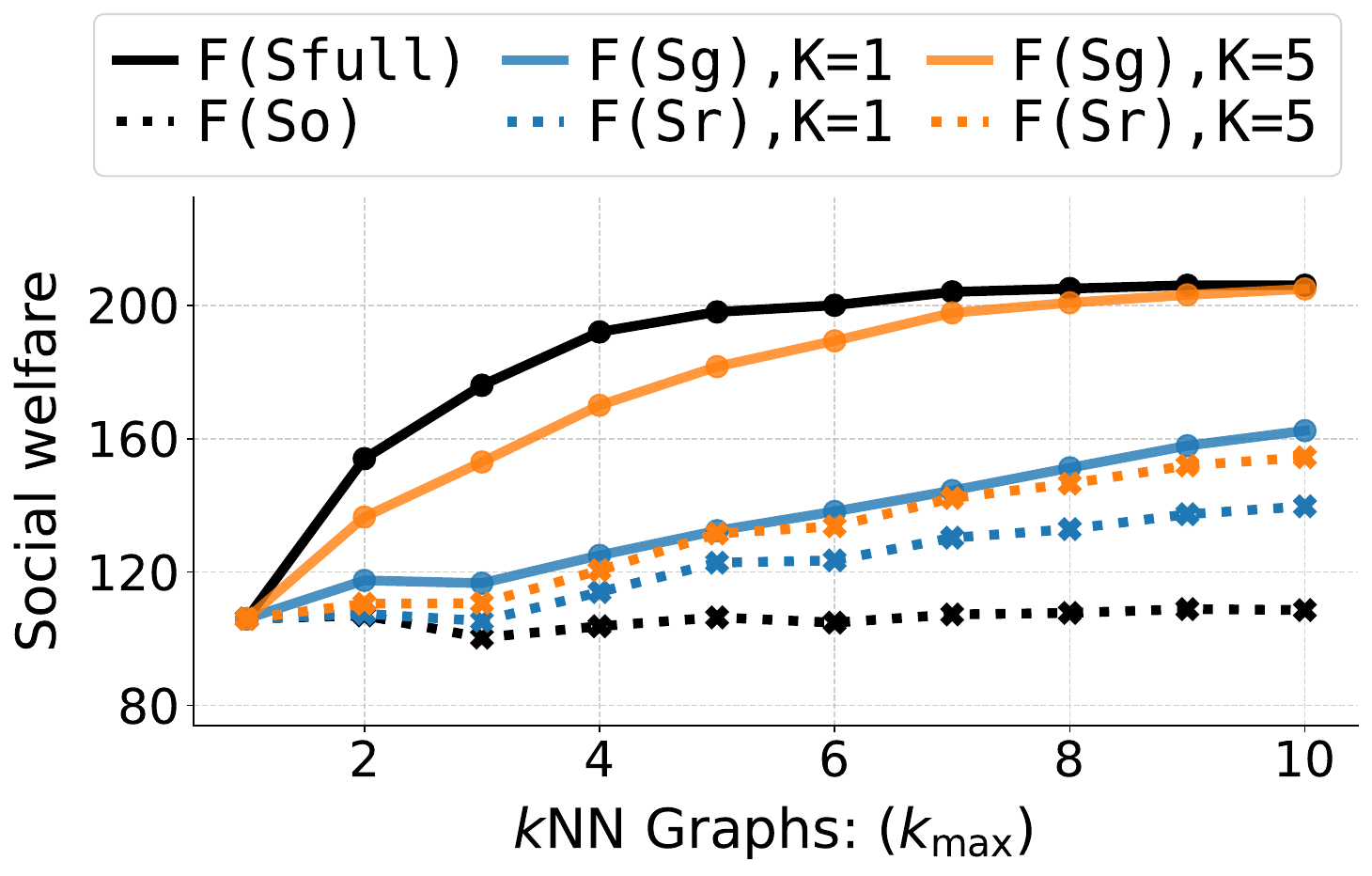}
    \caption{\(k\)NN graphs: $F(S_{\mathrm{g}})$ vs. $F(S_{\mathrm{r}})$}
    \label{fig:app_math_sr_sg_knn}    
\end{subfigure}
\begin{subfigure}[t]{0.24\textwidth}
\centering
    \includegraphics[width=\linewidth]{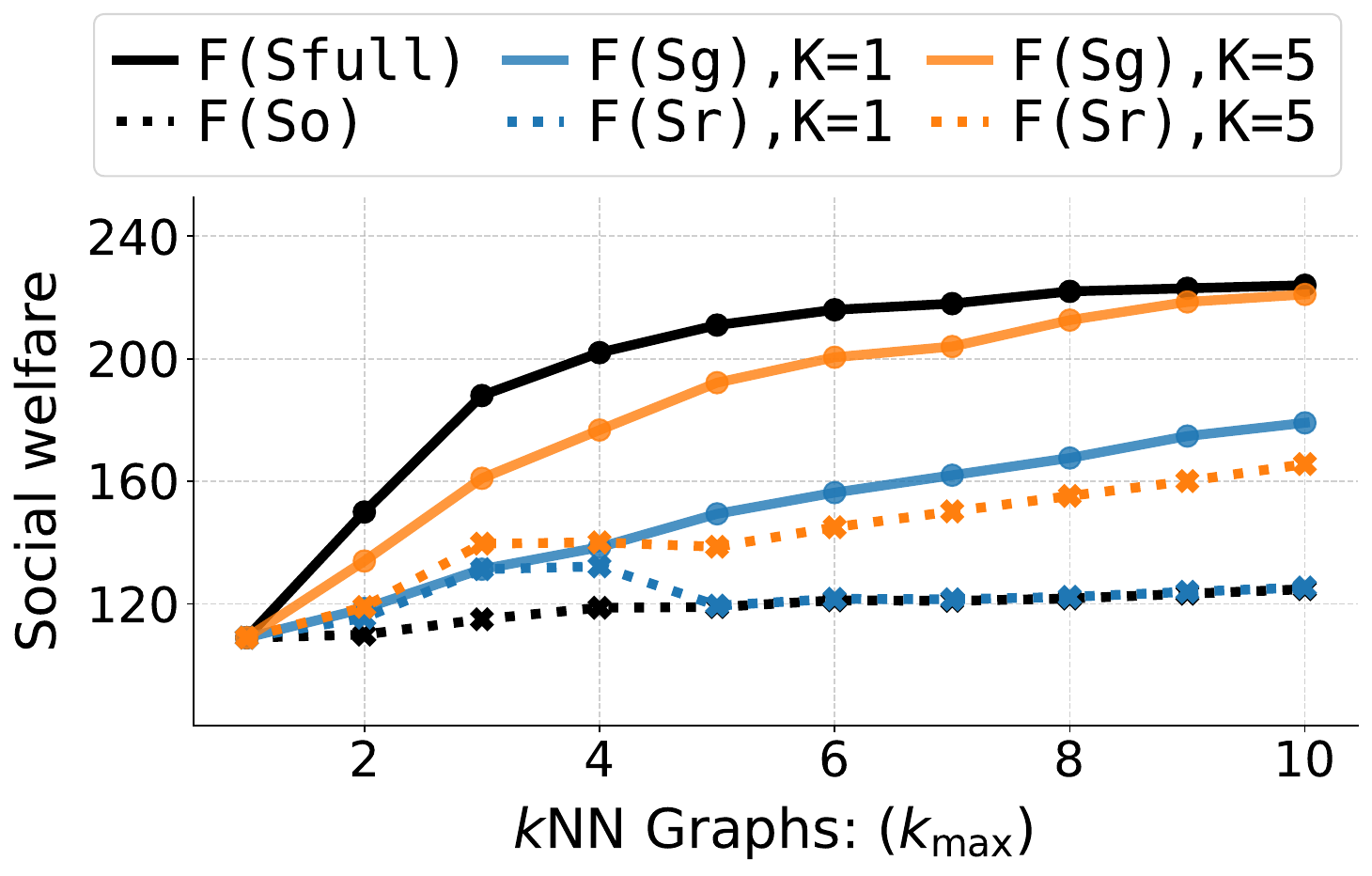}
    \caption{\(k\)NN graphs: $F(S_{\mathrm{g}})$ vs. $F(S_{\mathrm{r}})$}
    \label{fig:app_port_sr_sg_knn}
\end{subfigure}
\begin{subfigure}[t]{0.24\textwidth}
\centering
    \includegraphics[width=\linewidth]{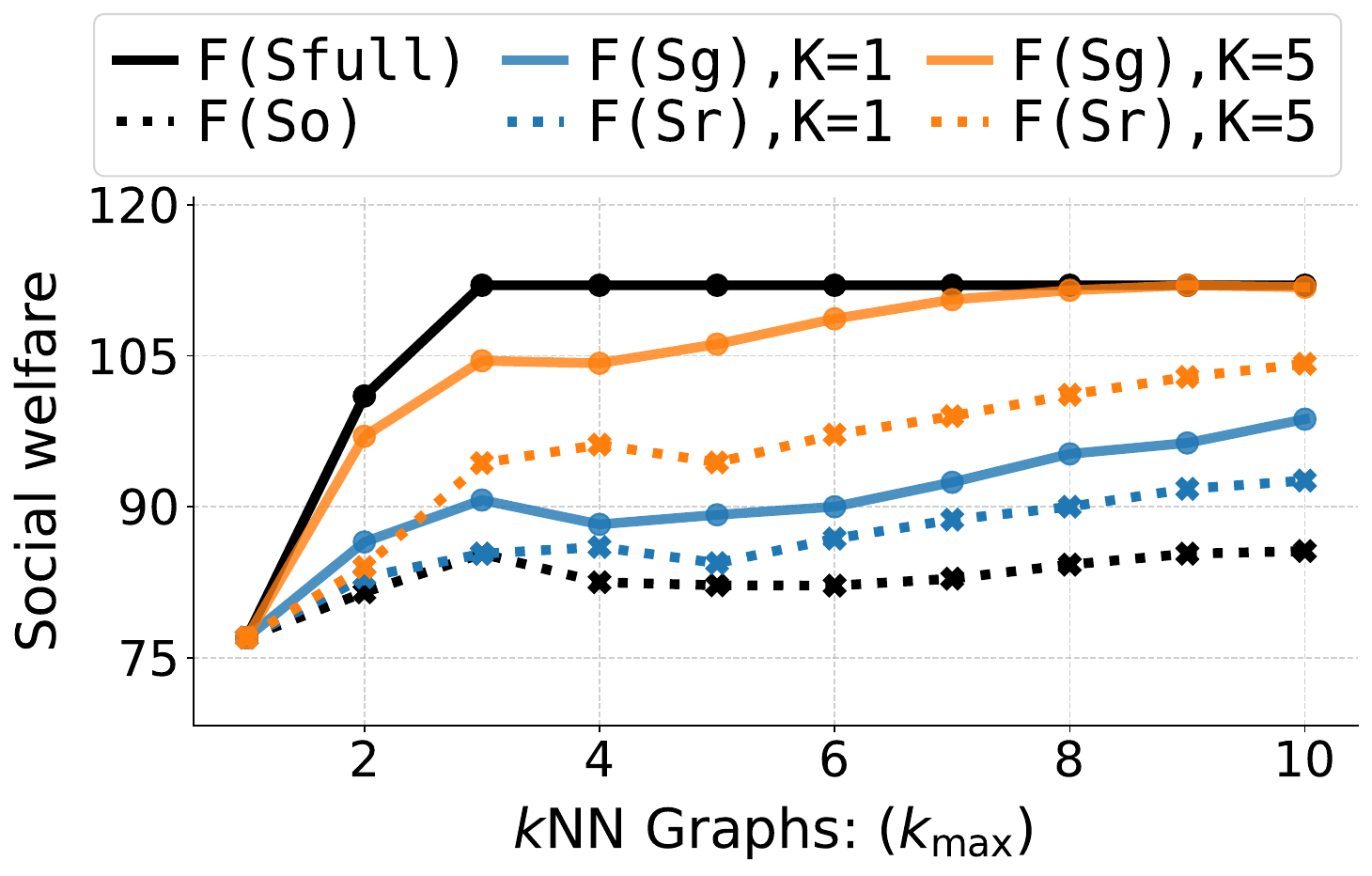}
    \caption{\(k\)NN graphs: $F(S_{\mathrm{g}})$ vs. $F(S_{\mathrm{r}})$}
    \label{fig:app_prod_sr_sg_knn}
\end{subfigure}\\[2ex]

\begin{subfigure}[t]{0.24\textwidth}
\centering
    \includegraphics[width=\textwidth]{reveal_rolemodels/revealrm_figures/aplot_Adult_sr_sg_thresh.pdf}
    \caption{Threshold graphs: $F(S_{\mathrm{g}})$ vs. $F(S_{\mathrm{r}})$}
    \label{fig:app_adult_sr_sg_thresh}
\end{subfigure}
\begin{subfigure}[t]{0.24\textwidth}
\centering
    \includegraphics[width=\linewidth]{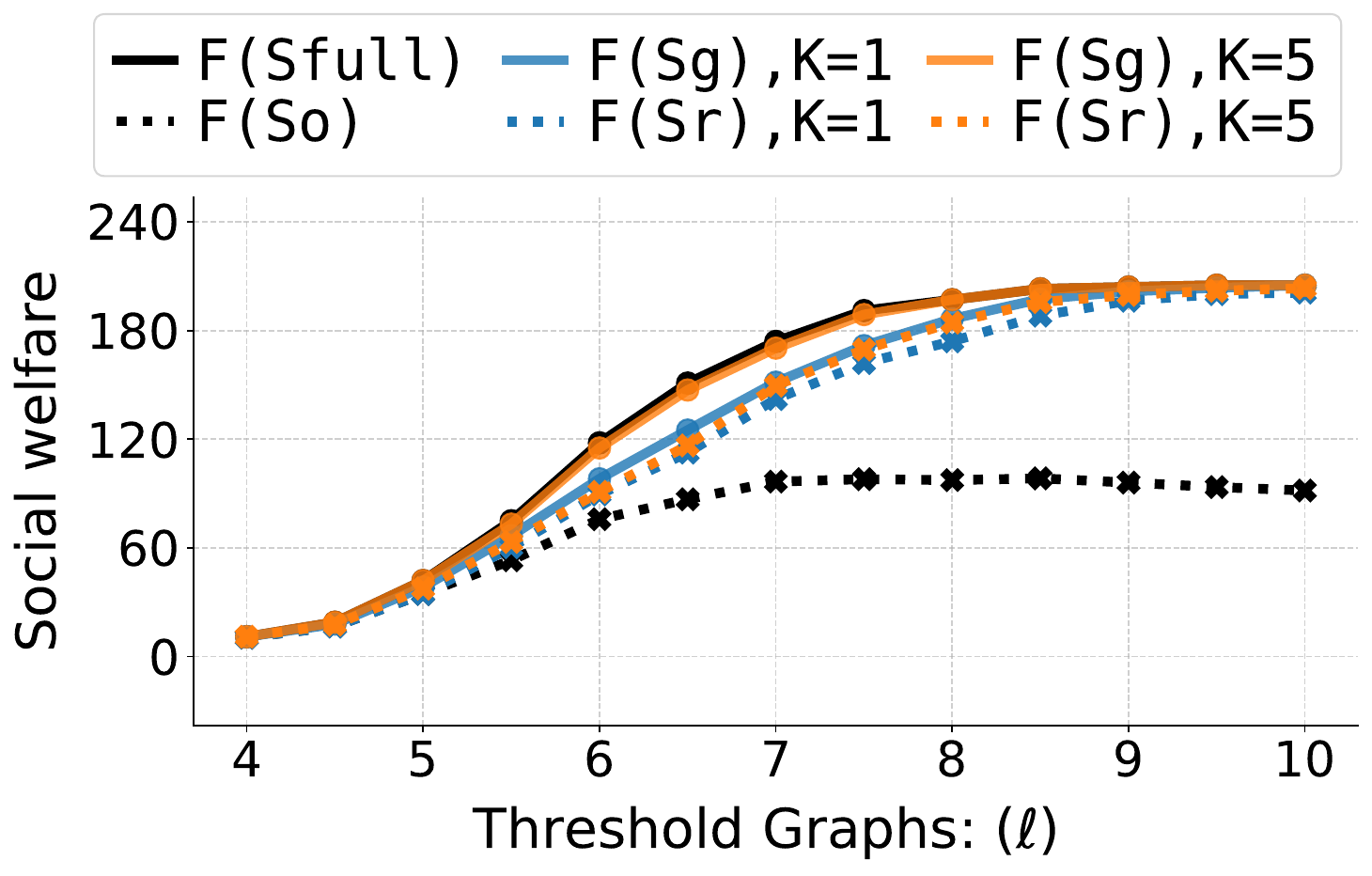}
    \caption{Threshold graphs: $F(S_{\mathrm{g}})$ vs. $F(S_{\mathrm{r}})$}
    \label{fig:app_math_sr_sg_thresh}
\end{subfigure}
\begin{subfigure}[t]{0.24\textwidth}
\centering
    \includegraphics[width=\linewidth]{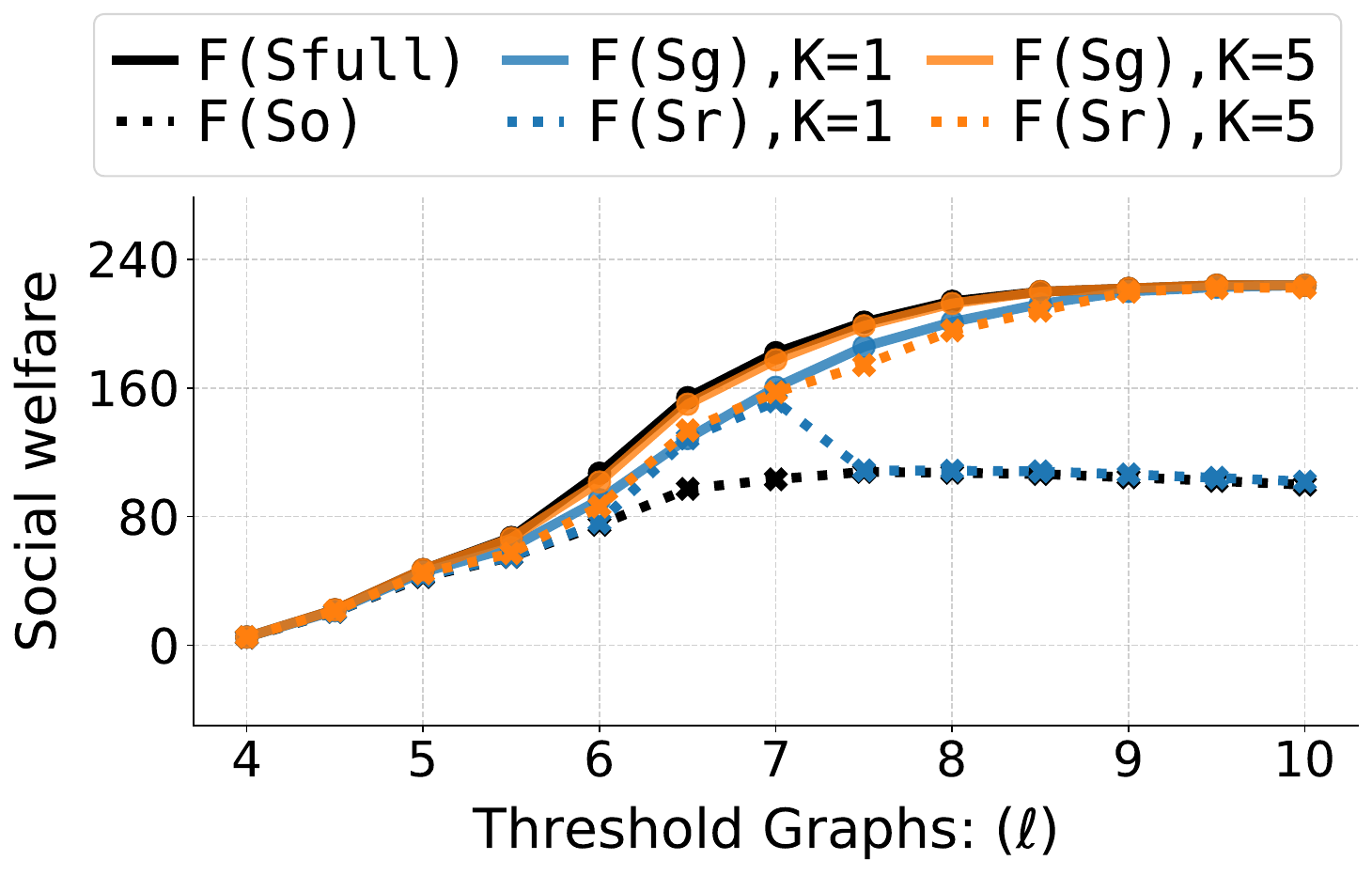}
    \caption{Threshold graphs: $F(S_{\mathrm{g}})$ vs. $F(S_{\mathrm{r}})$}
    \label{fig:app_port_sr_sg_thresh}
\end{subfigure}
\begin{subfigure}[t]{0.24\textwidth}
\centering
    \includegraphics[width=\linewidth]{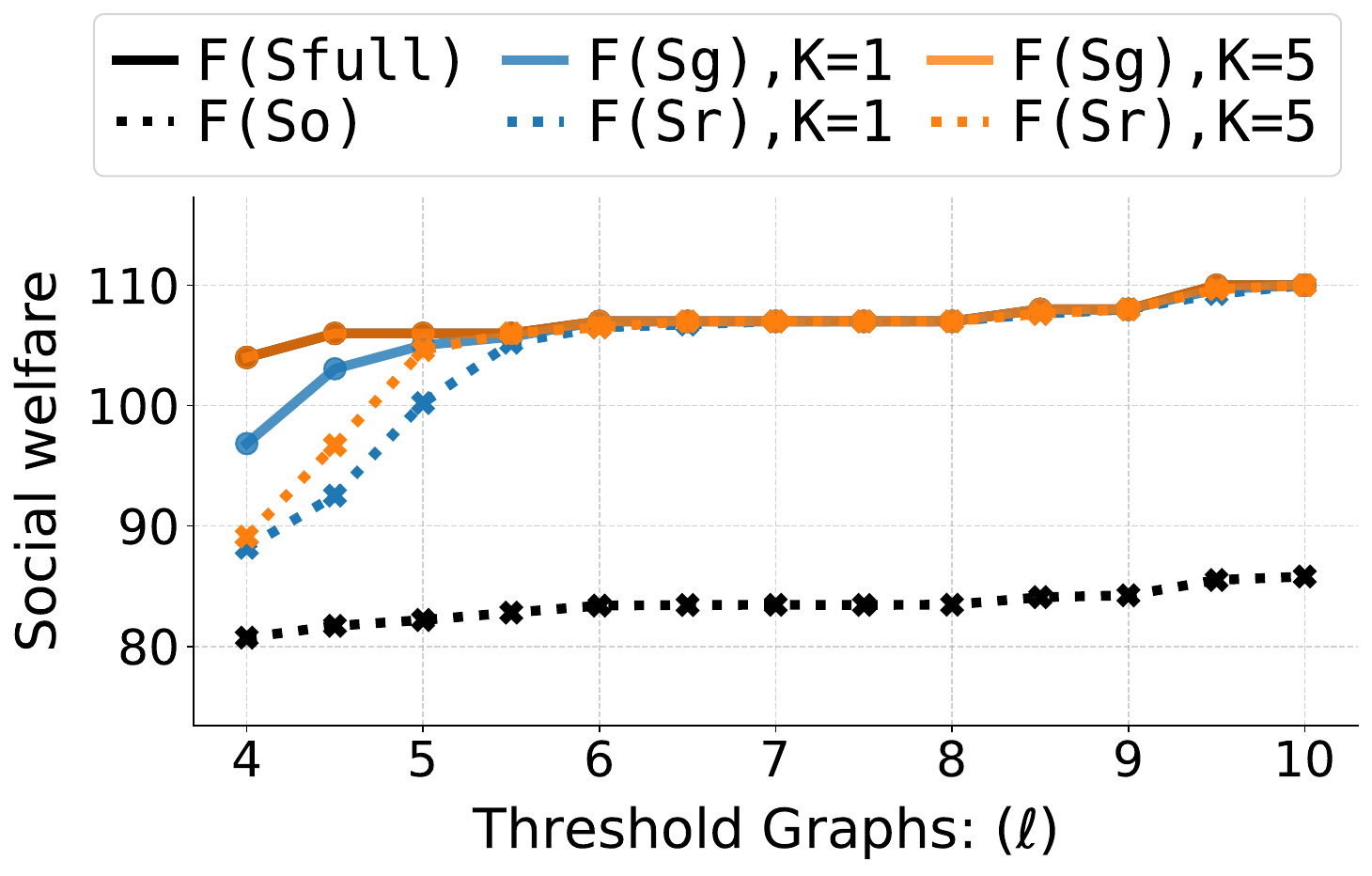}
    \caption{Threshold graphs: $F(S_{\mathrm{g}})$ vs. $F(S_{\mathrm{r}})$}
    \label{fig:app_prod_sr_sg_thresh}
\end{subfigure}
\caption[Comparison of \(F(S_{\mathrm{r}})\) with \(F(S_{\mathrm{g}})\)]{Comparative analysis of the social welfare generated from running the random \(F(S_{\mathrm{r}})\) and the classic greedy \(F(S_{\mathrm{g}})\) algorithms across \(k\)NN graphs (\subref{fig:app_adult_sr_sg_knn}--\subref{fig:app_prod_sr_sg_knn}) and the threshold graphs (\subref{fig:app_adult_sr_sg_thresh}--\subref{fig:app_prod_sr_sg_thresh}) from the \(4\) datasets (Tables~\ref{tab:adult_kmax_r_stats}--\ref{tab:productivity_kmax_r_stats}). 
Except on the Adult dataset generated graphs, when the graphs are nearly complete, budgets \(K=\{1,5\}\) have a low effect and social welfare returned by both algorithms is almost the same (\subref{fig:app_math_sr_sg_thresh}--\subref{fig:app_prod_sr_sg_thresh}). In contrast, in sparser settings, at similar budget levels, \(F(S_{\mathrm{g}})\) is consistently higher than \(F(S_{\mathrm{r}})\) (\subref{fig:app_adult_sr_sg_knn}--\subref{fig:app_prod_sr_sg_knn}).}
\label{fig:app_sr_sg}
\begin{picture}(0,0)
    \put(6,333){{\parbox{4cm}{\centering \textbf{Adult}}}}
    \put(120,333){{\parbox{4cm}{\centering \textbf{Math}}}}
    \put(230,333){{\parbox{4cm}{\centering \textbf{Portuguese}}}}
    \put(340,333){{\parbox{4cm}{\centering \textbf{Productivity}}}}
\end{picture}
\end{figure}

\begin{figure}[b!]
\vspace{0.66cm}
\captionsetup[subfigure]{justification=Centering}
\begin{subfigure}[t]{0.24\textwidth}
\centering
    \includegraphics[width=\textwidth]{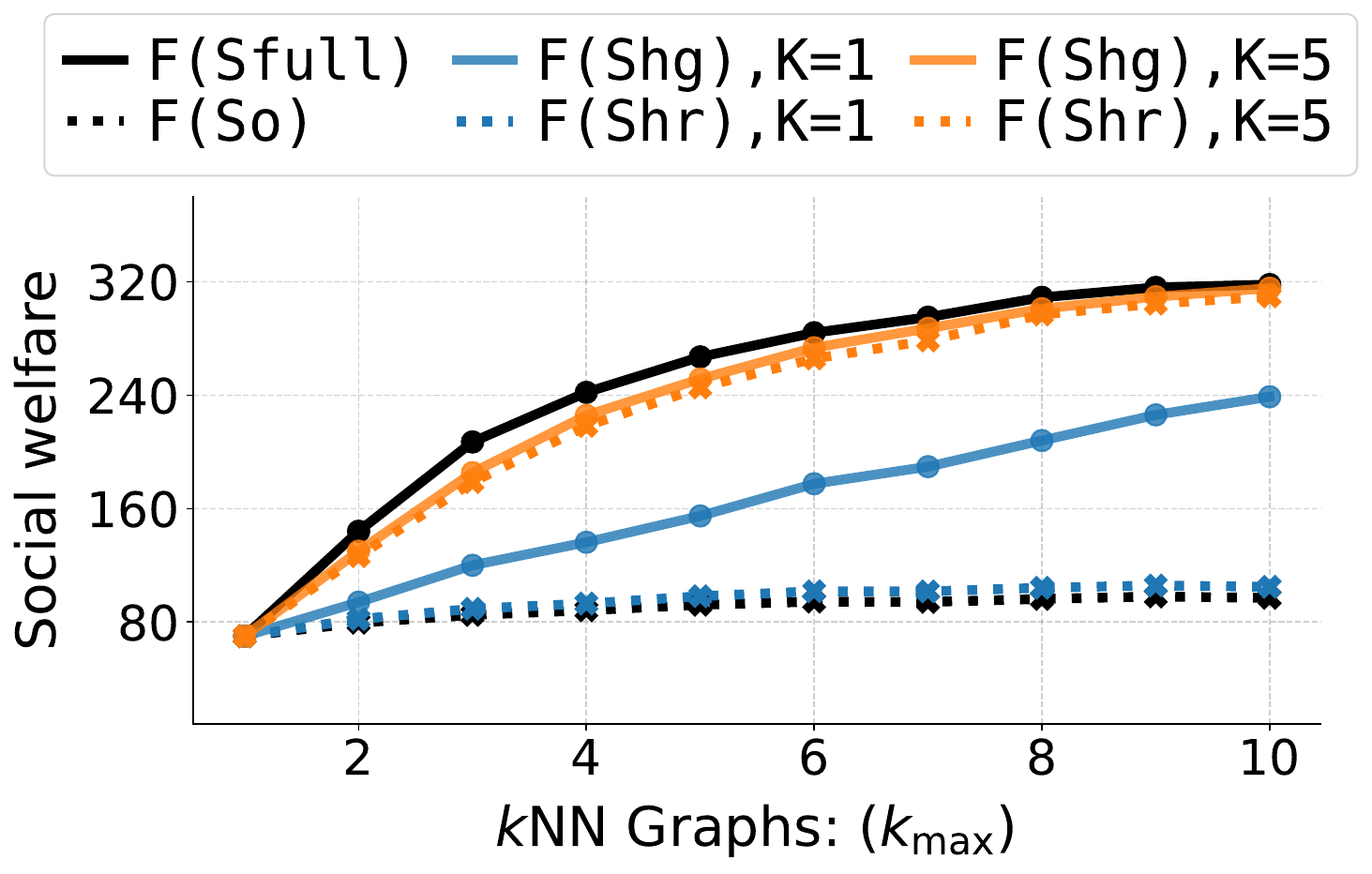}
    \caption{\(k\)NN graphs: \(F(S_{\mathrm{hg}})\) vs. \(F(S_{\mathrm{hr}})\)}
    \label{fig:app_adult_shg_shr_knn}
\end{subfigure}
\begin{subfigure}[t]{0.24\textwidth}
\centering
    \includegraphics[width=\linewidth]{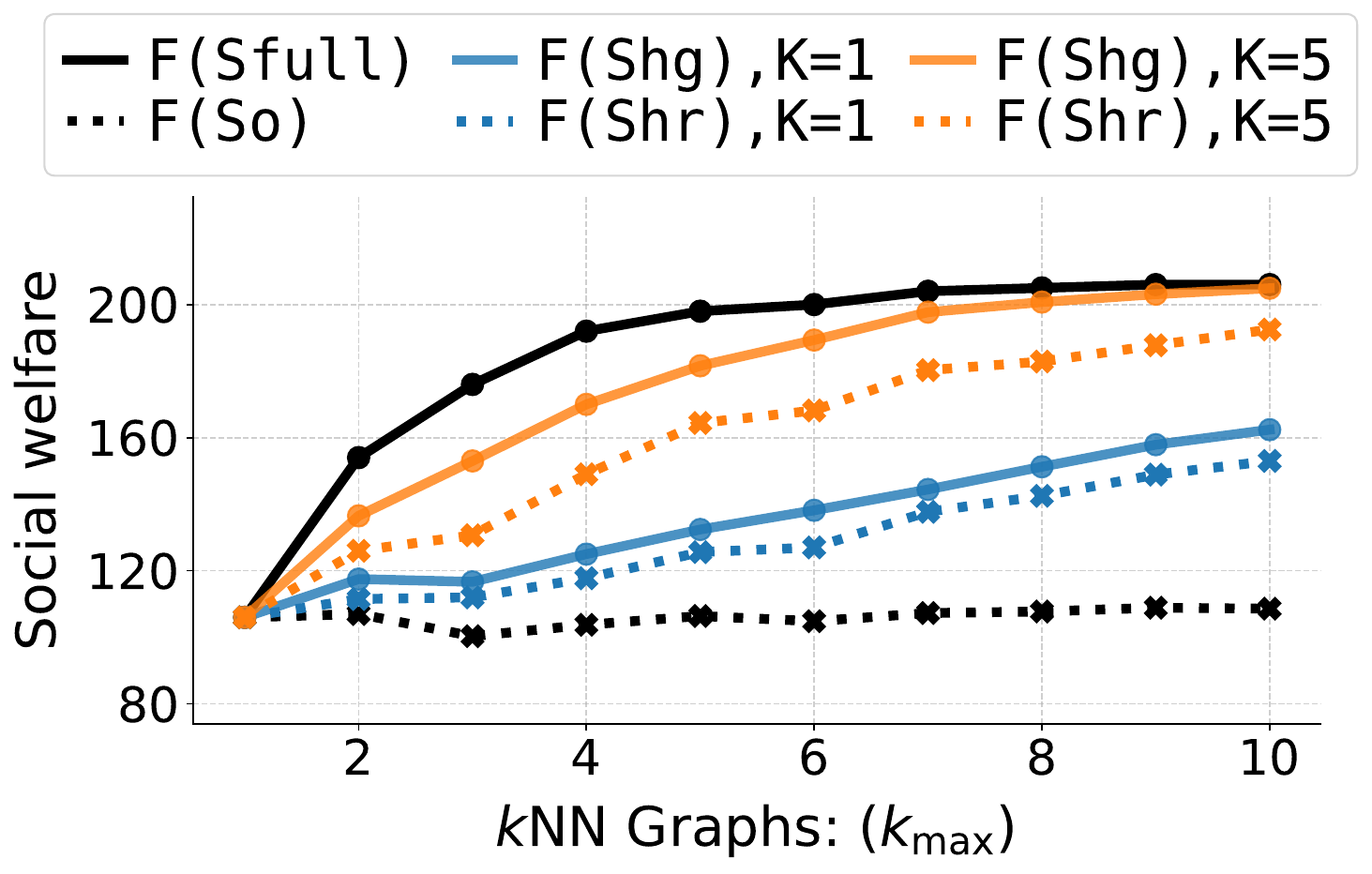}
    \caption{\(k\)NN graphs: \(F(S_{\mathrm{hg}})\) vs. \(F(S_{\mathrm{hr}})\)}
    \label{fig:app_math_shg_shr_knn}
\end{subfigure}
\begin{subfigure}[t]{0.24\textwidth}
\centering
    \includegraphics[width=\linewidth]{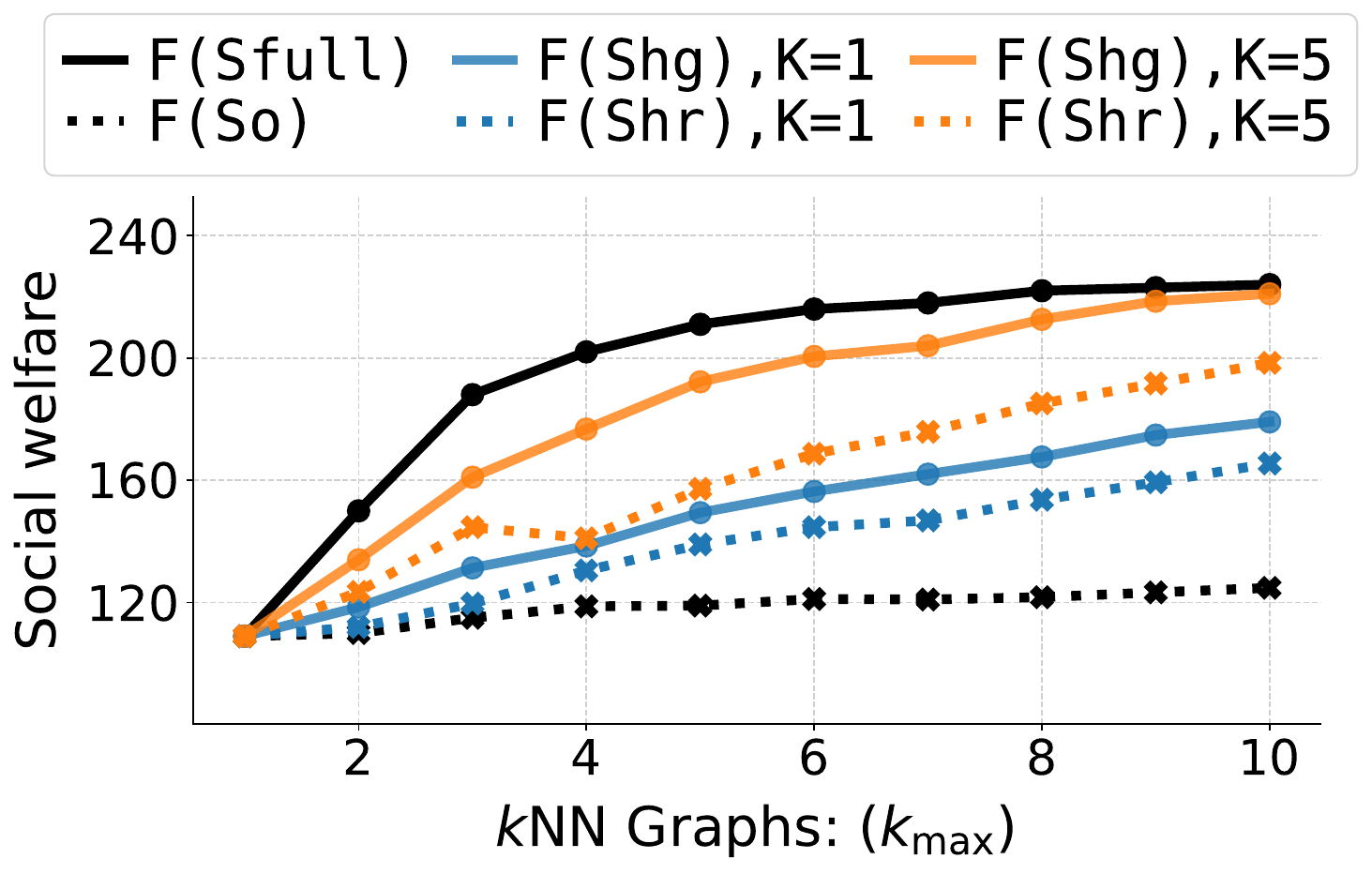}
    \caption{\(k\)NN graphs: \(F(S_{\mathrm{hg}})\) vs. \(F(S_{\mathrm{hr}})\)}
    \label{fig:app_port_shg_shr_knn}
\end{subfigure}
\begin{subfigure}[t]{0.24\textwidth}
\centering
    \includegraphics[width=\linewidth]{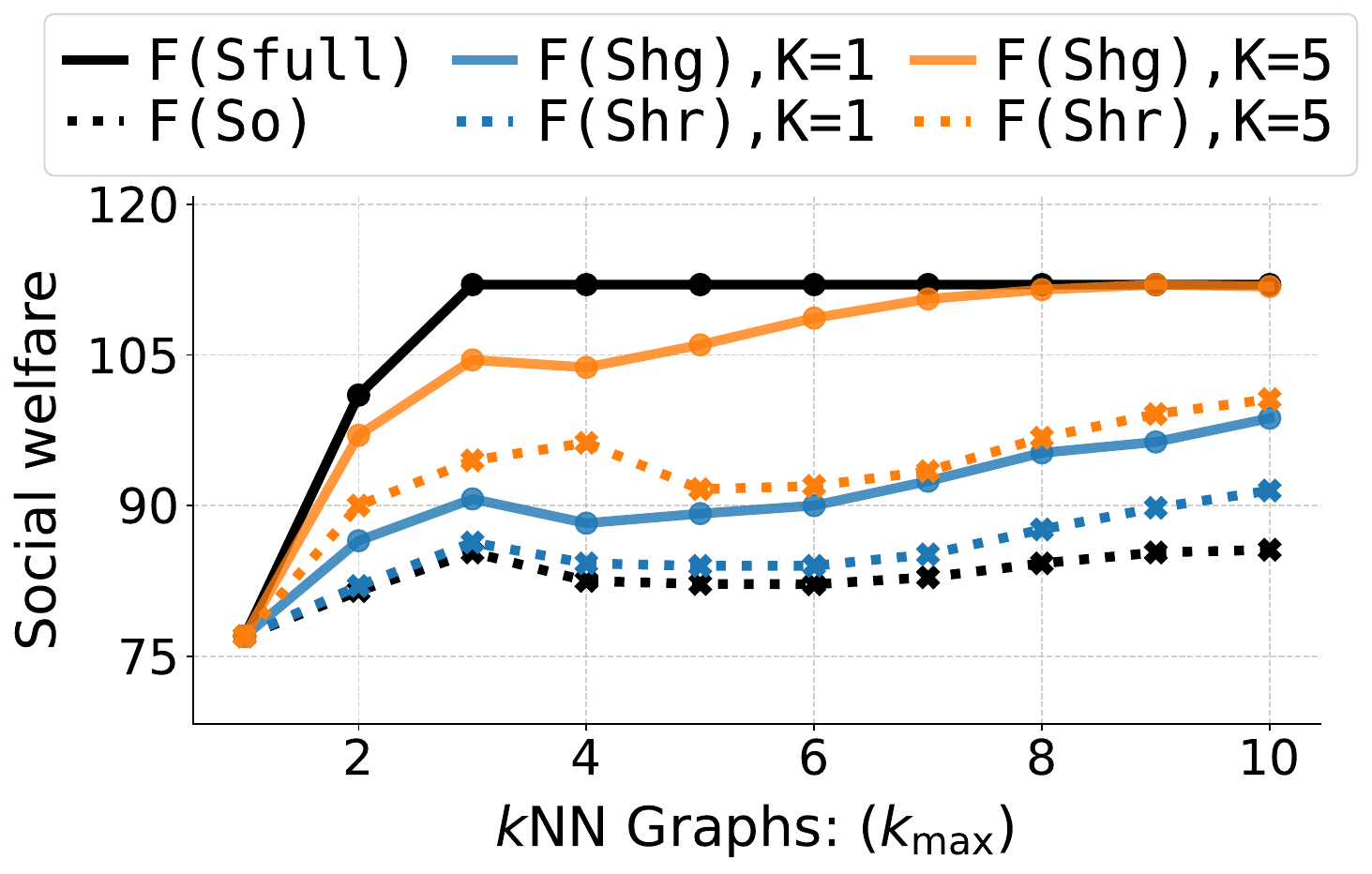}
    \caption{\(k\)NN graphs: \(F(S_{\mathrm{hg}})\) vs. \(F(S_{\mathrm{hr}})\)}
    \label{fig:app_prod_shg_shr_knn}
\end{subfigure}\\[2ex]

\begin{subfigure}[t]{0.24\textwidth}
\centering
    \includegraphics[width=\textwidth]{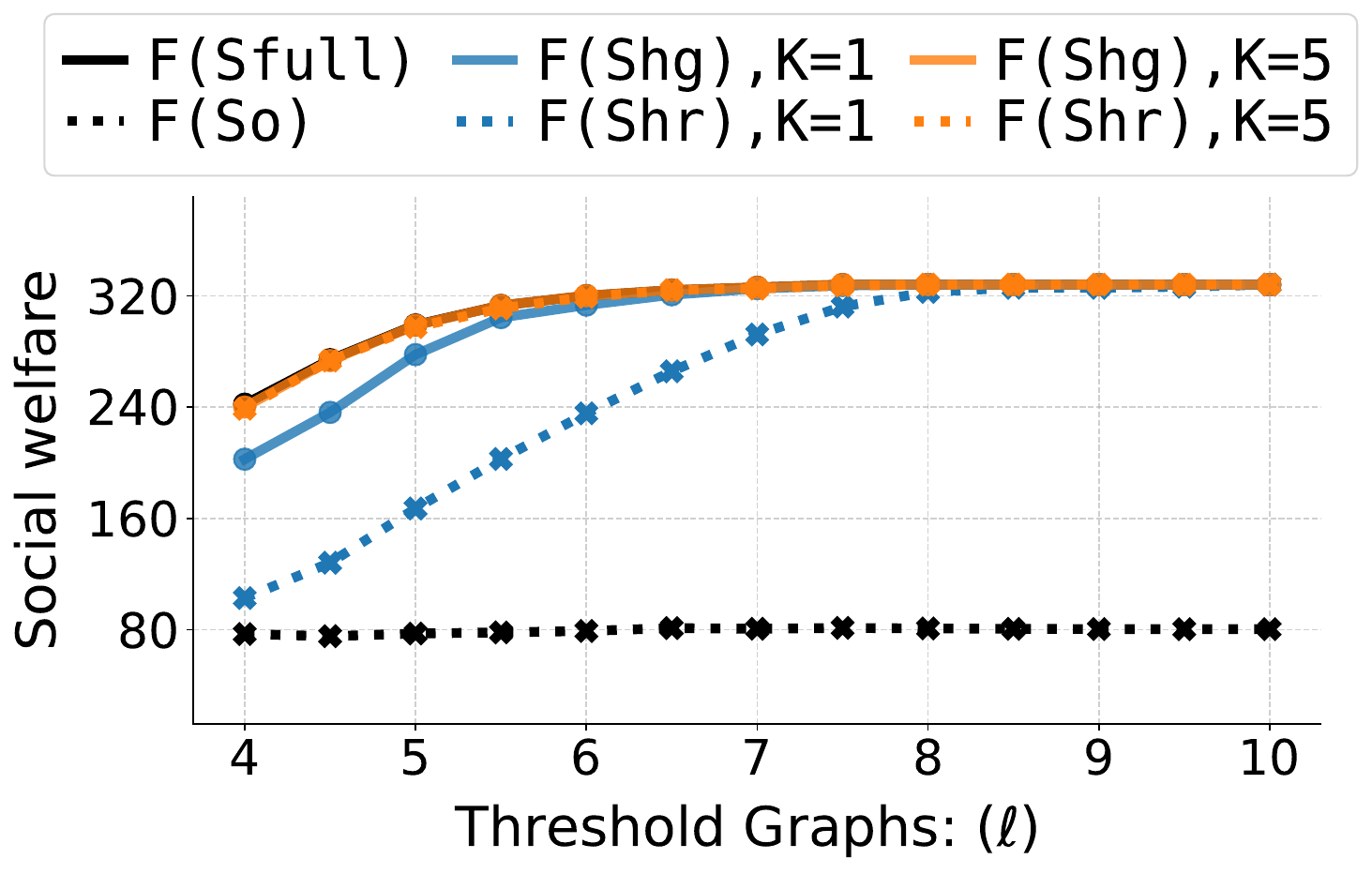}
    \caption{Threshold graphs: \(F(S_{\mathrm{hg}})\) vs. \(F(S_{\mathrm{hr}})\)}
    \label{fig:app_adult_shg_shr_thresh}
\end{subfigure}
\begin{subfigure}[t]{0.24\textwidth}
\centering
    \includegraphics[width=\linewidth]{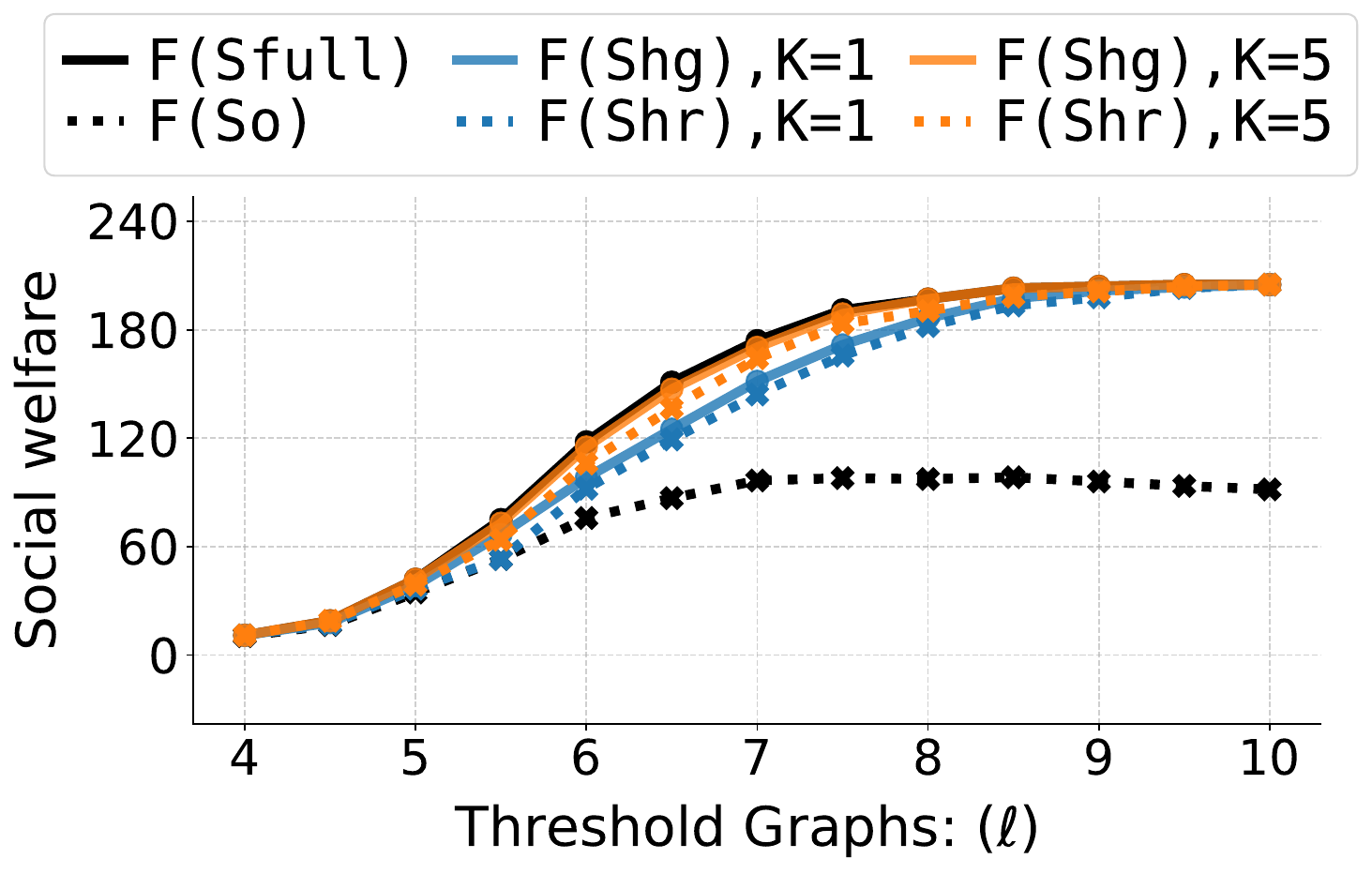}
    \caption{Threshold graphs: \(F(S_{\mathrm{hg}})\) vs. \(F(S_{\mathrm{hr}})\)}
    \label{fig:app_math_shg_shr_thresh}
\end{subfigure}
\begin{subfigure}[t]{0.24\textwidth}
\centering
    \includegraphics[width=\linewidth]{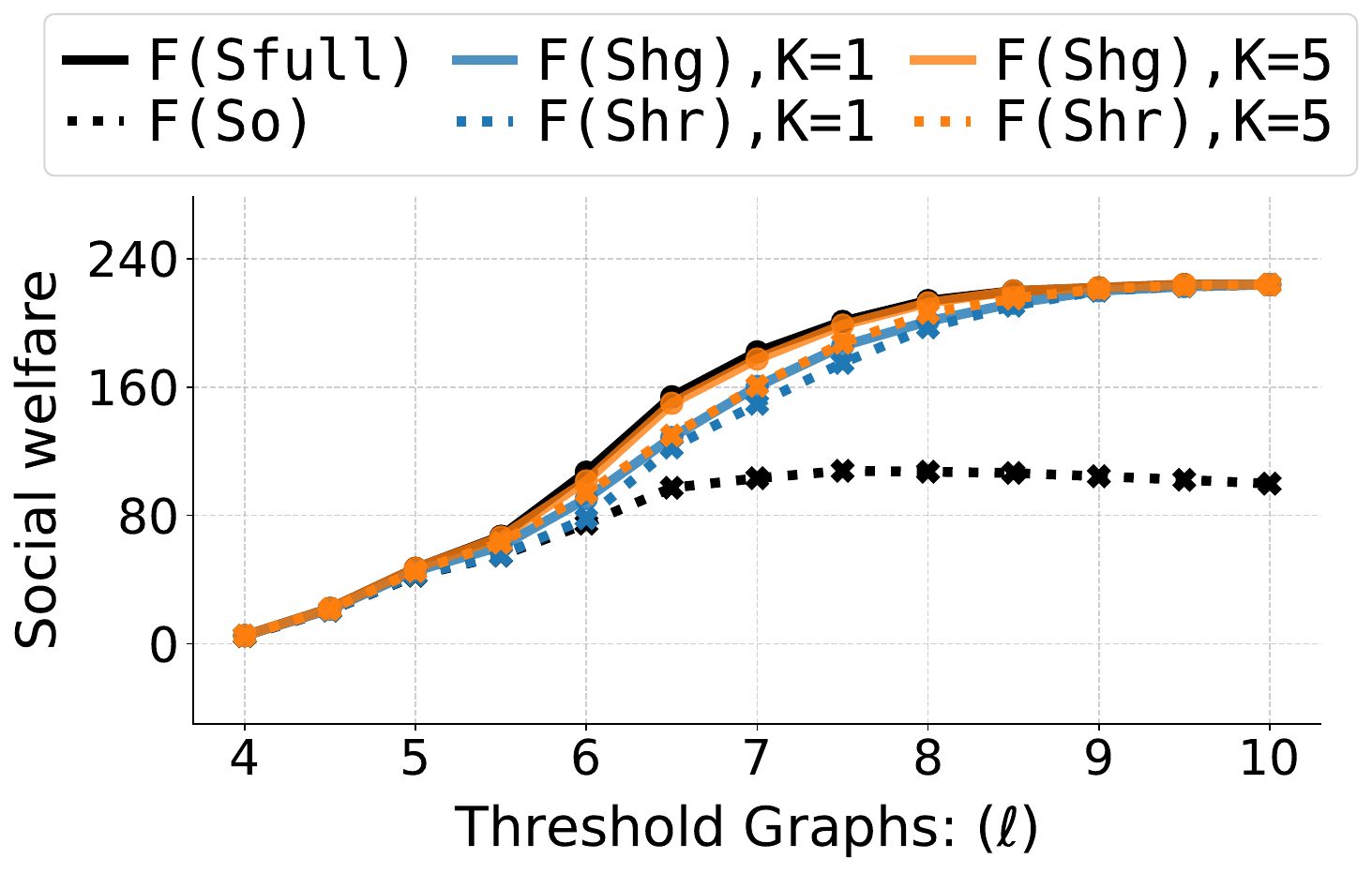}
    \caption{Threshold graphs: \(F(S_{\mathrm{hg}})\) vs. \(F(S_{\mathrm{hr}})\)}
    \label{fig:app_port_shg_shr_thresh}
\end{subfigure}
\begin{subfigure}[t]{0.24\textwidth}
\centering
    \includegraphics[width=\linewidth]{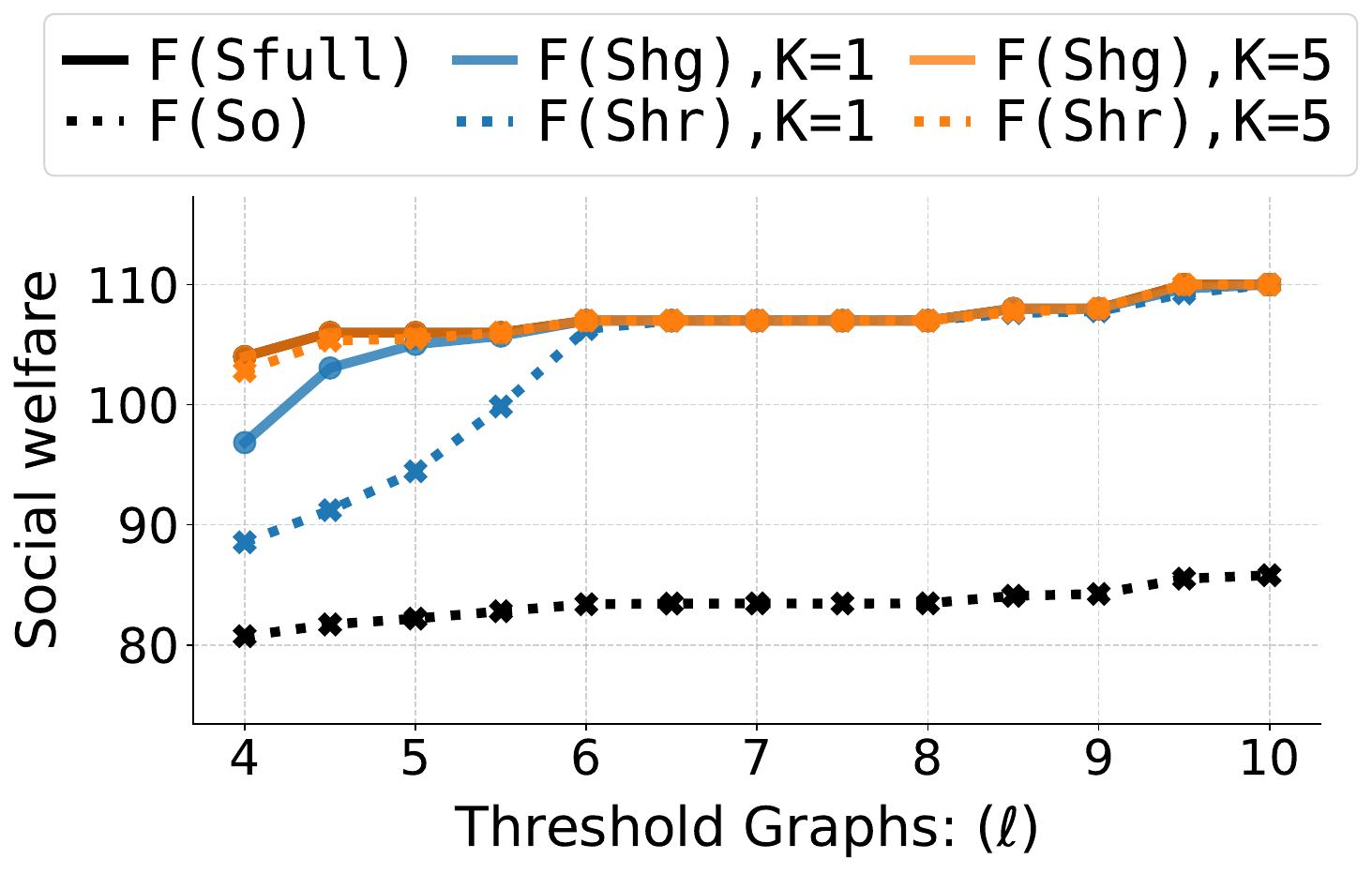}
    \caption{Threshold graphs: \(F(S_{\mathrm{hg}})\) vs. \(F(S_{\mathrm{hr}})\)}
    \label{fig:app_prod_shg_shr_thresh}
\end{subfigure}

\caption[Comparison of \(F(S_{\mathrm{hr}})\) with \(F(S_{\mathrm{hg}})\)]{Comparative analysis of the social welfare generated from running the heuristic random \(F(S_{\mathrm{hr}})\) and heuristic greedy \(F(S_{\mathrm{hg}})\) algorithms across the \(k\)NN graphs (\subref{fig:app_adult_shg_shr_knn}--\subref{fig:app_prod_shg_shr_knn}) and the threshold graphs (\subref{fig:app_adult_shg_shr_thresh}--\subref{fig:app_prod_shg_shr_thresh}) from the \(4\) datasets (Tables~\ref{tab:adult_kmax_r_stats}--\ref{tab:productivity_kmax_r_stats}). 
In general, when the graphs are nearly complete, across budget levels \(K=\{1,5\}\), social welfare returned by both heuristic algorithms is almost equal (\subref{fig:app_adult_shg_shr_thresh}--\subref{fig:app_prod_shg_shr_thresh}). In sparser settings, at similar budget levels, \(F(S_{\mathrm{hg}})\) is consistently higher than \(F(S_{\mathrm{hr}})\) (\subref{fig:app_adult_shg_shr_knn}--\subref{fig:app_prod_shg_shr_knn}).}

\label{fig:app_shg_shr}
\begin{picture}(0,0)
    \put(3,333){{\parbox{4cm}{\centering \textbf{Adult}}}}
    \put(117,333){{\parbox{4cm}{\centering \textbf{Math}}}}
    \put(230,333){{\parbox{4cm}{\centering \textbf{Portuguese}}}}
    \put(340,333){{\parbox{4cm}{\centering \textbf{Productivity}}}}
\end{picture}
\end{figure}

\begin{figure}[b!]
\centering
\captionsetup[subfigure]{justification=centering}
\begin{minipage}{0.48\textwidth}
\centering
\textbf{Adult Dataset}
\end{minipage}
\hfill
\begin{minipage}{0.48\textwidth}
\centering
\textbf{Productivity Dataset}
\end{minipage}
\vspace{0.55em}
\textit{kNN Graphs}\par\medskip
\begin{subfigure}[t]{0.24\textwidth}
    \centering
    \includegraphics[width=\linewidth]{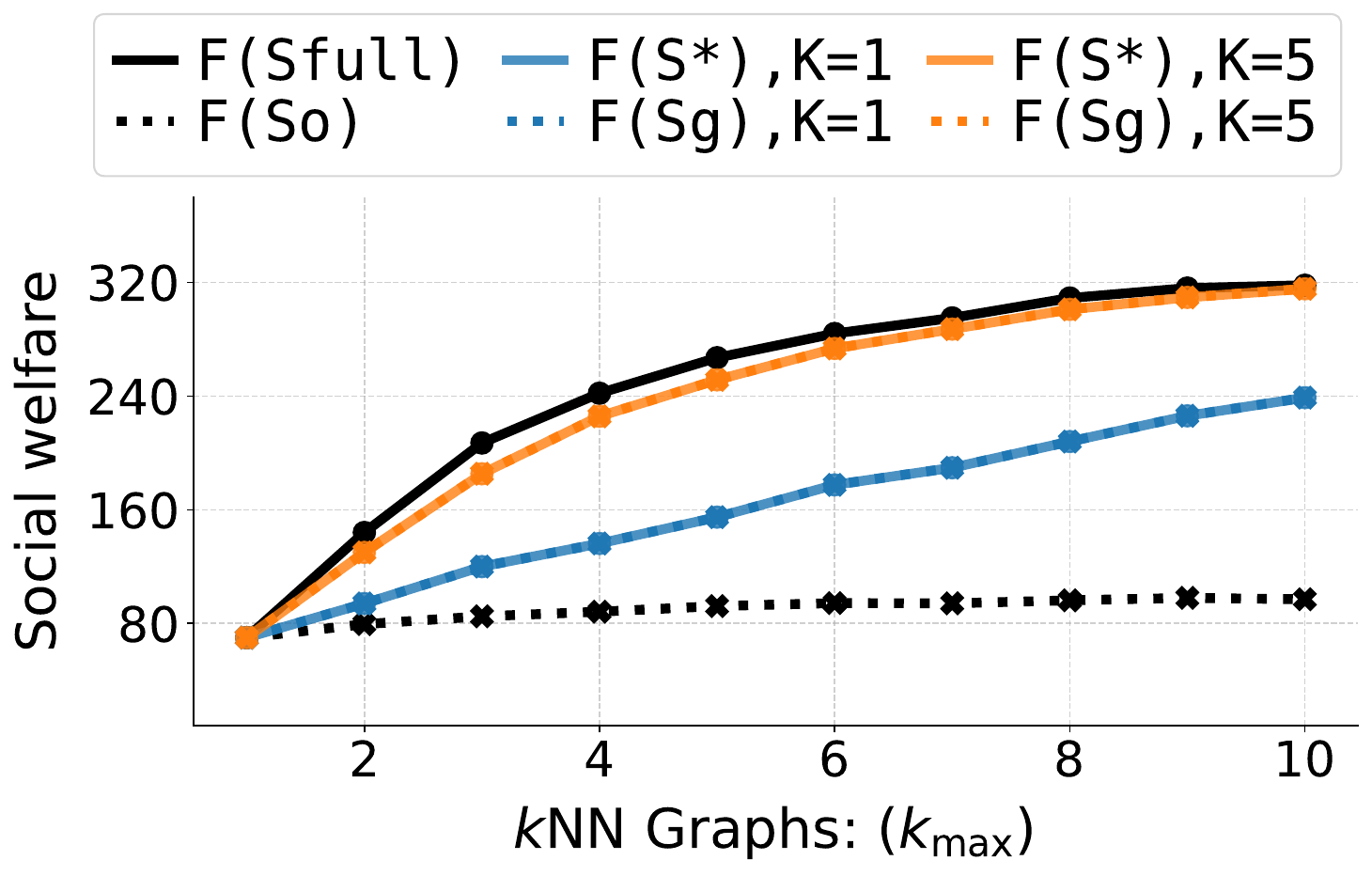}
    \caption{$F(S^{\star})$ vs.\ $F(S_{\mathrm{g}})$}
    \label{fig:app_adult_star_sg_knn}
\end{subfigure}
\begin{subfigure}[t]{0.24\textwidth}
    \centering
    \includegraphics[width=\linewidth]{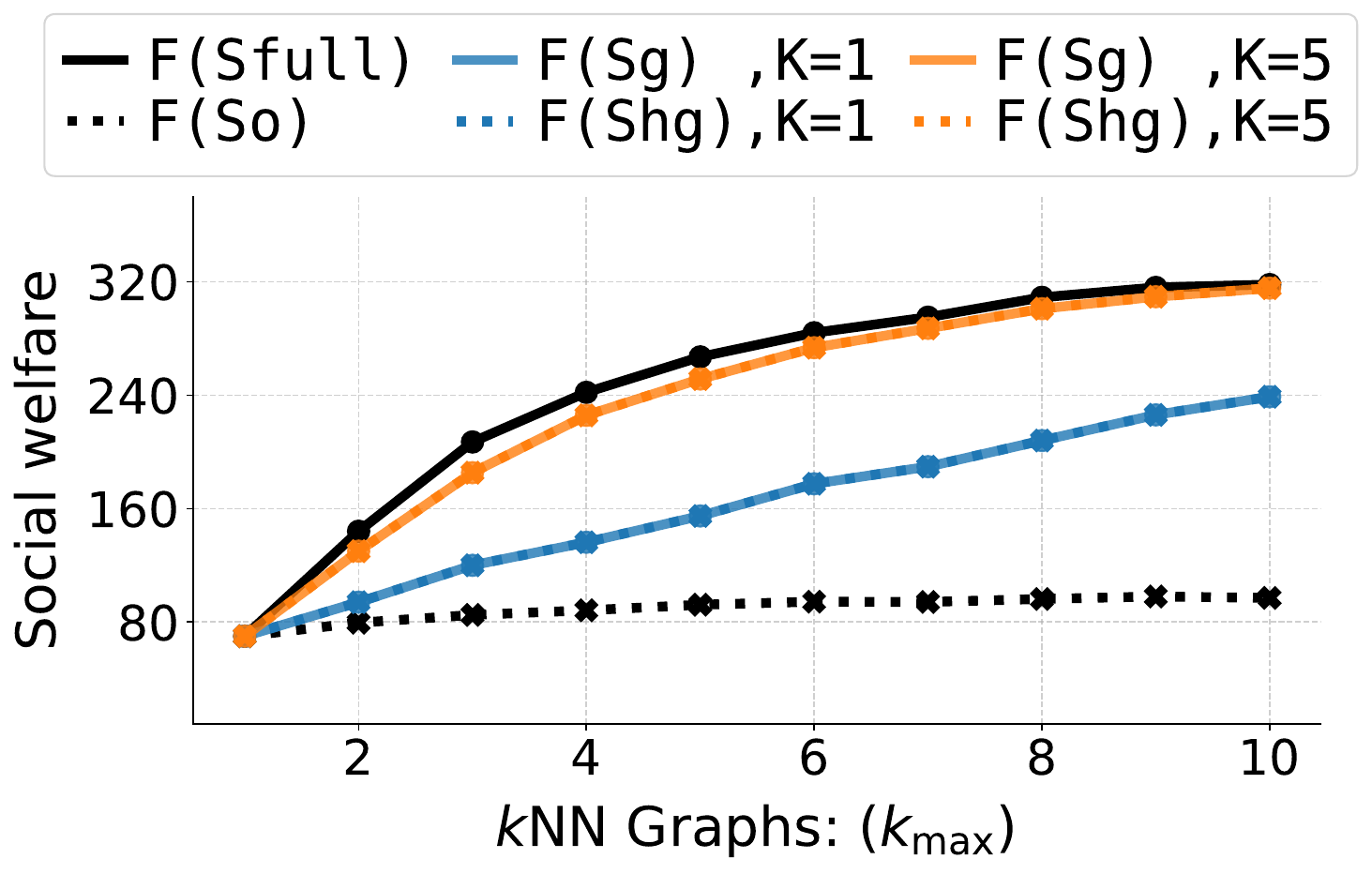}
    \caption{$F(S_{\mathrm{g}})$ vs.\ $F(S_{\mathrm{hg}})$}
    \label{fig:app_adult_sg_shg_knn}
\end{subfigure}
\hfill
\begin{subfigure}[t]{0.24\textwidth}
    \centering
    \includegraphics[width=\linewidth]{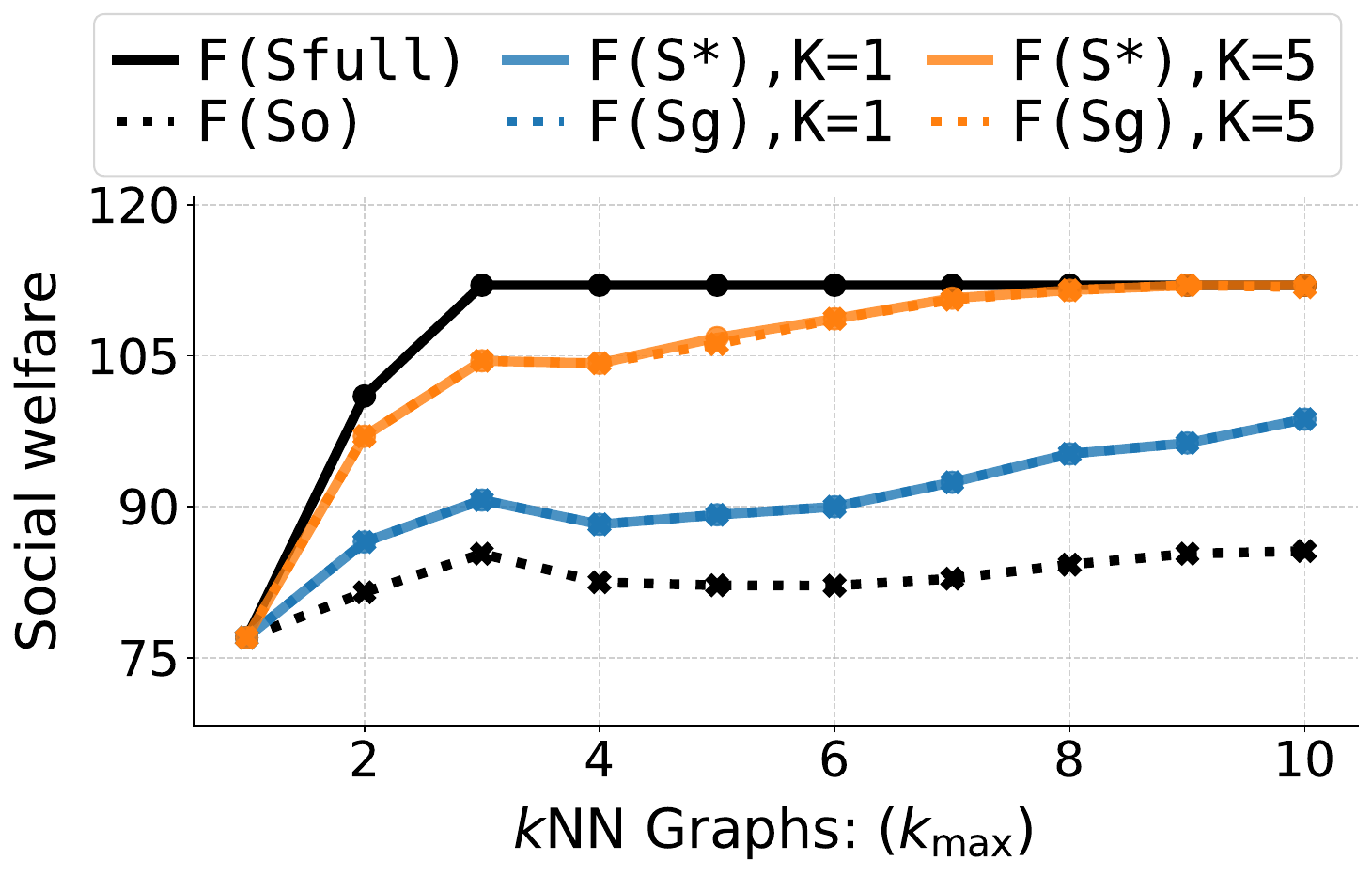}
    \caption{$F(S^{\star})$ vs.\ $F(S_{\mathrm{g}})$}
    \label{fig:app_prod_star_sg_knn}
\end{subfigure}
\begin{subfigure}[t]{0.24\textwidth}
    \centering
    \includegraphics[width=\linewidth]{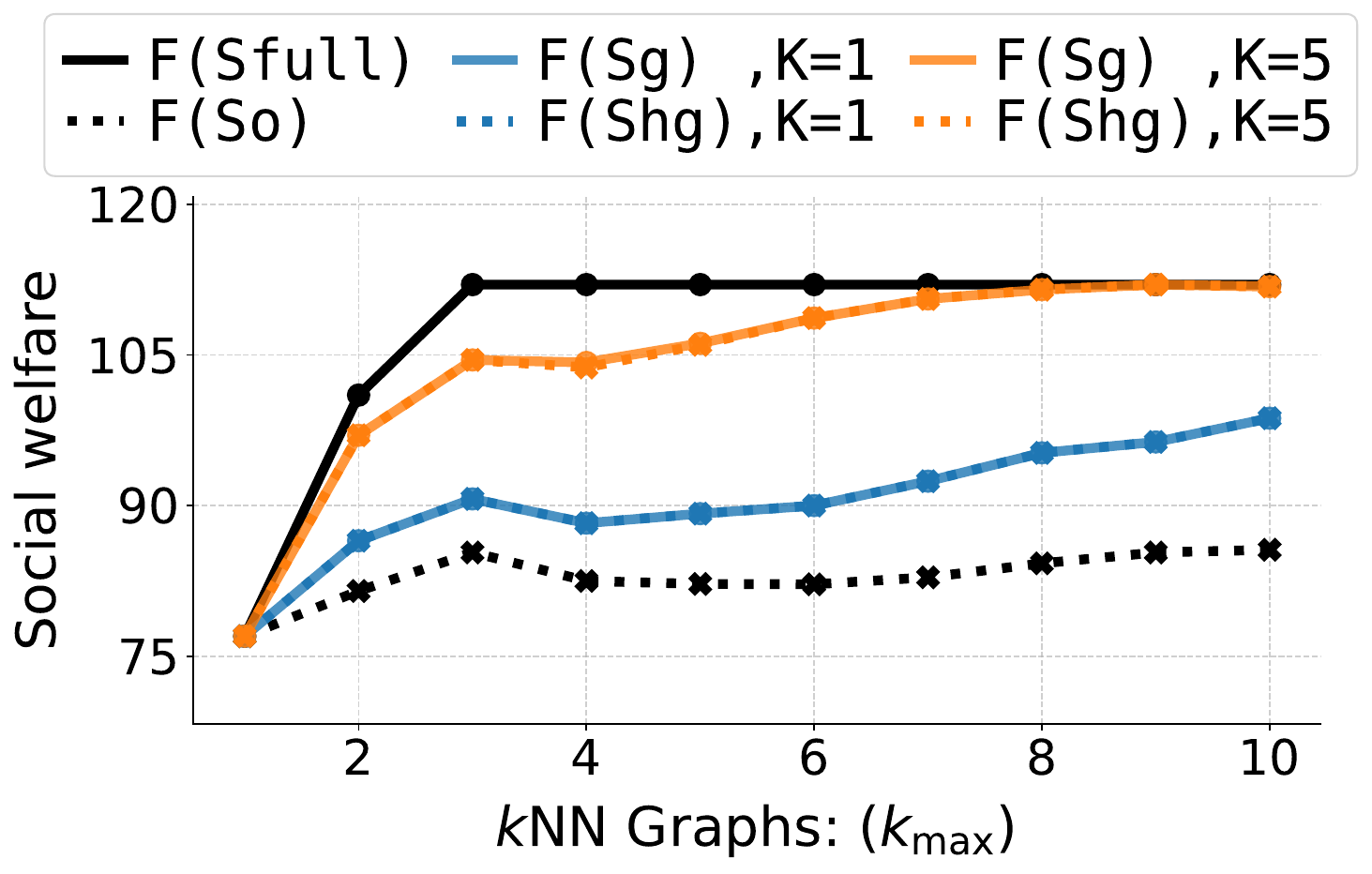}
    \caption{$F(S_{\mathrm{g}})$ vs.\ $F(S_{\mathrm{hg}})$}
    \label{fig:app_prod_sg_shg_knn}
\end{subfigure}\\
\vspace{0.99em}
\textit{Threshold Graphs}\par\medskip
\begin{subfigure}[t]{0.24\textwidth}
    \centering
    \includegraphics[width=\linewidth]{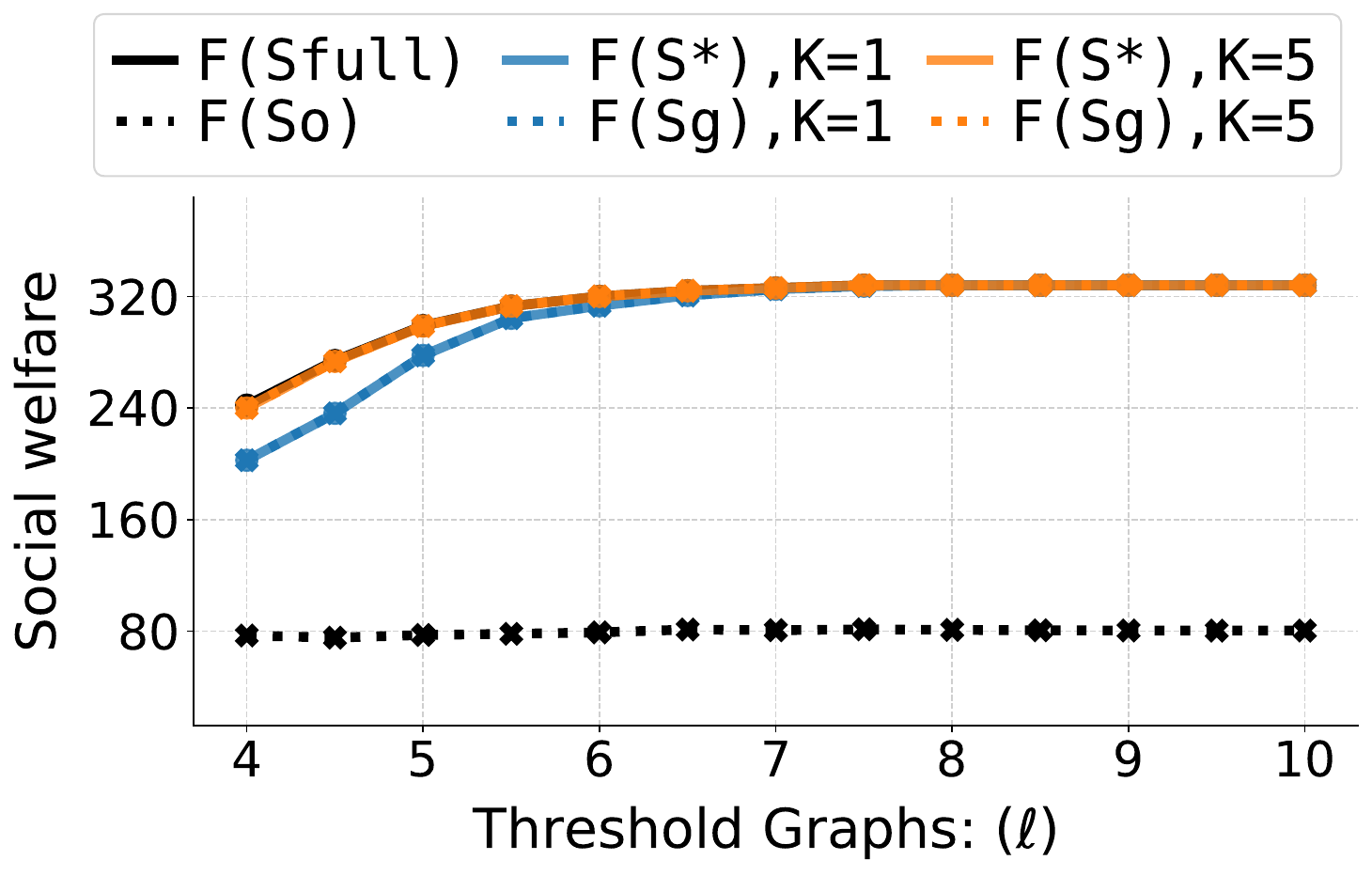}
    \caption{$F(S^{\star})$ vs.\ $F(S_{\mathrm{g}})$}
    \label{fig:app_adult_star_sg_thresh}
\end{subfigure}
\begin{subfigure}[t]{0.24\textwidth}
    \centering
    \includegraphics[width=\linewidth]{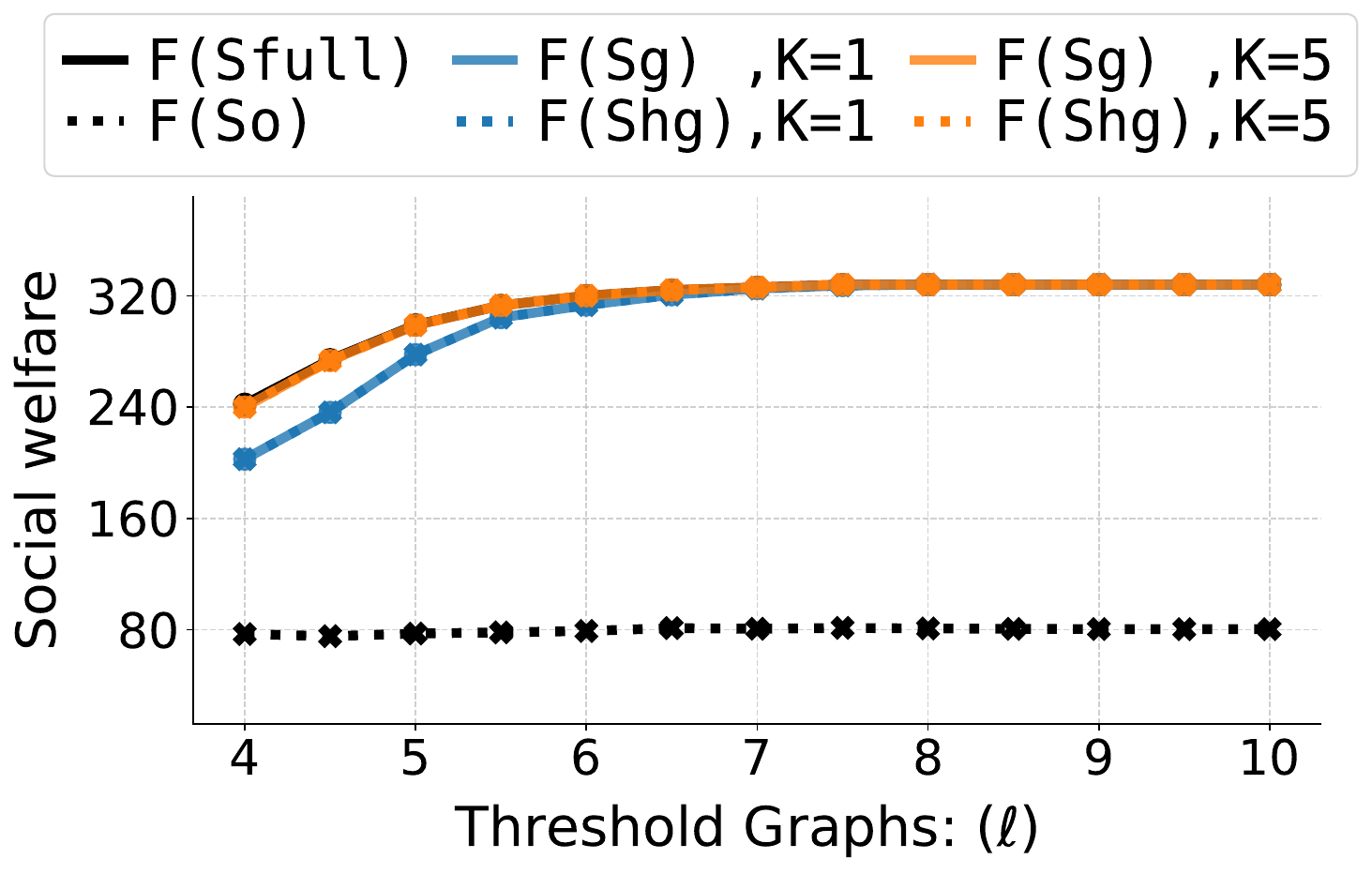}
    \caption{$F(S_{\mathrm{g}})$ vs.\ $F(S_{\mathrm{hg}})$}
    \label{fig:app_adult_sg_shg_thresh}
\end{subfigure}
\hfill
\begin{subfigure}[t]{0.24\textwidth}
    \centering
    \includegraphics[width=\linewidth]{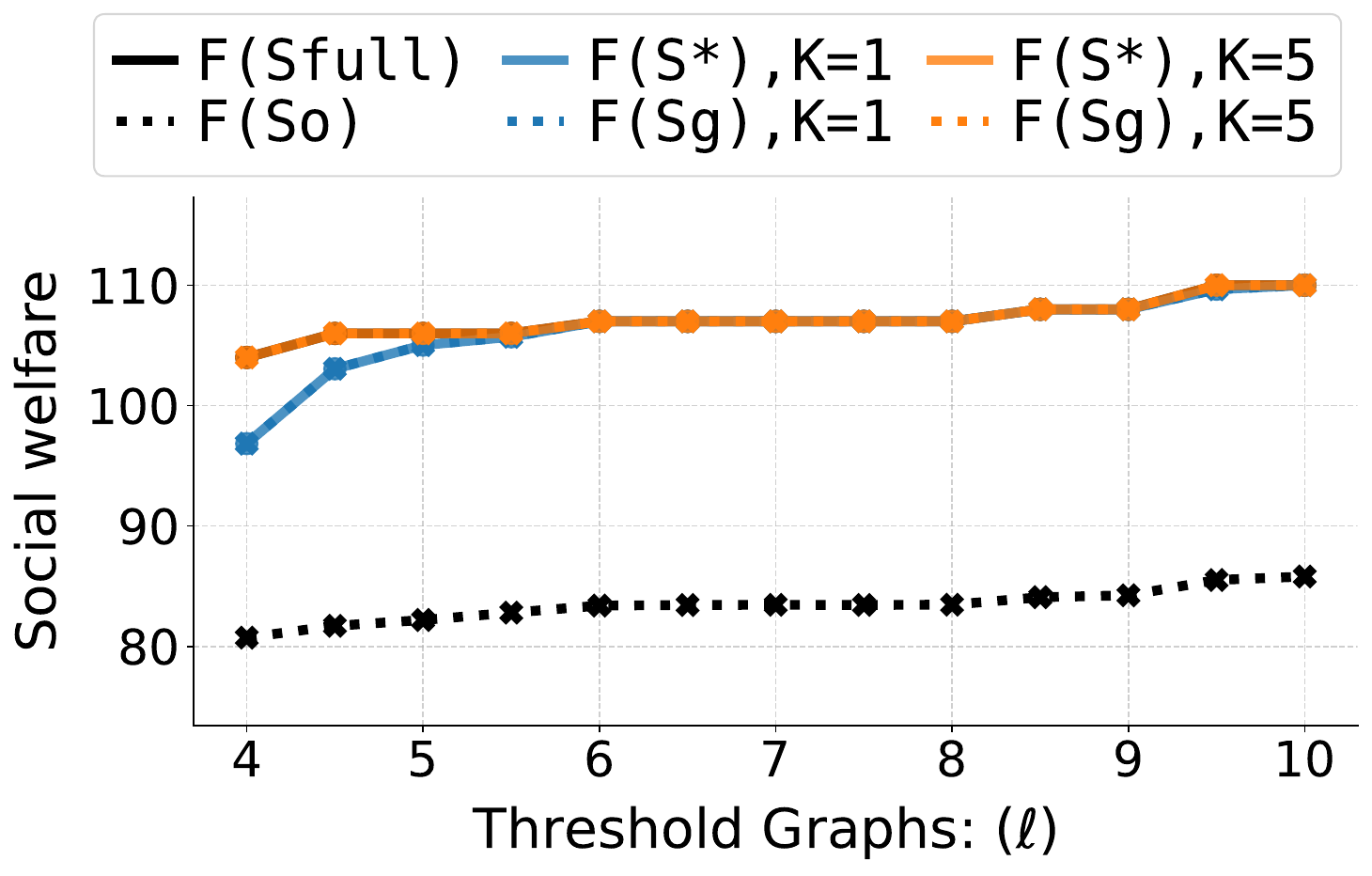}
    \caption{$F(S^{\star})$ vs.\ $F(S_{\mathrm{g}})$}
    \label{fig:app_prod_star_sg_thresh}
\end{subfigure}
\begin{subfigure}[t]{0.24\textwidth}
    \centering
    \includegraphics[width=\linewidth]{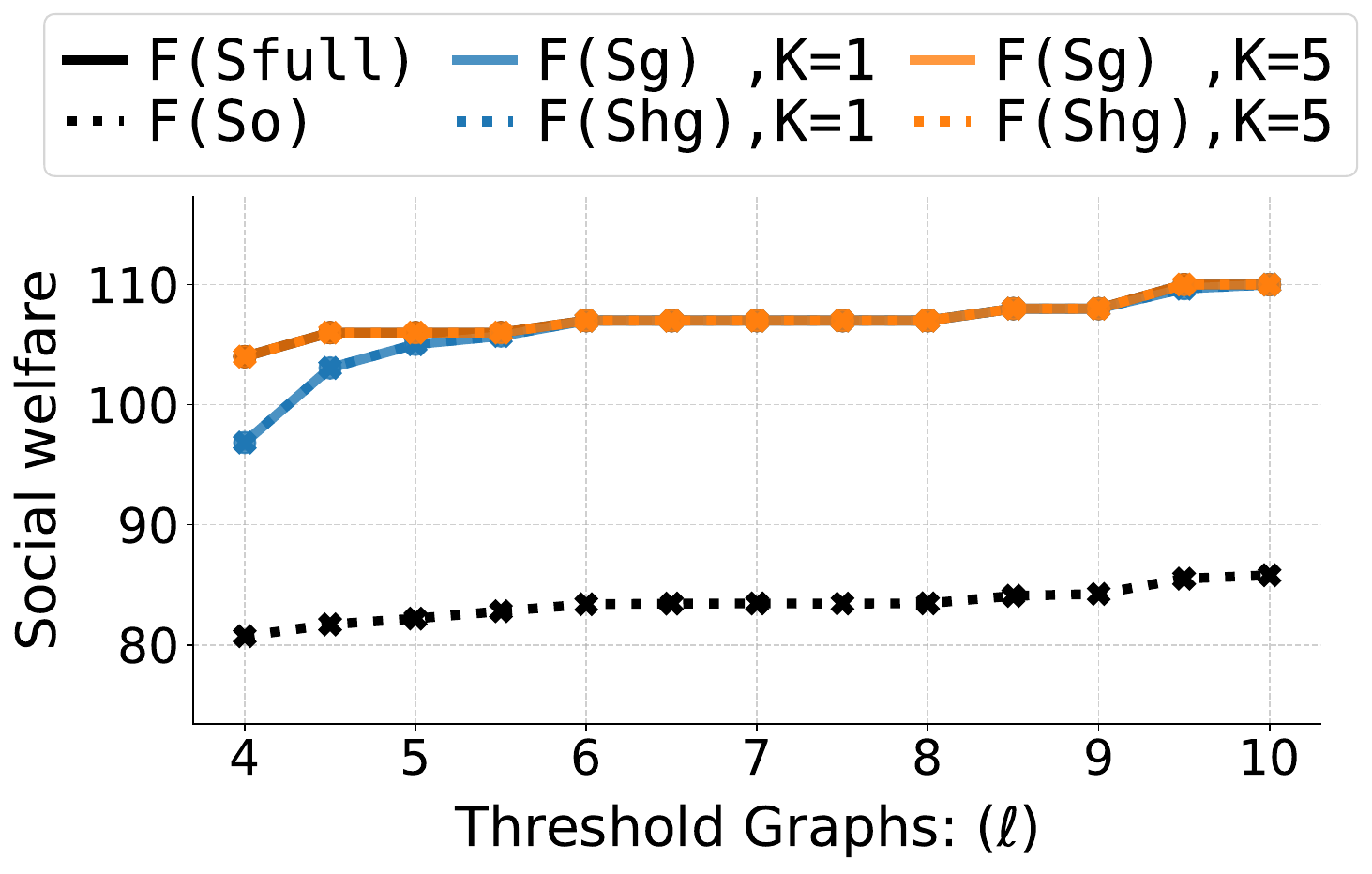}
    \caption{$F(S_{\mathrm{g}})$ vs.\ $F(S_{\mathrm{hg}})$}
    \label{fig:app_prod_sg_shg_thresh}
\end{subfigure}

\caption[Comparison of $F(S^{\star})$,  $F(S_{\mathrm{g}})$, and $F(S_{\mathrm{hg}})$]{Comparison of social welfare obtained via brute-force $F(S^{\star})$, classic greedy $F(S_{\mathrm{g}})$, and heuristic greedy $F(S_{\mathrm{hg}})$ on the Adult and Productivity datasets. Top row are $k$NN graphs and bottom row are threshold graphs. 
Across target reveal budgets $K=\{1,5\}$, in both datasets and graphs, both greedy variants get exact solutions ($F(S^{\star})$). As connectivity increases 
(cf. Tables~\ref{tab:adult_kmax_r_stats} and \ref{tab:productivity_kmax_r_stats}), both greedy variants closely  approximate the optimum $F(S_{\mathrm{full}})$ (Figures~\ref{fig:app_adult_star_sg_thresh}--\subref{fig:app_prod_sg_shg_thresh} where $\ell > 6$). Although we show only the Adult and Productivity datasets, we observe similar patterns on the Math and Portuguese datasets.}
\label{fig:adult-port_sstar_sg_shg}
\end{figure}

\paragraph{General observations.}
For all $k$NN generated graphs, when agents have atmost one target neighbor that is either positive or negative (Tables~\ref{tab:adult_kmax_r_stats}--\ref{tab:productivity_kmax_r_stats}, $k_{\max=1}$), 
revealing targets is unnecessary (Figures~\ref{fig:app_sr_sg}--\ref{fig:adult-port_sstar_sg_shg} subfigures (\textbf{a}--\textbf{d})). 

In sparse graphs, particularly when many agents have empty neighborhoods (Tables~\ref{tab:math_kmax_r_stats} and \ref{tab:portuguese_kmax_r_stats} where emptyNs $\ge 1$), overall social welfare remains close to zero regardless of the algorithm or budget 
(Figures~\ref{fig:app_math_sr_sg_thresh},\subref{fig:app_port_sr_sg_thresh} and 
\ref{fig:app_math_shg_shr_thresh},\subref{fig:app_port_shg_shr_thresh}).

As connectivity increases, especially in threshold-based graphs constructed with larger threshold values, resulting social welfare increases (e.g., Figures~\ref{fig:app_adult_sr_sg_thresh},\subref{fig:app_prod_sr_sg_thresh}) because of presence of positive target that are connected to all helpable agents (Tables~\ref{tab:adult_kmax_r_stats} and \ref{tab:productivity_kmax_r_stats}, $\ell\geq 8$ and $\textit{uni}^{+} \geq 1$).

\paragraph{Social welfare comparison: Algorithm~\ref{alg:greedy_lbreveal} vs. random selection ($F(S_{\mathrm{g}})$ vs. $F(S_{\mathrm{r}})$).}
In $k$NN generated graphs, the higher $k_{\max}$ is, the higher the social welfare achieved by the classic greedy algorithm, even at low budget levels (Figures~\ref{fig:app_adult_sr_sg_knn}--\subref{fig:app_prod_sr_sg_knn}). 
Although higher budgets generally lead to greater social welfare, particularly when using the classic greedy algorithm (Figures~\ref{fig:app_adult_sr_sg_knn}--\subref{fig:app_prod_sr_sg_knn}), and occasionally for the random algorithm as well (Figures~\ref{fig:app_math_sr_sg_knn},\subref{fig:app_prod_sr_sg_knn}), in some instances, such as in Figure~\ref{fig:app_adult_sr_sg_knn}, the budget appears to have little influence on the random algorithm's performance.
Overall, for the same budget \(K\), the Algorithm~\ref{alg:greedy_lbreveal} consistently attains social welfare that is at least that achieved by the random algorithm (Figures~\ref{fig:app_adult_sr_sg_knn}--\subref{fig:app_prod_sr_sg_knn}). But, when the random algorithm operates at a higher budget, it can occasionally result in higher social welfare than Algorithm~\ref{alg:greedy_lbreveal} at a lower budget 
(Figures~\ref{fig:app_port_sr_sg_knn}--\subref{fig:app_prod_sr_sg_knn}).
However, random selection generally performs poorly, and revealing additional targets often yields little to no increase in social welfare. E.g., see Figure~\ref{fig:app_adult_sr_sg_knn} for all values of \(k_{\max}\) and \(K\), and Figure~\ref{fig:app_port_sr_sg_knn} when \(k_{\max} \geq 5\) and \(K = 1\).

In threshold-based graphs, particularly those constructed with larger threshold values ($\ell$), the advantage of the greedy algorithm over the random baseline becomes more pronounced due to increased connectivity. Even with limited budgets, the greedy approach often attains near-maximal social welfare (Figures~\ref{fig:app_adult_sr_sg_thresh}--\subref{fig:app_prod_sr_sg_thresh}).
In contrast, the performance of the random algorithm varies substantially. In graphs where most targets are negative (Table~\ref{tab:adult_kmax_r_stats}), random selection yields very low social welfare (Figures~\ref{fig:app_adult_sr_sg_knn},\subref{fig:app_adult_sr_sg_thresh}). However, in graphs with a relatively large number of positive targets connected to nearly all helpable agents, where the probability of revealing one at random is higher (Tables~\ref{tab:math_kmax_r_stats}--\ref{tab:productivity_kmax_r_stats}), the random algorithm performs considerably better (Figures~\ref{fig:app_math_sr_sg_thresh}--\subref{fig:app_prod_sr_sg_thresh}).

\paragraph{Social welfare comparison: heuristic greedy vs. heuristic random ($F(S_{\mathrm{hg}})$ vs. $F(S_{\mathrm{hr}})$).}
Under low connectivity, particularly in graphs generated using the $k$NN method, the heuristic random algorithm still achieves a high social welfare but in some cases remains significantly below the heuristic greedy algorithm at a similar budget level 
(Figures~\ref{fig:app_adult_shg_shr_knn}--\subref{fig:app_prod_shg_shr_knn}). 
In settings where random algorithm produced low social welfare (Figures~\ref{fig:app_adult_sr_sg_knn},\subref{fig:app_port_sr_sg_knn}), the heuristic random algorithm performs markedly better (Figure~\ref{fig:app_adult_shg_shr_knn},\subref{fig:app_port_shg_shr_knn}). 
This improvement arises because the selection becomes localized, and randomly choosing among positive targets is more effective than selecting from all targets.
Performance further improves with higher connectivity: the social welfare returned by the heuristic random and heuristic greedy become nearly identical, even with graphs generated with a low threshold value (Figures~\ref{fig:app_math_shg_shr_thresh}--\subref{fig:app_prod_shg_shr_thresh}).

\paragraph{Social welfare comparison: bruteforce search vs. Algorithm~\ref{alg:greedy_lbreveal} vs. heuristic greedy ($F(S^\star)$ vs. $F(S_{\mathrm{g}})$ vs. $F(S_{\mathrm{hg}})$).}
Across bipartite graphs constructed via thresholding or $ k$NN, bruteforce search, Algorithm~\ref{alg:greedy_lbreveal}, and heuristic greedy achieve comparable social welfare across all budgets and datasets (Figure~\ref{fig:adult-port_sstar_sg_shg}). This similarity arises because the optimal target sets mainly consist of positive targets, leading all algorithms to converge on nearly identical solutions at similar budgets. These findings indicate that greedy approaches may still perform well in practice when there are no information disclosure restrictions, likely because graphs based on real-world tend to be well connected and balanced.

\subsection{Empirical Results for Fairness}
\label{sec:revealrm_app=fairnessexp}
\begin{figure}[b!]
\vspace{0.66cm}
\captionsetup[subfigure]{justification=Centering}
\begin{subfigure}[t]{0.32\textwidth}
\centering
    \includegraphics[width=\textwidth]{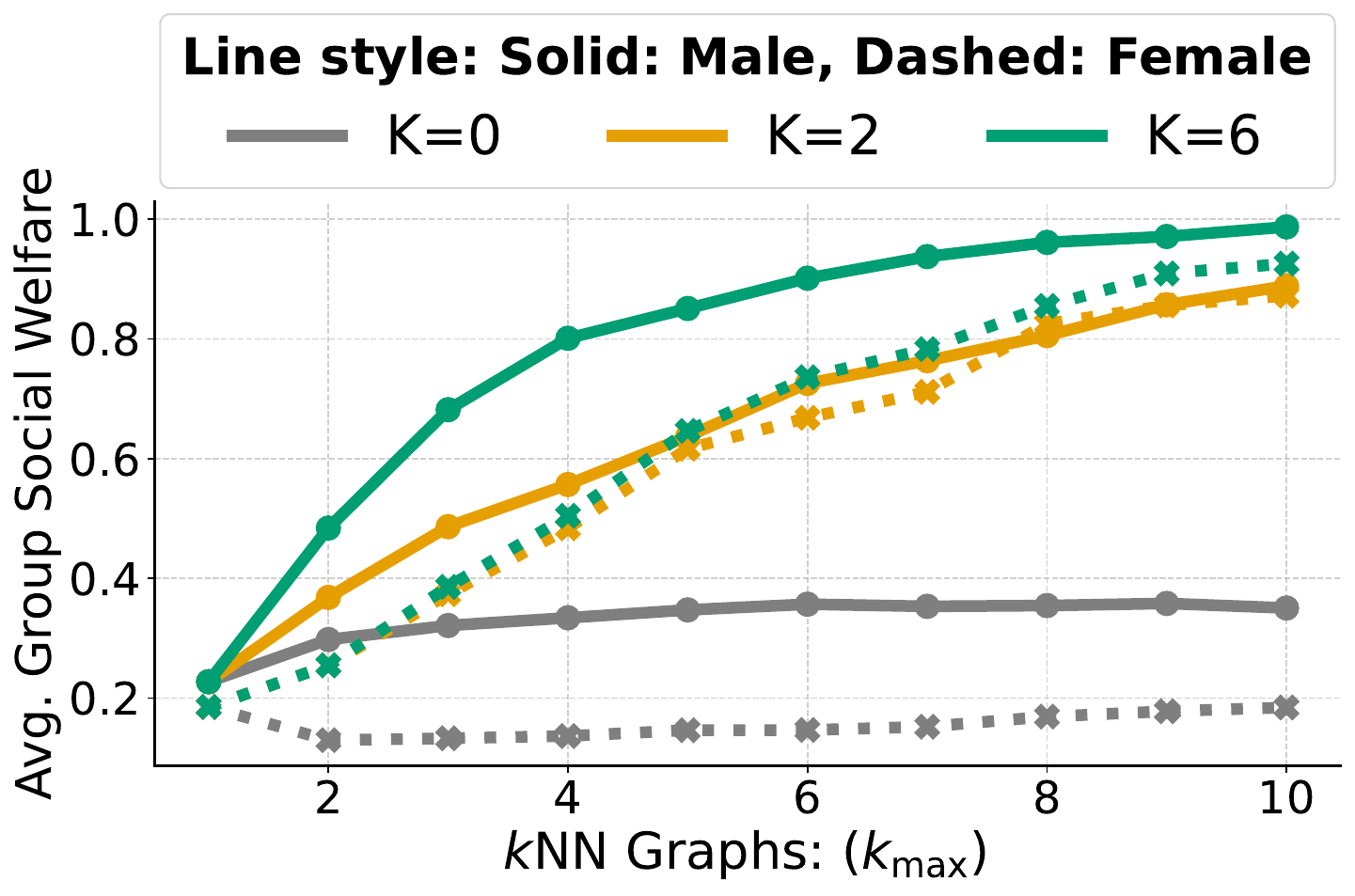}
    \caption{\(k\)NN graphs: (male vs. female)}
    \label{fig:app_adult_knn_undivided}
\end{subfigure}
\begin{subfigure}[t]{0.32\textwidth}
\centering
    \includegraphics[width=\linewidth]{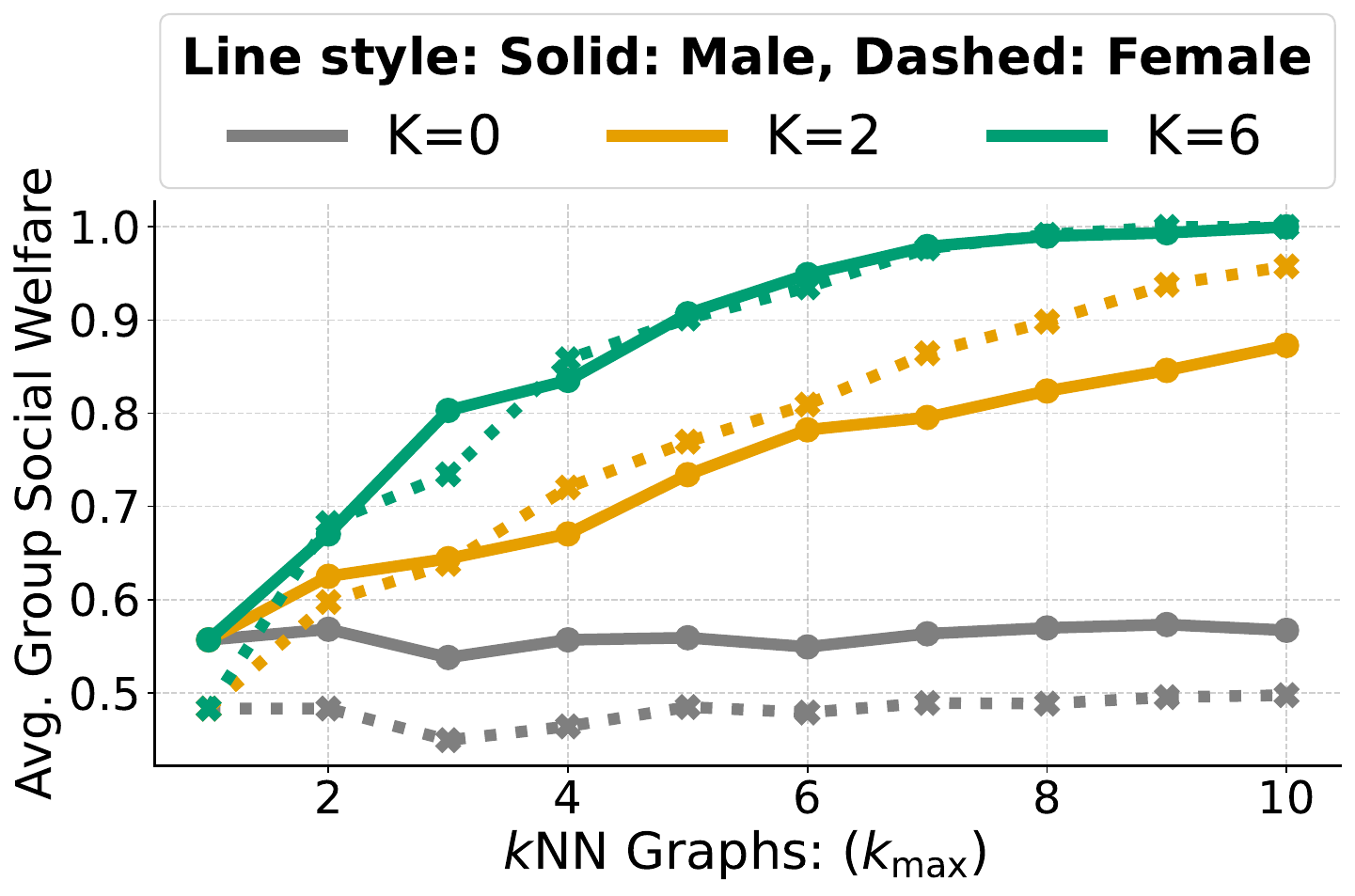}
    \caption{\(k\)NN graphs: (male vs. female)}
    \label{fig:app_math_knn_undivided}
\end{subfigure}
\begin{subfigure}[t]{0.32\textwidth}
\centering
    \includegraphics[width=\linewidth]{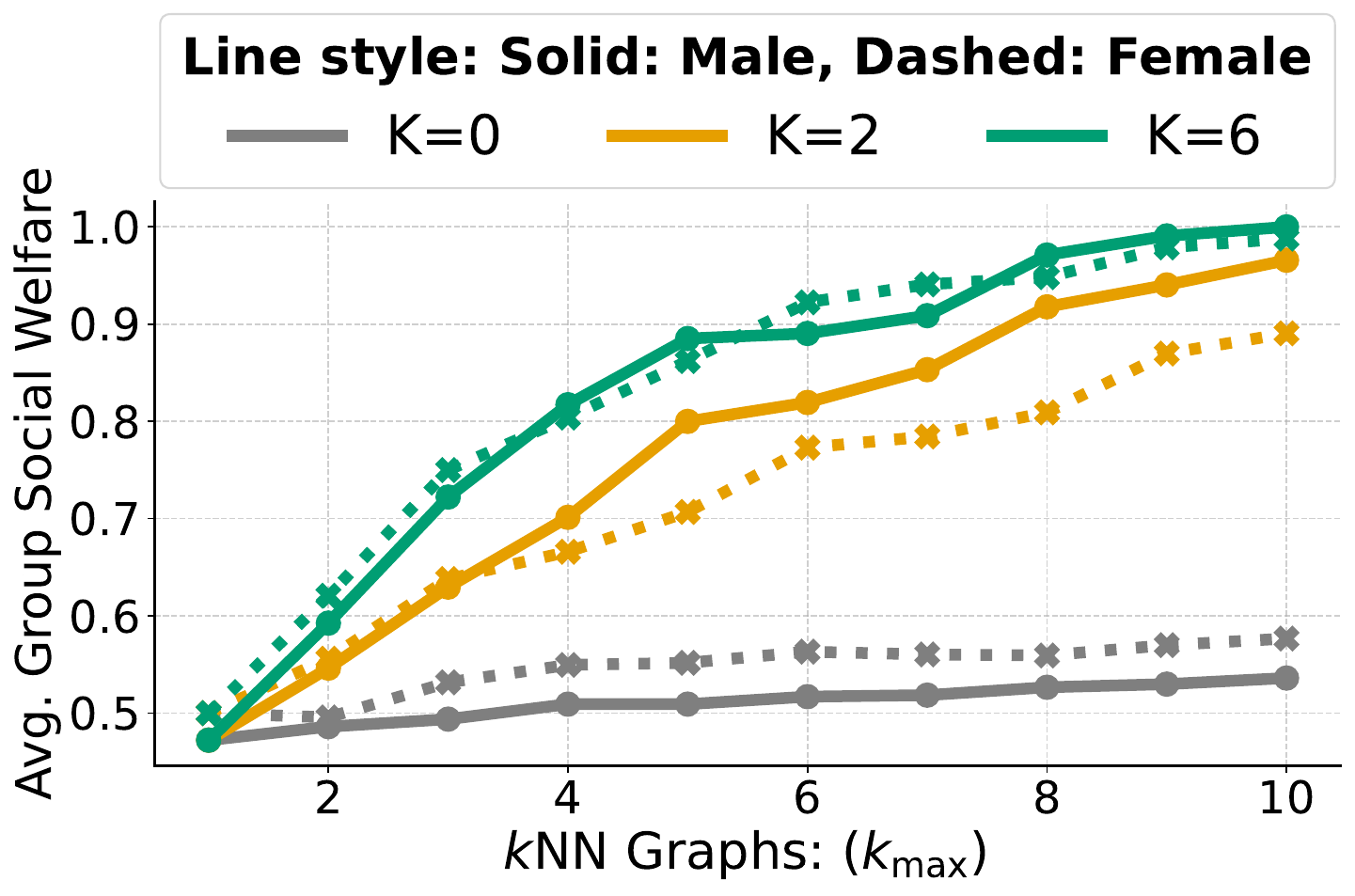}
    \caption{\(k\)NN graphs: (male vs. female)}
    \label{fig:app_port_knn_undivided}
\end{subfigure}\\[2ex]

\begin{subfigure}[t]{0.32\textwidth}
\centering
    \includegraphics[width=\textwidth]{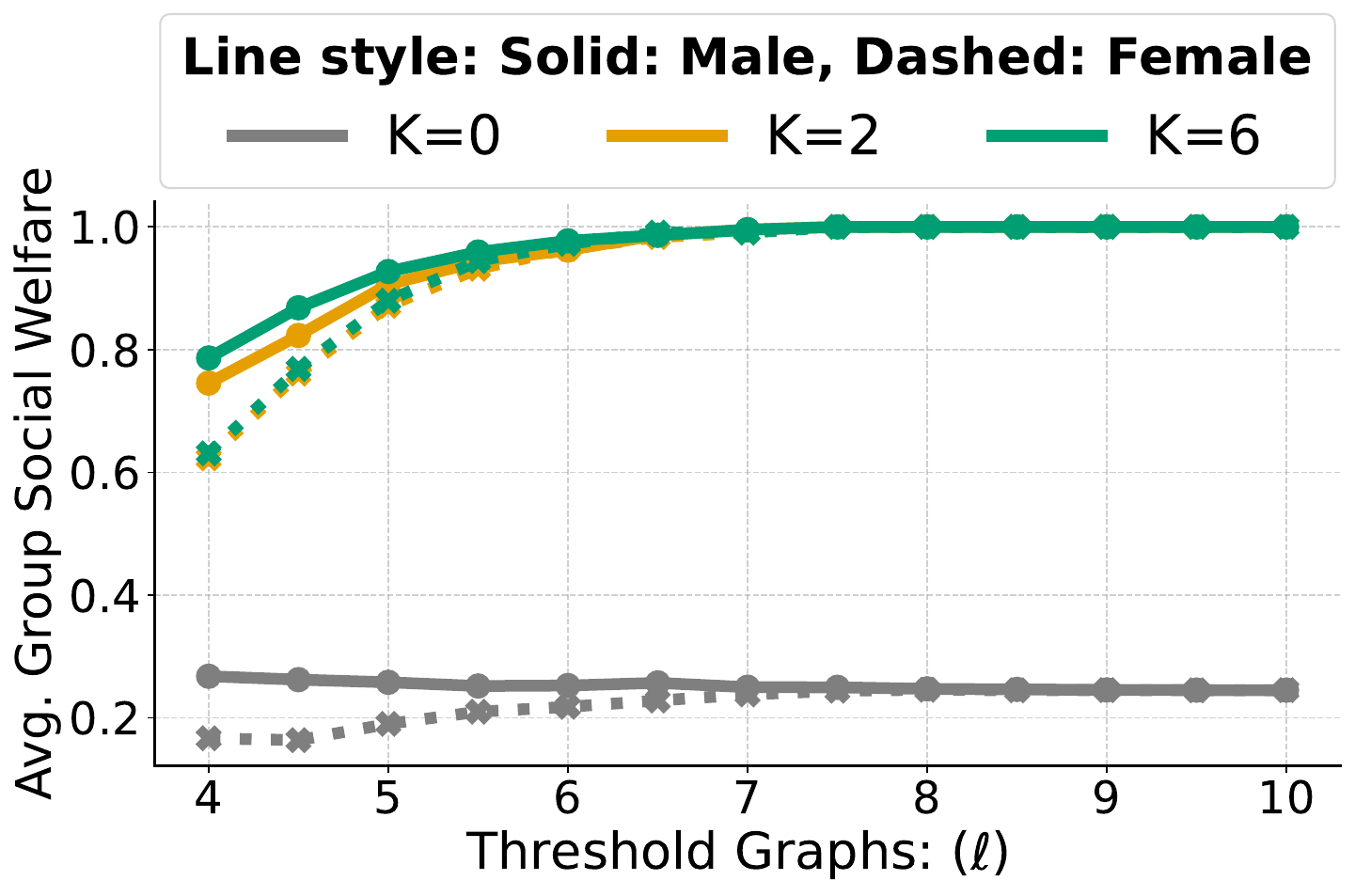}
    \caption{Threshold graphs: (male vs. female)}
    \label{fig:app_adult_thresh_undivided}
\end{subfigure}
\begin{subfigure}[t]{0.32\textwidth}
\centering
    \includegraphics[width=\linewidth]{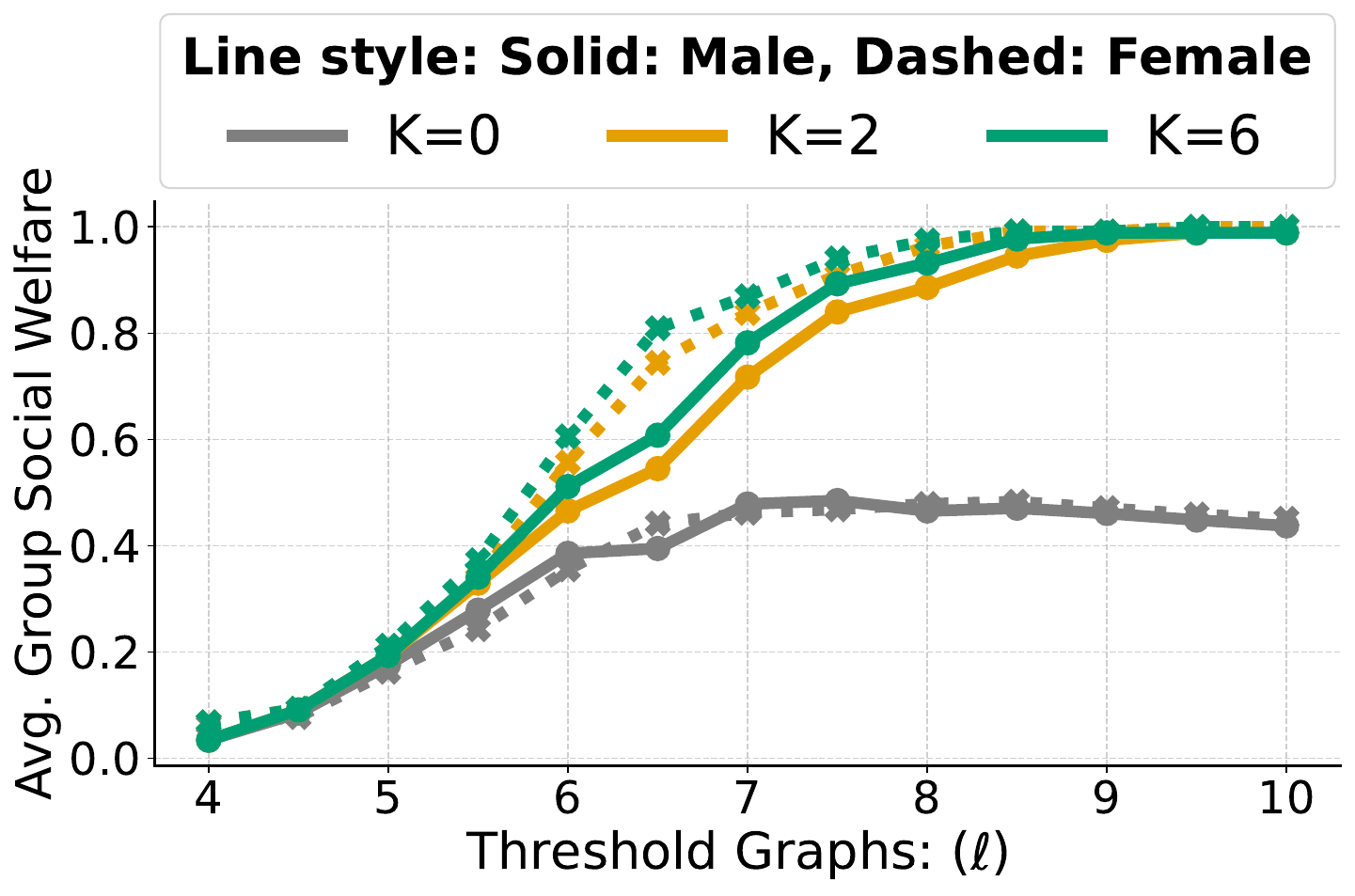}
    \caption{Threshold graphs: (male vs. female)}
    \label{fig:app_math_thresh_undivided}
\end{subfigure}
\begin{subfigure}[t]{0.32\textwidth}
\centering
    \includegraphics[width=\linewidth]{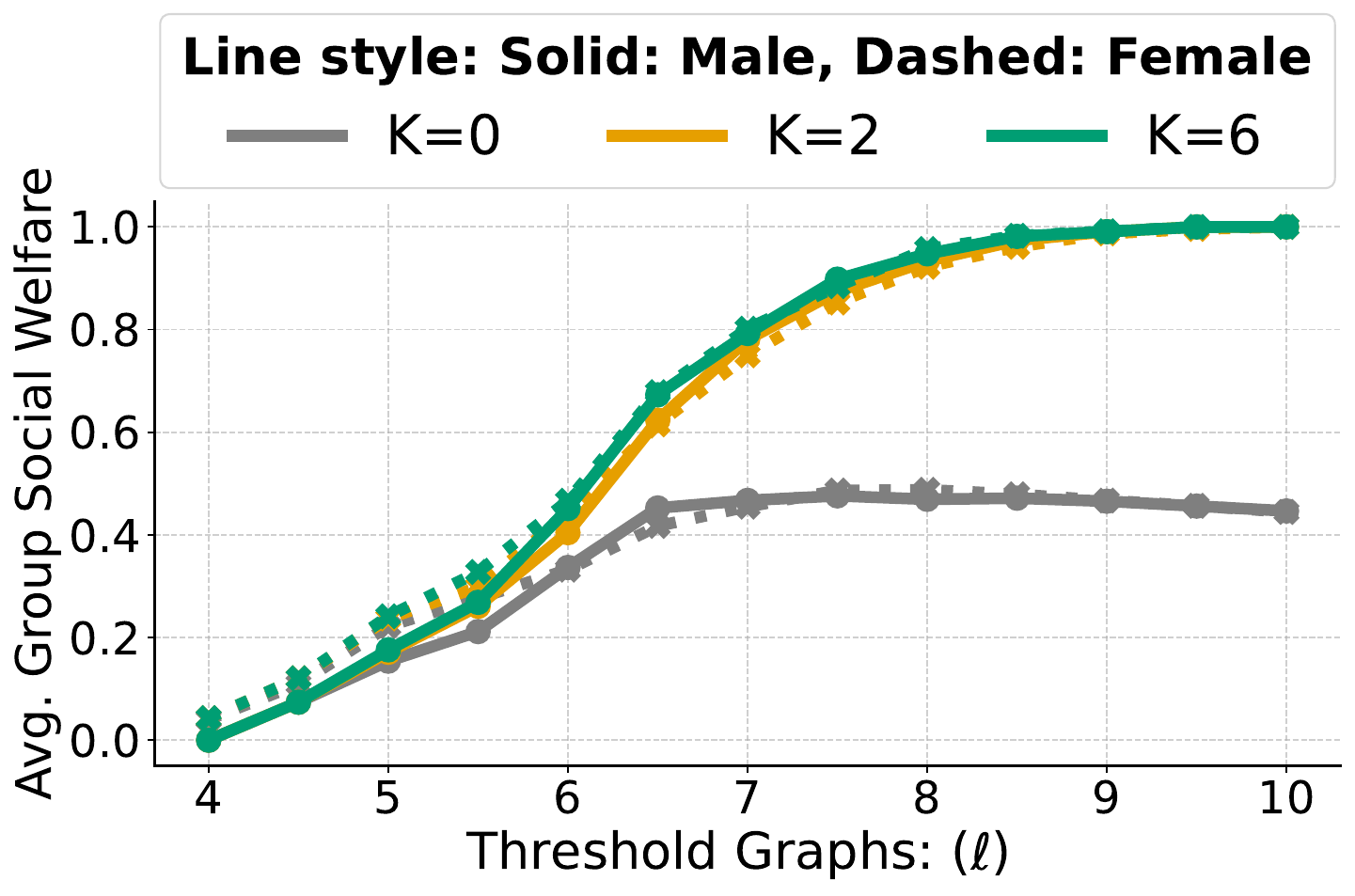}
    \caption{Threshold graphs: (male vs. female)}
    \label{fig:app_port_thresh_undivided}
\end{subfigure}

\caption[Comparative analysis of the average group social welfare when classic greedy approach is run on the whole graph]{Comparative analysis of the average group social welfare generated from running the classic greedy approach across \(k\)NN graphs (\subref{fig:app_adult_knn_undivided}--\subref{fig:app_port_knn_undivided}) and the threshold graphs (\subref{fig:app_adult_thresh_undivided}--\subref{fig:app_port_thresh_undivided}) from the \(3\) datasets (Tables~\ref{tab:adult_kmax_r_stats}--\ref{tab:portuguese_kmax_r_stats}) on the \textbf{whole graph}. When the graph connectivity is high, there is no disparity in social welfare generated for the 2 groups.}
\label{fig:app_knn_thresh_undivided}

\begin{picture}(0,0)
    \put(18,366){{\parbox{4cm}{\centering \textbf{Adult}}}}
    \put(178,366){{\parbox{4cm}{\centering \textbf{Math}}}}
    \put(325,366){{\parbox{4cm}{\centering \textbf{Portuguese}}}}
\end{picture}

\end{figure}

\begin{figure}[t!]
\captionsetup[subfigure]{justification=Centering}
\begin{subfigure}[t]{0.32\textwidth}
\centering
    \includegraphics[width=\textwidth]{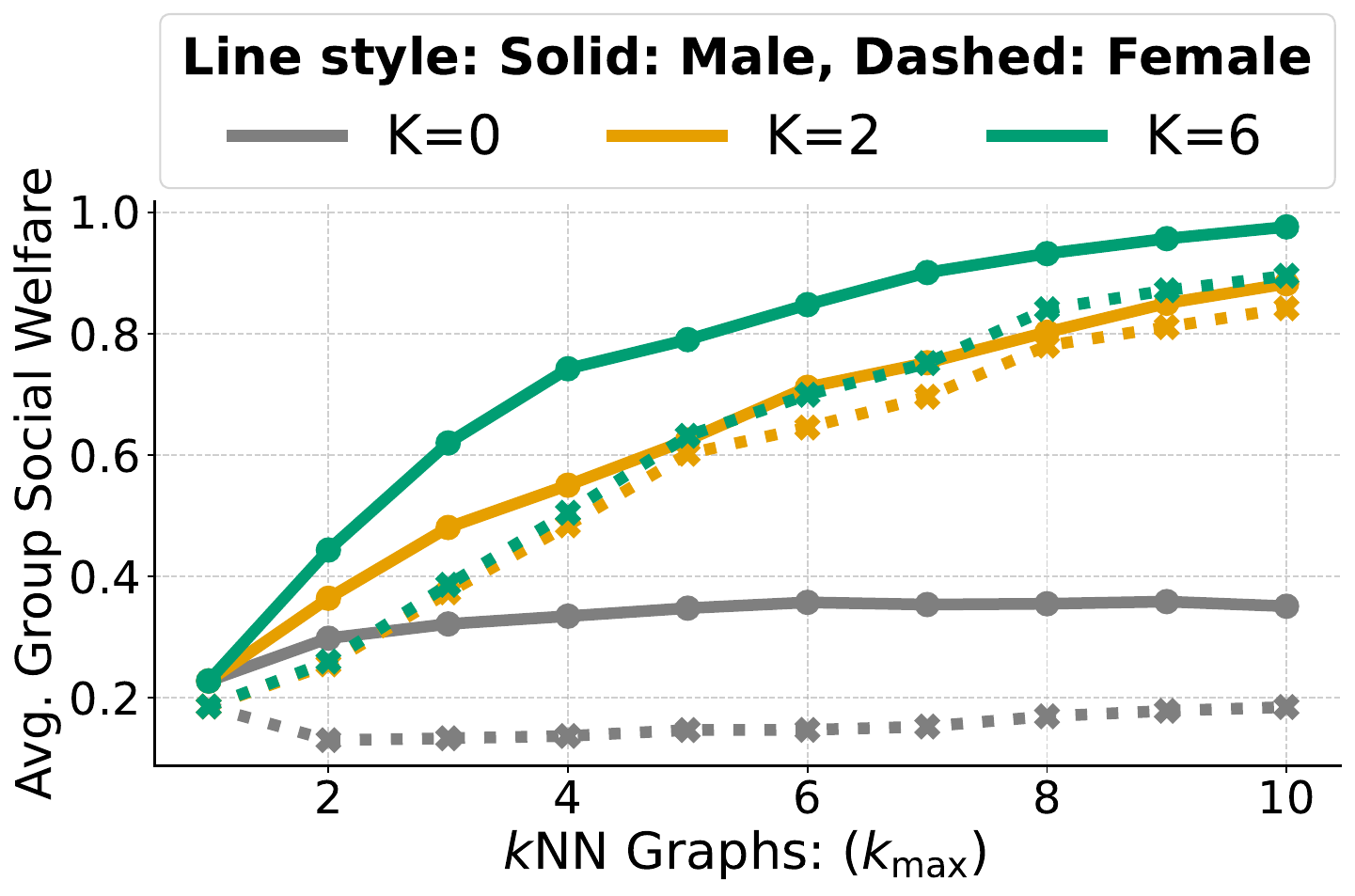}
    \caption{\(k\)NN graphs: (male vs. female)}
    \label{fig:app_adult_knn_divided}
\end{subfigure}
\begin{subfigure}[t]{0.32\textwidth}
\centering
    \includegraphics[width=\linewidth]{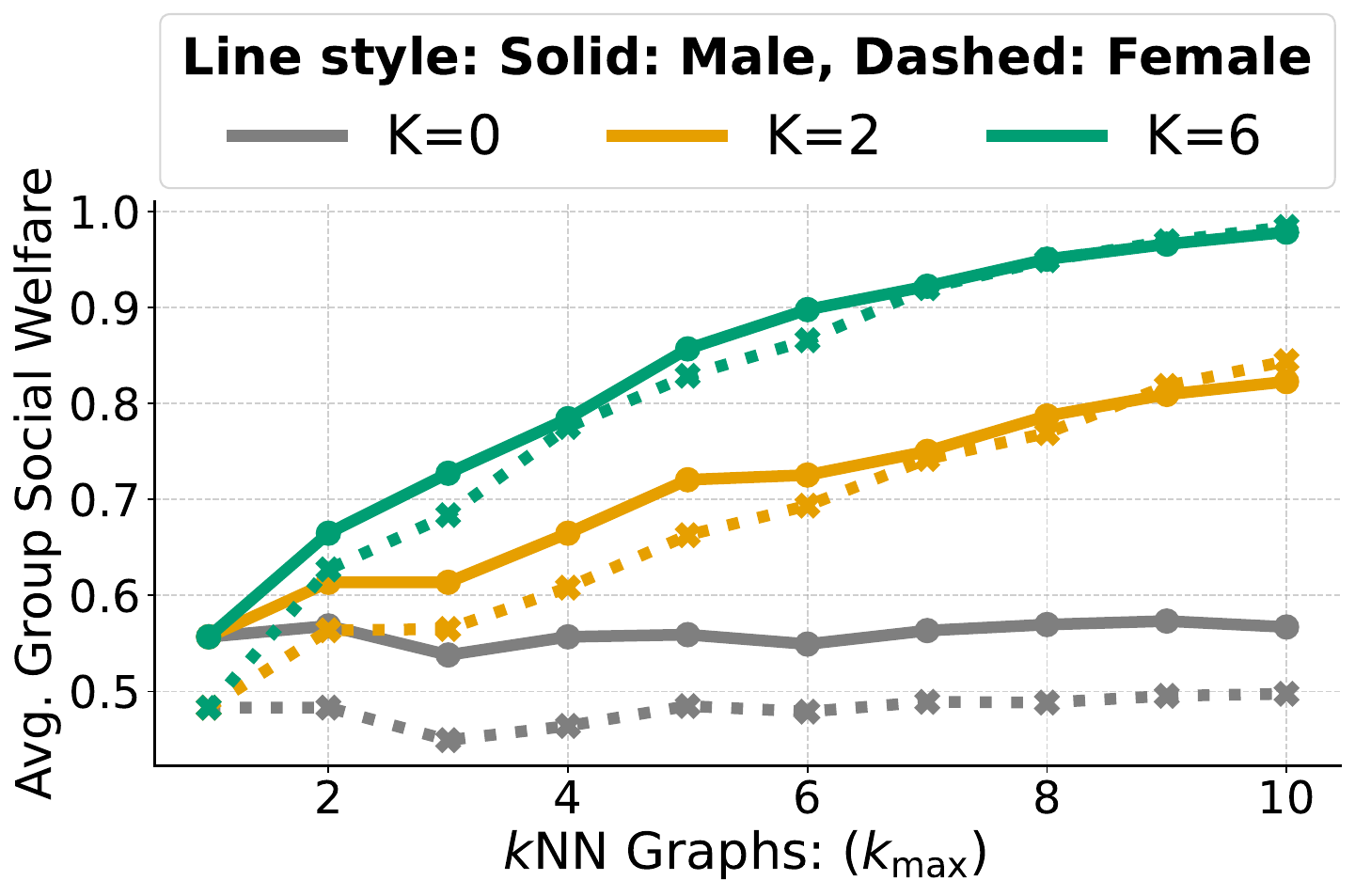}
    \caption{\(k\)NN graphs: (male vs. female)}
    \label{fig:app_math_knn_divided}
\end{subfigure}
\begin{subfigure}[t]{0.32\textwidth}
\centering
    \includegraphics[width=\linewidth]{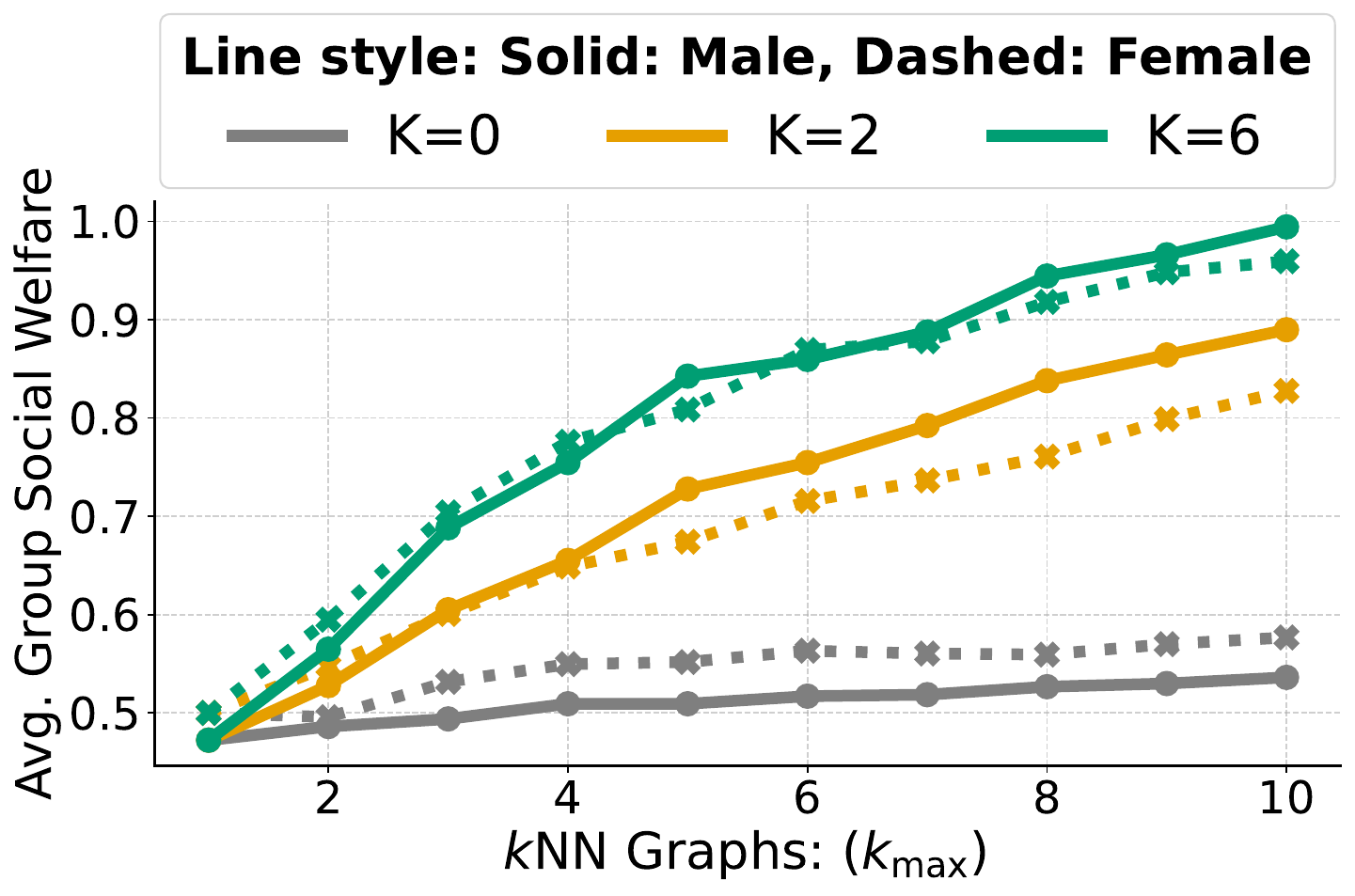}
    \caption{\(k\)NN graphs: (male vs. female)}
    \label{fig:app_port_knn_divided}
\end{subfigure}\\[2ex]

\begin{subfigure}[t]{0.32\textwidth}
\centering
    \includegraphics[width=\textwidth]{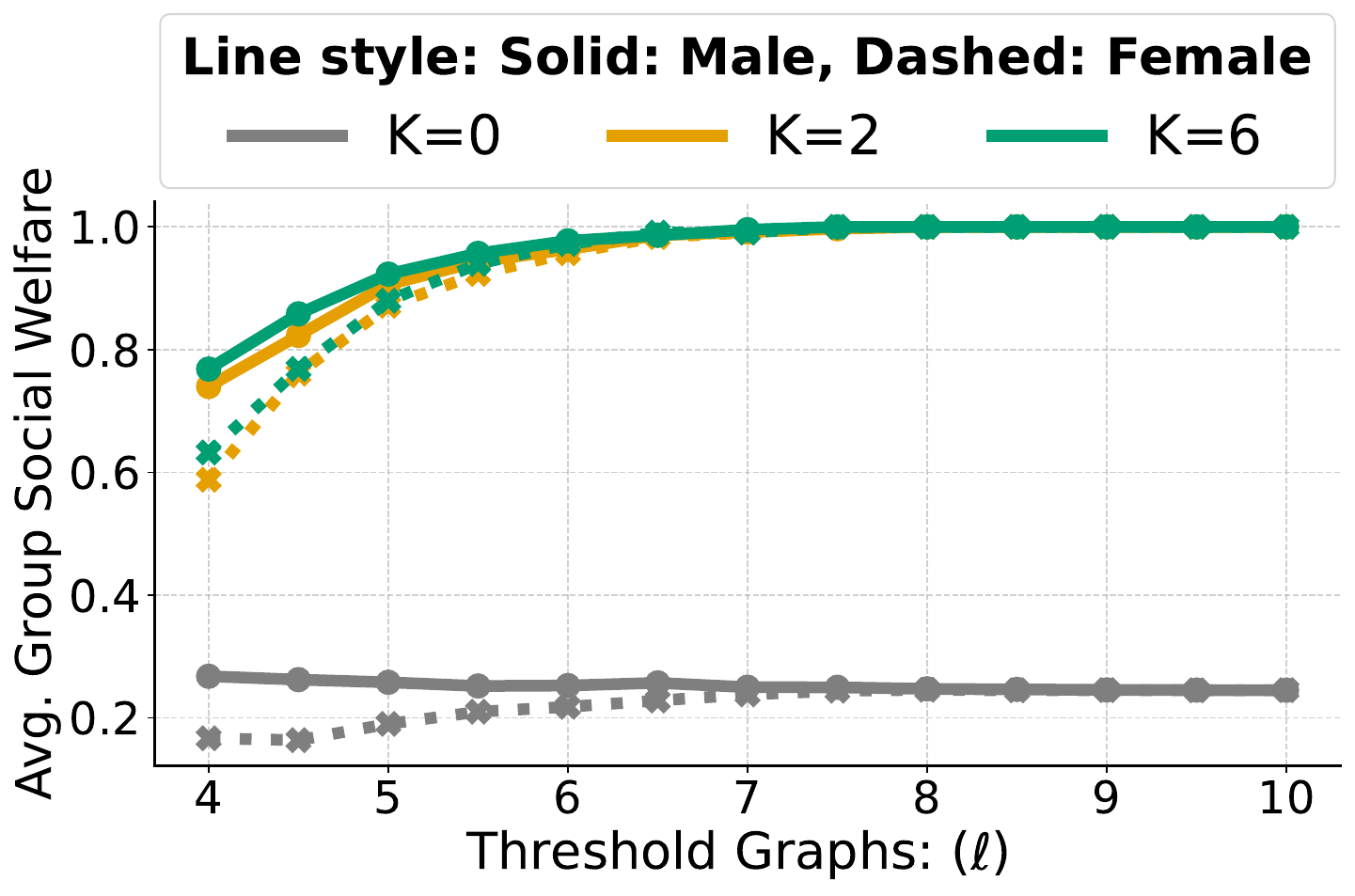}
    \caption{Threshold graphs: (male vs. female)}
    \label{fig:app_adult_thresh_divided}
\end{subfigure}
\begin{subfigure}[t]{0.32\textwidth}
\centering
    \includegraphics[width=\linewidth]{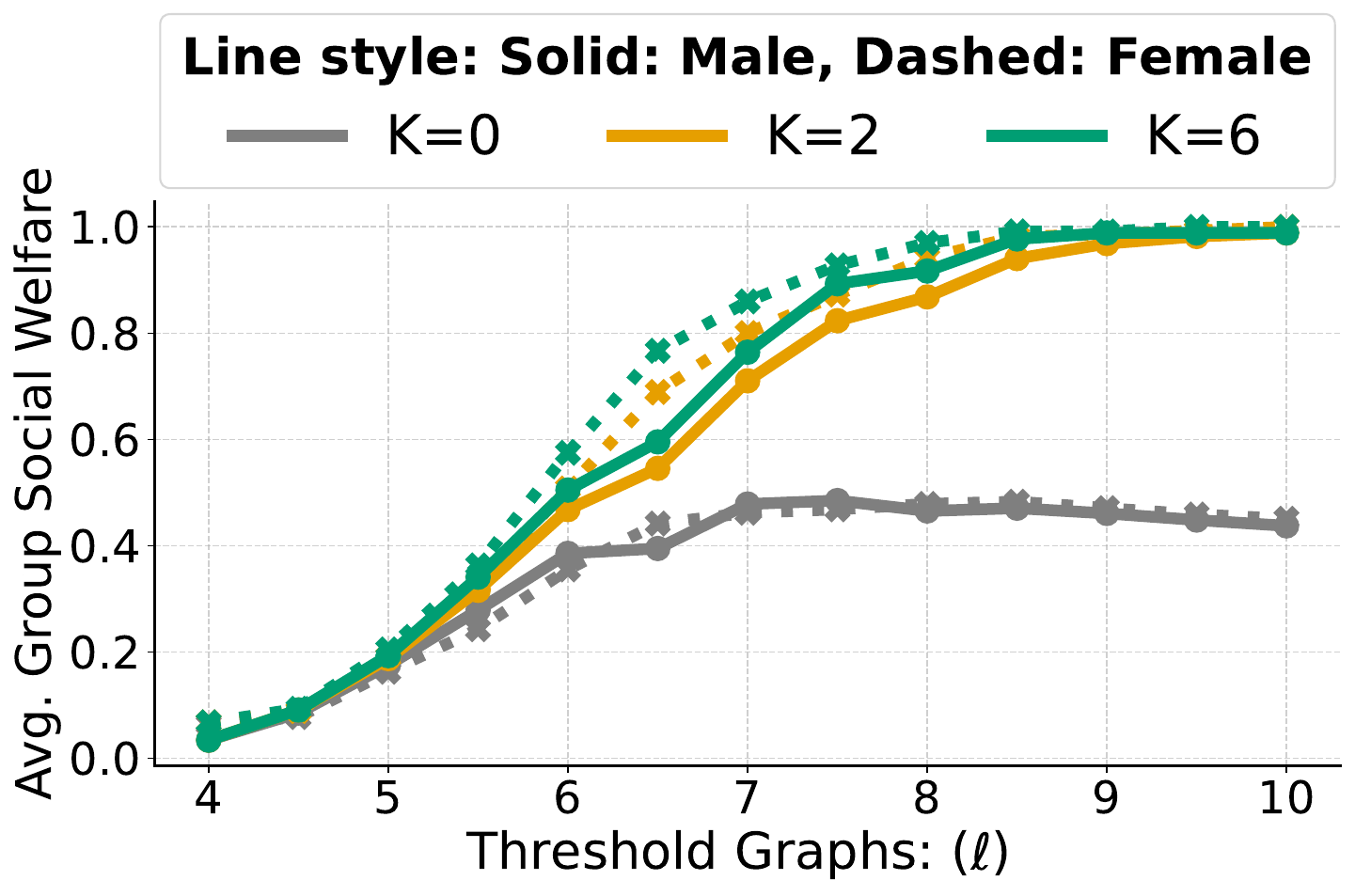}
    \caption{Threshold graphs: (male vs. female)}
    \label{fig:app_math_thresh_divided}
\end{subfigure}
\begin{subfigure}[t]{0.32\textwidth}
\centering
    \includegraphics[width=\linewidth]{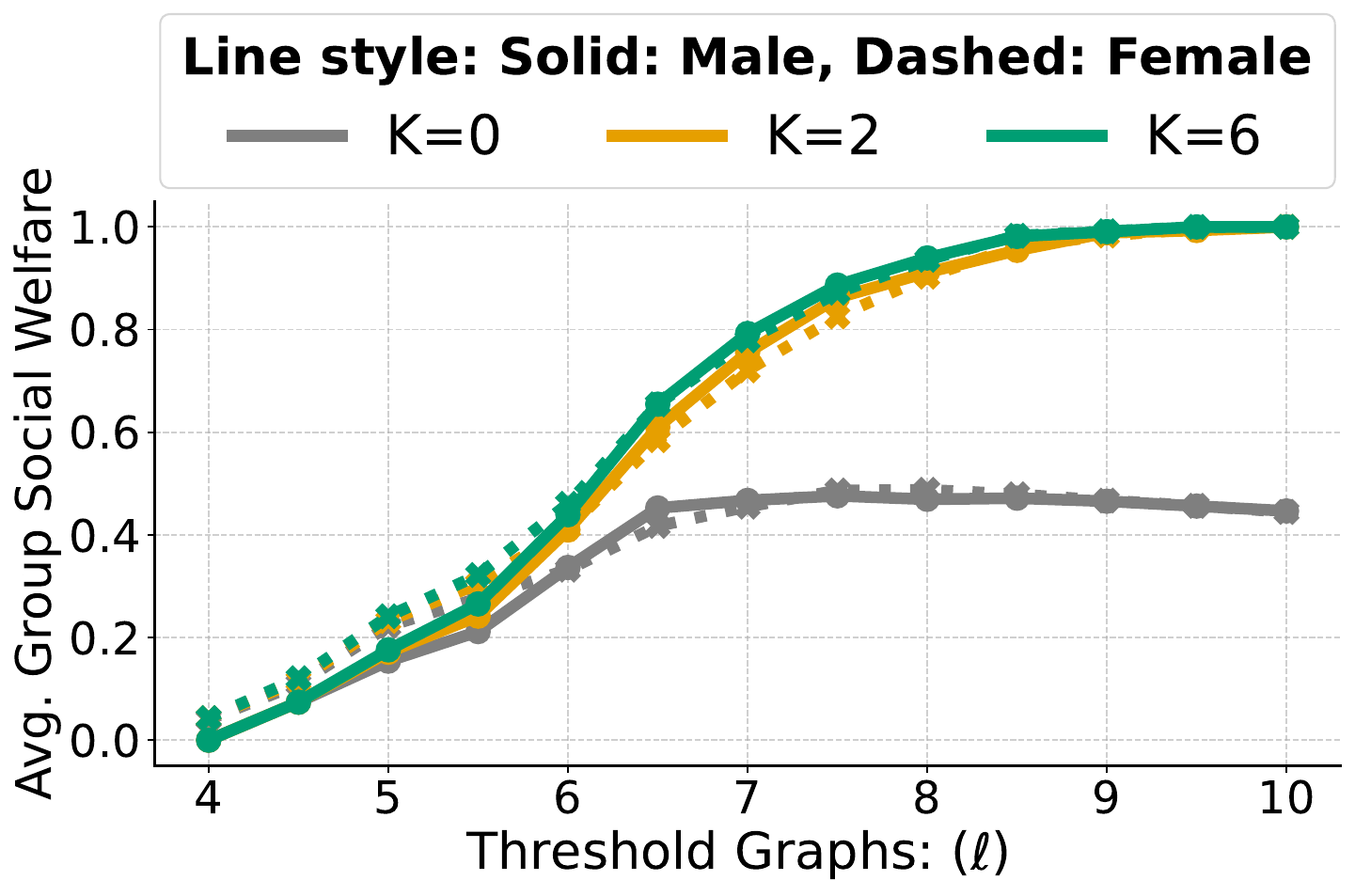}
    \caption{Threshold graphs: (male vs. female)}
    \label{fig:app_port_thresh_divided}
\end{subfigure}

\caption[Comparative analysis of the average group social welfare when classic greedy approach is run on the group subgraphs]{Comparative analysis of the average group social welfare generated from running the classic greedy approach across \(k\)NN graphs (\subref{fig:app_adult_knn_divided}--\subref{fig:app_port_knn_divided}) and the threshold graphs (\subref{fig:app_adult_thresh_divided}--\subref{fig:app_port_thresh_divided}) from the \(3\) datasets (Tables~\ref{tab:adult_kmax_r_stats}--\ref{tab:portuguese_kmax_r_stats}) \textbf{separately on each group}. That is, when we run greedy at a budget $K/2$ solely on each group's graph. When the groups' graph connectivity is high, there is no disparity in social welfare generated for the 2 groups.}
\label{fig:app_knn_thresh_divided}

\begin{picture}(0,0)
    \put(18,370){{\parbox{4cm}{\centering \textbf{Adult}}}}
    \put(178,370){{\parbox{4cm}{\centering \textbf{Math}}}}
    \put(325,370){{\parbox{4cm}{\centering \textbf{Portuguese}}}}
\end{picture}
\end{figure}

In the main chapter and in Figures~\ref{fig:app_knn_thresh_undivided} and \ref{fig:app_knn_thresh_divided}, we compared the average social welfare (gain) (i.e., the total group welfare (gain) divided by the number of agents in the group) achieved by the classic greedy algorithm when applied to the full graph and when applied separately to the male and female bipartite graphs derived from the Adult, Math, and Portuguese datasets. Here, we assess whether the group-prioritized greedy variant described below improves average group welfare and reduces inter-group disparities.

The group-prioritized classic greedy approach proceeds as follows. At each Algorithm~\ref{alg:greedy_lbreveal}  iteration, when multiple targets yield the same total marginal gain in social welfare, ties are resolved by selecting the target that maximizes the marginal gain for the prioritized group. For example, consider three targets with identical total marginal gains of $100$, but varied gains for the groups. Target $t_{1}$ yields $50$ for the male group and $50$ for the female group, $t_{2}$ yields $51$ for the male group and $49$ for the female group, and $t_{3}$ yields $100$ for the male group and $0$ for the female group. If the female group is prioritized, $t_{1}$ is selected, and if the male group is prioritized, $t_{3}$ is selected.

\paragraph{Group-prioritized classic greedy approach experimental results.}
Overall, the group-prioritized greedy variant rarely improves the average group welfare relative to the classic greedy algorithm, except in a few cases highlighted below.

Prioritizing a group can increase its average group welfare at the expense of the other group. 
On the Math $k$NN graph generated with $k_{\max}=2$, when classic greedy is run on the graph with a budget of $K=6$, it achieves an average welfare of $0.6822$  and $0.6705$ for the female and male groups, respectively. 
Prioritizing the male group raises their average welfare to $0.6818$ and lowers that of the female group $(0.6737)$.
Similarly, on the Portuguese $k$NN graph generated with $k_{\max}=2$, when classic greedy is run on the graph with a budget of $K=6$, it attains an average welfare of $0.6207$ and $0.5926$ for the female and male groups, respectively. 
Prioritizing the female group increases their average welfare to $0.6379$ and reduces that of the male group $(0.5787)$ while prioritizing the male group raises their average welfare to $0.5972$ and lowers that of the female group $(0.6164)$.

In other cases, prioritization harms the targeted group while benefiting the other.
On the Portuguese $k$NN graph generated with $k_{\max}=4$, when classic greedy is run on the graph with a budget of $K=6$, it attains an average welfare of $0.8039$ and $0.8171$ for the female and male groups, respectively. 
Prioritizing the male groups reduces their average welfare to $0.8009$ and increases that of the female group $(0.8211)$.
Similarly, on the Math $k$NN graph generated with $k_{\max}=2$, when classic greedy is run on the graph with a budget of $K=6$,  prioritizing the female group lowers their average welfare to $0.6737$ and raises the male group's average welfare to $0.6761$.

These findings show that group prioritization does not consistently improve outcomes over classic greedy and may introduce welfare trade-offs without a clear overall benefit.

\subsection{Empirical Results for Targeted Interventions}
\label{sec:revealrm_app-itmexps}
In this section, we compare pre- and post-reveal intervention gains and examine the effects of varying the intervention and target reveal budgets \((K, B)\).

\begin{figure}[ht!]
\captionsetup[subfigure]{justification=Centering}
\centering
\textbf{Adult Dataset}\\[2ex]
\begin{subfigure}[t]{0.45\textwidth}
\centering
    \includegraphics[width=0.788\linewidth]{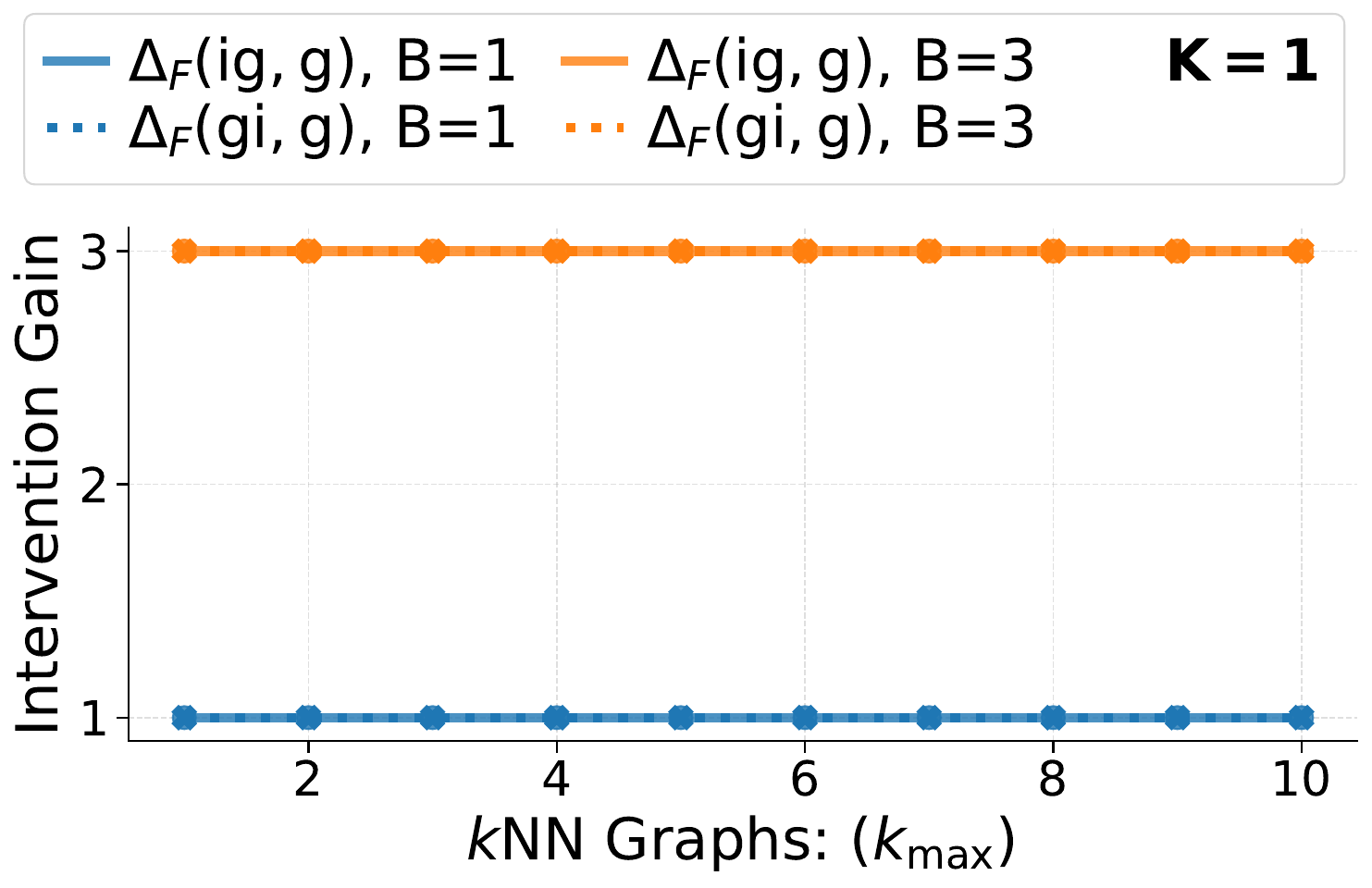}
    \caption{}
    \label{fig:adult_itm_knn1}
\end{subfigure}
\begin{subfigure}[t]{0.45\textwidth}
\centering
    \includegraphics[width=0.788\linewidth]{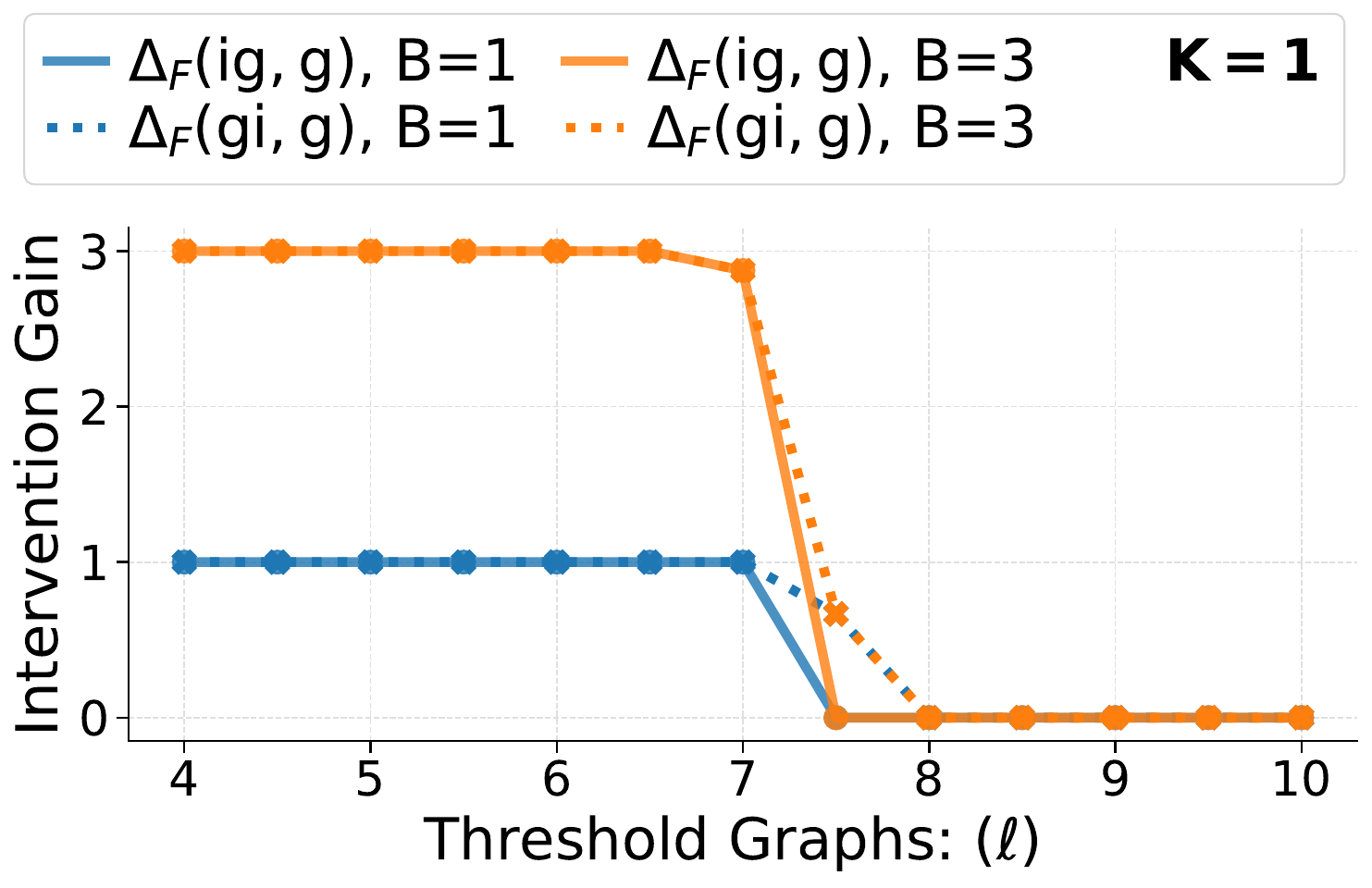}
    \caption{}
    \label{fig:adult_itm_thresh1}
\end{subfigure}\\[1ex]

\begin{subfigure}[t]{0.45\textwidth}
\centering
    \includegraphics[width=0.788\linewidth]{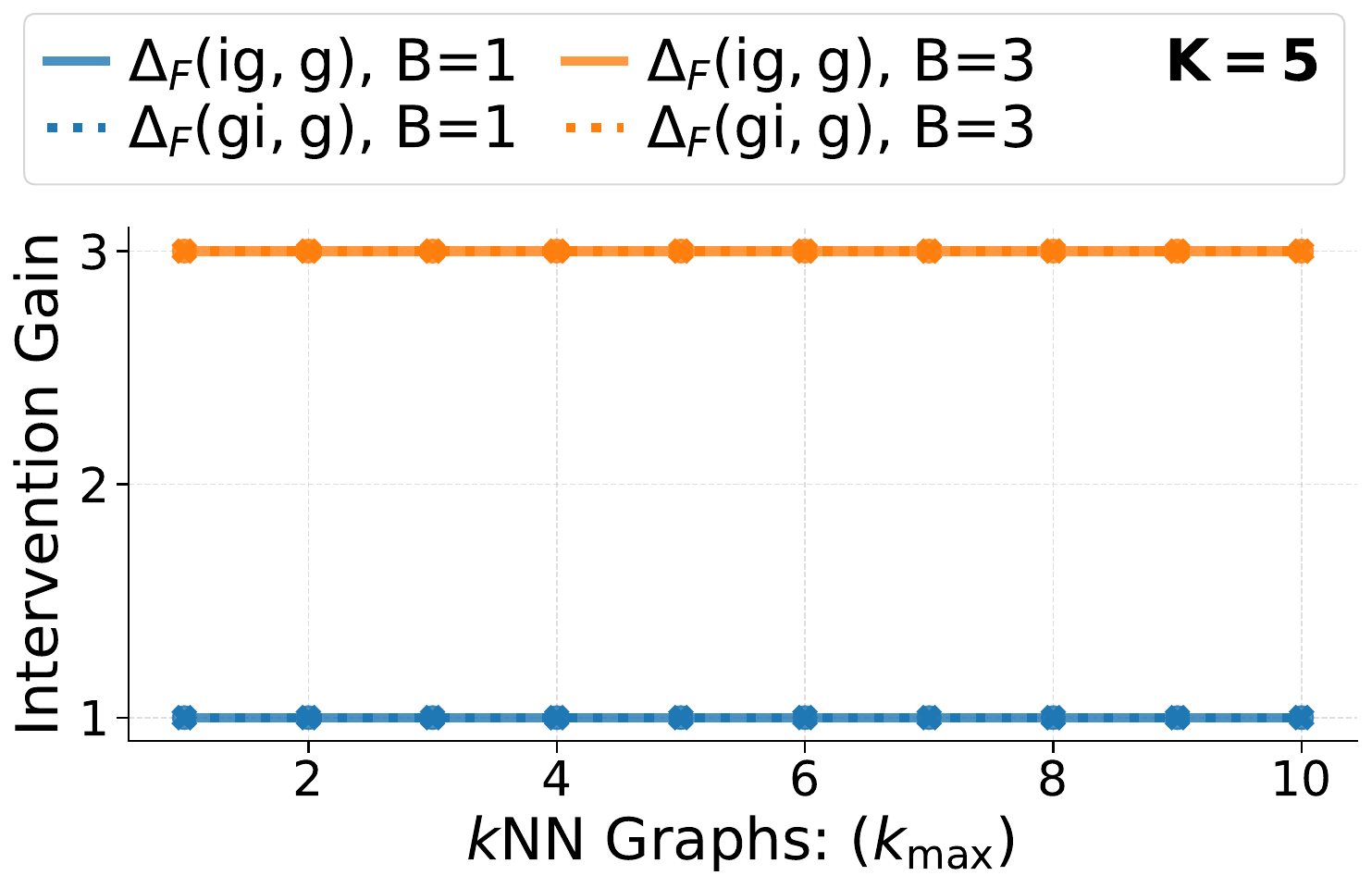}
    \caption{}
    \label{fig:adult_itm_knn5}
\end{subfigure}
\begin{subfigure}[t]{0.45\textwidth}
\centering
    \includegraphics[width=0.788\linewidth]{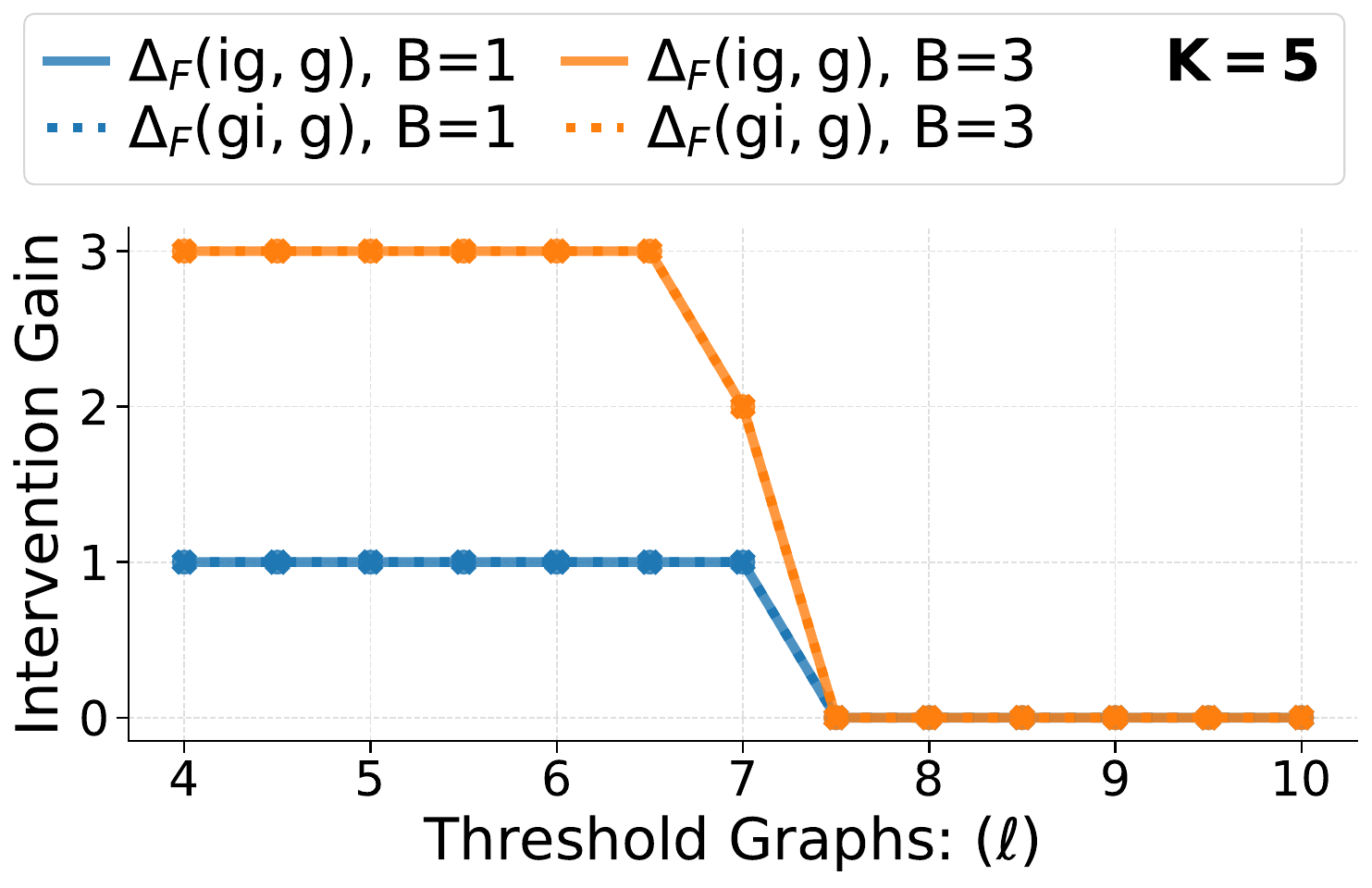}
    \caption{}
    \label{fig:adult_itm_thresh5}
\end{subfigure}\\
\vspace{3ex}
\textbf{Math Dataset}\\[2ex]
\begin{subfigure}[t]{0.45\textwidth}
\centering
    \includegraphics[width=0.788\linewidth]{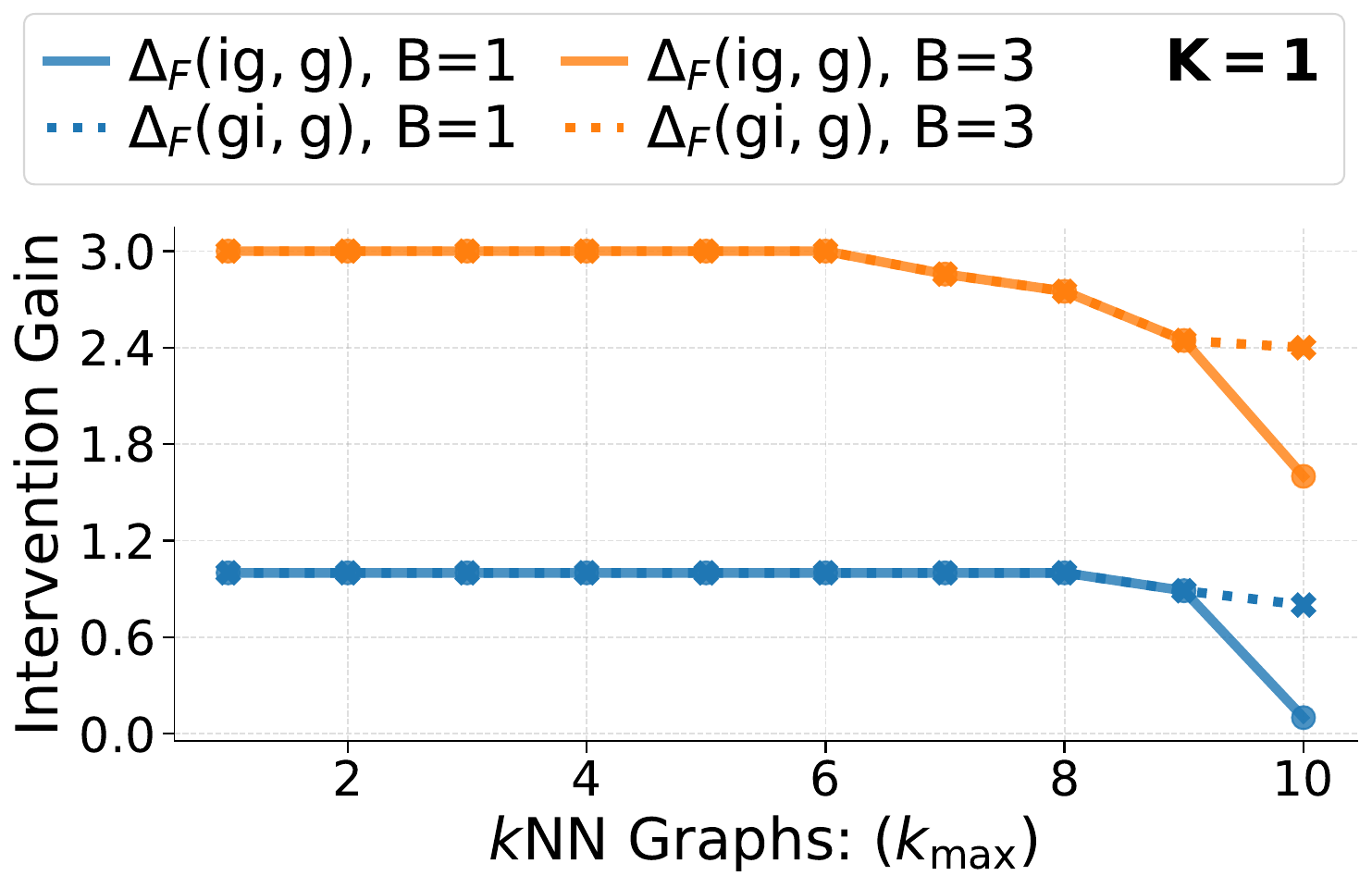}
    \caption{}
    \label{fig:math_itm_knn1}
\end{subfigure}
\begin{subfigure}[t]{0.45\textwidth}
\centering
    \includegraphics[width=0.788\linewidth]{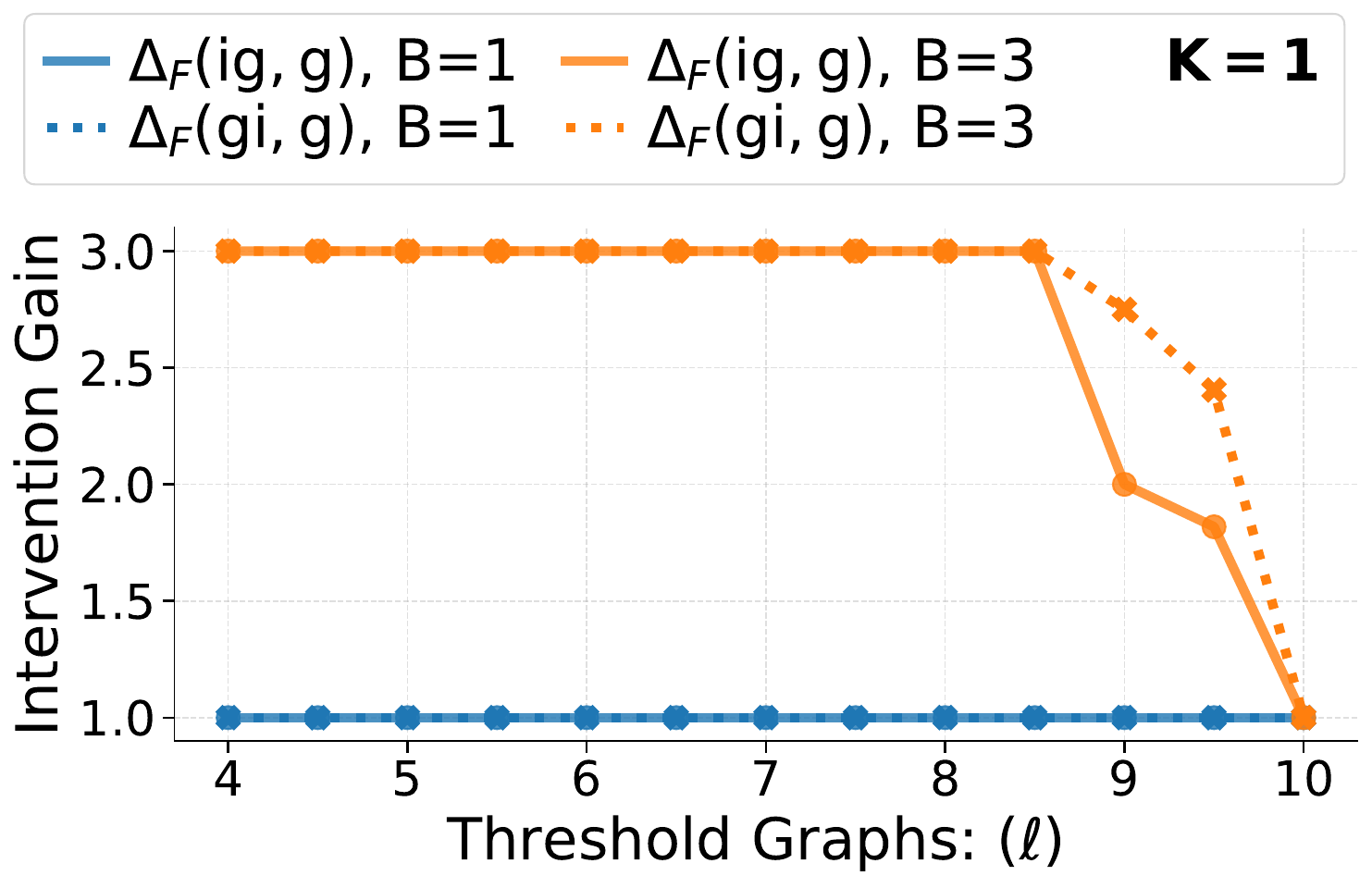}
    \caption{}
    \label{fig:math_itm_thresh1}
\end{subfigure}\\[1ex]

\begin{subfigure}[t]{0.45\textwidth}
\centering
    \includegraphics[width=0.788\linewidth]{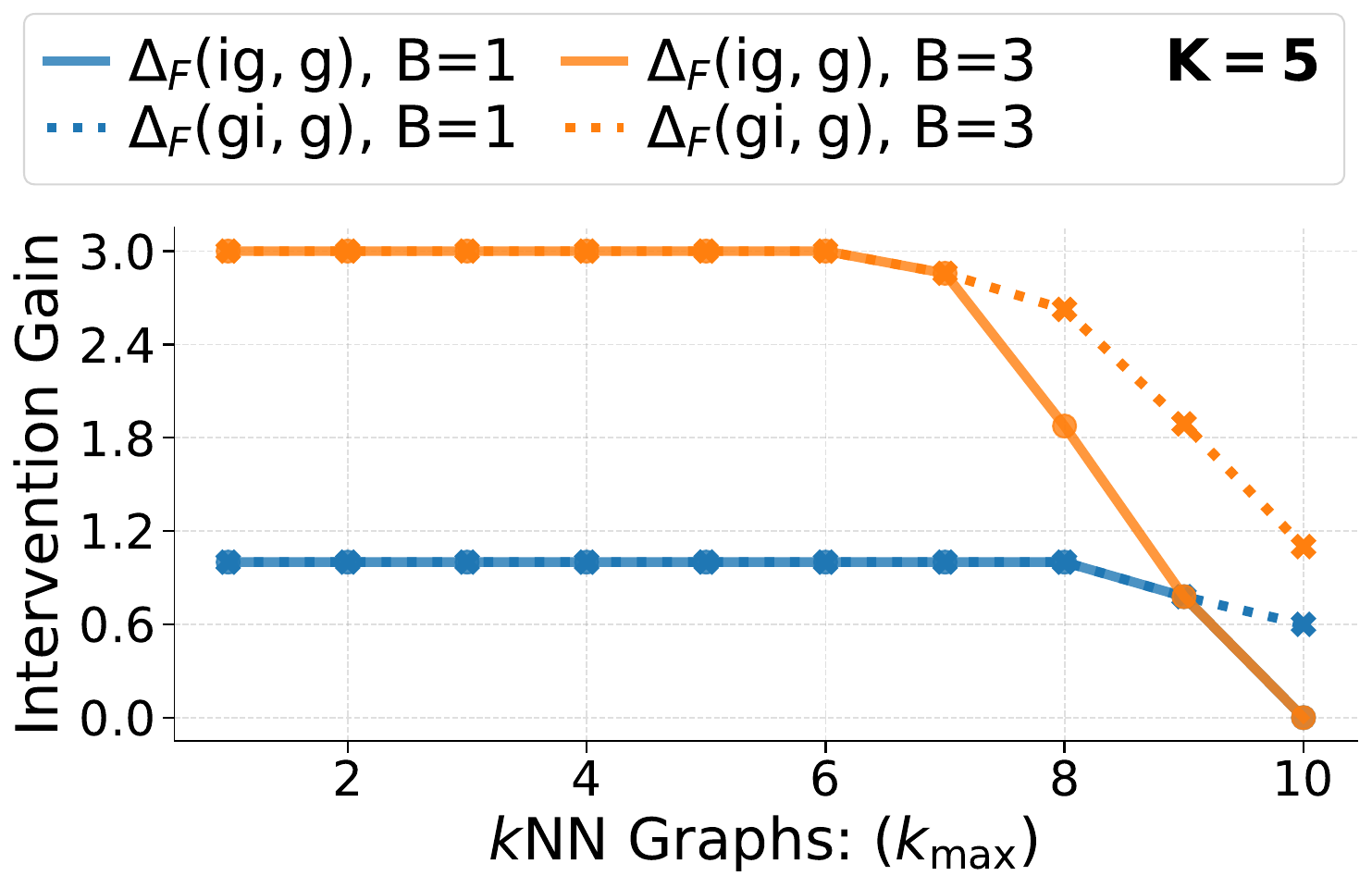}
    \caption{}
    \label{fig:math_itm_knn5}
\end{subfigure}
\begin{subfigure}[t]{0.45\textwidth}
\centering
    \includegraphics[width=0.788\linewidth]{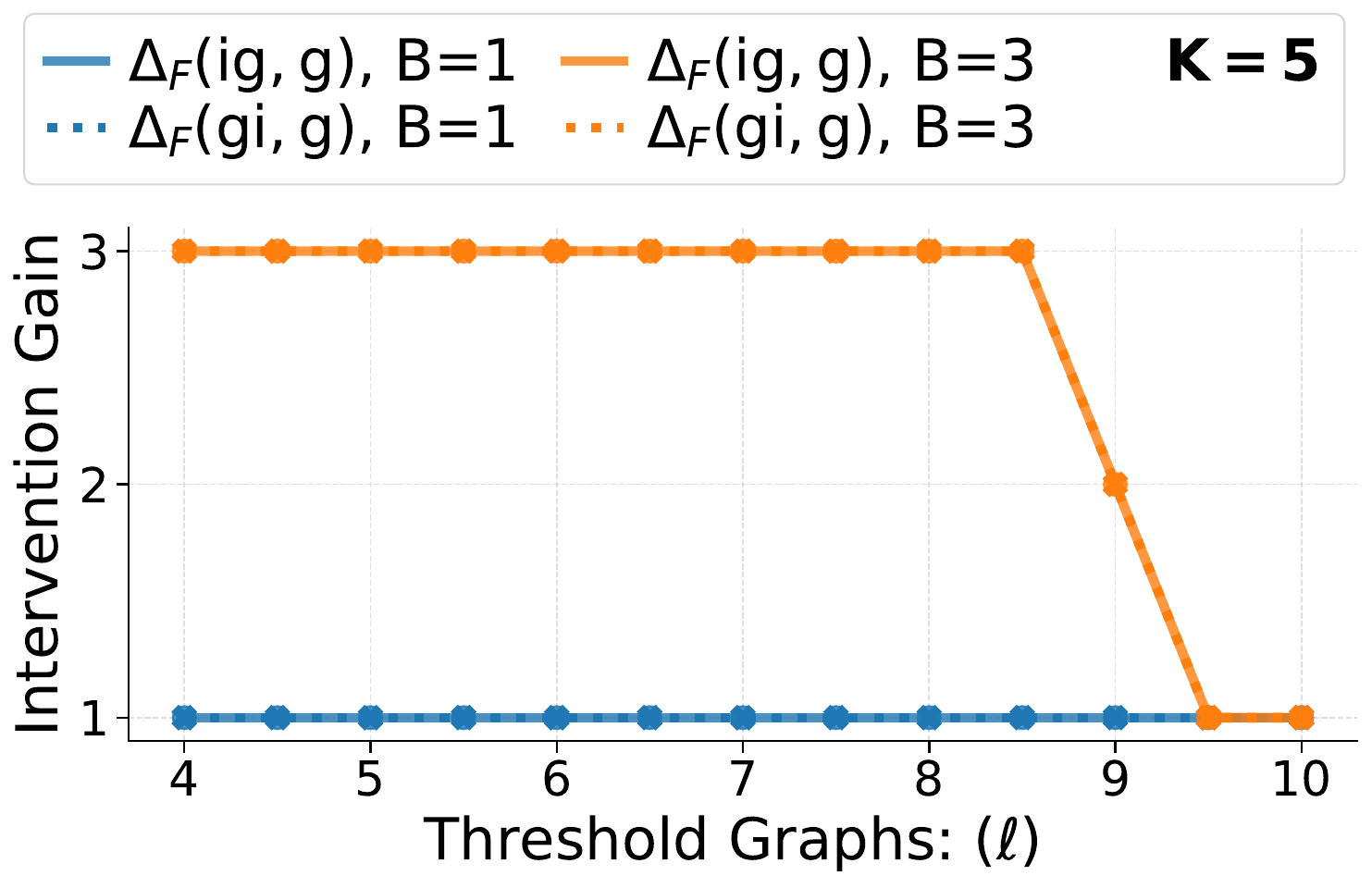}
    \caption{}
    \label{fig:math_itm_thresh5}
\end{subfigure}

\caption[Intervention gains (\(\Delta_F(ig,g)\) and \(\Delta_F(gi,g)\)) across Adult and Math datasets]{Pre- and post-reveal intervention gains (\(\Delta_F(ig,g)\) and \(\Delta_F(gi,g)\)) across datasets, target reveal budgets $K \in \{1,5\}$, intervention budgets $B \in \{1,3\}$, and graph generation methods ($k$NN and threshold) (Tables~\ref{tab:adult_kmax_r_stats} and \ref{tab:math_kmax_r_stats}).  
For both datasets, lower $K$ often yields larger intervention gains. A high $B$ can sometimes be redundant when agents  have all-positive neighborhoods, and/or very few agents have empty or all-negative neighborhoods and greedy is optimal (\subref{fig:adult_itm_thresh1}, \subref{fig:adult_itm_thresh5}, 
\subref{fig:math_itm_knn1}, \subref{fig:math_itm_knn5}). When the number of agents with all-negative neighborhoods is at least \(B\), both pre- and post-reveal interventions gains \(\Delta_{F}(ig,g)\) and \(\Delta_{F}(gi,g)\) are atmost \(B\) (\subref{fig:adult_itm_knn1}, \subref{fig:adult_itm_knn5}).}
\label{fig:adult_math_itm_knn_thresh}
\end{figure}

\begin{figure}[ht!]
\captionsetup[subfigure]{justification=Centering}
\centering
\textbf{Portuguese Dataset}\\[2ex]
\begin{subfigure}[t]{0.45\textwidth}
\centering
    \includegraphics[width=0.788\linewidth]{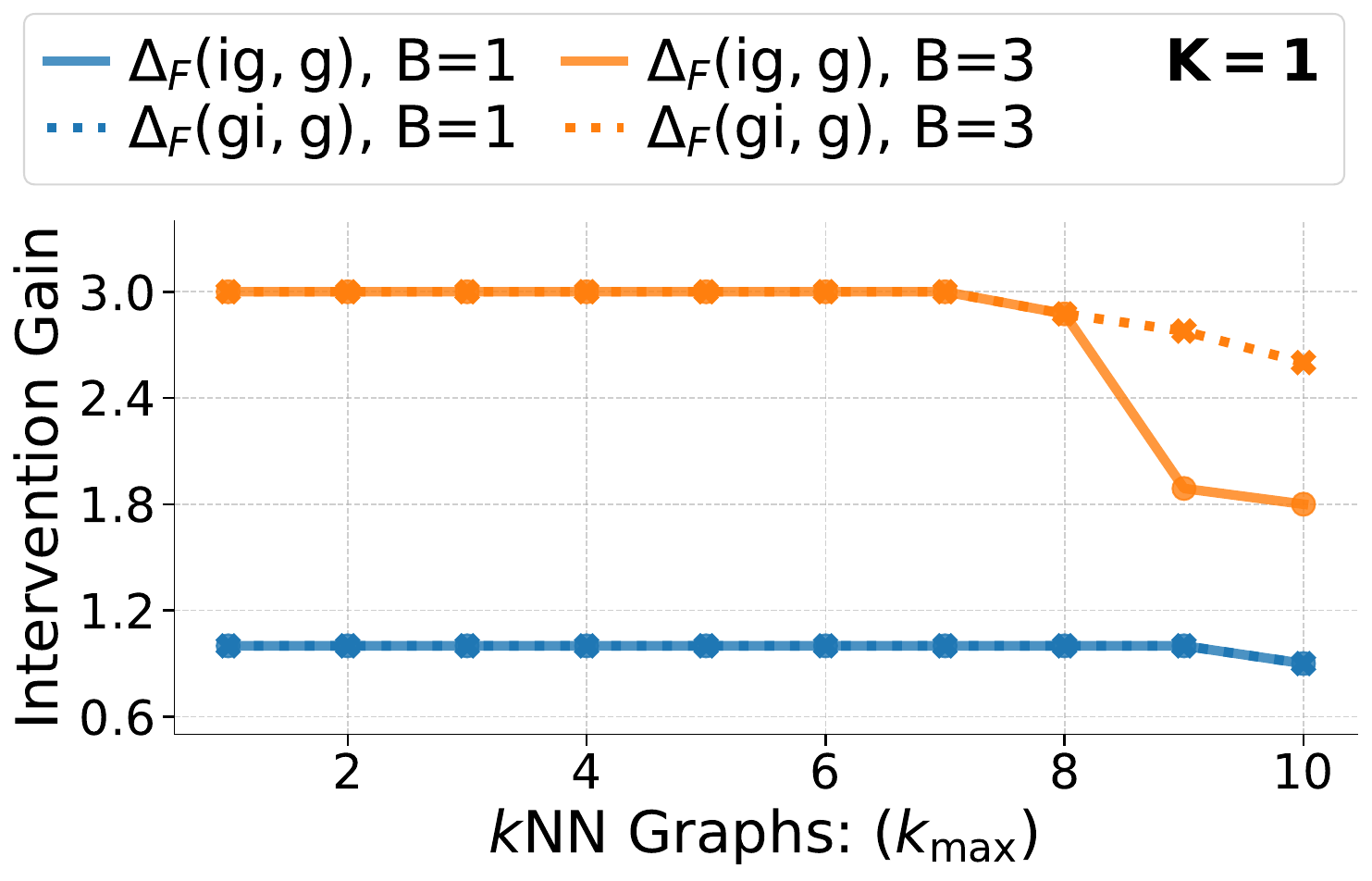}
    \caption{}
    \label{fig:portuguese_itm_knn1}
\end{subfigure}
\begin{subfigure}[t]{0.45\textwidth}
\centering
    \includegraphics[width=0.788\linewidth]{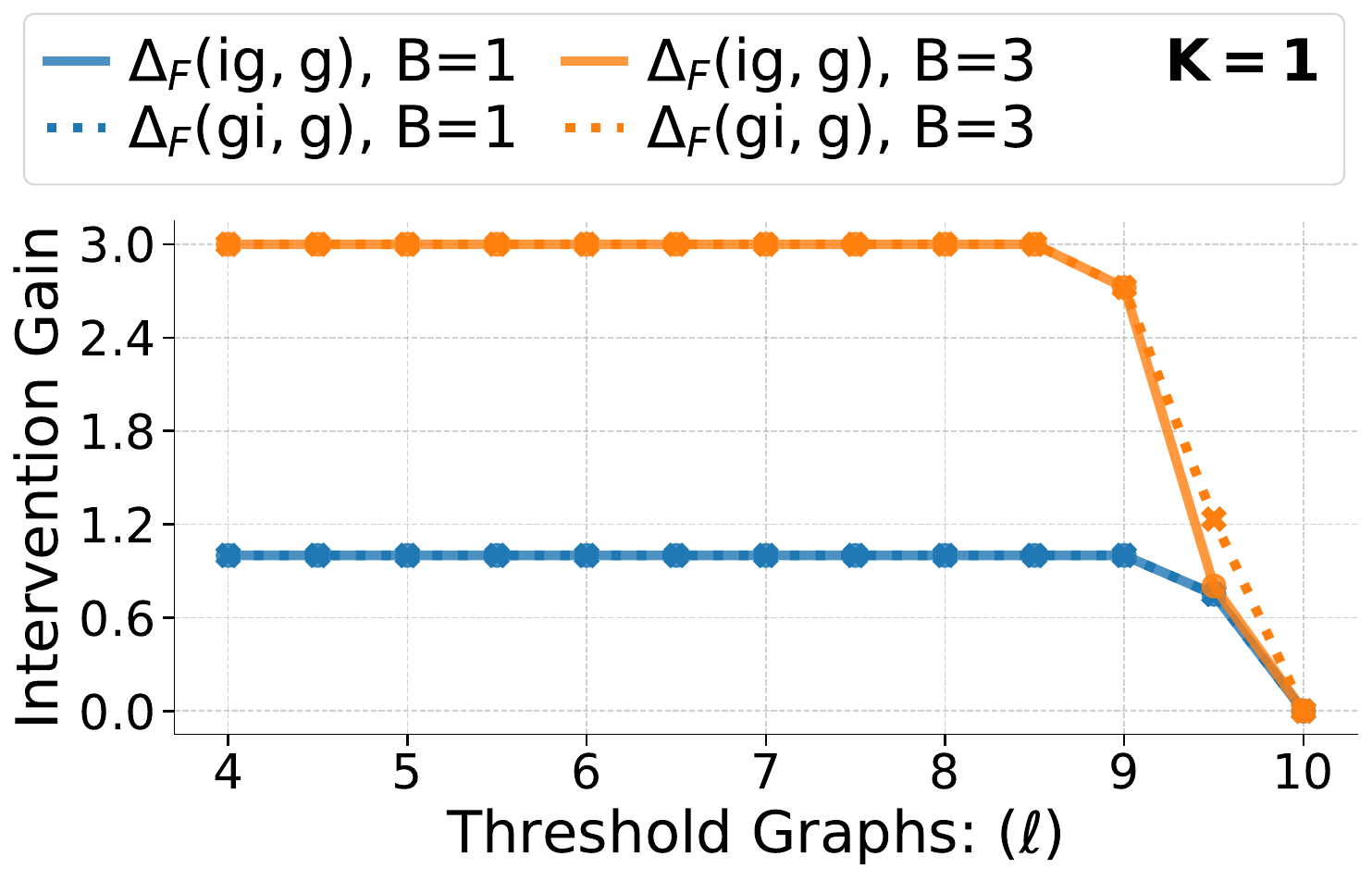}
    \caption{}
    \label{fig:portuguese_itm_thresh1}
\end{subfigure}\\[1ex]

\begin{subfigure}[t]{0.45\textwidth}
\centering
    \includegraphics[width=0.788\linewidth]{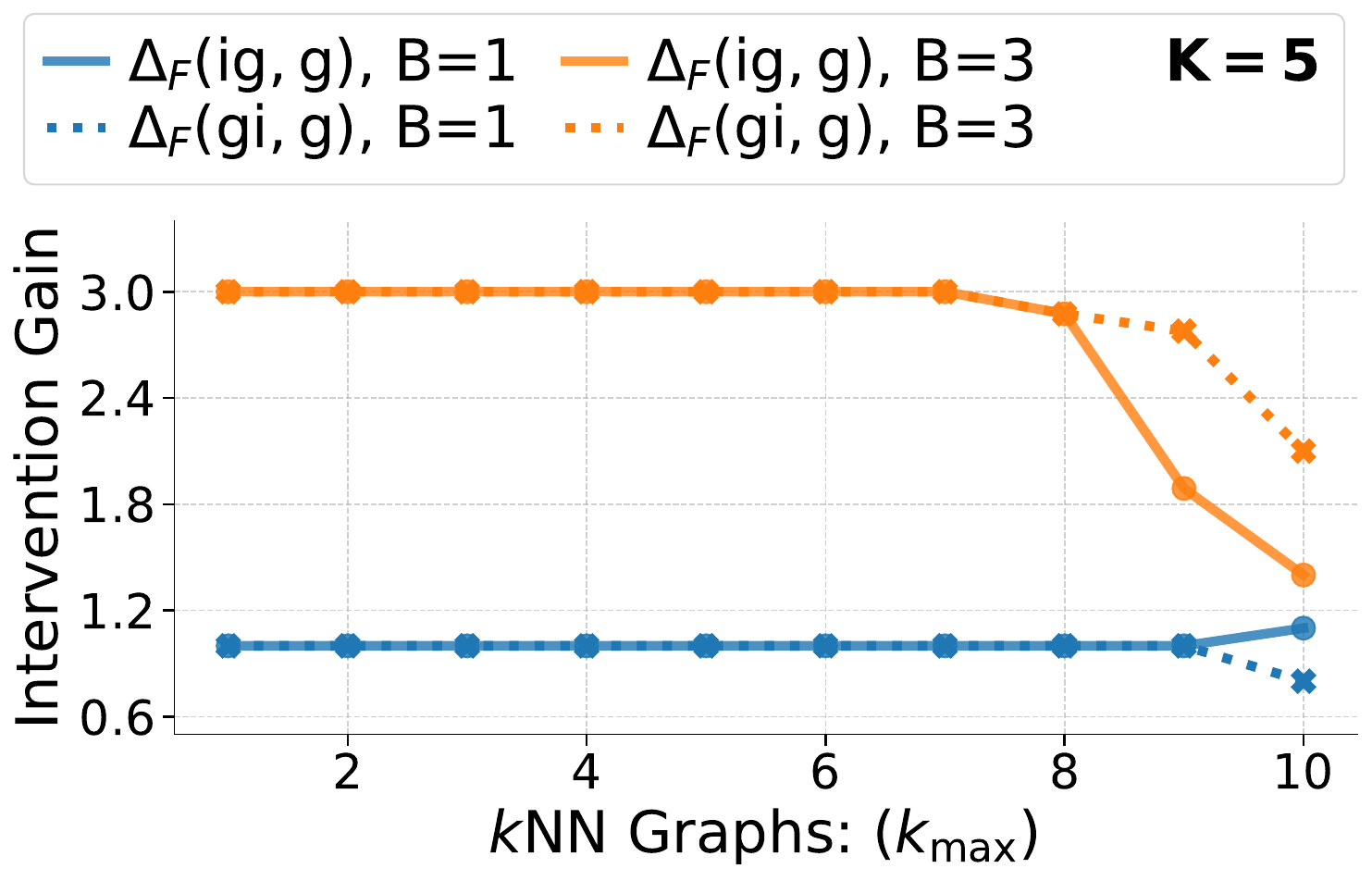}
    \caption{}
    \label{fig:portuguese_itm_knn5}
\end{subfigure}
\begin{subfigure}[t]{0.45\textwidth}
\centering
    \includegraphics[width=0.788\linewidth]{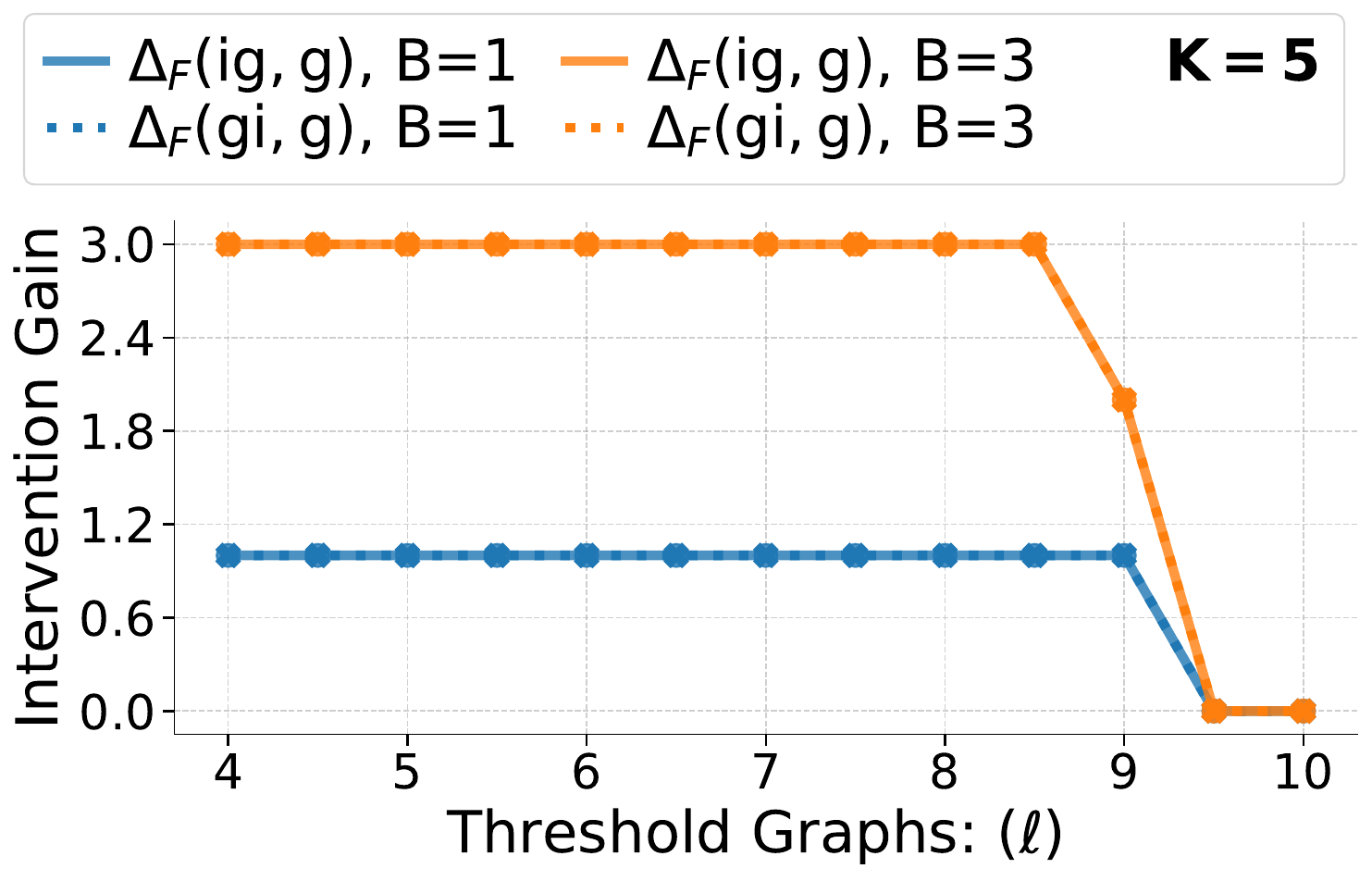}
    \caption{}
    \label{fig:portuguese_itm_thresh5}
\end{subfigure}
\vspace{4ex}
\textbf{Productivity Dataset}\\[2ex]
\begin{subfigure}[t]{0.45\textwidth}
\centering
    \includegraphics[width=0.788\linewidth]{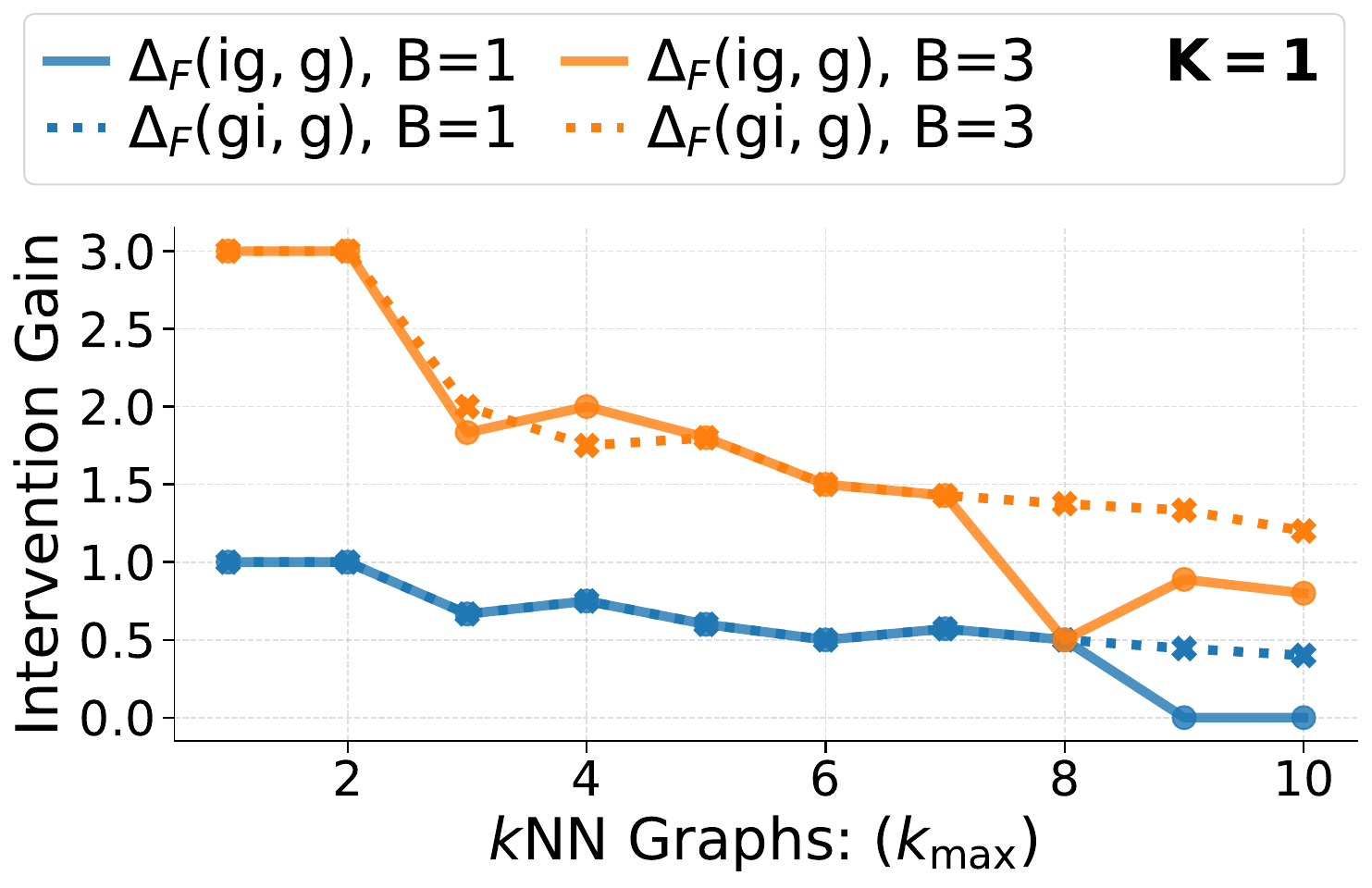}
    \caption{}
    \label{fig:app-prod_itm_knn1}
\end{subfigure}
\begin{subfigure}[t]{0.45\textwidth}
\centering
    \includegraphics[width=0.788\linewidth]{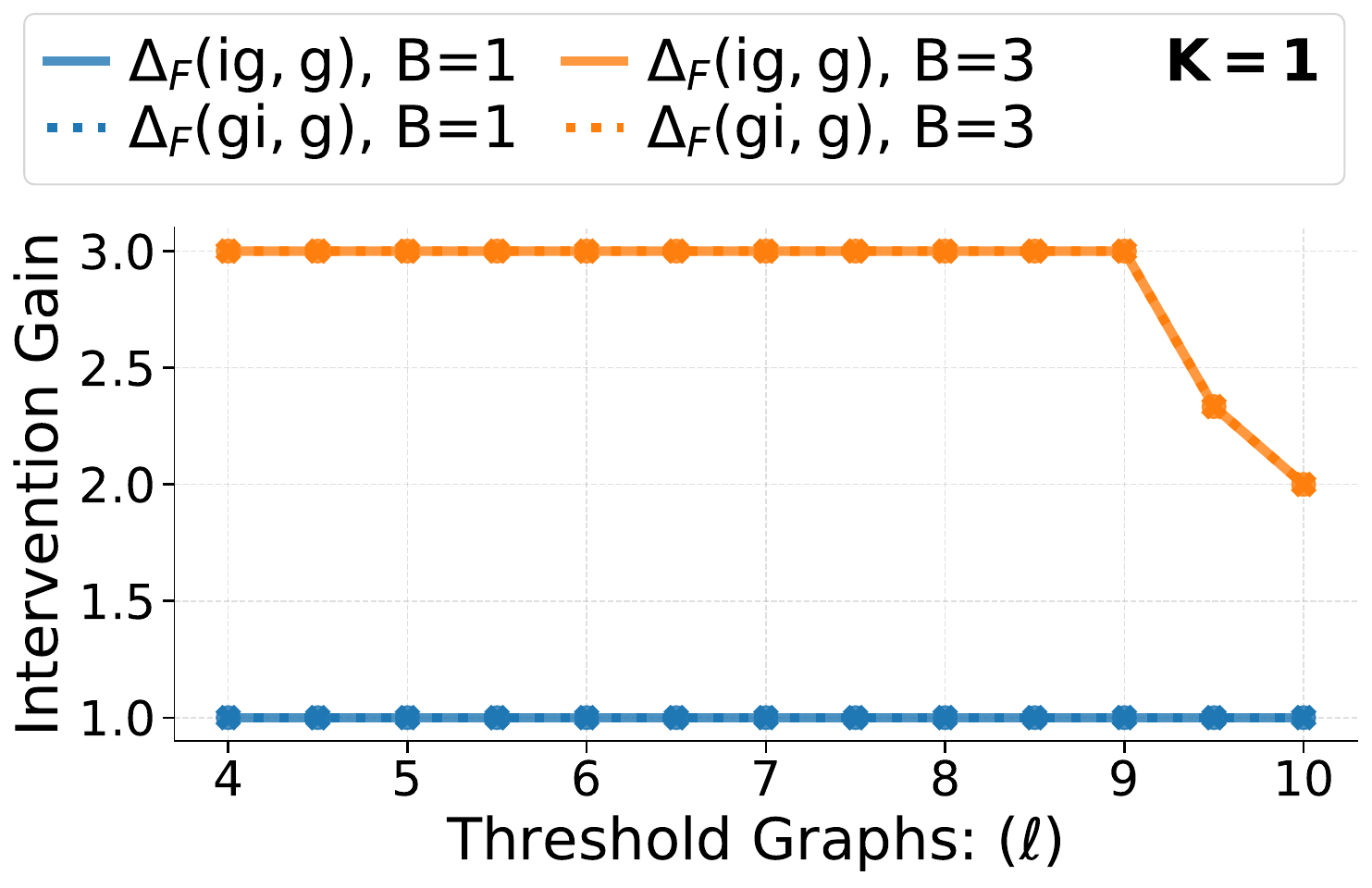}
    \caption{}
    \label{fig:app-prod_itm_thresh1}
\end{subfigure}\\[1ex]

\begin{subfigure}[t]{0.45\textwidth}
\centering
    \includegraphics[width=0.788\linewidth]{reveal_rolemodels/revealrm_figures/productivity_imb4aftersm_results_knn5.pdf}
    \caption{}
    \label{fig:app-prod_itm_knn5}
\end{subfigure}
\begin{subfigure}[t]{0.45\textwidth}
\centering
    \includegraphics[width=0.788\linewidth]{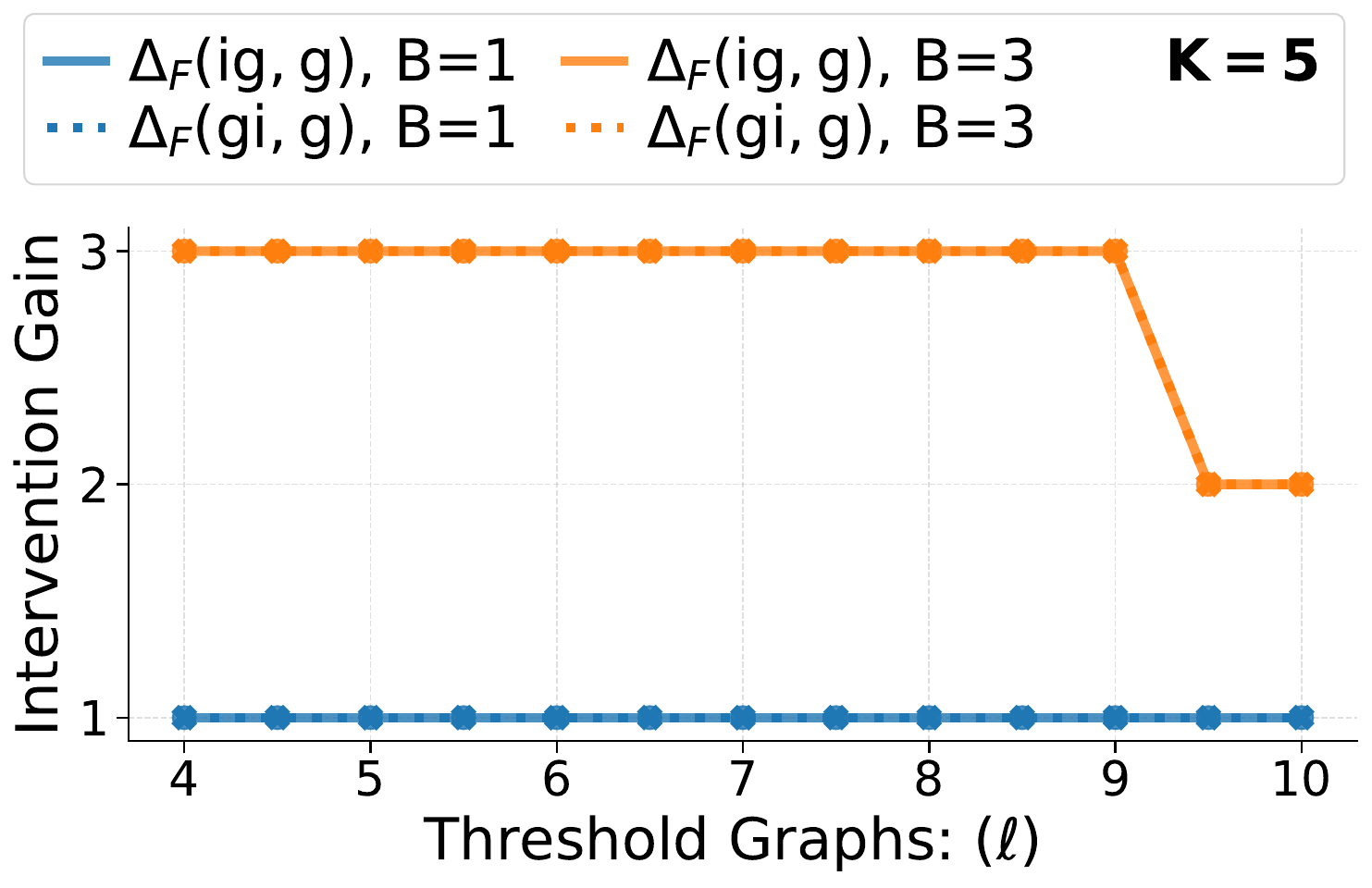}
    \caption{}
    \label{fig:app-prod_itm_thresh5}
\end{subfigure}
\caption[Intervention gains (\(\Delta_F(ig,g)\) and \(\Delta_F(gi,g)\)) across Portuguese and Productivity]{Pre- and post-reveal intervention gains (\(\Delta_F(ig,g)\) and \(\Delta_F(gi,g)\)) across datasets, target reveal budgets $K \in \{1,5\}$, intervention budgets $B \in \{1,3\}$, and graph generation methods ($k$NN and threshold) (Tables~\ref{tab:portuguese_kmax_r_stats} and \ref{tab:productivity_kmax_r_stats}).  
For both datasets, lower $K$ often yields larger gains. 
A high $B$ can be redundant when many agents have all-positive neighborhoods (\subref{fig:app-prod_itm_knn1}, \subref{fig:portuguese_itm_thresh1}, \subref{fig:portuguese_itm_thresh5}). Pre-reveal intervention can yield negative gains (Figure~\subref{fig:app-prod_itm_knn5}, $k_{\max}\!=\!\{9,10\}$) but may also sometimes outperform post-reveal intervention 
(Figure~\ref{fig:portuguese_itm_knn5}, $k_{\max}\!=\!10$).}
\label{fig:port_prod_itm_knn_thresh}
\end{figure}

\paragraph{General observations.}
Intervention gains are upper bounded by the intervention budget \(B\) (Figures~\ref{fig:adult_math_itm_knn_thresh} and \ref{fig:port_prod_itm_knn_thresh}).

Smaller target reveal budgets \(K\) tend to produce higher gains, as most agents may still have low probabilities of emulating a positive target, even after welfare-maximizing subset of targets is revealed. In Figures~\ref{fig:adult_math_itm_knn_thresh} and \ref{fig:port_prod_itm_knn_thresh}, intervening when greedy was run with a budget of \(K = 1\) (subfigures (\textbf{a,b,e,f})) yields gains at least as large as the those when run with \(K = 5\) (subfigures (\textbf{c,d,g,h})).

Intervention gains are highest when multiple agents have empty or all-negative neighborhoods. For instance, in Adult \(k\)NN-generated graphs, the number of agents with all-negative neighborhoods exceeds the intervention budget, that is, for each graph \(i \in [10], \text{only-Ns}^{(i)} > B = 4\) (Table~\ref{tab:adult_kmax_r_stats}). Consequently, the pre- and post-reveal intervention gains are capped by \(B\) (Figures~\ref{fig:adult_itm_knn1},\subref{fig:adult_itm_knn5}).

Intervention may be redundant or underutilized when many agents have all-positive neighborhoods, very few have all-negative or empty neighborhoods and the label reveal algorithm achieves an optimal solution. For instance, in Productivity threshold generated graphs, greedy is optimal (Table~\ref{tab:productivity_kmax_r_stats}; Figure~\ref{fig:app_prod_star_sg_thresh}), and intervention applies only to a shrinking set of agents with no neighbors (Figures~\ref{fig:app-prod_itm_thresh1},\subref{fig:app-prod_itm_thresh5}). A similar pattern is observed in Adult, Math, and Portuguese threshold-generated graphs 
(Tables~\ref{tab:adult_kmax_r_stats}--\ref{tab:portuguese_kmax_r_stats};
Figures~\ref{fig:app_adult_sr_sg_thresh}--\subref{fig:app_port_sr_sg_thresh}), where the intervention budget is underutilized/redundant (Figures~\ref{fig:adult_math_itm_knn_thresh} and \ref{fig:port_prod_itm_knn_thresh}, subfigures (\textbf{b,d,f,h})).
In addition, intervention gains are generally small when high-risk agents already have a high probability of emulating a positive target (e.g., in Figures~\ref{fig:app-prod_itm_knn1},\subref{fig:app-prod_itm_knn5}).

\paragraph{Comparison of pre- and post-reveal interventions.}
Post-reveal interventions consistently produce positive intervention gains, which are most times at least as large as those from pre-reveal intervention (Figures~\ref{fig:adult_math_itm_knn_thresh} and \ref{fig:port_prod_itm_knn_thresh}). 

Pre-reveal intervention can occasionally lead to negative intervention gains (Figure~\ref{fig:app-prod_itm_knn5}). If the label reveal algorithm (Algorithm~\ref{alg:greedy_lbreveal}) is effective and high-risk agents already have high probabilities of emulating positive targets, removing them before executing  Algorithm~\ref{alg:greedy_lbreveal} might distort the graph and lead to the algorithm selecting a target set with lower social welfare than if those agents had remained, resulting in a negative intervention gain \(\Delta_{F}(ig,g)\).

\subsection{Empirical Results under the Coverage Radius Model}
\label{sec:revealrm_app-cmexps}
\begin{figure}[ht!]
    \centering
    \includegraphics[width=0.66\textwidth]{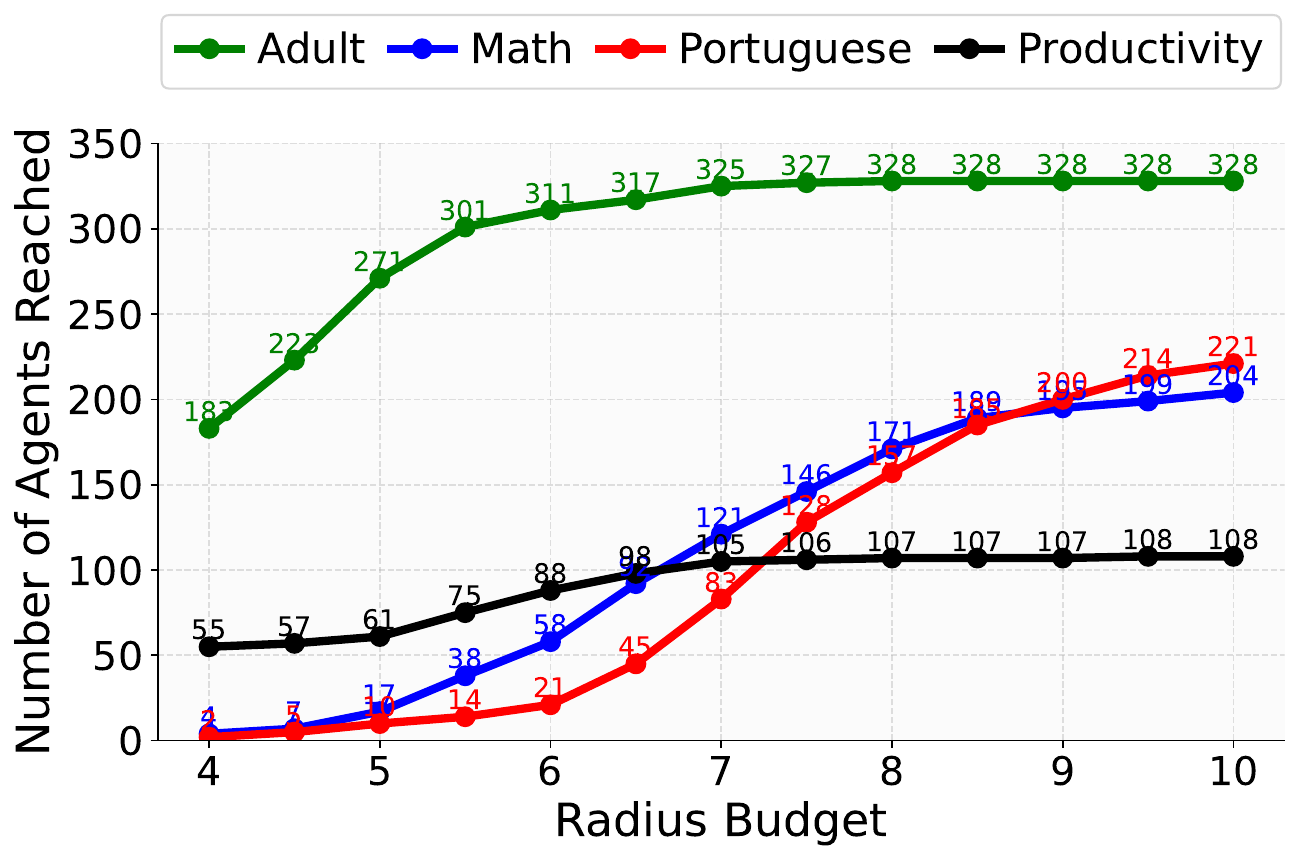}
    \caption[Performance of the greedy coverage radius algorithm]{Performance of the greedy coverage radius algorithm (Algorithm~\ref{alg:radius_greedy_coverage}) on the Adult, Math, Portuguese, and Productivity datasets. Number of agents reached increases with the radius budget.}
    \label{fig:alldatasets_radius}
\end{figure}

For all geometric graphs generated with zero initial target radius \(r_i = 0\) for all targets \(i\), increasing the radius budget increases the number of agents reached by positive targets (Figure~\ref{fig:alldatasets_radius}). When many agents have several positive targets within a comparable radius, an additional radius provides little to no gain (e.g., on Adult dataset, at \(R\geq 8\)). When distances to the nearest positive target vary substantially, larger radius budgets lead to broader coverage (e.g., on Math, Portuguese, and Productivity datasets). 
When agents are densely grouped at roughly the same large distance from positive targets, minor increases in radius produce little change while more substantial expansions broaden coverage, as seen in the Productivity dataset.

\subsection{Empirical Results under the Learning Setting}
\label{sec:revealrm_app-lsexps}
We analyze the empirical results for the learning setting. Training and testing scores are averaged over \(100\) independent train-test splits, each constructed with a different random seed. Performance is evaluated using three metrics, 
\(\mathrm{Perf}_{1}, \mathrm{Perf}_{2}, \mathrm{Perf}_{3} \in [0,100]\), defined in Appendix~\ref{subsec:revealrm_app-perfeval}.

\begin{figure}[b!]
\captionsetup[subfigure]{justification=Centering}
\centering
\textbf{Adult Dataset}\\[1.5ex]
\begin{subfigure}[t]{0.32\textwidth}
\centering
    \includegraphics[width=\textwidth]{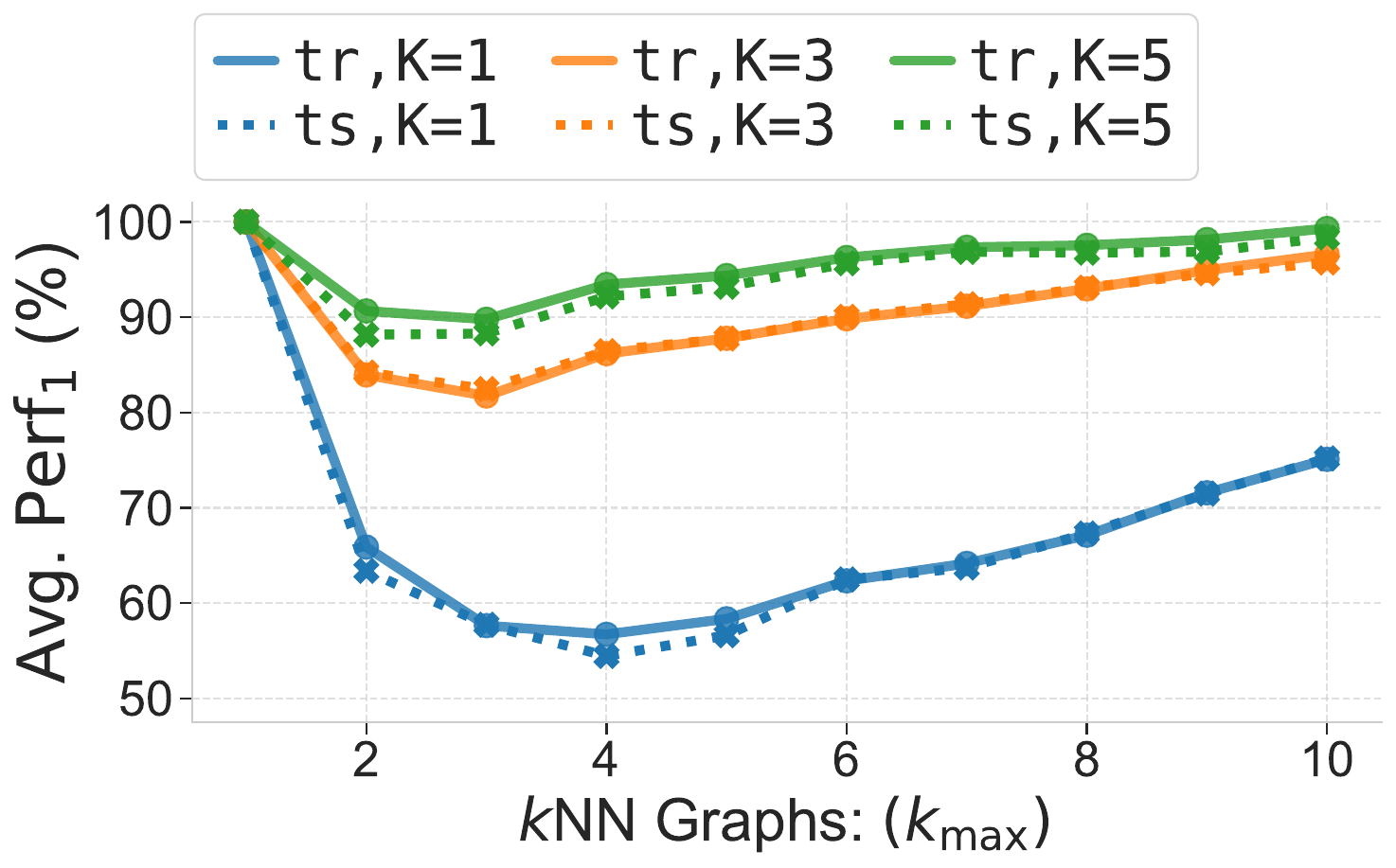}
    \caption{\(k\)NN: \(\mathrm{Perf}_{1}\)}
    \label{fig:app_adult_learn_knn_perf1}
\end{subfigure}
\begin{subfigure}[t]{0.32\textwidth}
\centering
    \includegraphics[width=\linewidth]{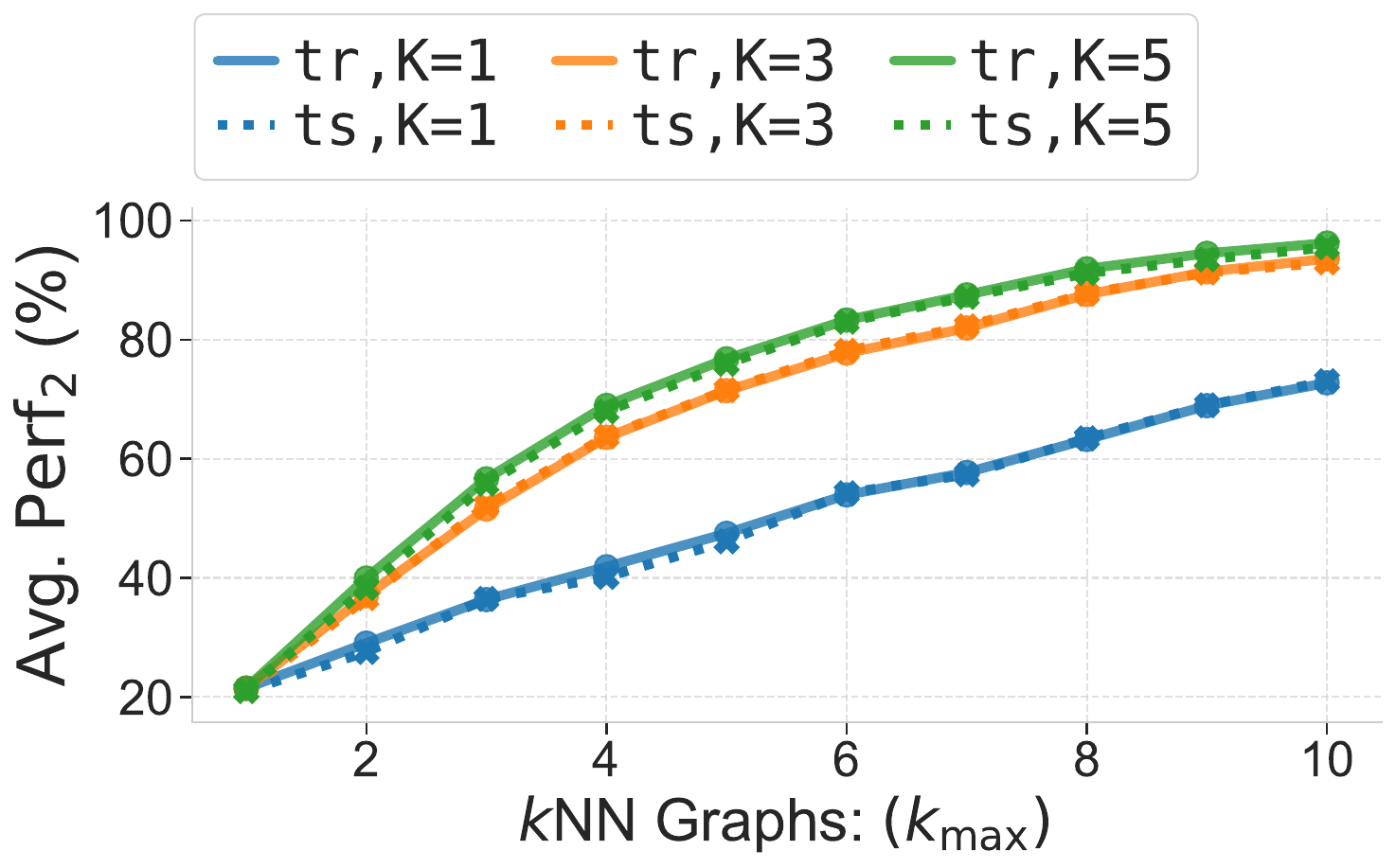}
    \caption{\(k\)NN: \(\mathrm{Perf}_{2}\)}
    \label{fig:app_adult_learn_knn_perf2}
\end{subfigure}
\begin{subfigure}[t]{0.32\textwidth}
\centering
    \includegraphics[width=\linewidth]{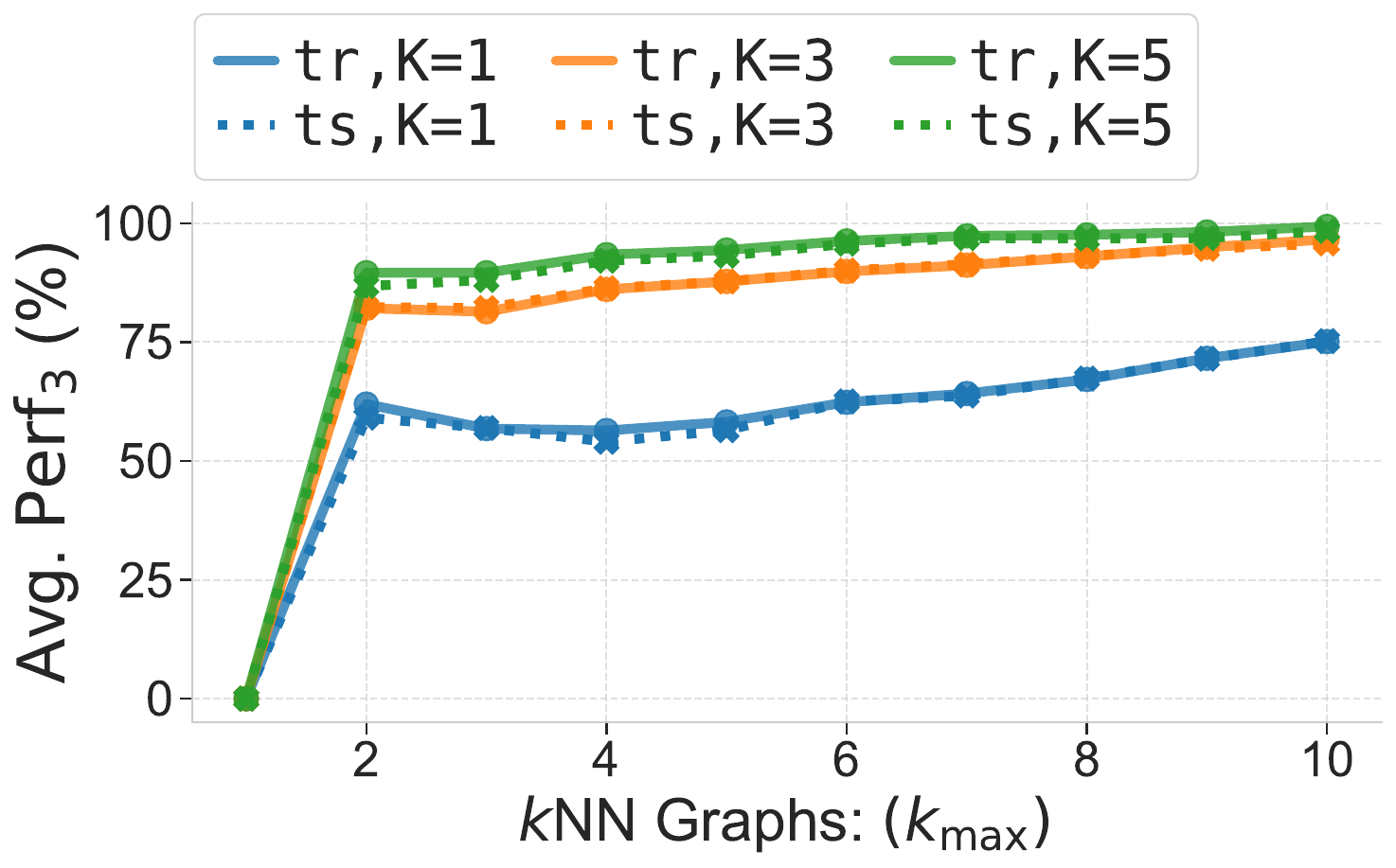}
    \caption{\(k\)NN: \(\mathrm{Perf}_{3}\)}
    \label{fig:app_adult_learn_knn_perf3}
\end{subfigure}\\[1ex]

\begin{subfigure}[t]{0.32\textwidth}
\centering
    \includegraphics[width=\textwidth]{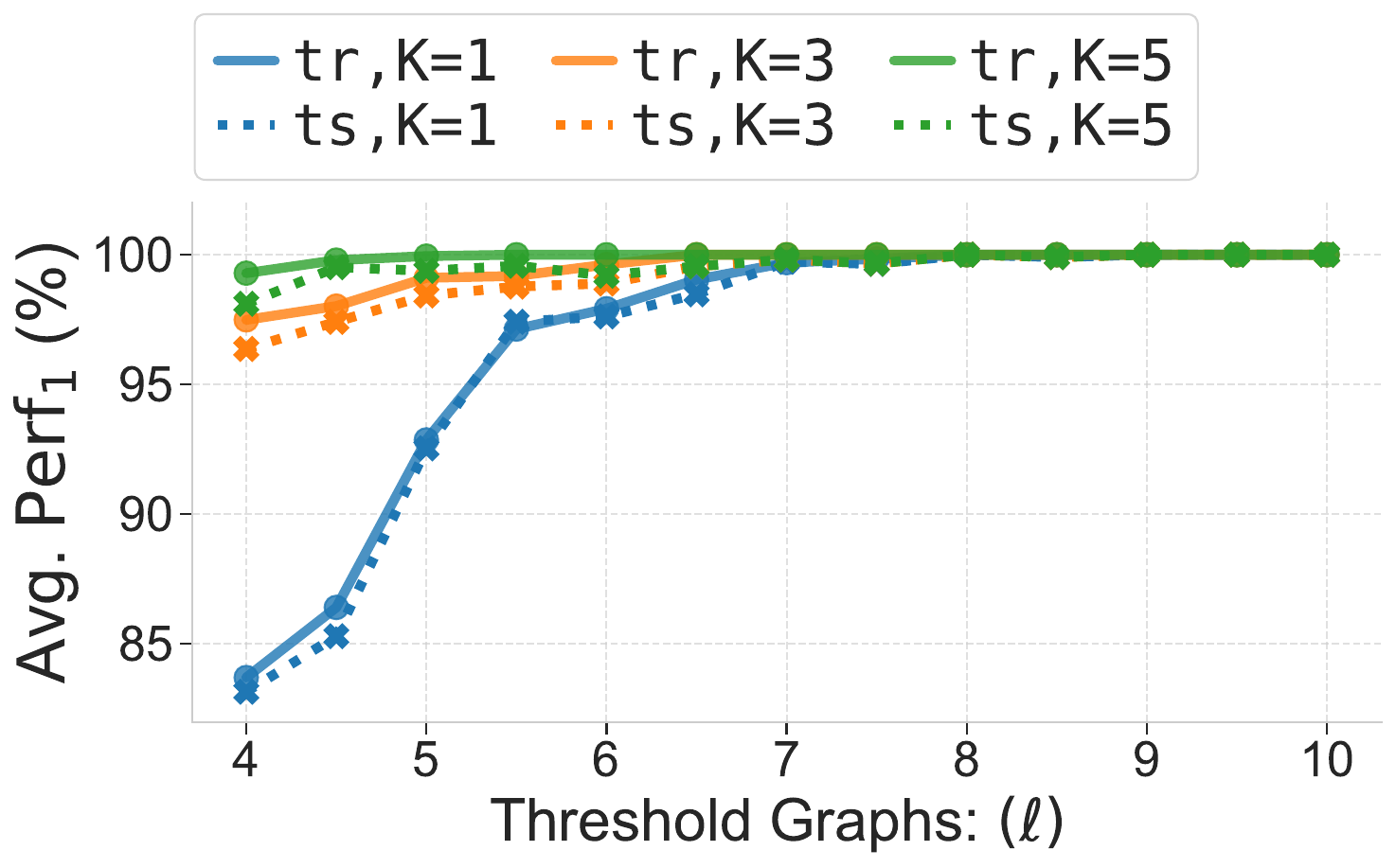}
    \caption{Threshold: \(\mathrm{Perf}_{1}\)}
    \label{fig:app_adult_learn_thresh_perf1}
\end{subfigure}
\begin{subfigure}[t]{0.32\textwidth}
\centering
    \includegraphics[width=\linewidth]{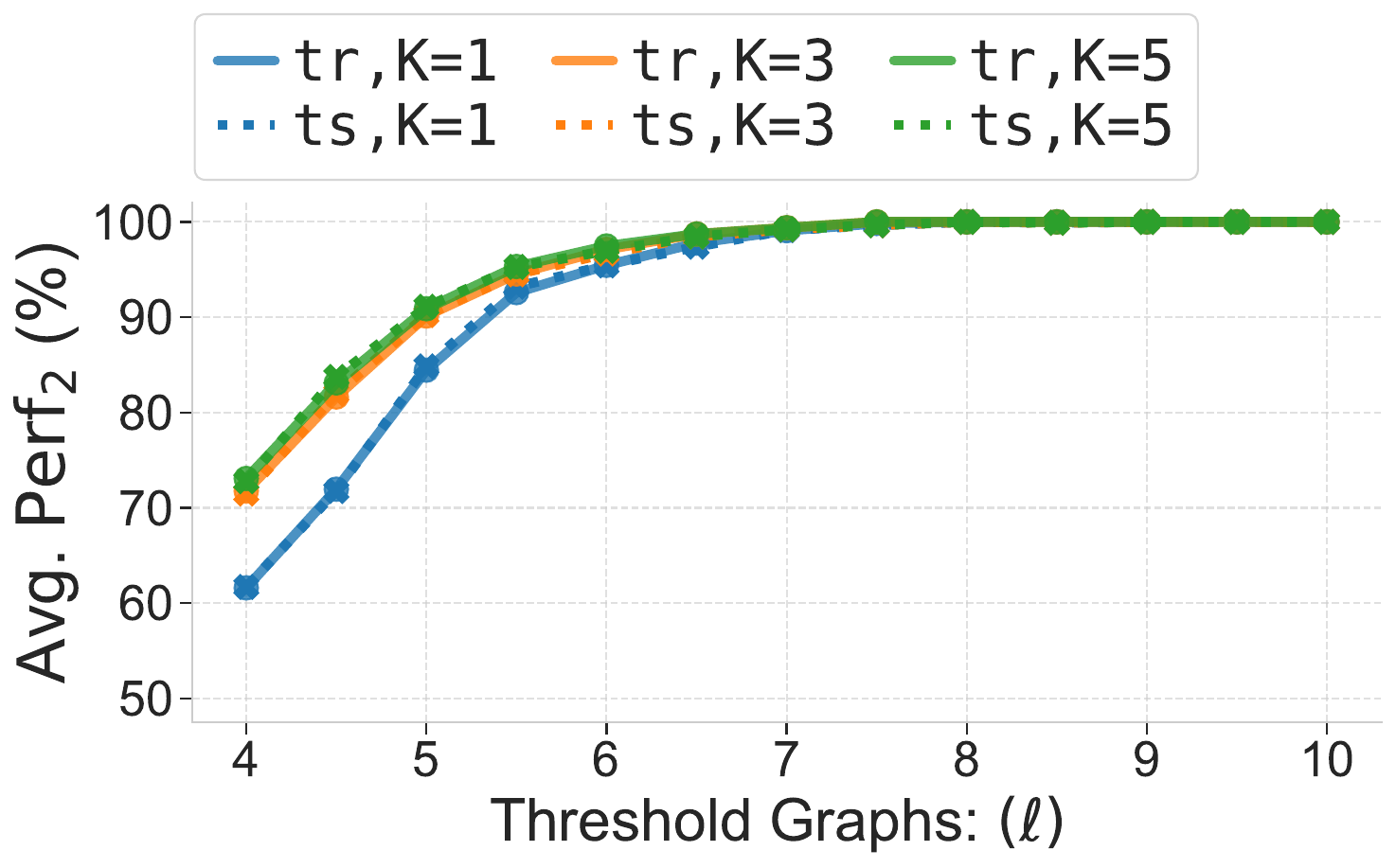}
    \caption{Threshold: \(\mathrm{Perf}_{2}\)}
    \label{fig:app_adult_learn_thresh_perf2}
\end{subfigure}
\begin{subfigure}[t]{0.32\textwidth}
\centering
    \includegraphics[width=\linewidth]{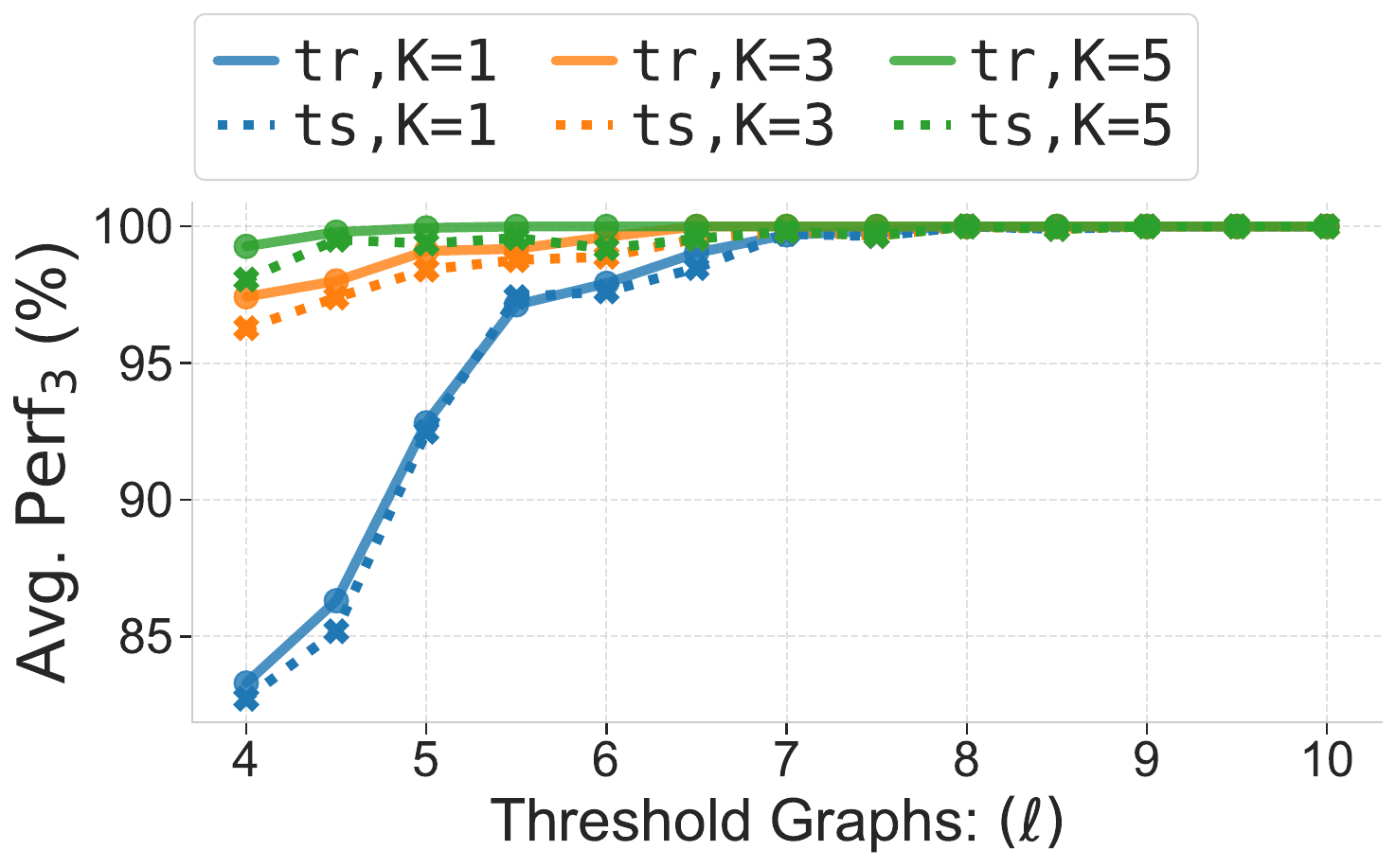}
    \caption{Threshold: \(\mathrm{Perf}_{3}\)}
    \label{fig:app_adult_learn_thresh_perf3}
\end{subfigure}
\\[2ex]
\textbf{Math Dataset}
\\[2ex]
\begin{subfigure}[t]{0.32\textwidth}
\centering
    \includegraphics[width=\textwidth]{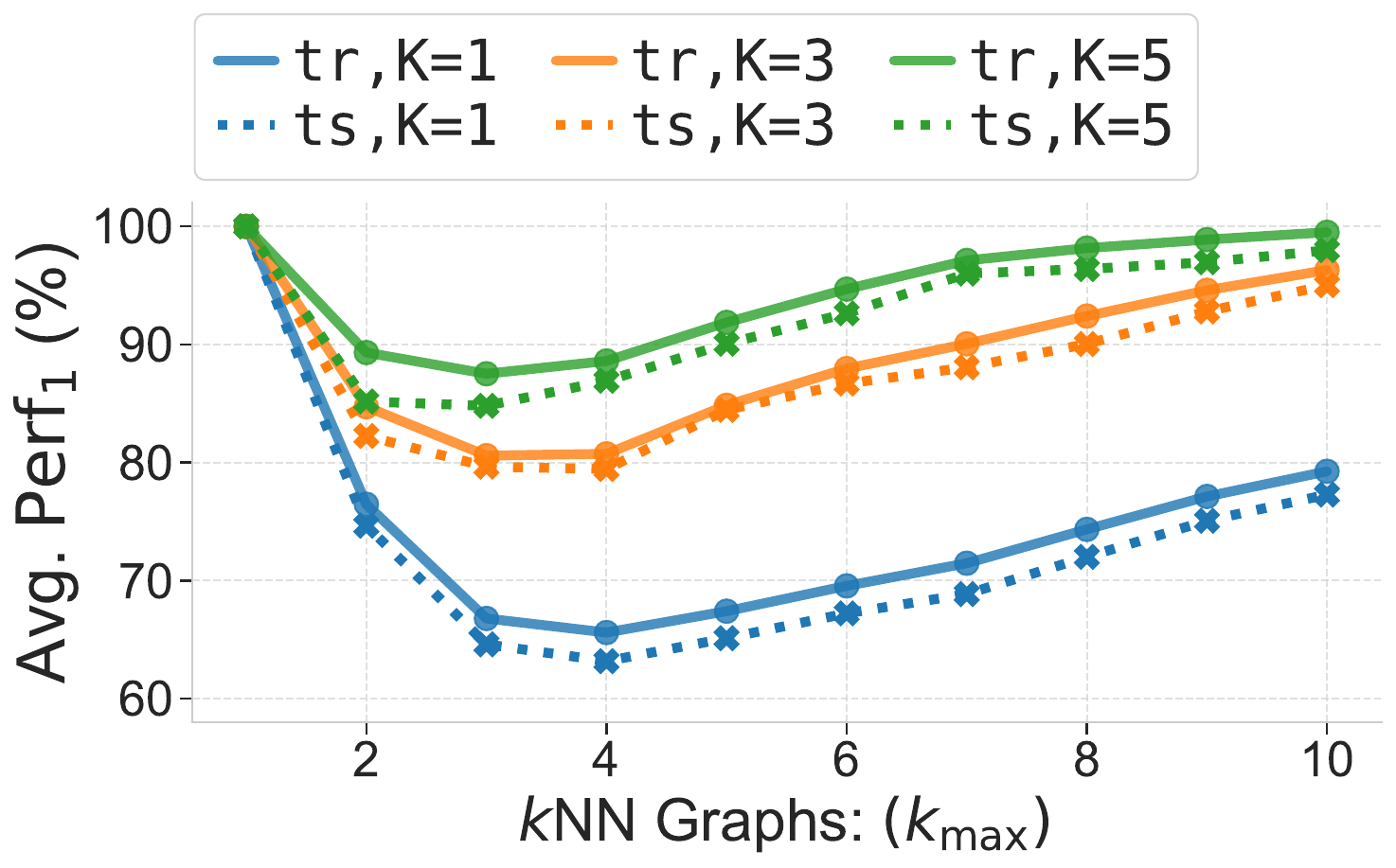}
    \caption{\(k\)NN: \(\mathrm{Perf}_{1}\)}
    \label{fig:app_math_learn_knn_perf1}
\end{subfigure}
\begin{subfigure}[t]{0.32\textwidth}
\centering
    \includegraphics[width=\linewidth]{reveal_rolemodels/revealrm_figures/studentmath_learn_both_results_knn_perf2.pdf}
    \caption{\(k\)NN: \(\mathrm{Perf}_{2}\)}
    \label{fig:app_math_learn_knn_perf2}
\end{subfigure}
\begin{subfigure}[t]{0.32\textwidth}
\centering
    \includegraphics[width=\linewidth]{reveal_rolemodels/revealrm_figures/studentmath_learn_both_results_knn_perf3.pdf}
    \caption{\(k\)NN: \(\mathrm{Perf}_{3}\)}
    \label{fig:app_math_learn_knn_perf3}
\end{subfigure}\\[1ex]

\begin{subfigure}[t]{0.32\textwidth}
\centering
    \includegraphics[width=\textwidth]{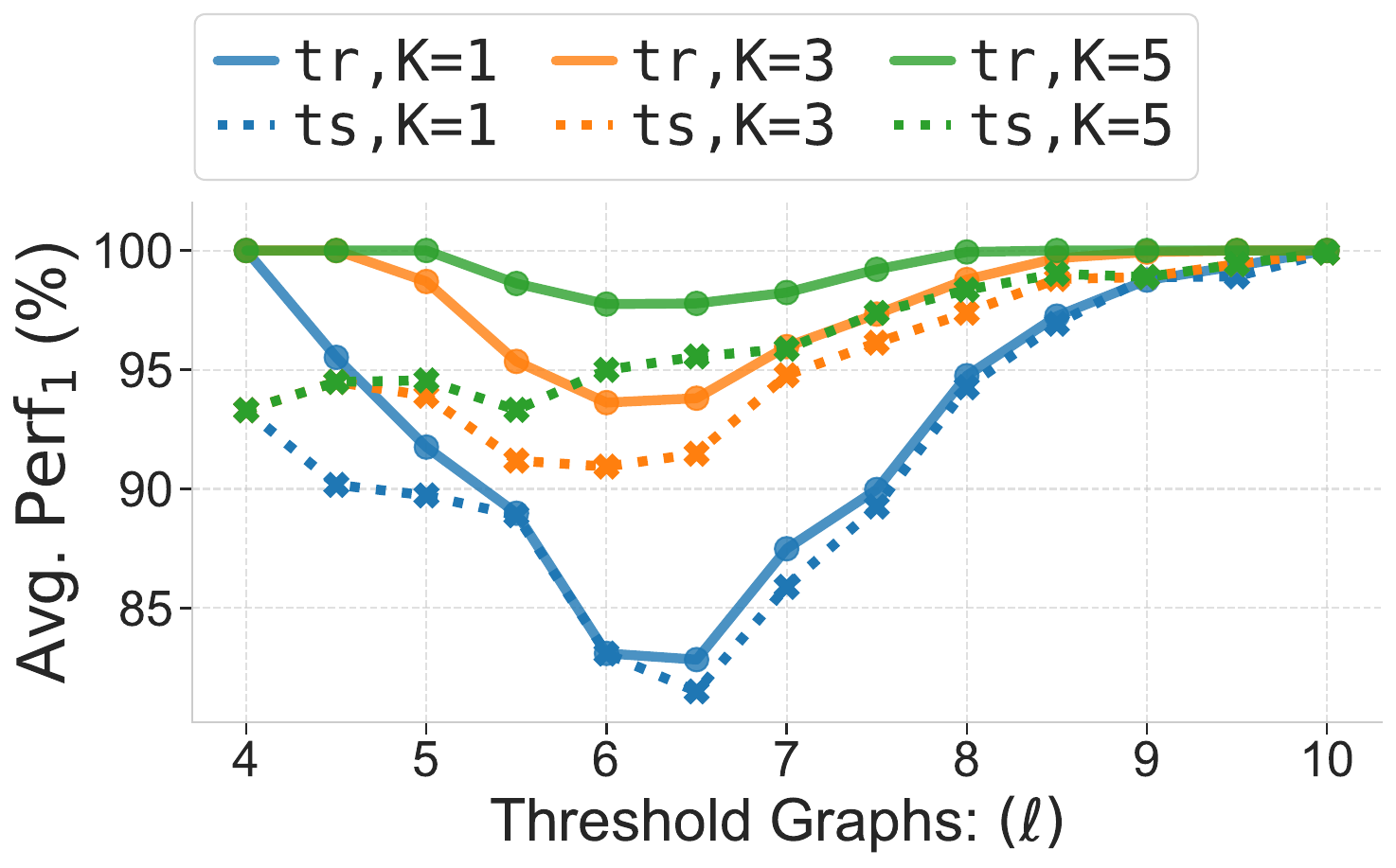}
    \caption{Threshold: \(\mathrm{Perf}_{1}\)}
    \label{fig:app_math_learn_thresh_perf1}
\end{subfigure}
\begin{subfigure}[t]{0.32\textwidth}
\centering
    \includegraphics[width=\linewidth]{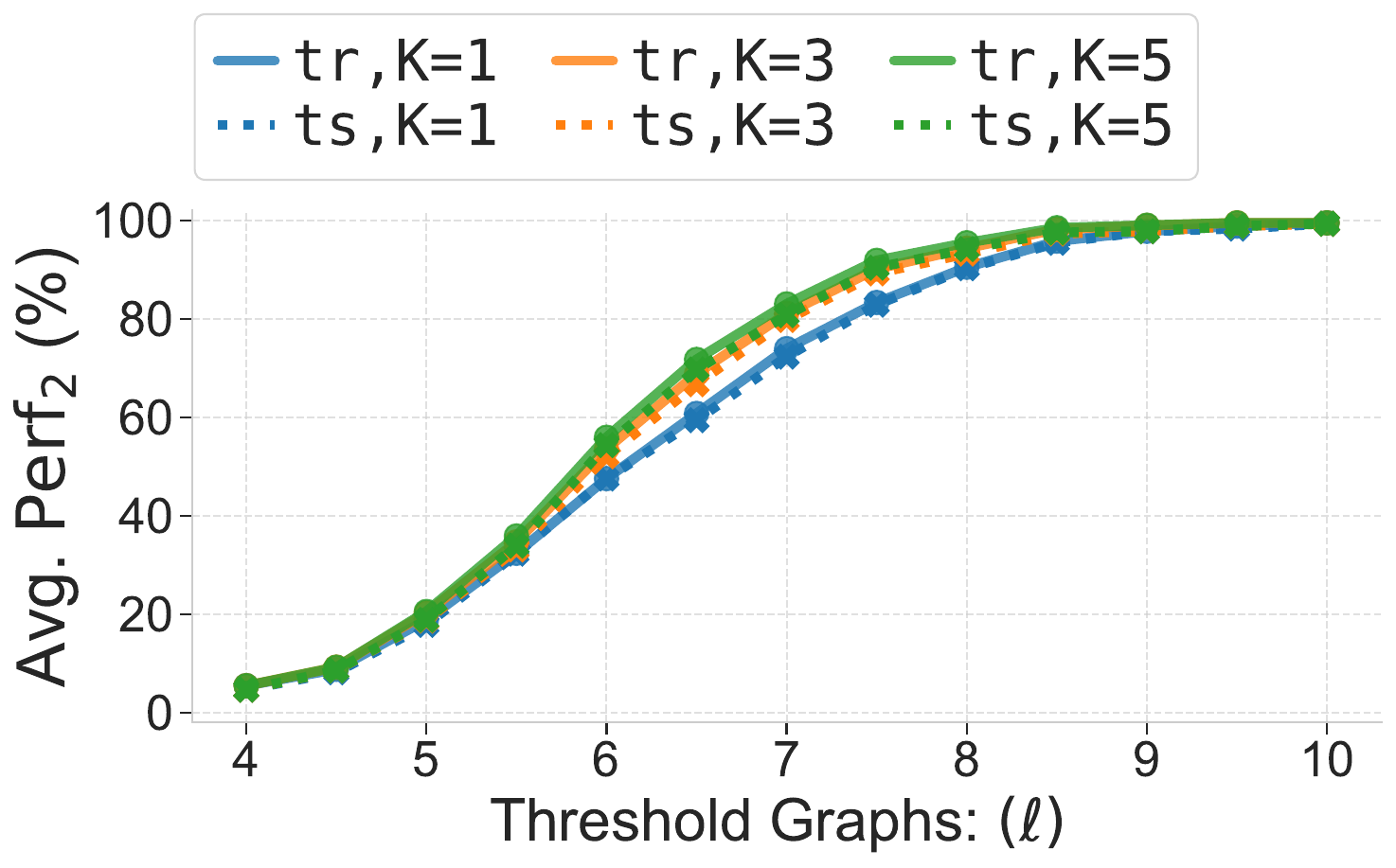}
    \caption{Threshold: \(\mathrm{Perf}_{2}\)}
    \label{fig:app_math_learn_thresh_perf2}
\end{subfigure}
\begin{subfigure}[t]{0.32\textwidth}
\centering
    \includegraphics[width=\linewidth]{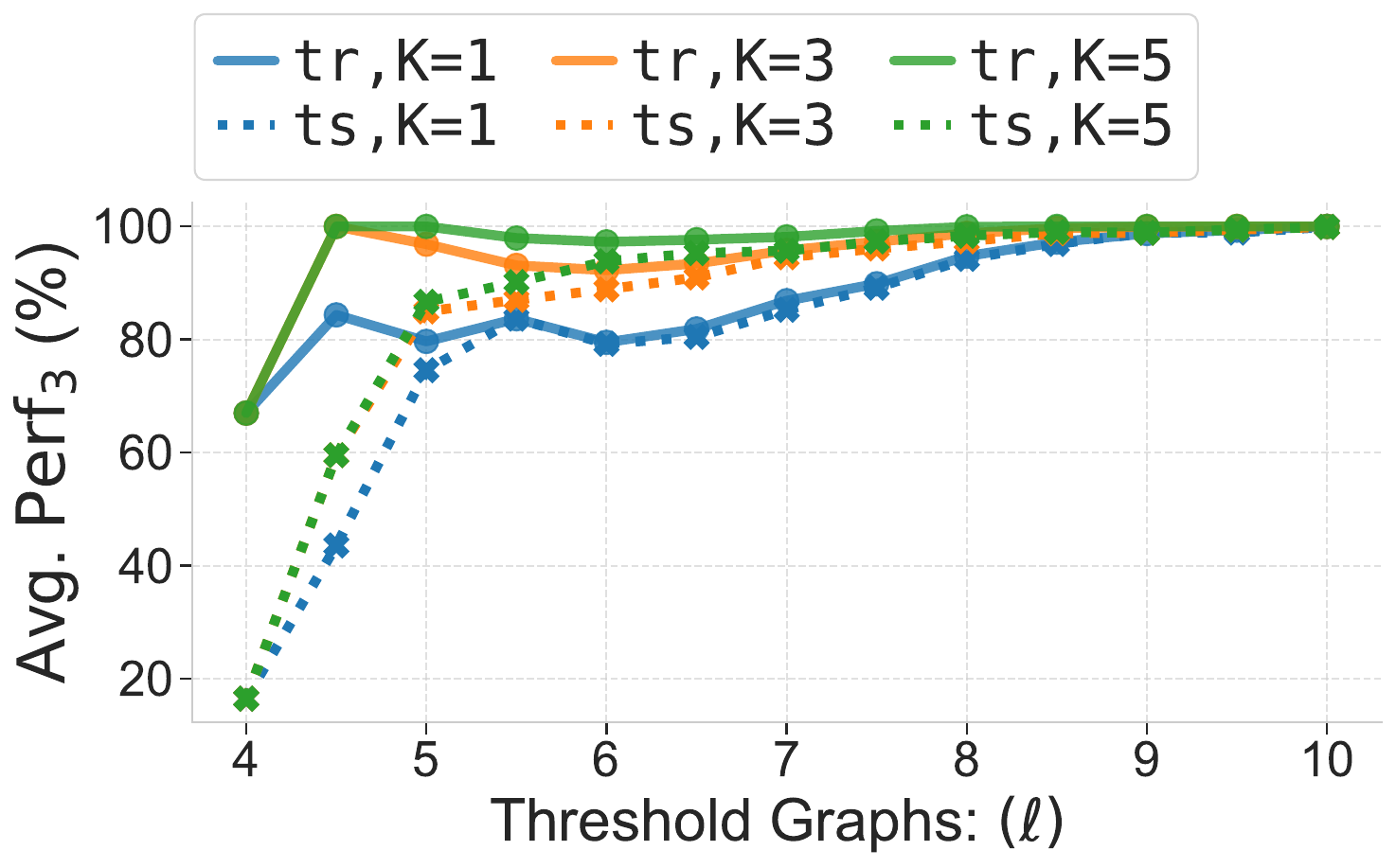}
    \caption{Threshold: \(\mathrm{Perf}_{3}\)}
    \label{fig:app_math_learn_thresh_perf3}
\end{subfigure}

\caption[Performance of Algorithm~\ref{alg:greedy_lbreveal} in the learning setting on the Adult and Math datasets]{Performance of Algorithm~\ref{alg:greedy_lbreveal} in the learning setting on the \textbf{Adult} and \textbf{Math} datasets under three metrics (\(\mathrm{Perf}_{1}, \mathrm{Perf}_{2}, \mathrm{Perf}_{3}\)) for \(k\)NN and threshold generated graphs  (Tables~\ref{tab:adult_kmax_r_stats} and \ref{tab:math_kmax_r_stats}).
Across both datasets, larger budget \(K\) lead to weakly higher performance. 
Increasing graph connectivity raises overall scores while reducing the performance gap between budgets (e.g., in Figures~\ref{fig:app_adult_learn_thresh_perf1}--\subref{fig:app_adult_learn_thresh_perf3}). 
Compared to Figures~\ref{fig:app_math_learn_thresh_perf1} and \ref{fig:app_math_learn_thresh_perf3}, the performance scores cluster more tightly across budgets in Figure~\ref{fig:app_math_learn_thresh_perf2} because unlike them, denominator in \(\mathrm{Perf}_{2}\) includes all sampled agents.}
\label{fig:app_adult-math_learn_knn_thresh}
\begin{picture}(0,0)
    \put(-240,533){\rotatebox{90}{\textit{$k$NN graphs}}}
    \put(-240,400){\rotatebox{90}{\textit{Threshold graphs}}}
    \put(-240,280){\rotatebox{90}{\textit{$k$NN graphs}}}
    \put(-240,158){\rotatebox{90}{\textit{Threshold graphs}}}
\end{picture}
\end{figure}

\begin{figure}[t!]
\captionsetup[subfigure]{justification=Centering}
\centering
\textbf{Portuguese Dataset}\\[1.5ex]
\begin{subfigure}[t]{0.32\textwidth}
\centering
    \includegraphics[width=\textwidth]{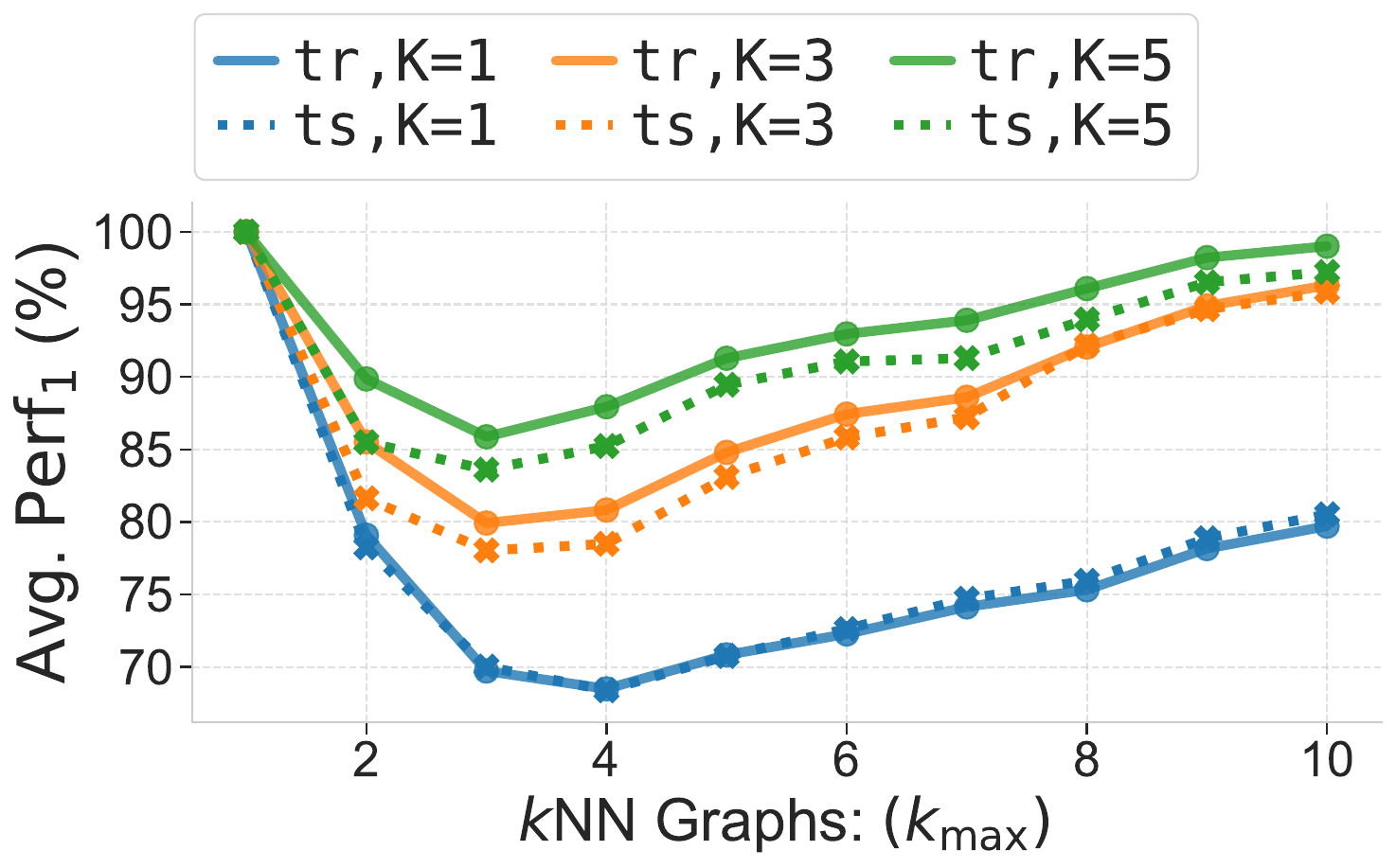}
    \caption{\(k\)NN: \(\mathrm{Perf}_{1}\)}
    \label{fig:app_port_learn_knn_perf1}
\end{subfigure}
\begin{subfigure}[t]{0.32\textwidth}
\centering
    \includegraphics[width=\linewidth]{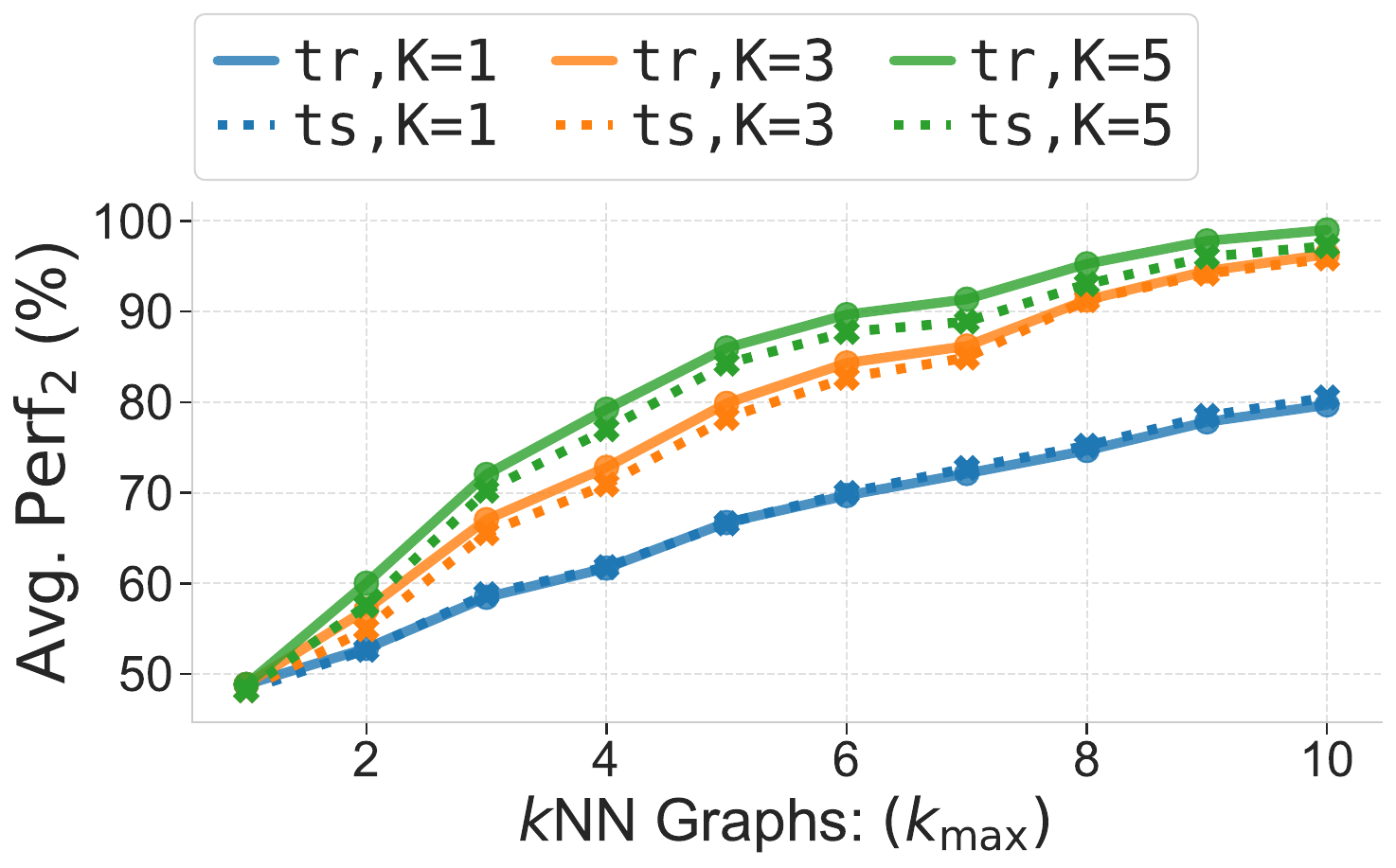}
    \caption{\(k\)NN: \(\mathrm{Perf}_{2}\)}
    \label{fig:app_port_learn_knn_perf2}
\end{subfigure}
\begin{subfigure}[t]{0.32\textwidth}
\centering
    \includegraphics[width=\linewidth]{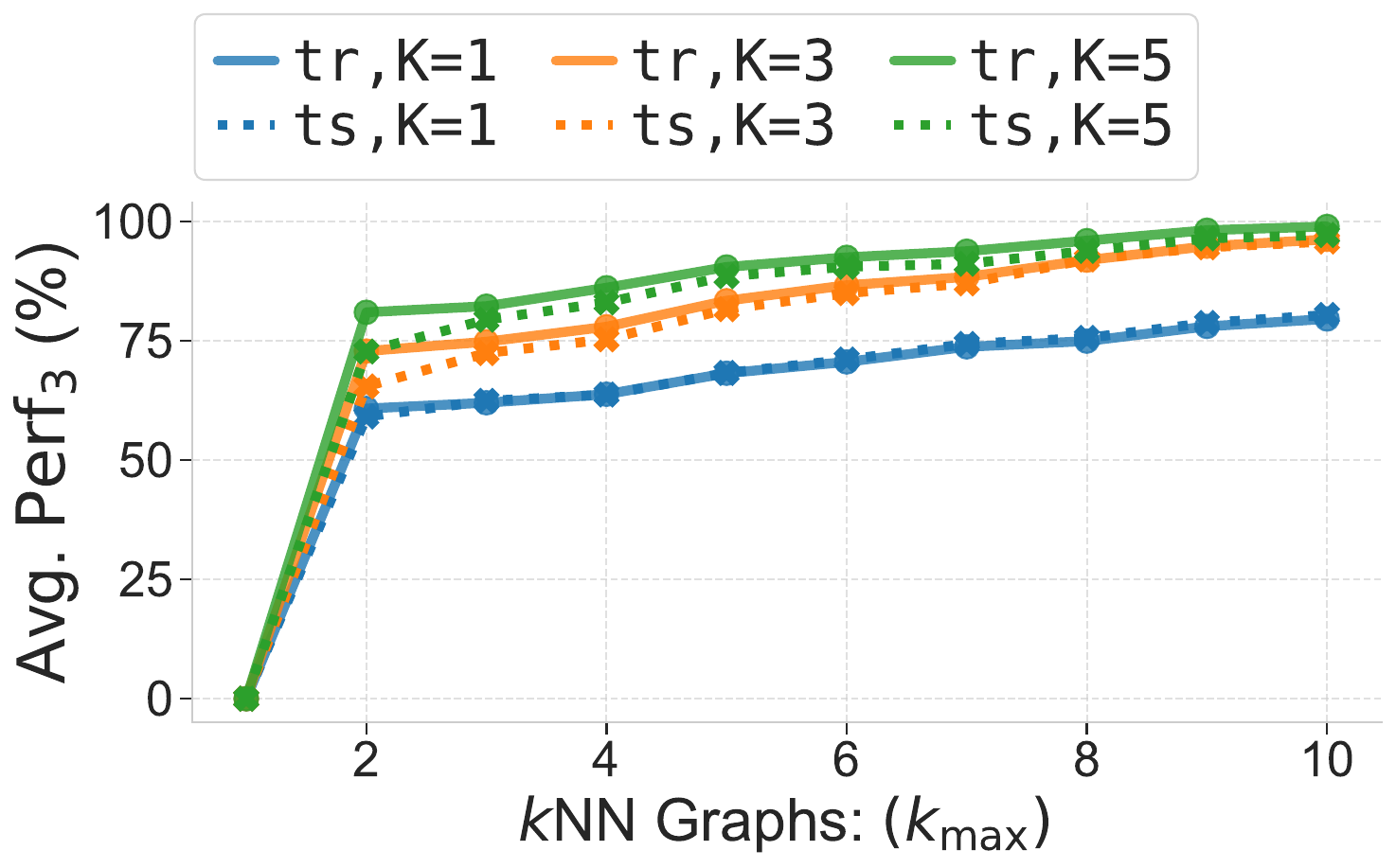}
    \caption{\(k\)NN: \(\mathrm{Perf}_{3}\)}
    \label{fig:app_port_learn_knn_perf3}
\end{subfigure}\\[1ex]

\begin{subfigure}[t]{0.32\textwidth}
\centering
    \includegraphics[width=\textwidth]{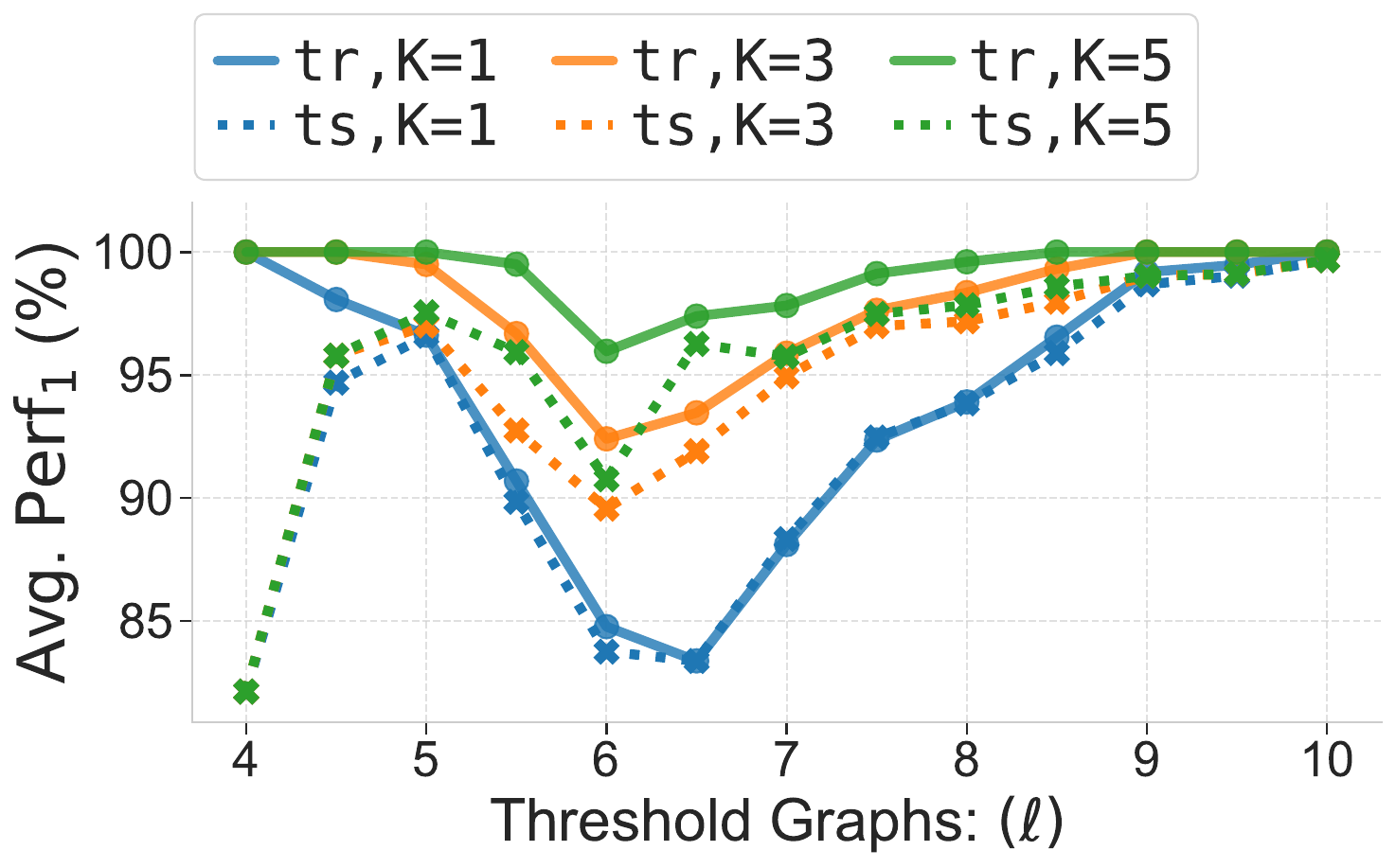}
    \caption{Threshold: \(\mathrm{Perf}_{1}\)}
    \label{fig:app_port_learn_thresh_perf1}
\end{subfigure}
\begin{subfigure}[t]{0.32\textwidth}
\centering
    \includegraphics[width=\linewidth]{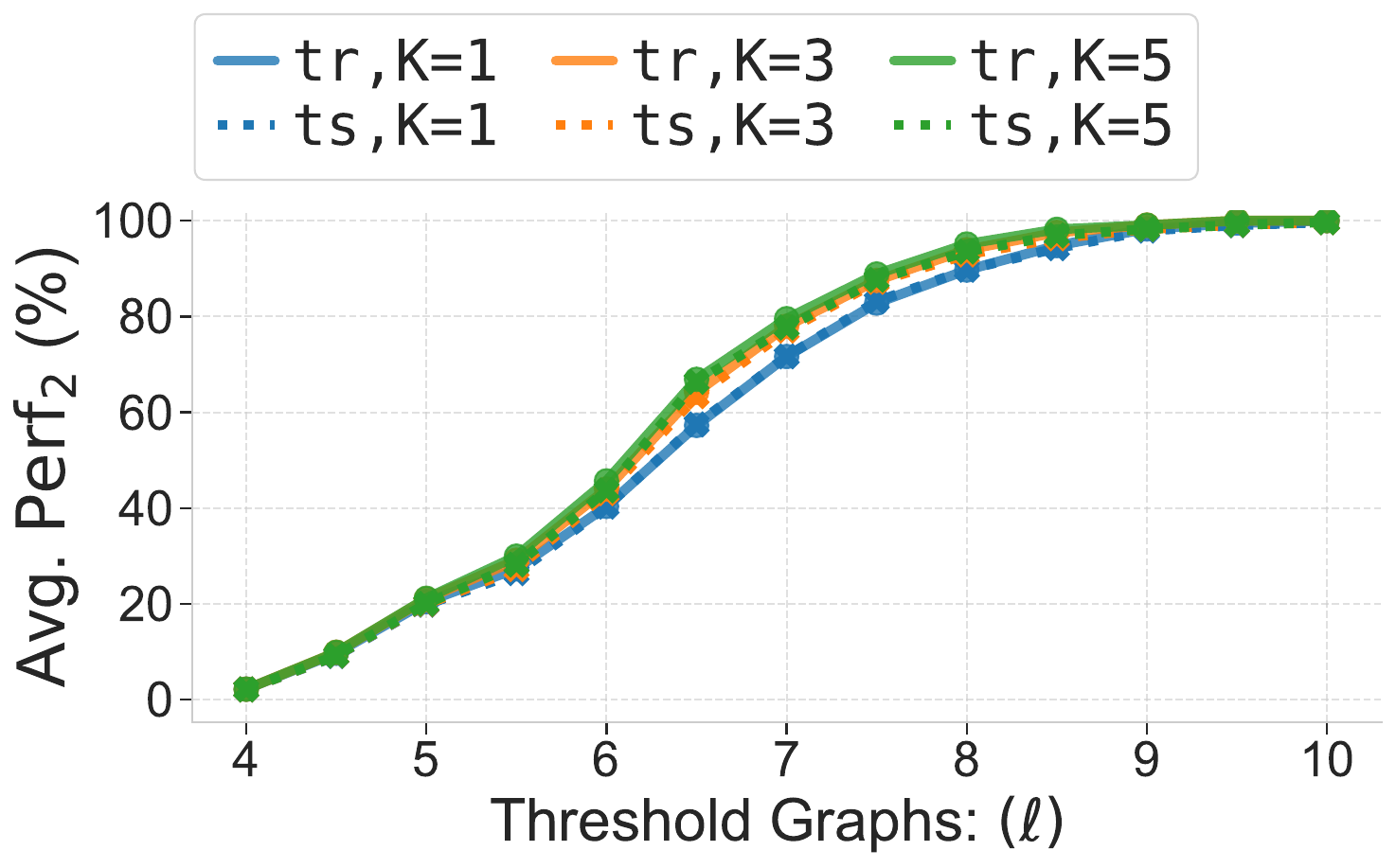}
    \caption{Threshold: \(\mathrm{Perf}_{2}\)}
    \label{fig:app_port_learn_thresh_perf2}
\end{subfigure}
\begin{subfigure}[t]{0.32\textwidth}
\centering
    \includegraphics[width=\linewidth]{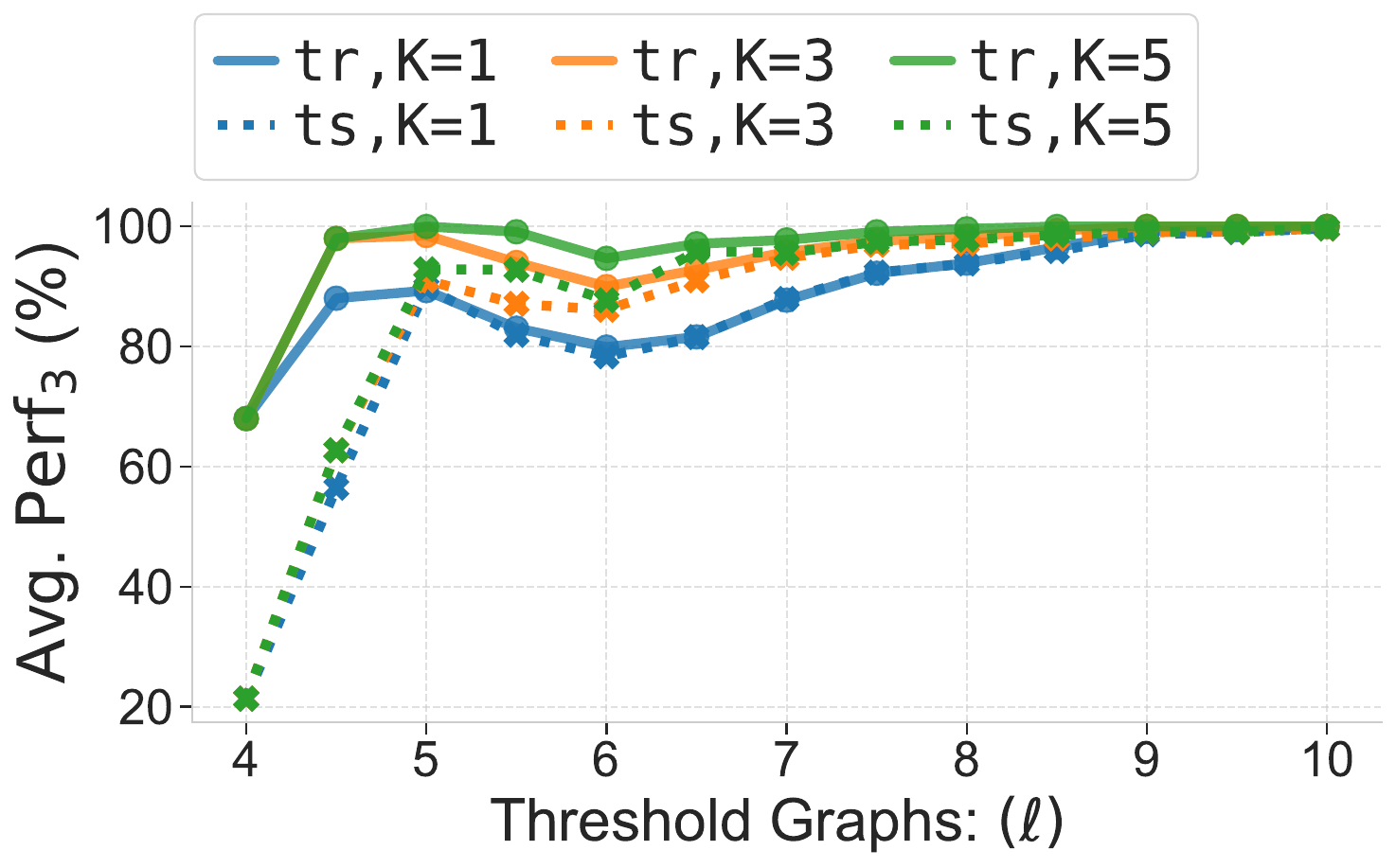}
    \caption{Threshold: \(\mathrm{Perf}_{3}\)}
    \label{fig:app_port_learn_thresh_perf3}
\end{subfigure}
\\[2ex]
\textbf{Productivity Dataset}
\\[2ex]
\begin{subfigure}[t]{0.32\textwidth}
\centering
    \includegraphics[width=\textwidth]{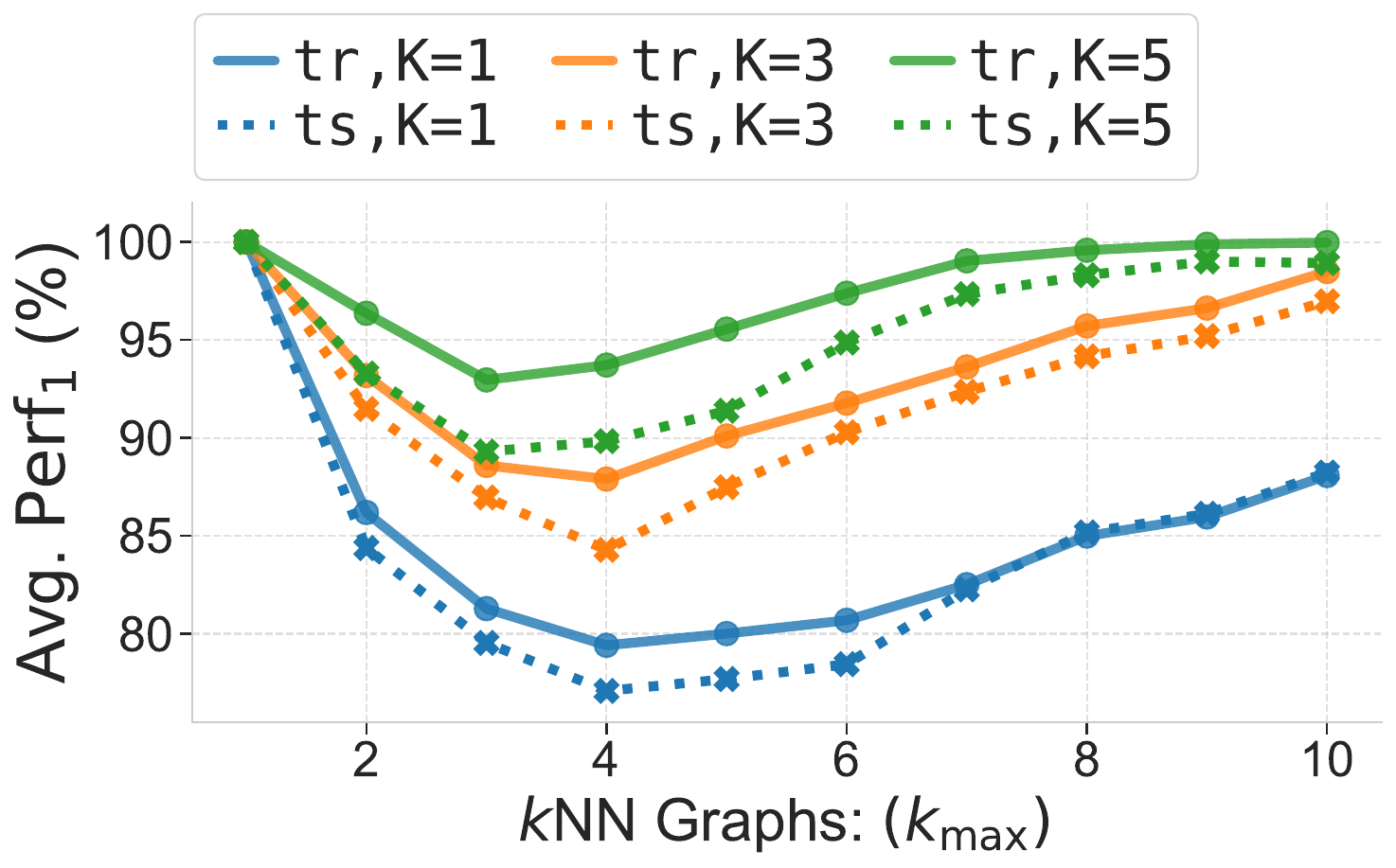}
    \caption{\(k\)NN: \(\mathrm{Perf}_{1}\)}
    \label{fig:app_prod_learn_knn_perf1}
\end{subfigure}
\begin{subfigure}[t]{0.32\textwidth}
\centering
    \includegraphics[width=\linewidth]{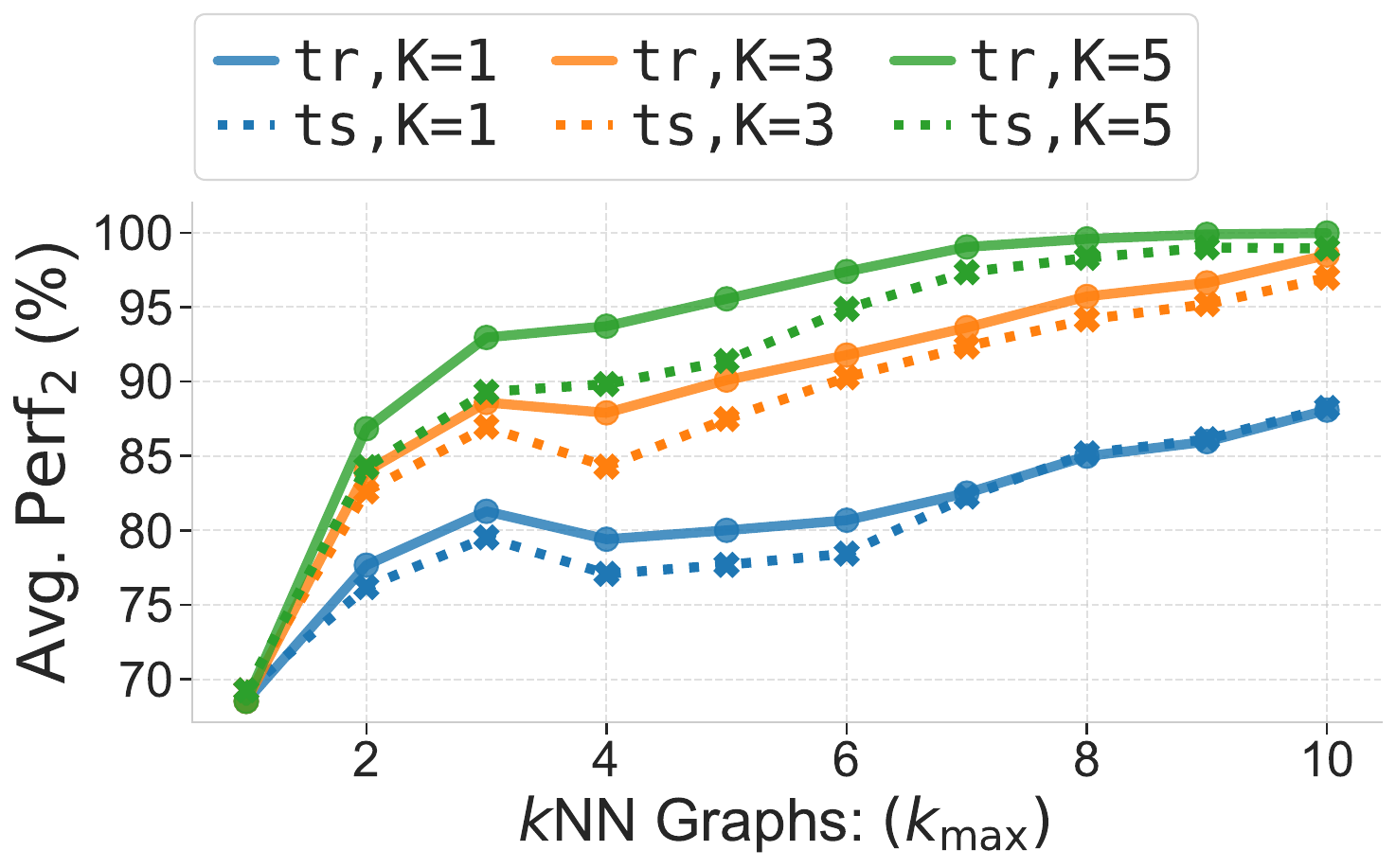}
    \caption{\(k\)NN: \(\mathrm{Perf}_{2}\)}
    \label{fig:app_prod_learn_knn_perf2}
\end{subfigure}
\begin{subfigure}[t]{0.32\textwidth}
\centering
    \includegraphics[width=\linewidth]{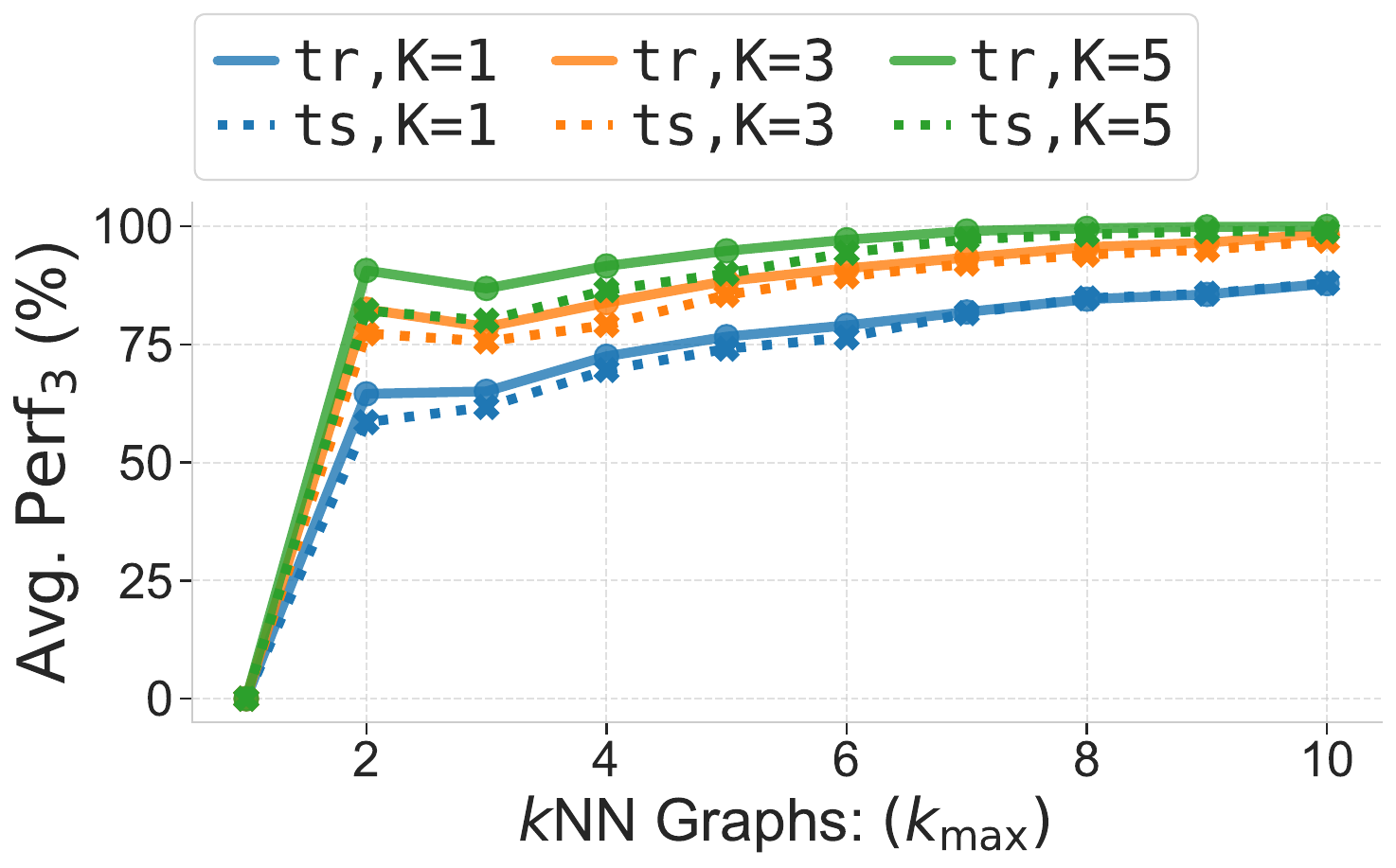}
    \caption{\(k\)NN: \(\mathrm{Perf}_{3}\)}
    \label{fig:app_prod_learn_knn_perf3}
\end{subfigure}\\[1ex]

\begin{subfigure}[t]{0.32\textwidth}
\centering
    \includegraphics[width=\textwidth]{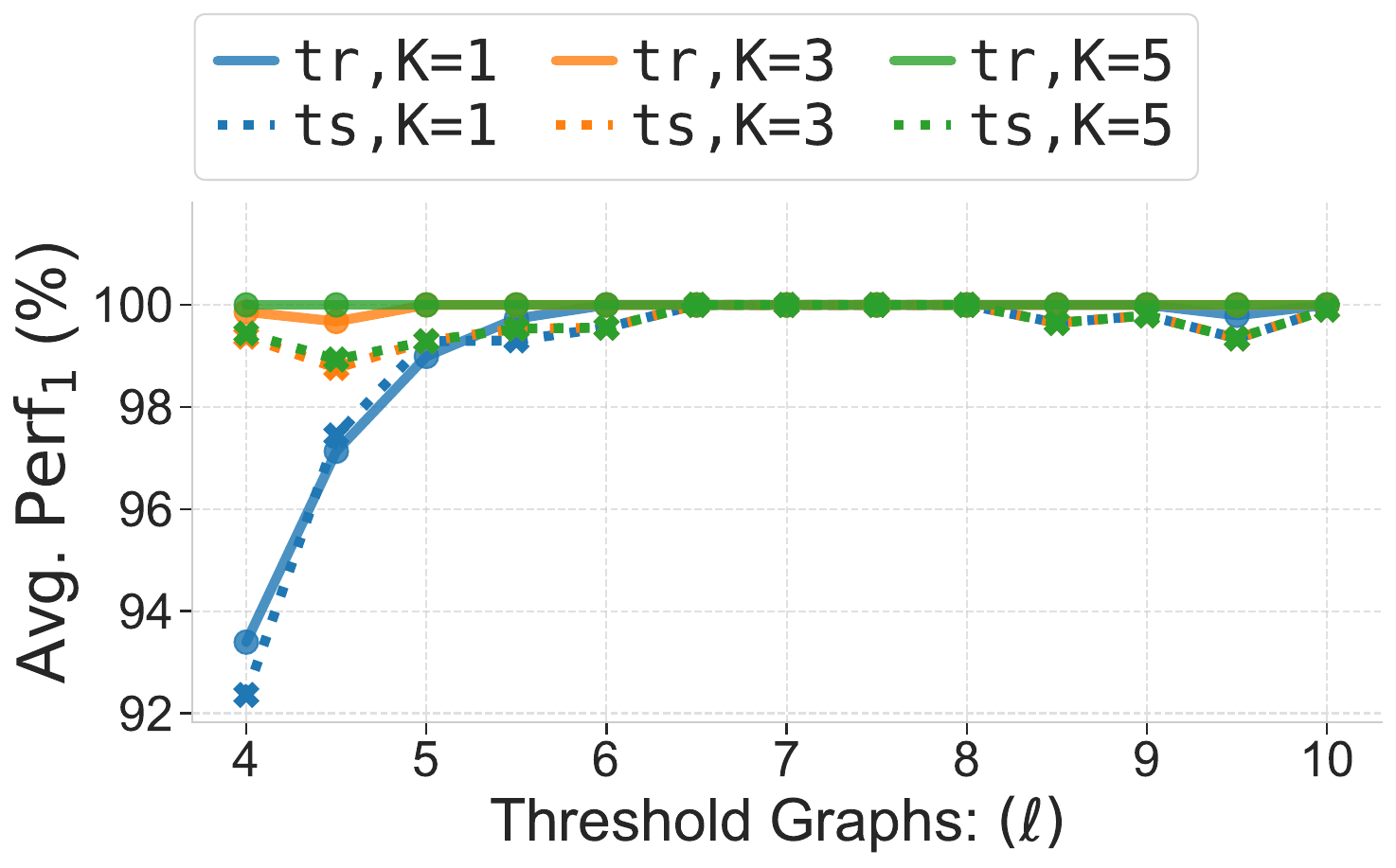}
    \caption{Threshold: \(\mathrm{Perf}_{1}\)}
    \label{fig:app_prod_learn_thresh_perf1}
\end{subfigure}
\begin{subfigure}[t]{0.32\textwidth}
\centering
    \includegraphics[width=\linewidth]{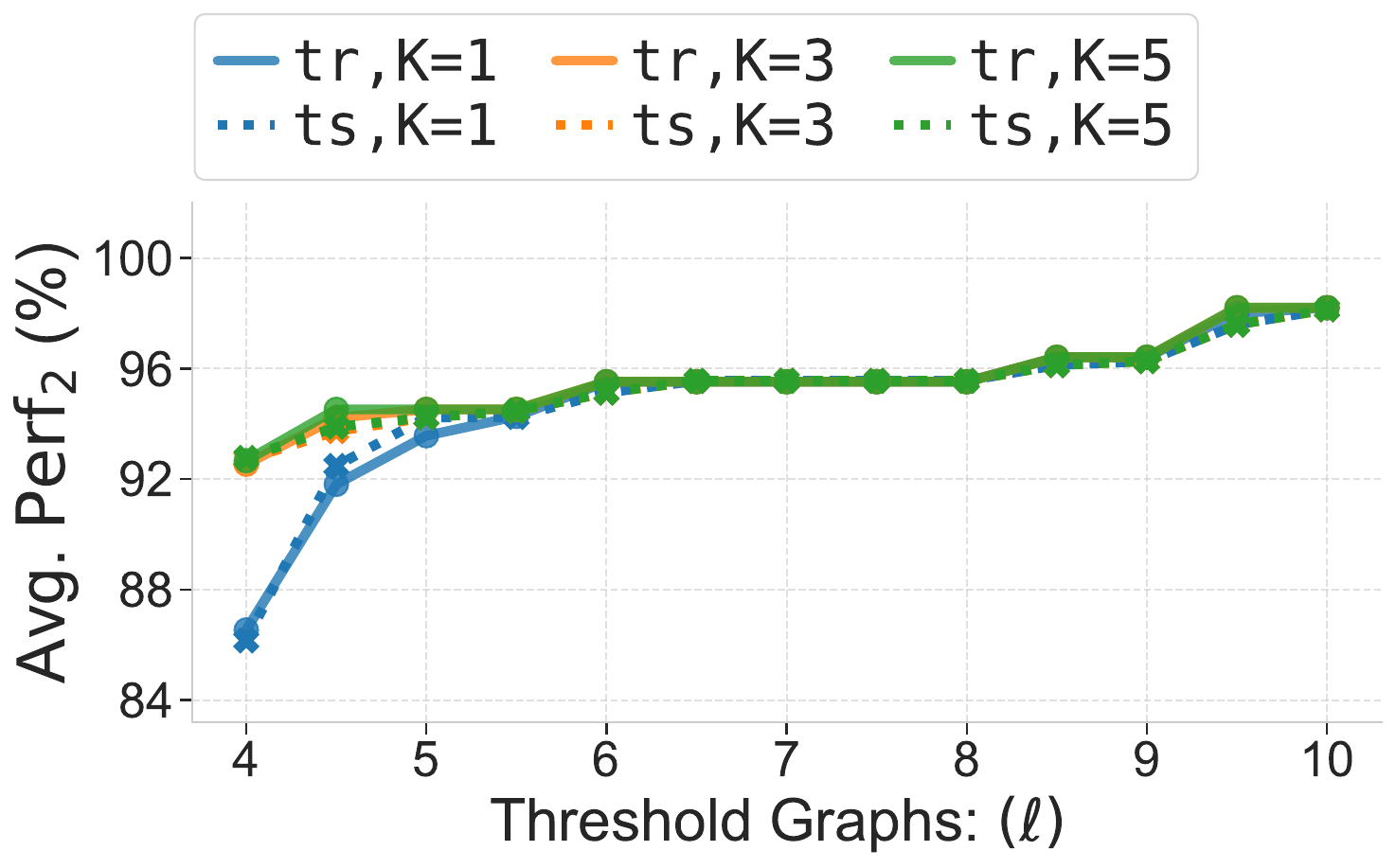}
    \caption{Threshold: \(\mathrm{Perf}_{2}\)}
    \label{fig:app_prod_learn_thresh_perf2}
\end{subfigure}
\begin{subfigure}[t]{0.32\textwidth}
\centering
    \includegraphics[width=\linewidth]{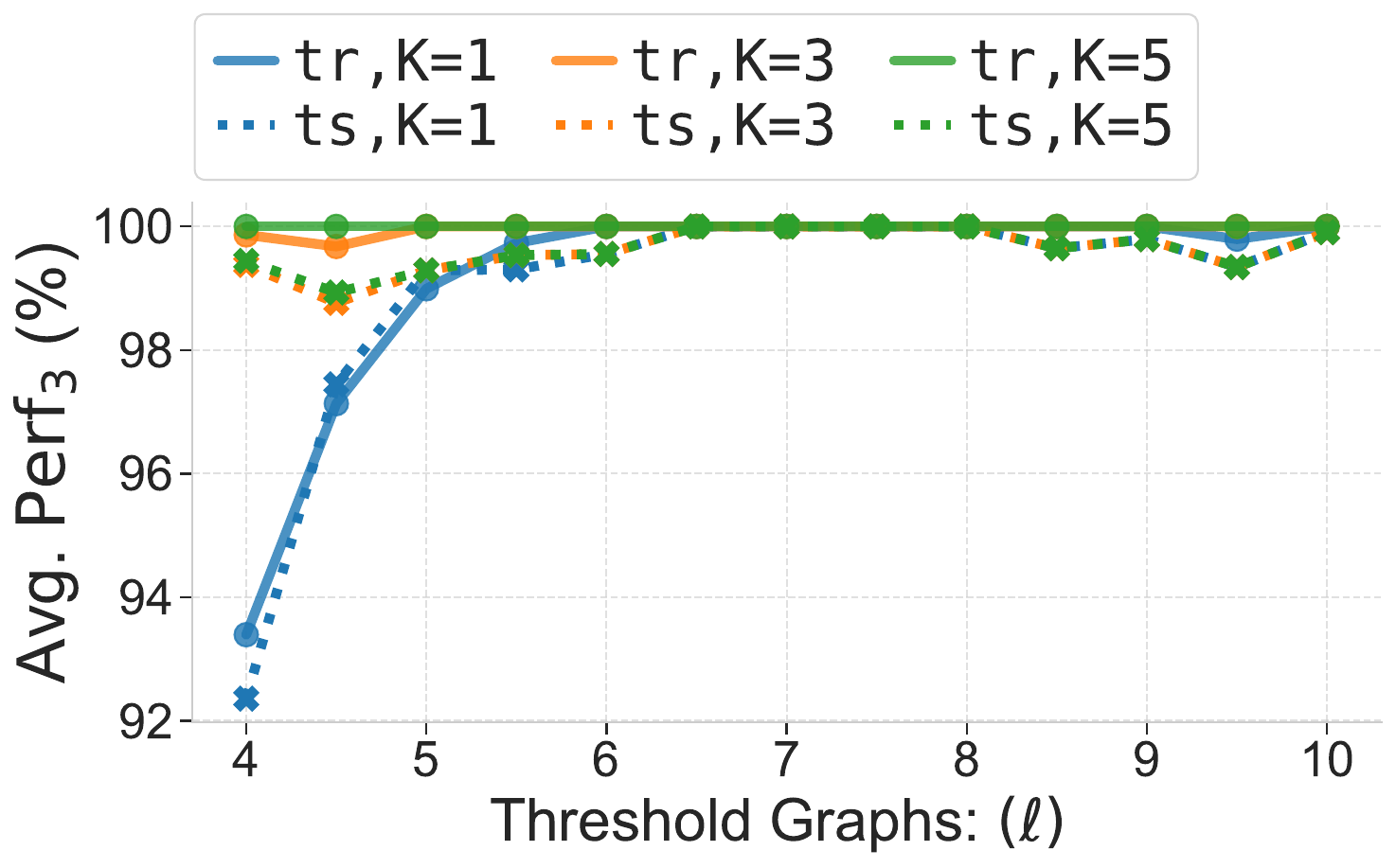}
    \caption{Threshold: \(\mathrm{Perf}_{3}\)}
    \label{fig:app_prod_learn_thresh_perf3}
\end{subfigure}

\caption[Performance of Algorithm~\ref{alg:greedy_lbreveal} in the learning setting on the Portuguese and Productivity datasets]{Performance of Algorithm~\ref{alg:greedy_lbreveal} in the learning setting on the \textbf{Portuguese} and \textbf{Productivity} datasets under three metrics (\(\mathrm{Perf}_{1}, \mathrm{Perf}_{2}, \mathrm{Perf}_{3}\)) for \(k\)NN and threshold generated graphs  (Tables~\ref{tab:portuguese_kmax_r_stats} and \ref{tab:productivity_kmax_r_stats}).
Across both datasets, larger budget \(K\) lead to weakly higher performance. 
Increasing graph connectivity raises overall scores while reducing the performance gap between budgets (e.g., in Figures~\ref{fig:app_prod_learn_thresh_perf1}--\subref{fig:app_prod_learn_thresh_perf3}). 
Because of presence of unhelpable agents in the graphs (Table~\ref{tab:productivity_kmax_r_stats}), $\mathrm{Perf}_{2}$ can remain below $100$ even when the algorithm is optimal (cf. Figure~\ref{fig:app_prod_learn_thresh_perf2}). Lastly, because the Math and Portuguese datasets were curated in a similar manner \citep{CortezP08}, learning setting results on the generated graphs are closely similar.}
\label{fig:app_port-prod_learn_knn_thresh}
\begin{picture}(0,0)
    \put(-240,563){\rotatebox{90}{\textit{$k$NN graphs}}}
    \put(-240,436){\rotatebox{90}{\textit{Threshold graphs}}}
    \put(-240,307){\rotatebox{90}{\textit{$k$NN graphs}}}
    \put(-240,177){\rotatebox{90}{\textit{Threshold graphs}}}
\end{picture}
\end{figure}

Across \(k\)NN and threshold graphs for all datasets, training performance is consistently at least as high as testing performance at comparable budget levels 
(Figures~\ref{fig:app_adult-math_learn_knn_thresh} and \ref{fig:app_port-prod_learn_knn_thresh}).
For \(k\)NN graphs in which each agent has at most one neighbor that is positive or negative 
(Tables~\ref{tab:adult_kmax_r_stats}--\ref{tab:productivity_kmax_r_stats}, $k_{\max}=1$), the performance scores are structure-dependent since there are no revealed targets. Here, \(\mathrm{Perf}_{1}=100\) since all agents with atleast one positive target neighbor are catered to, \(\mathrm{Perf}_{2}\) equals the fraction of agents connected to positive targets, and \(\mathrm{Perf}_{3}=0\) because no agents are helpable 
(Figures~\ref{fig:app_adult-math_learn_knn_thresh} and \ref{fig:app_port-prod_learn_knn_thresh}, 
subfigures (\textbf{a--c, g--i})).

For threshold graphs, particularly those generated with higher thresholds, increased connectivity 
(Tables~\ref{tab:adult_kmax_r_stats}--\ref{tab:productivity_kmax_r_stats} where $\ell \ge 6.5$) leads training and testing performance to converge to the same score, and the budget levels become less impactful 
(Figures~\ref{fig:app_adult-math_learn_knn_thresh} and \ref{fig:app_port-prod_learn_knn_thresh}, 
subfigures (\textbf{d--f, j--l}). These findings are attributed to an increase in the number of positive targets connected to all helpable agents.
When many agents have empty or all-negative target neighborhoods, 
\(\mathrm{Perf}_{2}\) is more affected than the other metrics, since those agents are excluded from the evaluation in \(\mathrm{Perf}_{1}\) and \(\mathrm{Perf}_{3}\). For example, in the Productivity threshold graphs (Table~\ref{tab:productivity_kmax_r_stats}), even when there is a high number of positive targets connected to all helpable agents, \(\mathrm{Perf}_{2}\) remains well below \(100\) because some agents are connected exclusively to negative targets (Figure~\ref{fig:app_prod_learn_thresh_perf2}).

Finally, the three metrics can differ markedly within the same graph, particularly when only one agent has both positive and negative targets in its neighborhood. For example, in the Portuguese \(\ell=4\) threshold graph, of \(224\) agents, \(4\) have only positive neighbors, \(219\) only negative neighbors, and \(1\) has both (Table~\ref{tab:portuguese_kmax_r_stats}, $\ell=4.0$). Likewise, in the Math \(\ell=4\) threshold graph, among \(206\) agents, \(10\) have only positive neighbors, \(5\) only negative neighbors, \(190\) have no neighbors, and \(1\) has both (Table~\ref{tab:math_kmax_r_stats}, $\ell=4.0$).
If this single mixed-neighborhood agent appears in the training set, Algorithm~\ref{alg:greedy_lbreveal} can reveal a target that ensures that the agent emulates a positive target with probability one. If it appears in the test set instead, the algorithm reveals no targets during training, yielding zero social welfare for that agent at test time. As a result, \(\mathrm{Perf}_{1}\) may reach \(100\%\) on the training set yet be very low on the test set. Overall, \(\mathrm{Perf}_{1}\) is reduced by the large number of agents who cannot be helped by Algorithm~\ref{alg:greedy_lbreveal}, whereas \(\mathrm{Perf}_{3}\) captures the average probability of helping the mixed agent across splits (Figures~\ref{fig:app_math_learn_thresh_perf1}--\subref{fig:app_math_learn_thresh_perf3} and \ref{fig:app_port_learn_thresh_perf1}--\subref{fig:app_port_learn_thresh_perf3}).

\section{Limitations and Future Works}
\label{sec:revealrm_limitations}

In addition to the recommendations outlined in the Section~\ref{sec:revealrm_conclusion}, we note several limitations and identify opportunities for future work below.

Currently, we focus on how the proxy social welfare function can help achieve a constant-factor approximation to the true social welfare when the revealed set can include negative targets. An interesting direction for future work is to further investigate our observation that an equal amount of proxy social welfare does not necessarily lead to similar emulation choices, and to examine the implications of this divergence for modeling fairness when agents are partitioned into distinct groups.  Additionally, further experiments could investigate the sensitivity of the greedy approach to parameter $c$, and also analyze how the divergence between the proxy and true welfare (through parameter $c$) is reflected in the performance of the greedy algorithm.

We consider a setting in which the social planner has complete knowledge of the graph, or observes an agent’s neighborhood upon sampling, when the left side of the graph is replaced with a probability distribution over agents.  Future work could explore learning in scenarios where the social planner only has access to partial information. In addition, as captured by the bipartite graph structure, we assume that there are no interactions among agents or among targets. Future work could relax this assumption by examining more general graph structures that allow for richer forms of interaction.

While our experiments are extensive, covering four datasets, multiple greedy variants, and 23 generated bipartite graphs per dataset, a viable next step is to evaluate our models on real-world bipartite networks. For instance, one could examine the performance of the greedy approaches on the social network of mentor-mentee relationships drawn from academic genealogy datasets \citep{academictree}.

Lastly, our work mainly focuses on a setting where role models or options are classified as either positive or negative. However, the core structure of our model naturally extends to settings with multiple types. For example, if “good” is defined relative to a threshold, a social planner could direct agents toward role models that exceed this benchmark, and during emulation, agents avoid those revealed to be below the desired threshold. For future work, it would be interesting to study the settings in which the role model or option quality is continuous and or can be ranked.

%
%
%
\chapter{PAC Learning with Improvements}
\label{app:paclearn}
\section{Additional Related Work}
\label{app:related-work}

\paragraph{Classification of gaming agents.}
\citet{hardt2016strategic} formalized the concept of strategic behavior, often referred to as ``gaming,'' where test-set agents who are negatively classified intentionally modify their features---within the bounds of a separable cost function---without altering their target label, to deceive the model into classifying them as positive. They theoretically and empirically showed that their strategy-robust algorithm outperforms the standard SVM algorithm under gaming. However, as the extent of gaming increases, overall model accuracy declines. 
\citet{revealed_preferences} also study a Stackelberg equilibrium where agents strategically respond to classification learners. However, unlike \citet{hardt2016strategic}, their model assumes that the learner lacks direct knowledge of the agents' utility functions and instead infers them through observed revealed preferences. Additionally, agents arrive sequentially, and only the true negatives strategically respond to the learner. The learner's objective is to minimize the Stackelberg regret. 
\citet{Chen2019LearningSL} also study a  learner whose goal is to minimize the Stackelberg regret, where gaming agents arrive sequentially. However, unlike \citet{revealed_preferences} who assumes a convex loss function, they deal with a less smooth agent utility function and learner loss function. They propose the Grinder algorithm, which adaptively partitions the learner's action space based on the agents' responses. 
Performative prediction~\citep{perdomo2021performativeprediction} considers a setting that involves a repeated interaction between the classifier and the agents, and as a result the underlying distribution of the gaming agents may change over time. 

\paragraph{Classification of  agents that can both game and improve.} 
Unlike earlier works in the strategic classification literature, which primarily focus on settings where agents engage in gaming behavior, \citet{Kleinberg2018HowDC} examine a scenario where agents can genuinely improve. In this context, the agent can modify their observable features and true label to achieve a positive model outcome. The authors demonstrate that a learner employing a linear mechanism can encourage rational agents, who optimize their allocation of effort, to prioritize actions that result in meaningful improvement. They show how to achieve this by selecting an evaluation rule that incentivizes a desirable effort profile. 

\citet{ahmadi2022classificationstrategicagentsgame}, like \citet{Kleinberg2018HowDC}, consider the agents' potentially truthful and actionable responses to the model. However, the primary objective of \citet{ahmadi2022classificationstrategicagentsgame} is to maximize true positive classifications while minimizing false positives. Notably, for the linear case, they show that the resulting classifier can become non-convex, depending on the agents' initial positions. 

On the other hand, \citet{ahmadi2022settingfairincentivesmaximize} design reachable sets of target levels such that they can incentivize effort-bounded agents within each group to improve optimally.

\paragraph{Theoretical guarantees of incentive-aware or incentive-compatible classifiers. } 
\citet{Zhang_Conitzer_2021} show that the vanilla ERM principle fails under strategic manipulation (gaming), even in simple scenarios that would otherwise be straightforward without gaming. To address this, they propose the concepts of incentive-aware and incentive-compatible ERMs, theoretically analyzing the corresponding classifiers, their sample complexity, and the impact of the VC dimension on the associated hypothesis class. Finally, they extend their analysis to ERM-based learning in environments with transitive strategic manipulation. 

Given adversarial data points wanting to receive an incorrect label, \citet{Cullina2018PAClearningIT} theoretically show that the sample complexity of PAC-learning a set of halfspace classifiers does not increase in the presence of adversaries bounded by convex constraint sets and that the adversarial VC dimension can be arbitrarily larger or smaller than the standard VC dimension. 
\citet{strategicPAC} provide theoretical guarantees for an offline, full-information strategic classification framework where data points have distinct preferences over classification outcomes (\(+\) or \(-\)) and incur varying manipulation costs, modeled using seminorm-induced cost functions. They propose a PAC-learning framework for strategic linear classifiers in this setting, providing a detailed analysis of their statistical and computational learnability. Additionally, they extend the concept of the adversarial VC dimension \citet{Cullina2018PAClearningIT} to this strategic context. They also show, among other things, that employing randomized linear classifiers can substantially improve accuracy compared to deterministic methods.

\paragraph{Reliable machine learning.} 
The concept of risk aversion in our work is closely related to selective classification or machine learning with a reject option~\citep{yaniv10a,NIPS2017_4a8423d5,Hendrickx2021MachineLW}, where the classifier balances the trade-off between risk and coverage, opting to abstain from making predictions when it is likely to make mistakes. Similarly, risk-averse classification aligns with aspects of reliable or learning with one-sided error~\citep{natarajan1987learning,onesidedagree}, particularly positive reliable learners~\citep{kalaiReliable}, which aim to achieve zero false positive errors while minimizing false negatives. Prior work has shown connections between strategic classification and adversarial learning (e.g. \citep{strategicPAC}), but it remains an interesting open question if similar connections can be established between learning with improvements and reliable learning in the presence of adversarial attacks~\citep{balcan2022robustly,balcan2023reliable,blum2024regularized}.

\paragraph{Adversarial robustness.} Our results indicate that the properties of the concept class that govern learnability with improvements are different from those established for adversarial robustness. In particular, finite VC dimension is not sufficient for PAC learnability with improvements (Example \ref{ex:example_notsufficient}) as opposed to adversarial robustness, where it is sufficient for (improper) learning~\citep{montasser2019vc}. Furthermore, our risk-averse learning algorithm is very different from techniques proposed in the adversarial literature, including robust Empirical Risk Minimization~\citep{Cullina2018PAClearningIT,attias2022improved}, boosting algorithms whose generalization is based on sample compression schemes \citep{montasser2019vc,montasser2020reducing,attias2022characterization,attias2023adversarially} and approaches developed for neural networks~\citep{AlexAdversarial,Cohen2019CertifiedAR,balcan2023analysis}. Similarly, our algorithm for learning halfspaces on a unit ball differs from classical approaches for learning from malicious noise~\citep{Awasthi2013ThePO,Balcan2020NoiseIC}. However, due to analogies between the frameworks, it is meaningful to ask future research questions along the lines of the directions pursued in the adversarial robustness literature---for example, extensions to unknown or uncertain improvement sets \citep{montasser2021adversarially,lechner2022learning}, and learning with tolerance \citep{ashtiani2023adversarially,raman2023proper,ashtiani2025simplifying}.

\section{Separation from Standard PAC Learning Model when 
\texorpdfstring{$\tilde{\mathcal{H}}\subset{\mathcal{H}}$}{H-tilde ⊂ H}}
\label{app:example-error-gap}

Example~\ref{ex:example4} is a proof of Theorem~\ref{thm:non-realizable-targets}

\begin{example}[Error gap when the learner's hypothesis space $\tilde{\mathcal{H}}$ is a strict subset of the concept space ${\mathcal{H}}$ that contains $f^*$]
    Let $\mathcal{X}=[-1,1]$ and $\tilde{\mathcal{H}}$ denote the set of concepts including unions of up to $k$ open intervals. The set of possible improvements for any point $x\in \mathcal{X}$ is given by $\mathcal{X} \cap \mathbb{Q}$, where $\mathbb{Q}$ denotes the set of rational numbers. Suppose the data distribution is uniform over $\mathcal{X}$. We set the target concept $f^*$ as follows

    \[f^*(x)=
    \begin{cases} 
            0, & \text{if } x<0, \text{or } x\in \mathbb{Q}, \\ 
            1, & \text{otherwise},
        \end{cases}
    \]
    
    Note that rationals are dense in $[0,1]$ and the set of all rationals have a Lebesgue measure zero.
    Thus, on any finite sample $S\in \mathcal{X}^m$, any sampled point $x$ will have a positive label according to $f^*$ iff $x\ge 0$ (with probability 1).
    In the standard PAC learning setting, the classifier $\Tilde{f}=\mathbb{I}\{x\in(0,1)\}$  achieves zero error w.r.t.\ the target $f^*$. This is because the misclassification error for points in $\mathbb{Q}$ is  zero.
    
    In our setting where agents have the ability to improve, for an $h\in \tilde{\mathcal{H}}$ which predicts any point $x'$ in $\mathbb{Q}$ as positive,  all negative agents in $[-1,0)$ can move to such a point $x'$ and be falsely classified as positive. This corresponds to a  lower bound of $\frac{1}{2}$ on the error.
    Since rationals are dense in the reals, any open interval which $h$ classifies as positive must contain a point in $\mathbb{Q}$.
    On the other hand, if $h$ classifies no point as positive, then error rate is again $\frac{1}{2}$ as all the positive points are misclassified.
\label{ex:example4}
\end{example}

\section{Missing Proofs from Section \ref{sec:paclearn_geometric-concepts}
\label{app:proofs-geometric}}

We include below missing proofs from Section \ref{sec:paclearn_geometric-concepts}.

\subsection{Learning Thresholds with the Uniform Distribution}
\label{app:thresholds-uniform}

Proof of Theorem \ref{thm:thresholds-uniform}

\begin{proof}
    Let $S\overset{\text{i.i.d.}}{\sim} \mathcal{D}^m$, where $\mathcal{D}$ is the uniform distribution over $[0,1]$. By using a standard calculation of the sample complexity of thresholds, 
    \begin{align}\label{eq:threshold-uniform}
    \begin{split}
    \underset{S\sim \mathcal{D}^m}{\mathbb{P}}\left[\underset{x\sim \mathcal{D}}{\mathbb{P}}\left[h_{t^{\star}}(x)\neq h_{S^+}(x)\right]>\epsilon\right]
    &\leq \prod^m_{i=1}\mathbb{P}\left[x_i \notin [t^{\star},t^{\star}+\epsilon]\right]     
    \\
    &\leq 
    (1-\epsilon)^m
    \\
    &\leq
    e^{-\epsilon m}
    \\
    &\leq \delta, 
    \end{split}
    \end{align}

    \noindent where the last inequality holds for $m\geq \frac{1}{\epsilon}\log\frac{1}{\delta}$. Since whenever $h_{S^+}$ classifies a point as positive, $h_{t^{\star}}$ also classifies it as positive, 
    any negative point that improves in response to $h_{S^+}$ must move to a true positive point, and the error can only decrease in the improvements setting for the choice of $h_{S^+}$.

    Now, since we allow improvements of distance $r$, the points in the interval $[t_{S^+}-r, t_{S^+}]$ that would have been classified negatively without improvement are able to improve under $h_{S^+}$ (and indeed improve to be positive with respect to $h_{t^{\star}}$) and are thus classified correctly. The points on which $h_{S^+}$ makes mistakes are those in the interval $[t^{\star}, t_{S^+}-r]$. Since $\mathcal{D}$ is uniform, our previous inequality implies that with probability at least $1-\delta$ we have $t_{S^+}\leq t^{\star}+\epsilon$. This implies the that error is at most $\max(\epsilon - r, 0)$ with probability $1 - \delta$ as desired.  
\end{proof}

\subsection{Learning Thresholds with An Arbitrary Distribution}
\label{app:thresholds-arbitrary}

\begin{theorem}[Thresholds, arbitrary distribution]
\label{thm:threshold-arbitrary-D}
    Let the improvement set $\Delta$ be the closed ball with radius $r$, $\Delta(x) =\{ x'\mid |x - x'| \leq r\} $.
    For any distribution $\mathcal{D}$, and any \(\epsilon,\delta \in (0,1/2)\), with probability $1-\delta$,
    \begin{align}
       \textsc{Loss}_{\mathcal{D}}(h_{S^+},h_{t^{\star}})
       \leq (\epsilon -p(h_{S^+};h_{t^{\star}},\mathcal{D},r))_+,
    \end{align}
    where 
    \begin{align}
        p(h_{S^+};h_{t^{\star}},\mathcal{D},r)= \mathbb{P}_{x\sim \mathcal{D}}\left[x\in[t_{S^+}-r,t_{S^+}]\right],
    \end{align}
    with sample complexity $M = O\left(\frac{1}{\epsilon} \log \frac{1}{\delta}\right).$

\label{thm:thresholds-general}
\end{theorem}

\begin{proof}

    Let $t_0$ be such that $\mathbb{P}_{x\sim\mathcal{D}}\left[x\in [t^{\star},t_0]\right]=\epsilon$. By following the same derivation as in Eqn.~\ref{eq:threshold-uniform} and replacing $t^{\star}+\epsilon$ with $t_0$, we get that 
    \begin{align}
        \underset{S\sim \mathcal{D}^m}{\mathbb{P}}\left[\underset{x\sim \mathcal{D}}{\mathbb{P}}\left[h_{t^{\star}}(x)\neq h_{S^+}(x)\right]\leq\epsilon\right],
    \end{align}
    with probability $1-\delta$ for $m\geq \frac{1}{\epsilon}\log\frac{1}{\delta}$.
    
    The points in the interval $[t_{S^+}-r, t_{S^+}]$ are able to improve under $h_{S^+}$ and thus classified correctly. The gain to the error of $h_{S^+}$ would be the probability mass of points in $\mathcal{D}$ that fall into this interval, defined as 
    \begin{align}
        p(h_{S^+};h_{t^{\star}},\mathcal{D},r)= \mathbb{P}_{x\sim \mathcal{D}}\left[x\in[t_{S^+}-r,t_{S^+}]\right].
    \end{align}
    The points on which $h_{S^+}$ makes mistakes are those in the interval $[t^{\star}, t_{S^+}-r]$. 
    
    We conclude that with probability at least $1-\delta$ we have $t_{S^+}\leq t_0$, and given the improvement of points in $[t^{\star}, t_{S^+}-r]$ we have an error at most $\max(\epsilon -  p(h_{S^+};h_{t^{\star}},\mathcal{D},r), 0)$ with probability $1 - \delta$ as desired. 
    
\end{proof}

\subsection{Learning Axis-Aligned  Hyperrectangles}
\label{app:rectangles}

Proof of Theorem \ref{thm:rectangles}

\begin{proof}

    Let $R_S^c=\clos_{\mathcal{H}_{\text{rec}}}\left(\{x_i \in S : y_i = 1\}\right)$ be the output of the closure algorithm.
    For the standard PAC setting, we have 
    \begin{align}
        \underset{S\sim \mathcal{D}^m}{\mathbb{P}}\left[\underset{x\sim \mathcal{D}}{\mathbb{P}}\left[R_S^c(x)\neq R^{\star}(x)\right]\leq\epsilon\right],
    \end{align}
    with probability $1-\delta$ for $m\geq \Omega\left(\frac{1}{\epsilon}\left(d+\log\frac{1}{\delta}\right)\right)$, see \citep{auer1997learning,darnstadt2015optimal}. Since whenever $R_S^c$ classifies a point as positive, $R^{\star}$ also classifies it as positive,
    any negative point that improves in response to $R_S^c$ must move to a true positive point and the error can only decrease in the improvements setting for the choice of $R_S^c$.
    
    Now, in order to quantify the gain in error from the improvements, we define the ``outer boundary strip" of $R_S^c$.
    Let the rectangle defined by
    $R_S^c=\prod_{i\in [d]}[a_i,b_i]$.
    The points that are able to improve under $R_S^c$ are exactly fall into the outer boundary strip of size $r$, defined as
    
    \begin{align}
        \bs(R_S^c,r)
        =
        \prod_{i\in [d]}[a_i+r,b_i+r]
        \setminus
        \prod_{i\in [d]}[a_i,b_i].
    \end{align}
    
    Note that this is exactly the improvement region of $R_S^c$: $\bs(R_S^c,r)=\ir(R_S^c;R^{\star},\Delta)$.
    Under general distribution $\mathcal{D}$, the probability mass of the improvement region is
    \begin{align}
        \mathbb{P}_{x \sim \mathcal{D}}\left[x \in \ir(R_S^c;R^{\star},\Delta) \right]
        =
        \mathbb{P}_{x\sim\mathcal{D}} \left[x\in \bs(R_S^c,r) \right],
    \end{align}
    since $R_S^c$ is the smallest rectangle that fits $S$, these points that are able to improve under $R_S^c$  indeed improve to be positive with respect to $R^{\star}$.
    This implies that 
    \begin{align}
       \textsc{Loss}_{\mathcal{D}}(R_S^c,R^{\star})
       \leq 
       \max\left(\epsilon - \mathbb{P}_{x \sim \mathcal{D}}\left[x \in \ir(R_S^c;R^{\star},\Delta) \right]\right), 0).
    \end{align}
    
    Now, for the uniform distribution, we can compute an exact expression of the improvement region. Let $l_i=b_i-a_i$, then 
    \begin{align}
        \mathbb{P}_{x \sim \mathcal{D}}\left[x \in \ir(R_S^c;R^{\star},\Delta) \right]
        &=
        \prod_{i\in [d]}(l_i+2r)-\prod_{i\in [d]}l_i.
    \end{align}
    For $d=2$, we get
    \begin{align}
      \mathbb{P}_{x \sim \mathcal{D}}\left[x \in \ir(R_S^c;R^{\star},\Delta) \right]  
      &=
      (l_1+2r)(l_2+2r)-l_1l_2
      \\
      &=
      2r(l_1+l_2)+4r^2.
    \end{align}

\end{proof}

\subsection{Hardness of Proper Learning in the Absence of the Intersection Closed Property}
\label{app:hardness-intersection-closed}

Proof of Theorem \ref{thm:hardness-intersection-closed}

\begin{proof}
    For any concept $h\in\mathcal{H}$, let $\mathcal{X}^+_h$ denote the set of points $\{x\in\mathcal{X}\mid h(x)=1\}$ positively classified by $h$. Since $\mathcal{H}':=\mathcal{H}\mid_{\mathcal{X}\setminus\{x'\}}$ is not intersection-closed, there must exist a set $S\subseteq \mathcal{X}\setminus \{x'\}$ such that $\clos_{\mathcal{H}'}(S)\notin \mathcal{H}'$. For the uniformly negative point $x'$, we have its improvement set as $\Delta(x')=\mathcal{X}\setminus S$. For points in $\clos(S)$ we have the improvement set as the empty set. We set the data distribution $\mathcal{D}$ as the uniform distribution over $\clos(S)\cup\{x'\}$. Let $h_1\in\mathcal{H}$ be a minimally consistent classifier w.r.t.\ $S$, i.e.\ if $h'\in\mathcal{H}$ and $\mathcal{X}^+_{h'}\subseteq \mathcal{X}^+_{h_1}$, then $h'=h_1$. By choice of $S$, there is a point $x_1\in \mathcal{X}^+_{h_1}\setminus \clos_{\mathcal{H}'}(S)$. By the definition of closure of $S$, there must exist $h_2\in\mathcal{H}$ consistent with $S$ (assumed minimally consistent WLOG) such that $h_2(x_1)=0$. Also, since $h_1$ was chosen to be minimally consistent, there must exist $x_2\in \mathcal{X}^+_{h_2}$ such that $h_1(x_2)=0$. We will set the target concept $f^{\star}$ to one of $h_1$ or $h_2$.
    
    Now any learner that picks a concept not consistent with $S$ will clearly suffer a constant error on the points in $\clos(S)$ which are incorrectly classified as negative and not allowed to improve. Suppose therefore that the learner selects a hypothesis $h$ consistent with $S$. Let $\tilde{h}$ denote a classifier which is minimally consistent with $S$ and $\mathcal{X}^+_{\tilde{h}}\subseteq \mathcal{X}^+_{h}$ ($\tilde{h}$ could possibly be the same as $h$). If $\tilde{h}=h_1$ (resp.\ $\tilde{h}=h_2$), the  learner suffers a constant error as $x'$ can improve to the false positive $x_1$ (resp.\ $x_2$) when the target concept is $h_2$ (resp.\ $h_1$). Else, there must exist $\tilde{x}\in \mathcal{X}^+_{\tilde{h}}$ such that $h_1(\tilde{x})=0$, since $h_1$ was chosen to be minimally consistent (and likewise for $h_2$). $h(\tilde{x})=1$ in this case, and $x'$ can now falsely ``improve'' to $\tilde{x}$. Since the learner has no way of knowing from the sample whether the target is $h_1$ or $h_2$, it must suffer a constant error for any $h$ it selects from $\mathcal{H}$. 

\end{proof}

\section{Missing Proofs and Additional Definitions from Section \ref{sec:paclearn_graph-model} 
\label{app:graph}}

\subsection{Additional Definitions}

\begin{definition}[Shortest path]
\label{shortestpathdef}
    The shortest path metric \( d_G : V \times V \to [0, |V|+1) \) is defined as follows: 
    
    \[
    d_G(x, x') =
    \begin{cases}
    \min \left\{ k \ \middle| \ \exists \ (x_0=x, x_1, \dots, x_k=x') \subseteq V, \ 
    (x_{i-1}, x_i) \in E \ \forall i \in [k] \right\}, & \!\!\text{if path exists,} \\
    |V|+1, & \!\!\text{otherwise}
    \end{cases}
    \]
    
    \noindent Here, \( k \) is the length of the shortest path between \(x\) and \(x'\) in terms of the number of edges. If there is no path connecting \(x\) and \(x'\), the distance is defined as \( |V|+1 \).
    \noindent The shortest path metric satisfies the following properties:
    \begin{itemize}
        \item \textbf{Non-negativity:} \( d_G(x, x') \geq 0 \) for all \( x, x' \in V \), with \( d_G(x, x) = 0 \).
        \item \textbf{Symmetry:} \( d_G(x, x') = d_G(x', x) \) for all \( x, x' \in V \), since \( G \) is undirected.
        \item \textbf{Triangle inequality:} \( d_G(x, x') \leq d_G(x, z) + d_G(z, x') \) for all \( x, x', z \in V \).
    \end{itemize}
    
     \noindent Our results in Section \ref{sec:paclearn_graph-model} extend to the more general improvement function $\Delta(x) = \{x' \in V \ | \ d_G(x, x') \le r\}$ by applying our arguments to $G^r$, the $r$th power of $G$.
\end{definition}

\begin{definition}[Dominating set]
    \label{defDS}
    Let \( G = (V, E) \) be an undirected graph, where \( V \) is the set of vertices and \( E \subseteq V \times V \) is the set of edges. A subset of vertices \( S \subseteq V \) is called a dominating set if every vertex in \( V \) is either in \( S \) or adjacent to at least one vertex in \( S \). Formally, \( S \) is a dominating set if:
    \[
    \forall x \in V, \quad x \in S \ \text{or} \ \exists x' \in S \ \text{such that} \ (x, x') \in E.
    \]
\end{definition}

\subsection{Enabling Improvement Whenever It Helps}
\label{app:improvwhenhelps}

\begin{theorem}
    Let \( G = (V, E) \) be an undirected graph with \( n = |V| \) vertices, and let \( f^{\star}: V \to \{0, +1\} \) denote the ground truth labeling function. Define: \( V^+ = \{x \in V \mid f^{\star}(x) = +1\} \), the set of positive vertices. Define \( V^- = \{x \in V \mid f^{\star}(x) = 0\} \) the set of negative vertices, and \( N = \{x \in V^- \mid \exists x' \in V^+ \text{ such that } (x, x') \in E\} \) denote the set of negative vertices that have a positive neighbor. Let \( d_{\min}^N = \min_{x \in N} |\{x' \in V^+ \mid (x, x') \in E\}| \) denote the minimum number of positive neighbors of vertices in $N$.  Assume that the data distribution \(\mathcal{D}\) is uniform on \( V \). For any $\delta>0$, 
and training sample $S \overset{\text{i.i.d.}}{\sim} \mathcal{D}^m$ of size  $m =  O \left(\frac{n (\log n + \log \frac{1}{\delta})}{d_{\min}^N}\right)$, there exists a learner that  outputs a hypothesis $h$ such that $\textsc{Loss}^{\textsc{e}}_{\mathcal{D}}(h,f^{\star})=0$, with probability at least $1-\delta$ over the draw of $S$. Moreover, there exists a graph $G$ for which any learner that always outputs $h$ with $\textsc{Loss}^{\textsc{e}}_{\mathcal{D}}(h,f^{\star})=0$ for any $\mathcal{D},f^{\star}$ must see at least $\Omega\left(\frac{n}{d_{\min}^N} \log \frac{n}{d_{\min}^N}\right)$ labeled points in the training sample, with high constant probability.
\label{thm:enable-improvement}
\end{theorem}

\begin{proof}
    The proof of the upper bound is technically similar to the proof of Theorem \ref{DSSamplecomplexity}. Essentially, to ensure that $\textsc{Loss}^{\textsc{e}}=0$, we need to {\textit cover} all the vertices in $N$ by some vertices in $V^+$. The probability that any fixed node in $N$ is covered by a random sample can be lower-bounded in terms of $d_{\min}^N$ as
    \[
p_{\text{cover}}(x) \geq \frac{d_{\min}^N }{n}.
\]
 Using the same argument as in Theorem \ref{DSSamplecomplexity}, we obtain an upper bound of $m =  O \left(\frac{n (\log n + \log \frac{1}{\delta})}{d_{\min}^N}\right)$ on the sample complexity of the classifier $h$ which outputs exactly the positively labeled points in its sample as positive to guarantee that $\textsc{Loss}^{\textsc{e}}_{\mathcal{D}}(h, f^{\star}) = 0$ with probability at least $1-\delta$.

 To establish the lower bound, consider a graph $G=(V,E)$ with two types of nodes, i.e.\ $V=V_1\cup V_2$, $|V_1|=k$, $|V_2|=n-k$. $f^{\star}$ labels all nodes in $V_1$ as negative. Each node $x_i\in V_1$ has  $d_{\min}^N=\frac{n-k}{k}$ neighboring nodes $\Delta_i$ in $|V_2|$ and the sets of these neighbors are pairwise disjoint. Now, suppose our training sample $S$ does not contain any point in $\Delta_i$ for some $i\in[k]$. If the learned hypothesis $h$ predicts any point  $\Delta_i$ as positive, we have $h(x_i)\ne\{x_i\}$ but if $f^{\star}$ labels all points in $\Delta_i$ negative, $f^{\star}(x_i)=\{x_i\}$ and we incur loss corresponding to $x_i$. Similarly, if $h$ labels all points in $\Delta_i$ as negative then $h(x_i)=\{x_i\}$ but we can label $\Delta_i$ consistent with $S$ such that $f^{\star}(x_i)\ne\{x_i\}$.

\begin{figure}[ht]
\vskip 0.2in
\begin{center}
\centerline{\includegraphics[width=0.45\columnwidth]{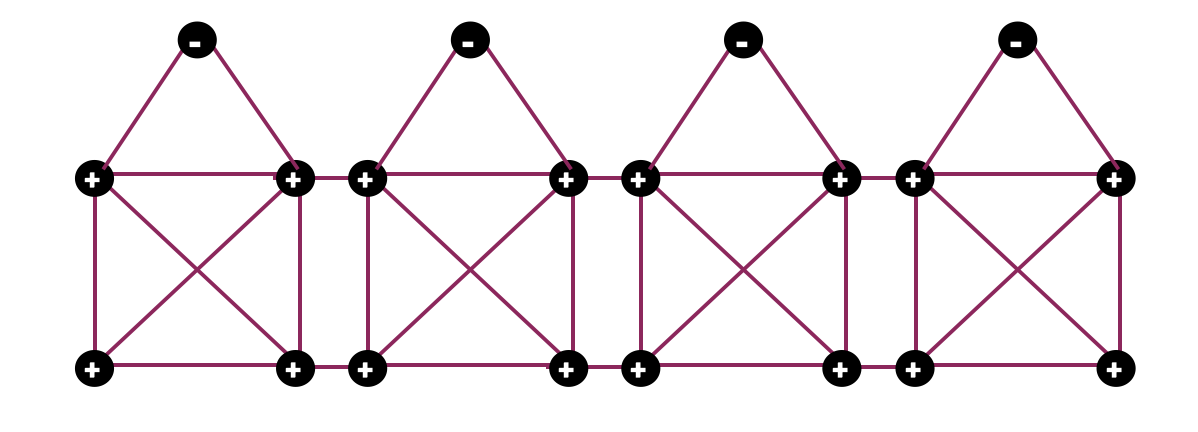}}
\caption{The graph $G$ used to establish our lower bound in Theorem \ref{thm:enable-improvement}.}
\label{posnegcl}
\end{center}
\vskip -0.2in
\end{figure}

Therefore it is sufficient  to determine a lower bound on the number of points required to ensure that every $\Delta_i$ has at least one of its vertices included in the training sample $S$. Using the standard coupon collector analysis, the number of trials needed to collect $k = \frac{n}{d_{\min}^N+1}$ coupons is $\Omega(k\log k)$ with high constant probability. 
\end{proof}

\begin{theorem}
    Let \( G = (V, E) \) be an undirected graph with \( n = |V| \) vertices, and let \( f^{\star}: V \to \{0, +1\} \) denote the ground truth labeling function. Define \( V^+ = \{x \in V \mid f^{\star}(x) = +1\} \), the set of positive vertices and $d_{\min}^+$ denote the minimum degree of a vertex in $V^+$ in the subgraph of $G$ induced by $V^+$. Define \( V^- = \{x \in V \mid f^{\star}(x) = 0\} \) the set of negative vertices, and \( N = \{x \in V^- \mid \exists x' \in V^+ \text{ such that } (x, x') \in E\} \) denote the set of negative vertices that have a positive neighbor. Let \( d_{\min}^N = \min_{x \in N} |\{x' \in V^+ \mid (x, x') \in E\}| \) denote the minimum number of positive neighbors of vertices in $N$.  Assume that the data distribution \(\mathcal{D}\) is uniform on \( V \). For any $\delta>0$, 
    and training sample $S \overset{\text{i.i.d.}}{\sim} \mathcal{D}^m$ of size  $m =  O \left(\frac{n (\log n + \log \frac{1}{\delta})}{\min\{d_{\min}^N,d_{\min}^+\}}\right)$, there exists a learner that  outputs a hypothesis $h$ such that $\textsc{Loss}_{\mathcal{D}}(h,f^{\star})=0$ and  $\textsc{Loss}^{\textsc{e}}_{\mathcal{D}}(h,f^{\star})=0$, with probability at least $1-\delta$ over the draw of $S$. Moreover, there exists a graph $G$ for which any learner that always outputs $h$ with $\textsc{Loss}_{\mathcal{D}}(h,f^{\star})=\textsc{Loss}^{\textsc{e}}_{\mathcal{D}}(h,f^{\star})=0$ for any $\mathcal{D},f^{\star}$ must see at least $\Omega\left(\max\{\frac{n}{d_{\min}^+} \log \frac{n}{d_{\min}^+}, \frac{n}{d_{\min}^N} \log{\frac{n}{d_{\min}^N}}\}\right)$ labeled points in the training sample, with high constant probability.
\label{thm:enable-improvement-and-zero-loss}
\end{theorem}
\begin{proof}
    See Theorems \ref{DSSamplecomplexity} and \ref{thm:enable-improvement}.
\end{proof}

\subsection{Teaching a Risk-Averse Student}\label{app:teaching-student}

Proof of Theorem \ref{thm:dominating-set-teaching}

\begin{proof}
    Since \( S^+ \) is a dominating set of \( G^+ =(V^+,E^+)\), for any \( x \in V^+ \), either \( x \in S^+ \) or there exists \( x' \in S^+ \) such that \( (x, x') \in E^+ \). In the first case, $h_{S^+}(x)=1$ and therefore $\Delta_{h_{S^+}}(x)=\{x\}$. Thus,  $h_{S^+}(x)=1=f^{\star}(x)$ implies that $\textsc{Loss}(x;h_{S^+},f^{\star})=0$ in this case.
    In the second case, \( x \notin S^+ \), but there exists a neighbor $\tilde{x}\in \Delta(x)$ such that $\tilde{x}\in S^+$ by the definition of $S^+$. Thus, for any point  $x'\in\Delta_{h_{S^+}}(x)\subseteq S^+$, we have that $h_{S^+}(x')=1=f^{\star}(x')$, ensuring that $\textsc{Loss}(x;h_{S^+},f^{\star})=0$ in this case as well.

    For \( x \in V \setminus V^+ \), if \( x \) has no positive neighbors in $S^+$, \( \Delta_{h_{S^+}}(x) = \{x\} \) because there is no neighboring vertex \( x' \in \Delta(x) \) that would induce a reaction. Thus, $h_{S^+}(x)=0=f^{\star}(x)$ in this case, implying the loss $\textsc{Loss}(x;h_{S^+},f^{\star})$ on $x$ is zero.
    
    Finally, if \( x \in V \setminus V^+\) has positive neighbors contained in the dominating set, i.e., \( \Delta(x) \cap S^+ \ne \emptyset\), then \( h_{S^+}(x') = +1 \). 
    The reaction set \( \Delta_{h_{S^+}}(x) \) ensures that \( x \) moves to one of these neighbors. Specifically, the reaction set allows \( x \) to improve and move to a neighboring vertex \( x' \in \Delta(x) \cap S^+ \) such that \( f^{\star}(x') = +1 \). Thus, \( h(x') = f^{\star}(x') = +1 \) for any $x'\in\Delta_{h_{S^+}}(x)$ implying $\textsc{Loss}(x;h_{S^+},f^{\star})=0$.
\end{proof}

\clearpage
\section{Evaluation: Supplementary Details}
\label{app:sec_eval}

This section includes supplementary details on the datasets and classifiers used, how improvement is done and results of the empirical evaluations.

\subsection{Datasets}
\label{app:sec_datasets}

We utilize three real-world datasets: the Adult Income dataset from UCI and the Open University Learning Analytics Dataset (OULAD) and Law School datasets sourced from \citet{le2022surveygit}. 
The preprocessing steps for all the datasets, similar to those described in \citet{le2022surveygit}, include removing missing data and applying label encoding to categorical variables.
In addition to the real-world datasets, we generate an 8-dimensional synthetic dataset with increased separability (\textit{class\_sep} = \(4\)) using the \textit{make\_classification} function from Scikit-learn. We clean the dataset by removing duplicates and outliers, with Z-scores applied with thresholds (\(0.9\) for class \(0\) and \(0.8\) for class \(1\)). The cleaned synthetic dataset is then balanced using SMOTE \citep{smote2002} to ensure class balance.

\begin{figure}[!ht]
    \centering
    \begin{subfigure}[t]{0.2\linewidth}
        \centering
        \includegraphics[width=\linewidth]{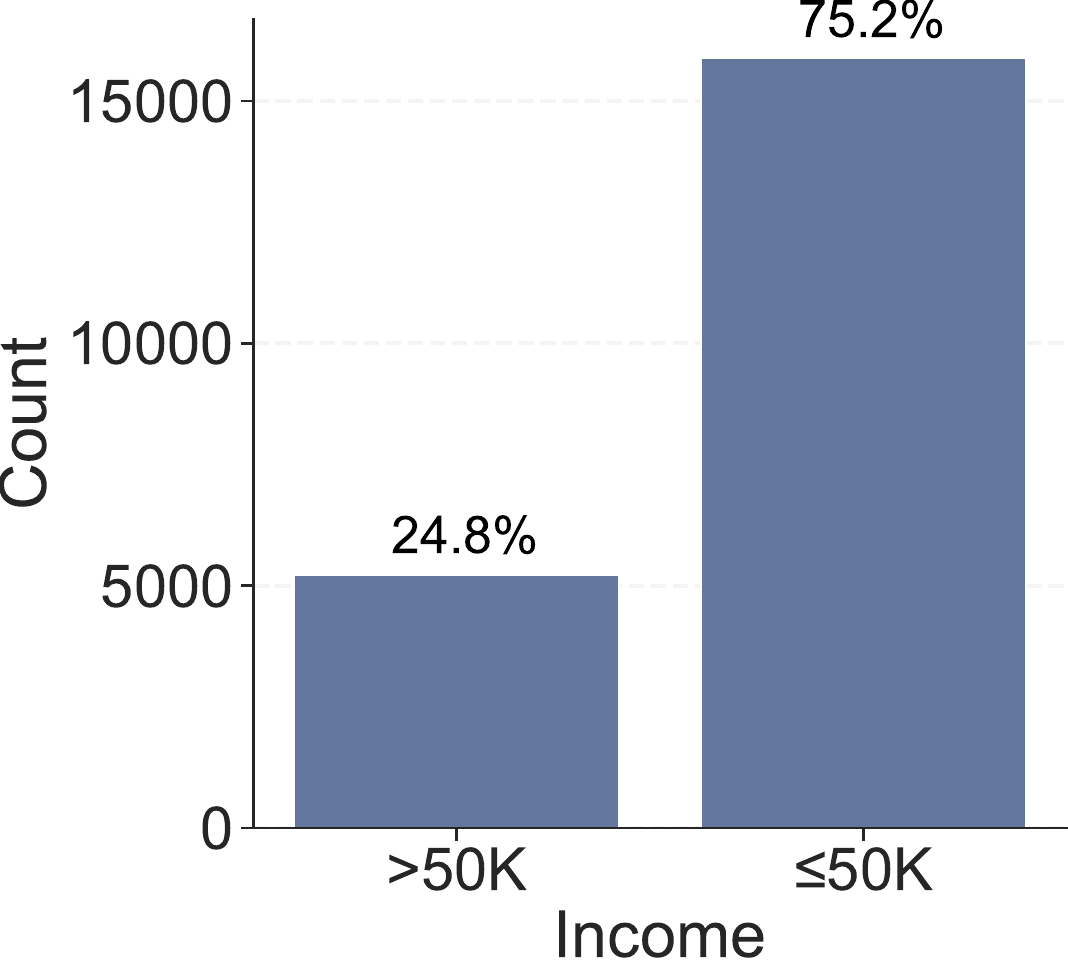}
    \caption{Adult}
    \label{fig:adult_y_hist}
    \end{subfigure}
    \hfill
    \begin{subfigure}[t]{0.2\linewidth}
        \centering
        \includegraphics[width=\linewidth]{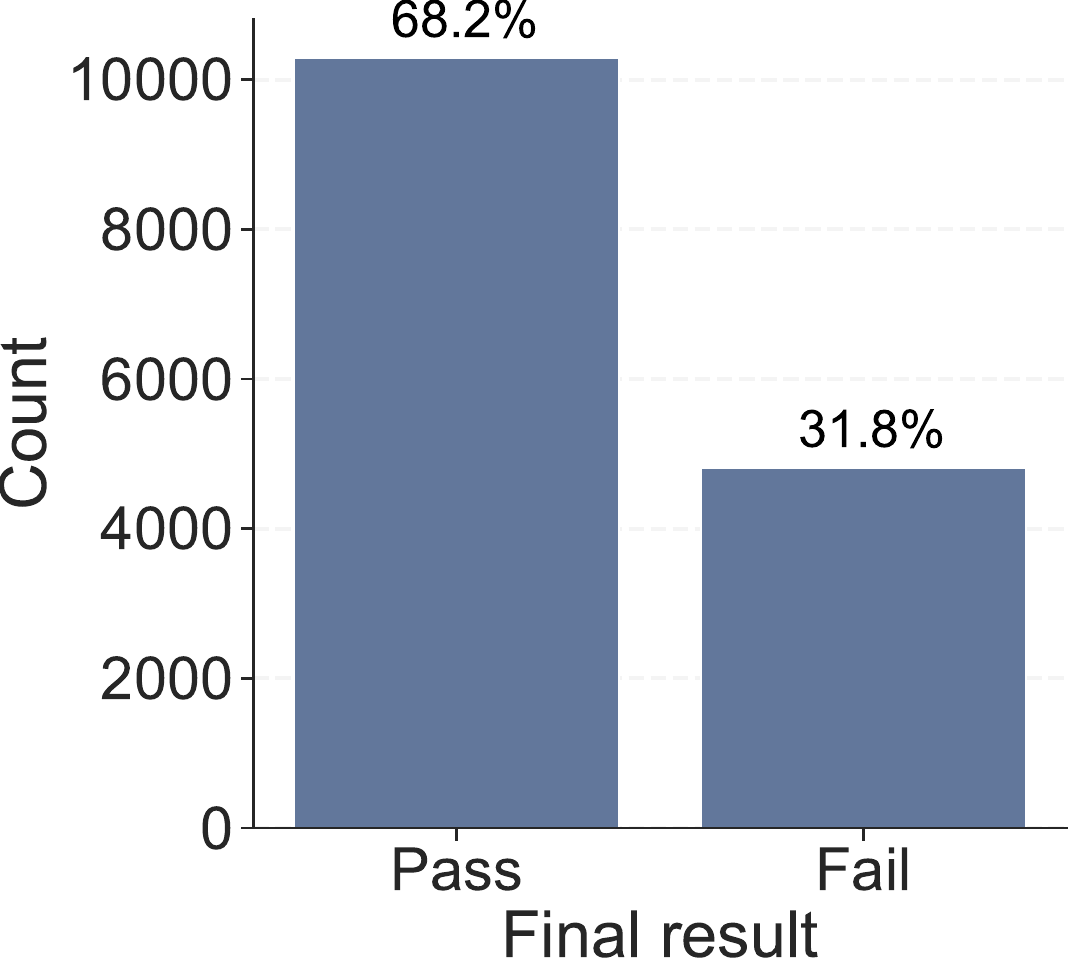}
    \caption{[OULAD}
    \label{fig:oulad_y_hist}
    \end{subfigure}
    \hfill
    \begin{subfigure}[t]{0.2\linewidth}
        \centering
        \includegraphics[width=\linewidth]{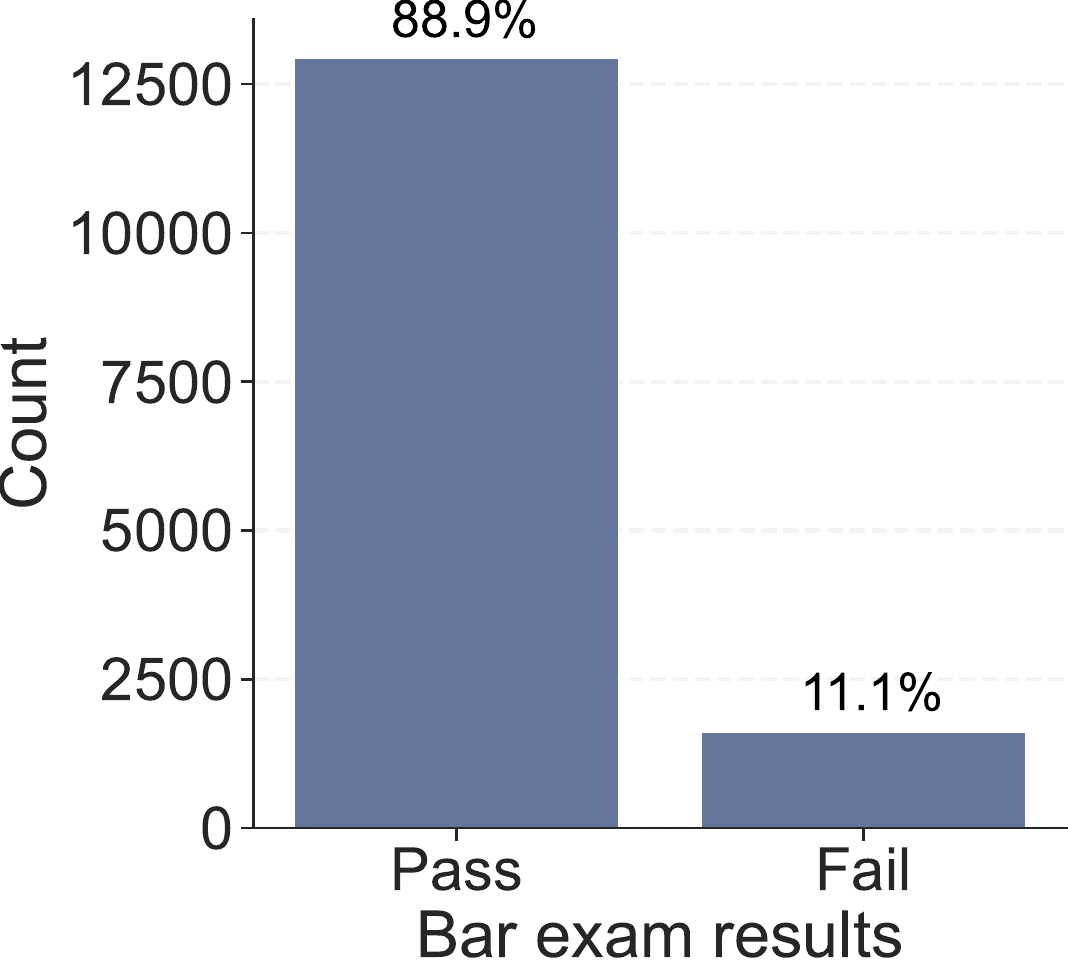}
    \caption{Law School}
    \label{fig:law_y_hist}
    \end{subfigure}
    \hfill
    \begin{subfigure}[t]{0.2\linewidth}
        \centering
        \includegraphics[width=\linewidth]{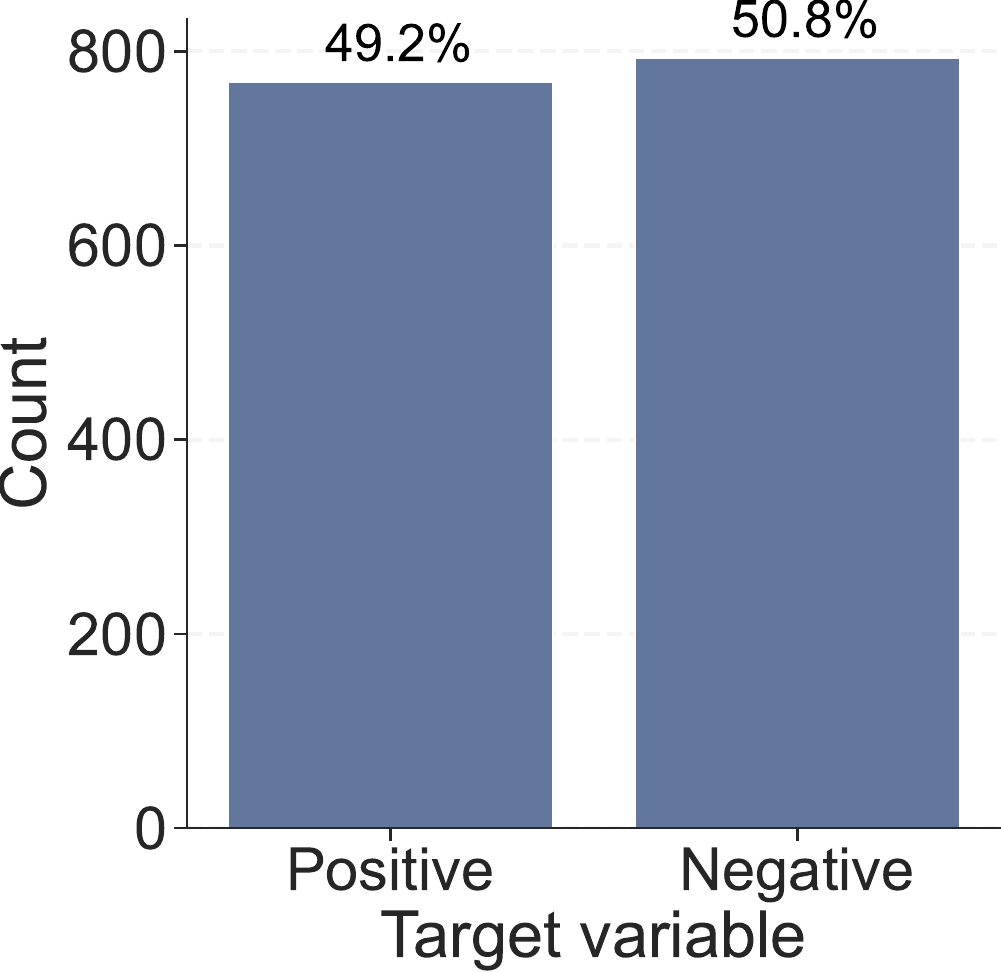}
    \caption{Synthetic}
    \label{fig:syn2_y_hist}
    \end{subfigure}
    
    \caption[Target variable distributions of synthetic and real-world train datasets]{Target variable distributions of synthetic and real-world train datasets: (\subref{fig:adult_y_hist}) Adult, (\subref{fig:oulad_y_hist}) OULAD,
    (\subref{fig:law_y_hist}) Law School, and (\subref{fig:syn2_y_hist}) Synthetic datasets.}
    \label{fig:orig_y_hist}
\end{figure}

\begin{figure}[htbp]
    \centering
    \includegraphics[width=0.9\linewidth]{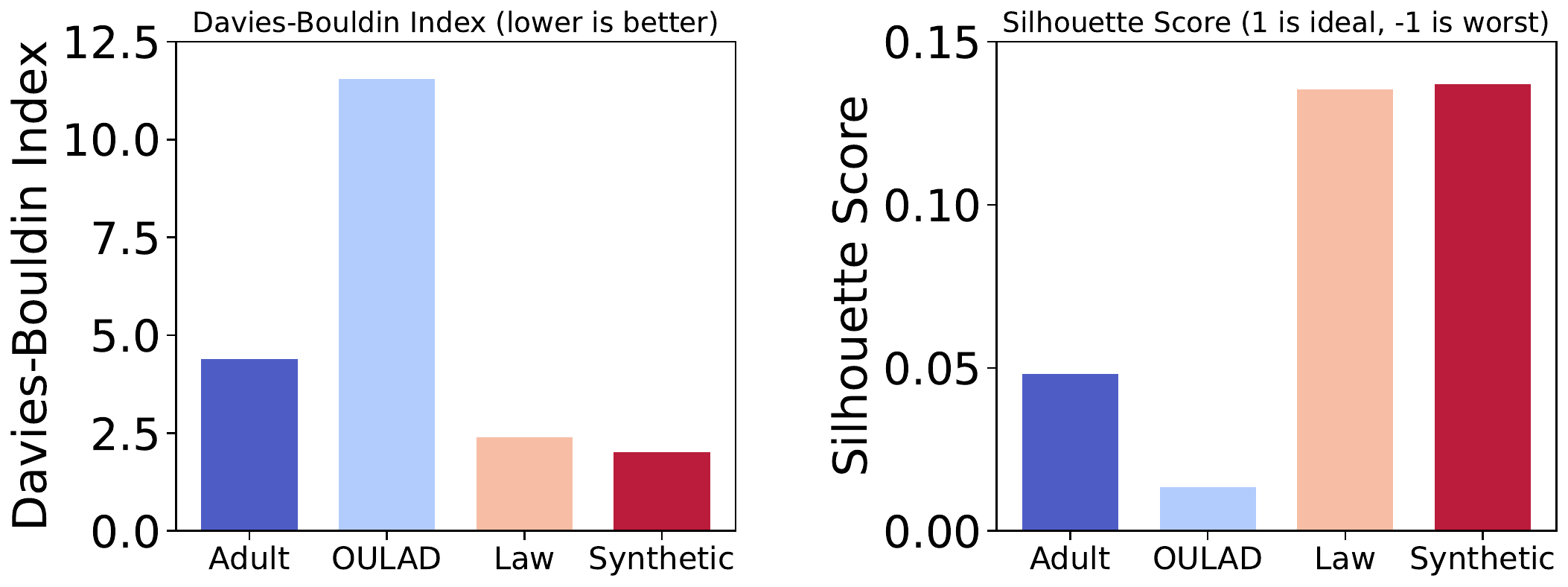}
    \caption[Inspection of clusteredness and class separation]{Inspection of clusteredness and class separation using Davies–Bouldin index \citep{davies_bouldin} and  Silhouettes scores \citep{ROUSSEEUW198753,silhouette}\label{fig:jumbleness}.}
\end{figure}

\begin{figure}[t!]
    \centering
    \begin{subfigure}[t]{0.45\linewidth}
        \centering
        \includegraphics[width=\linewidth]{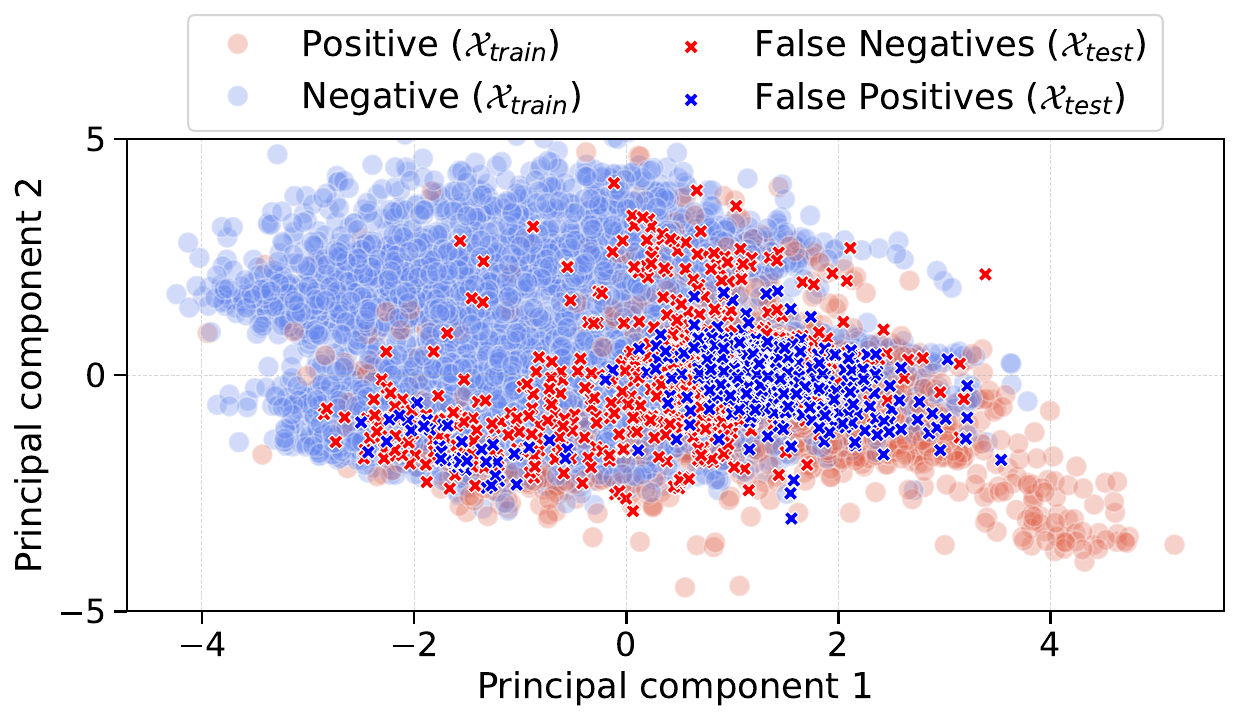}
    \caption{Adult}
    \label{fig:adult_knn}
    \end{subfigure}
    \hfill
    \begin{subfigure}[t]{0.45\linewidth}
        \centering
        \includegraphics[width=\linewidth]{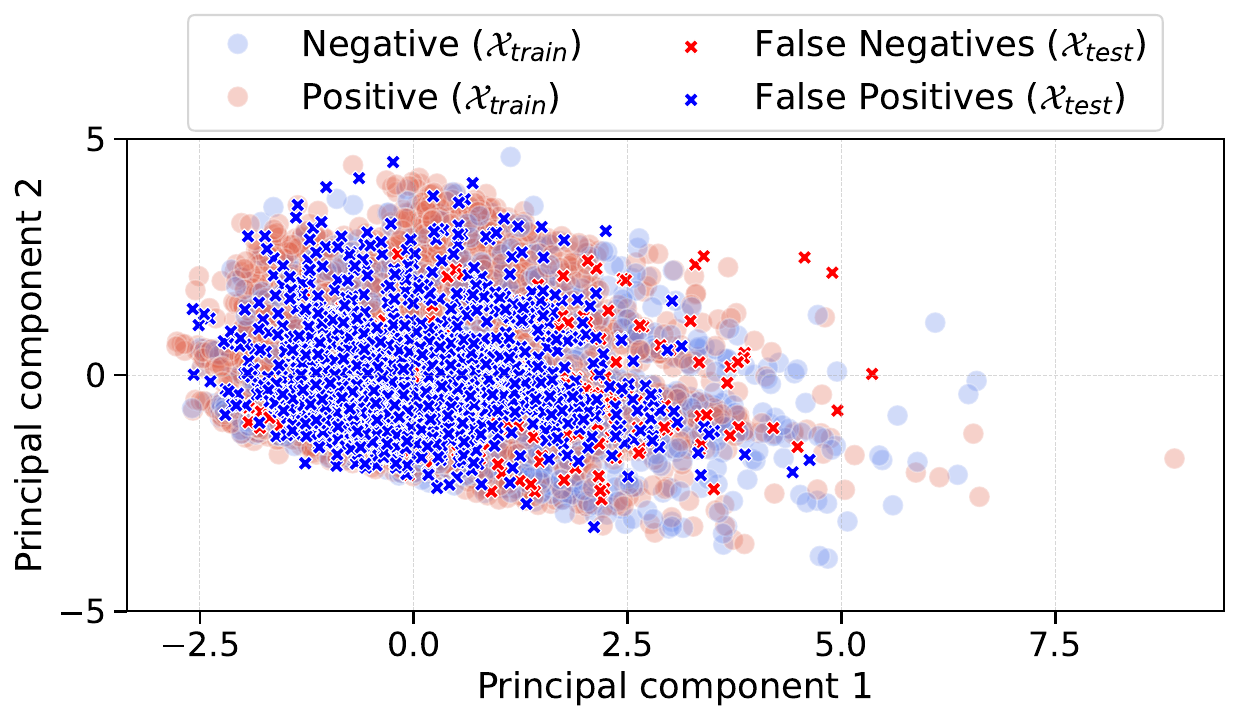}
    \caption{OULAD}
    \label{fig:oulad_knn}
    \end{subfigure}
    \hfill
    \begin{subfigure}[t]{0.45\linewidth}
        \centering
        \includegraphics[width=\linewidth]{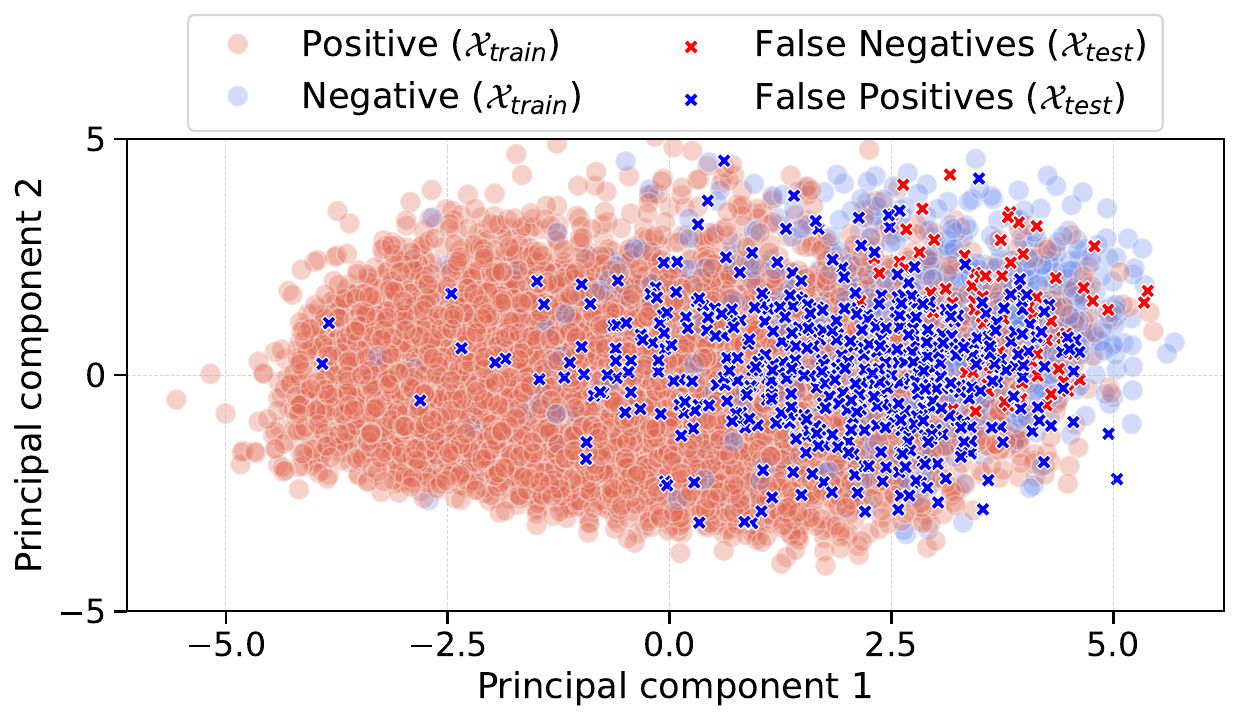}
    \caption{Law school}
    \label{fig:law_knn}
    \end{subfigure}
    \hfill
    \begin{subfigure}[t]{0.45\linewidth}
        \centering
        \includegraphics[width=\linewidth]{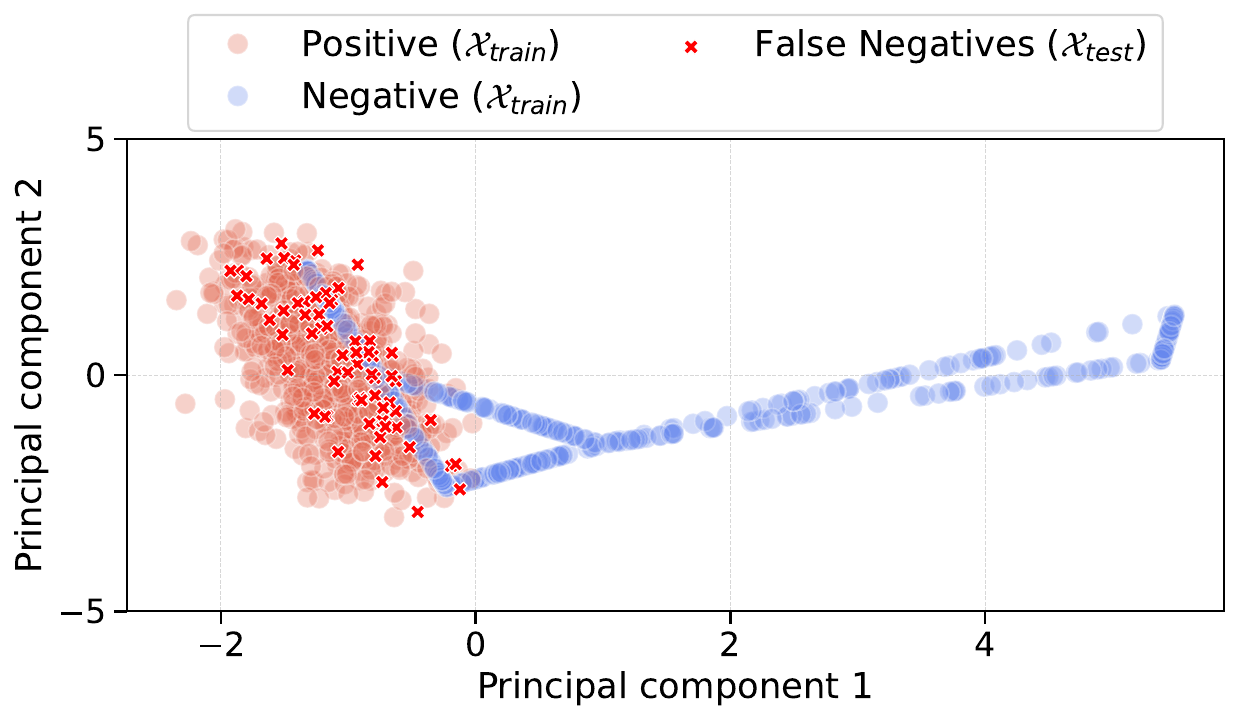}
    \caption{Synthetic}
    \label{fig:syn2_knn}
    \end{subfigure}
    
    \caption[Scatter plots of the two principal components]{Scatter plots of the two principal components of the training data and of the \(k\)-NN misclassification on test data for the (\subref{fig:adult_knn}) Adult (\(k\)-NN error: \(0.1670\), FNR: \(0.4610\), FPR: \(0.0678\)), (\subref{fig:oulad_knn}) OULAD (\(k\)-NN error: \(0.3291\), FNR: \(0.1195\), FPR: \(0.7645\)), (\subref{fig:law_knn}) Law School (\(k\)-NN error: \(0.1010\), FNR: \(0.0180\), FPR: \(0.7875\)), and  (\subref{fig:syn2_knn}) Synthetic (\(k\)-NN error: \(0.1016\), FNR: \(0.1960\), FPR: \(0.000\)) datasets.}
    \label{fig:orig_knn}
\end{figure}

Statistical details of the datasets, including test/train sizes and number of features, are in Table~\ref{tab:datasets_info}.  We examine the structural variations within datasets to gain deeper insights into how the characteristics influence the impact of improvements on error drop rates. Figure~\ref{fig:orig_y_hist} highlights the target distribution across training datasets: the Adult dataset has a higher proportion of negative examples, whereas the OULAD and Law School datasets have a higher percentage of positive examples. The synthetic dataset, by contrast, is balanced.
Figure~\ref{fig:jumbleness} and \ref{fig:orig_knn} illustrate dataset separability properties, showing that the synthetic dataset (\(k\)-NN error: \(0.1016\)) and the Law School dataset (\(k\)-NN error: \(0.1010\)) have the highest separability. However, as Figure~\ref{fig:orig_knn} shows, the Adult and synthetic datasets exhibit the lowest false positive (FP) outlier rates.
\renewcommand{\arraystretch}{1.0} 
\setlength{\abovecaptionskip}{3.6pt}
\setlength{\belowcaptionskip}{3.6pt}
\begin{table}[htbp]
\scriptsize
\centering
\begin{tabular}{p{1.8cm}p{3.7cm}p{0.6cm}p{1.7cm}p{6cm}}
    \toprule
    \textbf{Dataset} & \textbf{Target variable} & \(d\) & \textbf{Train/Test} & \textbf{Improvable features} \\ 
    \midrule
    Adult & \(\{1(>50K), \ 0(\leq50K)\}\) & \(14\) & \(21113/9049\) & \{``hours-per-week, capital-gain, capital-loss, fnlwgt, educational-num, workclass, education, occupation''\} \\ 
    OULAD & \(\{1(\text{pass}), \ 0(\text{fail})\}\) & \(11\) & \(15093/6469\) & \{``code\_module, code\_presentation, imd\_band, highest\_education, num\_of\_prev\_attempts, studied\_credits''\} \\ 
    Law school & \(\{1(\text{pass}), \ 0(\text{fail})\}\) & \(11\) & \(14558/6240\) & \{``decile1b, decile3, lsat, ugpa, zfygpa, zgpa, fulltime, fam\_inc, tier''\} \\ 
    Synthetic & \(\{1(\text{positive}), \ 0(\text{negative})\}\) & \(8\) & \(1561/669\) & \{all features are used\} \\ 
    \bottomrule
\end{tabular}  
\caption{Details of the tabular datasets, both synthetic and real, used in the experiments.}
\label{tab:datasets_info}
\end{table}
\clearpage
\begin{table}[h]
    \centering
    \scriptsize
    \setlength{\tabcolsep}{3pt} 
    \begin{tabular}{llllll}
        \toprule
        \textbf{Dataset} & \textbf{DTC1} & \textbf{DTC2} & \textbf{RFC1} & \textbf{RFC2} & \textbf{XGB} \\
        \midrule
        Adult      & \(0.999967 \pm 0.000049\) & \(0.999967 \pm 0.000049\) & \(0.999934 \pm 0.000079\) & \(0.999967 \pm 0.000049\) & \(0.999967 \pm 0.000049\) \\
        Law        & \(1.000000 \pm 0.000000\) & \(1.000000 \pm 0.000000\) & \(1.000000 \pm 0.000000\) & \(1.000000 \pm 0.000000\) & \(0.999952 \pm 0.000071\) \\
        OULAD      & \(1.000000 \pm 0.000000\) & \(1.000000 \pm 0.000000\) & \(1.000000 \pm 0.000000\) & \(1.000000 \pm 0.000000\) & \(1.000000 \pm 0.000000\) \\
        Synthetic & \(1.000000 \pm 0.000000\) & \(1.000000 \pm 0.000000\) & \(1.000000 \pm 0.000000\) & \(1.000000 \pm 0.000000\) & \(1.000000 \pm 0.000000\) \\
        \bottomrule
    \end{tabular}
    \caption[Accuracy score of the \(f^\star\) models]{Accuracy score of the \(f^\star\) models when trained and tested on \(\mathcal{S}_{T}\) across different datasets.}
    \label{tab:fstarclassifiers_gen_acc}
\end{table}
\begin{table}[h]
    \centering
    \scriptsize
    \setlength{\tabcolsep}{5pt} 
    \begin{tabular}{llllll}
        \toprule
        \textbf{Dataset} & \textbf{DTC1 (LOO)} & \textbf{DTC2 (LOO)} & \textbf{RFC1 (LOO)} & \textbf{RFC2 (LOO)} & \textbf{XGB (LOO)} \\
        \midrule
        Adult      & \(0.8084 \pm 0.0044\) & \(0.8053 \pm 0.0045\) & \(0.8541 \pm 0.0040\) & \(0.8545 \pm 0.0040\) & --- \\
        Law        & \(0.8507 \pm 0.0048\) & \(0.8428 \pm 0.0049\) & \(0.8962 \pm 0.0041\) & \(0.8972 \pm 0.0041\) & \(0.8894 \pm 0.0043\) \\
        OULAD      & \(0.5941\pm 0.0066 \) & \(0.5952 \pm 0.0066 \) & \(0.6684 \pm 0.00663\) & \(0.6689 \pm 0.0063\) & --- \\
        Synthetic & \(0.9955 \pm 0.0028\) & \(0.9951 \pm 0.0029\) & \(0.9969 \pm 0.0023\) & \(0.9973 \pm  0.0021\) & \(0.9955 \pm 0.0028\) \\
        \bottomrule
    \end{tabular}
    \caption[Average leave one out (LOO) score of the \(5, f^\star\) models]{Average leave one out (LOO) score of the \(5, f^\star\) models on \(\mathcal{S}_{T}\) across different datasets.}
    \label{tab:fstarclassifiers_loo}
\end{table}

\subsection{Classifiers}\label{app:sec_classifiers}

In all experiments we set the random seed to \(42\) to ensure reproducibility and consistency across all runs. 
All experiments were conducted on a laptop computer with the following hardware specifications: \(2.6\)-GHz 6-Core Intel Core i7 processor, \(16\) GB of \(2400\)-MHz DDR4 RAM, and an Intel UHD Graphics \(630\) graphics card with \(1536\) MB of memory. Below are supplementary details about the classification models used.
Below are supplementary details about the classification models used.

\subsubsection{The \texorpdfstring{\(f^\star\) model} .} 

The function \( f^\star \) served as the ground truth labeler, assessing whether the agent's modifications led to a successful improvement. We evaluated five standard machine learning binary classification models, each achieving near \(100\%\) accuracy when trained and tested on \(\mathcal{S}_{T}\) (see Table~\ref{tab:fstarclassifiers_gen_acc}). These models include two decision tree classifiers (DTC1 and DTC2), two random forest classifiers (RFC1 and RFC2), and a gradient boosting classifier (XGB). Descriptions of these models are provided below.
\begin{enumerate}
    \item \textbf{Model \(f^{\star}_{1}\)} (DTC1): A decision tree classifier with the following hyperparameters: criterion = ``entropy", min\_samples\_split = \(2\), min\_samples\_leaf = \(1\), and random\_state = \(42\).
    \item \textbf{Model \(f^{\star}_{2}\)} (DTC2): A decision tree classifier with the following hyperparameters: criterion = ``gini", min\_samples\_split = \(2\), min\_samples\_leaf = \(1\), and random\_state = \(42\).
    \item \textbf{Model \(f^{\star}_{3}\)} (RFC1): A random forest classifier with default settings and random\_state = \(42\).
    \item \textbf{Model \(f^{\star}_{4}\)} (RFC2): A random forest classifier with the following hyperparameters:\\ n\_estimators = \(500\), min\_samples\_split = \(2\), min\_samples\_leaf = \(1\), max\_features = ``sqrt", bootstrap = True, oob\_score = True, and random\_state = \(42\).
    \item \textbf{Model \(f^{\star}_{5}\)} (XGB): A gradient boosting classifier with the following hyperparameters: n\_estimators = \(500\), max\_depth = \(50\), learning\_rate = \(0.088\), min\_child\_weight = \(2\), gamma = \(0.088\), subsample=\(0.9\), and random\_state = \(42\).
\end{enumerate}
We define the ground truth labeler \(f^\star\) either as a singular near-\(100\%\) accuracy model (see Table~\ref{tab:fstarclassifiers_gen_acc}) or as an agreement among multiple near-\(100\%\) accuracy models.
\paragraph{The multi-defined \(f^\star\) model.} Although the five \(f^\star\) models described above achieve nearly \(100\%\) accuracy when trained and tested on \(\mathcal{S}_{T}\), we assessed their generalization using the leave-one-out (LOO) validation score. The observed differences in LOO validation scores (Table~\ref{tab:fstarclassifiers_loo}), despite similar and high accuracy scores (Table~\ref{tab:fstarclassifiers_gen_acc}), highlight potential generalization gaps. To account for this, we employ a multi-defined \(f^\star\) model to validate the experimental results.  \\

For a given data point \(x\), the five models: \(f^{\star}_{1}(x), f^{\star}_{2}(x), f^{\star}_{3}(x), f^{\star}_{4}(x)\) and \(f^{\star}_{5}(x)\) each make a prediction for the label of the data point. Based on these predictions, we define a boolean agreement mask \(M(x)\) that checks whether all four models agree on the prediction:
\[
M(x) = \mathbbm{1}( f^{\star}_{1}(x) = f^{\star}_{2}(x) = f^{\star}_{3}(x) = f^{\star}_{4}(x) = f^{\star}_{5}(x) )
\]
where \(\mathbbm{1}(\cdot)\) is the indicator function that outputs \(1\) if all four models agree, and \(0\) otherwise. 
Using this agreement mask, we define the ground truth labeling function \(f^{\star}(x)\) as follows:
\begin{equation}
    f^{\star}(x) = \begin{cases} 
        f^{\star}_{1}(x), & \text{if } M(x) = 1 \text{ (i.e., full agreement)} \\
        0, & \text{otherwise} 
    \end{cases}
\end{equation}

\paragraph{The singularly-defined \(f^\star\) model.} 
Alternatively, we define the labeling function  \(f^\star(x)\) using a single near-\(100\%\) accuracy model trained and tested on \(\mathcal{S}_{T}\), selected from the set \(\{f^{\star}_{1}(x), f^{\star}_{2}(x),\) \(f^{\star}_{3}(x),\) \(f^{\star}_{4}(x), f^{\star}_{5}(x)\}\).
Unless otherwise stated , all experimental results were obtained using the DTC2 model (\(f^{\star}_{2}(x)\)) as the designated singularly-defined \(f^\star(x)\) function.

\subsubsection{The decision-maker's model \texorpdfstring{(\(h\))}.}  

We trained two-layer neural networks, denoted as \(h\) functions, using PyTorch with Adam optimizer with a learning rate of \(0.001\) and a batch size of \(64\). These \(h\) functions generate decisions for the test set agents. In cases where the test agent receives a negative classification, they can, if within budget, improve their feature values to get the desired classification from the \(h\) function. Table~\ref{tab:h_classifiers} summarizes the performance metrics of the \(f^\star\) and \(h\) model functions, demonstrating their varied performance across the datasets.

Since the empirical setup evaluates the impact of improvement on \(h\)'s error drop rates, we vary the loss functions we train the model \(h\) function with. We use the standard binary cross entropy loss (BCE) and the risk-averse weighted-BCE (wBCE) loss functions defined in Equation~\ref{eq:losses}. In particular, because only negatively classified test-set agents improve, improvement (\(x'\)), if successful (that is, \(f^{\star}(x') = h(x') = 1\)) reduces the false negative rate and turns true negatives into true positives. On the other hand, when unsuccessful (that is, \(f^{\star}(x') = 0 \ \text{and} \ h(x') = 1\)), it increases the false positive rate by turning true negatives into false positives and false negatives into false positives. 

The model trained with the weighted-BCE loss corresponds to a more risk-averse classifier that penalizes the false positive (FP) errors more heavily than the false negative (FN) one, creating a more compact positive agreement region that ensures more successful improvements. We prioritize minimizing FPs by ensuring the false positive to false negative weight ratio \(\frac{w_{\textrm{FP}}}{w_{\textrm{FN}}} > 1 \) is high, for example, \(\frac{4.4}{0.001} = 4400\) for the adult dataset. Another form of risk-averse classification we consider is only classifying an agent as positive \textit{iff} the probability of being positive is high. That is to say, we use the standard threshold \(0.5\) for the standard classifier and a higher threshold \(0.9\) for a more risk-averse classifier.
\begin{table}[ht!]
    \centering
    \renewcommand{\arraystretch}{1.1} 
    \setlength{\tabcolsep}{7pt} 
    \begin{tabular}{l l l l l l}
        \toprule
        \textbf{Dataset} & \textbf{Model (kind)} & \textbf{Accuracy} & \textbf{Precision} & \textbf{Recall} & \textbf{F1 Score} \\
        \midrule
        Adult & 2-layer neural network  (\(h(x)\)) & 0.841087 & 0.699811 & 0.647677 & 0.672736 \\
        Law & 2-layer neural network (\(h(x)\)) & 0.901282 & 0.911691 & 0.984731 & 0.946805 \\
        OULAD & 2-layer neural network (\(h(x)\)) & 0.678003 & 0.698609 & 0.919853 & 0.794109 \\
        Synthetic & 2-layer neural network (\(h(x)\)) & 0.994021 & 1.000000 & 0.988473 & 0.994203 \\
        \hline
    \end{tabular}
    \caption[Average performance of the standard  models functions \(h\)]{Average performance of the standard  models functions \(h\) across different datasets' test sets \(\mathcal{S}_{\textrm{test}}\) when test-set agent cannot improve that is, \(r=0\).}
    \label{tab:h_classifiers}
\end{table}

\newpage

\subsection{Agents Improvement}
\label{app:sec_improve}

Given the feature vector of a negatively classified test-set agent, \(x_{\text{orig}}\) and it's negative label  \(h(x_{\text{orig}})\), the loss function \(\mathcal{L}\) (BCE or wBCE), improvement budget \(r\), step size \(\alpha\), number of iterations \(T\) and set of indices of improvement features \(S\), compute the agent's improvement features. For each dataset, we predefine the improvable features that the agents can change in order to get a desirable (positive) model outcome (see Table~\ref{tab:datasets_info}). We vary the improvement budget in the empirical setup to so as to assess the impact of improvement on the the error drop rates. 

Below are the steps of the improvement algorithm we used to compute each agent's improvement features.
\paragraph{Initialization:}
\[
x'_{(0)} = x_{\textrm{orig}}
\]

\paragraph{Iterative updates:}
For $t = 0, 1, \dots, T-1$:
\begin{enumerate}
    \item Compute the gradient of the loss $\mathcal{L}$ with respect to the agent's updates $x'_{(t)}$:
    \[
    \mathbf{g}_{(t)} = \nabla_{x'_{(t)}} \mathcal{L}\Big(h\big(x'_{(t)}\big), h\big(x_{\textrm{orig}}\big)\Big)
    \]
    \item Update the improvement features by taking a step in the direction of the sign of the gradient:
    \[
    \rho_{(t)}[i] =
    \begin{cases} 
        \alpha \cdot \text{sign}(\mathbf{g}_{(t)}[i]), & \text{if} \ i \in S \\
        0, & \text{otherwise}
    \end{cases}, \quad  \forall i \in [d]
    \]
    \[
    x'_{(t+1)} = x'_{(t)} + \rho_{(t)}
    \]
    \item Project the updated improvement features back onto the $r$-ball around the original features  \(x_{\text{orig}}\):
    \[
    x'_{(t+1)} = \big(x_{\textrm{orig}} + \textrm{clip}_{[-r, r]}(x'_{(t+1)} - x_{\textrm{orig}})\big)
    \]
\end{enumerate}
\paragraph{Improvement vector:} After $T$ iterations, the final agent's improvement is given by:
\[
x' = x'_{(T)}
\]

\subsection{Evaluation Results}
\label{app:sec_results}

Following the key insights mentioned in Section~\ref{sec:paclearn_experiments}, below are the detailed observations from the experimental evaluation and the supplementary figures of the experimental results. 

\paragraph{Effect of dataset characteristics.} For all datasets we consider, Figure~\ref{fig:main_thresh0.5} shows that as the improvement budget increases, error rates drop significantly, particularly when agents improve in response to a risk-averse model trained using the \(\mathcal{L}_{\textrm{wBCE}}\) loss function. Dataset characteristics notably influence performance. For instance, as shown in Figure~\ref{fig:jumbleness}, the Law school and Synthetic datasets exhibit the highest separability and require relatively less risk aversion to achieve substantial and close to zero error reductions. Among all datasets, the Synthetic dataset shows a sharper error decline, reaching zero as the improvement budget increases (refer to Figure~\ref{fig:main_thresh0.5}). Furthermore, as depicted in Figure~\ref{fig:orig_knn}, the Adult and Synthetic datasets demonstrate the lowest \(k\)-NN test-set false positive rates (FPRs) of \(0.0678\) and \(0.000\), respectively, compared to the OULAD and Law School datasets. As seen in Figure~\ref{fig:main_thresh0.5}, for both datasets, the error drops close to \(0\) as \(r\) increases.

\begin{figure}[htb!]
    \centering
    
    \begin{subfigure}[t]{0.42\linewidth}
        \centering
        \includegraphics[width=\linewidth]{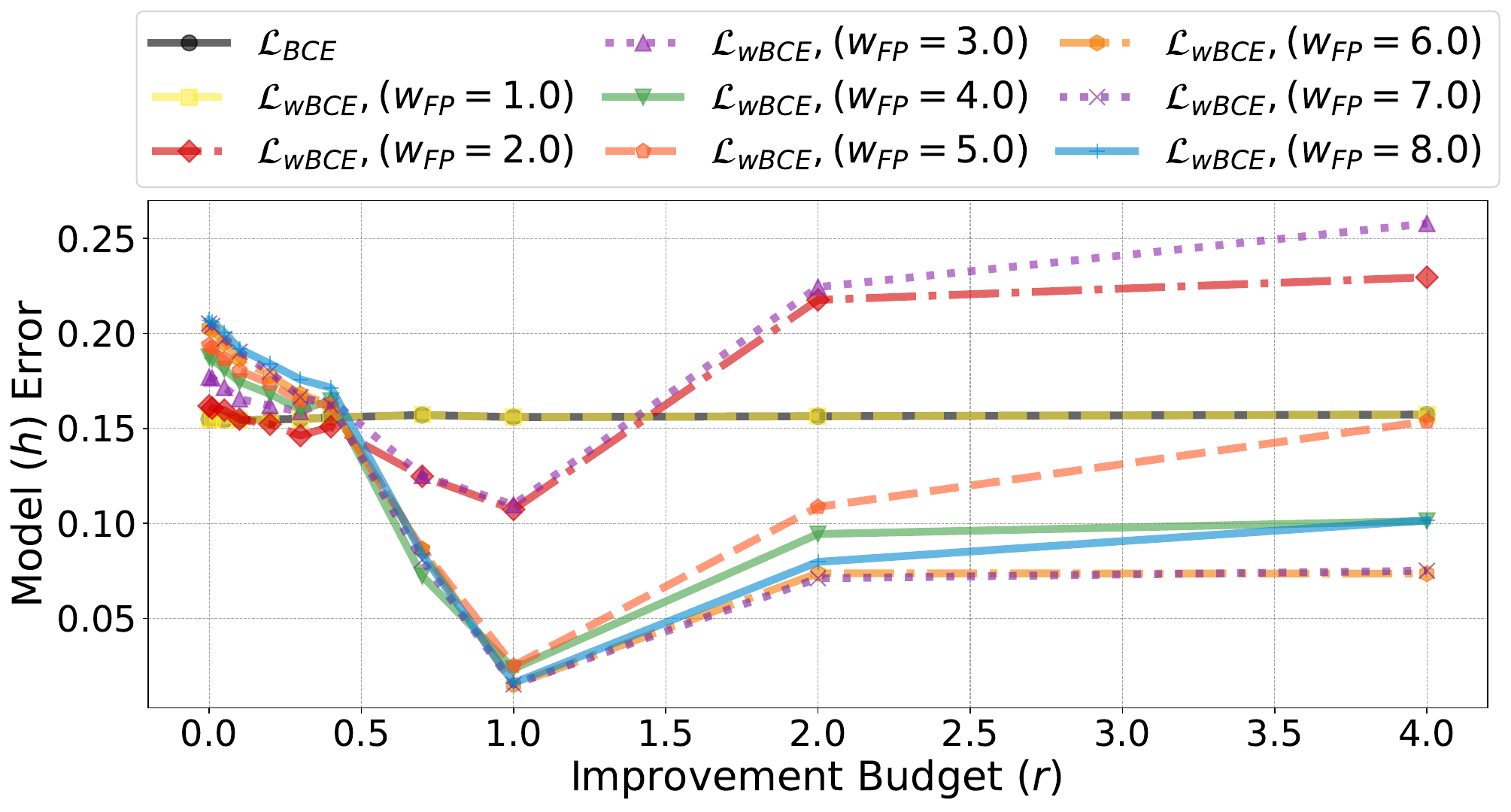}
    \caption{Adult \(\big(\mathcal{L}_{\textrm{wBCE}} (w_{\textrm{FN}}=1.0)\big)\)}
    \label{fig:app_adult_fn1_th0.5}
    \end{subfigure}
    ~
    \begin{subfigure}[t]{0.42\linewidth}
        \centering
        \includegraphics[width=\linewidth]{paclearn_improves/paclearn_figures/lfconlyw_0.5_error_droprate_linfx_adult.pdf}
    \caption{Adult \(\big(\mathcal{L}_{\textrm{wBCE}} (w_{\textrm{FN}}=0.001)\big)\)}
    \label{fig:app_adult_fnz_th0.5}
    \end{subfigure}
    \vskip\baselineskip
    \begin{subfigure}[t]{0.42\linewidth}
        \centering
        \includegraphics[width=\linewidth]{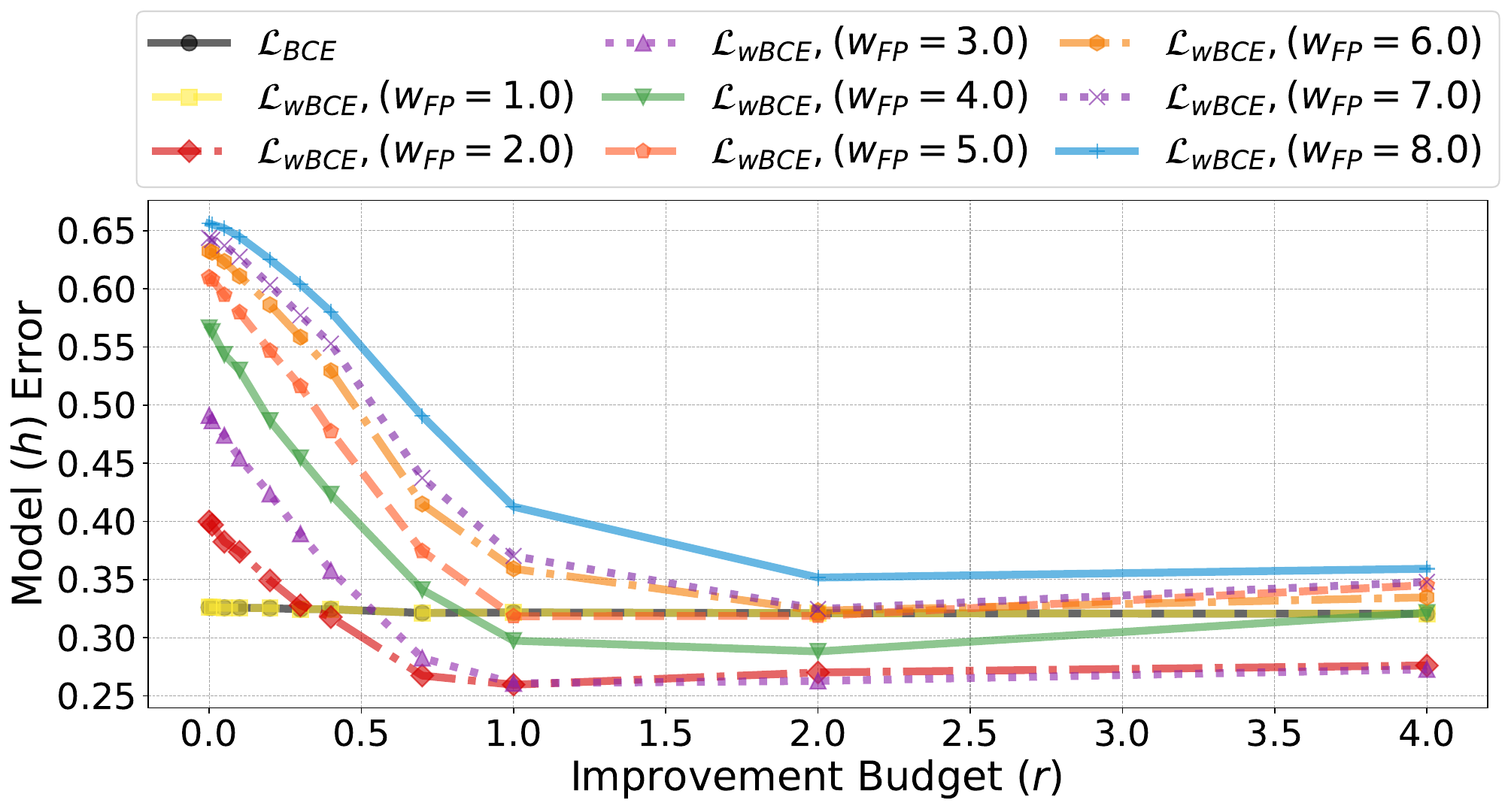}
    \caption{OULAD \(\big(\mathcal{L}_{\textrm{wBCE}} (w_{\textrm{FN}}=1.0)\big)\)}
    \label{fig:app_oulad_fn1_th0.5}
    \end{subfigure}
    ~
    \begin{subfigure}[t]{0.42\linewidth}
        \centering
        \includegraphics[width=\linewidth]{paclearn_improves/paclearn_figures/lfconlyw_0.5_error_droprate_linfx_oulad.pdf}
    \caption{OULAD \(\big(\mathcal{L}_{\textrm{wBCE}} (w_{\textrm{FN}}=1.33)\big)\)}
    \label{fig:app_oulad_fnz_th0.5}
    \end{subfigure}
    \vskip\baselineskip
    \begin{subfigure}[t]{0.42\linewidth}
        \centering
        \includegraphics[width=\linewidth]{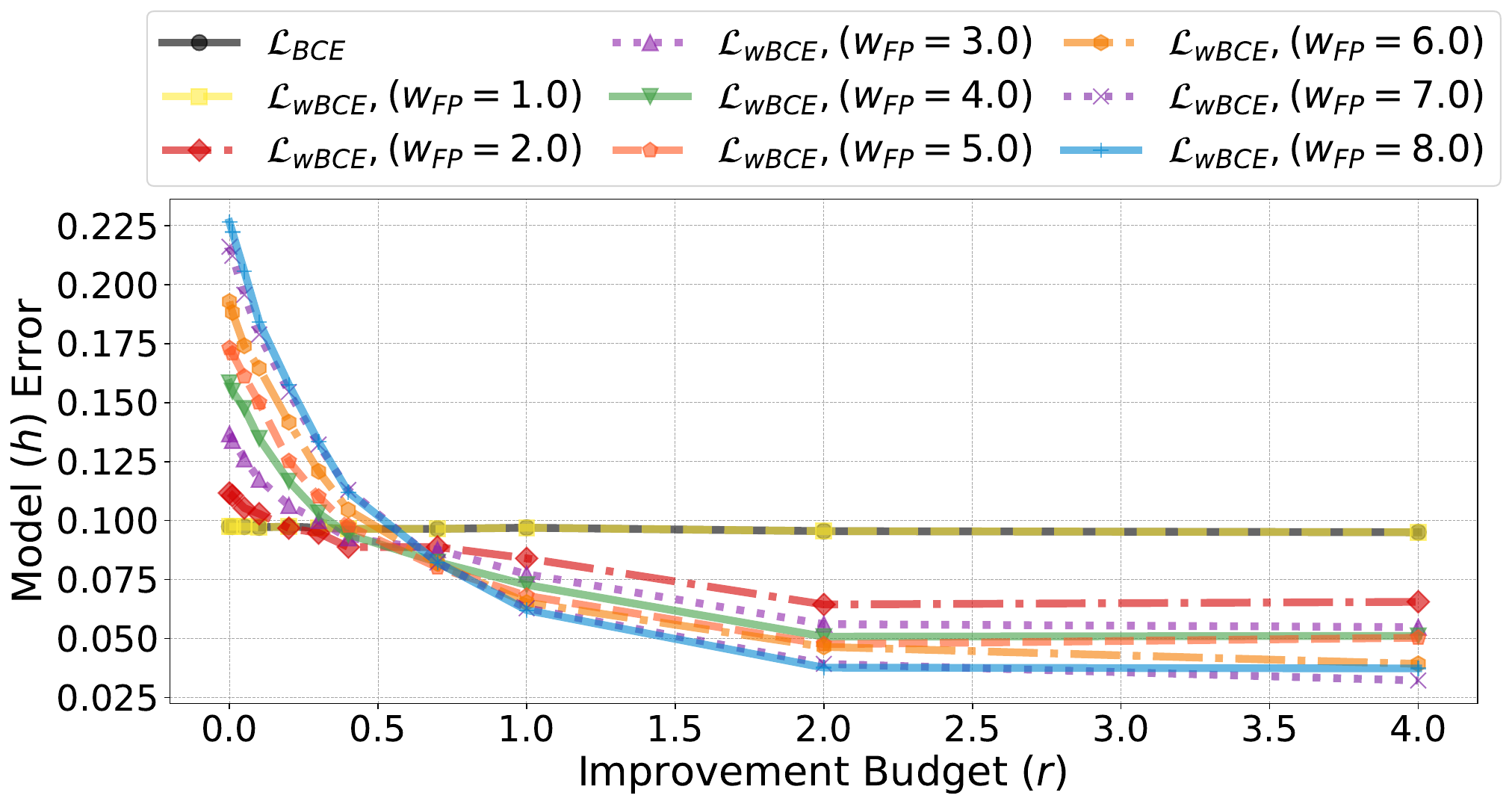}
    \caption{Law school \(\big(\mathcal{L}_{\textrm{wBCE}} (w_{\textrm{FN}}=1.0)\big)\)}
    \label{fig:app_law_fn1_th0.5}
    \end{subfigure}
    ~
    \begin{subfigure}[t]{0.42\linewidth}
        \centering
        \includegraphics[width=\linewidth]{paclearn_improves/paclearn_figures/lfconlyw_0.5_error_droprate_linfx_law.pdf}
    \caption{Law school \(\big(\mathcal{L}_{\textrm{wBCE}} (w_{\textrm{FN}}=0.009)\big)\)}
    \label{fig:app_law_fnz_th0.5}
    \end{subfigure}
    \vskip\baselineskip
    \begin{subfigure}[t]{0.42\linewidth}
        \centering
        \includegraphics[width=\linewidth]{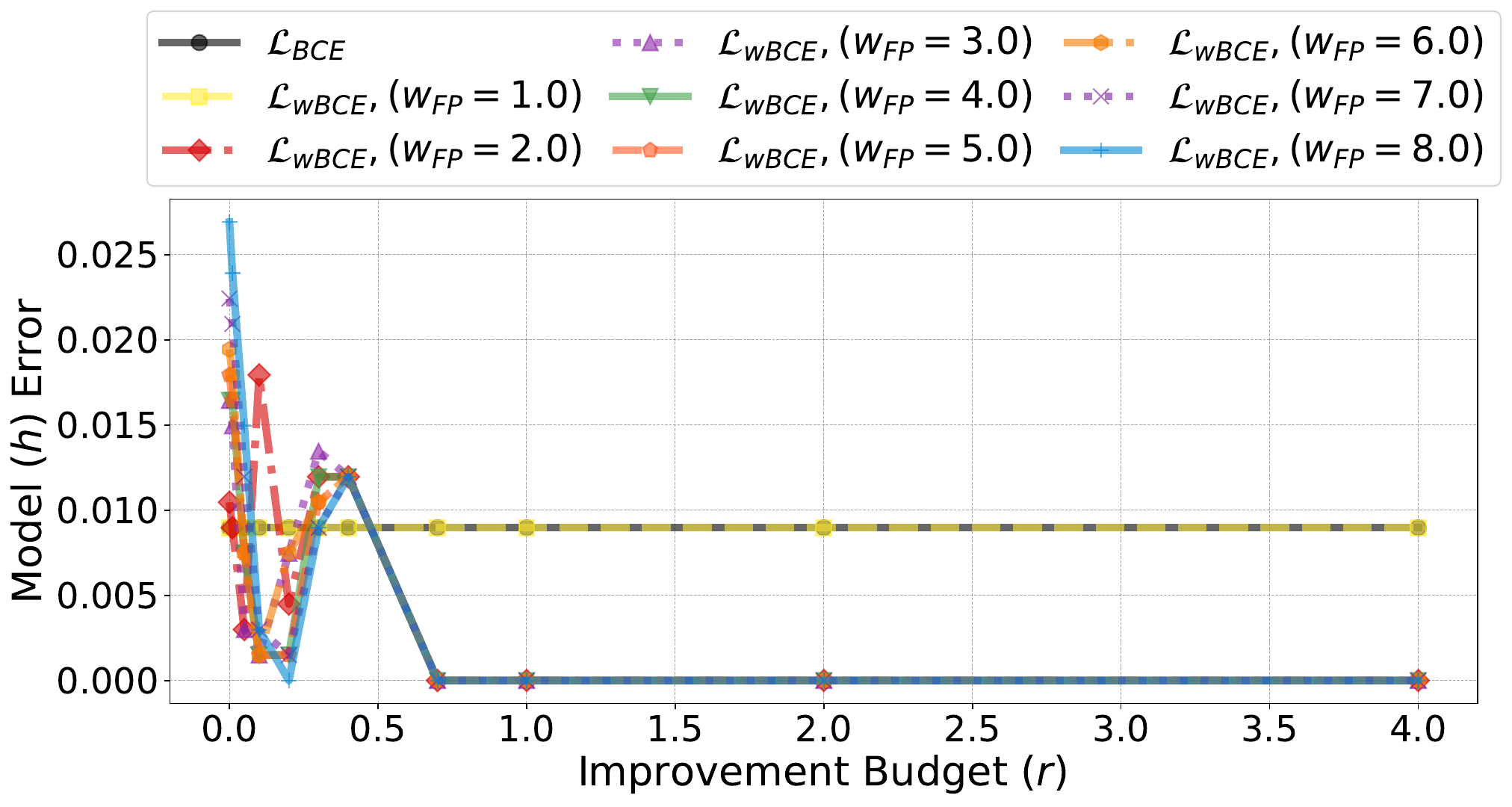}
    \caption{Synthetic \(\big(\mathcal{L}_{\textrm{wBCE}} (w_{\textrm{FN}}=1.0)\big)\)}
    \label{fig:app_syn2_fn1_th0.5}
    \end{subfigure}
    ~
    \begin{subfigure}[t]{0.42\linewidth}
        \centering
        \includegraphics[width=\linewidth]{paclearn_improves/paclearn_figures/lfconlyw_0.5_error_droprate_linfx_synthetic2.pdf}
    \caption{Synthetic \(\big(\mathcal{L}_{\textrm{wBCE}} (w_{\textrm{FN}}=0.009\big)\)}
    \label{fig:app_syn2_fnz_th0.5}
    \end{subfigure}
    
    \caption[Comparative analysis of error drop rate given loss-based risk-aversion]{Comparison of the error drop rate when agents improve to the risk-averse models trained with \(\mathcal{L}_{\textrm{wBCE}}\) where \(w_{\textrm{FN}}=1.0\) (\subref{fig:app_adult_fn1_th0.5}, \subref{fig:app_oulad_fn1_th0.5}, \subref{fig:app_law_fn1_th0.5}, and \subref{fig:app_syn2_fn1_th0.5}) and where \(w_{\textrm{FN}}\) (\subref{fig:app_adult_fnz_th0.5}, \subref{fig:app_oulad_fnz_th0.5}, \subref{fig:app_law_fnz_th0.5}, and \subref{fig:app_syn2_fnz_th0.5}) is optimized and false positive weight is varied \(w_{\textrm{FP}} = \{i\}_{i=1}^{8}\) across different datasets (Adult, OULAD, Law school and Synthetic). The standard model (black line) trained with \(\mathcal{L}_{\textrm{BCE}}\) loss function is such that \(w_{\textrm{FP}} = w_{\textrm{FN}}=1\), and in all cases an agent is classified as positive if the probability of being positive is above \(0.5\).}
    \label{fig:bce_wbce_threshvar}

    \begin{picture}(0,0)
        \put(-155,670){{\parbox{4cm}{\centering \(\mathcal{L}_{\textrm{wBCE}} (w_{\textrm{FN}}=1.0)\)}}}
        \put(55,670){{\parbox{4cm}{\centering \(\mathcal{L}_{\textrm{wBCE}} (w_{\textrm{FN}}=z)\)}}}
        \put(-210,585){\rotatebox{90}{Adult}}
        \put(-210,435){\rotatebox{90}{OULAD}}
        \put(-210,290){\rotatebox{90}{Law school}}
        \put(-210,155){\rotatebox{90}{Synthetic}}
        
    \end{picture}
\end{figure}

\begin{figure}[ht!]
    \centering
    \begin{subfigure}[t]{0.46\linewidth}
        \centering
        \includegraphics[width=\linewidth]{paclearn_improves/paclearn_figures/adult_all_movements_percentages_logscale.pdf}
    \caption{Adult (Movement of agents from TN/FN to TP/FP)}
    \label{fig:app_adult_move_0.5}
    \end{subfigure}
    \hfill
    \begin{subfigure}[t]{0.49\linewidth}
        \centering
        \includegraphics[width=\linewidth]{paclearn_improves/paclearn_figures/adult_FNsFPs_percentages.pdf}
    \caption{Adult (FNR/FPR before and after agents' improvement)}
    \label{fig:app_adult_move_fpr_fnr_0.5}
    \end{subfigure}
    \vspace{1.6em}
    \begin{subfigure}[t]{0.46\linewidth}
        \centering
        \includegraphics[width=\linewidth]{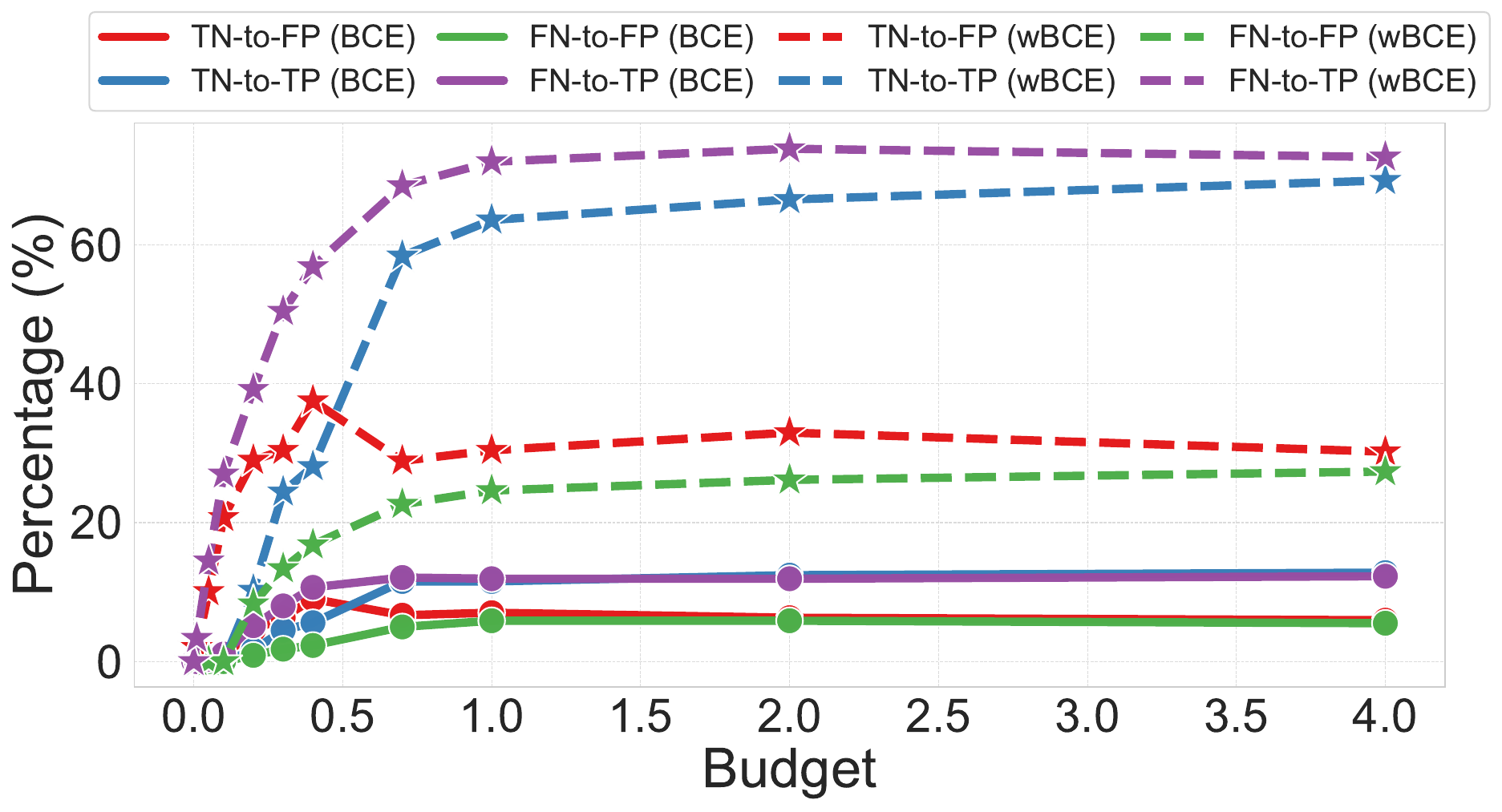}
    \caption{OULAD (Movement of agents from TN/FN to TP/FP)}
    \label{fig:app_oulad_move_0.5}
    \end{subfigure}
    \hfill
    \begin{subfigure}[t]{0.49\linewidth}
        \centering
        \includegraphics[width=\linewidth]{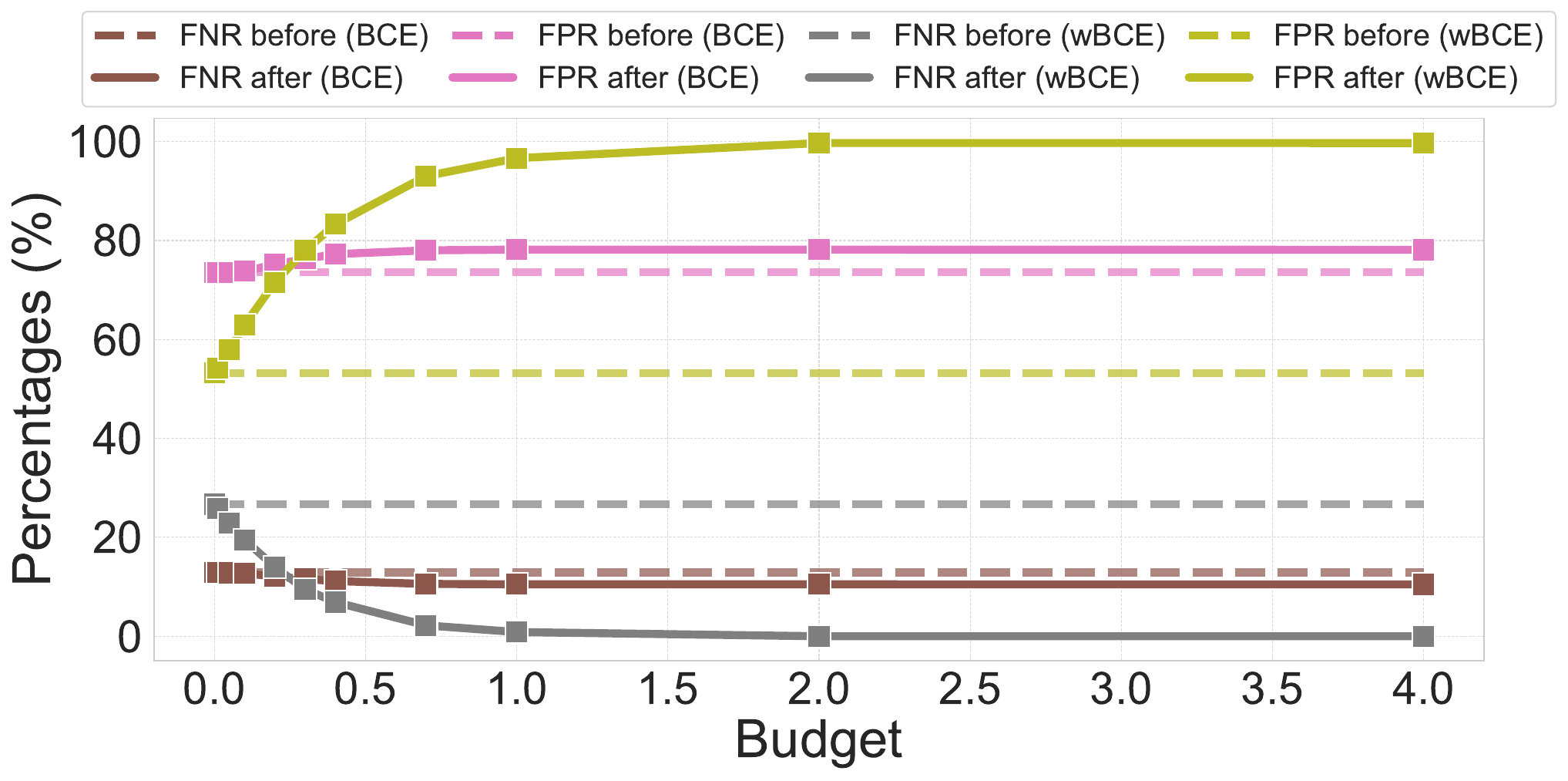}
    \caption{OULAD (FNR/FPR before and after agents' improvement)}
    \label{fig:app_oulad_move_fpr_fnr_0.5}
    \end{subfigure}
    \vspace{1em}
    \begin{subfigure}[t]{0.46\linewidth}
        \centering
        \includegraphics[width=\linewidth]{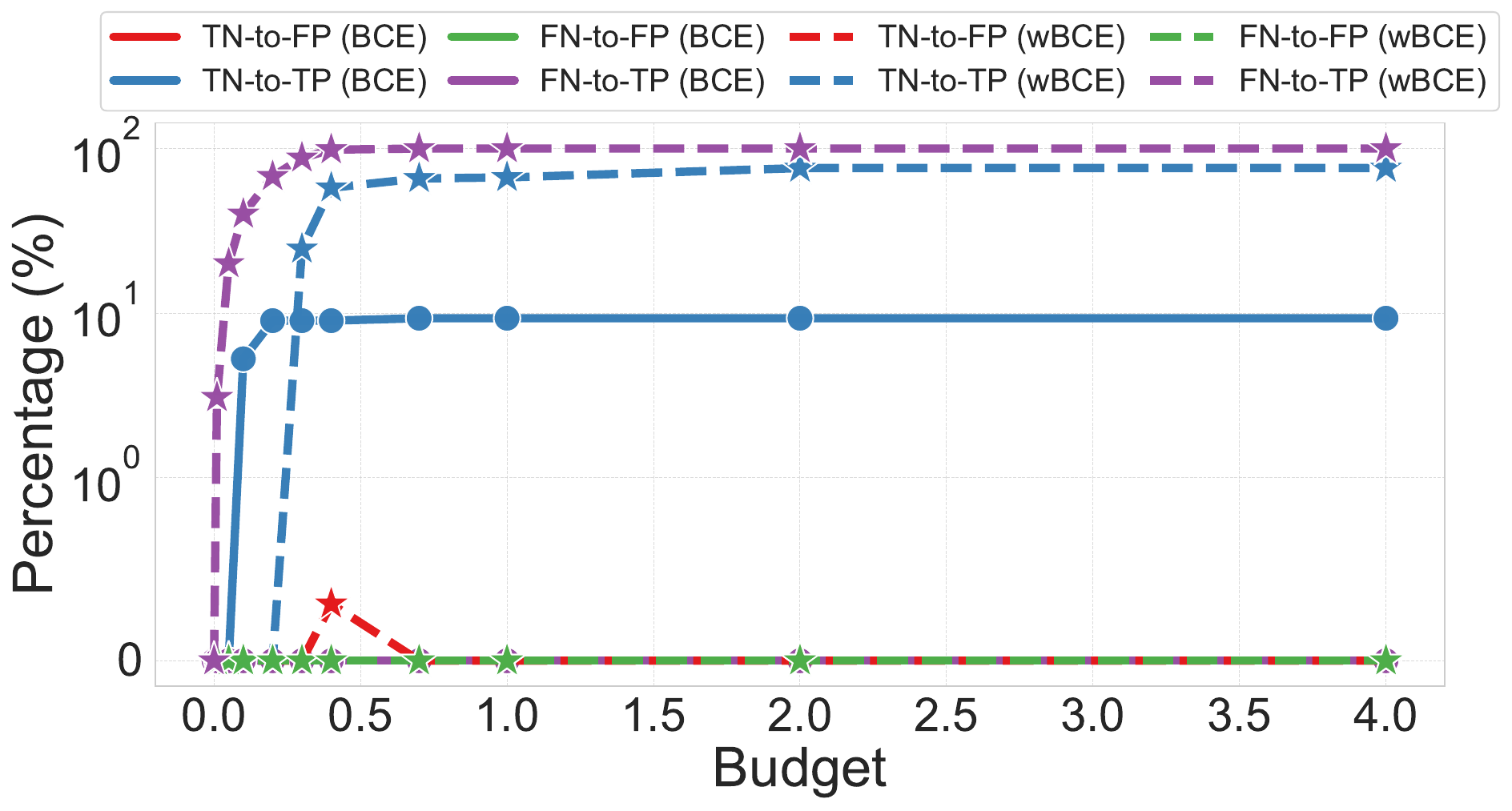}
    \caption{Synthetic (Movement of agents from TN/FN to TP/FP)}
    \label{fig:app_synthetic_move_0.5}
    \end{subfigure}
    \hfill
    \begin{subfigure}[t]{0.49\linewidth}
        \centering
        \includegraphics[width=\linewidth]{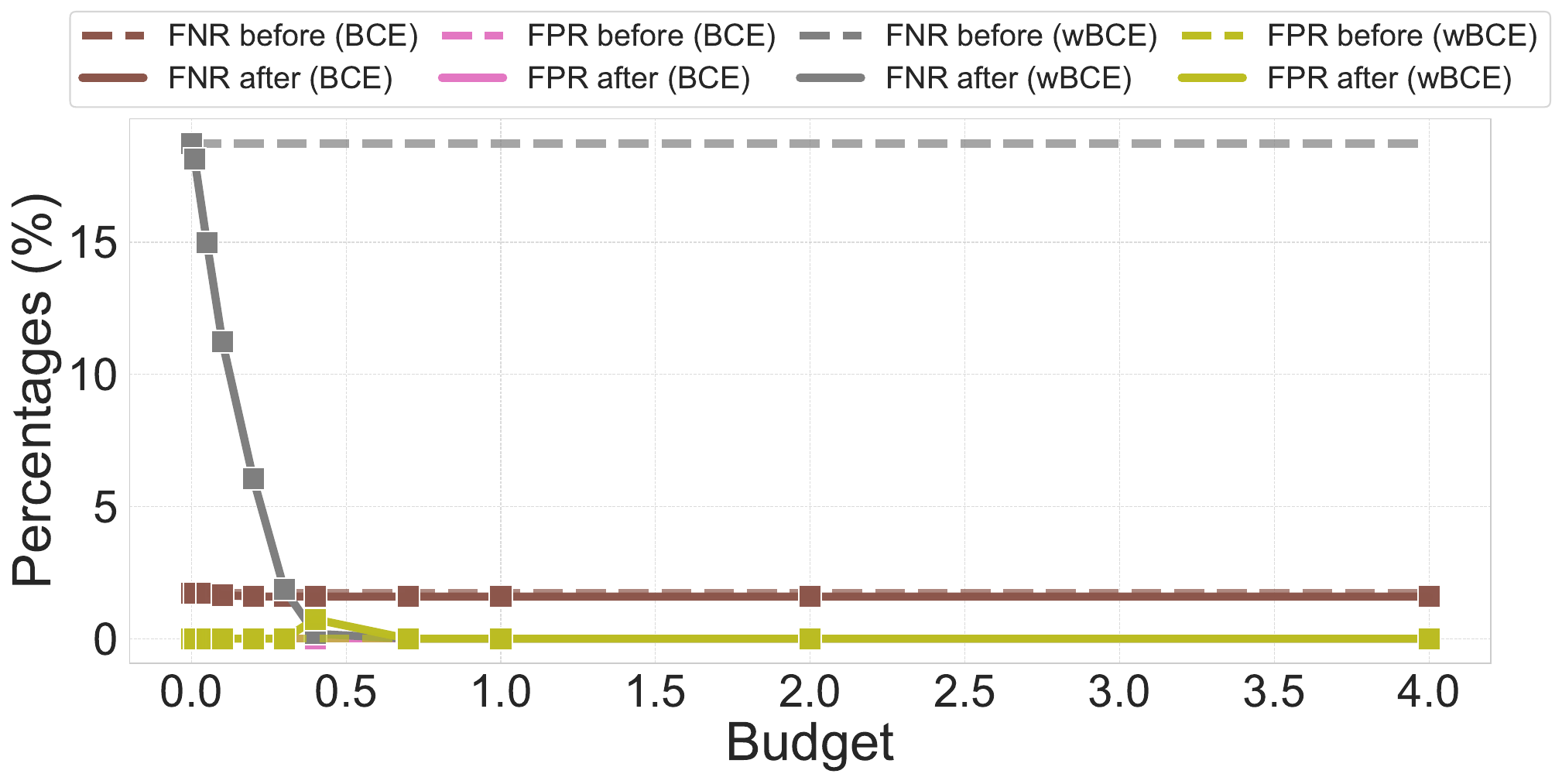}
    \caption{Synthetic (FNR/FPR before and after agents' improvement)}
    \label{fig:app_synthetic_move_fpr_fnr_0.5}
    \end{subfigure}
    
    \caption[The percentage of TN and FN agents that transition to TP and FP]{(\subref{fig:app_adult_move_0.5}, \subref{fig:app_oulad_move_0.5}, and \subref{fig:app_synthetic_move_0.5}) The percentage of negatively classified agents (TN and FN) that transition to TP and FP after responding to the classifier (\(h(x)\)). 
    (\subref{fig:app_adult_move_fpr_fnr_0.5}, \subref{fig:app_oulad_move_fpr_fnr_0.5}, and \subref{fig:app_synthetic_move_fpr_fnr_0.5}) The FPR and FNR before and after agents move. On the adult dataset, the wBCE-trained model used \(w_{\textrm{FP}}=0.001\) and \(w_{\textrm{FN}} = 4.4\), while on the OULAD dataset, it used \(w_{\textrm{FP}}=1.33\) and \(w_{\textrm{FN}} = 2\), and on the synthetic dataset, \(w_{\textrm{FP}}=0.009\) and \(w_{\textrm{FN}} = 1.0\). In all cases, BCE-trained models used \(w_{\textrm{FP}} = w_{\textrm{FN}} = 1\), and in all cases an agent is classified as positive if the probability of being positive is above \(0.5\).}
    \label{fig:oulad_synthetic_move_erroreval_0.5}
\end{figure}

\begin{figure}[htb!]
    \centering
    \begin{subfigure}[t]{0.42\linewidth}
        \centering
        \includegraphics[width=\linewidth]{paclearn_improves/paclearn_figures/lfconlyw_0.5_error_droprate_linfx_adult.pdf}
    \caption{Adult \(\big(\mathcal{L}_{\textrm{wBCE}} (w_{\textrm{FN}}=0.001)\big)\)}
    \label{fig:app_adult_0.5}
    \end{subfigure}
    ~
    \begin{subfigure}[t]{0.42\linewidth}
        \centering
        \includegraphics[width=\linewidth]{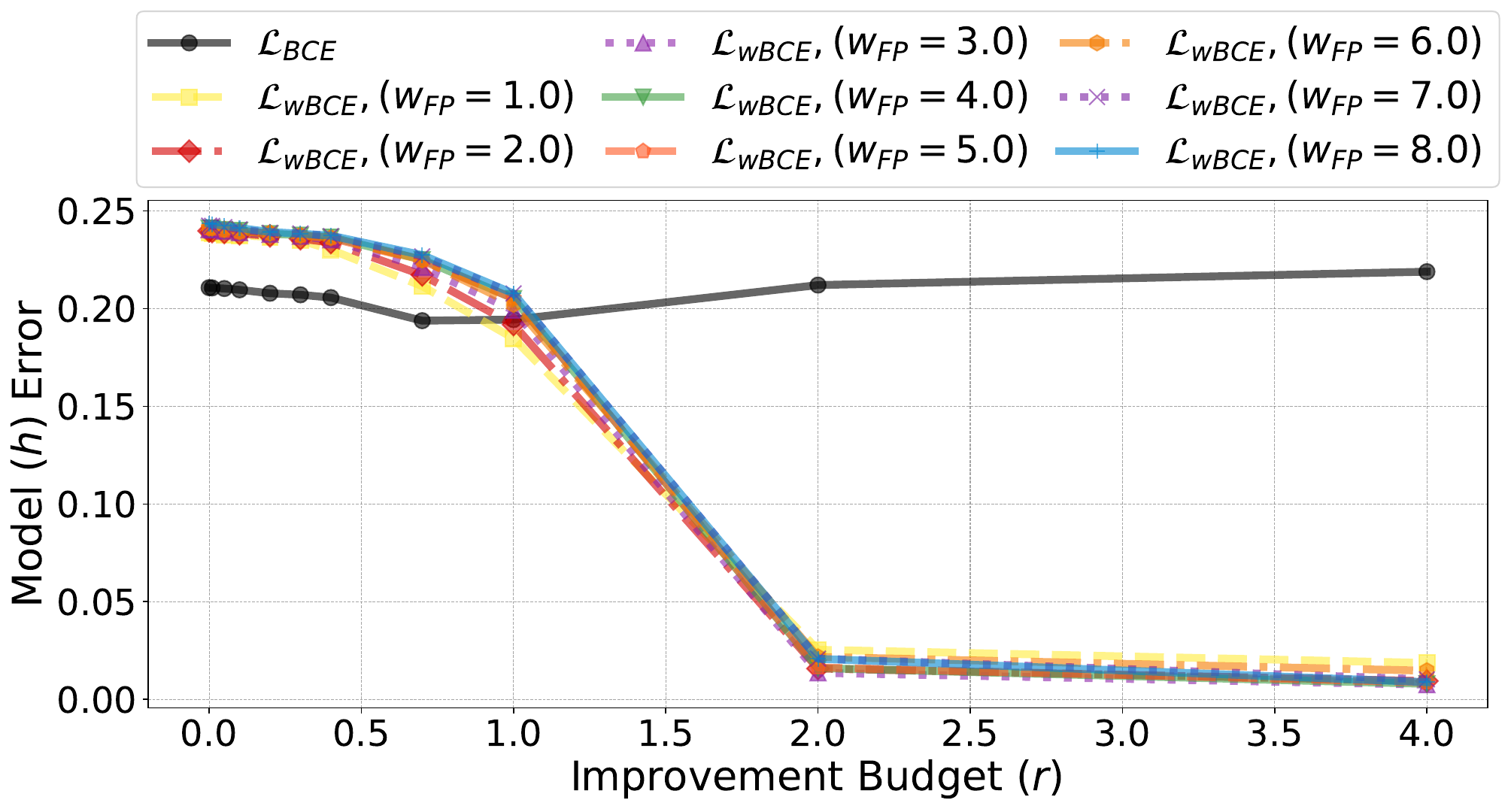}
    \caption{Adult \(\big(\mathcal{L}_{\textrm{wBCE}} (w_{\textrm{FN}}=0.001)\big)\)}
    \label{fig:app_adult_0.9}
    \end{subfigure}    
    \vskip\baselineskip
    \begin{subfigure}[t]{0.42\linewidth}
        \centering
        \includegraphics[width=\linewidth]{paclearn_improves/paclearn_figures/lfconlyw_0.5_error_droprate_linfx_oulad.pdf}
    \caption{OULAD \(\big(\mathcal{L}_{\textrm{wBCE}} (w_{\textrm{FN}}=1.33)\big)\)}
    \label{fig:app_oulad_0.5}
    \end{subfigure}
    ~
    \begin{subfigure}[t]{0.42\linewidth}
        \centering
        \includegraphics[width=\linewidth]{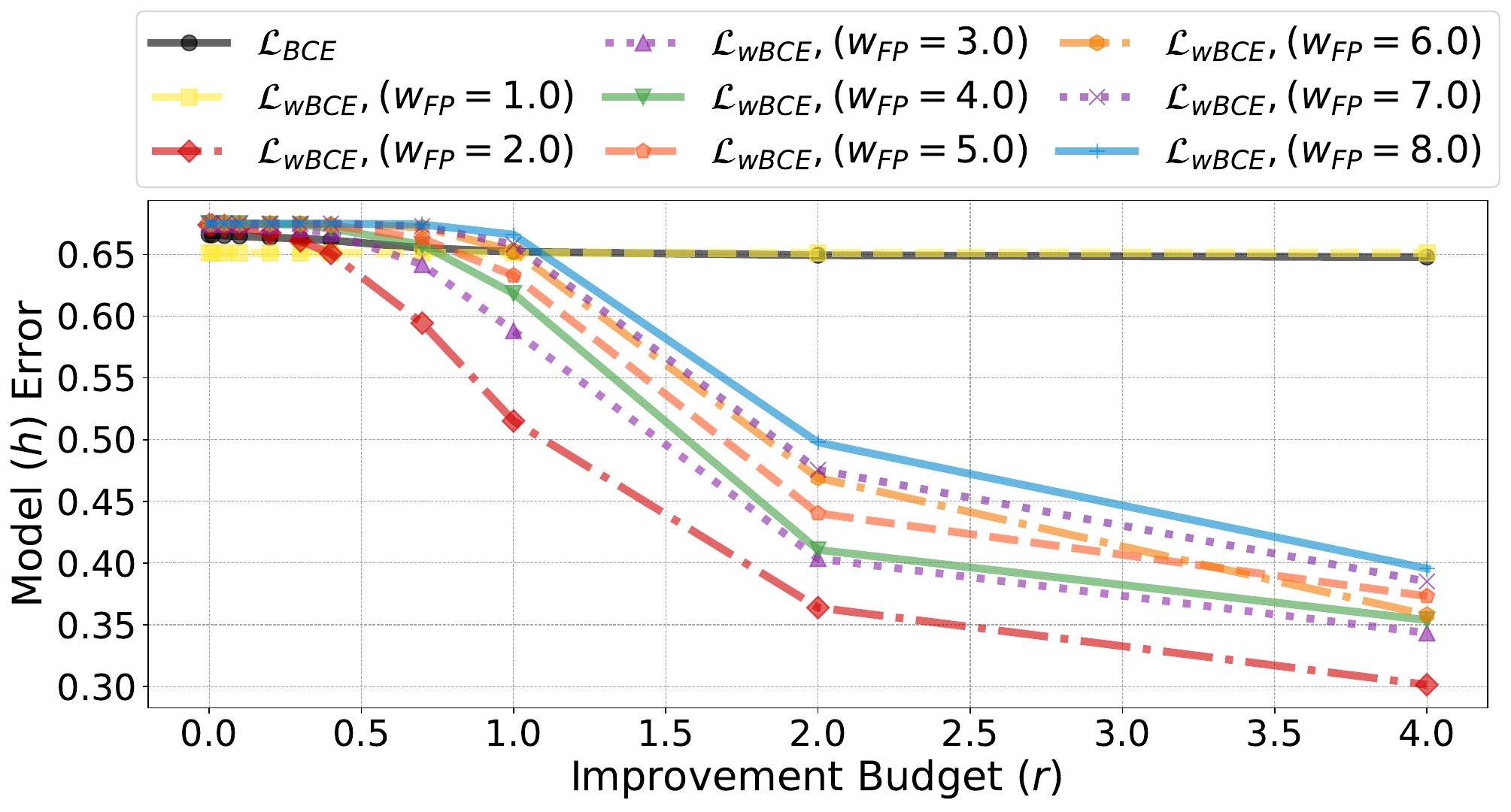}
    \caption{OULAD \(\big(\mathcal{L}_{\textrm{wBCE}} (w_{\textrm{FN}}=1.33)\big)\)}
    \label{fig:app_oulad_0.9}
    \end{subfigure}
    \vskip\baselineskip
   
    \begin{subfigure}[t]{0.42\linewidth}
        \centering
        \includegraphics[width=\linewidth]{paclearn_improves/paclearn_figures/lfconlyw_0.5_error_droprate_linfx_law.pdf}
    \caption{Law school \(\big(\mathcal{L}_{\textrm{wBCE}} (w_{\textrm{FN}}=0.009)\big)\)}
    \label{fig:app_law_0.5}
    \end{subfigure}
    ~
    \begin{subfigure}[t]{0.42\linewidth}
        \centering
        \includegraphics[width=\linewidth]{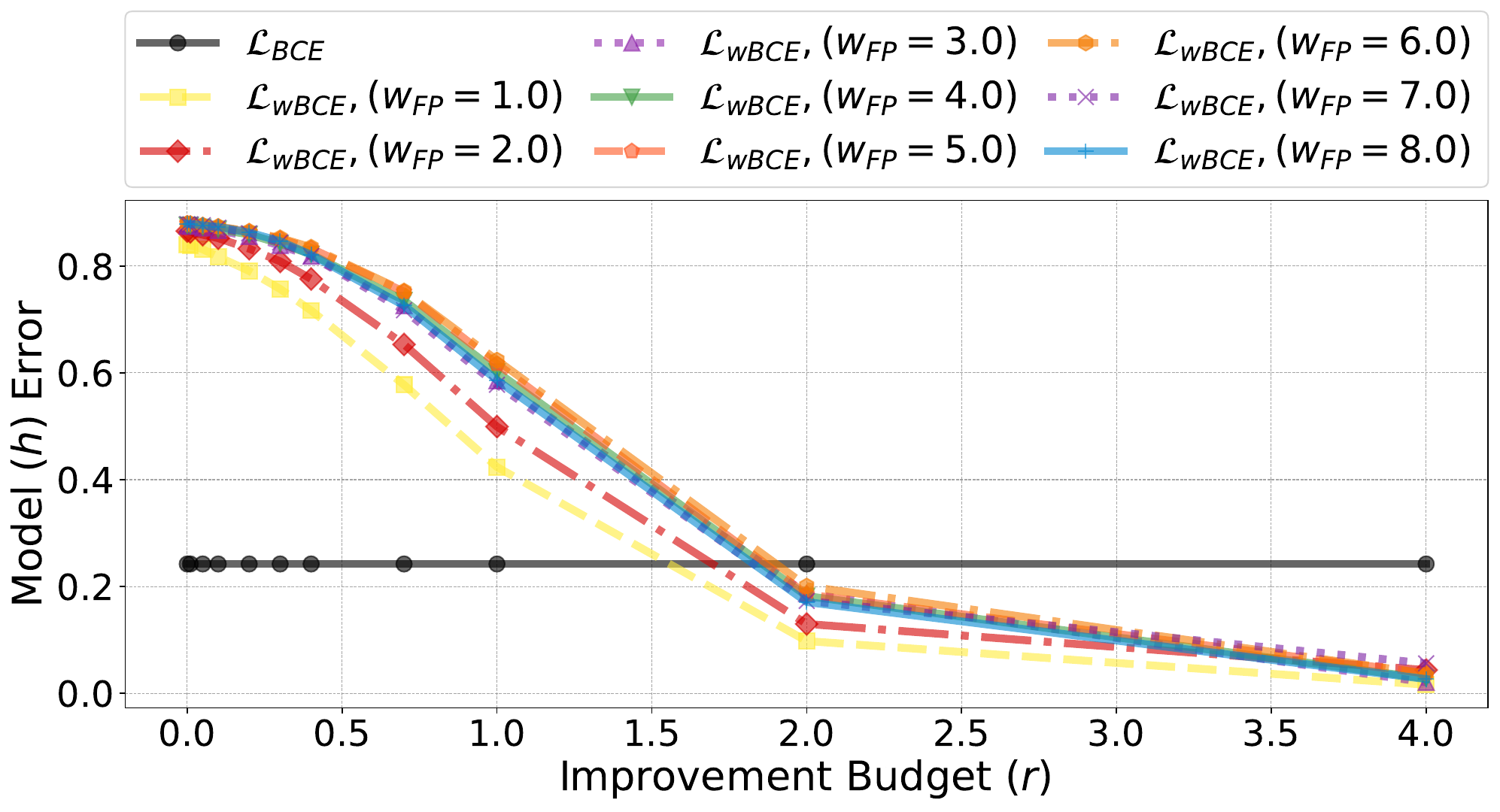}
    \caption{Law school \(\big(\mathcal{L}_{\textrm{wBCE}} (w_{\textrm{FN}}=0.009)\big)\)}
    \label{fig:app_law_0.9}
    \end{subfigure}     
    \vskip\baselineskip
   
    \begin{subfigure}[t]{0.42\linewidth}
        \centering
        \includegraphics[width=\linewidth]{paclearn_improves/paclearn_figures/lfconlyw_0.5_error_droprate_linfx_synthetic2.pdf}
    \caption{Synthetic \(\big(\mathcal{L}_{\textrm{wBCE}} (w_{\textrm{FN}}=0.009)\big)\)}
    \label{fig:app_synthetic_0.5}
    \end{subfigure}
    ~
    \begin{subfigure}[t]{0.42\linewidth}
        \centering
        \includegraphics[width=\linewidth]{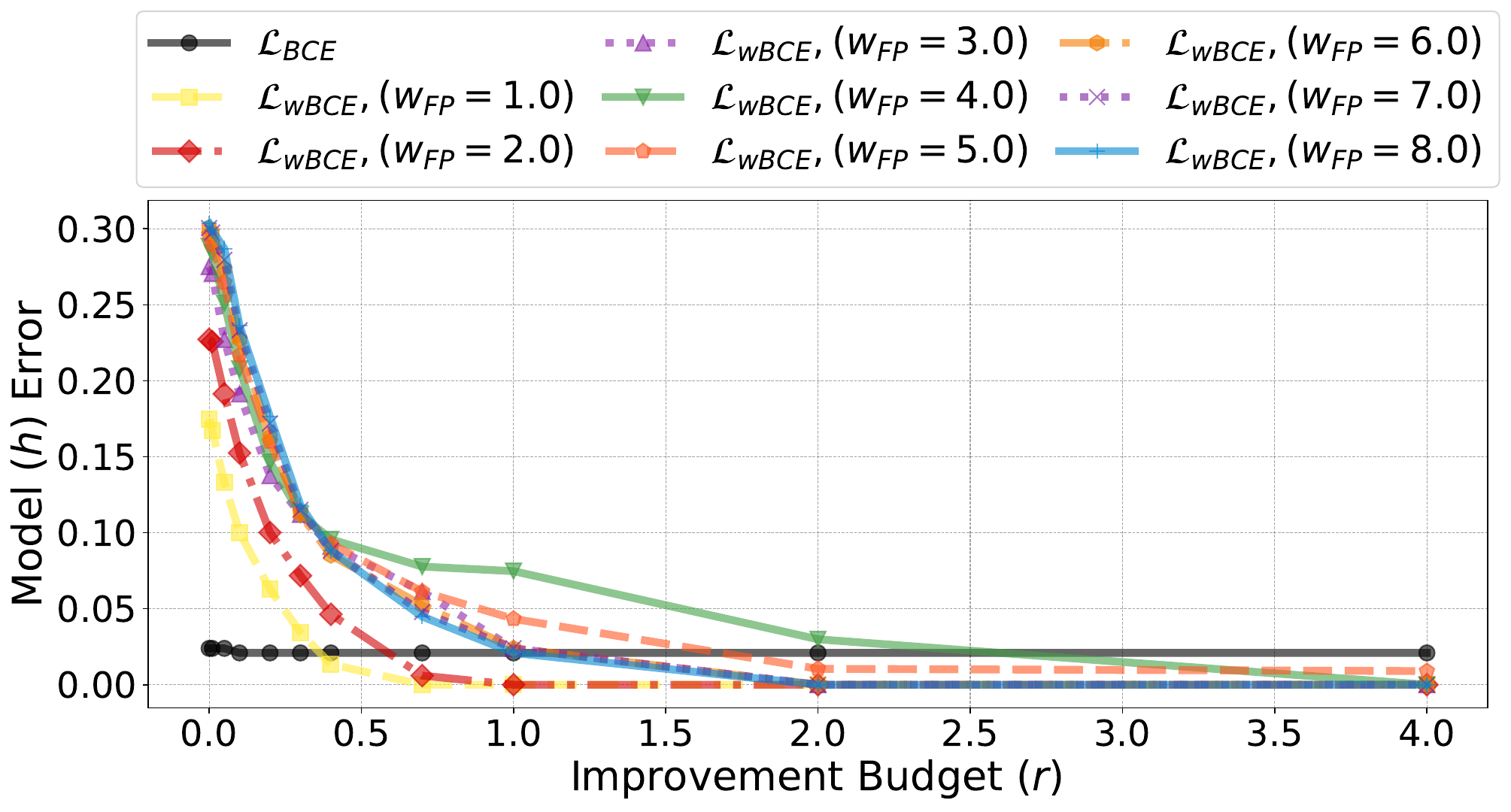}
    \caption{Synthetic \(\big(\mathcal{L}_{\textrm{wBCE}} (w_{\textrm{FN}}=0.009)\big)\)}
    \label{fig:app_synthetic_0.9}
    \end{subfigure}
    
    \caption[Comparative analysis of error drop rate given threshold-based risk-aversion]{Comparison of the error drop rate when agents improve to the risk-averse \(\big(\mathcal{L}_{\textrm{wBCE}}, \frac{w_\textrm{FP}}{w_\textrm{FN}}>1,  w_{\textrm{FP}}=\{i\}_{i=1}^{8}\big)\) and standard  (\(\mathcal{L}_\textrm{BCE}, w_{\textrm{FP}}= w_{\textrm{FN}}=1\)) models across four datasets (Adult, OULAD, Law school, and Synthetic). \textbf{Column one} (\subref{fig:app_adult_0.5}, \subref{fig:app_oulad_0.5}, \subref{fig:app_law_0.5} and \subref{fig:app_synthetic_0.5}) considers models where an agent is classified as positive if the probability of being positive is above \(0.5\) and \textbf{column two}  (\subref{fig:app_adult_0.9}, \subref{fig:app_oulad_0.9}, \subref{fig:app_law_0.9} and \subref{fig:app_synthetic_0.9}) considers models where a higher threshold is used \(0.9\). Increasing the improvement budget and classifier risk-aversion (high \(\frac{w_{\textrm{FP}}}{w_{\textrm{FN}}}\)) leads to a sharper error drop rate, and loss-based risk aversion is more effective than the threshold-based risk-aversion.}
    \label{fig:app_thresh0.5thresh0.9}

    \begin{picture}(0,0)
        \put(-150,690){{\parbox{4cm}{\centering \(\text{Threshold}=0.5\)}}}
        \put(66,690){{\parbox{4cm}{\centering \(\text{Threshold}=0.9\)}}}
        \put(-210,610){\rotatebox{90}{Adult}}
        \put(-210,460){\rotatebox{90}{OULAD}}
        \put(-210,318){\rotatebox{90}{Law school}}
        \put(-210,182){\rotatebox{90}{Synthetic}}        
    \end{picture}
\end{figure}

\paragraph{Effect of risk-aversion and improvement budget.} Figures~\ref{fig:app_adult_0.5}, \ref{fig:app_oulad_0.5}, \ref{fig:app_law_0.5} and \ref{fig:app_synthetic_0.5}
show that when agents improve to a risk-averse model trained using \(\mathcal{L}_{\textrm{wBCE}}\) with \(\frac{w_{\textrm{FP}}}{w_{\textrm{FN}}} > 1\), the error rate decreases rapidly, particularly as the improvement budget increases. Notably, the higher the false positive to false negative weight ratio \(\frac{w_{\textrm{FP}}}{w_{\textrm{FN}}}\), the faster the reduction in error (see  Figure~\ref{fig:bce_wbce_threshvar}). 
In contrast, models trained with the standard \(\mathcal{L}_{\textrm{BCE}}\) loss function exhibit a slower error reduction rate, almost looking like a line, under the same conditions as the effects of improvement are minimal and cancel each other out (see Figure~\ref{fig:oulad_synthetic_move_erroreval_0.5}). 

Furthermore, the false negative rate (FNR) decreases as the improvement budget \(r\) grows when the agents respond (improve) to an \(\mathcal{L}_{\textrm{wBCE}}\)--trained model (see Figures~\ref{fig:app_adult_move_fpr_fnr_0.5},  \ref{fig:app_oulad_move_fpr_fnr_0.5}, and \ref{fig:app_synthetic_move_fpr_fnr_0.5}). This is because almost \(100\%\) of the false negatively classified agents improve to become true positives as shown in Figures~\ref{fig:app_adult_move_0.5}, \ref{fig:app_oulad_move_0.5}, and \ref{fig:app_synthetic_move_0.5}.
On the other hand, on datasets where the weighted-BCE loss function effectively removed false positives (close to \(0\) false positive rate (FPR)) before agents' improvement, remains very low,  in some cases close to \(0\) (e.g., in Figure~\ref{fig:app_synthetic_move_fpr_fnr_0.5}), since no or few agents become false positives after improvement (see Figure~\ref{fig:app_synthetic_move_0.5}). However, when the weighted-BCE loss function wasn't as effective, the false positive rate increases as the improvement budget \(r<2.0\) grows (see Figure~\ref{fig:app_oulad_move_fpr_fnr_0.5}). 

Although we observe similar trends when the threshold for classifying an agent as positive increases to \(0.9\) (instead of \(0.5\)), the error is slightly higher and the reduction becomes slower (Figure~\ref{fig:app_thresh0.5thresh0.9}). Additionally, while Figures~\ref{fig:main_thresh0.5} and \ref{fig:bce_wbce_threshvar} demonstrate diminishing returns for \(r > 2.0\), a different trend emerges in Figures~\ref{fig:app_adult_0.9}, \ref{fig:app_oulad_0.9}, \ref{fig:app_law_0.9} and \ref{fig:app_synthetic_0.9} where we classify agents positive with high probability (\(0.9\)).

\begin{figure}[t!]
    \centering
    \begin{subfigure}[t]{0.48\linewidth}
        \centering
        \includegraphics[width=\linewidth]{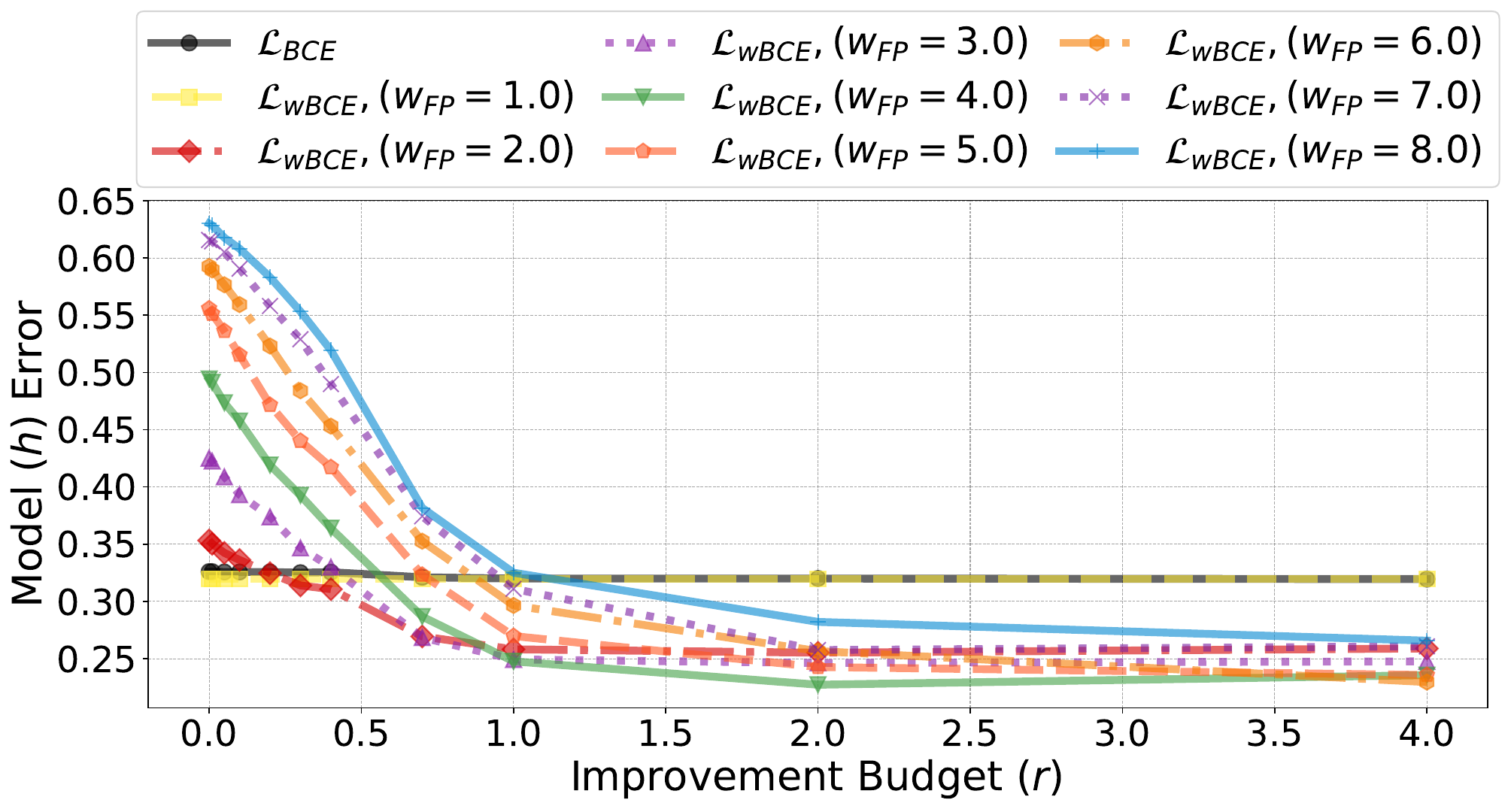}
    \caption{OULAD (\textbf{DTC1})}
    \label{fig:app_oulad_dtc1_0.5}
    \end{subfigure}
    \hfill
    \begin{subfigure}[t]{0.48\linewidth}
        \centering
        \includegraphics[width=\linewidth]{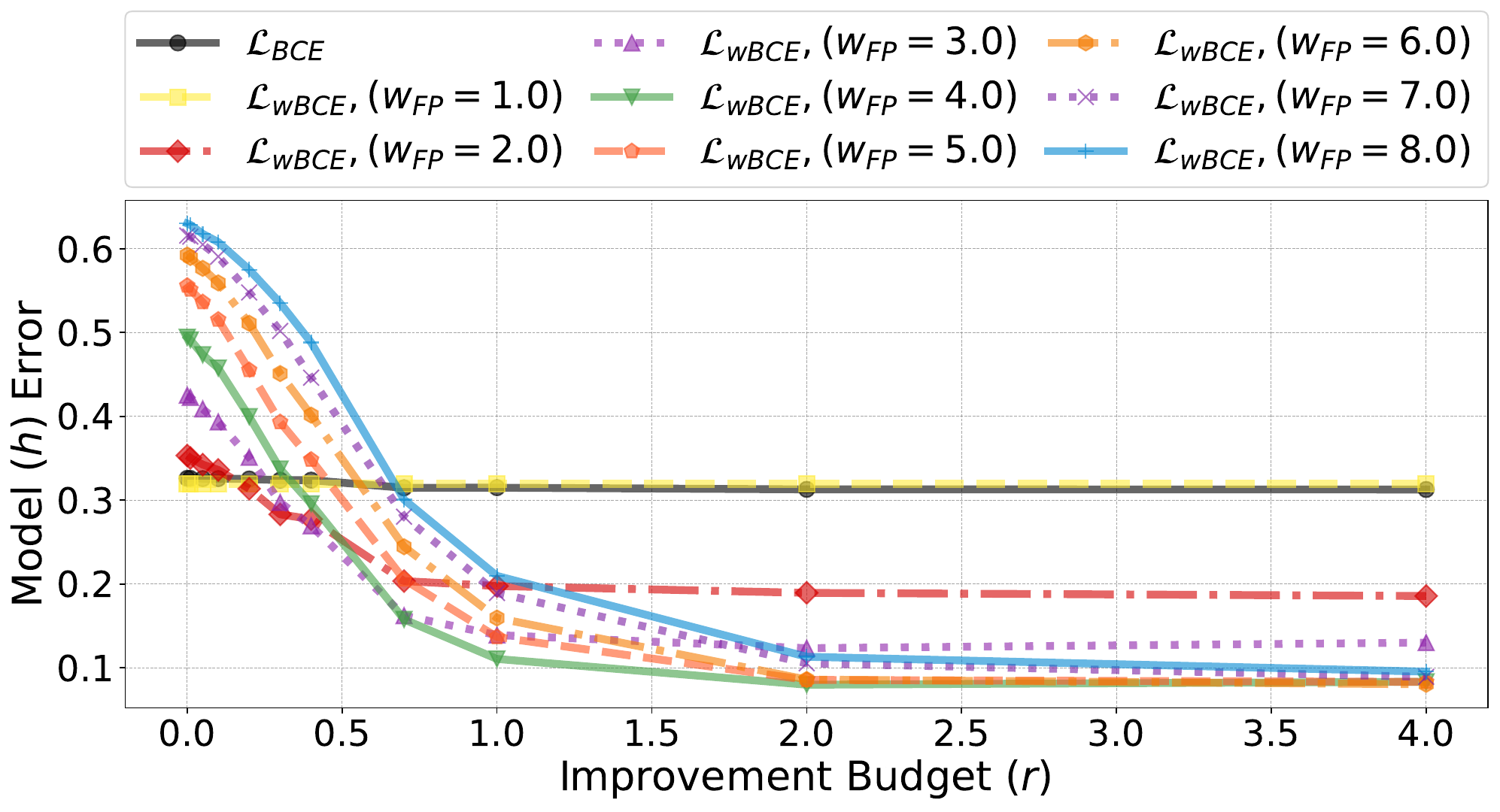}
    \caption{OULAD (\textbf{RFC1})}
    \label{fig:app_oulad_rfc1_0.5}
    \end{subfigure}
    \hfill
    \begin{subfigure}[t]{0.48\linewidth}
        \centering
        \includegraphics[width=\linewidth]{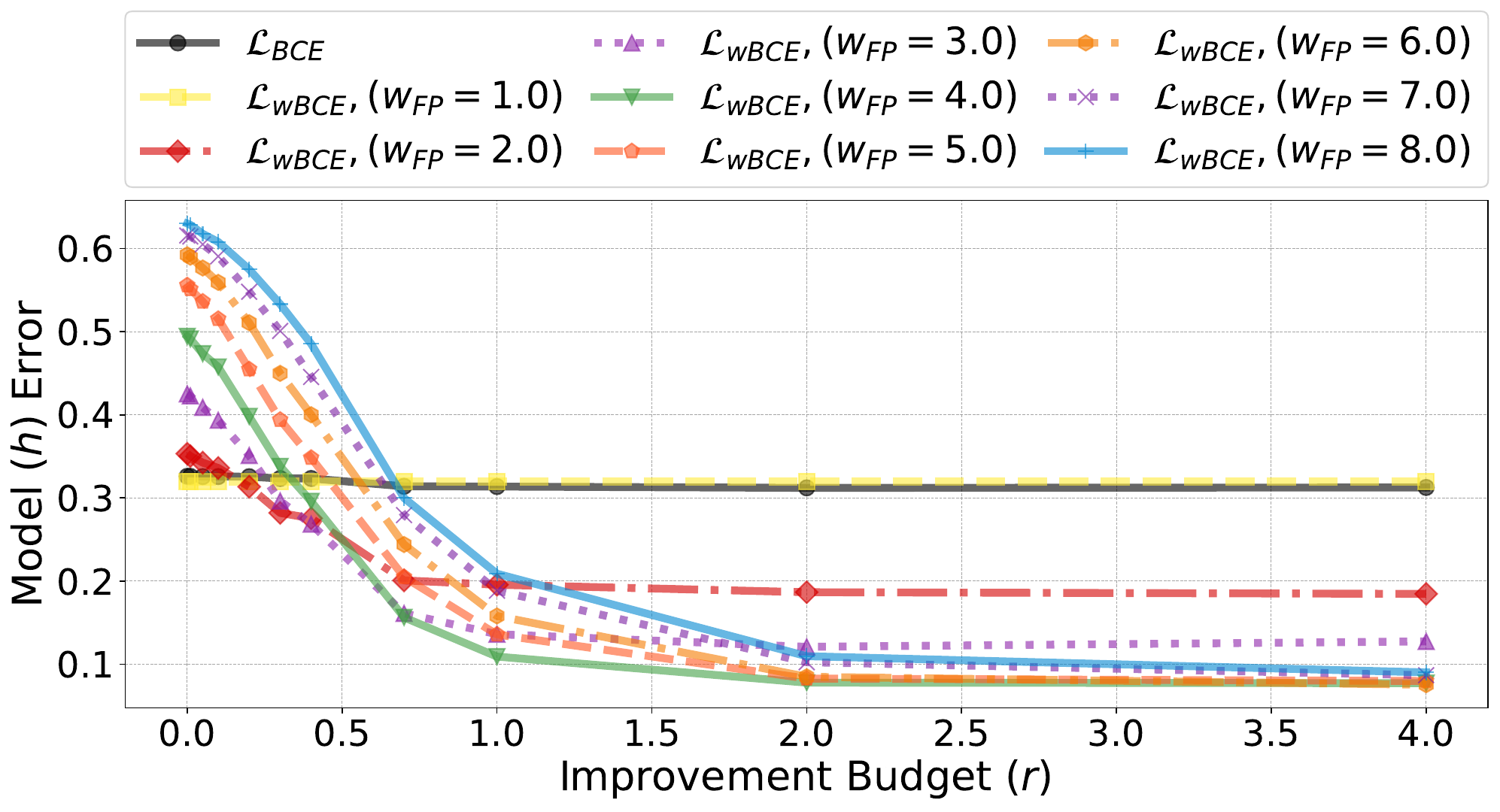}
    \caption{OULAD (\textbf{RFC2})}
    \label{fig:app_oulad_rfc2_0.5}
    \end{subfigure}
    \hfill
    \begin{subfigure}[t]{0.48\linewidth}
        \centering
        \includegraphics[width=\linewidth]{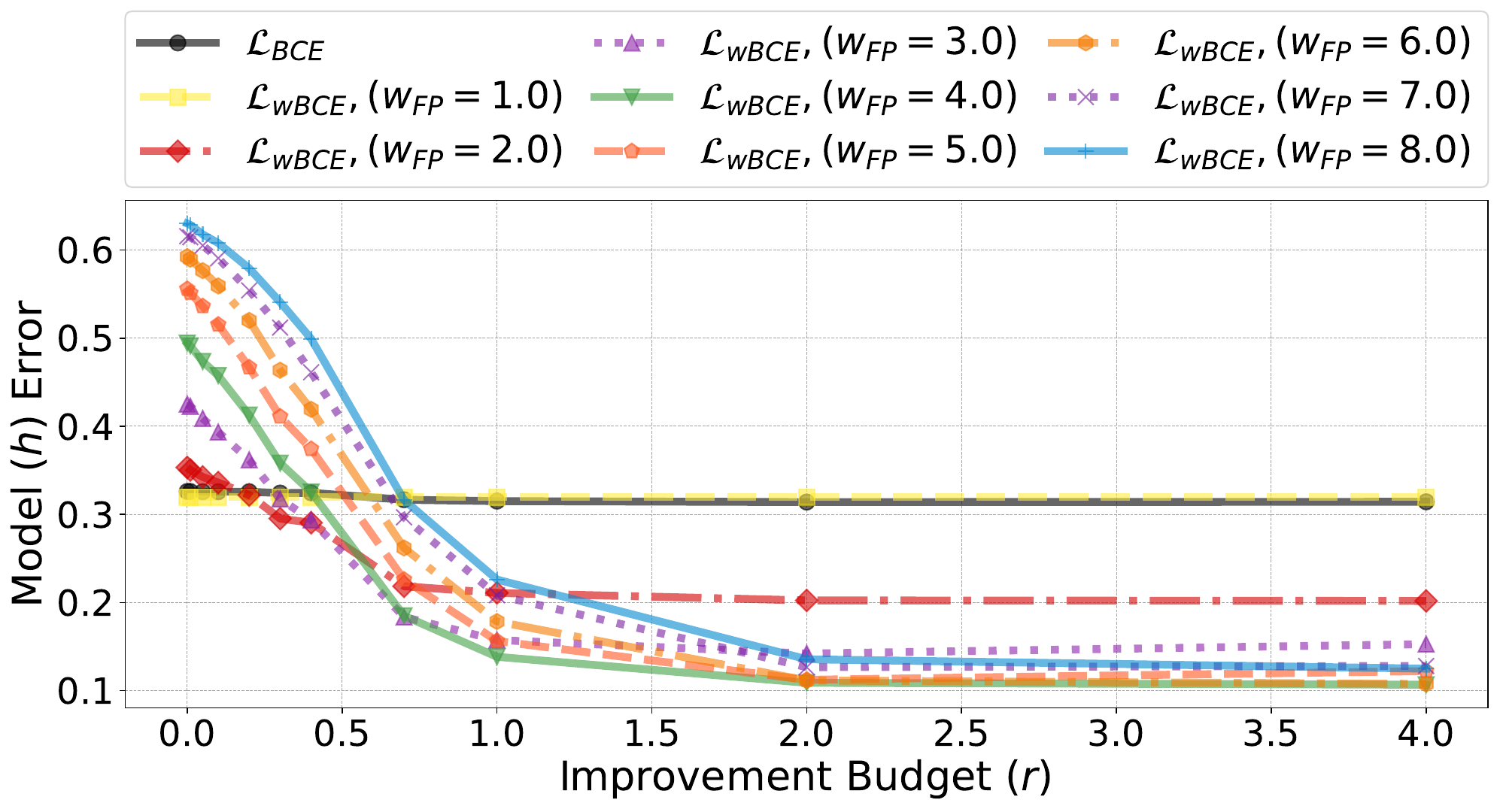}
    \caption{OULAD (\textbf{XGB})}
    \label{fig:app_oulad_xgb_0.5}
    \end{subfigure}

    \caption[Error drop rate versus budget \(r\), given evaluation with \textbf{singularly-defined} \(f^\star\)]{Risk-averse \(\big(\mathcal{L}_{\textrm{wBCE}} \ \text{with} \ w_{\textrm{FP}}=\{i\}_{i=1}^{8}, w_{\textrm{FN}}=1.33\big)\) and standard (\(\mathcal{L}_{\textrm{BCE}}, w_{\textrm{FP}}=w_{\textrm{FN}}=1\)) trained model function (\(h\)) error drop rates versus improvement budget (\(r\)) on the Adult dataset where different \textbf{singularly-defined} \(f^\star\) are used to verify successful-ness of improvement: (\subref{fig:app_oulad_dtc1_0.5}) with \(f^{\star}_{1}\) , (\subref{fig:app_oulad_rfc1_0.5}) with \(f^{\star}_{3}\), (\subref{fig:app_oulad_rfc2_0.5}) with \(f^{\star}_{4}\), and (\subref{fig:app_oulad_xgb_0.5}) with \(f^{\star}_{5}\). For \textbf{DTC2}, \(f^{\star}_{2}\) see Figure~\ref{fig:app_oulad_0.5}. In all cases, the threshold for classifying an agent as positive is \(0.5\).}
    \label{fig:app_oulad_multi_thresh0.5}
\end{figure}

\paragraph{Effect of choice of \(f^\star\) model.}
Although different \(f^\star\) models achieved \(\sim 100\%\) accuracy on a given dataset while exhibiting varied LOO accuracy (cf. Tables~\ref{tab:fstarclassifiers_gen_acc} and \ref{tab:fstarclassifiers_loo}), the error drop rate showed consistent patterns when different \(f^\star\) models are used. As shown in Figure~\ref{fig:app_oulad_multi_thresh0.5}, although some \(f^\star\) models, such as RFC2 (\(f^\star_{4}\)) (cf. Figure~\ref{fig:app_oulad_rfc2_0.5}), had higher performance gains than others, in all cases, the error drops rapidly as improvement budget increases and agents respond (improve) to wBCE-trained models.

\begin{figure}[ht!]
    \centering
    \begin{subfigure}[t]{0.48\linewidth}
        \centering
        \includegraphics[width=\linewidth]{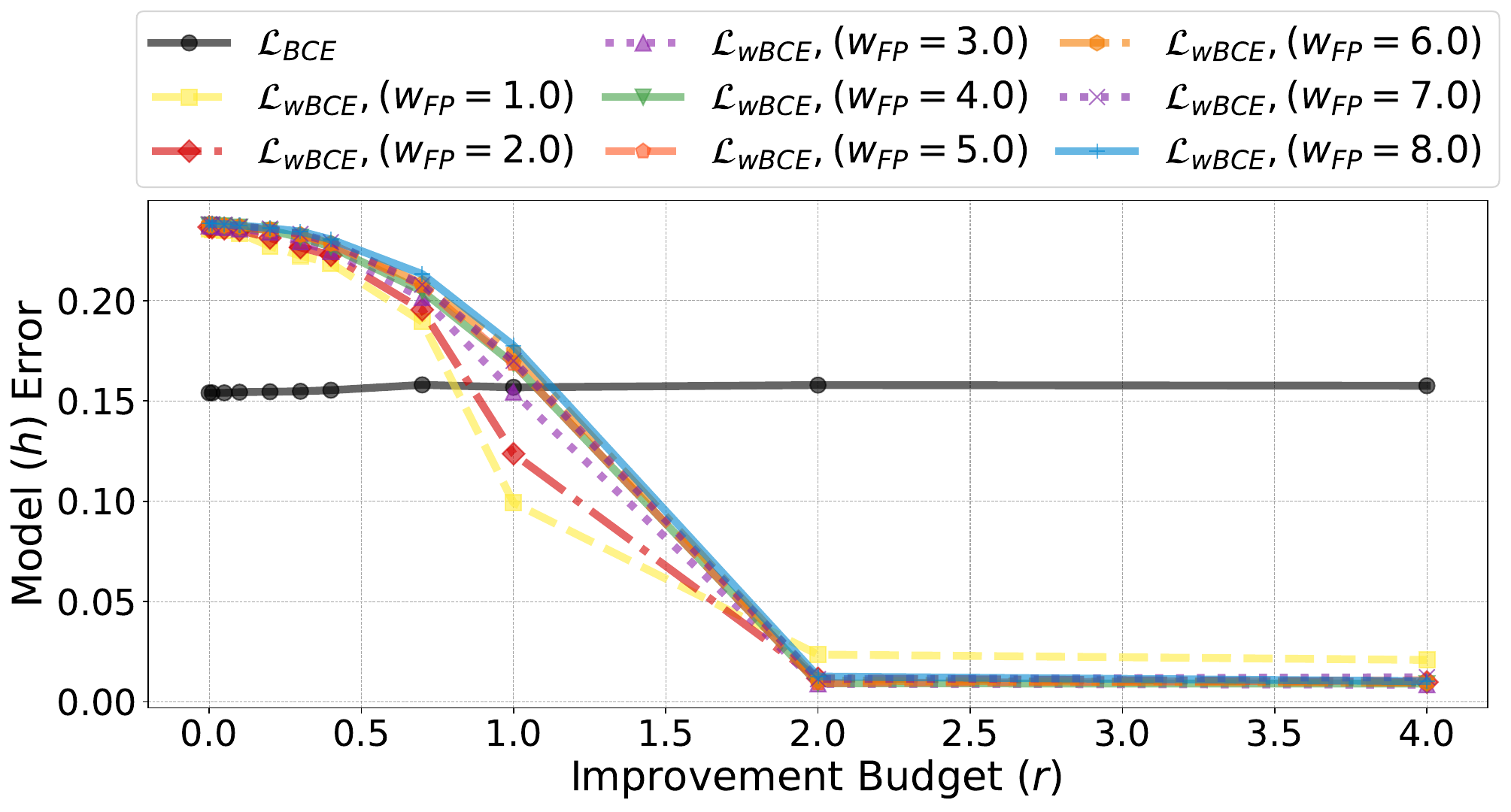}
    \caption{Adult \(\big(w_{\textrm{FN}}=0.001\big)\)}
    \label{fig:app_adult_multi_0.5}
    \end{subfigure}
    \hfill
    \begin{subfigure}[t]{0.48\linewidth}
        \centering
        \includegraphics[width=\linewidth]{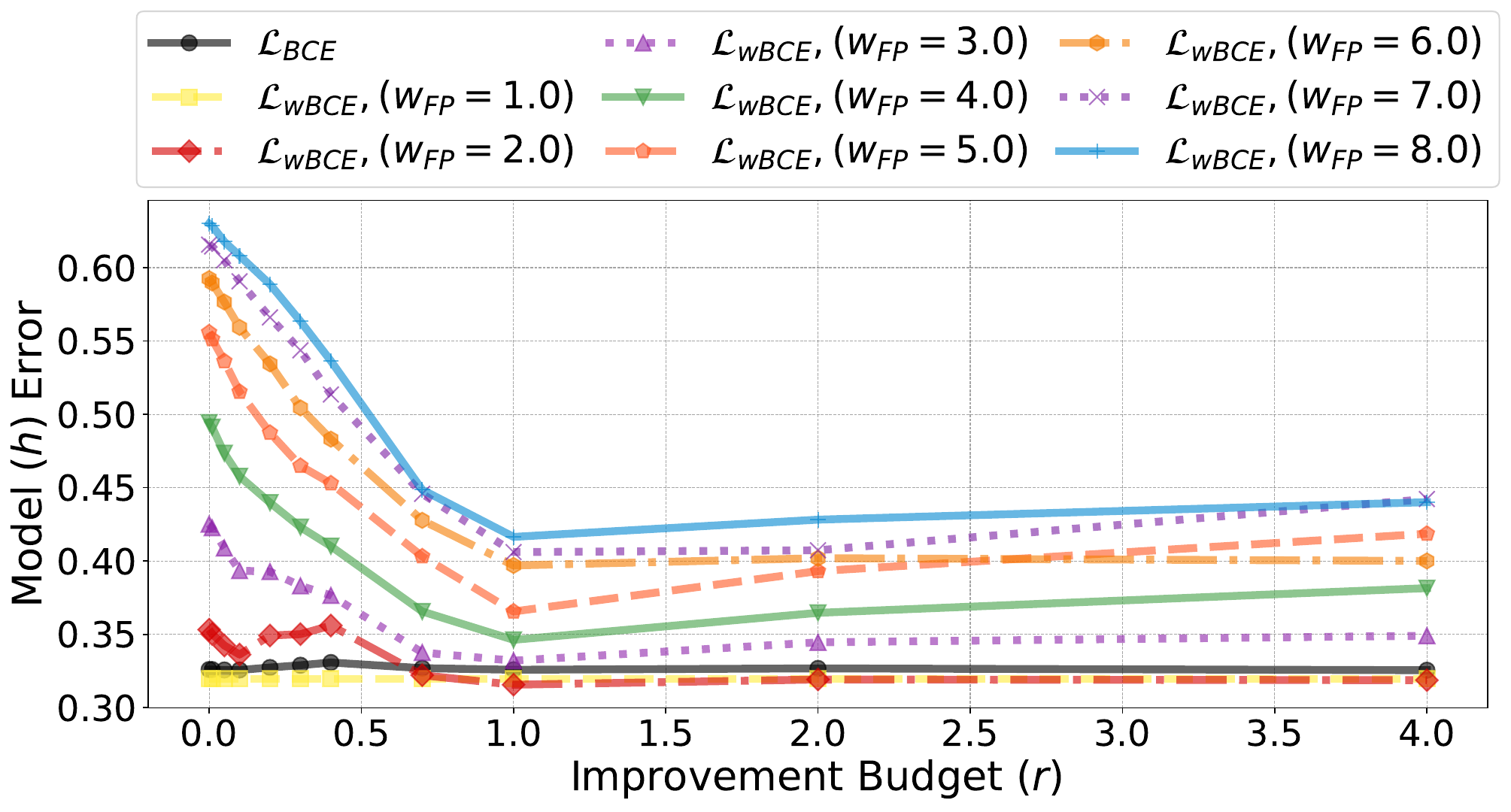}
    \caption{OULAD \(\big(w_{\textrm{FN}}=1.33\big)\)}
    \label{fig:app_oulad_multi_0.5}
    \end{subfigure}
    \hfill
    \begin{subfigure}[t]{0.48\linewidth}
        \centering
        \includegraphics[width=\linewidth]{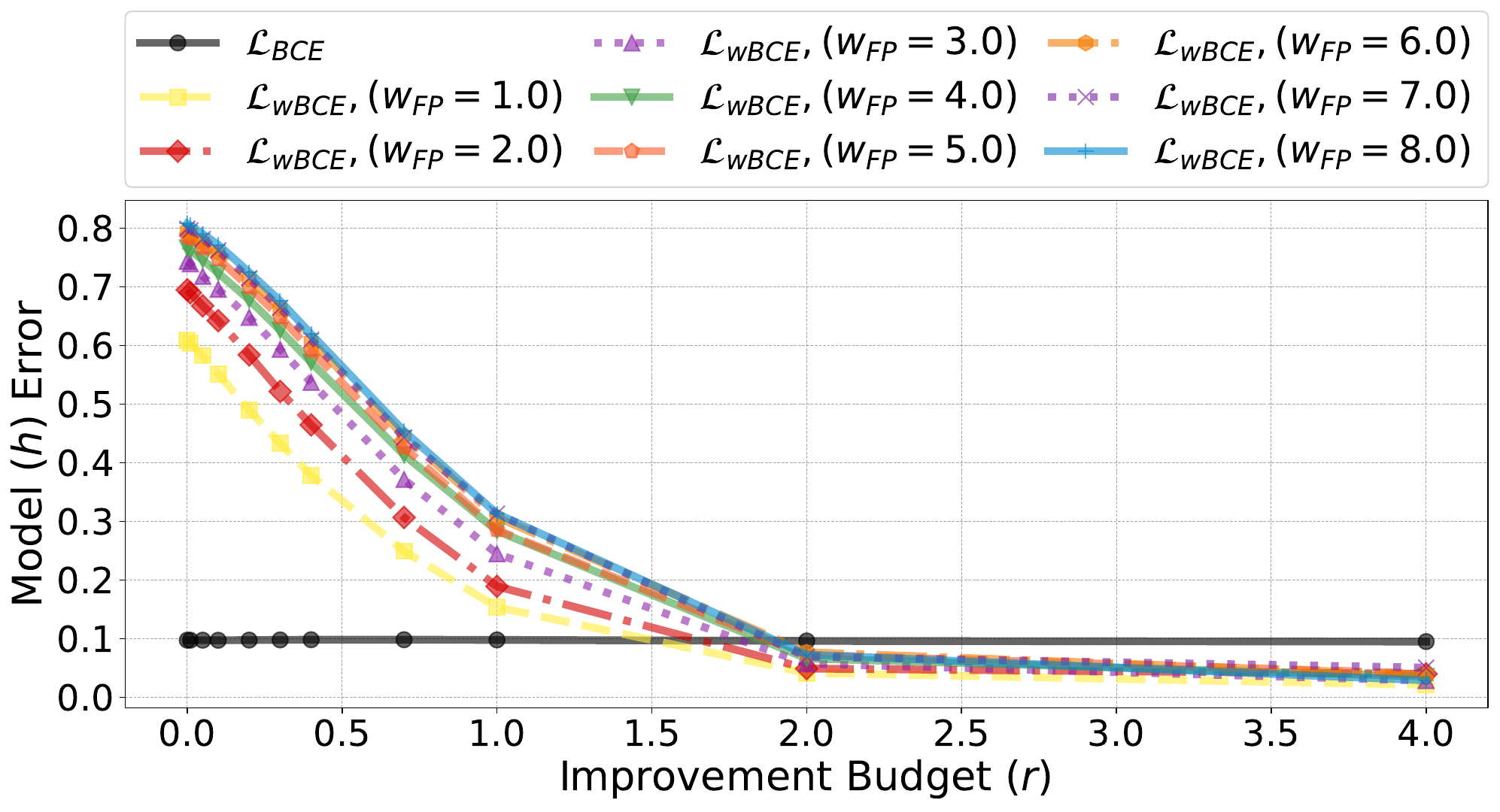}
    \caption{Law school \(\big(w_{\textrm{FN}}=0.009\big)\)}
    \label{fig:app_law_multi_0.5}
    \end{subfigure}
    \hfill
    \begin{subfigure}[t]{0.48\linewidth}
        \centering
        \includegraphics[width=\linewidth]{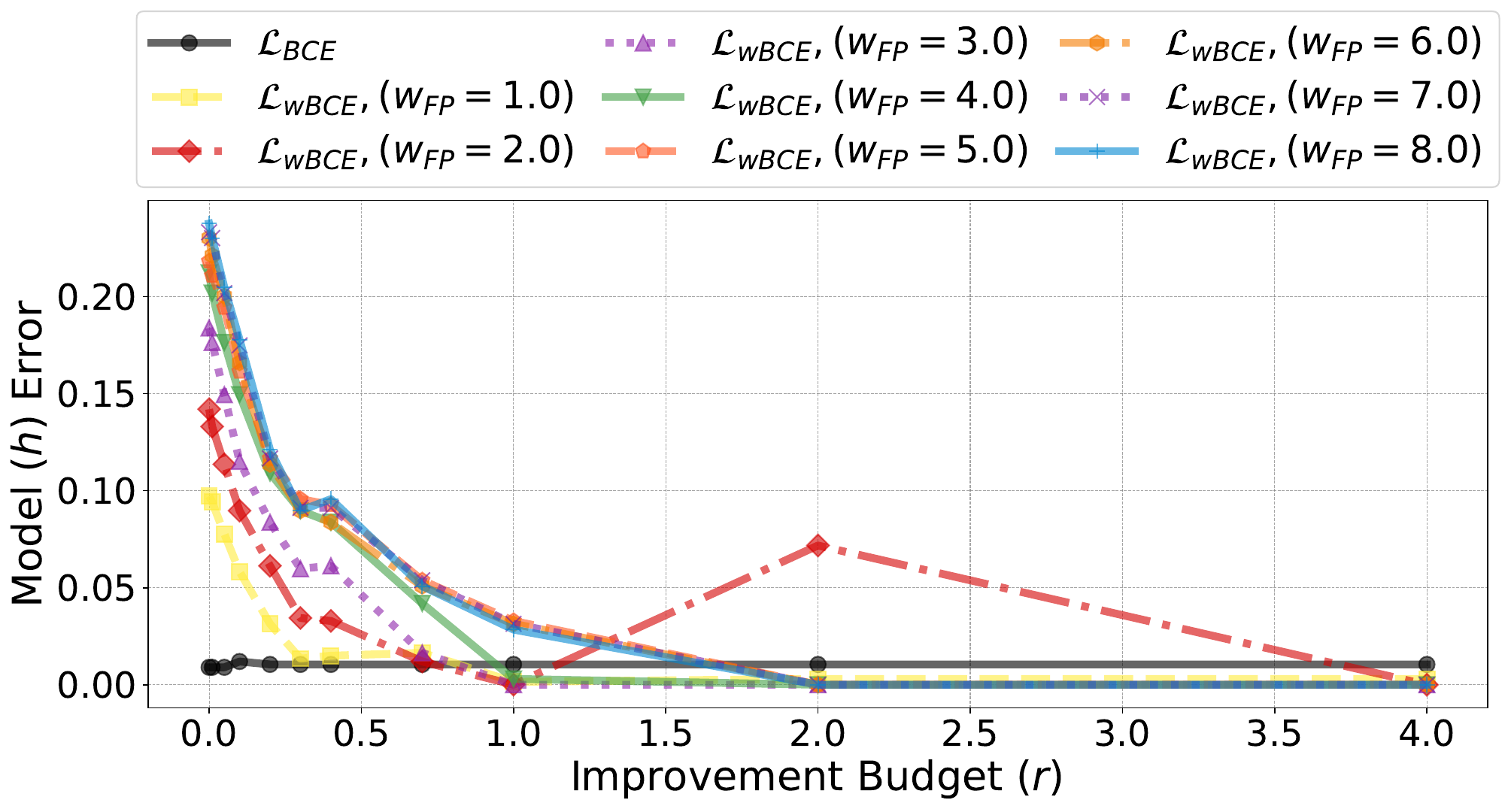}
    \caption{Synthetic \(\big(w_{\textrm{FN}}=0.009\big)\)}
    \label{fig:app_synthetic_multi_0.5}
    \end{subfigure}
    
    \caption[Error drop rate versus budget \(r\), given evaluation with \textbf{multi-defined} \(f^\star\)]{Risk-averse \(\big(\mathcal{L}_{\textrm{wBCE}} \ \text{with} \ w_{\textrm{FP}}=\{i\}_{i=1}^{8}\big)\) and standard (\(\mathcal{L}_{\textrm{BCE}}\)) trained model function (\(h\)) error versus improvement budget (\(r\)) across four datasets (Adult, OULAD, Law school, and Synthetic). For all cases, the threshold for classifying an agent as positive is \(0.5\) and a \textbf{multi-defined} \(f^\star\) model is used to verify successful-ness of improvement. Increasing the improvement budget and classifier risk-aversion (high \(\frac{w_{\textrm{FP}}}{w_{\textrm{FN}}}\)) leads to a faster error drop rate. }
    \label{fig:app_multi_thresh0.5}
\end{figure}

Additionally, performance gains observed with evaluation of successful improvement using the multi-defined \(f^\star\) (cf. Figure~\ref{fig:app_multi_thresh0.5}) were quite similar in trend and gains to those when a singularly-defined model function \(f^\star = f^\star_{2}\) was used (cf. Figure~\ref{fig:app_thresh0.5thresh0.9}, column one (\subref{fig:app_adult_0.5}, \subref{fig:app_oulad_0.5}, \subref{fig:app_law_0.5} and \subref{fig:app_synthetic_0.5})).

Our results indicate that while the \(f^\star\) models achieve \(\sim 100\%\) accuracy when trained and tested on the unsplit dataset \(\mathcal{S}_{T}\), they often overfit, as shown by the LOO scores (cf. Table~\ref{tab:fstarclassifiers_loo}). Nevertheless, they yield comparable performance gains when assessing agents' improvement (cf. Figures~\ref{fig:app_multi_thresh0.5} and \ref{fig:app_oulad_multi_thresh0.5}). 

\chapter{Theory of Classification of Improving-and-Gaming Agents}
\label{app:ocagi_theory}
\section{Missing Proofs of Section~\ref{sec:ocagi_general}}
\label{app:ocagi_hardness}

\begin{proof}[Proof of Proposition~\ref{prop:running-time-greedy-alg}]
The size of $\agents$ is $n$, and within the for loop each computation takes $O(1)$ time since the edges for each  $x_i$ are already sorted. When the flag is set to $1$, at least one point in $\points$ is removed, and when the flag is $0$ at the end of the inner loop, the algorithm returns. Therefore, the outer loop is run at most $|\points|$ times while the inner loop is run $n$ times; resulting in a running time of $O(|\points|n)$.
\end{proof}

\begin{proof}[Proof of Theorem~\ref{thm:atmost_k_game}]
We show the following problem is NP-hard.

\begin{problem}
\label{pr:atmost_k_game}
Suppose we are given a set of $n$ agents where $\init{x}_1, \init{x}_2, \ldots, \init{x}_n$ denote their initial feature vectors, and a set $\points$ of potential criteria also called target points in the linear model. Find a subset $\pfinal \subseteq \points$ that maximizes the number of true positives subject to at most $k$ false positives. 
\end{problem}

We prove the NP-hardness by reducing the Max-$k$-Cover problem with equal-sized sets of size $3$ to this problem. In the Max-$k$-Cover problem, we are given a set $\mathcal{E}$ of elements $e_i$, and sets $S_j \subseteq \mathcal{E}$, and the goal is to select at most $k$ sets out of $S_j$ that maximize the number of elements they cover. 

First, we show how to construct an instance of Problem~\ref{pr:atmost_k_game} from an instance of the Max-$k$-Cover problem. To do so, we determine the number of dimensions, initial positions of the agents, the target points, and the movement costs. Let $n$ be the number of elements of the Max-$k$-Cover instance, we construct an $n+1$-dimensional space where the first $n$ dimensions are improvement and the last dimension is gaming. Consider elements $e_1, e_2, \ldots, e_n$ in the Max-$k$-Cover instance. For every element, we consider an agent; and for every set, we consider an agent and a target point. For $e_i$, the corresponding agent is at initial point $\init{x}_i$, an $n+1$-dimensional vector whose $i^{th}$ and $n+1^{st}$ coordinates are $1$ and the other coordinates are $0$. For every set $S_j$, we consider a target point $\vec{p}_j$ and an agent with initial point $\init{x}_{n+j}$. In $\vec{p}_j$, the coordinates corresponding to the elements in $S_j$ and the $n+1^{st}$ coordinate are set to $1$ and the rest of the coordinates are $0$. In $\init{x}_{n+j}$, the coordinates corresponding to the elements in $S_j$ are set to $1$, the $n+1^{st}$ coordinate is set to $-1$, and the rest of the coordinates are $0$. Finally, let the movement cost in any dimension be  $1/2$. Note that this construction fits into the framework of a linear model and $\f^{\star}:\sum_{j=1}^{n+1} \vec{x}[j] \geq 4$ is the linear threshold function for the truly qualified agents. All the target points $\vec{p}_j$ satisfy the threshold and all the agents are initially unqualified and do not meet the threshold. 

Next, we discuss what target point each agent selects and whether they become truly qualified (true positive) or not (false positive). Because the cost per unit of movement equals $1/2$, each agent can only afford to reach to target points with distance at most $2$. Agents $\init{x}_i$ for $i \in \{1,\ldots, n\}$ can only afford to reach a target point whose $i^{th}$ coordinate is $1$ since they are at distance $2$. They are at distance $3$ to any other target points. Since all dimensions $1,\ldots,n$ are improving dimensions these agents become truly qualified when they reach such target points. Agents $\init{x}_i$ for $i>n$ can only afford to reach $\vec{p}_i$ since they have distance $2$. They have distance more than $2$ to any other target points. Agents $\init{x}_i$ for $i>n$ can only reach to $\vec{p}_i$. To do so, these agents move in a gaming dimension and do not become truly qualified. 

Finally, we show how the solutions of these two problems coincide. Consider the problem of maximizing the true positives subject to including at most $k$ false positives. Including each $\vec{p}_j$ in the final set of target points, $\pfinal$, causes exactly one agent, $\init{x}_j$, to be a false positive. Therefore, having at most $k$ false positive is equivalent to including at most $k$ target points. Maximizing the true positives subject to at most $k$ target points is exactly equivalent to selecting at most $k$ sets that maximize the elements they cover. This completes the reduction.
\end{proof}

\begin{proof}[Proof of Theorem~\ref{thm:relaxed-no-destination-pts}]
We show the following problem is NP-hard.

\begin{problem}
\label{pr:relaxed-no-destination-pts}
Suppose we are given a set of $n$ agents where $\init{x}_1, \init{x}_2, \ldots, \init{x}_n$ denote their initial feature vectors.
Does there exist a set of target points $\pfinal \subseteq \mathbb{R}^d$ for which all the agents become truly qualified?
\end{problem}

We prove the NP-hardness by a reduction from the approximate version of the hitting set with equal-sized sets problem. As an instance of the hitting set problem we are given, $(\mathcal{F}, \mathcal{E})$ where $\mathcal{F} = \{S_1, \cdots, S_m\}$ is a collection of the subsets of $\mathcal{E}=\{e_1,e_2,\cdots,e_n\}$, and each set $S_i$ has a size of $0 < s < n$, and our goal is to find a minimum size set $S^{\star}\subseteq \mathcal{E}$ that intersects every set in $\mathcal{F}$.  
In order to show NP-hardness, we construct an instance of Problem~\ref{pr:relaxed-no-destination-pts} and prove: (1) If all the agents can become true positives by reaching to a set of target points $\pfinal\subseteq \mathbb{R}^d$ that the mechanism designer selects, then there exists a hitting set of size at most $2k$. (2) If there exists a hitting set of size $k$ then the mechanism designer can select a set of target points that encourages all the agents to become true positives. 
Since hitting set and set cover problems are equivalent and approximating set cover within a constant factor is NP-hard~\citep{feige1998threshold}, this implies that Problem~\ref{pr:relaxed-no-destination-pts} is NP-hard.

First, we show how to construct an instance of Problem~\ref{pr:relaxed-no-destination-pts} from an instance of the Hitting Set problem. To do so, we determine the number of dimensions, initial positions of the agents, the movement costs, and a linear threshold function for the truly qualified. Let $n$ be the number of elements of the Hitting Set instance, we construct an $n+1$-dimensional space where the first $n$ dimensions are improvement and the last dimension is gaming. Consider sets $S_1, S_2, \ldots, S_m$ in the Hitting Set instance. For every set $S_i$, we consider agent $i$ at initial point $\init{x}_i$. In $\init{x}_i$, the $j^{th}$ coordinates such that $e_j \in S_i$ is set to $1$. The rest of the first $n$ coordinates are set to $2k$ and the last coordinate is $0$. Also consider an extra agent $m+1$ at initial point $\init{x}_{m+1}$ where all the first $n$ coordinates are $0$ and the last coordinate is $2k(n-s)+s$. Note that for all the agents $\sum_{j=1}^{n+1} \init{x}_i[j] = 2k(n-s)+s$. Let the movement cost in all the dimensions $1 \leq j \leq n$ be  $\frac{1}{2k}$ and in dimension $n+1$ be $c$ such that $\frac{1}{2k(n-s)+s+1}<c<\frac{1}{2k(n-s)+s}$. Let $\f^{\star}:\sum_{j=1}^{n+1} \vec{x}[j] \geq 2k(n-s)+s+2k$. Therefore, all the agents are initially unqualified and at $\ell_1$ distance of $2k$ from $\f^{\star}$.

Now we prove the first direction, i.e., if all the agents can become true positives by reaching to a set of target points $\pfinal\subseteq \mathbb{R}^d$ that the mechanism designer selects, then there exists a hitting set of size at most $2k$. For all $1\leq i\leq m+1$, let $\vec{p}_i \in \pfinal$ denote the target point that $\init{x}_i$ moves to and becomes true positive.

It consists of the following arguments: (i) For all $1\leq i\leq m+1$, agent $i$ receives utility $0$ by reaching to $\vec{p}_i$. (ii) For all $1\leq i\leq m$, agent $m+1$ does not afford to reach to $\vec{p}_i$. 
(iii) If $\vec{p}_{m+1}[j] \leq 1$ for all $e_j \in S_i$, agent $i$ moves to $\vec{p}_{m+1}$ and becomes a false positive. Therefore if all agents improve, for each $1 \leq i \leq m$, there exists $e_j \in S_i$ such that $\vec{p}_{m+1}[j] > 1$. (iv) In order for agent $m+1$ to afford to reach to target point $\vec{p}_{m+1}$, the number of coordinates $1 \leq j \leq m$ with value at least $1$ must be at most $2k$. (v) These elements constitute a hitting set of size at most $2k$.

First, we prove argument (i). Each agent $1 \leq i \leq m+1$, is at $\ell_1$ distance of $2k$ to $\f^{\star}$. To become qualified it needs to move $2k$ in the improvement dimensions.
Since moving for a distance of $2k$ along the improvement dimensions costs a value of $(2k)\times(\frac{1}{2k})=1$, agent $i$ makes a utility of $0$. 

Now, we move to argument (ii). Following up on the previous claim, to reach $\vec{p}_i$, agent $1 \leq i \leq m$ spends all of their movement budget in the improvement dimensions and cannot move a positive amount in the gaming dimension $n+1$. Therefore, $\vec{p}_i[n+1] = 0$ and $\sum_{j=1}^{n} \vec{p}_i[j] = 2k(n-s)+s+2k$. In order for agent $m+1$ to reach such a target point, it needs to move a total of $2k(n-s)+s+2k > 2k$ in the improvement dimensions, which costs more than $1$ and it cannot afford.

Next, we prove argument (iii). Since $\init{x}_{m+1}$ has an $\ell_1$ distance of $2k$ from $f^{\star}$ and costs exactly a value of $1$ to reach there, it can only afford to move along the improvement dimensions. Therefore, $\vec{p}_{m+1}[n+1] \leq 2k(n-s)+s$.
Additionally, for $1 \leq j \leq n$, $\vec{p}_{m+1}[j] \leq 2k$; otherwise, agent $m+1$ cannot afford to reach to $\vec{p}_{m+1}$. Suppose $\vec{p}_{m+1}[j] \leq 1$ for all $e_j \in S_i$. Using this assumption, for agent $i$ to reach $\vec{p}_{m+1}$ it only needs to pay cost of movement in dimension $n+1$, moving $2k(n-s)+s$ units and paying $c$ per unit of movement. Since $(2k(n-s)+s)\times c < 1$, agent $i$ makes a strictly positive utility. Therefore agent $i$ prefers $\vec{p}_{m+1}$ over any other target point that makes it true positive which by argument (i) achieves utility $0$.

Argument (iv) is straight-forward. To achieve non-negative utility each agent can afford to move at most $2k$ units along the improvement dimensions. Therefore, for the target point  $\vec{p}_{m+1}$, the number of coordinates $1 \leq j \leq n$ with value at least $1$ must be at most $2k$.

Argument (v) is a direct implication of the two previous arguments. By argument (iii), for each $1 \leq i \leq m$ there is an element $e_j \in S_i$ such that $\vec{p}_{m+1}[j] > 1$. By argument (iv), the number of coordinates $j \leq n$ such that $\vec{p}_{m+1}[j] > 1$ is at most $2k$ since otherwise agent $m+1$ cannot afford to reach to $\vec{p}_{m+1}$. Therefore, elements $e_j$ such that $\vec{p}_{m+1}[j] > 1$ constitute a hitting set of size at most $2k$.

Now, we prove the reverse direction: if there exists a hitting set $S^{\star}$ of size $k$, the mechanism designer can select a set of target points that encourages all the agents to become true positives. To do so, we construct a set of target points $\pfinal = \{\vec{p}_1, \ldots, \vec{p}_{m+1}\}$ that makes every agent to become true positive. For each agent $i$, $1 \leq i \leq m$, put a target point $\vec{p}_i$ whose first coordinate is $2k$ more than $\init{x}_i$. For agent $m+1$, put a target point $\vec{p}_{m+1}$ whose coordinates $j$ where $e_j \in S^{\star}$ are set to $2$ and the remaining agree with $\init{x}_{m+1}$. Each target point $\init{x}_i$ is set such that $\sum_{j=1}^{n+1} \init{x}_i[j] = 2k(n-s)+s+2k$. In order to show that every agent is able to improve, we argue that: (i) For all $1\leq i \leq m$, agent $i$ can afford to move to $\vec{p}_i$. Additionally, if agent $i$ moves to any of the target points $\vec{p}_j$ where $1\leq j\leq m$, it becomes true positive.  (ii) For all $1\leq i\leq m$, agent $i$ cannot reach to $\vec{p}_{m+1}$. (iii) Agent $m+1$ moves to $\vec{p}_{m+1}$ and becomes true positive. 

First, we prove argument (i): Agent $i$ is at a distance of $2k$ from $\vec{p}_i$. It can afford to reach to $\vec{p}_i$ by paying a cost of $(2k)\times(\frac{1}{2k})=1$ and become true positive. In addition, if it moves to any of the other target points $\vec{p}_j$ where $1\leq j\leq m$, since it has only moved along the improvement dimensions, it would become true positive.

Next, we prove argument (ii): We know that for each $S_i$, there exists an element $e_j\in S_i$ such that $\vec{p}_{m+1}[j]=2$. As a result, the $\ell_1$ distance of $\init{x}_i$ and $\vec{p}_{m+1}$ is at least $(2k(n-s)+s+1)\times c > 1$. Therefore, for each $1\leq i\leq m$, $\init{x}_i$ cannot afford to reach to $\vec{p}_{m+1}$.

Finally, we prove argument (iii): First, we argue that agent $m+1$ cannot afford to reach to any of the target points $\vec{p}_i$ where $1\leq i\leq m$. For each target point $\vec{p}_i$ where $1\leq i\leq m$, $\vec{p}_i[n+1] = 0$ and $\sum_{j=1}^{n} \vec{p}_i[j] = 2k(n-s)+s+2k$. In order for agent $m+1$ to reach such a target point, it needs to move a total of $2k(n-s)+s+2k > 2k$ units in the improvement dimensions, which costs more than $1$ and it cannot afford. In addition, agent $m+1$ can afford to move to $\vec{p}_{m+1}$, and by reaching there it becomes true positive.

As a result of the above arguments, given a hitting set of size $k$, the mechanism designer can select a set of target points that encourages all the agents to become true positives.

Combining the above two directions, shows that the problem of selecting a set of target points for which all the agents become truly qualified is NP-hard.

\end{proof}

\section{Missing Proofs of Section~\protect\ref{sec:ocagi_two-dimension}}

\subsection{Proof of Lemma~\protect\ref{lemma:2-dim-triangle-inequalities}}
\label{appendix:proof-lemma-2-dim-triangle-inequalities}

\begin{proof}
Initially, if $\vec{p}[2]< \init{x}_{i}[2]$, $\vec{p}$ is replaced with $(\vec{p}[1], \init{x}_i[2])$. By doing so, $cost(\init{x}_j, \vec{p})$ would not decrease. Hence, without loss of generality, we can assume $\vec{p}[2]\geq \init{x}_{i}[2]$.

First, we show that $cost(\init{x}_j,\vec{p})\leq cost(\init{x}_j,\vec{x}_{i,min})+cost(\vec{x}_{i,min},\vec{p})$, where the inequality holds when $\vec{x}_{i,min}[1]<\init{x}_j[1]\leq \vec{p}[1]$.
\begin{align}
&cost(\init{x}_j,\vec{x}_{i,min})+cost(\vec{x}_{i,min},\vec{p})\\
&=max\Big\{\vec{x}_{i,min}[1]-\init{x}_{j}[1],0\Big\}+\Big(\vec{x}_{i,min}[2]-\init{x}_j[2]\Big)+\Big(\vec{p}[1]-\vec{x}_{i,min}[1]\Big)+\Big(\vec{p}[2]-\vec{x}_{i,min}[2]\Big)\\
&=max\Big\{\vec{x}_{i,min}[1]-\init{x}_{j}[1],0\Big\}+\Big(\vec{p}[1]-\vec{x}_{i,min}[1]\Big)+\Big(\vec{p}[2]-\init{x}_j[2]\Big)
\end{align}
If $\init{x}_j[1]\leq \vec{x}_{i,min}[1]$, the last equation above gets equal to $\Big(\vec{p}[1]-\init{x}_j[1]\Big)+\Big(\vec{p}[2]-\init{x}_j[2]\Big) = cost(\init{x}_j,\vec{p})$. Otherwise, $\init{x}_j[1] > \vec{x}_{i,min}[1]$ and the last equation above gets equal to $\Big(\vec{p}[1]-\vec{x}_{i,min}[1]\Big)+\Big(\vec{p}[2]-\init{x}_j[2]\Big)> \Big(\vec{p}[1]-\init{x}_{j}[1]\Big)+\Big(\vec{p}[2]-\init{x}_j[2]\Big) = cost(\init{x}_j,\vec{p})$. In any case, $cost(\init{x}_j,\vec{p})\leq cost(\init{x}_j,\vec{x}_{i,min})+cost(\vec{x}_{i,min},\vec{p})$.

Next we argue that $cost(\vec{x}_{i,min},\vec{p})= cost(\vec{x}_{i,min},\vec{q})$. First, since $\vec{p}[1]\geq \vec{x}_{i,min}[1]$ and $\vec{p}[2]\geq \vec{x}_{i,min}[2]$, then $cost(\vec{x}_i,\vec{p}) = cost(\vec{x}_i,\vec{x}_{i,min})+ cost(\vec{x}_{i,min},\vec{p})$. Similarly, $cost(\vec{x}_i,\vec{q}) = cost(\vec{x}_i,\vec{x}_{i,min})+ cost(\vec{x}_{i,min},\vec{q})$. Since $cost(\vec{x}_i, \vec{p}) = cost(\vec{x}_i, \vec{q})$, it is the case that $cost(\vec{x}_{i,min},\vec{p})= cost(\vec{x}_{i,min},\vec{q})$.

Therefore,
\begin{align}
& cost(\init{x}_j,\vec{p})\leq cost(\init{x}_j,\vec{x}_{i,min})+cost(\vec{x}_{i,min},\vec{p})\\
&\leq cost(\init{x}_j,\vec{x}_{i,min})+cost(\vec{x}_{i,min},\vec{q})\\
&=max\Big\{\vec{x}_{i,min}[1]-\init{x}_j[1], 0\Big\}+ max\Big\{\vec{x}_{i,min}[2]-\init{x}_j[2], 0\Big\}+\\
&max\Big\{\vec{q}[1]-\vec{x}_{i,min}[1],0\Big\}+max\Big\{\vec{q}[2]-\vec{x}_{i,min}[2],0\Big\}\\
&=max\Big\{\vec{x}_{i,min}[1]-\init{x}_j[1], 0\Big\}+ \Big(\vec{x}_{i,min}[2]-\init{x}_j[2]\Big)+\Big(\vec{q}[2]-\vec{x}_{i,min}[2]\Big)\\
&=max\Big\{\vec{x}_{i,min}[1]-\init{x}_j[1], 0\Big\}+ \Big(\vec{q}[2]-\init{x}_j[2]\Big)\\
&=max\Big\{\vec{q}[1]-\init{x}_j[1], 0\Big\}+ \Big(\vec{q}[2]-\init{x}_j[2]\Big)\\
&= cost(\init{x}_j, \vec{q})
\end{align}
\end{proof}

\chapter{Empirical Study on the Classification of Improving-and-Gaming Agents}
\label{app:ocagi_exp1}
\section{Supplemental Results for Section~\ref{subsec:ocagiexp_results_errorvutility}}

\begin{figure*}[ht!]
    \centering

    \begin{subfigure}[t]{0.47\linewidth}
        \centering
        \includegraphics[width=\linewidth]{ocagi_exp/ocagi_figures/errordrop_thresh_0.5_wfn_0.001wfpvaried_lp_linf_improveonly_no_fstar_c2_adult.pdf}
    \caption{The \textbf{error drop} rate variations when agents respond to a BCE-trained model and wBCE-trained models with weight configurations \(\big({w_\textrm{FN}=0.001}, w_\textrm{FP}=\{i\}_{i=1}^{8}\big)\)}
    \label{fig:ocagi_adult_error_0.5linfno}
    \end{subfigure}
    \hfill
    \begin{subfigure}[t]{0.47\linewidth}
        \centering
        \includegraphics[width=\linewidth]{ocagi_exp/ocagi_figures/tpfp_percentage_thresh_0.5_wfn_0.001wfpvaried_lp_linf_improveonly_no_fstar_c2_adult.pdf}
    \caption{The \textbf{utility score} variations when agents respond to a BCE-trained model and wBCE-trained models with weight configurations \(\big({w_\textrm{FN}=0.001}, w_\textrm{FP}=\{i\}_{i=1}^{8}\big)\)}
    \label{fig:ocagi_adult_tpfp_0.5linfno}
    \end{subfigure}

    \vspace{1.2em}

    \begin{subfigure}[t]{0.47\linewidth}
        \centering
        \includegraphics[width=\linewidth]{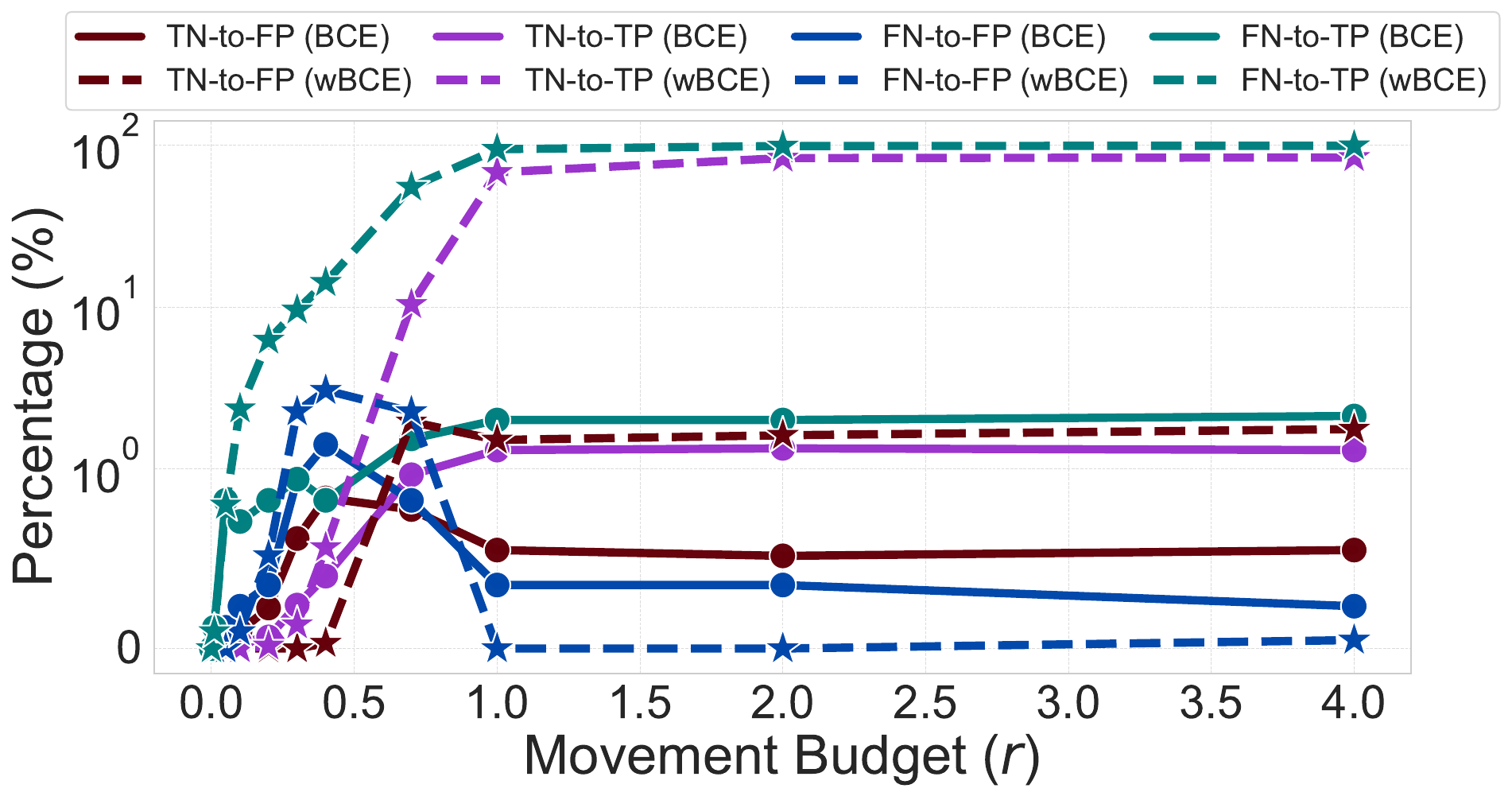}
    \caption{Percentage of agents that transition \textbf{from TN/FN to TP/FP} in response to a BCE- and wBCE-trained models with \(\big(\mathbf{w_\textrm{FN}=0.001, w_\textrm{FP}=1.0}\big)\)}
    \label{fig:ocagi_adult_move_0.5linf1.0no}
    \end{subfigure}
    \hfill
    \begin{subfigure}[t]{0.47\linewidth}
        \centering
        \includegraphics[width=\linewidth]{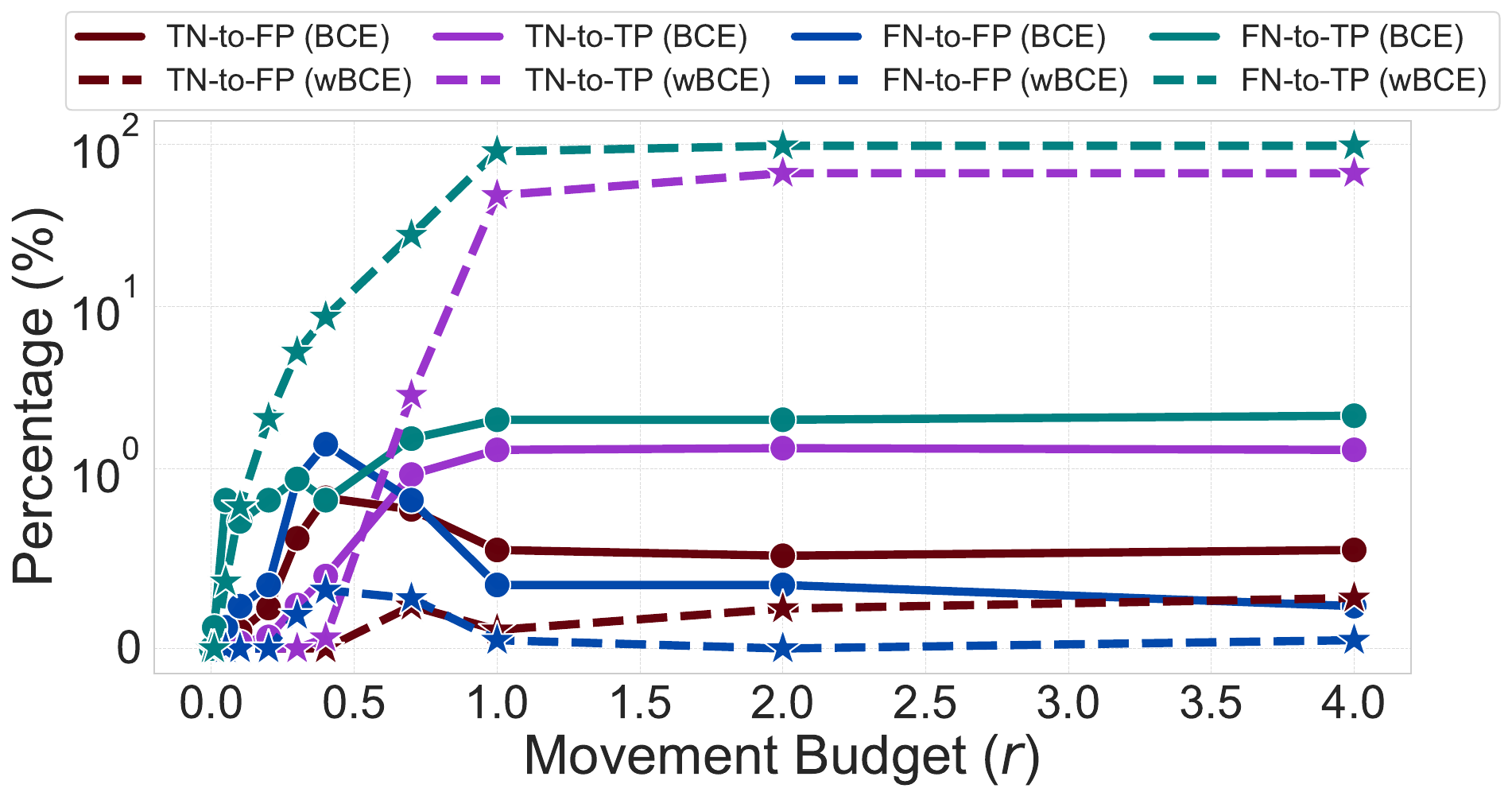}
    \caption{Percentage of agents that transition \textbf{from TN/FN to TP/FP} in response to a BCE- and wBCE-trained models with \(\big(\mathbf{w_\textrm{FN}=0.001, w_\textrm{FP}=8.0}\big)\)}
    \label{fig:ocagi_adult_move_0.5linf8.0no}
    \end{subfigure}

    \caption[On the Adult dataset, a comparative analysis to study the effect of level of risk-aversion]{On the \textbf{Adult} dataset, we compare the performance of a BCE-trained model with wBCE-trained models under different weights \(\big(w_\textrm{FN}=0.001, w_\textrm{FP}=\{i\}_{i=1}^{8}\big)\), focusing on the error drop rate and the \(\textrm{utility}\) score increment rate as agents modify their features (either by gaming or improving) in response to the models. Figures~\subref{fig:ocagi_adult_error_0.5linfno} and \subref{fig:ocagi_adult_tpfp_0.5linfno} illustrate the model's error reduction and the corresponding \(\textrm{utility}\) score increment as a function of the agents' movement budget \(r\). Figures~\subref{fig:ocagi_adult_move_0.5linf1.0no} and \subref{fig:ocagi_adult_move_0.5linf8.0no} show the percentage of agents transitioning between states (e.g., from true negative to false positive) after modifying their features in response to a BCE-trained models and wBCE-trained models with weight settings of \(\big(w_\textrm{FN}=0.001, w_\textrm{FP}=1.0\big)\) and \(\big(w_\textrm{FN}=0.001, w_\textrm{FP}=8.0\big)\), respectively. Zero error doesn't imply \(100\%\) \(\textrm{utility}\) score. In all cases agents move within an \(\ell_{\infty}\) ball and they are classified as positive if the probability is higher than \(0.5\).}
    \label{fig:adult_errors_tpfp_th0.5linf}
\end{figure*}

\begin{figure*}[ht!]
    \centering
    \begin{subfigure}[t]{0.47\linewidth}
        \includegraphics[width=\linewidth]{ocagi_exp/ocagi_figures/errordrop_thresh_0.5_wfn_0.009wfpvaried_lp_linf_improveonly_no_fstar_c2_law.pdf}
        \caption{The \textbf{error drop} rate variations when agents respond to a BCE-trained model and wBCE-trained models with weight configurations \(\big({w_\textrm{FN}=0.009}, w_\textrm{FP}=\{i\}_{i=1}^{8}\big)\)}
        \label{fig:ocagi_law_error_0.5linfno}
    \end{subfigure}
    \hfill
    \begin{subfigure}[t]{0.47\linewidth}
            \centering
            \includegraphics[width=\linewidth]{ocagi_exp/ocagi_figures/tpfp_percentage_thresh_0.5_wfn_0.009wfpvaried_lp_linf_improveonly_no_fstar_c2_law.pdf}
    \caption{The \textbf{utility score} variations when agents respond to a BCE-trained model and wBCE-trained models with weight configurations \(\big({w_\textrm{FN}=0.009}, w_\textrm{FP}=\{i\}_{i=1}^{8}\big)\)}
    \label{fig:ocagi_law_tpfp_0.5linfno}
    \end{subfigure}

    \vspace{1.2em}

    \begin{subfigure}[t]{0.47\linewidth}
        \centering
        \includegraphics[width=\linewidth]{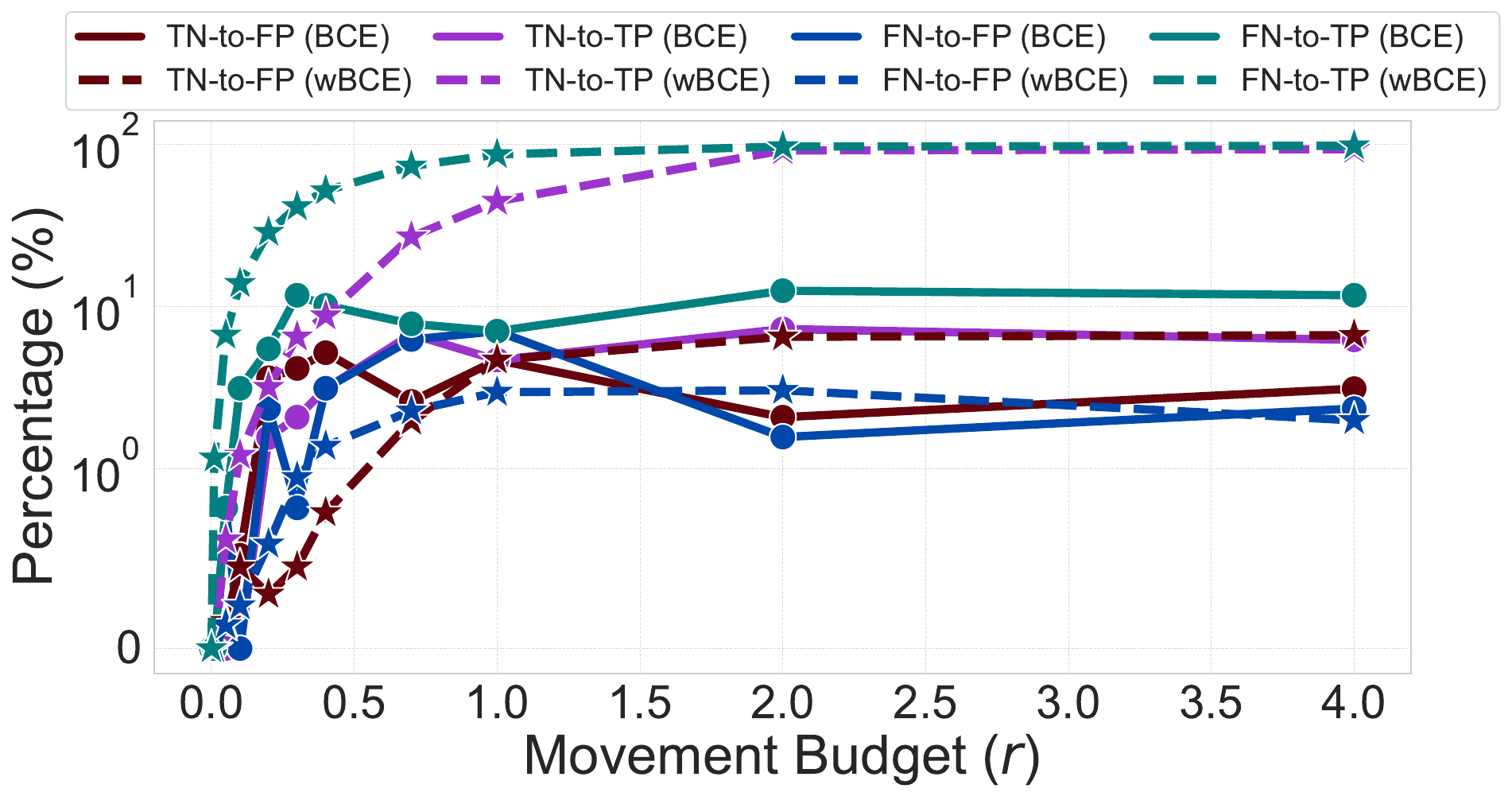}
    \caption{Percentage of agents that transition \textbf{from TN/FN to TP/FP} in response to a BCE- and wBCE-trained models with \(\big(\mathbf{w_\textrm{FN}=0.009, w_\textrm{FP}=1.0}\big)\)}
    \label{fig:ocagi_law_move_0.5linf1.0no}
    \end{subfigure}
    \hfill
    \begin{subfigure}[t]{0.47\linewidth}
        \centering
        \includegraphics[width=\linewidth]{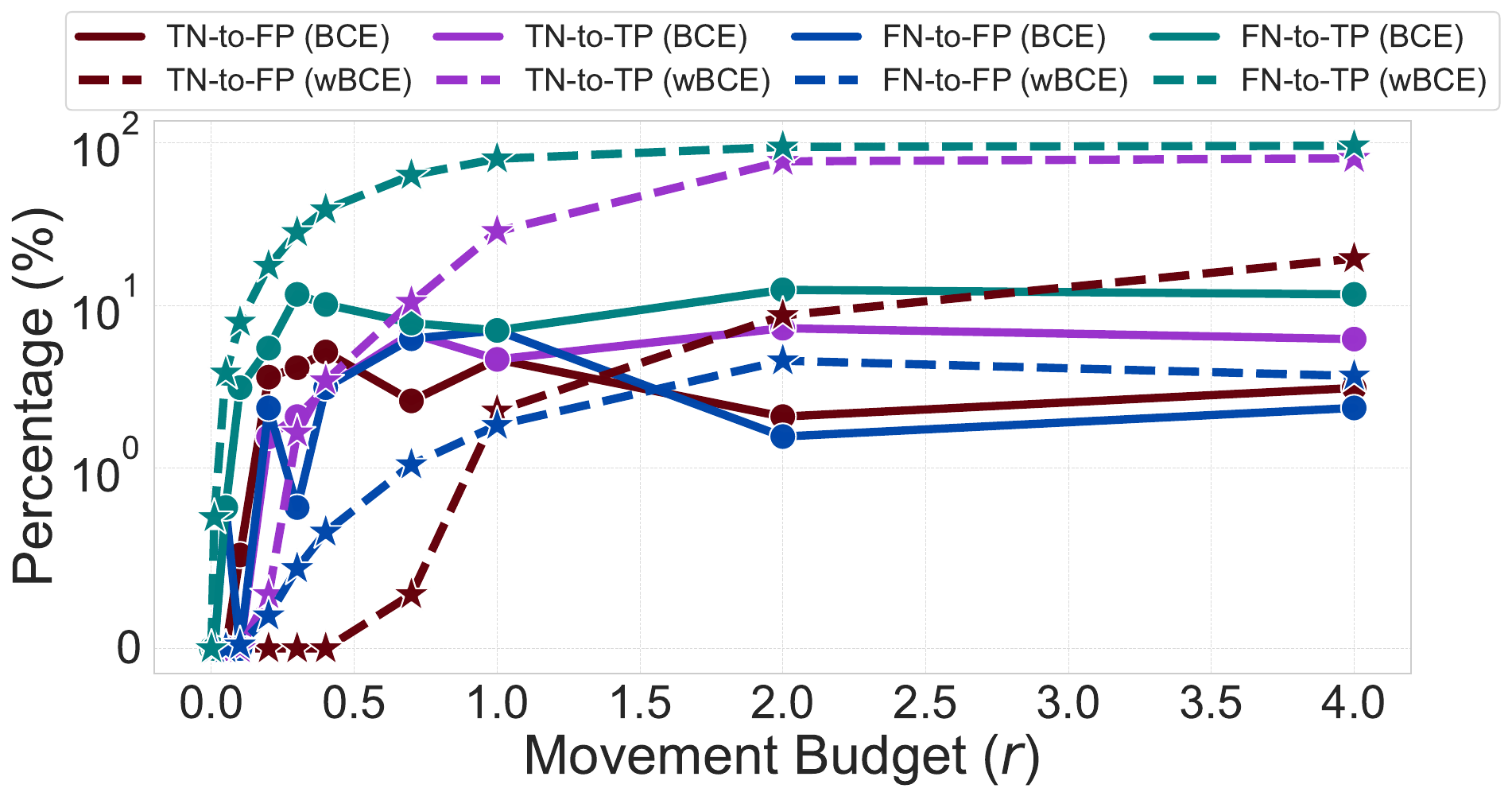}
    \caption{Percentage of agents that transition \textbf{from TN/FN to TP/FP} in response to a BCE- model and wBCE-trained models with \(\big(\mathbf{w_\textrm{FN}=0.009, w_\textrm{FP}=6.0}\big)\)}
    \label{fig:ocagi_law_move_0.5linf6.0no}
    \end{subfigure}

    \caption[On the Law School dataset, a comparative analysis to study the effect of level of risk-aversion]{On the \textbf{Law School} dataset, we compare the performance of a BCE-trained model with wBCE-trained models under different weights \(\big(w_\textrm{FN}=0.009, w_\textrm{FP}=\{i\}_{i=1}^{8}\big)\), focusing on the error drop rate and the \(\textrm{utility}\) score increment rate as agents modify their features (either by gaming or improving) in response to the models. Figures~\subref{fig:ocagi_law_error_0.5linfno} and \subref{fig:ocagi_law_tpfp_0.5linfno} illustrate the model's error reduction and the corresponding \(\textrm{utility}\) score increment as a function of the agents' movement budget \(r\). Figures~\subref{fig:ocagi_law_move_0.5linf1.0no} and \subref{fig:ocagi_law_move_0.5linf6.0no} show the percentage of agents transitioning between states (e.g., from true negative to false positive) after modifying their features in response to a BCE-trained models and wBCE-trained models with weight settings of \(\big(w_\textrm{FN}=0.009, w_\textrm{FP}=1.0\big)\) and \(\big(w_\textrm{FN}=0.009, w_\textrm{FP}=6.0\big)\), respectively. Zero error doesn't imply \(100\%\) \(\textrm{utility}\) score. In all cases agents move within an \(\ell_{\infty}\) ball and they are classified as positive if the probability is higher than \(0.5\).}
    \label{fig:law_errors_tpfp_th0.5linf}
\end{figure*}

\begin{figure*}[ht!]
    \centering

    \begin{subfigure}[t]{0.47\linewidth}
        \centering
        \includegraphics[width=\linewidth]{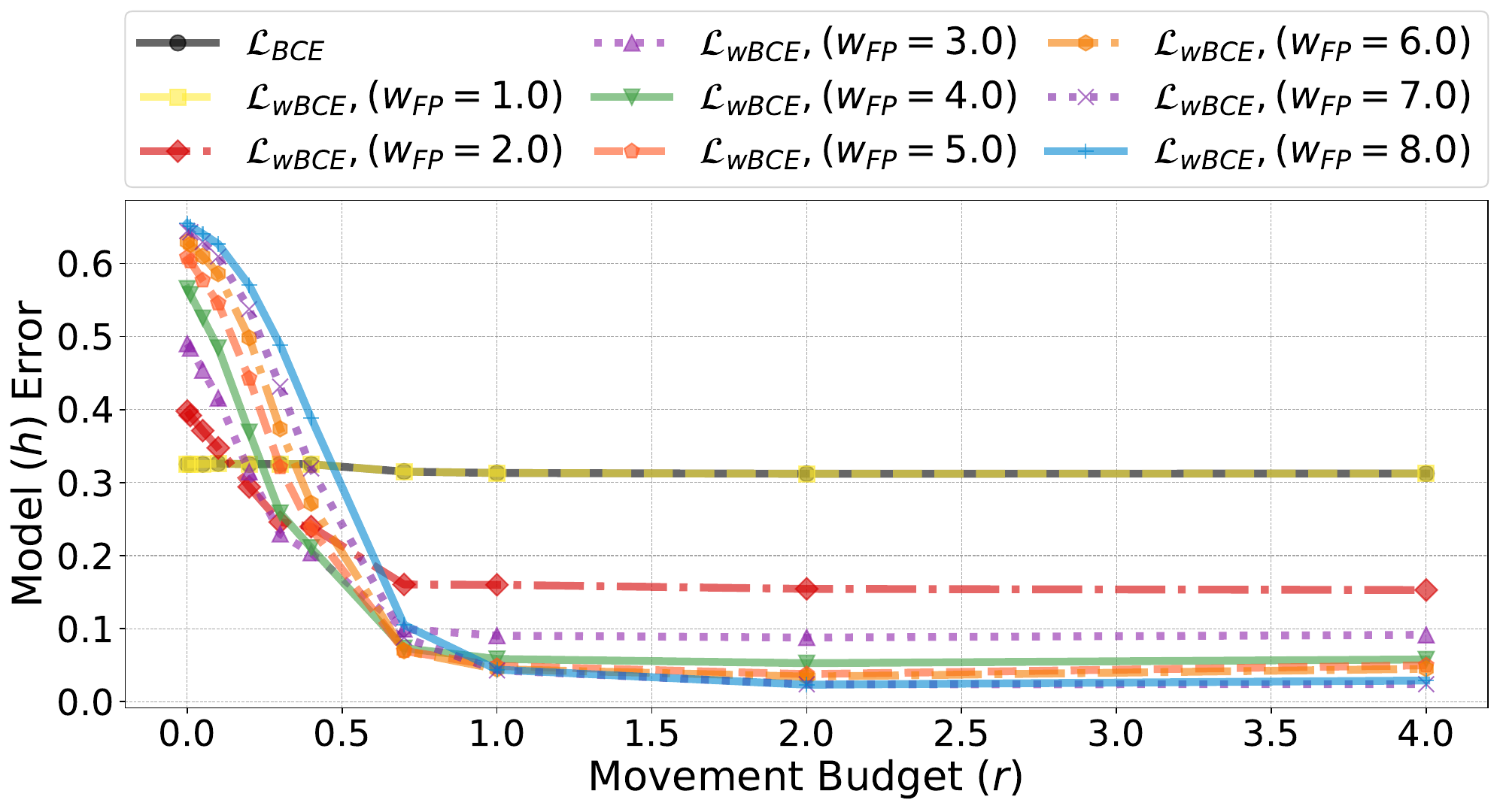}
    \caption{The \textbf{error drop} rate variations when agents respond to a BCE-trained model and wBCE-trained models with weight configurations \(\big({w_\textrm{FN}=1.0}, w_\textrm{FP}=\{i\}_{i=1}^{8}\big)\)}
    \label{fig:ocagi_oulad_error_0.5linfno}
    \end{subfigure}
    \hfill
    \begin{subfigure}[t]{0.47\linewidth}
        \centering
        \includegraphics[width=\linewidth]{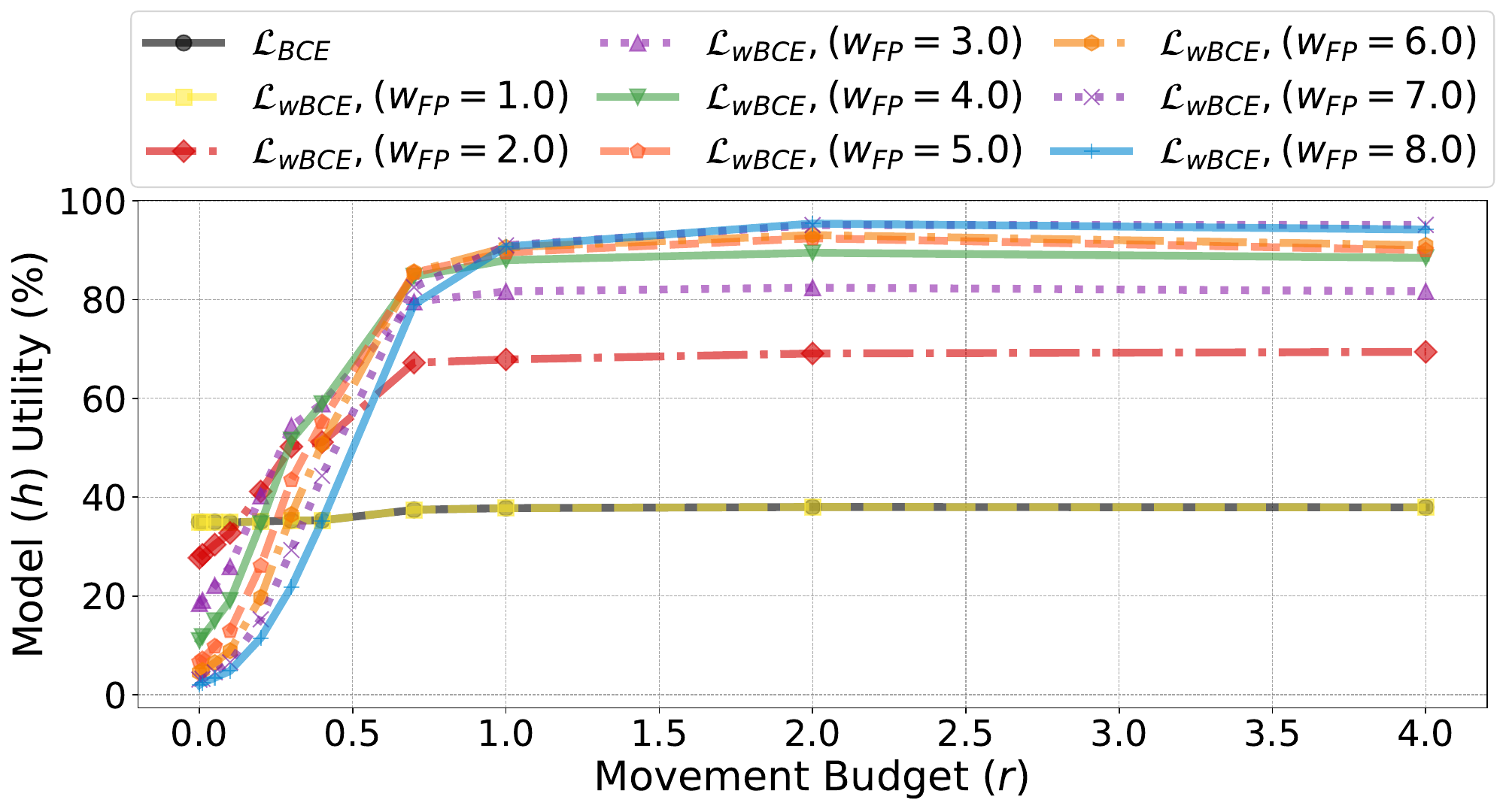}
    \caption{The \textbf{utility score} variations when agents respond to a BCE-trained model and wBCE-trained models with weight configurations \(\big({w_\textrm{FN}=1.0}, w_\textrm{FP}=\{i\}_{i=1}^{8}\big)\)}
    \label{fig:ocagi_oulad_tpfp_0.5linfno}
    \end{subfigure}

    \vspace{1.2em}

    \begin{subfigure}[t]{0.47\linewidth}
        \centering
        \includegraphics[width=\linewidth]{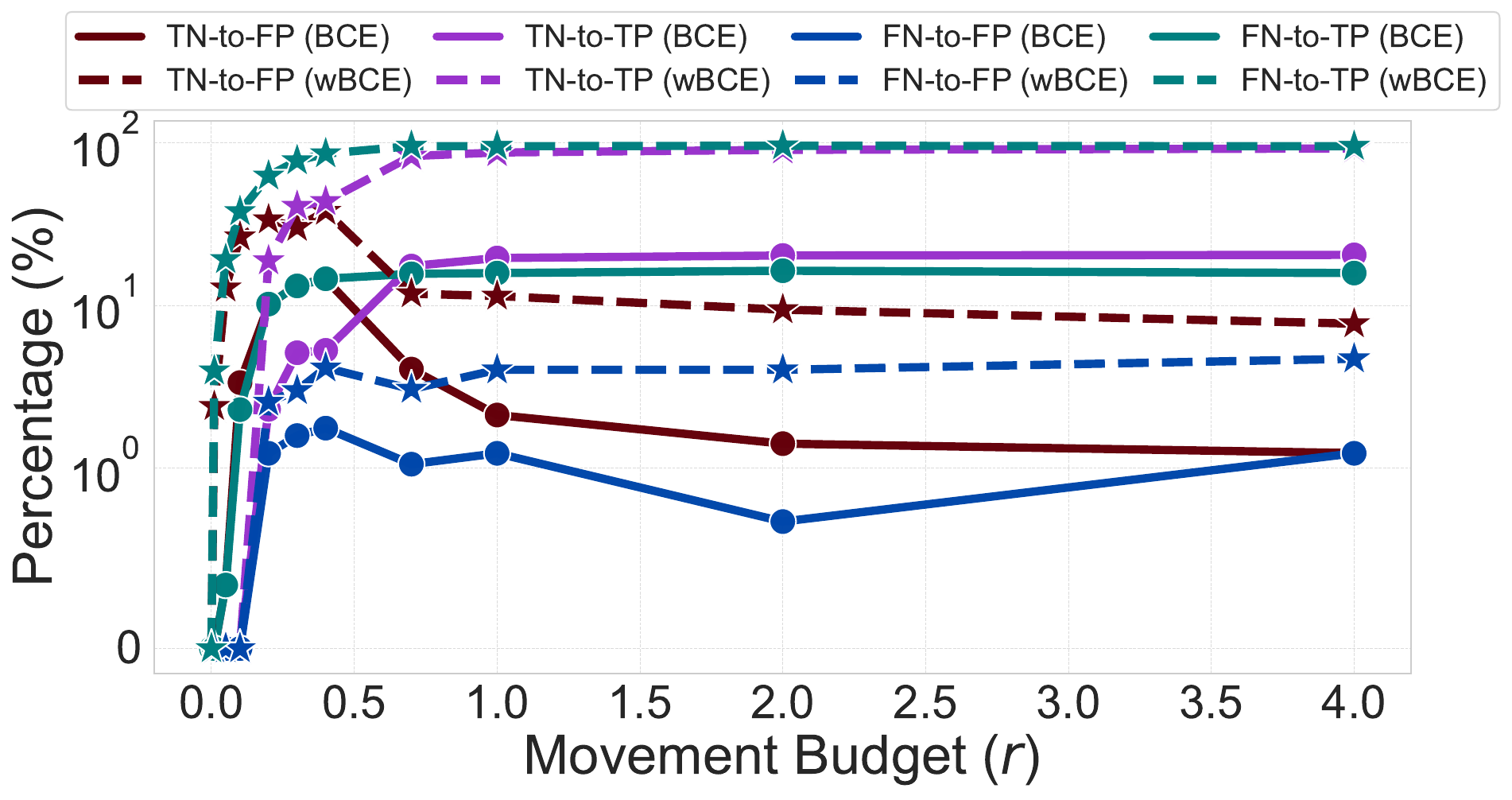}
    \caption{Percentage of agents that transition \textbf{from TN/FN to TP/FP} in response to a BCE- and wBCE-trained models with \(\big(\mathbf{w_\textrm{FN}=1.0, w_\textrm{FP}=2.0}\big)\)}
    \label{fig:ocagi_oulad_move_0.5linf2.0no}
    \end{subfigure}
    \hfill
    \begin{subfigure}[t]{0.47\linewidth}
        \centering
        \includegraphics[width=\linewidth]{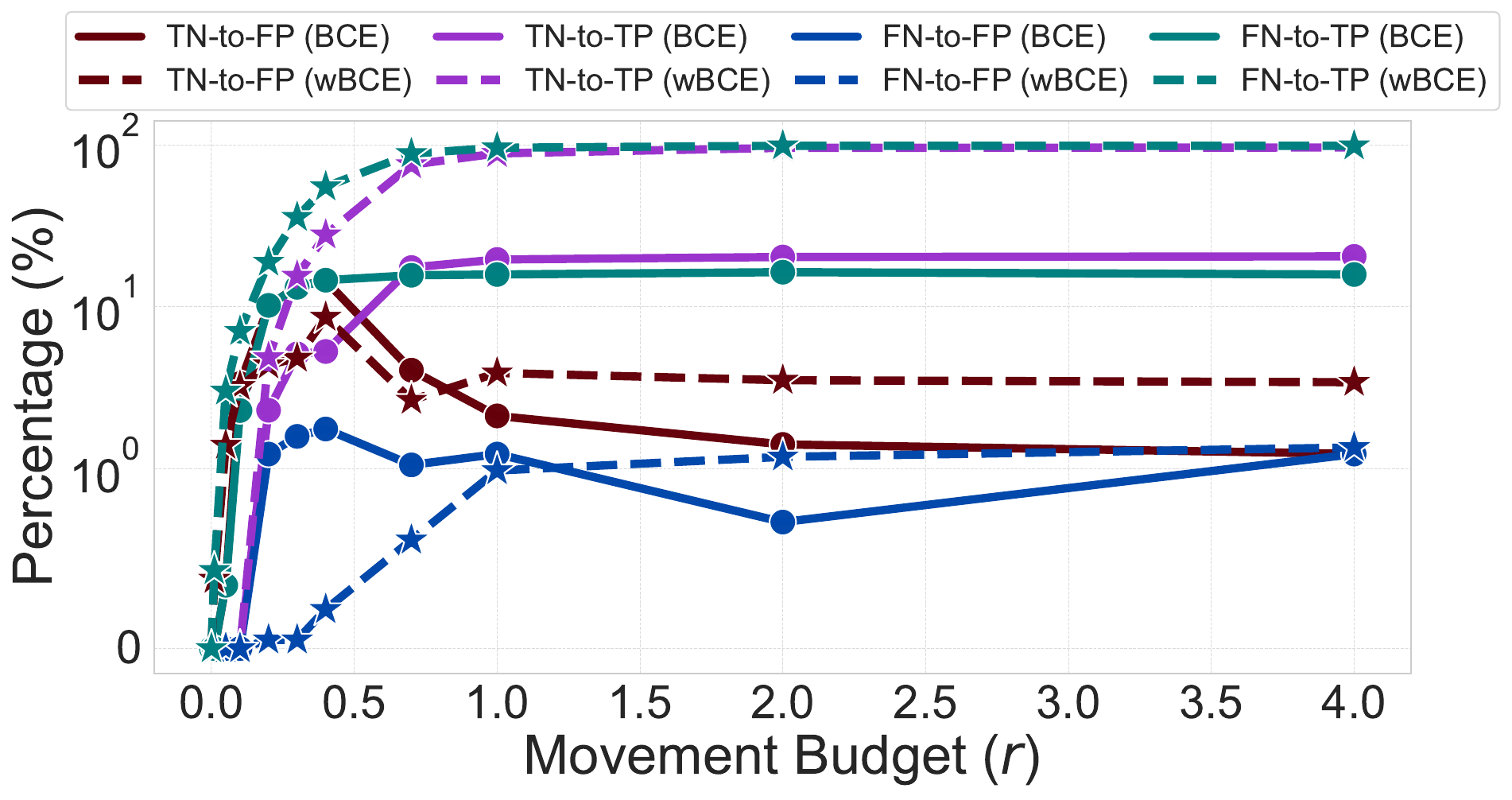}
    \caption{Percentage of agents that transition \textbf{from TN/FN to TP/FP} in response to a BCE- and wBCE-trained models with \(\big(\mathbf{w_\textrm{FN}=1.0, w_\textrm{FP}=7.0}\big)\)}
    \label{fig:ocagi_oulad_move_0.5linf7.0no}
    \end{subfigure}

    \caption[On the OULAD dataset, a comparative analysis to study the effect of level of risk-aversion]{On the \textbf{OULAD} dataset, we compare the performance of a BCE-trained model with wBCE-trained models under different weights \(\big(w_\textrm{FN}=1.0, w_\textrm{FP}=\{i\}_{i=1}^{8}\big)\), focusing on the error drop rate and the \(\textrm{utility}\) score increment rate as agents modify their features (either by gaming or improving) in response to the models. Figures~\subref{fig:ocagi_oulad_error_0.5linfno} and \subref{fig:ocagi_oulad_tpfp_0.5linfno} illustrate the model's error reduction and the corresponding \(\textrm{utility}\) score increment as a function of the agents' movement budget \(r\). Figures~\subref{fig:ocagi_oulad_move_0.5linf2.0no} and \subref{fig:ocagi_oulad_move_0.5linf7.0no} show the percentage of agents transitioning between states (e.g., from true negative to false positive) after modifying their features in response to a BCE-trained models and wBCE-trained models with weight settings of \(\big(w_\textrm{FN}=1.0, w_\textrm{FP}=2.0\big)\) and \(\big(w_\textrm{FN}=1.0, w_\textrm{FP}=7.0\big)\), respectively. Zero error doesn't imply \(100\%\) \(\textrm{utility}\) score. In all cases agents move within an \(\ell_{\infty}\) ball and they are classified as positive if the probability is higher than \(0.5\).}
    \label{fig:oulad_errors_tpfp_th0.5linf}
\end{figure*}

\clearpage

\section{Supplemental Results for Section~\ref{subsec:ocagiexp_results_profits}}

\begin{figure*}[htb!]
    \centering

    \begin{subfigure}[t]{0.47\linewidth}
        \centering
        \includegraphics[width=\linewidth]{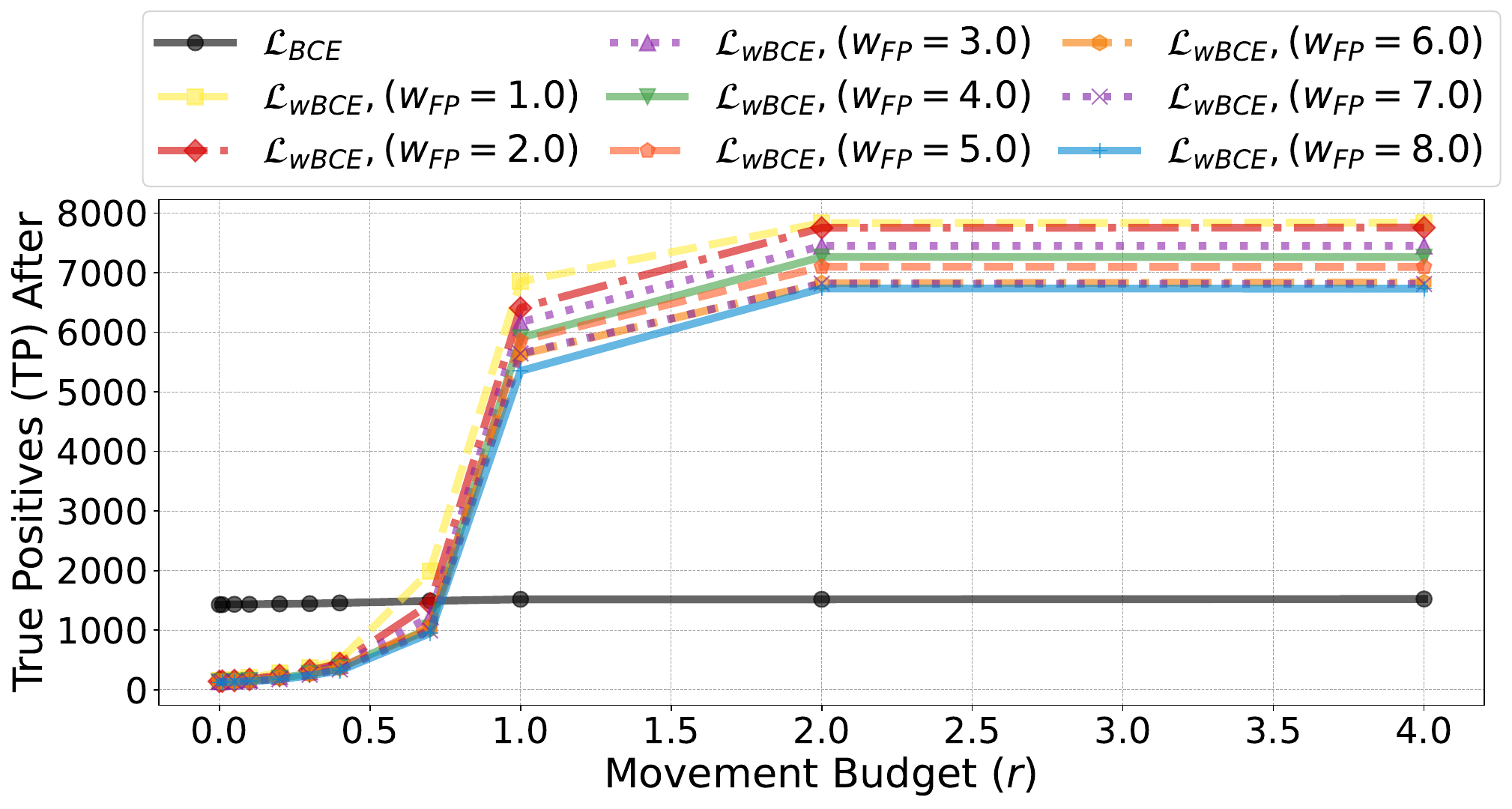}
    \caption{On the \textbf{Adult} dataset, the \textbf{number of true positives} when agents respond to a BCE-trained model and wBCE-trained models with weight configurations \(\big(\mathbf{w_\textrm{FN}=0.001}, w_\textrm{FP}=\{i\}_{i=1}^{8}\big)\)}
    \label{fig:ocagi_adult_tp_0.5linfno}
    \end{subfigure}
    \hfill
    \begin{subfigure}[t]{0.47\linewidth}
        \centering
        \includegraphics[width=\linewidth]{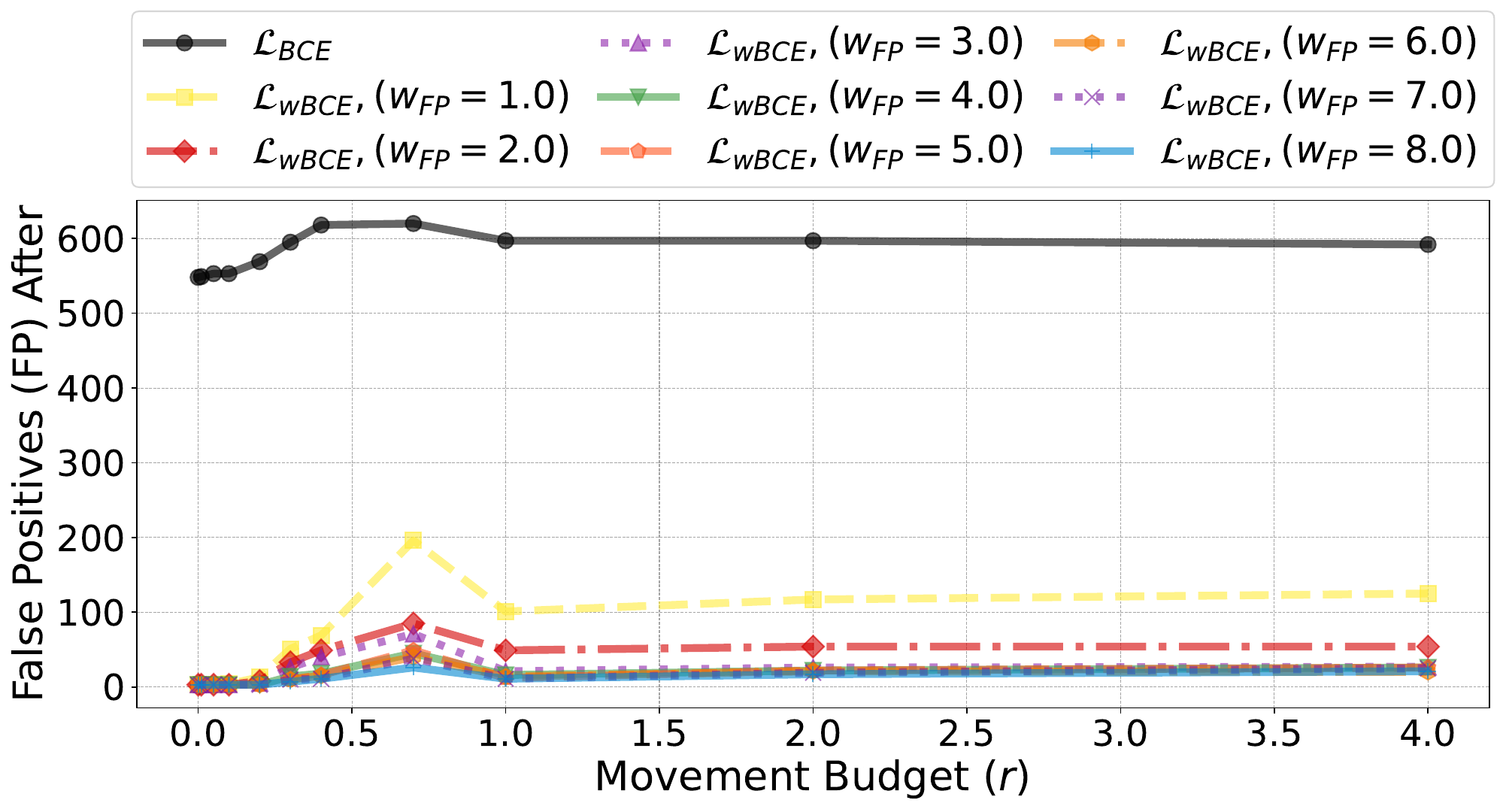}
    \caption{On the \textbf{Adult} dataset, the \textbf{number of false positives} when agents respond to a BCE-trained model and wBCE-trained models with weight configurations \(\big(\mathbf{w_\textrm{FN}=0.001}, w_\textrm{FP}=\{i\}_{i=1}^{8}\big)\)}
    \label{fig:ocagi_adult_fp_0.5linfno}
    \end{subfigure}

    \vspace{1.2em}

    \begin{subfigure}[t]{0.47\linewidth}
        \centering
        \includegraphics[width=\linewidth]{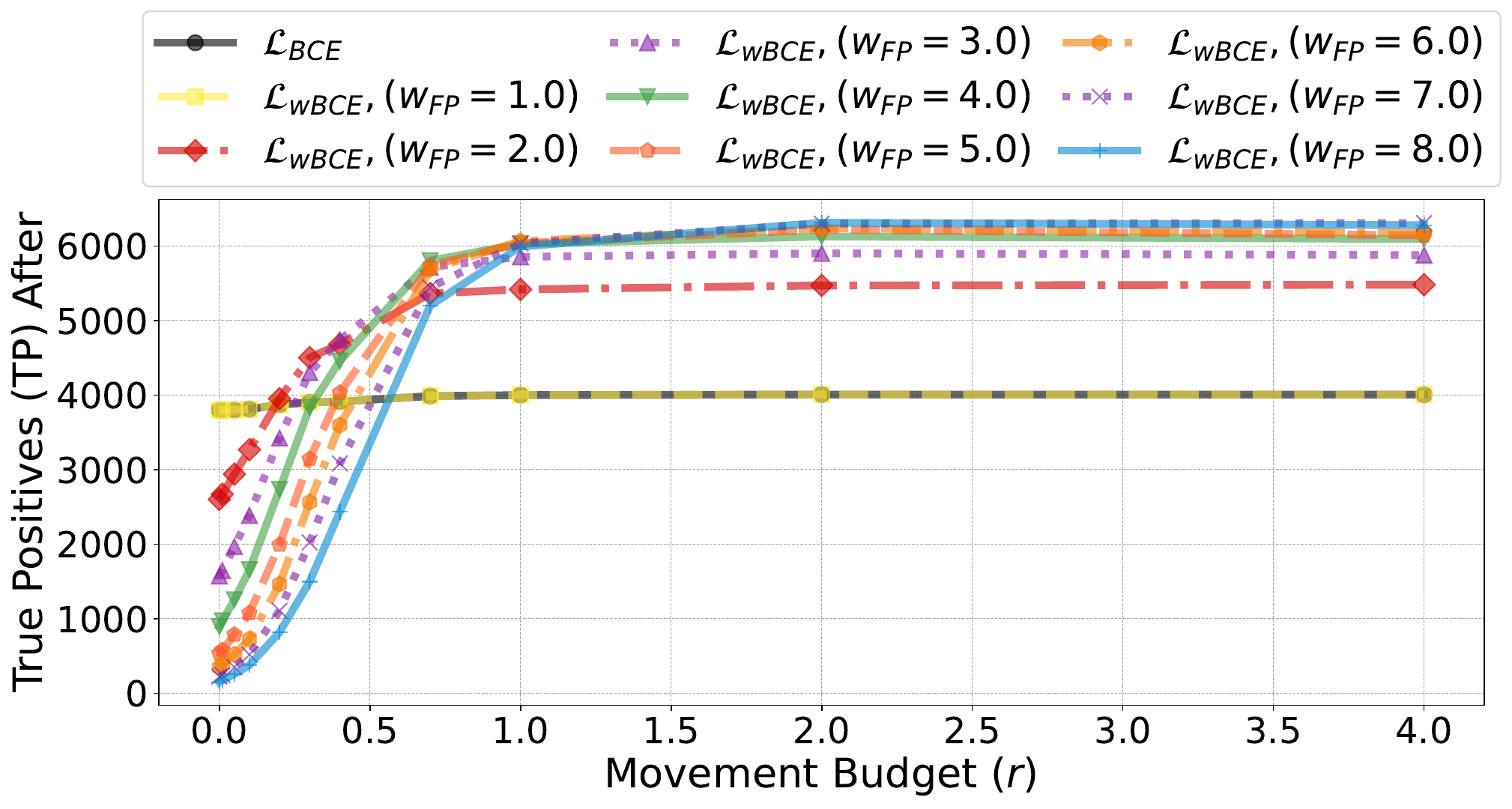}
    \caption{On the \textbf{OULAD} dataset, the \textbf{number of true positives} when agents respond to a BCE-trained model and wBCE-trained models with weight configurations \(\big(\mathbf{w_\textrm{FN}=1.0}, w_\textrm{FP}=\{i\}_{i=1}^{8}\big)\)}
    \label{fig:ocagi_oulad_tp_0.5linfno}
    \end{subfigure}
    \hfill
    \begin{subfigure}[t]{0.47\linewidth}
        \centering
        \includegraphics[width=\linewidth]{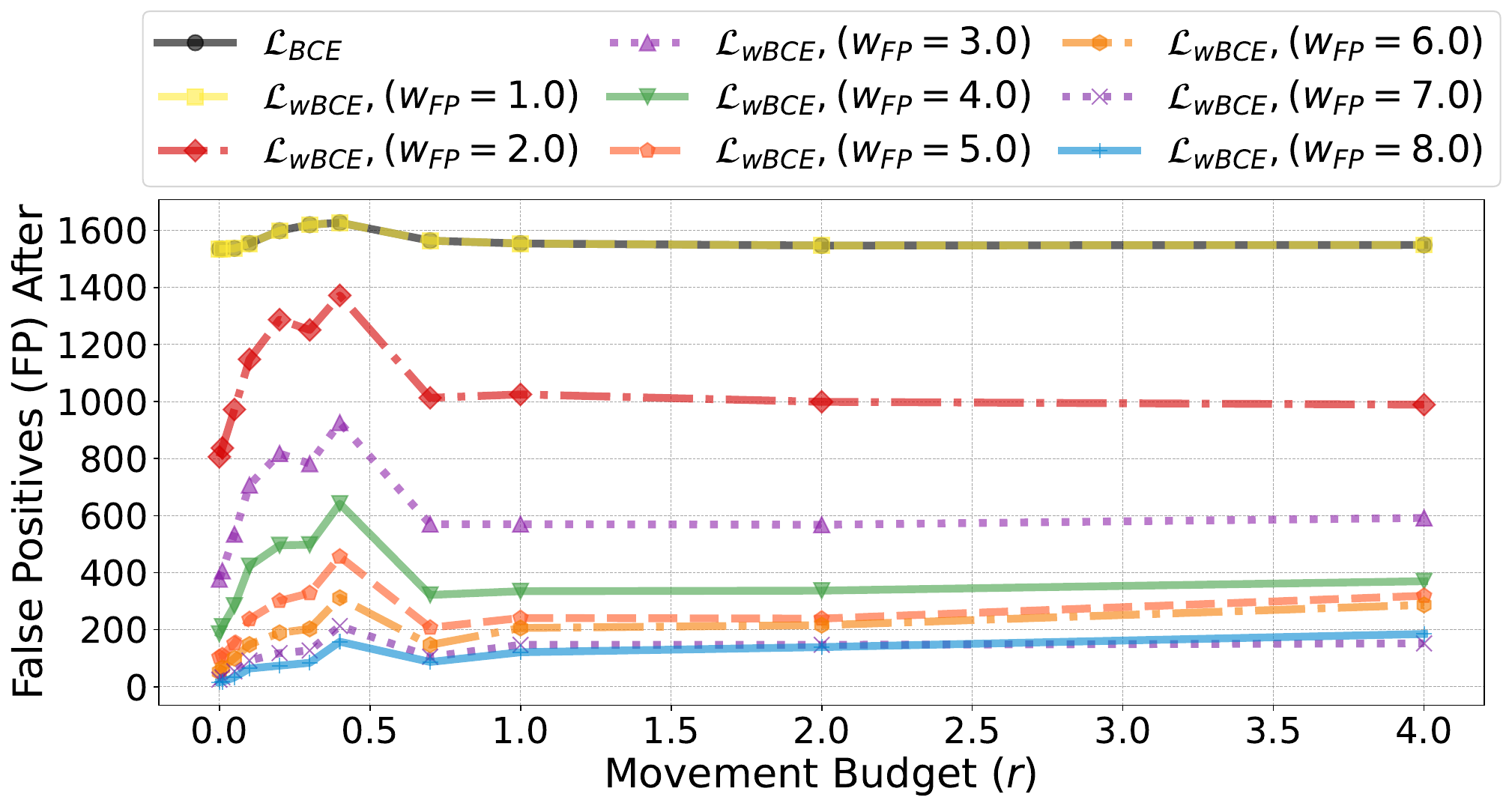}
    \caption{On the \textbf{OULAD} dataset, the \textbf{number of false positives} when agents respond to a BCE-trained model and wBCE-trained models with weight configurations \(\big(\mathbf{w_\textrm{FN}=1.0}, w_\textrm{FP}=\{i\}_{i=1}^{8}\big)\)}
    \label{fig:ocagi_oulad_fp_0.5linfno}
    \end{subfigure}

    \caption[A comparative analysis to study the variation in the true-to-false positives ratio]
    {Comparative analysis of the variation in the number of
    true positives (\subref{fig:ocagi_adult_tp_0.5linfno} and \subref{fig:ocagi_oulad_tp_0.5linfno}) and number of
    false positives (\subref{fig:ocagi_adult_fp_0.5linfno} and \subref{fig:ocagi_oulad_fp_0.5linfno}) after agents modify their features under different movement budgets \(r\) in response to BCE- and wBCE-trained models with varying weights: \(\big(w_\textrm{FN}=0.001, w_\textrm{FP}=\{i\}_{i=1}^{8}\big)\) on Adult and \(\big(w_\textrm{FN}=1.0, w_\textrm{FP}=\{i\}_{i=1}^{8}\big)\) on OULAD dataset. 
    In all cases, and all agents move within an \(\ell_{\infty}\) ball and are classified as positive if the probability is higher than \(0.5\).}
    \label{fig:oulad_adult_tpfp_th0.5linfno}
\end{figure*}

\end{appendices}

\printbibliography

\end{document}